\documentclass[
11pt, 
english, 
singlespacing, 
nolistspacing, 
parskip, 
chapterinoneline, 
consistentlayout, 
]{mythesis} 

\usepackage{multirow}
\usepackage{multicol}
\usepackage{adjustbox}
\usepackage{lscape}
\usepackage{longtable}

\usepackage[titles]{tocloft} 
\usepackage{etoc} 

\usepackage{graphicx}
\usepackage{xcolor} 
\usepackage{caption}
\usepackage{subcaption}

\usepackage[most]{tcolorbox}

\usepackage[utf8]{inputenc} 
\usepackage[T1]{fontenc} 
\usepackage{pifont}
\usepackage[T1]{fontenc}  
\usepackage{microtype}  

\usepackage{amsmath, amssymb, amsthm}  
\usepackage{mathtools}  
\usepackage{bm}  

\usepackage{listings}  
\usepackage{graphicx}  
\usepackage{subcaption}  
\usepackage{booktabs}  

\usepackage{hyperref}  
\hypersetup{
    colorlinks=true,
    linkcolor=blue,
    citecolor=blue,
    urlcolor=blue,
    linktoc=page
}

\usepackage{etoolbox}

\usepackage{scrlayer-scrpage}

\usepackage{enumitem} 
\usepackage[autostyle=true]{csquotes} 
\usepackage{epigraph} 
\usepackage[normalem]{ulem} 
\useunder{\uline}{\ul}{}
\usepackage{nicefrac} 

\usepackage[LGR,T1]{fontenc}

\usepackage{natbib}
\setcitestyle{authoryear,round,citesep={;},aysep={,},yysep={;}}

\usepackage{cite} 

\usepackage{setspace}

\definecolor{epistemicblue}{rgb}{0.0, 0.439, 0.753}

\definecolor{epistemicdarkblue}{rgb}{0.0, 0.2352, 0.4424}

\usepackage{bookmark}
\usepackage{titling}

\usepackage{todonotes}

\usepackage{mfirstuc}

\usepackage{titlesec}
\let\cleardoublepage\clearpage

\usepackage{svg}

\usepackage{float}
\usepackage{algorithm}
\usepackage{algpseudocode}  

\usepackage[utf8]{inputenc}  
\usepackage{lmodern}         
\usepackage{fix-cm} 

\usepackage{tikz}
\usetikzlibrary{positioning, intersections, arrows.meta, shapes}

\definecolor{myblue}{RGB}{3,52,108}
\definecolor{myorange}{RGB}{243,146,0}

\usepackage[most]{tcolorbox}
\tcbset{colframe=orange, colback=orange!5, coltitle=black, boxrule=0.8pt, arc=2mm}

\usepackage{xurl}  
\usepackage[most]{tcolorbox}

\usepackage[acronym]{glossaries}
\makeglossaries

\usepackage{xspace}

\definecolor{lightblue}{rgb}{0.93, 0.95, 1.0}

\newif\ifshowtoc
\showtoctrue 

\newcommand{\resbox}[2][]{
\begin{tcolorbox}[enhanced,attach boxed title to top center={yshift=-3mm,yshifttext=-1mm},
  colback=cyan!5!white,colframe=cyan!75!black,colbacktitle=blue!60!black,
  title=Research Question\ifx&#1&\else\ (#1)\fi,fonttitle=\bfseries,
  boxed title style={size=small,colframe=blue!40!black} ]
  #2
\end{tcolorbox}
}

\newlength\tocrulewidth
\newcommand{\addtoc}[0]{
\hypersetup{linkcolor=black}
\begingroup
{\noindent\large\bfseries Content }
\vspace{-3 mm}
\parindent=0em
\etocsettocstyle
{\rule{\linewidth}{\tocrulewidth}\vskip0.5\baselineskip}
{\rule{\linewidth}{\tocrulewidth}}
\etocsetstyle{section}
   {}
   {}
   {\bfseries\hspace*{1.5em}\etocnumber\hspace*{1.3em}\etocname\leaders\hbox to 1em{\hss.\hss}\hfill\etocpage\par\nobreak\kern0.5ex}
   {\vspace{-0.5em}}
\etocsetstyle{subsection}
   {}
   {}
   {\mdseries\hspace*{4em}\etocnumber\hspace*{1.3em}\etocname\leaders\hbox to 1em{\hss.\hss}\hfill\etocpage\par\nobreak\kern0.5ex}
   {}
\etocsettocdepth{subsection} 
\localtableofcontents 
\endgroup
}

\let\caps\capitalisewords
\MFUnocap{are}
\MFUnocap{or}
\MFUnocap{a}
\MFUnocap{an}
\MFUnocap{and}
\MFUnocap{the}
\MFUnocap{with}
\MFUnocap{of}
\MFUnocap{in}
\MFUnocap{vs}

\newcommand{\paper}[5]{{#1}. \textit{{#2}}. {#3} (\underline{{#4} {#5}}).}

\titleformat{\section}
{\normalfont\Large\bfseries}
{\thesection}{1em}{\caps}

\titleformat{\subsection}
{\normalfont\large\bfseries}
{\thesubsection}{1em}{\caps}

\titleformat{\subsubsection}
{\normalfont\normalsize\bfseries}
{\thesubsubsection}{1em}{\caps}

\titleformat{\part}[display]
{\normalfont\Huge\bfseries\centering}
{\partname\ \thepart}
{1em}
{\color{blue!50!black}}

\let\cite\citep

\thesistitle{Epistemic Deep Learning: Enabling Machine Learning Models to `Know When They Do Not Know'}

\firstsupervisor{Prof. Fabio Cuzzolin}
\secondsupervisor{Dr. Andrew {Bradley}}
\thirdsupervisor{Dr. Matthias {Rolf}}

\examiner{} 

\degree{{Doctor of Philosophy}} 

\author{\textbf{Shireen Kudukkil Manchingal}} 
\addresses{} 

\subject{Uncertainty Quantification in Deep Learning} 
\university{\href{https://www.brookes.ac.uk/}{Oxford Brookes University}} 
\department{\href{https://www.brookes.ac.uk/ecm/}{{Department of Engineering, Computing, and Mathematics}}} 
\faculty{\href{https://www.brookes.ac.uk/tde/}{Faculty of Technology, Design and Environment}} 

\AtBeginDocument{
\hypersetup{pdftitle=\ttitle} 
\hypersetup{pdfauthor=\authorname} 
\hypersetup{pdfkeywords=\keywordnames} 
}

\makeatletter
\providecommand{\subsubsubsection}{\@startsection{paragraph}{4}{\z@}
  {3.25ex \@plus 1ex \@minus .2ex}
  {1ex}
  {\normalfont\normalsize\bfseries}}
\providecommand{\l@subsubsubsection}{\@dottedtocline{4}{7em}{4em}}  
\makeatother

\begin{document}

\frontmatter 

\pagestyle{plain} 


\begin{titlepage}
\begin{center}

\hypersetup{
    urlcolor=blue!70!black!100, 
}


\vspace*{1cm}

{\fontsize{20}{28}\selectfont \bfseries \ttitle\par}\vspace{2cm} 


{\LARGE \href{https://scholar.google.com/citations?user=RioUSBEAAAAJ&hl=en}{\authorname}}\\[1cm] 


 
\vspace*{3cm}

\large {\textit{This thesis is submitted in fulfillment of the requirements for the degree of} \\
[0.6cm]} {\LARGE \textbf{\degreename}} \\
[0.5cm] {in} \textbf{{Computer Science and Mathematics}}\\
[0.7cm] {\textit{at the}}\\[0.7cm] 
\deptname\\[0.2cm]
\univname 
 
\vspace{3cm}
{\large May 31, 2025}

 


\end{center}
\end{titlepage}


\newpage
\thispagestyle{empty}
\mbox{}

\begin{flushright}

{\Huge \bfseries{{Thesis Committee}}
}\\[1cm]
\addchaptertocentry{Thesis Committee}

{\LARGE\bfseries{{Supervisors}}}\\[1cm]

{\Large\textbf{\href{https://www.brookes.ac.uk/profiles/staff/fabio-cuzzolin/}{\fsupname} }} \\{\itshape
 Professor of Artificial Intelligence \\
 Director of AIDAS Institute \\
 Oxford Brookes University \\[1cm]
}

{\Large\textbf{\href{https://www.brookes.ac.uk/profiles/staff/andrew-bradley}{\ssupname}}} \\
{\itshape
 Reader in Automotive \& Motorsport Engineering \\
 Lead, Autonomous Driving and Intelligent Transport \\
 Oxford Brookes University \\[1cm]
}

{\Large\textbf{\href{https://www.brookes.ac.uk/profiles/staff/matthias-rolf}{\tsupname}}} \\
{\itshape
 Reader in Artificial Intelligence \& Robotics \\
 Postgraduate Research Tutor, PhD Computing \\
 Oxford Brookes University \\[1cm]
}

{\LARGE\bfseries{{Examiners}}}\\[1cm]

{\Large\textbf{\href{https://www.brookes.ac.uk/profiles/staff/alex-rast}{Dr. Alexander Rast}}} \\{\itshape
 \textit{(Internal examiner)}\\
{\itshape
 Senior Lecturer in Artificial Intelligence \\
 Leadership at AIDAS Institute \\
 Oxford Brookes University \\[1cm]
}}


{\Large\textbf{\href{https://www.hds.utc.fr/~sdesterc/dokuwiki/}{Dr. Sébastien Destercke}}} \\{\itshape
 \textit{(External examiner)}\\
 {\itshape
 Senior AI Researcher \\
 Centre National de la Recherche Scientifique (CNRS)
 \\
 Laboratoire Heuristique et Diagnostic des Systèmes Complexes (Heudiasyc)\\
 Paris, France \\[1cm]
}
}

\end{flushright}

\cleardoublepage

\newpage
\thispagestyle{empty}
\vspace*{\fill}


\begin{declaration}
\addchaptertocentry{\authorshipname}
\vspace{0.1cm}

    I hereby declare that except where specific reference is made to the work of others, the contents of this thesis are original and have not been submitted in whole or in part for consideration for any other degree or qualification in this, or any other university.
    This thesis is my own work, except for contributions explicitly specified in the text, Acknowledgements, and any mentioned collaborations.
    This thesis contains less than 65,000 words excluding bibliography, footnotes, tables, and equations and has fewer than 80 figures.

\vspace{0.5cm}
\begin{flushright}
    Shireen Kudukkil Manchingal 
    
    May 2025
\end{flushright}

\end{declaration}

\newpage
\thispagestyle{empty}
\mbox{}

\begin{acknowledgements}
\addchaptertocentry{\acknowledgementname} 

I am profoundly grateful to my supervisor, \textit{Fabio}, whose unwavering support, patience, and insightful guidance have been the foundation of both this research and my development as a researcher. His steadfast belief in my potential, and the intellectual freedom he granted me, has been instrumental in shaping both this work and my academic identity. The impact he has had on my career is immense, and his mentorship is a guiding influence I will carry with me throughout my life. 


I am also deeply thankful to my second supervisor, \textit{Andy}, for his thoughtful advice, constant encouragement and support throughout this journey, which has greatly enriched both this work and my growth as a researcher.
My sincere gratitude goes to my third supervisor, \textit{Matthias}, whose guidance throughout my PhD journey, also in the form of my PGR tutor, provided me with the clarity and motivation needed during challenging phases.


I feel incredibly fortunate to have worked alongside brilliant colleagues throughout this journey. Special thanks to the entire Epistemic AI project team—\textit{Kaizheng}, \textit{Mubashar}, \textit{Maryam}, \textit{Matthijs}, \textit{Hans}, \textit{Neil}, \textit{Julian}, \textit{Keivan}, \textit{David}, \textit{Salman}, \textit{Moritz}, \textit{Noah}, \textit{Pascal}, \textit{Adam}, and \textit{Guopeng}—for several insightul discussions and constant motivation. I am also grateful to collaborators like \textit{Michele}, and VAIL members \textit{Izzeddin}, \textit{Vivek} and \textit{Ajmal}, whose insights and camaraderie made this experience truly rewarding.


I am also sincerely thankful to the staff at the \textit{Department of Engineering, Computing, and Mathematics} for providing a supportive academic environment and invaluable resources that made this research possible.


Most importantly, I owe my deepest gratitude to my family---my parents \textit{Assia} and \textit{Mohammed}, \textit{Shafna}, \textit{Shakeer}, \textit{Rayyan} and \textit{Raha}. \textit{Safeer}, my partner, has been my greatest support and strength throughout this journey. Their unconditional love and patience have been my anchor through the challenges and triumphs of this process. 

\vspace{4mm}

\textit{This thesis is based on work conducted as part of the \textbf{European Union’s Horizon 2020 Research and Innovation Programme} under Grant Agreement No. 964505 (E-pi), which has also funded my PhD and supported this research throughout.
}

\end{acknowledgements}

\newpage
\thispagestyle{empty}
\mbox{}


\dedicatory{
  \begin{onehalfspace}
  \textit{In loving memory of} \\[1em]
  Ayisha, \\[1em]
  \textit{my grandmother, \\
  and the strongest woman I know, \\
  who passed away in 2023 \\
  during my PhD journey.}
  \end{onehalfspace}
}

\newpage
\thispagestyle{empty}
\mbox{}


\begin{abstract}
\addchaptertocentry{\abstractname} 

\noindent Machine learning has achieved remarkable successes, particularly with deep learning, yet its deployment in safety-critical domains remains hindered by an inherent inability to manage uncertainty, resulting in overconfident and unreliable predictions when models encounter out-of-distribution data, adversarial perturbations, or naturally fluctuating environments. 

\vspace{1mm}

\noindent This thesis, titled \textit{Epistemic Deep Learning: Enabling Machine Learning Models to `Know When They Do Not Know'},
addresses these critical challenges by advancing the paradigm of \textbf{Epistemic Artificial Intelligence}, which explicitly models and quantifies \emph{epistemic uncertainty}: the uncertainty arising from limited, biased, or incomplete training data, as opposed to the irreducible randomness of \emph{aleatoric uncertainty}, thereby empowering models to acknowledge their limitations and refrain from overconfident decisions when uncertainty is high.

\vspace{1mm}

\noindent Central to this work is the development of the \textbf{Random-Set Neural Network (RS-NN)}, a novel methodology that leverages random set theory to predict belief functions over sets of classes, capturing the extent of epistemic uncertainty through the width of associated credal sets, and demonstrating superior performance in terms of robustness and out-of-distribution detection compared to traditional approaches. The final part of the thesis explores applications of RS-NN, including its adaptation to Large Language Models (LLMs) and its deployment in weather classification for autonomous racing.

\vspace{1mm}

\noindent In addition, the thesis proposes a \textbf{unified evaluation framework for uncertainty-aware classifiers}, motivated by the observation that existing methods employ heterogeneous uncertainty measures, such as mutual information in Bayesian models, variance in deep ensembles, thereby impeding systematic comparisons; the proposed framework bridges this gap by providing a common metric that balances prediction accuracy with precision or imprecision, enabling objective assessment and model selection for safety-critical applications.

\vspace{1mm}

\noindent Extensive experiments validate that integrating epistemic awareness into deep learning not only mitigates the risks associated with overconfident predictions but also lays the foundation for a paradigm shift in artificial intelligence, where the ability to `know when it does not know' becomes a hallmark of robust and dependable systems. The title encapsulates the core philosophy of this work, emphasizing that true intelligence involves recognizing and managing the limits of one’s own knowledge.

\end{abstract}



\tableofcontents

\newpage
\listoffigures 

\newpage
\listoftables 




\glsaddallunused

\newacronym{ai}{AI}{\textbf{A}rtificial \textbf{I}ntelligence}
\newacronym{dl}{DL}{\textbf{D}eep \textbf{L}earning}
\newacronym{ml}{ML}{\textbf{M}achine \textbf{L}earning}
\newacronym{llm}{LLM}{\textbf{L}arge \textbf{L}anguage \textbf{M}odel}
\newacronym{ood}{OoD}{\textbf{O}ut-\textbf{o}f-\textbf{D}istribution detection}
\newacronym{agis}{AGI}{\textbf{A}rtificial \textbf{G}eneral \textbf{I}ntelligence}
\newacronym{uq}{UQ}{\textbf{U}ncertainty \textbf{Q}uantification}
\newacronym{rsnn}{RS-NN}{\textbf{R}andom-\textbf{S}et \textbf{N}eural \textbf{N}etworks}
\newacronym{rsllm}{RS-LLM}{\textbf{R}andom-\textbf{S}et \textbf{L}arge \textbf{L}anguage \textbf{M}odels}
\newacronym{ae}{AE}{\textbf{A}leatoric \textbf{U}ncertainty}
\newacronym{au}{EU}{\textbf{E}pistemic \textbf{U}ncertainty}
\newacronym{bnn}{BNN}{\textbf{B}ayesian \textbf{N}eural \textbf{N}etworks}
\newacronym{mcd}{MCD}{\textbf{M}onte \textbf{C}arlo \textbf{D}ropout}
\newacronym{mcmc}{MCMC}{\textbf{M}arkov \textbf{C}hain \textbf{M}onte \textbf{C}arlo}
\newacronym{edl}{EDL}{\textbf{E}vidential \textbf{D}eep \textbf{L}earning}
\newacronym{vi}{VI}{\textbf{V}ariational \textbf{I}nference}
\newacronym{laplace}{LAPLACE}{\textbf{L}aplace \textbf{A}pproximation for Bayesian Inference}
\newacronym{mc}{MC}{\textbf{M}onte \textbf{C}arlo sampling}
\newacronym{dir}{DIR}{\textbf{D}irichlet \textbf{D}istribution}
\newacronym{ddu}{DDU}{\textbf{D}eep \textbf{D}eterministic \textbf{U}ncertainty}
\newacronym{cp}{CP}{\textbf{C}onformal \textbf{P}rediction}
\newacronym{gmm}{GMM}{\textbf{G}aussian \textbf{M}ixture \textbf{M}odel}
\newacronym{cnn}{CNN}{\textbf{C}onvolutional \textbf{N}eural \textbf{N}etworks}
\newacronym{rs}{RS}{\textbf{R}andom-\textbf{S}ets}
\newacronym{dempster}{DST}{\textbf{D}empster-\textbf{S}hafer \textbf{T}heory}
\newacronym{bpa}{BPA}{\textbf{B}asic \textbf{P}robability \textbf{A}ssignment}
\newacronym{kl}{KL}{\textbf{K}ullback-\textbf{L}eibler {D}ivergence}
\newacronym{ns}{NS}{\textbf{N}on-\textbf{S}pecificity}
\newacronym{vi}{VI}{\textbf{V}ariational \textbf{I}nference} 
\newacronym{dp}{DP}{\textbf{D}irichlet \textbf{P}rocess}
\newacronym{mse}{MSE}{\textbf{M}ean \textbf{S}quared \textbf{E}rror}
\newacronym{map}{MAP}{\textbf{M}aximum \textbf{A} \textbf{P}osteriori}
\newacronym{mle}{MLE}{\textbf{M}aximum \textbf{L}ikelihood \textbf{E}stimation}
\newacronym{auc}{AUC}{\textbf{A}rea \textbf{U}nder the \textbf{C}urve}
\newacronym{roc}{ROC}{\textbf{R}eceiver \textbf{O}perating \textbf{C}haracteristic}
\newacronym{lstm}{LSTM}{\textbf{L}ong \textbf{S}hort-\textbf{T}erm \textbf{M}emory}
\newacronym{gan}{GAN}{\textbf{G}enerative \textbf{A}dversarial \textbf{N}etwork}
\newacronym{vae}{VAE}{\textbf{V}ariational \textbf{A}utoencoder}
\newacronym{auroc}{AUROC}{\textbf{A}rea \textbf{U}nder the \textbf{R}eceiver \textbf{O}perating \textbf{C}haracteristic curve}
\newacronym{auprc}{AUPRC}{\textbf{A}rea \textbf{U}nder the \textbf{P}recision-\textbf{R}ecall \textbf{C}urve}
\newacronym{ece}{ECE}{\textbf{E}xpected \textbf{C}alibration \textbf{E}rror}
\newacronym{tpr}{TPR}{\textbf{T}rue \textbf{P}ositive \textbf{R}ate (also called sensitivity or recall)}
\newacronym{fpr}{FPR}{\textbf{F}alse \textbf{P}ositive \textbf{R}ate}

\printglossary[title = List of Abbreviations, type=\acronymtype]

\newpage
\newglossaryentry{inputx}{
    name={\textbf{\boldmath{$\mathbf{x}$}}},
    sort={0},
    description={Input data point in the input space $\mathbf{X}$}
}

\newglossaryentry{labely}{
    name={\textbf{\boldmath{$y$}}},
    sort={1},
    description={Class label corresponding to input $\mathbf{x}$, belonging to $\mathbf{Y}$}
}

\newglossaryentry{trainingdata}{
    name={\textbf{\boldmath{$\mathbb{D}$}}},
    sort={2},
    description={Training dataset $\{(\mathbf{x}_i, y_i)\}_{i=1}^{\mathcal{N}}$ of size $\mathcal{N}$}
}

\newglossaryentry{predictedprob}{
    name={\textbf{\boldmath{$\hat{p}(y \mid \mathbf{x}, \mathbb{D})$}}},
    sort={3},
    description={Predictive distribution of label $y$ given input $\mathbf{x}$ and training data $\mathbb{D}$}
}

\newglossaryentry{softmaxpred}{
    name={\textbf{\boldmath{$\hat{p}_s(y \mid \mathbf{x}, \mathbb{D})$}}},
    sort={4},
    description={Predictive distribution produced by a softmax-based standard classifier (CNN)}
}

\newglossaryentry{ddupred}{
    name={\textbf{\boldmath{$\hat{p}_{ddu}(y \mid \mathbf{x}, \mathbb{D})$}}},
    sort={5},
    description={Prediction from Deterministic Uncertainty Quantification (DDU)}
}

\newglossaryentry{bnn}{
    name={\textbf{\boldmath{$\hat{p}_b(y \mid \mathbf{x}, \mathbb{D})$}}},
    sort={6},
    description={Prediction from a Bayesian Neural Network by marginalizing over the posterior of model parameters}
}

\newglossaryentry{deensemble}{
    name={\textbf{\boldmath{$\hat{p}_{de}(y \mid \mathbf{x}, \mathbb{D})$}}},
    sort={7},
    description={Prediction from Deep Ensembles (DEs), obtained by averaging outputs from multiple models}
}

\newglossaryentry{edl}{
    name={\textbf{\boldmath{$\hat{p}_e(y \mid \mathbf{x}, \mathbb{D})$}}},
    sort={8},
    description={Prediction from an Evidential Deep Learning model, as parameters of a second-order Dirichlet distribution}
}

\newglossaryentry{classspace}{
    name={\textbf{\boldmath{$\mathcal{C}$}}},
    sort={9},
    description={The set of all possible class labels}
}

\newglossaryentry{powerset}{
    name={\textbf{\boldmath{$\mathbb{P}(\mathcal{C})$}}},
    sort={10},
    description={The powerset of $\mathcal{C}$, i.e., the set of all its subsets}
}

\newglossaryentry{massvalue}{
    name={\textbf{\boldmath{$m(A)$}}},
    sort={11},
    description={Mass value assigned to a subset $A \subseteq \mathcal{C}$ in belief function theory}
}

\newglossaryentry{focalsets}{
    name={\textbf{\boldmath{$\mathcal{O}$}}},
    sort={12},
    description={Set of focal class subsets, $\{A_k \subset \mathcal{C}\}$}
}

\newglossaryentry{beliefvec}{
    name={\textbf{\boldmath{$\mathbf{bel}$}}},
    sort={13},
    description={Vector of belief values for each subset $A$ of $\mathcal{C}$}
}

\newglossaryentry{detclassifier}{
    name={\textbf{\boldmath{$e : X \rightarrow \mathcal{C}$}}},
    sort={14},
    description={Classifier mapping inputs to single class labels}
}

\newglossaryentry{rsclassifier}{
    name={\textbf{\boldmath{$e : X \rightarrow \mathbb{P}(\mathcal{C})$}}},
    sort={15},
    description={Classifier mapping inputs to subsets of class labels}
}

\newglossaryentry{numinstances}{
    name={\textbf{\boldmath{$\mathcal{N}$}}},
    sort={16},
    description={Number of training instances in dataset $\mathbb{D}$}
}

\newglossaryentry{subsetA}{
    name={\textbf{\boldmath{$A$}}},
    sort={17},
    description={A subset of class labels, i.e., $A \subseteq \mathcal{C}$}
}

\newglossaryentry{beliefvalue}{
    name={\textbf{\boldmath{$Bel(A)$}}},
    sort={18},
    description={Belief value assigned to subset $A$ of class labels}
}

\newglossaryentry{numclasses}{
    name={\textbf{\boldmath{$N$}}},
    sort={19},
    description={Number of classes in the classification problem}
}

\newglossaryentry{focalset}{
    name={\textbf{\boldmath{$A$}}},
    sort={20},
    description={A subset of classes $A \subseteq \mathcal{C}$, also called a focal set}
}

\newglossaryentry{classlabelspace}{
    name={\textbf{\boldmath{$\mathbf{Y}$}}},
    sort={21},
    description={The class label space, i.e., the set of all possible class labels}
}

\newglossaryentry{beliefvector}{
    name={\textbf{\boldmath{$\mathbf{bel}$}}},
    sort={22},
    description={Vector of belief values $\{Bel(A) : A \in \mathbb{P}(\mathcal{C})\}$ representing ground truth or predicted belief}
}

\newglossaryentry{budget}{
    name={\textbf{\boldmath{$\mathcal{O}$}}},
    sort={23},
    description={Budget of selected focal sets: $\mathcal{O} = \{A_1, \ldots, A_K\}$}
}

\newglossaryentry{numbudget}{
    name={\textbf{\boldmath{$K$}}},
    sort={24},
    description={Number of focal sets selected in the budgeting process}
}

\newglossaryentry{overlap}{
    name={\textbf{\boldmath{$overlap(A)$}}},
    sort={25},
    description={Overlap ratio of subset $A \subseteq \mathcal{C}$, computed as intersection over union of clusters representing classes in $A$: $\frac{\cap_{c \in A} A^c}{\cup_{c \in A} A^c}$}
}

\newglossaryentry{massvalueshat}{
    name={\textbf{\boldmath{$\hat{m}$}}},
    sort={26},
    description={Predicted mass values assigned to focal sets}
}

\newglossaryentry{pignistic}{
    name={\textbf{\boldmath{$BetP$}}},
    sort={27},
    description={Pignistic probability computed from the predicted belief function, used for decision making}
}

\newglossaryentry{lossrs}{
    name={\textbf{\boldmath{$\mathcal{L}_{RS}$}}},
    sort={28},
    description={Loss function used to train RS-NN by comparing predicted and ground truth belief vectors}
}

\newglossaryentry{credalset}{
    name={\textbf{\boldmath{$\hat{\mathbb{C}r} (y \mid \mathbf{x}, \mathbb{D})$}}},
    sort={29},
    description={Credal set defined by predicted probability intervals for class $y$ given input $\mathbf{x}$ and data $\mathbb{D}$: set of probability distributions bounded by lower and upper probabilities $\hat{\underline{p}}(y), \hat{\overline{p}}(y)$}
}

\newglossaryentry{lowerprob}{
    name={\textbf{\boldmath{$\underline{P}$}}},
    sort={30},
    description={Lower probability function mapping subsets $A \subseteq \mathbf{Y}$ to $[0,1]$, monotone with $\underline{P}(\emptyset)=0$ and $\underline{P}(\mathbf{Y})=1$}
}

\newglossaryentry{upperprob}{
    name={\textbf{\boldmath{$\overline{P}$}}},
    sort={31},
    description={Upper probability function conjugate to lower probability $\underline{P}$, defined as $\overline{P}(A) = 1 - \underline{P}(A^c)$ for all $A \subseteq \mathbf{Y}$}
}

\newglossaryentry{probmeasureset}{
    name={\textbf{\boldmath{$\mathcal{P}$}}},
    sort={32},
    description={Set of all probability distributions over the class space $\mathbf{Y}$}
}

\newglossaryentry{credalsetlower}{
    name={\textbf{\boldmath{$\mathbb{C}r(\underline{P})$}}},
    sort={33},
    description={Credal set associated with a lower probability $\underline{P}$, i.e., all probability measures $P \in \mathcal{P}(\mathbf{Y})$ such that $P(A) \geq \underline{P}(A)$ for all $A \subseteq \mathbf{Y}$}
}

\newglossaryentry{mobiuslowerprob}{
    name={\textbf{\boldmath{$m_{\underline{P}}(A)$}}},
    sort={34},
    description={Möbius inverse of the lower probability $\underline{P}$ for subset $A \subseteq \mathbf{Y}$, computed as $m_{\underline{P}}(A) = \sum_{B \subseteq A} (-1)^{|A \setminus B|} \underline{P}(B)$}
}

\newglossaryentry{massfunction}{
    name={\textbf{\boldmath{$m$}}},
    sort={35},
    description={Mass function or basic probability assignment, mapping subsets $A \subseteq \mathbf{Y}$ to $[0,1]$ with $m(\emptyset) = 0$ and $\sum_{A} m(A) = 1$}
}

\newglossaryentry{non_specificity}{
    name={\textbf{\boldmath{$NS[m]$}}},
    sort={36},
    description={Degree of non-specificity of a mass function $m$, defined as $NS[m] = \sum_{A \subseteq \mathbf{Y}} m(A) \log |A|$}
}

\newglossaryentry{metric}{
    name={\textbf{\boldmath{$\mathcal{E}$}}},
    sort={37},
    description={Combined metric for evaluation, $\mathcal{E} = d(y, \hat{y}) + \lambda \cdot NS[m]$, where $d$ is a divergence (e.g., KL divergence) and $NS[m]$ is non-specificity}
}

\newglossaryentry{kl}{
    name={\textbf{\boldmath{$D_{KL}(y || \hat{y})$}}},
    sort={38},
    description={Kullback-Leibler divergence between ground truth distribution $y$ and prediction $\hat{y}$: $D_{KL}(y||\hat{y}) = \int_x y(x) \log \frac{y(x)}{\hat{y}(x)} dx$}
}

\printglossary[title=Notations]


\mainmatter 

\pagestyle{thesis} 


\chapter{Introduction} 
\label{ch:intro} 


\section{Motivation}\label{app:motivation}

The success of artificial intelligence (AI), particularly in the realm of deep learning, is widely acknowledged and continues to redefine the boundaries of what machines can achieve \cite{lecun2015deep}. AI models have demonstrated impressive capabilities across a wide spectrum of tasks, from image recognition and natural language processing to game playing and protein folding \citep{jumper2021highly}. Large language models (LLMs) such as GPT-3 \cite{brown2020language} exemplify the growing capacity of neural networks to manipulate and generate human-like language. Generative models, including DALL·E 2 \cite{ramesh2022hierarchical}, have advanced into creative domains, generating novel content across modalities. Moreover, multimodal systems like CLIP \cite{radford2021learning} have bridged vision and language, achieving strong performance in tasks requiring cross-domain understanding. Despite these advancements, the deployment of AI in real-world systems often falls short of expectations, revealing a gap between benchmark success and operational reliability.

This gap has fueled both public enthusiasm and critical reflection. Some of the most anticipated AI applications, such as fully autonomous vehicles, remain underdeveloped for worldwide deployment despite significant investment and media attention \cite{amodei2016concrete}. While self-driving prototypes can perform well under controlled conditions, they consistently struggle to generalize across diverse, unstructured environments. Similarly, generative AI models, while compelling, often exhibit hallucination, inconsistency, and brittleness under distributional shift \cite{bender2021dangers}. Meanwhile, discourse around artificial general intelligence (AGI) and the speculative risk of uncontrollable AI systems often diverts attention from the immediate, pressing challenge of ensuring robustness, reliability, and interpretability in today’s systems \cite{russell2015research}. 

Neural networks, while powerful, are also often overconfident in their predictions, even when presented with adversarial inputs, noise, or samples drawn from distributions not represented in the training data \cite{papernot2016limitations, hendrycks2016baseline, hullermeier2021aleatoric, zhang2021evaluating}. This overconfidence is not just a theoretical concern; it manifests in high-stakes applications such as autonomous driving, where models fail to generalize to unexpected scenarios like unusual weather, rare traffic patterns, or sensor failures \cite{bojarski2016end}. Despite mitigation strategies such as dropout regularization \cite{srivastava2014dropout}, domain adaptation \cite{ganin2015unsupervised}, adversarial training \cite{goodfellow2014explaining}, and various ensemble methods, the core problem persists. 

There is a growing consensus that the
accurate estimation of \textit{uncertainty} \citep{cuzzolin2024uncertainty} is vital to improve machine learning models' reliability \citep{senge2014reliable, kendall2017uncertainties}.
However, one of the most pressing limitations of modern machine learning (ML) systems is their inability to handle uncertainty in a principled and efficient manner. 
This has fueled increasing interest in the field of uncertainty quantification (UQ), which aims to enable models to assess and express uncertainty about their predictions. Accurate uncertainty estimation is critical for deploying ML models in safety-critical environments where incorrect or overconfident decisions can result in significant harm. For instance, in autonomous driving, predictive uncertainty can be used to trigger fallback strategies or human intervention \cite{fort2019large}; in medical diagnosis, uncertainty-aware models can defer decisions to clinicians when confidence is low \cite{lambrou2010reliable}; and in climate and infrastructure applications such as flood risk estimation \cite{chaudhary2022flood} and structural health monitoring \cite{vega2022variational}, quantifying uncertainty informs probabilistic risk assessments. Machine learning techniques for UQ span a wide methodological spectrum, including Bayesian Neural Networks (BNNs) \cite{blundell2015weight, gal2016dropout, DBLP:journals/corr/abs-2007-06823}, Deep Ensembles \cite{lakshminarayanan2017simple}, Monte Carlo dropout \cite{gal2016dropout}, Evidential Deep Learning \cite{sensoy}, and approaches grounded in Conformal Prediction \cite{angelopoulos2021gentle}. However, these existing UQ methods have key limitations. \textit{Bayesian} models are sensitive to prior mis-specification (with the risk of biasing the whole process) and incur heavy computational overhead \citep{https://doi.org/10.48550/arxiv.2105.06868, caprio2023imprecise}, and Bayesian Model Averaging (BMA) may dilute useful predictive information \citep{hinne2020conceptual, graefe2015limitations}. \textit{Ensemble} methods are computationally demanding \citep{DBLP:journals/corr/abs-2007-06823, he2020bayesian}.
{
\textit{Conformal} predictors lose validity and efficiency in the presence of distributional shift or model misspecification without adjustments for epistemic uncertainty \citep{bates2023testing}.}
\textit{Evidential} approaches \cite{sensoy} violate asymptotic assumptions, struggle with out-of-distribution data \citep{bengs2022pitfalls, ulmer2023prior, kopetzki2021evaluating, stadler2021graph} and exhibit high inference times (Tab. \ref{tab:uq-comparison-table}). These shortcomings underscore the inadequacy of current methods in capturing the full spectrum of uncertainty, as well as their inefficiency in deployment within complex, real-world environments.


This thesis addresses that gap by exploring epistemic uncertainty quantification in deep learning (for classification tasks), focusing on techniques that enable models to express \textit{ignorance} (Sec. \ref{ch:knowledge}) in meaningful, interpretable, and actionable ways. It operates within the paradigm of \textbf{Epistemic Artificial Intelligence} (Chapter \ref{ch:epi_ai}), which emphasizes the importance of learning from ignorance, in line with the Socratic principle: \textit{`know that you do not know.'}
A central contribution of Epistemic Artificial Intelligence (Epistemic AI) is its principled use of second-order uncertainty measures (Sec. \ref{app:uncertainty-measures}) to represent and reason about epistemic uncertainty, \textit{i.e.}, uncertainty arising from a lack of knowledge or insufficient information. Unlike traditional probabilistic models that assign precise probabilities to outcomes (first-order uncertainty), Epistemic AI explicitly models uncertainty about uncertainty. This second-order framework enables systems to quantify their own ignorance, rather than being forced to commit to possibly overconfident or ill-informed predictions.
Second-order uncertainty measures such as random sets \citep{Molchanov05} and credal sets  \citep{levi1980enterprise,cuzzolin2008credal} are rooted in imprecise probability theory (Sec. \ref{sec:imprecise-models}) and offer a richer language for expressing model uncertainty in situations where the data is ambiguous, scarce, or non-representative. 

To realize this Epistemic AI perspective, this thesis proposes principled frameworks for UQ in machine learning. It tackles two main challenges:

\begin{itemize}
    \item \textbf{Developing a rigorous and mathematically grounded method for estimating epistemic uncertainty by leveraging second-order uncertainty measures}, which better represent ignorance and the inherent uncertainty about unknowns. This is achieved through the introduction of \textbf{Random-Set Neural Networks (RS-NN)} (Chapter \ref{ch:rsnn}), a novel classification approach based on the theory of random sets that overcomes limitations of existing UQ methods. Contributions to credal set models are also briefly discussed in Chapter \ref{ch:cre_models}.
    \item \textbf{Enabling fair and consistent comparison of different UQ methods}, as their epistemic predictions often take disparate forms. This challenge is addressed by proposing \textbf{A Unified Evaluation Framework for Epistemic Predictions} (Chapter \ref{ch:eval_framework}), which standardizes the evaluation process and helps practitioners select the most suitable model for their application.
\end{itemize}

\section{Research Questions}\label{sec:research-ques}

Motivated by these challenges and the existing academic literature, which is detailed in Chapter \ref{ch:unc_models}, this thesis aims to answer the following key research questions:

\begin{figure}[!h]
\centering
\begin{tikzpicture}[
    question/.style={
        rectangle, rounded corners, draw=orange, text width=0.95\textwidth, minimum height=1.1cm, font=\bfseries\color{myblue},
        fill=none, inner sep=8pt
    },
    explanation/.style={
        rectangle, rounded corners,
        draw=none, fill=none,
       text width=0.98\textwidth, minimum height=1.1cm,
        font=\color{black}, inner sep=8pt
    },
    connector/.style={
        draw=orange, line width=1pt,
        -{Latex[length=3mm]}
    },
    title/.style={
        font=\bfseries\color{black}
    },
        separator/.style={
        draw=orange, line width=1pt
    }
]

\node[question] (q1) at (0cm, 0.1) {\textbf{\textcolor{black}{Q1:}} How can epistemic uncertainty be formally and reliably quantified in deep learning models using second-order uncertainty representations such as random sets while tackling set complexity?};
\node[explanation] (e1) at ([xshift=0cm, yshift=-1.4cm]q1.south) {
    Standard probabilistic models, including Bayesian approaches, struggle to represent epistemic uncertainty because a single probability distribution cannot capture ignorance. This question investigates whether more expressive, set-based frameworks can provide a principled alternative for modeling uncertainty {(Chapter \ref{ch:rsnn})}, and if they can be made computationally feasible {(Sec. \ref{sec:rsnn-budgeting})}.
};


\node[question] (q3) at (0cm, -4.5) {\textbf{\textcolor{black}{Q2:}} How do epistemic models that employ second-order uncertainty measures, such as random sets, behave when confronted with previously unseen or adversarial inputs at test time?};
\node[explanation] (e3) at ([xshift=0cm, yshift=-0.9cm]q3.south) {
    This question explores whether such models can more effectively distinguish between familiar and unfamiliar data, thereby improving out-of-distribution (OoD) detection {(Sec. \ref{sec:ood-detection})} and providing greater robustness against adversarial perturbations {(Secs. \ref{app:fgsm} \& \ref{app:noisy})}.
};

\node[question] (q4) at (0cm, -8.2) {\textbf{\textcolor{black}{Q3:}} How can a unified and principled evaluation framework be designed to compare epistemic uncertainty predictions across methods that produce fundamentally different output structures?};
\node[explanation] (e4) at ([xshift=0cm, yshift=-0.6cm]q4.south) {
    Given that different approaches yield probability vectors, intervals, belief functions, or sets, this question seeks a unified evaluation methodology {(Chapter \ref{ch:eval_framework})} for fair comparison.
};

\end{tikzpicture}
\vspace{-7mm}
\end{figure}

\section{Thesis Structure}

\begin{figure}[!h]
\vspace{2mm}
\centering
\small
\begin{tikzpicture}[
    box/.style={
        rectangle, draw=black, rounded corners, align=center,
        text width=3.5cm, minimum height=1.8cm, font=\scriptsize
    },
arrow/.style={
    thick,
    draw=orange!30,          
    -{Stealth[length=6mm, width=6mm]},    
    line width=9pt
    },
    connector/.style={
        draw=orange, line width=1pt
    },
    dot/.style={
        circle, fill=orange, minimum size=4pt, inner sep=0pt
    },
    title/.style={
        font=\bfseries\color{black}
    },
subtitle/.style={
    font=\bfseries\color{myblue}, align=center, text width=3.5cm
    }
]

\def\xstep{4}

\node[title] (title1) at (0, 5.2) {PART I};
\node[subtitle] (sub1) at (0, 4.3) {Knowing What\\ We Do Not Know};
\node[box] (box1) at (0, 1.4) {Introduces aleatoric vs. epistemic uncertainty, and surveys existing approaches.};
\node[dot] (dot1) at ([yshift=-1mm]sub1.south) {};
\draw[connector] (dot1) -- ([yshift=0mm]box1.north) node[midway] (conn1) {};

\node[box] (box2) at (\xstep, 4.5) {Introduces Epistemic AI and a novel uncertainty method called Random-Set Neural Networks.};
\node[subtitle] (sub2) at (\xstep, 1) {Epistemic Deep\\ Learning};
\node[title] (title2) at (\xstep, 1.9) {PART II};
\node[dot] (dot2) at ([yshift=8mm]sub2.north) {};
\draw[connector] (dot2) -- ([yshift=0mm]box2.south) node[midway] (conn2) {};

\node[title] (title3) at ({2*\xstep}, 5.2) {PART III};
\node[subtitle] (sub3) at ({2*\xstep}, 4.3) {Evaluating Epistemic\\ Predictions};
\node[box] (box3) at ({2*\xstep}, 1.4) {Proposes a unified evaluation framework to compare epistemic predictions across methods.};
\node[dot] (dot3) at ([yshift=-1mm]sub3.south) {};
\draw[connector] (dot3) -- ([yshift=0mm]box3.north) node[midway] (conn3) {};

\node[box] (box4) at ({3*\xstep}, 4.5) {Demonstrates applicability on Large Language Models (LLMs) and weather classification in autonomous racing.};
\node[subtitle] (sub4) at ({3*\xstep}, 1.2) {Applications};
\node[title] (title4) at ({3*\xstep}, 1.9) {PART IV};
\node[dot] (dot4) at ([yshift=8mm]sub4.north) {};
\draw[connector] (dot4) -- ([yshift=0mm]box4.south) node[midway] (conn4) {};

\draw[arrow] ([xshift=2.5mm]conn1.center) -- ++(3.5cm, 0);
\draw[arrow] ([xshift=2.5mm, yshift=0mm]conn2.center) -- ++(3.5cm, 0);
\draw[arrow] ([xshift=2.5mm, yshift=1mm]conn3.center) -- ++(3.5cm, 0);
\end{tikzpicture}
\caption{Overview of the thesis structure.}
\label{fig:thesis-structure}
\vspace{2mm}
\end{figure}
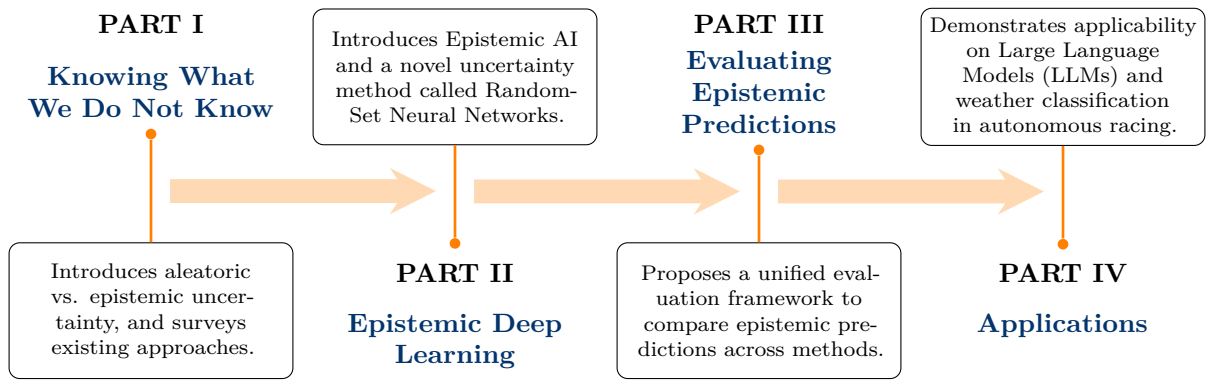

The thesis is organized into \textbf{four} main parts as shown in Fig. \ref{fig:thesis-structure}. \textbf{Part \ref{part:unc} (\textit{Knowing What We Do Not Know})} introduces the conceptual foundations of uncertainty in machine learning (Chapter \ref{ch:unc_quant}), why uncertainty quantification matters  (Sec. \ref{ch:why_unc}), and the two main sources of uncertainty: \textit{aleatoric} uncertainty (stems from inherent data noise) and \textit{epistemic} uncertainty (due to limited knowledge) (Sec. \ref{sec:sources-of-unc}). It surveys a broad spectrum of existing and emerging models for representing epistemic uncertainty (Chapter \ref{ch:unc_models}). This part lays the foundation for robust epistemic modeling and motivates the central research questions addressed throughout the thesis.

\textbf{Part \ref{part:epi} (\textit{Epistemic Deep Learning})} introduces the paradigm of \textbf{Epistemic Artificial Intelligence} (Chapter \ref{ch:epi_ai}), which emphasizes learning from ignorance, and details the use of second-order uncertainty measures (Chapter \ref{ch:knowledge}) to model a lack of knowledge (ignorance). The thesis then presents a novel \textbf{Random-Set Neural Network (RS-NN)} (Chapter \ref{ch:rsnn}) approach to classification which predicts \textit{belief functions} (rather than classical probability vectors) over the class list using the mathematics of random sets. RS-NN encodes the {`epistemic'} uncertainty induced by training sets that are insufficiently representative or limited in size via the size of the convex set of probability vectors associated with a predicted belief function. RS-NN outperforms state-of-the-art Bayesian and Ensemble methods (Sec. \ref{sec:rsnn-experiments}) in terms of \textit{accuracy}, \textit{uncertainty estimation} and \textit{out-of-distribution (OoD) detection} on multiple benchmarks (CIFAR-10 \textit{vs.} SVHN/Intel-Image, MNIST \textit{vs.} FMNIST/KMNIST, ImageNet \textit{vs.} ImageNet-O). RS-NN also scales up effectively to large-scale architectures (\textit{e.g.} WideResNet-28-10, VGG16, Inception V3, EfficientNetB2 and ViT-Base-16), exhibits remarkable robustness to \textit{adversarial attacks} and can provide statistical guarantees in a conformal learning setting (\S\ref{app:statistical}). RS-NN is extended to the domain of text generation through the development of \textbf{Random-Set Large Language Models (RS-LLMs)} in Part \ref{part:app} \textit{(Applications)}, Chapter \ref{ch:rsllm}. Further collaborative research on \textbf{credal set models}, another family of second-order uncertainty representations, is briefly discussed in Chapter \ref{ch:cre_models}.


\textbf{Part \ref{part:eval} (\textit{Evaluating Epistemic
Predictions})} focuses on the evaluation of uncertainty-aware models, a largely underexplored topic. It discusses related work on evaluation under uncertainty (Chapter \ref{ch:eval_unc}) and introduces \textbf{a unified evaluation framework for assessing epistemic predictions} (Chapter \ref{ch:eval_framework}). This includes new metrics for gauging both \textit{accuracy} (distance-based) and \textit{imprecision} from predictions, supported by experimental analysis and ablation studies (Sec. \ref{sec:experiments}).

\textbf{Part \ref{part:app} \textit{(Applications)}} showcases the practical implementation of Random-Set Neural Networks (RS-NNs) through two case studies: first, their extension to Large Language Models (LLMs) for text generation, leading to the creation of \textbf{Random-Set Large Language Models (RS-LLMs)} (Chapter \ref{ch:rsllm}); and second, their use in \textbf{weather classification for autonomous racing} (Chapter \ref{ch:weather}). These applications highlight the value of modeling epistemic uncertainty to develop more robust, transparent, and safer machine learning systems in real-world scenarios.

The final chapter \textbf{concludes} the thesis (Chapter \ref{ch:conclusion}), providing summaries of key contributions and methods (Sec. \ref{sec:thesis-summary}), as well as discussions on \textbf{limitations} (Sec. \ref{sec:limitations}) and \textbf{future research directions} (Sec. \ref{sec:future-directions}) for each method. Additionally, the broader \textbf{impact of the epistemic deep learning models} proposed in this thesis on the emerging field of Epistemic AI is explored in Sec. \ref{sec:impact-epi}. The chapter further elaborates on \textbf{open challenges} (Sec. \ref{sec:challenges}) and prospects for the \textbf{future of Epistemic AI} (Sec. \ref{sec:opportunities}). The Appendix contains detailed algorithms for the methods (\S\ref{app:algo}) and statistical guarantees for RS-NN under a conformal setting (\S\ref{app:statistical}). The code for the experiments presented in this work is available at \url{https://github.com/shireenkmanch}.

\section{Contributions}

This thesis contributes a novel class of models based on random sets and establishes a principled framework for evaluating uncertainty-aware models in deep learning. The key contributions are as follows:

\begin{itemize}
    \item \textbf{\textcolor{myblue}{Random-Set Neural Networks (RS-NN):}} A novel class of classifiers grounded in the theory of random sets, which enables models to represent and reason about epistemic uncertainty in a mathematically rigorous way. RS-NNs allow models to express structured ignorance by outputting sets of plausible labels rather than single-point predictions, a capability unavailable in standard probabilistic approaches. Importantly, this random-set framework is not limited to classification tasks; it can also be applied to complex problems such as text generation in large language models, providing a new avenue for uncertainty-aware natural language processing.

    \item \textbf{\textcolor{myblue}{A scalable framework for controlling set complexity:}} This thesis introduces an innovative budgeting mechanism to manage and limit the complexity of random sets, making second-order uncertainty representations both practical and interpretable. Rather than relying on the full power set of possibilities (which is often computationally infeasible), this approach strategically selects a representative subset of sets. This enables random-set models to maintain strong performance and reliable uncertainty quantification without the overhead of handling all possible combinations.

    \item \textbf{\textcolor{myblue}{A Unified Evaluation Framework for Epistemic Predictions:}} To enable fair comparison between epistemic uncertainty quantification methods that produce heterogeneous outputs (\textit{e.g.}, distributions, intervals, sets), this thesis proposes a model-agnostic evaluation framework. This framework standardizes key performance metrics for epistemic models and provides guidance for selecting appropriate methods based on task-specific needs.
\end{itemize}

\section{Dissemination}

This thesis is based in part on the following publications, preprints, and dissemination activities:

\subsection*{Peer-reviewed Publications}

\begin{enumerate}
    \item \paper{\textbf{Manchingal, S. K.}, Mubashar, M., Wang, K., Shariatmadar, K., \& Cuzzolin, F}{Random-Set Neural Networks}{Proceedings of the Thirteenth International Conference on Learning Representations}{ICLR}{2025}

    \item \paper{\textbf{Manchingal, S. K.}, Mubashar, M., Wang, K., \& Cuzzolin, F}{A Unified Evaluation Framework for Epistemic Predictions}{Proceedings of the 28th International Conference on Artificial Intelligence and Statistics}{AISTATS}{2025} 
    
    \item \paper{Wang, K., Shariatmadar, K., \textbf{Manchingal, S. K.}, Cuzzolin, F., Moens, D., \& Hallez, H}{CreINNs:Credal-Set Interval Neural Networks for Uncertainty Estimation in Classification Tasks}{}{Neural Networks}{2025}

    \item \paper{Wang, K., Cuzzolin, F., \textbf{Manchingal, S. K.}, Shariatmadar, K., Moens, D., \& Hallez, H}{Credal Deep Ensembles for Uncertainty Quantification}{Advances in Neural Information Processing Systems}{NeurIPS}{2024}

    \item \paper{\textbf{Manchingal, S. K.}, \& Cuzzolin, F}{Epistemic Deep Learning}{ICML 2022 Workshop on Distribution-free Uncertainty Quantification}{ICML-DFUQ}{2022}

\end{enumerate}





\subsection*{Workshops}

In addition to the publications, {the author} also participated in organizing the following workshop:

\begin{itemize}
    \item The First Workshop on Epistemic Uncertainty in Artificial Intelligence (E-pi), International Conference on Uncertainty in Artificial Intelligence (UAI) 2023.
    \subitem \emph{Workshop link}: \url{https://sites.google.com/view/epi-workshop-uai-2023/home}
    \subitem \emph{Workshop proceedings}: \url{https://link.springer.com/book/10.1007/978-3-031-57963-9}
\end{itemize}

\subsection*{Software and Reproducibility}

To promote transparency and reproducibility, all developed codebases and experiments have been made available:

\begin{itemize}
    \item Random-Set Neural Networks (\textit{Codebase}): \url{https://github.com/shireenkmanch/Random-Set-Neural-Networks}
    \item A Unified Evaluation Framework for Epistemic Predictions (\textit{Codebase}): \url{https://github.com/shireenkmanch/Evaluation-Epistemic-Preds}
    \item CreINNs: Credal-Set Interval Neural Networks for Uncertainty Estimation in Classification Tasks (\textit{Codebase}): \url{https://github.com/WangKaizheng/CreINNs}
\end{itemize}

\part{{{\MakeUppercase{Knowing What We Do Not Know}}}}
\label{part:unc}
\thispagestyle{plain}
\chapter{Background: Uncertainty Quantification in Deep Learning} 
\label{ch:unc_quant} 


Machine learning is fundamentally about building models from data, often with the goal of making predictions. A core challenge in this process is dealing with \textit{uncertainty}. When a machine learning model `learns', it generalizes from the specific examples it has seen to broader patterns or rules that apply beyond the training data. This process, called induction, inherently involves uncertainty because the model is only an approximation—it can never be proven absolutely correct. Instead, models are hypotheses about how the data were generated, so their predictions come with some degree of uncertainty.

Moreover, uncertainty in machine learning does not just come from the inductive nature of learning. Other sources include incorrect assumptions within the model and noise or errors in the data itself. These factors can affect the reliability of the model’s predictions.
Representing and managing this uncertainty accurately is essential, especially in safety-critical applications such as autonomous driving \cite{fort2019large}, medical diagnosis \cite{lambrou2010reliable}, flood uncertainty estimation \cite{chaudhary2022flood}, {and} structural health monitoring \cite{vega2022variational}, where incorrect decisions can have serious consequences. 

Uncertainty is also a fundamental concept within machine learning techniques themselves. For example, in active learning \cite{aggarwal2014active}, algorithms focus on reducing uncertainty by selectively querying the most informative data points. Similarly, decision tree algorithms \cite{mitchell1980need} use measures of uncertainty (like information gain) to decide how to split data and build the model.
Machine learning models often
struggle to
provide reliable predictions when confronted with unfamiliar data \cite{guo2017calibration, ovadia2019can, minderer2021revisiting}, {be it} 
noisy samples \cite{papernot2016limitations} deliberately designed to deceive models, or out-of-distribution data (OoD) beyond the model's training distribution. 
An ideal learning system, especially when deployed in safety-critical applications, should be aware of how confident it is
in its own predictions, acknowledge the limits of its knowledge and gauge these limitations to make informed decisions by modelling the {uncertainty} associated with it, in essence, \textit{`to know when it does not know’}.

\section{Sources of Uncertainty: Aleatoric vs. Epistemic}
\label{sec:sources-of-unc}

The defining issue for AI now is how it manages uncertainty and leverages data, with a key distinction between \emph{epistemic} (reducible) and \emph{aleatoric} (irreducible) uncertainty. The latter refers to the different outcomes a random experiment produces, for example, when we play roulette, we cannot know in advance what the outcome of each single spin is, but can predict that they will arise (over time) with a known frequency (1/36). {By contrast}, the former express a more essential lack of knowledge about what underlying process drives the observed phenomenon. A player faced with the choice of playing one of a number of roulette wheels, without knowing the workings of each wheel (\textit{i.e.}, what probability distribution models it), will be subject to such `epistemic’ uncertainty. A similar distinction between `common cause’ and `special cause’ \cite{snee1990statistical} is central in the philosophy of probability \cite{keynes2013treatise}. Unlike uncertainty caused by randomness (aleatoric), uncertainty due to ignorance (epistemic) can, in principle, be reduced with the acquisition of additional information \cite{hullermeier2021aleatoric}. {However, this reduction is theoretically guaranteed only in the limit of infinite data, assuming that the data sufficiently covers the input space and that the learning algorithm has sufficient capacity to approximate the true underlying distribution.}

{That said, the relevance of this distinction is not universally agreed upon. In many real-world machine learning scenarios, models are trained to optimize performance rather than disentangle the origins of their uncertainty. When predictions must be made regardless of data limitations, the source of uncertainty, whether due to inherent noise or lack of knowledge, is often treated uniformly. Recent work \cite{de2024disentangled, mucsanyi2024benchmarking} questions whether aleatoric and epistemic uncertainties are cleanly separable in high-capacity models, where incomplete knowledge and stochastic variability often interact.}

The main source of uncertainty in AI (but also in its science and engineering applications) is the lack of a sufficient amount of data to train a model, in both quantity and quality (\textit{i.e.}, data fairly describing all regions of operation, including rare events). This uncertainty is \textbf{epistemic in nature}, as it concerns the model itself, and can be reduced by collecting more data or information. 

Epistemic uncertainty can be modeled at two levels: (i) a \textit{target} level, where the network outputs an uncertainty measure on the target space, while its parameters (weights) remain deterministic; (ii) a \textit{parameter} level, where uncertainty is modeled on the parameter space (\textit{i.e.}, weights and biases). This thesis presents a novel method to feasibly implement random sets, during training and inference, to represent uncertainty at the target level. The extension of this approach to model uncertainty at the parameter level using random sets is reserved for future work (Sec. \ref{sec:future-directions}).
In this thesis, we focus mainly on epistemic uncertainty, especially in the context of \textbf{predictive uncertainty over the target space}, as accurately quantifying it remains a central challenge in AI.

\section{Predictive Uncertainty}\label{sec:predictive-undertainty}

While aleatoric uncertainty may stem from inherent ambiguities in the data or inaccuracies in class annotations, epistemic uncertainty arises from our incomplete knowledge about the true underlying model parameters governing the relationship between input features and class labels. Epistemic uncertainty represents the disparity between the true distribution $p^{*}$ and and the model's estimate $\hat{p}$. In practice, {{aleatoric uncertainty is almost always computed by the UQ community as the \textbf{difference between predictive (total) uncertainty and epistemic uncertainty}}}. Consequently, the primary focus of this thesis is on computing predictive and epistemic uncertainty.

In 
supervised learning tasks, e.g, 
classification, 
where the objective is to assign class labels $y \in \mathbf{Y}$ to input data points $\mathbf{x} \in \mathbf{X}$, \emph{predictive uncertainty} can be represented using the predictive distribution $\hat{p}(y \mid \mathbf{x}, \mathbb{D})$, where $y \in \mathbf{Y}$ is a label, $\mathbf{x} \in \mathbf{X}$ an input instance, and $\mathbb{D} = \{ (\mathbf{x}_i, y_i)\}_{i=1}^\mathcal{N} \in \mathbf{X} \times \mathbf{Y}$ 
the available training set, $\mathcal{N}$ being the number of 
training instances.
The true data-generating distribution
remains unknown to us, in turn implying uncertainty about the true predictive distribution.
A model will provide its best estimate 
$\hat{p}(y | \mathbf{x}, \mathbb{D})$
based on the available data $\mathbb{D}$.

As introduced earlier, this thesis proposes a novel model based on random sets that predicts epistemic uncertainty over sets of probability distributions. Since \textbf{{epistemic uncertainty is defined differently across various uncertainty-aware model}s}, direct comparison of these measures is not straightforward. Therefore, we propose an evaluation framework that enables comparison between models predicting both predictive entropy and epistemic uncertainty, despite differences in their formulations, by leveraging their predictions within a unified metric space.

\section{Why Uncertainty Quantification Matters} 
\label{ch:why_unc} 

As mentioned in Sec. \ref{app:motivation}, uncertainty quantification is crucial for improving the reliability and safety of machine learning models, particularly in real-world, safety-critical applications. Without a clear understanding of prediction confidence, models can exhibit \textit{overconfidence}, \textit{fail to adapt to new domains}, and \textit{make unsafe decisions}. This section highlights key motivations for uncertainty quantification, including enhancing adversarial robustness, improving domain adaptation, calibrating confidence estimates, and supporting safer sequential decision-making. Together, these aspects demonstrate why accurate modeling of uncertainty is essential for trustworthy AI systems.

\subsection{Adversarial Robustness}

Traditional neural networks often suffer from overconfidence, producing highly confident predictions even when incorrect, which can lead to serious consequences in safety-critical applications.
This issue arises because softmax probabilities represent relative confidence rather than true uncertainty, causing these models to assign high confidence to out-of-distribution (OoD) inputs or adversarially perturbed data \cite{guo2017calibration, hendrycks2016baseline}. 

\begin{figure*}[!h]
\centering
    \vspace{-15pt}
    \includegraphics[width=0.7\textwidth]{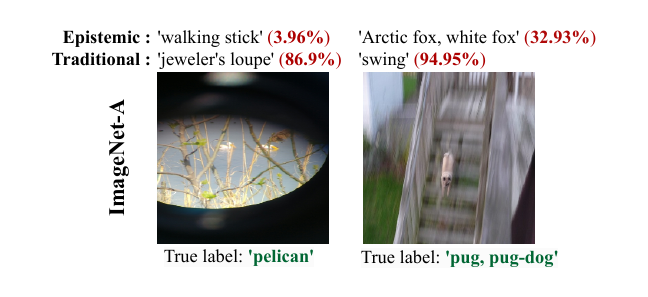}
    \vspace{-15pt}
    \caption[Confidence scores of uncertainty-aware (Epistemic) and standard model with no uncertainty estimation (Traditional) for samples of ImageNet-A (adversarial) dataset.]{Confidence scores of uncertainty-aware (Epistemic) and standard model with no uncertainty estimation (Traditional) for samples of ImageNet-A (adversarial) dataset. Traditional model is overconfident in a misclassification with confidence scores of 86.9\% and 94.95\%, whereas the epistemic model exhibits lower confidence with scores of 32.93\% and 3.96\% for misclassifications.}
      \label{fig:imagenet-a}
\end{figure*}

For instance, Fig. \ref{fig:imagenet-a} shows incorrect predictions by a standard ResNet50 (Traditional) and an uncertainty-aware model (Epistemic) based on ResNet50 over a classification problem on the ImageNet-A dataset. The uncertainty-aware (Epistemic) model used here is Random-Set Neural Network (Chapter \ref{ch:rsnn}). ImageNet-A (adversarial) \cite{hendrycks2021natural} is a challenging adversarially filtered dataset designed to deliberately expose the overconfidence and poor generalization of models. As shown in Fig. \ref{fig:imagenet-a}, when a standard ResNet50 with no uncertainty estimation and an uncertainty-aware model based on ResNet50, both trained on ImageNet are tested on ImageNet-A, the standard model tends to exhibit high confidence in incorrect predictions, whereas the uncertainty-aware model assigns lower confidence to misclassifications, avoiding overconfidence.

\subsection{Robustness and Domain Adaptation}

Robustness in machine learning often focuses on domain adaptation \cite{awais2021adversarial, alijani2024vision}, using methods like minimax learning \cite{azar2017minimax}, counterfactual error bounding \cite{swaminathan2015self}, and custom loss functions to adapt models to target domains. In unsupervised domain adaptation, adversarial approaches reduce the gap between source and target domain features. While reinforcement learning has less focus on robustness, some work includes adversarial methods \cite{pinto2017robust} and Bayesian Bellman formulations \cite{derman2020bayesian}. Recent research on out-of-distribution (OoD) generalization and domain generalization (DG) addresses discrepancies between training and test distributions \cite{gulrajani2020search, koh2021wilds, singh2024domain}, with theoretical work on kernel methods \cite{blanchard2011generalizing, deshmukh2019generalization, hu2020domain, 10.5555/3042817.3042820} and domain-adversarial learning using H-divergence \cite{albuquerque2019generalizing}. However, these methods struggle when training and test distributions differ significantly \cite{rosenfeld2022online}, and managing uncertainty can improve adaptation and OoD detection.

\subsection{Out-of-distribution (OoD) detection}

Out-of-distribution (OoD) detection refers to a model’s ability to identify inputs that deviate significantly from the training data distribution. In real-world scenarios, a model may encounter inputs that lie outside the support of its training distribution \cite{hendrycks2016baseline}. Making overconfident predictions on such OoD inputs can lead to erroneous and potentially harmful outcomes. Thus, incorporating uncertainty estimation is essential to enable models to recognize unfamiliar inputs and express appropriate caution \cite{amodei2016concrete}.

Quantitatively, OoD detection performance is commonly evaluated using metrics such as the Area Under the Receiver Operating Characteristic curve (AUROC) and the Area Under the Precision-Recall Curve (AUPRC). AUROC assesses the model’s ability to distinguish between in- and out-of-distribution samples based on uncertainty scores, while AUPRC is particularly informative in imbalanced settings, where OoD instances are rare.

The OoD detection procedure (detailed in Algorithm \S\ref{app:algo-auroc}) begins by aggregating uncertainty scores from in-distribution (iD) and OoD inputs. Binary labels are assigned (0 for iD, 1 for OoD), and these are combined with the corresponding scores to compute the ROC curve, which plots the true positive rate (TPR) against the false positive rate (FPR) across various thresholds. The AUROC is then computed as the area under this curve. Likewise, a precision-recall (PR) curve is generated, and the AUPRC is obtained. Higher AUROC and AUPRC values indicate better separation between iD and OoD samples, reflecting the model’s ability to assign lower confidence (higher uncertainty) to unfamiliar inputs.

\begin{figure}[!h]
\centering
    \includegraphics[width=0.7\textwidth]{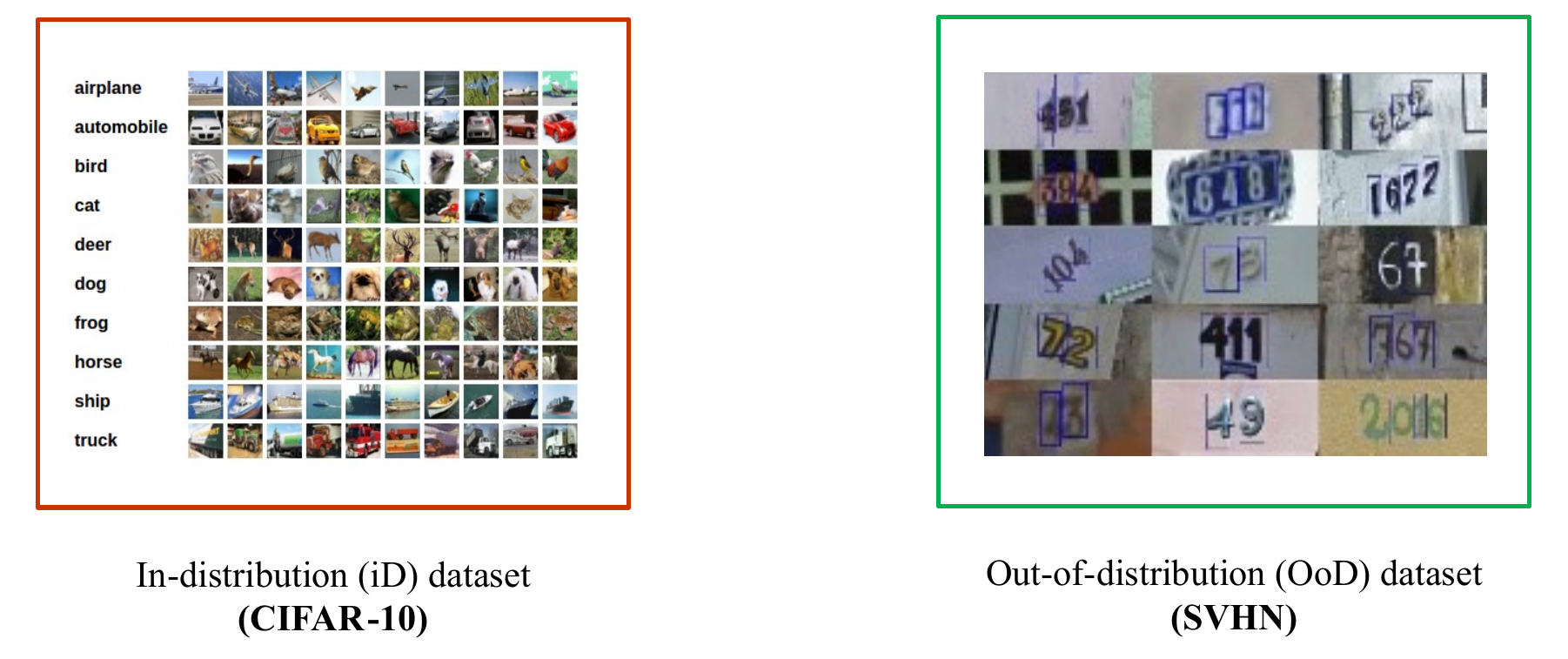}
    \caption{In-distribution (\textit{left}) \textit{vs.} out-of-distribution (\textit{right}) datasets.}
      \label{fig:ood}
\end{figure}

{In this thesis, out-of-distribution (OoD) data refers to inputs that differ significantly from the distribution of the training data, meaning they represent scenarios or classes that the model has not encountered during training. For example, if a model is trained on CIFAR-10 \cite{cifar10}, which consists of natural images of objects like cats, cars, and airplanes, then inputs from the SVHN \cite{netzer2011reading} dataset composed of street-view house number digits would be considered OoD (Fig. \ref{fig:ood}), since they represent a fundamentally different visual domain. }

{
It is important to note that there are multiple perspectives on what constitutes OoD data: some define OoD strictly as samples from entirely new classes not present during training, while others also consider shifts in data style, context, or domain as forms of OoD, even if the underlying classes overlap. }

\subsection{Calibration}

Neural networks are typically uncalibrated, meaning that when a network predicts $\hat{y}$ with confidence $\delta$, the fraction of correct predictions at confidence $\delta$ should be equal to $\delta$, which often does not occur \cite{guo2017calibration}. To address this, various calibration techniques \cite{ojeda2023calibrating} like histogram binning, Bayesian binning, and Platt scaling have been proposed \cite{platt1999probabilistic}, with extensions to regression \cite{kuleshov2018accurate}. 
{Metrics such as Expected Calibration Error (ECE) \cite{naeini2015obtaining} and Adaptive Calibration Error (ACE) \cite{nixon2019measuring} are commonly used to quantify calibration error. Loss-based adjustments \cite{mukhoti2020calibrating, luo2022local, tao2023dual} aim to actively improve model calibration, though they often overlook deeper issues in uncertainty representation. }

\subsection{Sequential decision-making} 

In sequential decision-making tasks such as autonomous driving, the decisions made at each step influence future actions and outcomes. Unlike single-shot predictions, where uncertainty might only impact a single decision, sequential decision-making requires a model to account for uncertainty over time. This is particularly important because each decision made in a sequence can have long-term consequences \citep{schwarting2018planning}, and failure to properly manage uncertainty can result in compounded errors. For instance, in autonomous driving \citep{teeti2023temporal}, a vehicle must make real-time decisions about route planning, vehicle control, and responding to dynamic environments (\textit{e.g.}, pedestrians, traffic signals, or road hazards). If uncertainty in perception or state estimation is not adequately modeled, the vehicle might make decisions that could lead to unsafe situations, such as misinterpreting a pedestrian’s motion or failing to account for out-of-date maps. Effective uncertainty quantification \citep{kendall2017uncertainties, depeweg2018decomposition}, particularly epistemic uncertainty, helps the system understand when it is uncertain about its predictions, enabling more cautious and informed decision-making.

\chapter{Related Work: Machine Learning Models For Representing Uncertainty} 
\label{ch:unc_models} 
\label{app:uncertainty-models}
\label{sec:uncertainty-models}
\label{app:A}

The machine learning community has recognized the challenge of estimating uncertainty in model predictions, leading to the development of several 
Bayesian approximations \cite{gal2016dropout, charpentier2020posterior}, evidential Dirichlet models \cite{sensoy, gao2024comprehensive} and conformal prediction \cite{shafer2008tutorial, 10.5555/1712759.1712773, balasubramanian2014conformal, crossconformal, angelopoulos2021gentle}.
Ensemble-based approaches, such as Deep Ensembles (DE) \cite{lakshminarayanan2017simple} and Epistemic Neural Networks (ENN) \cite{osband2024epistemic}, estimate uncertainty by leveraging multiple models. However, the computational cost of training ensembles, especially for large models, is often impractical.
Some methods rely on prior knowledge \cite{https://doi.org/10.48550/arxiv.2105.06868}, whereas others require setting a desired threshold on predictions \cite{angelopoulos2021gentle}.
Some
\cite{baron1987second, hullermeier2021aleatoric} have argued that classical probability is not equipped to model `second-level' uncertainty on the probabilities themselves. This has led to the formulation of numerous
uncertainty calculi \cite{cuzzolin2020geometry}, 
including
possibility theory \cite{Dubois90}, probability intervals
\cite{halpern03book}, credal sets \cite{levi1980enterprise}, 
random sets \cite{Nguyen78} or imprecise probability \cite{walley91book}. 

\begin{figure*}[!ht]
\hspace{-18pt}
    \includegraphics[width=1.06\textwidth]{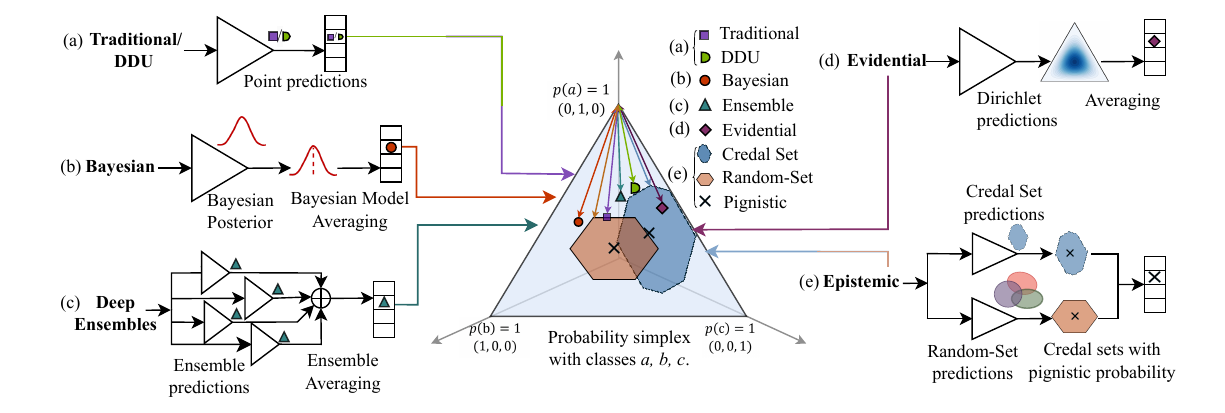}
    \caption[Major approaches to uncertainty in AI: (a) Traditional networks and Deterministic uncertainty models, (b) Bayesian neural networks, (c) Deep ensembles, (d) Evidential approaches, and (e) Epistemic approaches.]{\textbf{Major approaches to uncertainty in AI.} While traditional networks and Deterministic uncertainty models \cite{mukhoti2023deep} (a) have deterministic weights, and output either a determistic output value or a probability vector in the output space, Bayesian neural networks \cite{blundell2015weight, gal2016dropout, rudner2022tractable} (b) compute a predictive distribution there by integrating over a learnt posterior distribution of model parameters given training data. As this is often infeasible due to the complexity of the posterior, Bayesian Model Averaging (BMA) \cite{hinne2020conceptual, graefe2015limitations} is often used to approximate the predictive distribution by averaging over predictions from multiple samples. The same (averaging) technique can be applied to (deep) ensembles of networks \cite{lakshminarayanan2017simple} (c), in which a number of models are trained independently using different initializations. Evidential approaches \cite{sensoy} (d) make predictions as parameters of a second-order Dirichlet distribution on the output space, instead of softmax probabilities. Finally, epistemic approaches (e) employ second-order probability representations, either in the output space or the target space, \textit{e.g.} in the form of interval probabilities, credal sets or random sets \cite{manchingal2025randomsetneuralnetworksrsnn}. A pignistic probability \cite{SMETS2005133} estimate can be obtained from credal sets for comparison to other models \cite{manchingal2025unifiedevaluationframeworkepistemic}.
}
      \label{fig:uncertainty-models}
      \vspace{-2mm}
\end{figure*}

This chapter surveys related work in uncertainty-aware  ML models for classification tasks, outlining the various types of uncertainty-aware approaches, their methods for computing predictive uncertainty, and the nature of their resulting predictions.
The predictions of a classifier can be plotted in the simplex (convex hull) of the one-hot probability vectors assigning probability 1 to a particular class.
For instance, in a $3$-class classification scenario ($\mathbf{Y}= \{a, b, c\}$), 
the simplex would be a 2D simplex (triangle) connecting three points, each representing one of the classes, as shown in Fig. \ref{fig:uncertainty-models} (center).
Fig. \ref{fig:uncertainty-models} shows major uncertainty-aware models in AI and their predictions on a 3-class probability simplex.

\section{Traditional and Deterministic methods}

Traditional machine learning models provide deterministic predictions for classification (categorical labels) or regression tasks (continuous values), and inherently lacks mechanisms to capture epistemic uncertainty, as they assume precise knowledge of the dependency between inputs and outputs. {One type of} standard model is denoted as CNN (Convolutional Neural Network) in this thesis, as it mainly focuses on image classification.
Traditional/standard neural networks (CNN)
predict a vector of $N$ scores, one for each class, duly \emph{calibrated} to a probability vector representing a (discrete, categorical)
probability distribution over the list of classes $\mathbf{Y}$, 
\begin{equation}
    \hat{p}_{s}(y \mid \mathbf{x}, \mathbb{D}),
\end{equation}
which represents the probability of observing class $y$ given the input $\mathbf{x}$ and training data $\mathbb{D}$.

A recent development, Deep Deterministic Uncertainty (DDU) \cite{mukhoti2023deep} estimates epistemic uncertainty by analyzing latent representations or using distance-sensitive functions, rather than predicted softmax probabilities \cite{alemi2018uncertainty, wu2020simple, liu2020simple, mukhoti2023deep, van2020uncertainty}. However, regularization techniques based on feature space distances, such as bi-Lipschitz regularization (used by most of these models), do not effectively improve OoD detection or calibration \cite{postels2021practicality}. This is due to the limitations of distance metrics in high-dimensional data, such as images. These methods have been applied to areas like object detection \cite{gasperini2021certainnet} but overlooks calibration: how well uncertainty reflects model performance under shifting distributions. Calibration is crucial for safe deployment but remains underexplored, as {DDUs} are typically tested only on simple tasks and datasets, leaving their effectiveness in more complex scenarios unproven \cite{postels2021practicality}. Uncertainty estimates are poorly calibrated under distributional shifts, especially when compared to scalable Bayesian methods. 

Moreover, unlike other uncertainty-aware baselines, DDU represents uncertainty in the input space (by identifying an input sample as in- or out-of-distribution) rather than the prediction space.  
As a result, DDU provides predictions $\hat{p}_{ddu}(y \mid \mathbf{x}, \mathbb{D})$ in the form of softmax probabilities akin to traditional neural networks. Both DDU and traditional models make point predictions (see Fig. \ref{fig:uncertainty-models}).


\section{Bayesian Methods}
\label{app:A_BNN}

Bayesian Deep Learning (BDL) \cite{Buntine1991BayesianB, 10.1162/neco.1992.4.3.448, neal2012bayesian} is the most well-known approach to uncertainty quantification in AI. It leverages Bayesian neural networks (BNNs) \cite{blundell2015weight, gal2016dropout, DBLP:journals/corr/abs-2007-06823} to model network parameters (\textit{i.e.}, weights and biases) as distributions, predicting a `second-order' distribution, a distribution of distributions \cite{hullermeier2021aleatoric}, with predictions often generated by running the network on sample parameters extracted from an approximated posterior.

Bayesian uncertainty quantification in deep neural networks is commonly approached through three main methods. (1) Variational inference methods aim to approximate the intractable posterior by minimizing divergence to a simpler, tractable distribution \citep{blundell2015weight, https://doi.org/10.48550/arxiv.1312.6114, gal2016dropout, louizos2017multiplicative, zhang2018noisy}. A well-known example is Monte Carlo Dropout \citep{gal2016dropout}, which interprets dropout as a Bernoulli-distributed random variable and enables efficient posterior sampling at inference time. (2) Sampling-based approaches typically rely on Markov Chain Monte Carlo (MCMC) methods \citep{neal2012bayesian, welling2011bayesian, chen2016bridging}, which iteratively generate samples that approximate the posterior distribution, providing asymptotically exact inference. (3) Laplace approximation methods simplify the posterior by performing a second-order Taylor expansion around the mode of the log-posterior \citep{10.1162/neco.1992.4.3.448, osawa2019practical, dusenberry2020efficient, ritter2018scalable}. Recognizing that many eigenvalues of the Hessian are near zero, \citet{ritter2018scalable} propose a low-rank approximation that yields efficient and scalable posterior covariance estimation, enabling the application of Laplace methods to large-scale models.
Notable techniques include 
R-BNN (Reparameterisation) \cite{https://doi.org/10.48550/arxiv.1506.02557},
variational inference with reparameterisation
and Laplace Bridge Bayesian approximation (LB-BNN) \cite{hobbhahn2022fast}, which uses the Laplace Bridge to map between Gaussian and Dirichlet distributions. Various approximations of full Bayesian inference exist, such as Markov chain Monte Carlo (MCMC) \cite{NIPS1992_f29c21d4,4767596},  variational inference (VI) \cite{NIPS2011_7eb3c8be}, function-space BNNs \cite{sun2019functional} such as function-space variational inference (FSVI) \cite{rudner2022tractable}, and Dropout Variational Inference \cite{https://doi.org/10.48550/arxiv.1506.02158}.
In our experiments, we do not consider older Bayesian models such as R-BNN, MCMC and Dropout VI, since they 
have been superseded by {higher} performing approaches
and are computationally expensive to train on larger datasets and architectures.

Despite the development of efficient training techniques such as variational inference \cite{blundell2015weight, gal2016dropout, https://doi.org/10.48550/arxiv.1506.02557, https://doi.org/10.48550/arxiv.1506.02158, sun2019functional, rudner2022tractable, https://doi.org/10.48550/arxiv.1506.02158}, Laplace approximations \cite{daxberger2021laplace, osawa2019practical, dusenberry2020efficient} and sampling methods \cite{hoffman2014no, neal2011mcmc} and achieving some success in real-world tasks
\cite{vega2022variational, kwon2020uncertainty},
practical challenges remain. This includes the significant computational complexity associated with training and inference \cite{DBLP:journals/corr/abs-2007-06823}, establishing appropriate prior distributions before training \cite{https://doi.org/10.48550/arxiv.2105.06868}, handling complex network architectures, and ensuring real-time applicability.
Furthermore, several studies have indicated that the use of single probability distributions to model the lack of knowledge is insufficient \cite{caprio2023imprecise}. This is further discussed in this thesis in Sec. \ref{sec:why-epi}. Recent work addresses challenges in computational cost \cite{hobbhahn2022fast, daxberger2021laplace} and model priors \cite{tran2020all}. However, Bayesian models will always face challenges when the model prior is misspecified.

Bayesian Neural Networks (BNNs) \cite{lampinen2001bayesian, titterington2004bayesian, goan2020bayesian, hobbhahn2022fast} 
compute a predictive distribution $\hat{p}_{b}(y \mid \mathbf{x}, \mathbb{D})$ by integrating over a learnt posterior distribution of model parameters $\theta$ given training data $\mathbb{D}$. 
This
is often infeasible due to the complexity of the posterior, leading to the use of \emph{Bayesian Model Averaging} (BMA), which approximates the predictive distribution by averaging over predictions from multiple samples. When applied to classification, BMA yields
point-wise predictions. 


Bayesian inference integrates over the posterior distribution $p(\theta \mid \mathbb{D})$
over model parameters $\theta$ given training data $\mathbb{D}$ to compute the predictive distribution $\hat{p}_{b}(y \mid \mathbf{x}, \mathbb{D})$, reflecting updated beliefs after observing the data:
\begin{equation}
    \hat{p}_{b}(y \mid \mathbf{x}, \mathbb{D}) = \int p(y \mid \mathbf{x}, \theta) p(\theta \mid \mathbb{D}) d\theta,
\end{equation}
where $p(y \mid \mathbf{x}, \theta)$ represents the likelihood function of observing label $y$ given $x$ and $\theta$. 
To overcome the infeasibility of this integral, direct sampling from $\hat{p}_{b}(y \mid \mathbf{x}, \mathbb{D})$ using methods such as Monte-Carlo \cite{hastings1970monte} are applied to obtain a large set of sample weight vectors, $\{ \theta_k, k\}$, from the posterior distribution. These sample weight vectors are then used to compute a set of possible outputs $y_k$,
namely: 
\begin{equation}
    \hat{p}_{b}(y_k\mid \mathbf{x}, \mathbb{D}) = \frac{1}{|\Theta|} \sum_{\theta_k \in \Theta} \Phi_{\theta_k}(\mathbf{x}),
\end{equation}
where $\Theta$ is the set of sampled weights, $\Phi_{\theta_k}(\mathbf{x})$ is the prediction made by the model with weights $\theta_k$ for input $\mathbf{x}$, and $\Phi$ is the function for the model. This process is called \emph{Bayesian Model Averaging (BMA)}.

\begin{figure}[!ht]
    \centering
    \begin{minipage}[t]{0.49\textwidth}
        \centering
        \includegraphics[width=\textwidth]{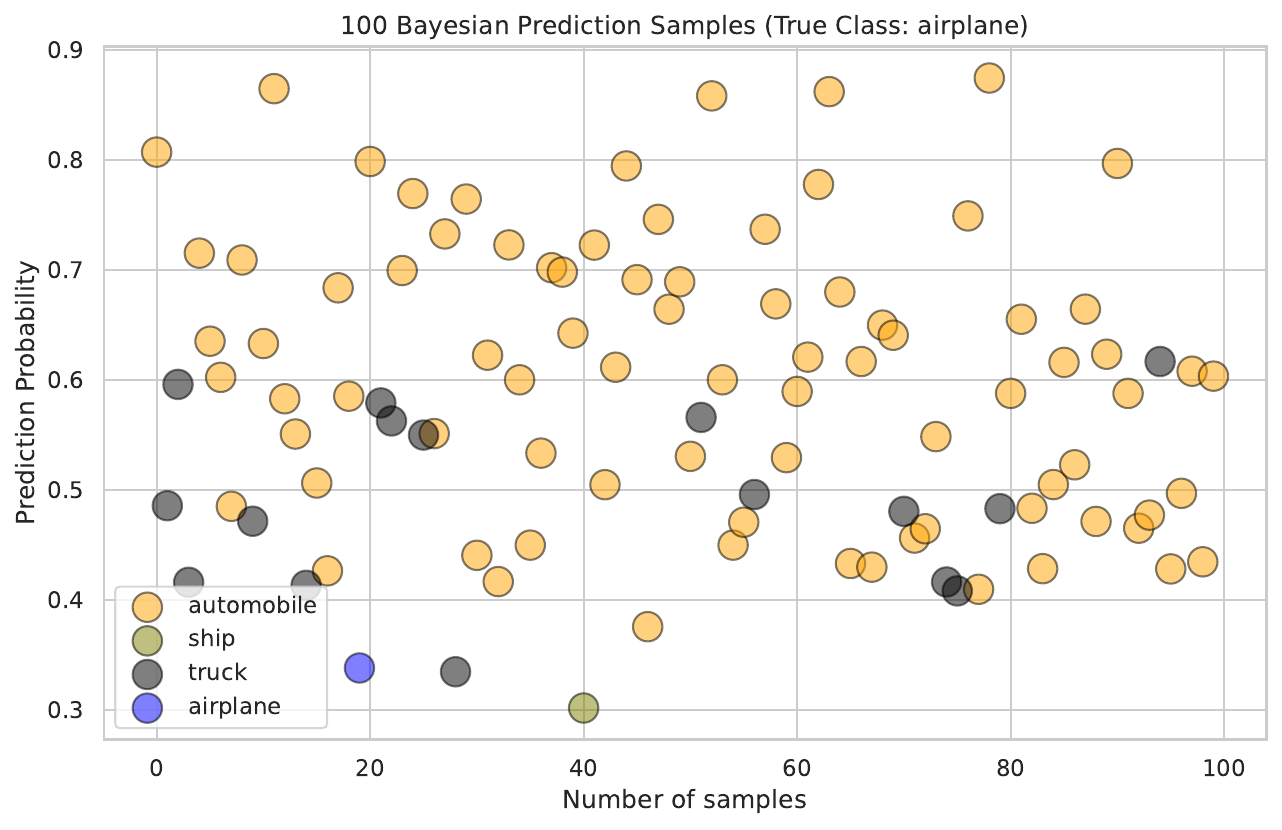}
    \end{minipage}\hspace{0.2cm}
    \begin{minipage}[t]{0.49\textwidth}
        \centering
        \includegraphics[width=\textwidth]{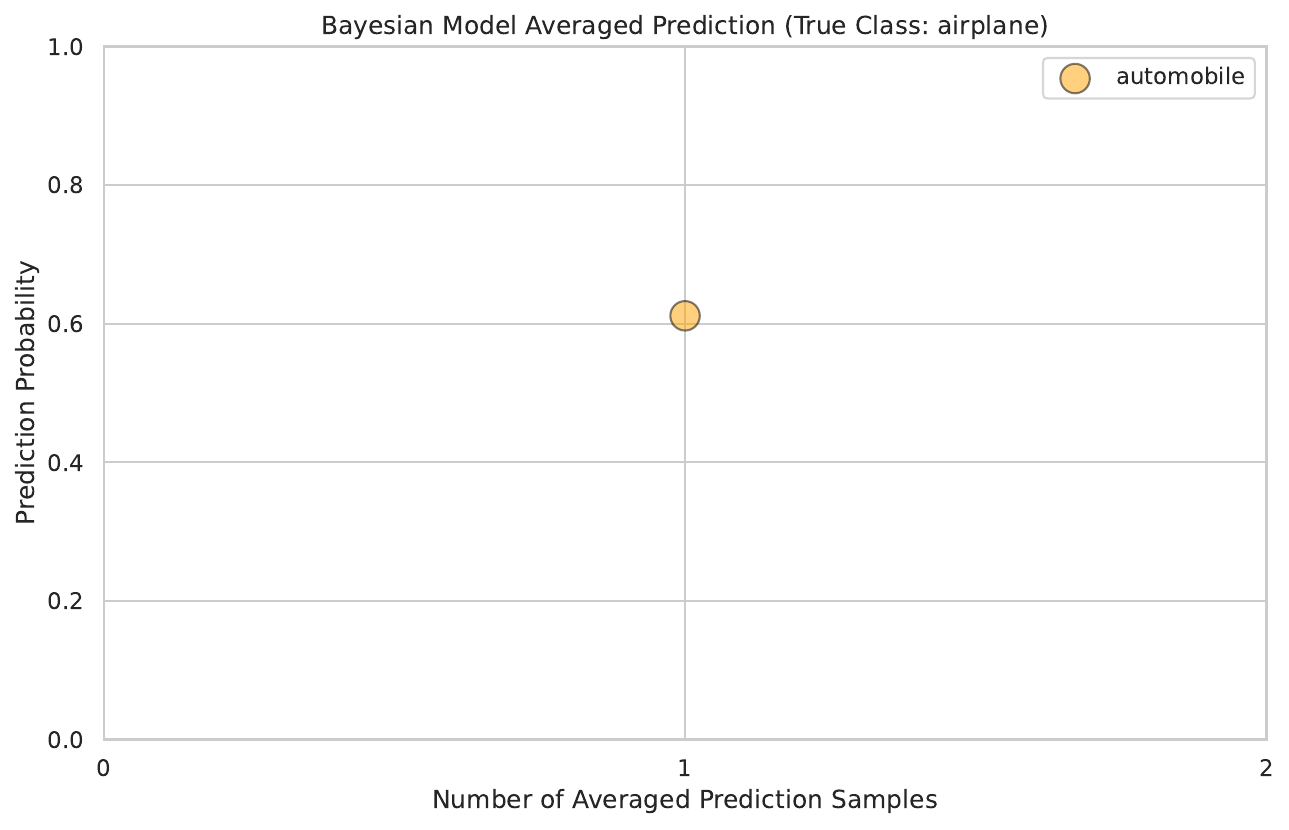}
    \end{minipage}\hfill
    \begin{minipage}[t]{0.49\textwidth}
        \centering
        \includegraphics[width=\textwidth]{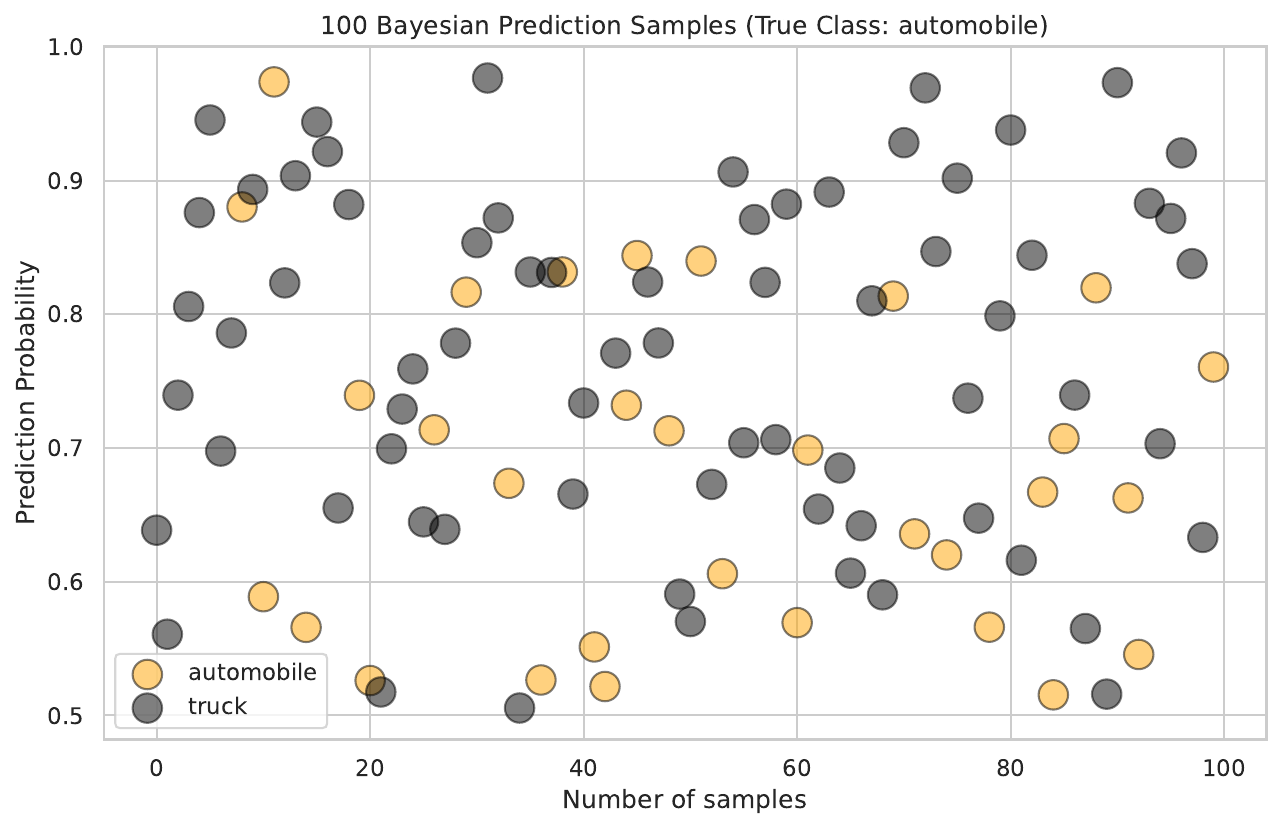}
    \end{minipage}\hspace{0.2cm}
    \begin{minipage}[t]{0.49\textwidth}
        \centering
        \includegraphics[width=\textwidth]{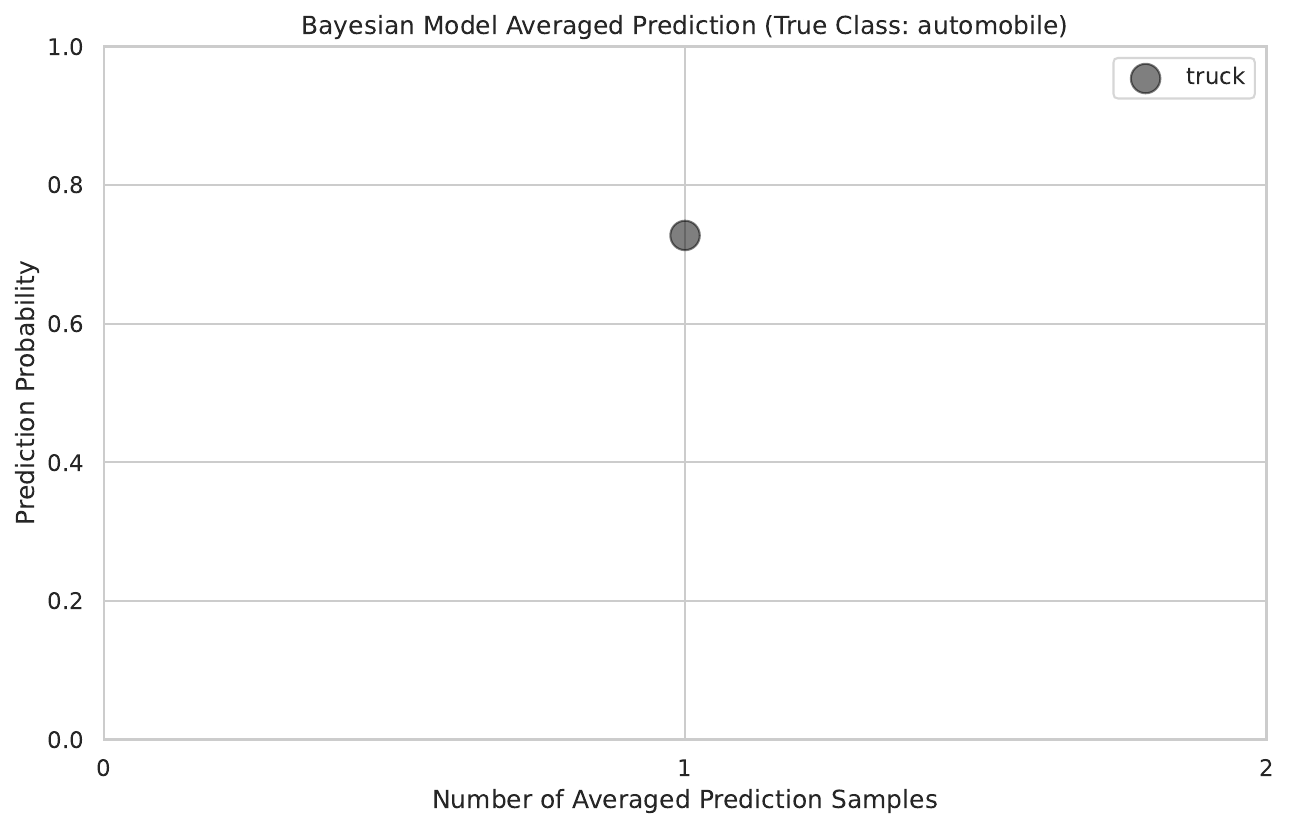}
    \end{minipage}
    \caption{Visualizations of 100 prediction samples obtained prior to Bayesian Model Averaging and corresponding Bayesian Model Averaged prediction in two real scenarios from CIFAR-10.}
    \label{fig:BMA_vs_nonBMA}
\end{figure}

{Fig. \ref{fig:BMA_vs_nonBMA} illustrate key limitations of Bayesian model averaging (BMA).
In Fig. \ref{fig:BMA_vs_nonBMA} (\textit{top}), the true class is `airplane', but most posterior samples assign high probability mass to the `automobile' class. Although one sample assigns a non-negligible probability ($\approx$ 0.34) to `airplane', the majority vote across samples overwhelms this minority, and the BMA prediction incorrectly favors automobile with final averaged probability. This highlights how averaging may suppress informative outlier samples.
In Fig. \ref{fig:BMA_vs_nonBMA} (\textit{bottom}), the true class is `automobile', and the posterior samples are more mixed: `truck' receives the highest number of samples, but automobile is a close second. However, the averaging process again results in the incorrect prediction (`truck') despite there being substantial support for the correct class.}
This can limit a BNN's ability to accurately represent complex uncertainty patterns, potentially undermining its effectiveness in scenarios requiring reliable uncertainty quantification.
BMA may inadvertently smooth out predictive distributions, diluting the inherent uncertainty present in individual models \cite{hinne2020conceptual, graefe2015limitations} as shown in Fig. \ref{fig:BMA_vs_nonBMA}. 
When applied to classification, BMA yields
point-wise predictions. For fair comparison and to overcome BMA's limitations, 
in Chapter \ref{ch:eval_framework} of this thesis, we use sets of prediction samples obtained from the different posterior weights before averaging. {This method of analyzing individual prediction samples before averaging appears to be relatively underexplored.}

In BNNs,
\emph{predictive uncertainty} is measured by the predictive entropy of the model’s output distribution, which captures the total uncertainty in predictions.
Epistemic uncertainty is sometimes represented using the mutual information (MI) between the model predictions and the model parameters \citep{hullermeier2021aleatoric, hullermeier2022quantification}. MI measures the reduction in uncertainty about the model parameters after observing the prediction and can be interpreted as the difference between the entropy of the predictive distribution and the expected entropy of the individual predictive samples.
Despite its theoretical appeal, MI as an epistemic uncertainty measure is not always directly comparable with epistemic uncertainty estimates produced by other methods, such as deep ensembles or other non-Bayesian approaches. This is because different methods define and approximate epistemic uncertainty in different ways, leading to variations in scale and interpretation. Consequently, in this thesis, it is also challenging to directly compare the epistemic uncertainty measures we employ with those based on mutual information.


The predictive entropy can be calculated as: 
\begin{equation}
H(\hat{p}_{b}(y \mid \mathbf{x}, \mathbb{D})) = - \int \hat{p}_{b}(y \mid \mathbf{x}, \mathbb{D}) \log \hat{p}_{b}(y \mid \mathbf{x}, \mathbb{D}) dy ,
\end{equation}
where $\hat{p}_{b}(y \mid \mathbf{x}, \mathbb{D})$ is the predictive distribution and $H(\cdot)$ denotes the Shannon entropy function. This equation represents the average uncertainty associated with the predictions across different possible values of $y$, considering the variability introduced by the parameter uncertainty captured in the posterior distribution $p(\theta \mid \mathbb{D})$.

\section{Ensemble methods}
\label{app:A_DE}

Ensemble methods such as Deep Ensembles (DE) \cite{lakshminarayanan2017simple},  Epistemic Neural Networks (ENN) \cite{osband2024epistemic} and several other studies \cite{pearce2020uncertainty, wen2020batchensemble, beluch2018power} quantify prediction uncertainty by leveraging the prediction of multiple models (\textit{e.g.}, multiple deep networks). Thus, they represent uncertainty by aggregating a limited set of point estimates while averaging a set of single probability distributions as predictions. Deep ensembles serve as one of the top methods in literature for estimating predictive uncertainty \cite{ovadia2019can, gustafsson2020evaluating, abe2022deep}, however, they have been criticized because of their lack of robust theoretical foundations and their demand for substantial memory resources. Computational cost of training ensembles, especially for large models, is often impractical \cite{liu2020simple, ciosek2019conservative, he2020bayesian}.

In Deep Ensembles (DEs) \cite{lakshminarayanan2017simple}, 
a prediction 
$\hat{p}_{de}(y \mid \mathbf{x}, \mathbb{D})$ for an input $\mathbf{x}$ is obtained by averaging the predictions of $K$ individual models:
\begin{equation}
       \hat{p}_{de}(y \mid \mathbf{x}, \mathbb{D}) = \frac{1}{K} \sum_{k=1}^{K} \hat{p}_{de}(y_k \mid \mathbf{x}, \mathbb{D}) ,
\end{equation}
where $\hat{p}_{k}$ represents the prediction of the $k$-th model, trained independently with different initialisations or architectures.

In Deep Ensembles,
{predictive uncertainty} is assessed via the predictive entropy, averaged entropy of each ensemble's prediction, while {epistemic uncertainty} is encoded by the predictive variance, 
the difference between the entropy of all ensembles and the averaged entropy of each ensemble.

Let $\mathcal{M} = \{ M_1, M_2, \ldots, M_K \}$ denote the ensemble of $K$ neural network models for $k = 1, 2, \ldots, K$. Given an input $\mathbf{x}$, the prediction $y_{\mathcal{M}}$ is obtained by averaging the predictions of individual models.
The \emph{predictive entropy} 
represents the entropy of the averaged prediction given the input 
$\mathbf{x}$ and the observed data $\mathbb{D}$:
\begin{equation}
H(\hat{p}_{de}(y \mid \mathbf{x}, \mathbb{D})) = H\biggl(\frac{1}{K} \sum_{k=1}^{K} \hat{p}_{de}(y_{k} \mid \mathbf{x}, \mathbb{D})\biggl),
\end{equation}
where $y_{k}$ represents the prediction of the $k$-th model $M_k$.

The \emph{predictive variance} 
is measured as the difference between the entropy of all the ensembles, $H(\hat{p}_{de}(y \mid \mathbf{x}, \mathbb{D}))$, and the averaged entropy of each ensemble.
\begin{equation}
H(y_{\mathcal{M}}) = H(\hat{p}_{de}(y \mid \mathbf{x}, \mathbb{D})) - \frac{1}{K} \sum_{k=1}^{K} H(\hat{p}_{de}(y_{k} \mid \mathbf{x}, \mathbb{D})).
\end{equation}

The predictive variance in DEs is considered an approximation of mutual information \cite{hullermeier2021aleatoric}. This formulation captures both the model uncertainty inherent in the ensemble predictions and the uncertainty due to the variance among individual model predictions. 

\section{Evidential Methods}
\label{sec:sota-evidential}

Evidential Deep Learning (EDL) models represent predictions as Dirichlet distributions, as explored in various works over the last few years \cite{sensoy, malinin2018predictive, malinin2019reverse, malinin2019ensemble, charpentier2020posterior}. 
A significant amount of work in the evidential framework \cite{155943} has focused on neural networks \cite{ROGOVA1994777}, decision trees \cite{elouedi00decision}, K-nearest neighbour classifiers \cite{Denoeux2008classic}, and more recently, evidential deep learning classifiers able to quantify uncertainty \cite{tong2021evidential}. Notably, \citet{sensoy} introduced a classifier estimating second-order uncertainty in the Dirichlet representation by forming subjective opinions. 

Evidential Deep Learning (EDL) methods and Prior Networks \cite{malinin2018predictive} both leverage the Dirichlet distribution to model uncertainty, but differ in focus \cite{ulmer2023prior}. The Dirichlet, as a conjugate prior to the categorical distribution, allows both methods to maintain Dirichlet-form outputs. Prior Networks \citep{malinin2018predictive} parameterize the Dirichlet prior directly to model uncertainty before observing data, whereas EDL \citep{sensoy} infers the posterior Dirichlet by interpreting network outputs as class evidence under subjective logic \citep{josang2016subjective}. Thus, EDL can be viewed as a type of posterior network that learns uncertainty from data, capturing both aleatoric and epistemic uncertainty through evidence accumulation.

However, many commonly used loss functions for these networks are flawed, as they fail to ensure that epistemic uncertainty diminishes with more data, violating basic asymptotic assumptions \cite{bengs2022pitfalls}. Additionally, some approaches require out-of-distribution (OoD) data for training, which may not always be available and doesn't guarantee robustness against all types of OoD data. Studies \cite{ulmer2023prior} have shown that OoD detection degrades in certain models under adversarial conditions, and even techniques like normalizing flows (NFs) in posterior networks \cite{charpentier2020posterior}, while effective, can struggle with OoD data when relying on learned features \cite{kopetzki2021evaluating, stadler2021graph}.

Evidential models \cite{sensoy} make predictions $\hat{p}_{e}(y \mid \mathbf{x}, \mathbb{D})$ as parameters of a second-order Dirichlet distribution on the class space, instead of softmax probabilities. EDL uses these parameters to obtain a pointwise 
prediction.
Similar to BNNs, averaged DE and EDL predictions are point-wise predictions and averaging may not always be optimal.

\section{Conformal prediction} \label{sec:sota-conformal}

Conformal prediction \cite{vovk2005algorithmic} provides a framework for estimating uncertainty \cite{shafer2008tutorial} by applying a threshold on the error the model can make to produce prediction sets, irrespective of the underlying prediction problem. Different variants of conformal predictors are described in papers by \citet{Saunders1999TransductionWC}, \citet{Nouretdinov01ridgeregression}, \citet{10.1007/3-540-36755-1_32} and \citet{10.5555/1712759.1712773}. It is a wrapper method applicable to any model, generating prediction sets (for accuracy guarantees) by computing empirical cumulative distributions and applying hypothesis testing to them. Several variants exist (\textit{e.g.}, conditional, full conformal prediction) \cite{10.1007/3-540-36755-1_29, 10.5555/1712759.1712773, ICP, Saunders1999TransductionWC, Nouretdinov01ridgeregression, NIPS2003_10c66082, crossconformal, https://doi.org/10.48550/arxiv.1211.0025, campos2024conformal}. 
Since the computational inefficiency of conformal predictors posed a problem for their use in neural networks, Inductive Conformal Predictors (ICPs) were proposed by \citet{ICP}. Venn Predictors \cite{NIPS2003_10c66082}, cross-conformal predictors \cite{crossconformal} and Venn-Abers predictors \cite{https://doi.org/10.48550/arxiv.1211.0025} were introduced in distribution-free uncertainty quantification using conformal learning. 

{Conformal predictors primarily capture aleatoric uncertainty in a frequentist framework by guaranteeing marginal coverage under the assumption of data exchangeability \citep{vovk2005algorithmic}. While coverage is formally guaranteed regardless of model misspecification, the efficiency of the prediction sets (i.e., their width) depends heavily on the quality of the underlying model. Moreover, under distributional shift, conformal methods may fail to maintain valid coverage unless explicitly extended to handle epistemic uncertainty \citep{bates2023testing}.
Standard CP methods rely on calibration via a held-out set and do not model uncertainty about the model itself. \textit{i.e.}, they assume the underlying model is fixed and reasonably accurate.}
In addition to this, since conformal prediction is a wrapper method applicable on top of any underlying model, it is not directly compared with any of the uncertainty quantification methods described in this thesis (including the related methods in this section). Instead, it can be combined with any of these methods, including the model proposed in this thesis. As demonstrated in \S{\ref{app:statistical}}, the model introduced in Chapter \ref{ch:rsnn} is compatible with conformal prediction and can serve as its underlying predictive model.

\vspace{-2pt}

\section{Imprecise models}
\label{sec:imprecise-models}

Imprecise Probability (IP) theory models uncertainty using sets of probability distributions (\textit{e.g.}, credal sets, random sets, \textit{etc.}) rather than single distributions, explicitly capturing epistemic uncertainty \citep{levi1980enterprise, hullermeier2021aleatoric}. \citet{walley1996inferences} introduced the imprecise Dirichlet model, which uses a set of Dirichlet priors to model prior ignorance; as a result, the predictive probabilities become intervals. Credal inference, including models such as the Naive Credal Classifier (NCC) \citep{corani2008learning, Zaffalon, corani2008learning} and credal random forests \citep{shaker2021ensemble}, provides a principled way to quantify epistemic uncertainty by predicting convex sets of probability distributions \citep{corani2012bayesian}. The NCC, extending naive Bayes classifiers by outputting sets of classes rather than single predictions, offers robustness especially when data are scarce or incomplete \citep{corani2008learning, Zaffalon, corani2008learning}. Extensions to multilabel problems and credal networks have further generalized this approach \citep{ANTONUCCI2017320, corani2012bayesian}.

Random sets provide a natural framework for modeling missing data \citep{cuzzolin2020geometry}, while belief function models, rooted in the foundational work of Dempster \citep{Dempster67} and Shafer \citep{shafer1976mathematical}, have been widely applied in machine learning tasks such as ensemble classification \citep{liu2019evidence, 155943, ROGOVA1994777}, regression \citep{gong2017belief, Laanaya2010338, gong2017belief, https://doi.org/10.48550/arxiv.2208.00647}, and generalized max-entropy classification \citep{cuzzolin18belief-maxent}. These models facilitate the representation of uncertainty through transferable belief models and convex sets, combining aleatoric and epistemic components within a unified mathematical framework \citep{cuzzolin10ida}.
Significant contributions have been made by Denoeux and co-authors \citep{Denoeux2008classic, denoeux} on belief function-based unsupervised learning and clustering \citep{Liu2012291}, and on ensemble classification, including methods based on neural networks \citep{ROGOVA1994777}, decision trees \citep{elouedi00decision}, and K-nearest neighbor classifiers \citep{Denoeux2008classic}. 

Research on the use of deterministic intervals in neural networks to represent and quantify uncertainty, known as interval neural networks (INNs), focuses primarily on regression tasks. One line of INN research \citep{Khosrav2011DataDrivenIntervals, pearce2018high, pmlrsalem20a, lai2022exploring} emphasizes generating deterministic interval predictions while keeping the network weights and biases fixed at point estimates. To account for epistemic uncertainty, some researchers, \textit{e.g.}, \citep{pearce2018high, pmlrsalem20a, lai2022exploring} have proposed using ensembles of INNs. For example, the variances of the upper and lower prediction bounds across ensemble INN members can be calculated to quantify the EU associated with each prediction bound \citep{pearce2018high}. Unlike traditional INNs that only represent predictions as intervals, an alternative approach models both network weights and biases as intervals \citep{Ishibuchi1993, Garczarczyk2000, oala2021, betancourt2022interval, tretiak2023neural, cao2024interval}. This design allows the INN to capture both aleatoric and epistemic uncertainty within an interval framework. A further advantage of these models is their capacity to handle interval-valued input data.

A limited number of studies have explored the use of INNs for classification tasks, primarily due to the practical challenges discussed earlier in the introduction. For example, \cite{kowalski2017interval} extended the probabilistic neural network framework by incorporating interval representations to enhance robustness. 

Expected utility maximization methods, such as Bayes-optimal prediction \citep{Mortier} and classification with reject options for risk-averse decision-making \citep{ijcai2018-706}, extend credal and belief frameworks to support decision-theoretic applications. These approaches enable partial or set-valued classification outputs that improve robustness by avoiding overconfident or risky predictions. Practical heuristics often classify all classes whose lower probabilities exceed a threshold, producing `safe' classifications that balance risk and informativeness \citep{corani2008learning, Zaffalon}. This flexibility outperforms rigid single-label decisions or naive reject options, especially in uncertain or low-data regimes.

Classification with partial data has been studied by various authors \cite{10.1007/11518655_80}, 
while a large number of papers have been published on decision trees in the belief function framework \cite{elouedi}. Significant work in the neural network area was conducted in \cite{denoeux} and \cite{10.1007/11829898_18}. 
Classification approaches based on rough sets were proposed in \cite{Trabelsi2011ClassificationSB}. Ensemble classification \cite{burger} is another area in which uncertainty theory has been quite impactful, as the problem is one of combining the results of different classifiers. Important contributions have been made by \cite{155943} and \cite{ROGOVA1994777}. Regression, on the other hand, has only recently been considered by belief theorists \cite{Laanaya2010338,gong2017belief, https://doi.org/10.48550/arxiv.2208.00647}. For tasks like image classification in large datasets, most of these approaches have low performance metrics including training time and test accuracy. 

Recent progress in deep learning has integrated IP frameworks to enhance uncertainty quantification. For instance, Credal Bayesian Deep Learning (Credal-BDL) trains Bayesian neural networks with finitely generated credal sets of priors and likelihoods, yielding sets of posterior distributions that separate aleatoric from epistemic uncertainty and improve robustness against prior misspecification and distribution shifts \citep{caprio2023credal}. Similarly, Credal Wrappers construct convex hulls over ensemble model outputs to provide interval-valued class probabilities, enhancing calibration and out-of-distribution detection on benchmarks such as CIFAR-10 and ImageNet \citep{wang2024credalwrapper}. Methods optimizing interval-valued losses or leveraging conformal prediction with imprecise probabilities produce set-valued predictions with rigorous coverage guarantees, addressing ambiguous or partially labeled data common in medical and human-annotated vision datasets \citep{Javanmardi2024Conformalized}. More recent advancements leverage Dempster–Shafer theory and subjective logic to quantify second-order uncertainty, enabling set-valued predictions that incorporate utility functions from decision theory \citep{tong2021evidential}.

\section{Discussion on existing methods and motivating this thesis}
\label{sec:uncertainty-models-discussion}

As mentioned above, despite their popularity, many uncertainty quantification methods suffer from significant limitations. {Deep Deterministic Uncertainty (DDU)} relies on feature-space distances, which are only as good as the learned embeddings and often degrade in high dimensions, limiting their reliability \citep{postels2021practicality}. Moreover, DDU often lacks proper calibration and fail in out-of-distribution (OoD) detection. {Bayesian Neural Networks (BNNs)} offer principled uncertainty estimates but face \textbf{high computational costs}, complex training procedures, and sensitivity to prior choices \citep{DBLP:journals/corr/abs-2007-06823, https://doi.org/10.48550/arxiv.2105.06868}. Deep Ensembles are empirically strong \citep{lakshminarayanan2017simple}, but \textbf{scale poorly}, with training and inference costs increasing linearly with the number of ensemble members \citep{ovadia2019can, gustafsson2020evaluating}. They also rely on sufficient model diversity, without which their uncertainty estimates can become unreliable \citep{ciosek2019conservative}.

{Evidential methods}, which model second-order uncertainties using Dirichlet distributions, often \textbf{fail to reduce epistemic uncertainty with more data}, violating a core assumption of learning under uncertainty \citep{bengs2022pitfalls}. These models are also \textbf{sensitive to training loss formulations and regularization} choices, and some require access to OoD data during training, which is something not always available in practice \citep{ulmer2023prior}. {Conformal prediction}, while providing \textbf{distribution-free prediction sets with guaranteed coverage}, fundamentally captures \textbf{aleatoric} rather than \textbf{epistemic uncertainty} since it relies on building cumulative
distribution functions of ‘nonconformity scores’ to which it applies classical hypothesis testing \citep{bates2023testing}. It also assumes data exchangeability, and when this assumption is violated, \textbf{coverage guarantees break down}. Together, these limitations highlight the need for approaches that are both \textbf{computationally tractable} and capable of expressing \textbf{epistemic uncertainty in a faithful and comparable manner} across different models.

While Bayesian neural networks do claim to offer a principled second-order treatment by maintaining distributions over parameters, they suffer from high computational costs, prior sensitivity, and scalability issues in real-world applications. Crucially, they also face conceptual shortcomings in modeling ignorance, detailed further in Sec. \ref{sec:why-epi}. Deep ensembles and evidential methods often approximate second-order reasoning, but do so heuristically or unreliably, relying on architectural diversity or uncertain loss formulations, which makes them sensitive and difficult to generalize. Together, these limitations suggest that while many existing methods capture aspects of uncertainty, they struggle {to represent epistemic ignorance robustly}, motivating the need for more expressive alternatives (Sec. \ref{app:uncertainty-measures}).

Some critics, including this thesis, argue that classical probability theory cannot fully address `second-level' uncertainty \citep{hullermeier2021aleatoric}, suggesting the use of more generalized frameworks (Sec. {\ref{app:uncertainty-measures}}), such as possibility theory \citep{Dubois90}, probability intervals \citep{halpern03book}, credal sets \citep{levi1980enterprise} or imprecise probabilities \citep{walley91book} (Sec. \ref{sec:imprecise-models}). 
This thesis presents imprecise models based on random sets, {belief functions,} and credal {sets, motivated} by the fact that these frameworks offer substantial unexplored potential. It demonstrates that significant performance improvements can be achieved without necessarily leveraging the full expressive power of second-order uncertainty measures (see Sec. \ref{sec:rsnn-budgeting}). Instead, scalability can be effectively addressed by designing model structures that are sufficiently expressive to capture key aspects of second-order uncertainty, while remaining computationally tractable.

\textbf{The way epistemic uncertainty is managed and data is leveraged is a defining issue for AI. 
This thesis contributes to the debate by advocating a paradigm shift towards an \emph{Epistemic Artificial Intelligence} in the next part (Chapter \ref{ch:epi_ai}, Part \ref{part:epi}) that explicitly emphasizes learning under ignorance through the use of second-order uncertainty measures (Chapter \ref{ch:knowledge}, Part \ref{part:epi}).}

\part{{{\MakeUppercase{Epistemic Deep Learning}}}}
\label{part:epi}
\thispagestyle{plain}
\chapter{Epistemic Artificial Intelligence} 
\label{ch:epi_ai} 

As discussed in Chapter \ref{sec:uncertainty-models-discussion}, existing models struggle {to capture epistemic uncertainty faithfully}, largely because a {single probability distribution cannot adequately express ignorance} about the data-generating process \citep{cuzzolin2020geometry}. Addressing this limitation requires moving beyond classical probabilistic frameworks and embracing richer, \textbf{\textit{second-order uncertainty measures}} (Sec. \ref{app:uncertainty-measures}), mathematical formalisms designed to explicitly represent and reason under partial knowledge.

This chapter introduces the paradigm of \emph{\textbf{epistemic artificial intelligence}}, in which models are designed to reason not only from what they know but also from what they do not know. By explicitly recognizing and managing uncertainty, epistemic AI aims to enhance the resilience and robustness of AI systems, enabling them to perform more reliably in unpredictable real-world environments. Building on this foundation, this thesis presents a novel approach (Chapter \ref{ch:rsnn}) to effectively model, learn, and act under such uncertainty.

\section{Concept}
\label{sec:epistemic-ai}

\begin{figure}[!h]
\centering
\vspace{5pt}
    \includegraphics[width=0.7\linewidth]{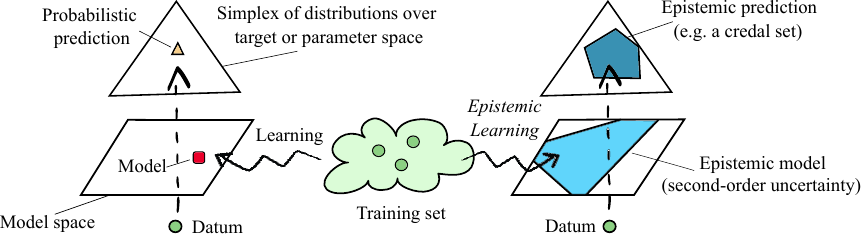}
    \caption[Epistemic Learning.]{\textbf{Epistemic Learning.} Contrary to a traditional learning process in which a single model is learned from a training set to map new data to predictions, in the form of a probability distribution over the target space (left), epistemic learning outputs a second-order uncertainty measure (right).
}
      \label{fig:epistemicai}
\end{figure}

\textbf{Epistemic AI rests on the `paradoxical’ principle that one should first and foremost learn from (or be ready for) the data it cannot see.} 
Prior to observing any data, the task at hand is thus completely unknown (albeit prior knowledge can be utilized to formalize the task and set a model space of solutions). The (limited) available evidence should only be used to temper our ignorance, to avoid `catastrophically forgetting’
how much we ignore about the problem. The problem can be formalized as one of learning a mapping (epistemic model) from input data points to predictions in the form of an uncertainty measure (\textit{e.g.} a credal set), either on the target space or on the parameter space of the model itself (Fig. \ref{fig:epistemicai}). Later, this prediction may be updated in the light of new data. For comparison with other models, a probability-like estimate called the pignistic probability \cite{SMETS2005133} can be computed as the center of mass of the credal set (see Fig. \ref{fig:uncertainty-models}).

\section{Why Epistemic AI is Essential}
\label{sec:why-epi}

In addition to the limitations discussed in Chapter \ref{sec:uncertainty-models}, existing models fundamentally struggle to capture epistemic uncertainty.
Bayesian methods, though widely used, {falter particularly} in data-sparse or ambiguous settings because they must assign fixed belief mass even when knowledge is lacking. Uninformative priors such as Jeffreys’ \citep{jeffreys1998theory} are not invariant under reparameterization and can be improper, violating objectivity and the strong likelihood principle \citep{shafer84tech}. Moreover, priors must be specified even for systems without past data, leading to arbitrary modeling choices that can \textbf{bias the learning process for a long time} (Bernstein-von Mises theorem \citep{kleijn2012bernstein}). Bayesian \textbf{posteriors may appear similar whether we have no knowledge or weak evidence}, conflating ignorance with imprecise belief and potentially causing misleading overconfidence. For example, if we are predicting the probability of a number in a roulette, Bayesian inference treats the uncertainty in the wheel’s bias as part of a single posterior, rather than explicitly representing how much we lack knowledge about the underlying bias.

Model selection and prior choice lack objective criteria and \textbf{prior sensitivity worsens with scarce data} \citep{kass1996selection, berger2013statistical}. 
Further, Bayesian models \textbf{cannot naturally represent set-valued or propositional evidence}, because the additivity of probability forces allocation to individual outcomes, even when evidence supports sets of hypotheses, in {contrast} to random-sets which can naturally model missing data \citep{cuzzolin2020geometry}. Bayes’ rule also \textbf{assumes that new evidence is sharp and definitive}, which is unrealistic in many real-world cases.
Hierarchical Bayesian models, which place priors over priors, can model epistemic uncertainty and potentially address some of these issues, but are very computationally expensive in high-dimensional or open-world settings.

Moreover, Bayesian inference tends to smooth out epistemic uncertainty by averaging over models, collapsing diverse possibilities into a single estimate and failing to distinguish knowns from unknowns \citep{hinne2020conceptual, graefe2015limitations, hullermeier2021aleatoric}. Computationally, Bayesian models also often suffer from slow convergence and large inference times (Tab. \ref{tab:uq-comparison-table}), limiting their suitability for real-time safety-critical systems like autonomous vehicles \citep{DBLP:journals/corr/abs-2007-06823}.
\begin{tcolorbox}
Epistemic AI advocates for the use of second-order uncertainty measures, such as probability intervals, credal sets \citep{levi1980enterprise,cuzzolin2008credal} or random-sets, as they generalize classical probability using set-based representations and can richly encapsulate imprecision to model the epistemic uncertainty about an underlying, shifting data distribution, possibly the central challenge in machine learning. 
\end{tcolorbox}


Indeed, most second-order measures contain classical probability as a special case \citep{cuzzolin2024uncertainty}, with random-set reasoning subsuming Bayesian reasoning as a special case \citep{shafer1981b}. The relationship between Bayesian inference and other mathematical frameworks for uncertainty is discussed in several works \citep{cuzzolin2020geometry,cuzzolin07smcb,cuzzolin11isipta-consonant,cuzzolin09ecsqaru}, also using the language of geometry \citep{cuzzolin2001geometric,cuzzolin2008geometric,cuzzolin2003geometry,cuzzolin04smcb,cuzzolin04ipmu,cuzzolin2010geometry,cuzzolin2013l,cuzzolin2010geometry}.

\section{Modeling A Lack Of Knowledge} 
\label{ch:knowledge} 

The following chapter lays the theoretical foundation by providing an overview of these alternative uncertainty frameworks, including \textbf{random sets} (Sec. \ref{sec:random-sets}), \textbf{belief functions} (Sec. \ref{sec:belief-functions}), and \textbf{credal sets} (Sec. \ref{sec:credal-sets}). These frameworks underpin the models developed throughout this thesis, offering the expressive capacity needed to reason under ignorance. Together, they support the paradigm shift toward \emph{Epistemic Artificial Intelligence} discussed in Chapter \ref{ch:epi_ai}.

\subsection{Theories of uncertainty: Beyond Bayesian Probabilities}
\label{app:uncertainty-measures}

\begin{figure}[!h]
\centering
    \includegraphics[width=0.95\textwidth]{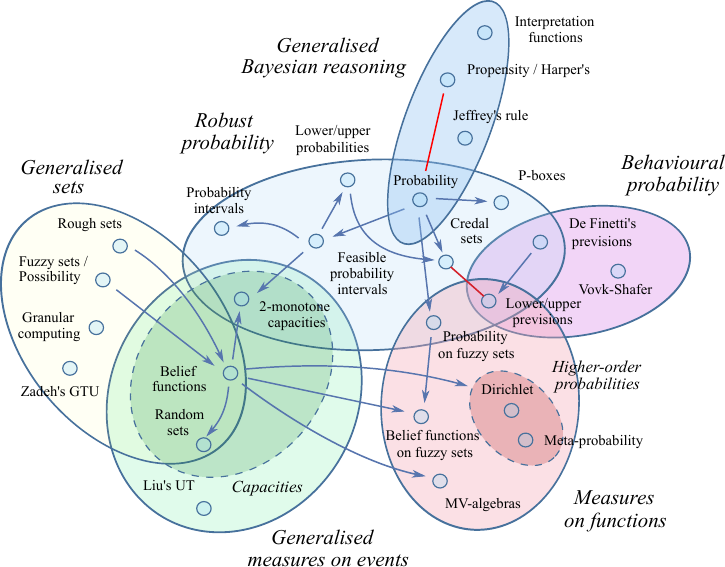}
    \caption[Clusters of uncertainty theories.]{\textbf{Clusters of uncertainty theories.} Uncertainty theories can be arranged into various clusters based on the objects they quantify and their rationale. Arrows indicate the level of generality, with more general theories encompassing less general ones. Note that the quality and rigour of different approaches can vary significantly.
}
      \label{fig:uncertaintymeasures}
\end{figure}

Uncertainty theory (UT){, an} array of theories devised to encode `second-order’, `epistemic’ uncertainty, \textit{i.e.}, uncertainty about what probabilistic process actually generates the data, can provide a principled solution to this conundrum \cite{augustin2014introduction, walley91book}. This is the situation ML is in, for we usually ignore the form of the data-generating process at hand, even accepting that it should be modelled by a probability distribution. Many (but not all) uncertainty measures amount to convex sets of distributions or `credal sets’ (\textit{e.g.}, p-boxes) \cite{troffaes07}, while random sets and belief functions directly assign probability values to sets of outcomes \cite{shafer1976mathematical}, modelling the fact that observations often come in the form of sets. The paramount principle in UT is to {refine continually} one’s degree of uncertainty (measured, \textit{e.g.}, by how wide a convex set of models is) in the light of new evidence. All uncertainty theories are equipped with operators (playing the role of Bayes’ rule in classical probability) allowing one to reason with such measures (\textit{e.g.} Dempster’s combination for belief functions) \cite{dempster2008upper, smets1994transferable}.

The various theories of uncertainty form `clusters’ characterized by a common rationale \cite{cuzzolin2020geometry} (see Fig. \ref{fig:uncertaintymeasures}). A first set of methods can be seen as ways of `robustifying’ classical probability: The most general such approach is Walley’s theory of imprecise probability \cite{walley91book}, a behavioral approach whose roots can be found in the ground-breaking work of de Finetti \cite{DeFinetti74}. In behavioral probability, the latter is a measure of an agent’s propensity to gamble on the uncertain outcomes. A different cluster of approaches hinges on generalizing the very notion of set: these include, for instance, the theory of rough sets \cite{Pawlak1982}, possibility theory \cite{dubois2012possibility}, and Dempster-Shafer theory \cite{shafer1976mathematical}. More general still are frameworks generalizing measure theory, \textit{e.g.} the theory of monotone capacities or `fuzzy measures’ \cite{grabisch2000book}. Some proposals (including Popper’s propensity) aim at generalizing Bayesian reasoning \cite{popper1959propensity} in terms of either the measures used or the inference mechanisms. The theory of set-valued random variable or `random sets’ extends the notion of set \cite{Molchanov_2017} and generalizes Bayesian reasoning. Frameworks which completely replace events by scoring functions \cite{vovk2005algorithmic} in a functional space form arguably the most general class of methods. The diagram also illustrates the relationships between different theories, indicating which are more general and which are more specific. An arrow from formalism 1 to formalism 2 suggests that the former is a less general case of the latter. 

\subsubsection{Random Sets}
\label{sec:random-sets}
\label{sec:background}

\begin{figure}[!h]
    \centering
    \includegraphics[width =0.6\linewidth]{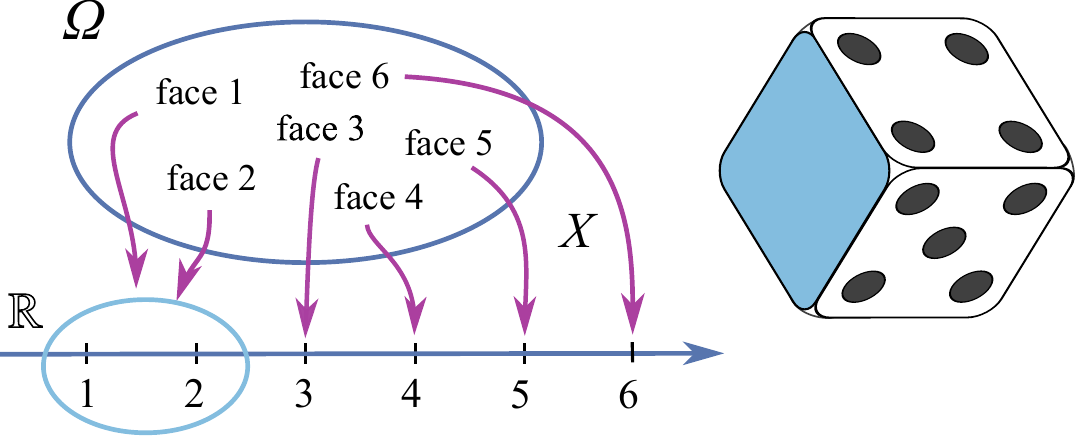} 
    \vspace{-4pt}
    \caption{The random set associated with a cloaked die in which faces 1 and 2 are not visible.} \label{fig:cloaked-die}
\end{figure}

A die is 
a simple example
of a (discrete) random variable. Its probability space is defined on the sample space $\Theta = \{ \text{face1}, \text{face 2}, \ldots , \text{face 6}\}$, where elements are mapped to the real numbers $1,2,\ldots,6$, respectively.
Now, imagine that faces 1 and 2 are cloaked, and we roll the die. How do we model this new experiment, mathematically? Actually, the probability space has not changed (as the physical die has not been altered, its faces still have the same probabilities). What has changed is the mapping: since we cannot observe the outcome when a cloaked face is shown 
(assuming that only the top face is observable),
both face 1 and face 2 (as elements of sample space $\Theta$) are mapped to the set of possible values $\{1,2\}$ on the real line $\mathbb{R}$ (Fig. \ref{fig:cloaked-die}). Mathematically, this is a \emph{\textbf{random set}} \cite{Matheron75,Kendall74foundations,Nguyen78,Molchanov05}, \textit{i.e.}, a set-valued
random variable, modelling random experiments in which observations come in the form of sets. Random sets \cite{Matheron75,Kendall74foundations,Nguyen78,Molchanov05} are \emph{non-additive} measures (\textit{i.e.}, the additivity property does not hold). Their additional degrees of freedom allow to express the `epistemic' uncertainty about probability values themselves. 

\subsubsection{Belief functions}
\label{sec:belief-functions}

Random sets have been proposed by \citet{dempster2008upper} and \citet{shafer1976mathematical} as a mathematical model for subjective belief, alternative to Bayesian reasoning. 
Thus,
on finite domains (\textit{e.g.}, for classification) they assume the name of \emph{\textbf{belief functions}}.
While classical discrete mass functions assign normalised, non-negative values to \emph{elements} $\theta \in \Theta$ of their sample space, a belief function independently assigns normalised, non-negative mass values to its \emph{subsets} $A \subset \Theta$: 
\vspace{-4pt}
\begin{equation}\label{eq:mass-non-neg}
m(A) \geq 0, \:\:\:\:\: 
\sum_A m(A)=1, \:\: \forall A \subset \Theta \vspace{-6pt}
\end{equation} 
The {belief function} 
associated with a mass function $m$ 
measures the total mass of the subsets of each `focal set' $A$. Mass functions can be recovered from belief functions via Moebius inversion \cite{shafer1976mathematical}, 
which, in combinatorics, plays a role similar to that of the derivative:
\vspace{-2pt}
\begin{equation} \label{eq:belief-mobius}
Bel(A) = \sum_{B\subseteq A} m(B), \:\:\:\:\:
m(A) = \sum_{B\subseteq A}{(-1)^{|A\setminus B|} {Bel}(B)}.
\vspace{-6pt}
\end{equation}

\textbf{Pignistic probability.} 
Given a belief function $Bel$, its \emph{pignistic probability} is the precise probability distribution obtained by re-distributing the mass of its focal sets $A$ to its constituent elements, $\theta \in A$:
\vspace{-7pt}
\begin{equation}\label{eq:pignistic}
    BetP(\theta) = \sum_{A \ni \theta} \frac{m(A)}{|A|}.
\end{equation}
\citet{SMETS2005133} originally proposed to use the pignistic probability for decision making using belief functions, by applying expected utility to it. Notably, the pignistic probability is geometrically the centre of mass of the credal set (see Fig. \ref{fig:belief-credal}) associated with a belief function \cite{cuzzolin2018visions}. 
\\
\textbf{Obtaining a credal prediction from belief functions.} 
A belief function is also associated with a convex set of probability distributions, a \emph{credal set} (Sec. \ref{sec:credal-sets}) \cite{levi1980enterprise,zaffalon-treebased,cuzzolin2010credal,antonucci2010credal,cuzzolin2008credal}, on the same domain. 
This is the set:
\begin{equation} \label{eq:consistent}
Cre
=  \left \{ P 
: 
\Theta
\rightarrow [0,1]
| Bel(A) \leq P(A) 
\right \},
\end{equation}
of probability distributions $P$ on $\Theta$ which dominate the belief function on each focal set $A$.
The size of the resulting credal prediction measures the extent of the related epistemic uncertainty 
arising
from lack of evidence (see Sec. {\ref{app:ent-vs-credal}}). 
The use of credal set size as a measure of epistemic uncertainty is well-supported in literature \cite{hullermeier2021aleatoric,bronevich2008axioms}, as it aligns with established concepts of uncertainty such as conflict and non-specificity \cite{yager2008entropy,kolmogorov1965three}.
A credal prediction, by encompassing multiple potential distributions, reflects the model's acknowledgment of this uncertainty. A wider credal set indicates higher uncertainty, as the model refrains from committing to a specific probability distribution due to limited or conflicting evidence. In contrast, a narrower credal set implies lower uncertainty, signifying a more confident prediction based on substantial, consistent evidence.

\subsubsection{Credal Sets}
\label{sec:credal-sets}

In decision theory and probabilistic reasoning, \textit{credal sets} provide a robust approach to modeling epistemic uncertainty by generalizing the traditional Bayesian framework. Unlike standard probabilistic models that assign precise probabilities to events, credal sets represent \textbf{convex sets of probability distributions}, allowing for a more flexible and cautious representation of uncertainty \cite{walley91book}.

A \textit{credal set} is a closed and convex set of probability distributions over a finite space $\Omega$.
This formulation allows one to express \textit{imprecise probabilities}, where instead of a single probability value $P(A)$, we consider a range $[P^-(A), P^+(A)]$ that characterizes the lower and upper bounds of belief for an event $A$. This approach is particularly useful in settings where data is scarce or conflicting, making precise probability assignments unreliable~\cite{augustin2014introduction}.
Credal sets have been extensively used in \textit{robust Bayesian inference}, \textit{classification}, and \textit{decision-making under ambiguity}. In machine learning, credal classifiers~\cite{Zaffalon} extend Bayesian classifiers by considering sets of posterior probabilities rather than single estimates, improving robustness to small-sample uncertainties.

\subsubsection{A unified example}

To concretely illustrate how mass functions, belief functions, and credal sets relate to one another, we present a small, unified example. This example serves to clarify the abstract concepts discussed earlier by grounding them in a simple, interpretable setting with three classes. It highlights how belief values can be computed from mass functions and how these, in turn, define a credal set.

{Consider a 
sample space (class list)
$\Theta = \mathcal{C} = \{\text{$c_1$}, \text{$c_2$}, \text{$c_3$}\}$ and let 
$\mathbb{P}(\Theta)$
be its power set (collection of all subsets). As shown in Fig. \ref{fig:belief-credal}, one can define a mass function 
on 
$\mathbb{P}(\Theta)$
as: $m(\{\text{$c_1$}\}) = 0.5$, 
$m(\{\text{$c_3$}\}) = 0.1$, $m(\{\text{$c_1$}, \text{$c_2$}\}) = 0.4$ (Fig. \ref{fig:belief-credal}, top), and
{the masses for all other sets (unspecified) equal to zero.
Note that 
$m$
is normalised: 
$\sum_{B \subseteq \Theta} m(B) = 1$.
By Eq. \ref{eq:belief-mobius}, the belief value of the composite class $A = \{\text{$c_2$}, \text{$c_3$}\}$ is:
$\text{Bel}(\{\text{$c_2$}, \text{$c_3$}\}) = m(\{\text{$c_3$}\}) = 0.1$ (Fig. \ref{fig:belief-credal}, left). Similarly, the belief value of composite class $\{\text{$c_1$}, \text{$c_2$}\}$ can by computed as $\text{Bel}(\{\text{$c_1$}, \text{$c_2$}\}) = m(\{\text{$c_1$}\}) + m(\{\text{$c_1$}, \text{$c_2$}\}) = 0.5 + 0.4 = 0.9$.} 



\begin{figure}[!h]
    \centering
    \includegraphics[width = 0.92\linewidth]{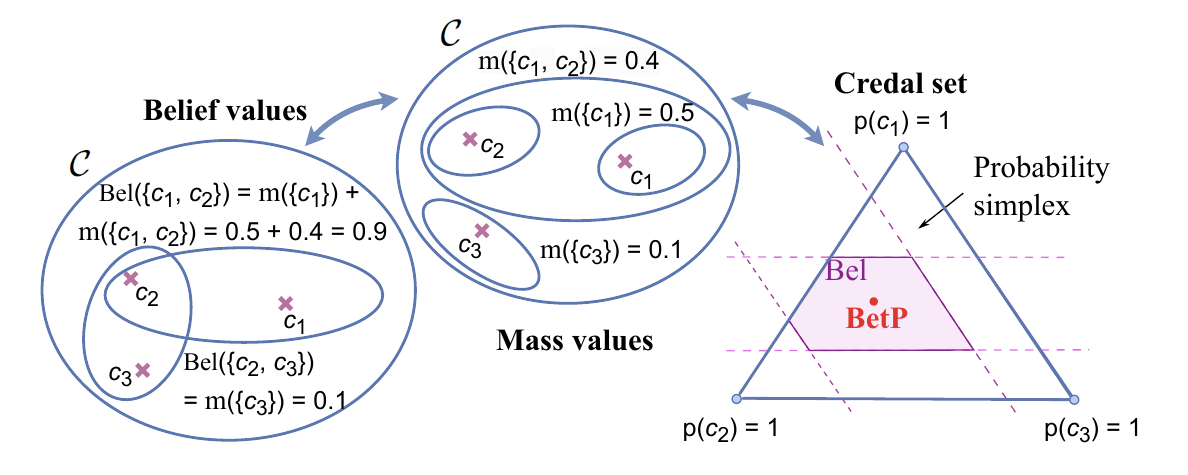}
    \vspace{-8pt}
    \caption{A belief function is equivalent to a credal set with boundaries determined by lower bounds (Eq. \ref{eq:consistent}) on probability values. 
    \label{fig:belief-credal}
    }
\end{figure}

\section{Classical Probabilities {vs.} Random Sets \& Belief Functions}\label{app:classicalvsbelief}

In this section, we explain the difference between classical probability measures and non-additive measures, such as random sets. 

In a standard classification task, where each input is classified into one of several mutually exclusive classes, the classical probability framework does assume that each class is distinct and independent. The output of a softmax model provides the probabilities for each class, and these probabilities sum to 1. 
This approach works well when you are 
confident
about the classification, 
for
you have sufficient evidence to assign a clear probability to each class. 
However, it does not capture the uncertainty or ambiguity that exists when the evidence (as provided by the training set) is insufficient (\textit{e.g.}, because you did not sample similar points at training time, or because the test data is affected by distribution shift). 

Relying solely on confidence measures (like softmax probabilities) is insufficient for a comprehensive analysis of model predictions. Fig. \ref{fig:imagenet-a} shows that standard neural networks are often overconfident \cite{Nguyen_2015_CVPR} even for misclassifications. 

Random sets, as the name suggests, assign probability mass values to sets \emph{independently}, and are therefore more general than classical probability measures (which are a special case of random set). As a result, 
belief functions offer a way to handle uncertainty and make decisions when information is incomplete or ambiguous. While in classical probability, the sum of probabilities for individual events and their unions adheres to strict additive rules, belief functions relax these constraints to better capture uncertainty. 
In particular, the belief in the union of two hypotheses, 
Bel(\{A,B\}), can be greater than the sum of the beliefs in the individual hypotheses, 
Bel(\{A\}) + Bel(\{B\}), i.e, Bel(\{A\}) + Bel(\{B\}) $\leq$ Bel(\{A,B\}).  This inequality captures the idea that, 
in some situations, the available evidence may support the true class being in the set $\{A, B\}$, without being enough to distinguish between the two alternatives.
If Bel(\{A,B\}) $>$ Bel(\{A\}) + Bel(\{B\}), it indicates that the model has significant uncertainty or confusion about distinguishing between these classes for a particular input.

Unlike classical probability theory, belief functions operate in a higher space than probability vectors, i.e, it operates in a framework that generalizes classical probability theory.
Suppose we know with 100\% certainty that an event is either A or B. In a classical probability setting, we might say P(A) = 0.5 and P(B) = 0.5, or we might assign different probabilities to A and B if we have some reason to favor one over the other. With belief functions, we can represent the same scenario differently. We could assign a belief value of 1 to the set \{A, B\} and 0 to both A and B individually (Bel(\{A\}) = Bel(\{B\}) = 0, Bel(\{A, B\}) = 1). This means that while we know the outcome must be either A or B, we are entirely uncertain about which one it is. Alternatively, we could have Bel(\{A\}) = 0.5, Bel(\{B\}) = 0.5, Bel(\{A, B\}) = 1, reflecting a different state of knowledge or evidence, where we're somewhat more informed but still not fully certain. In the behavioural interpretation of probability \cite{walley2000towards}, the belief value of an event A is the upper bound to the price one is willing to pay for betting on an outcome A.

Consider the following example \cite{cuzzolin2020geometry}: Suppose there is a murder, and three people—Peter, John, and Mary—are suspects. Our hypothesis space is therefore $\Theta$ = \{Peter, John, Mary\}. A witness testifies that the person he saw was a man, supporting the proposition A = \{Peter, John\} $\subseteq \Theta$. However, the witness was tested, and the machine reported a 20\% chance that he was drunk when he gave his testimony. As a result, we assign 80\% belief to the proposition A, representing our uncertainty about whether any of the two could be the murderer.

In classical probability, Kolmogorov’s additive probability theory \cite{kolmogorov} forces us to specify support for \textit{individual outcomes}, \textit{i.e.} to distribute this 80\% probability between Peter and John individually, even when the evidence (our data) supports \textit{set propositions}. This is unreasonable – an artificial constraint due to a mathematical model that is not general enough. In the example, we have do not have enough evidence to assign this 80\% probability to either Peter or John, nor information on how to distribute it amongst them. The cause is the additivity constraint that probability measures are subject to. This constraint reflects a broader limitation of measure-theoretical probability, which, while effective in many scenarios, struggles with second-order uncertainties and incomplete information.

This means that \textit{belief functions allow us to model situations where we have uncertainty not just about which class an input belongs to, but also about whether we have enough evidence to distinguish between certain classes}. This added flexibility is what makes belief functions suitable for single-label classification tasks.

\section{Leveraging Epistemic AI}

This thesis proposes an epistemic deep learning model, called Random-Set Neural Network (RS-NN), as a step toward realizing the vision of Epistemic AI, detailed in Chapter \ref{ch:rsnn}. In addition, {the author} contributed to the development of two credal set models through collaborative research, which are briefly outlined in Chapter \ref{ch:cre_models}; however, a full exposition of these co-authored models is beyond the scope of this thesis. The conclusion of this thesis aso provides a comprehensive comparison of Epistemic AI models against baselines such as Bayesian and Ensemble models. The comparison (Sec. \ref{sec:impact-epi}) spans key performance metrics, including test accuracy, out-of-distribution (OoD) detection, expected calibration error (ECE), and training and inference times, demonstrating the advantages of the proposed approach in uncertainty-aware learning.

\chapter{Random-Set Neural Networks} 
\label{ch:rsnn} \label{sec:belief-function-models}\label{app:A_BF}


The \textit{main contribution} of this thesis, presented in this chapter, is a novel approach to classification that models epistemic uncertainty using a \emph{random set} framework \cite{Molchanov05,Molchanov_2017}, introduced here as \textbf{Random-Set Neural Networks (RS-NN)}.
This method serves as a concrete realization of the Epistemic AI perspective (Chapter \ref{ch:epi_ai}) by explicitly modeling uncertainty through second-order uncertainty measures (Sec. \ref{app:uncertainty-measures}), specifically belief functions (Sec. \ref{sec:belief-functions}), random sets (Sec. \ref{sec:random-sets}), and credal sets (Sec. \ref{sec:credal-sets}).

As they assign probability values to sets of outcomes directly, random sets can naturally model the fact that observations 
often
come in the form of sets (in particular, when missing data occurs), and  accommodate
ambiguity, incomplete data, and non-probabilistic uncertainties.
As 
classification 
involves only
a finite list of classes,
we model uncertainty using \emph{belief functions} \cite{shafer1976mathematical}, the finite incarnation of random sets \cite{cuzzolin2018visions},
whose theory \cite{shafer1976mathematical}
is, in fact, a generalisation of Bayesian inference \cite{smets86bayes}. Classical (discrete) probabilities can be seen as special belief functions, and Bayes' rule as a special case of the Dempster's rule of combination \cite{dempster2008upper}
originally proposed for 
aggregating
belief functions. 
{For readers less familiar with the topic,
in Sec. {\ref{app:classicalvsbelief}} we recall the distinction between 
classical probabilities and belief functions and the way they handle uncertainty in more detail.

Fig. \ref{fig:rscnn_bayesian} contrasts the inference processes of our \emph{Random-Set Neural Network} (RS-NN) and of a Bayesian Neural Network \cite{DBLP:journals/corr/abs-2007-06823}. 
In hierarchical Bayesian inference (top), 
a posterior distribution over the network's weights is learned from a training set. At prediction time, a predictive distribution is generated in the target space (left) by sampling from this weight posterior (middle), which amounts to a second-order probability distribution there \cite{hullermeier2021aleatoric}.
A single (mean) prediction is then typically derived by Bayesian Model Averaging (BMA) \cite{hoeting1999bayesian} (right), while 
uncertainty is measured by the entropy of the mean prediction and the variance of the predictive 
distribution. 

\begin{figure}[!ht]
    \centering
        \hspace{0.6cm}
    \includegraphics[width=0.8\textwidth]{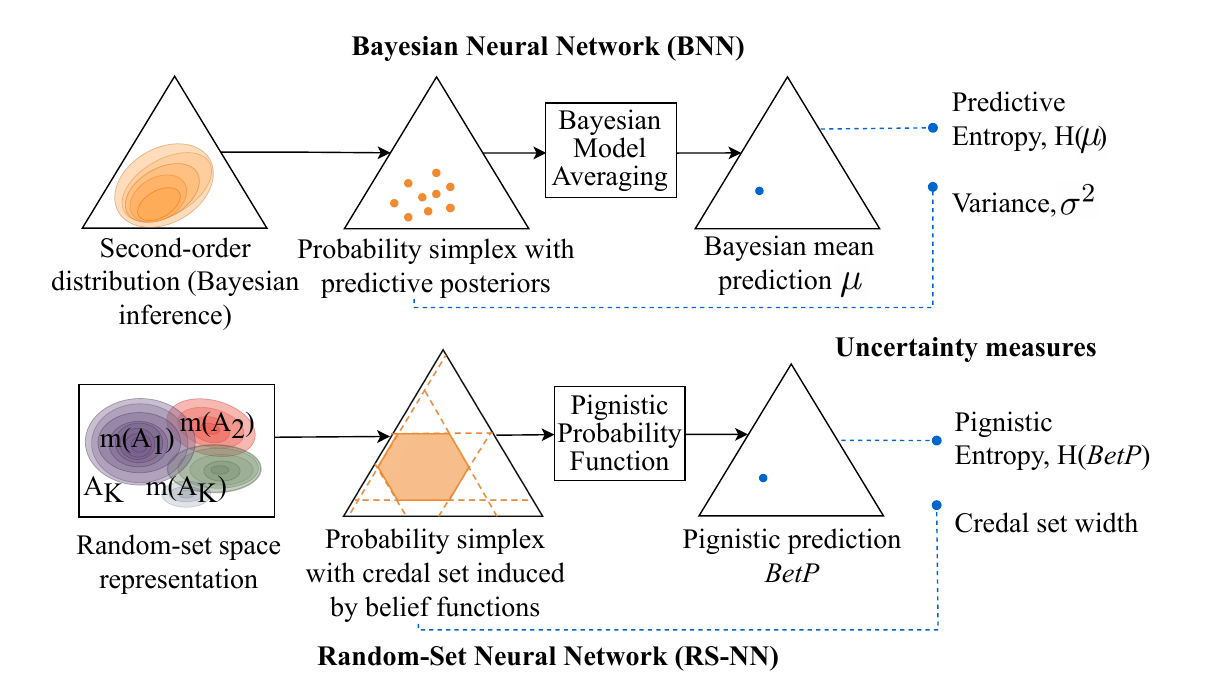}
        \caption[Inference in a Bayesian Neural Network (top) as opposed to a Random-Set Neural Network (bottom), with corresponding measures of uncertainty and their sources.]{Inference in a Bayesian Neural Network (top) as opposed to a Random-Set Neural Network (bottom), with corresponding measures of uncertainty and their sources.
        The triangle represents the set of probability vectors ({probability simplex}) one can define on the target space (\textit{e.g.}, a set of 3 classes).
         }
        \label{fig:rscnn_bayesian}
\end{figure} 
    

In a Random-Set Neural Network
(Fig. \ref{fig:rscnn_bayesian}, bottom), each test input at inference time is mapped to a belief function \cite{shafer1976mathematical} on the collection of classes $\mathcal{C}$. 
This belief function is encoded by a mass value $m(A) \in [0,1]$ 
assigned to a finite budget $\mathcal{O} = \{A_k \subset \mathcal{C}\}$ of 
\emph{sets} of classes 
(left).
The
most relevant 
such sets 
are identified from training data by fitting a Gaussian Mixture Model (GMM) to 
the labelled data
and computing the overlap among the resulting clusters \cite{spruyt2014draw}. 
Such a predicted belief function is mathematically equivalent to a convex set of probability vectors (\emph{credal set} \cite{levi1980enterprise,cuzzolin2008credal}) on the 
class list $\mathcal{C}$ (middle).
A pointwise prediction (right) can then be obtained by computing the centre of mass of this credal set, termed \emph{pignistic probability} \cite{smets1994transferable}. In 
RS-NN,
uncertainty can be expressed using either the entropy of the pignistic prediction 
(analogously to the Bayesian case),
or the width of the credal set prediction \cite{antonucci2010credal}.
We find empirically that the pignistic entropy better separates in-distribution (iD) and out-of-distribution (OoD) samples than Bayesian entropy does (Tab. \ref{tab:ood-table});
the width of the credal prediction, as it encodes the epistemic uncertainty about the prediction itself, is empirically much less correlated with the confidence score than entropy (Sec. {\ref{app:ent-vs-credal}}). 


\begin{figure}[!b]
        \centering
        \hspace{0.6cm}
        \includegraphics[width=0.7\textwidth]{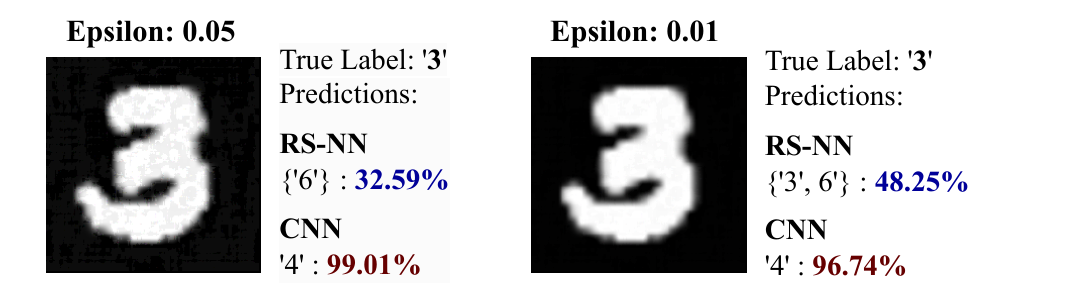}
        \caption{Confidence scores of RS-NN and CNN for FGSM adversarial attack for different perturbations ($\epsilon$ = 0.05, 0.01) on MNIST dataset.}
        \label{fig:confidence_scores}
\end{figure}

Using a random-set representation prevents the need to discard information, a challenge observed in Bayesian Model Averaging \cite{hinne2020conceptual, graefe2015limitations}.
While 
Bayesian inference requires
defining prior distributions for model parameters even in the absence of relevant information,
in belief theory
priors are not required for the inference process, thus avoiding the selection bias risks that can seriously condition Bayesian reasoning \cite{freedman1999wald}. 
Based on our extensive experiments on multiple datasets, including large-scale ones, RS-NNs not only demonstrate superior accuracy compared to state-of-the-art Bayesian and Ensemble models (Sec. \ref{sec:acc}), 
but also arguably better encode the epistemic uncertainty associated with the predictions (Sec. \ref{sec:uncertainty}) and 
better distinguish in-distribution and out-of-distribution data (Sec. \ref{sec:ood-detection}). 
Furthermore, 
RS-NNs effectively circumvent the tendency of standard networks to generate overconfident incorrect predictions, as illustrated in Fig. \ref{fig:confidence_scores} in an experiment on Fast Gradient Sign Method (FGSM) \cite{goodfellow2014explaining} adversarial attacks on MNIST \cite{LeCun2005TheMD}.
CNN misclassifies with high confidence scores of 99.01\% and 96.74\%, while RS-NN shows lower confidence at 32.59\% and 48.25\%. 
{This calibrated uncertainty helps avoid overconfident errors and supports risk-aware decisions, rather than leading to decision paralysis. In high-stakes, real-time decision contexts such as avoidance maneuvers in autonomous driving, overconfident yet incorrect predictions, as seen in the CNN, could lead to hazardous outcomes. In contrast, RS-NN’s lower, more calibrated confidence enables the system to recognize uncertainty and either defer action, engage contingency protocols, or request external intervention, thereby promoting safer and more cautious responses under uncertainty.}

\section{Key Contributions} 

\begin{itemize}[label=\textbullet, leftmargin=*]
  \setlength\itemindent{10pt}
  \setlength\parindent{0pt}  

\item \textbf{Firstly,} 
    a novel \emph{Random-Set Neural Network (RS-NN)} approach is proposed based on the principle that a deep neural network 
    predicting belief values for \emph{sets} of classes, rather than individual classes, has the potential to be
    a more faithful representation of the epistemic uncertainty induced by
    the limited quantity and quality of the training data. 
    RS-NN acts as a `wrapper' technique that can be applied on top of any existing baseline network model, by changing only the output layers and loss function.
    Statistical guarantees for RS-NN predictions can be provided by applying conformal prediction on the pignistic probabilities (\S{\ref{app:statistical}}).
\item \textbf{Secondly,} 
    a \emph{budgeting} method is outlined for efficiently selecting a limited budget of 
    relevant sets of classes for the task at hand given the available training data, 
    by fitting Gaussian Mixture Models to the labelled training points and computing their clusters.
    This overcomes the 
    exponential complexity of 
    vanilla random-set implementations,
    ensures the scalability of the approach to large datasets, and helps the network learn by limiting the available degrees of freedom
    {(Sec. \ref{app:ablation_k}).}
\item \textbf{Thirdly,} 
    two new methods are introduced for assessing the uncertainty associated with a random-set prediction: the Shannon entropy of the pignistic prediction 
    and the width of the credal prediction itself, 
    which prove to be more robust uncertainty measures. For instance, pignistic entropy provides a clearer distinction between in-distribution (iD) and out-of-distribution (OoD) entropy (Fig. \ref{fig:entropy_comparison}, Sec. {\ref{app:entropy}}), while credal set width excels in separating iD and OoD samples, particularly for challenging datasets like ImageNet vs. ImageNet-O (Tab. \ref{tab:vit-table}).
\item \textbf{Finally,} 
    a large body of experimental results is presented
    (based on a fair comparison principle in which all competing models are trained from scratch) demonstrating
    how RS-NN outperforms both state-of-the-art Bayesian (LB-BNN \cite{hobbhahn2022fast}, FSVI \cite{rudner2022tractable}) and Ensemble (DE \cite{lakshminarayanan2017simple}, ENN \cite{osband2024epistemic}) methods in terms of: \\
    \hangindent=1.5em (i) performance (test accuracy, inference time) (Sec. \ref{sec:acc}); \\
    \hangindent=1.5em (ii) 
    results 
    on various out-of-distribution (OoD) benchmarks (Sec. \ref{sec:ood-detection}), including CIFAR-10 vs. SVHN/Intel-Image, MNIST vs. FMNIST/KMNIST, and ImageNet vs. ImageNet-O; \\
    \hangindent=1.5em (iii) 
    ability to provide reliable measures of uncertainty quantification (Sec. \ref{sec:uncertainty}) in the form of pignistic entropy and credal set width, verified on OoD benchmarks; \\
    \hangindent=1.5em (iv) scalability to large-scale architectures (WideResNet-28-10, Inception V3, EfficientNet B2, ViT-Base-16) and datasets (\textit{e.g.} ImageNet) (Sec. \ref{sec:scalability}). 
\item Additionally, 
    RS-NN is shown to be robust to adversarial attacks (Sec. {\ref{app:fgsm}}) and noisy data (Sec. {\ref{app:noisy}}),
    and to circumvent the overconfidence problem in CNNs 
    (Sec. {\ref{app:cnn-overconfidence}}). 
    A qualitative assessment of entropy \textit{vs.} credal set width is given in Sec. {\ref{app:ent-vs-credal}}
    and in-distribution \textit{vs.} out-of-distribution entropy scores are shown in Fig. \ref{fig:entropy_comparison}. 
\end{itemize}

\section{Methodology}\label{sec:rscnn}

This section intoduces the proposed Random-Set Neural Network (RS-NN) approach, ground truth representation, budgeting algorithm and loss function, and explains how RS-NN utilizes the concepts introduced in the previous Chapter \ref{ch:knowledge} to efficiently model classification uncertainty.

\subsection{Ground-truth Representation}\label{sec:arch}

\begin{figure}[!h]
 \centering
              \hspace{1.2cm}
       \includegraphics[width=0.9\linewidth]{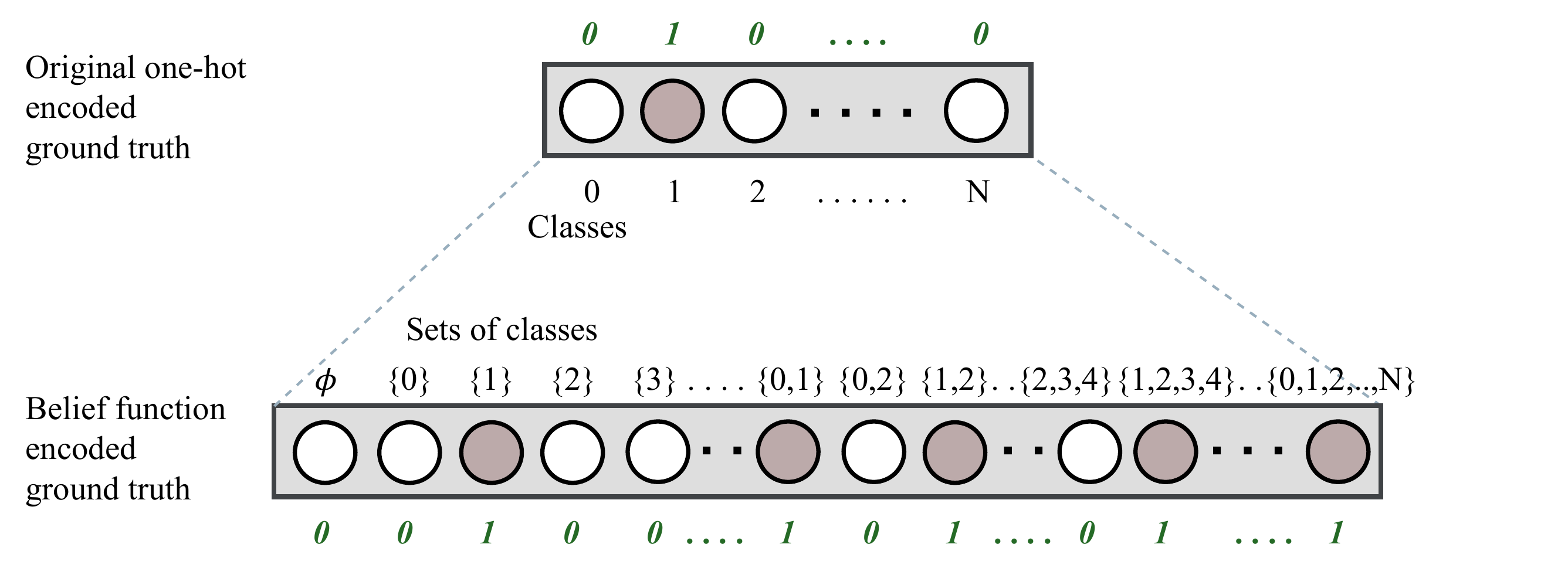}
       \caption{Standard one-hot encoded ground truth in traditional neural networks vs. belief function-based encoding in Random-Set Neural Networks (RS-NN).}
       \label{fig:groundtruth-rep}
   \end{figure}
   
A classifier $e$ (\textit{e.g.}, a neural network) is a mapping from an input space $X$ to a 
target space 
$\mathcal{C}$ (the list of classes), 
\textit{i.e.}, 
$e: X \rightarrow \mathcal{C}$.
In our set-valued (Sec. \ref{sec:random-sets}) classification setting, on the other hand, $e$ is a mapping from $X$ to the set of all \emph{subsets} of $\mathcal{C}$, the powerset $\mathbb{P}(\mathcal{C})$, namely: $e: X \rightarrow 
\mathbb{P}(\mathcal{C})$.

As shown in Fig. \ref{fig:rsnnapproach} (b), RS-NN predicts for each input data point a 
belief function,
rather than a vector of softmax probabilities as in a traditional CNN. 
For $N$ classes, a `vanilla' RS-NN would have $2^N$ outputs (as $2^N$ is the cardinality of $\mathbb{P}(\mathcal{C})$), each being the belief value (Eq. \ref{eq:belief-mobius}) of the focal set of classes $A \in\mathbb{P}(\mathcal{C})$ corresponding to that output neuron.
{Our architecture focuses primarily on the final layers 
(Fig. \ref{fig:rsnnapproach} (b), in grey), acting as a wrapper 
applicable on top of
\textit{any} baseline representation layers from existing models 
(Fig. \ref{fig:rsnnapproach} (b), in blue). This enables the integration of pre-trained networks while fine-tuning only the final decision layers. Hence, RS-NN is easily scalable to any model architecture, as demonstrated in Sec. \ref{sec:scalability}, Tab. \ref{tab:scalability}.

Given a training datapoint with a true class attached, its
ground truth is encoded by the vector 
$\mathbf{bel} = \{ Bel(A), A \in \mathbb{P}(\mathcal{C}) \}$
of belief values for each focal set of classes $A\in\mathbb{P}(\mathcal{C})$.
$Bel(A)$ is set to 1 iff the true class is contained in the subset $A$, 0 otherwise\footnote{Note that the belief encoding of ground-truth is not related to label smoothing. It maps ground-truth labels from the original class space to a set space without adding noise, thus preserving label `hardness'.} as shown in Fig. \ref{fig:groundtruth-rep}.
This corresponds to full certainty that the element belongs to that set and there is complete confidence in this proposition 
(see Sec. {\ref{app:rsnn-learning}} for an example). 

{In this thesis, the empty set is not included in the representation of the focal set. This is because there is no ground-truth label corresponding to the empty set, \textit{i.e.}, the training process assumes that each input belongs to at least one of the known classes. As a result, no belief is ever explicitly allocated to the empty set during training.
Including the empty set would imply allocating belief to the hypothesis that none of the known classes apply, which cannot be learned or supervised in the current closed-world setting. Its exclusion is therefore a modeling decision aligned with the structure of the training data and the learning objective.
}

\subsection{Budgeting}\label{sec:rsnn-budgeting}

\begin{figure}[!h]
    \centering
    \begin{minipage}{0.34\textwidth}
    \hspace{-0.75cm}   
        \includegraphics[width=1.25\linewidth]{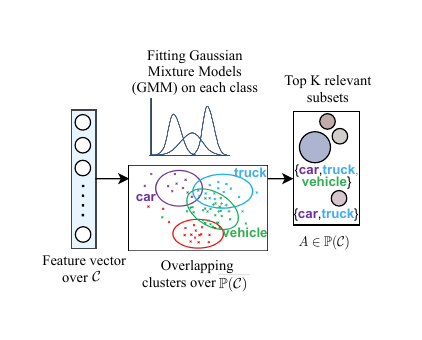}
        \caption*{\centering(a)}
    \end{minipage}
    \hspace{0.001pt}
    \begin{minipage}{0.64\textwidth}
            \hspace{-0.3cm}   
        \includegraphics[width=1.12\linewidth]{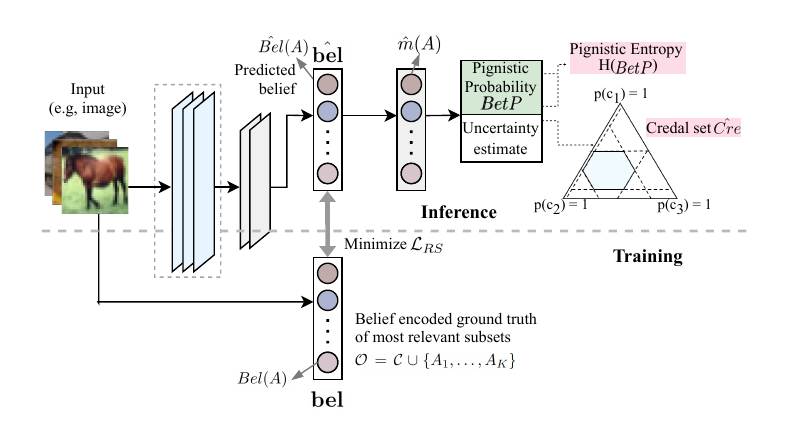}
        \vspace{-30pt}
        \caption*{\centering(b)}
    \end{minipage}
    \caption[RS-NN model architecture: (a) {Budgeting}, (b) {Training and Inference}.]{\textbf{RS-NN model architecture.} (a) \textit{Budgeting}: Given a collection $\mathcal{C}$ of $N$ classes, the top $K$ relevant (focal) sets of classes $\{A_1, \ldots, A_K\}$ are selected from the powerset $\mathbb{P}(\mathcal{C})$ (Algorithm in \S{\ref{app:algo-budget}})
    and added to the singleton classes to form the budget $\mathcal{O}$. (b) \textit{Training and Inference}: Ground truth classes are encoded as belief vectors $\mathbf{bel}$ and used to predict a belief function $\mathbf{\hat{bel}}$ for each 
    training
    data point by minimising the loss $\mathcal{L}_{RS}$ (\ref{final_loss}), to train the output layers (in grey) producing $\mathbf{\hat{bel}}$. Mass values $\hat{m}$ and pignistic probability estimates $BetP$ are computed from the predicted belief function. Uncertainty is estimated as described in Sec. \ref{uncertainty-estimation}.} 
    \label{fig:rsnnapproach}
    \vspace{-1mm}
\end{figure}

{To overcome the exponential complexity of using $2^N$ sets of classes (especially for large $N$), a fixed budget of $K$ relevant non-singleton (of cardinality $>1$) focal sets are used.
These focal sets are obtained by clustering the original classes $\mathcal{C}$, fitting ellipses over them, and selecting the top $K$ focal sets of classes with the highest overlap ratio (Fig. \ref{fig:rsnnapproach} (a)). This is computed as the intersection over union for each subset $A$ in $\mathbb{P}(\mathcal{C})$: $ overlap(A) = {\cap_{c \in A}A^c}/{\cup_{c \in A}A^c}, \: A_1, \ldots, A_K \in \mathbb{P}(\mathcal{C})$. 
Clustering is performed on feature vectors of images
generated by any feature extractor trained on the original classes $\mathcal{C}$. In our experiments, we used features from a trained standard ResNet50 model.
These feature vectors are further reduced to 3 dimensions using t-SNE (t-Distributed Stochastic Neighbor Embedding) \cite{JMLR:v9:vandermaaten08a} before applying a Gaussian Mixture Model (GMM) to them.
Ellipsoids \cite{spruyt2014draw}, covering $95\%$ of data, are generated using eigenvectors and eigenvalues of the covariance matrix $\Sigma_c$ and the mean vector $\mu_c$, $\forall c \in \mathcal{C}, P_c \sim \mathcal{N}(x_c;\mu_c, \Sigma_c)$ obtained from the GMM to calculate the overlaps.
To avoid computing a degree of overlap for all $2^N$ subsets, the algorithm is early stopped when increasing the cardinality does not alter the list of 
most-overlapping sets of classes.
The $K$ non-singleton focal sets so obtained, along with the $N$ original (singleton) 
classes, form our network's 
budget of outputs
$\mathcal{O} = \mathcal{C} \cup \{ A_1,\ldots, A_K \}$. 
E.g., in a $100$-class scenario, the powerset contains $2^{100}$ subsets ($10^{30}$ possibilities). 
Setting a budget of $K = 200$, for instance, results in 
$100 + K = 300$ outputs, a far more manageable number.

As shown in our experiments, budgeting also helps the network converge without overfitting. While, in theory, a complete belief function model would be more powerful, in practice a larger number of focal sets impedes the learning process. As shown in Tab. \ref{tab:full-rsnn}, Sec. {\ref{app:ablation_k}}, RS-NN with a limited number of well-representative sets often 
performs better than RS-NN with a full power set of classes, and is more efficient at uncertainty estimation.
The budgeting step is a one-time procedure, applied to any given dataset before training. It requires 
just 2 minutes for CIFAR-10, 7 minutes for CIFAR-100, and {60 minutes}
for ImageNet ($\approx$ 1.1M images, 1000 classes).}

Further, our budgeting procedure is not specific to t-SNE, but can use any dimensionality reduction technique 
\cite{mackiewicz1993principal}. For instance, Uniform Manifold Approximation and Projection (UMAP) \cite{mcinnes2018umap} requires approximately 1 minute for the CIFAR-10 dataset, 2 minutes for CIFAR-100, and around 23 minutes for ImageNet to generate embeddings 
{(Tabs. \ref{tab:umap1}, \ref{tab:umap2} in Sec. \ref{app:budget} for a t-SNE \textit{vs.} UMAP ablation study).}
This is including the overlap computation which only takes a few seconds and is efficiently parallelised across 150 CPU cores. 

\begin{figure}[!htbp]
 \centering
       \centering       
              \vspace{-3pt}
       \includegraphics[width=0.8\linewidth]{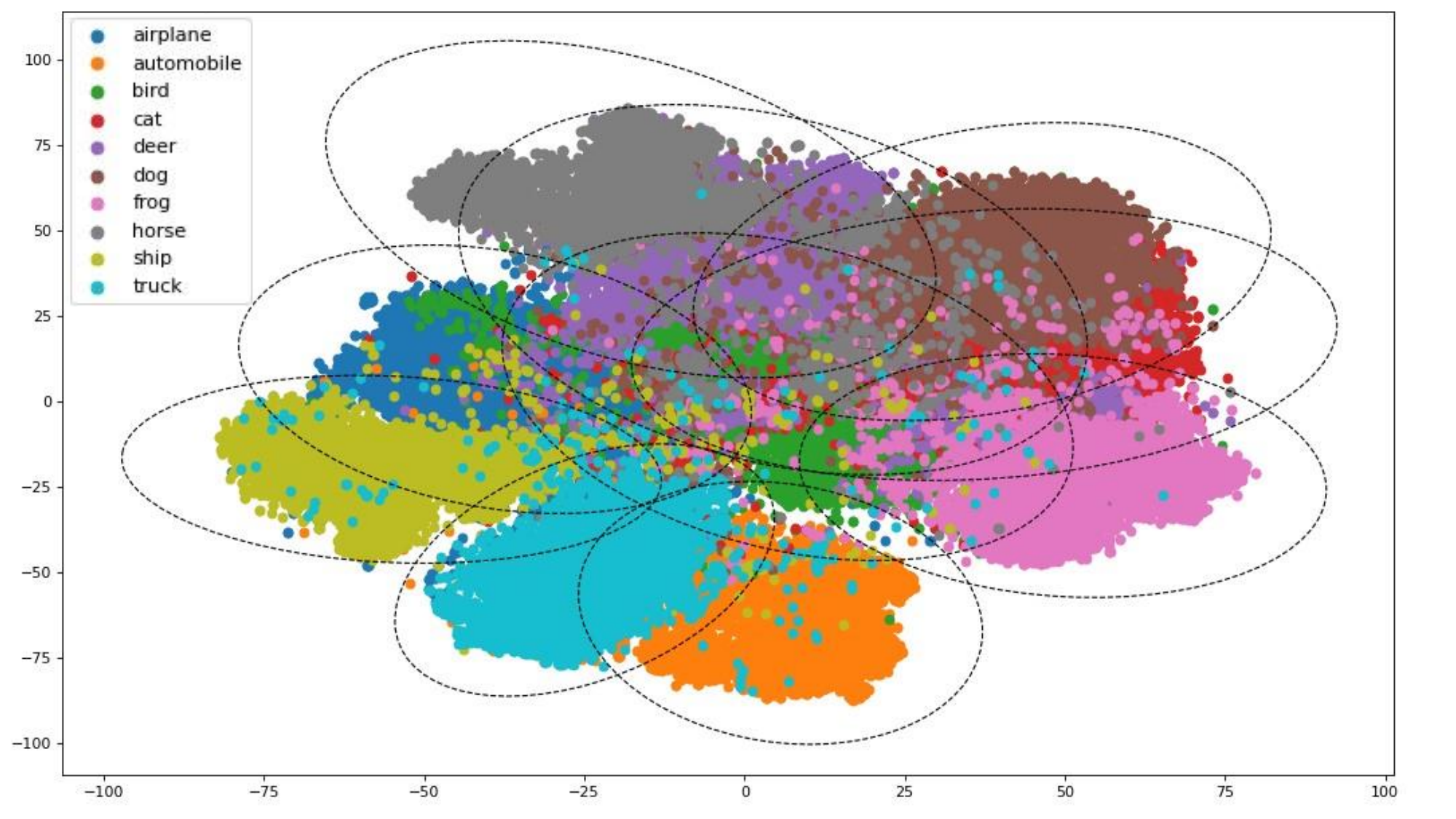}
              \vspace{-4pt}
       \caption{2D visualization of the clusters of 10 classes of CIFAR-10 dataset and the ellipses formed by RS-NN based on the hyperparameters of Gaussian Mixture Models.}
       \label{2Dclusters}
   \end{figure}

Fig. \ref{2Dclusters} shows a 2D visualization of the ellipses formed by computing the principal axes and their lengths from the eigenvectors and eigenvalues of the Gaussian Mixture Models (GMMs) fitted to each class $c$ \cite{spruyt2014draw}. Note that this is for visualization purposes only, in practice, the method operates in 3D.



\subsection{Training RS-NN: Loss Function}\label{sec:loss}

A random-set prediction problem is mathematically similar to the multi-label classification problem, for in both cases 
the ground truth vector contains several 1s. In the former case, these correspond to sets all containing the true class; in the latter, to all the class labels attached to same data point.
Despite the different semantics, we can thus adopt as loss Binary Cross-Entropy (BCE) \cite{bceloss} (Eq. \ref{bce-loss}) 
with sigmoid activation, 
to drive the prediction of a belief value
for each 
focal set in the identified budget:
\begin{equation}\label{bce-loss}
\begin{array}{lll}
    \mathcal{L}_{BCE} = \displaystyle
    - \frac{1}{b_{size}} \sum_{i=1}^{b_{size}} 
    \frac{1}{|\mathcal{O}|}
    \sum_{A \in \mathcal{O}}
    \Big [
    Bel_i(A) \log(\hat{Bel}_i(A)) + 
   (1 - Bel_i(A)) \log(1 - \hat{Bel}_i(A)) \Big ].
\end{array}
\end{equation}
Here, $i$ is the index of the training point within a batch of cardinality $b_{size}$, 
$A$ is a focal set of classes in the budget $\mathcal{O}$,
$Bel_i(A)$ is the $A$-th component of the vector $\mathbf{bel}_i$ encoding the ground truth belief values
for the $i$-th training point,
and $\hat{Bel}_i(A)$ is the corresponding belief value in the predicted vector $\hat{\mathbf{bel}}_i$ for the same training point. 
Both $\mathbf{bel}_i$ and $\hat{\mathbf{bel}}_i$ are vectors of cardinality $|\mathcal{O}|$ for all $i$.

A valid belief function satisfies Eq. \ref{eq:mass-non-neg} which states that mass values derived from belief functions should be non-negative and should sum up to 1 \cite{shafer1976mathematical}. To {promote} this, we incorporate
a mass regularization term $M_r$ and a mass sum term $M_s$ in the loss function:
\begin{equation} \label{mass-terms}
\begin{array}{l}
\displaystyle
    M_r = \frac{1}{b_{size}} \sum_{i=1}^{b_{size}}\sum_{A \in \mathcal{O}} \max(0, - \hat{m}_i(A)), \hspace{0.5em}
    \displaystyle
    M_s = \max \left(0, \frac{1}{b_{size}} \sum_{i=1}^{b_{size}}\sum_{A \in \mathcal{O}} \hat{m}_i(A) - 1 \right).
\end{array}
\end{equation}
$M_r$ encourages non-negativity of the (predicted) mass values $\hat{m}(A)$, $A \in \mathcal{O}$. These mass values are obtained from the predicted belief function $\hat{\mathbf{bel}}$ via the Moebius inversion formula (Eq. \ref{eq:belief-mobius}).
For it to be valid, the sum of the masses of the predicted belief function must be equal to 1 (Eq. \ref{eq:mass-non-neg}), which is encouraged by the mass 
sum term $M_s$.
All loss components $\mathcal{L}_{BCE}$, $M_r$ and $M_s$ are computed during batch training. 
Two hyperparameters, $\alpha$ and $\beta$, control the relative importance of the two mass terms, 
yielding as the total loss for RS-NN:
\begin{equation}\label{final_loss}
    \mathcal{L}_{RS} = \mathcal{L}_{BCE} + \alpha M_r + \beta M_s.
\end{equation}

The regularisation terms aim to penalise deviations from valid belief functions,
in line with, \textit{e.g.}, the way training time regularisation in neurosymbolic learning encourages predictions to be commonsense \cite{giunchiglia2023road}.
In general, soft constraints \cite{marquez2017imposing} have been shown {to be not} inferior to hard ones.
Still, when 
$\alpha$ and $\beta$ are too small, 
this
may not be sufficient to ensure that predictions are valid belief functions.
In such cases,\footnote{At any rate, improper belief functions are normally used in the literature \cite{denoeux2021distributed}, \textit{e.g.} for conditioning
\cite{cuzzolin2020geometry}.}
post-training, 
we 
set any negative masses
to zero and  
add the `universal' set of all classes to the final budget.
This subset is assigned all the remaining mass, ensuring that the sum of masses across all focal sets in $\mathcal{O}$ equals 1. 
This approach mimics classical approximation schemes (\textit{e.g.} \citet{cuzzolin2020geometry}, Part III).


\subsection{Accuracy \& Uncertainty Estimation}\label{uncertainty-estimation}

\textbf{Pignistic prediction}.
The pignistic probability (Eq. \ref{eq:pignistic}) is the central prediction associated with a belief function (seen as a credal set): standard performance metrics such as accuracy can then be calculated after extracting the most likely class according to the pignistic prediction 
(\textit{e.g.} in Sec. {\ref{app:pignistic}}).
Notably, the RS-NN architecture and training mechanism are designed to facilitate set-based learning by incorporating class set information during training.
When pignistic probabilities are computed during inference, they reflect the learning derived from the masses and beliefs of various subsets, making it more reliable than traditional softmax probabilities, as shown in Sec. {\ref{app:cnn-overconfidence}}.

\textbf{Entropy of the pignistic prediction}.  
The Shannon entropy of the pignistic 
prediction
$BetP$ 
can then be used as a measure of the 
uncertainty associated with the predictions, 
in some way analogous to the entropy of a Bayesian mean average prediction:
\begin{equation}\label{entropy-pred}
    H_{RS} = -\sum_{c \in \mathcal{C}}{BetP}(c)\log {BetP}(c).
\end{equation} 
A higher entropy value (Eq. \ref{entropy-pred}) indicates greater uncertainty in the model's predictions.

\textbf{Size of the credal prediction}. As discussed above, and further elaborated upon in Sec. \ref{app:ent-vs-credal},
a sensible measure of the epistemic uncertainty attached to a random-set prediction $\mathbf{\hat{bel}}$ is the size of the corresponding credal set.
Several ways exist of measuring the size of a convex polytope such as a credal set \cite{sale2023volume}.
Given the upper and lower bounds to the probability assigned to each class $c$ by
distributions within the predicted credal set $\hat{Cre}$ (Eq. \ref{eq:consistent}), 
\begin{equation} \label{eq:size-credal}
\overline{P}(c) = \max_{P \in 
\hat{Cre}
} P(c), \quad \underline{P}(c) = \min_{P \in 
\hat{Cre}
} P(c),
\end{equation}
we propose a simple way to measure the size of the credal set as the difference between the lower and upper bounds (Eq. \ref{eq:size-credal}) associated with the most likely class, according to the pignistic prediction.
Note that the predicted pignistic 
estimate $BetP(c)$ falls within the interval $[\underline{P}(c), \overline{P}(c)]$
for each class $c$, not just for the most likely class $\hat{c}$. Its width $\overline{P}(c) - \underline{P}(c)$
indicates the epistemic uncertainty associated with the prediction. 

\begin{figure}[!ht]
    \centering   \includegraphics[width=1\textwidth]{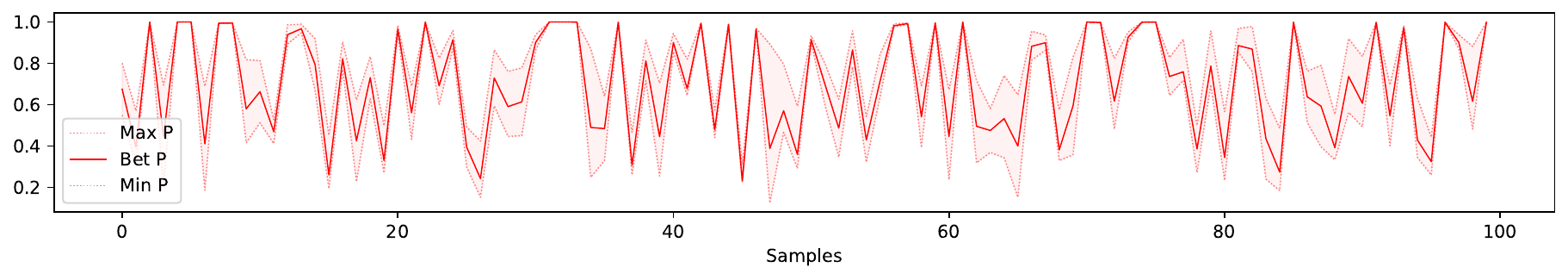}
    \caption{Upper 
    and lower 
    bounds (Eq. \ref{eq:size-credal}), in dotted red, to the probability of the predicted most likely class according to the pignistic prediction (in solid red)
    for 100 samples of CIFAR-10. 
    }
    \label{credal-set}
\end{figure}

As they have belief values as lower bounds, credal sets induced by belief functions have a peculiar shape (Fig. \ref{fig:belief-credal}, right, for a case with 3 classes). 
Their vertices are induced by the permutations of the elements of the sample space (for us, the set of classes) \cite{Chateauneuf89,cuzzolin2008credal}. Given one such permutation, \textit{e.g.}, $(c_2, c_4, c_1, c_3)$ for a set of 4 classes,
the corresponding probability vector 
(vertex of the credal prediction) assigns to each 
class 
the mass of all the focal sets containing it, 
\emph{but} not containing any class preceding it in the permutation order \cite{wallner2005}.

For additional illustration, 
the lower and upper bounds (Eq. \ref{eq:size-credal}) to the probability of the top predicted class are plotted in Fig. \ref{credal-set}, together with the pignistic probability, for 100 samples of CIFAR-10.
The bounds
Eq. \ref{eq:size-credal} can be efficiently computed from the finite number of vertices of the credal prediction (\cite{cuzzolin2008credal}, Fig. \ref{fig:belief-credal}).

There has been considerable research on uncertainty measures for credal sets. One of the earlier works by \citet{yager2008entropy} makes the distinction between two types of uncertainty within a credal set: conflict (also known as randomness or discord) and non-specificity. Non-specificity essentially varies with the size of the credal set \cite{kolmogorov1965three}. Since by definition, aleatoric uncertainty refers to the inherent randomness or variability in the data, while epistemic uncertainty relates to the lack of knowledge or information about the system \cite{hullermeier2021aleatoric}, conflict and non-specificity directly parallel these concepts. These measures of uncertainty are axiomatically justified \cite{bronevich2008axioms}. 

\textit{Predictive uncertainty} in RS-NN is represented as the pignistic entropy of predictions $H_{BetP}$, whereas \textit{epistemic uncertainty} can be modelled by the `size' of the credal set (Eq. \ref{eq:consistent}).

\subsection{RS-NN Learning Mechanism}\label{app:rsnn-learning}

Here we wish to expand on the rationale behind our choices for ground-truth representation $\mathbf{bel}$, loss function and training, and the way in which RS-NN learns sets of outcomes.

Recall from Sec. \ref{sec:arch} that the ground-truth for a RS-NN is the belief-encoded vector $\mathbf{bel} = \{ Bel(A), A \in \mathbb{P}(\mathcal{C}) \}$, where $Bel(A)$ is the belief function of a focal set $A$ in the power set $\mathbb{P}$ of classes $\mathcal{C}$. In our method, $Bel(A)$ is 1 if the true class is in subset $A$ and 0 otherwise. Consequently, the belief-encoded ground truth $\mathbf{bel}$ will include multiple occurrences of 1. 
For instance, in a digit-classification task such as MNIST, if $\{3\}$ is the true class, the belief-encoded ground truth would contain 1s in all sets where $\{3\}$ is present, such as \{1, 3\}, \{0, 3\}, \{1, 2, 3\}, and so forth, 0 for sets not containing $\{3\}$ at all, such as \{0, 1\}, \{1, 2\},  \{0, 1, 2\}, \textit{etc.} 

Since we assign precise labels for belief containing different focal sets, our model begins by penalizing predictions that differ from the observed label in the same manner, regardless of the set's composition or relationship to the true label. Using MNIST as an example, this means that the loss incurred for predicting label $\{3\}$ is equivalent to predicting label $\{3, 7\}$, as both sets contain the true label. This equivalence in losses might seem counterintuitive at first glance. However, despite the identical loss values, the probabilities output by the sigmoid activation function will vary due to differences in the input logit values for each label. For instance, if the correct label is $\{3\}$, the loss for predicting $\{3\}$ or $\{3, 7\}$ would be the same. Still, the loss for predicting $\{7\}$ would differ, allowing the model to discern the set structure during training.

During the training process, the model learns to capture and understand these relationships by observing patterns and dependencies in the training data. As the model optimizes its parameters based on the training objective (\textit{e.g.}, minimizing the loss function), it gradually adjusts its internal representations to better reflect these relationships. We use as the basis for our loss function Binary cross-entropy with sigmoid activation. It allows the model to predict the presence or absence of each label separately as a binary classification problem, producing probabilities between 0 and 1 for each class. While the model is unaware that it is learning for sets of outcomes, we leverage this technique to extract mass functions and pignistic predictions from the learnt belief functions. By re-distributing masses to original classes, we obtain the best pignistic predictions that are often more accurate than standard CNN predictions, attributing to the higher accuracy in Tab. \ref{tab:accuracy-table}. 

\section{Experiments}
\label{sec:rsnn-experiments}

\textbf{Firstly}, we assess the \emph{test accuracy} (\%) and \emph{inference time} (ms) of all baselines on all the multi-class classification datasets. Tab. \ref{tab:accuracy-table} provides a comparison of test accuracies and inference times of RS-NN against the baselines. 
Our \textbf{second} set of experiments (Sec. \ref{sec:ood-detection}) concerns \emph{out-of-distribution (OoD) detection}. 
Tab. \ref{tab:ood-table} shows our results on OoD detection metrics AUROC (Area Under Receiver Operating Characteristic curve) and AUPRC (Area Under Precision-Recall curve) for all the models on the iD \textit{vs.} OoD datasets listed above. We also report the Expected Calibration Error (ECE) for all models on datasets CIFAR-10, MNIST and ImageNet (Tab. \ref{tab:ood-table}).
In a \textbf{third} set of experiments (Sec. \ref{sec:uncertainty}), we test the \emph{uncertainty estimation} capabilities of RS-NN using both the entropy of the pignistic prediction and the size of the predicted credal set. Tab. \ref{tab:ood-table} presents the predicted pignistic entropy of RS-NN alongside entropies of all baselines on iD and OoD datasets.
Tab. \ref{tab:credal-table} shows credal set widths for RS-NN predictions across the same
datasets.
In our \textbf{final} set of experiments, we explore the \emph{scalability of RS-NN} (Sec. \ref{sec:scalability}) to large-scale architectures, employing it on models such as WideResNet-28-10, VGG16, Inception V3, EfficientNetB2 and ViT-Base-16. Further, to underscore the 
model's ability to leverage 
transfer learning, we train and test on a pre-trained ResNet50 model initialised with ImageNet weights 
(Tab. \ref{tab:scalability}). 

\textbf{Additional experiments} 
concern obtaining statistical guarantees for RS-NN using non-conformity scores (Sec. {\ref{app:motivation}}), 
evaluating the robustness of RS-NN to adversarial attacks (Sec. {\ref{app:fgsm}}), 
noisy and rotated in-distribution samples (Appendix Sec. {\ref{app:noisy}}), showing how RS-NN circumvents the overconfidence problem in CNNs (Sec. {\ref{app:cnn-overconfidence}}). 
Results showing how credal set width is less correlated with confidence scores than entropy are discussed in
in Sec. {\ref{app:ent-vs-credal}}. We also conduct ablation studies on $\alpha$ and $\beta$ 
(Sec. {\ref{app:hyperparam}}) and the number of non-singleton focal sets $K$ (Sec. {\ref{app:ablation_k}}), which 
show
that the best accuracy is obtained at small values ($1e-3$) of $\alpha$ and $\beta$, and for $K=20$ (on the CIFAR-10 dataset).




\subsection{Implementation}\label{sec:implementation}
\subsubsection{Datasets}

Our experiments are performed on multi-class image classification datasets, including MNIST \cite{LeCun2005TheMD}, CIFAR-10 \cite{cifar10}, Intel Image \cite{intelimage}, CIFAR-100 \cite{krizhevsky2009learning}, and ImageNet \cite{deng2009imagenet}. For out-of-distribution (OoD) experiments, we assess several in-distribution (iD) \textit{vs.} OoD datasets: CIFAR-10 \textit{vs.} SVHN \cite{netzer2011reading}/Intel-Image  \cite{intelimage}, MNIST \textit{vs.} F-MNIST \cite{xiao2017fashion}/K-MNIST \cite{clanuwat2018deep}, and ImageNet \textit{vs.} ImageNet-O \cite{hendrycks2021natural}. The data is split into 40000:10000:10000 samples for training, testing, and validation respectively for CIFAR-10 and CIFAR-100, 50000:10000:10000 samples for MNIST, 13934:3000:100 for Intel Image, 1172498:50000:108669 for ImageNet. For OoD datasets, we use 10,000 testing samples, except for Intel Image (3,000) and ImageNet-O (2,000).
Training images are resized to $224 \times 224$ pixels.

\subsubsection{Baselines and backbones}\label{app:trainingdetails}

Our baselines include state-of-the-art Bayesian methods 
LB-BNN \cite{hobbhahn2022fast} and FSVI \cite{rudner2022tractable}, Ensemble classifiers DE (5 ensembles) \cite{lakshminarayanan2017simple} and ENN (3 ensembles) \cite{osband2024epistemic}, 
and standard neural network (CNN). 
All models, including RS-NN, are trained on ResNet50 (on NVIDIA A100 80GB GPUs) with a learning rate scheduler initialized at 1e-3 with 0.1 decrease at epochs 80, 120, 160 and 180. 
{Standard data augmentation \cite{NIPS2012_c399862d}}, including random horizontal/vertical shifts with a magnitude of 0.1 and horizontal flips, is applied to all models. 

Laplace Bridge Bayesian approximation (LB-BNN) \cite{hobbhahn2022fast} uses the Laplace Bridge to map efficiently between Gaussian and Dirichlet distributions and enhance computational efficiency. In our experiments, we obtain multiple samples from the LB-BNN posterior and average these samples using Bayesian Model Averaging (BMA). Function-Space Variational Inference (FSVI) \cite{rudner2022tractable} utilizes variational inference to derive an approximation of the posterior distribution over the function space. FSVI proposes approximating distributions over functions as Gaussian by linearizing their mean parameters and derived a tractable and well-defined variational posterior. Deep Ensembles (DEs) \cite{lakshminarayanan2017simple} and Epistemic Neural Networks (ENNs) \cite{osband2024epistemic} are ensemble methods that train ensembles of models to efficiently estimate uncertainty. The mean and variance of ensemble predictions are calculated and the mean is softmaxed to obtain predictions from DE and ENN. The training hyperparameters for all models are outlined in Tab. \ref{tab:training_hyperparameters}.

\begin{table}[!h]
\centering
\caption{Training Hyperparameters for All Models}
    \vspace{-10pt}
\label{tab:training_hyperparameters}
\resizebox{1\linewidth}{!}{%
\begin{tabular}{l l}
\toprule
\textbf{Parameter}                   & \textbf{Value}                                      \\ 
\midrule
\textbf{Architecture}                & ResNet50 (excluding top classification layer)      \\ 
\textbf{Additional Dense Layers}     & 1024 and 512 neurons (ReLU activation)             \\ 
\textbf{Output Layer Activation (RS-NN)} & Sigmoid (multi-label classification)          \\ 
\textbf{Output Layer Activation (Other Models)} & Softmax (multi-class classification)        \\ 
\midrule
\textbf{Learning Rate}               & Initial: 1e-3                                      \\ 
\textbf{Learning Rate Scheduler}     & Decrease by 0.1 at epochs 80, 120, 160, 180       \\ 
\textbf{Training Epochs}             & 200                                                \\ 
\textbf{Batch Size}                  & 128                                                \\ 
\textbf{Optimizer}                   & RS-NN: Adam, LB-BNN: Adam, ENN: Adam, DE: Adam, FSVI: SGD \\ 
\midrule
\textbf{Training Dataset Sizes}      & \begin{tabular}[c]{@{}l@{}}CIFAR-10: 40,000 \\ CIFAR-100: 40,000 \\ MNIST: 50,000 \\ Intel Image: 13,934 \\ ImageNet: 1,172,498\end{tabular} \\ 
\midrule
\textbf{Testing Dataset Sizes}       & \begin{tabular}[c]{@{}l@{}}CIFAR-10: 10,000 \\ CIFAR-100: 10,000 \\ MNIST: 10,000 \\ Intel Image: 3,000 \\ ImageNet: 2,000\end{tabular} \\ 
\midrule
\textbf{Testing Samples for OoD Datasets} & 10,000 (except Intel Image: 3,000)              \\ 
\midrule
\textbf{Input Image Size}            & 224 × 224                                         \\ 
\textbf{Data Augmentation}           & Random horizontal/vertical shifts (magnitude 0.1), horizontal flips \\ 
\textbf{GPU}                         & NVIDIA A100 80GB                                   \\ 
\bottomrule
\end{tabular}%
}
\end{table}

\subsubsection{Training details}\label{sec:training-rsnn}

ResNet50, excluding the top classification layer, serves as the common architecture. 
ResNet was originally designed for classification on the ImageNet dataset (1000 classes). To accommodate a reduced number of classes as in smaller datasets (\textit{e.g.}, CIFAR-10/MNIST with 10 classes, Intel Image with 7 classes, and CIFAR-100 with 100 classes), two additional dense layers (1024 and 512 neurons, ReLU activation) are added to ResNet50.
Similar techniques are commonly applied in deep learning to adapt model architectures to datasets \cite{He_2016_CVPR, zagoruyko2016wide}.
The output layer of RS-NN on ResNet50 has the same number of units as the number of (selected) focal sets $|\mathcal{O}|$, and uses sigmoid activation, since the ground truth encoding resembles a multi-label classification problem 
(Sec. \ref{sec:loss}).

For all other models, 
the final output layer simply consists of a softmax activation for multi-class classification. 
All the models were trained from scratch for 200 epochs (recommended by most), with a batch size of 128.
{\emph{Our objective is to ensure a fair comparison across all models for all experiments.} Tuning each model separately to maximise performance would not guarantee 
that,
as some models use pre-trained weights while others train for larger number of epochs, which could result in the over-training of another model. We used pre-trained weights for all models only when trained on ImageNet for efficiency.}
For all other training hyperparameters (e.g, optimizer), we use what is specified for each model in their respective papers (more in Sec. {\ref{app:trainingdetails}}, Tab. \ref{tab:training_hyperparameters}). 
Consequently, the performance metrics reported in the original papers may vary from those presented here, as they were obtained under different training conditions.

In the budgeting phase, we obtain features directly from a pre-trained ResNet50 classifier.
We use 
50 CPU cores for t-SNE dimensionality reduction, and 150 CPU cores for computing the overlap of classes.
We set
a budget $K$ of 20 focal sets (ablation study on K in Sec. {\ref{app:ablation_k}}) for CIFAR-10/ MNIST/ Intel Image, 200 for CIFAR-100 and 3000 for ImageNet. 
RS-NN is trained from scratch on ground-truth belief encoding of sets using the $\mathcal{L}_{RS}$ loss function (Eq. \ref{final_loss}) over 200 epochs, with a batch size ($b_{size}$) of 128 and $\alpha = \beta = 1e-3$ as hyperparameter values. 


\subsubsection{Sample prediction}

\begin{table}[!h]
\setlength{\tabcolsep}{3pt}
\label{pred-table}
\small
\centering
\caption{The predicted belief, mass values, pignistic probabilities, and entropy for two CIFAR-10 predictions. The figure on the top (True Label = `\texttt{horse}') is a certain prediction with 99.9\% confidence and a low entropy of 0.0017, whereas the figure on the bottom  (True Label = `\texttt{cat}') is an uncertain prediction with 33.3\% confidence and a higher entropy of 2.6955. }
\vspace{-8pt}
\label{tab:prediction-table}
\resizebox{\textwidth}{!}{
\begin{tabular}{lllll}
\toprule
 Sample & Belief & Mass & Pignistic & Entropy \\         
\midrule \\
\multirow{5}{1.9cm}[2.9ex]{\raisebox{-0.2\totalheight}{\includegraphics[width=0.1\textwidth]{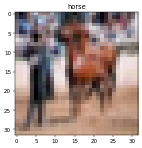}}}
& \begin{tabular}[t]{ll}
  \{`\texttt{horse}', `\texttt{bird}'\} & 0.9999402 \\
  \{`\texttt{horse}', `\texttt{dog}'\} & 0.9999225 \\
  \{`\texttt{horse}'\} & 0.9999175 \\
  \{`\texttt{horse}', `\texttt{deer}'\} & 0.9998697 \\
  \{`\texttt{cat}', `\texttt{truck}'\} & 7.0380207e-05
\end{tabular}
& \begin{tabular}[t]{ll}
  \{`\texttt{horse}'\} & 0.9999175 \\
  \{`\texttt{cat}', `\texttt{truck}'\} & 6.859753e-05 \\
  \{`\texttt{ship}', `\texttt{bird}'\} & 4.094290e-05 \\
  \{`\texttt{horse}', `\texttt{bird}'\} & 2.250525e-05\\
  \{`\texttt{dog}'\} & 1.717869e-05
\end{tabular}
& \begin{tabular}[t]{ll}
  horse & 0.9998833 \\
  truck & 3.5826409e-05 \\
  cat & 3.3859180e-05 \\
  bird & 3.3015738e-05\\
  ship & 2.3060647e-05
\end{tabular}
& 0.0017040148 \\\\\\
\multirow{5}{1.9cm}[2.9ex]{\raisebox{-0.1\totalheight}{\includegraphics[width=0.1\textwidth]{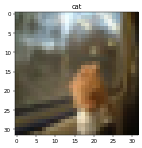}}} 
 & \begin{tabular}[t]{ll}
  \{`\texttt{deer}', `\texttt{cat}'\} & 0.4787728 \\
  \{`\texttt{deer}', `\texttt{airplane}', `\texttt{bird}'\} & 0.4126398 \\
  \{`\texttt{horse}', `\texttt{deer}'\} & 0.3732957 \\
  \{`\texttt{deer}', `\texttt{bird}'\} & 0.3658997 \\
  \{`\texttt{deer}', `\texttt{dog}'\} & 0.3651531
\end{tabular}
& \begin{tabular}[t]{ll}
  \{`\texttt{deer}'\} & 0.3104962 \\
  \{`\texttt{cat}', `\texttt{truck}'\} & 0.1762222 \\
  \{`\texttt{dog}', `\texttt{bird}'\} & 0.0998060 \\
  \{`\texttt{horse}', `\texttt{bird}'\} & 0.0954350\\
  \{`\texttt{bird}'\} & 0.0524873
\end{tabular}
& \begin{tabular}[t]{ll}
    deer		&	0.3332411 \\
    cat		&	0.2230723 \\
    horse	&		0.1153417 \\
    bird	&		0.1086245 \\
    dog		&	0.1039505 \\
\end{tabular}
& 	2.6955228261
\\
\bottomrule
\end{tabular} 
}
\end{table}

\begin{figure}[!h]
    \centering
    \includegraphics[width=0.9\textwidth]{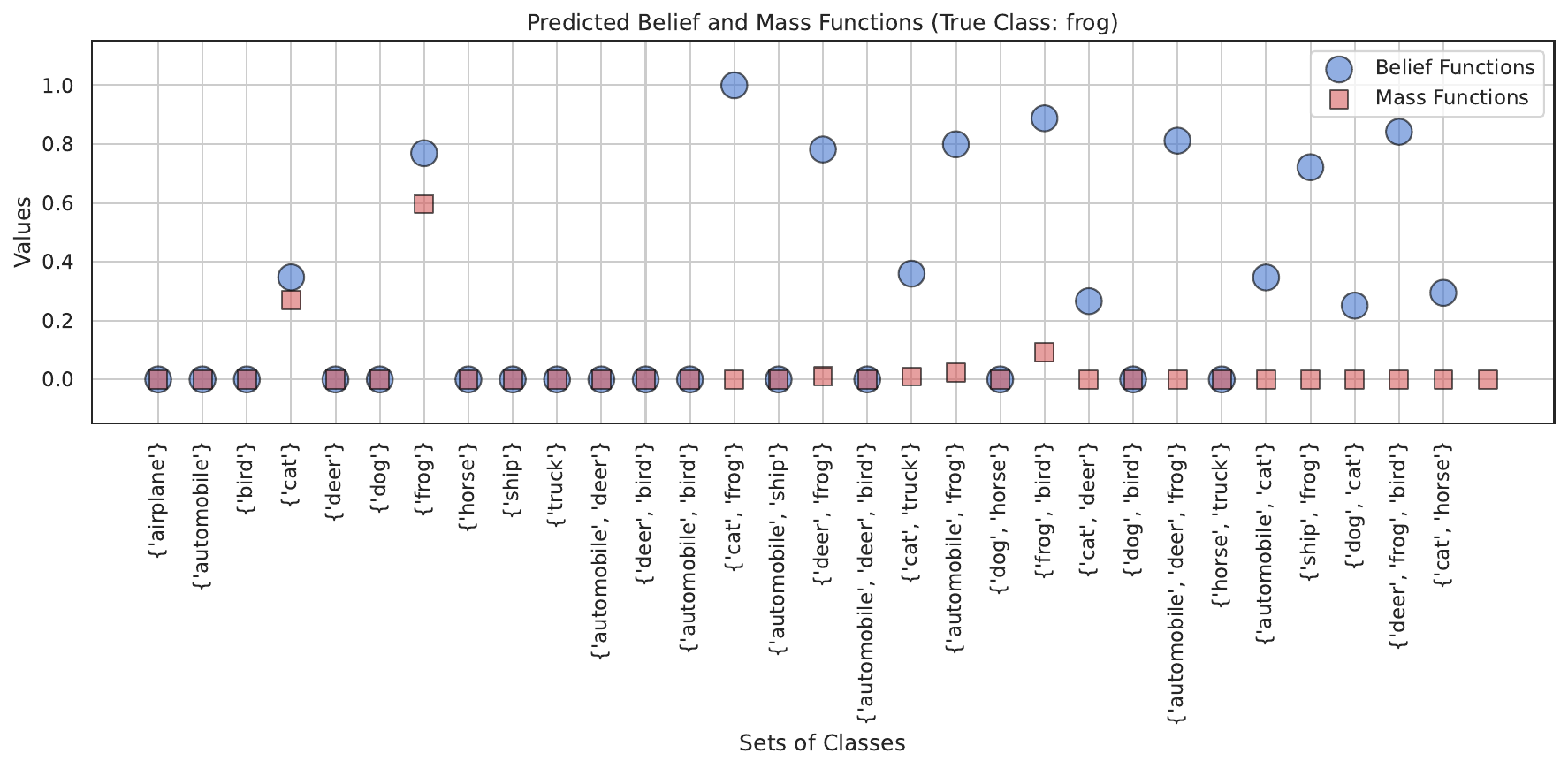}
    \centering
    \includegraphics[width=0.44\textwidth]{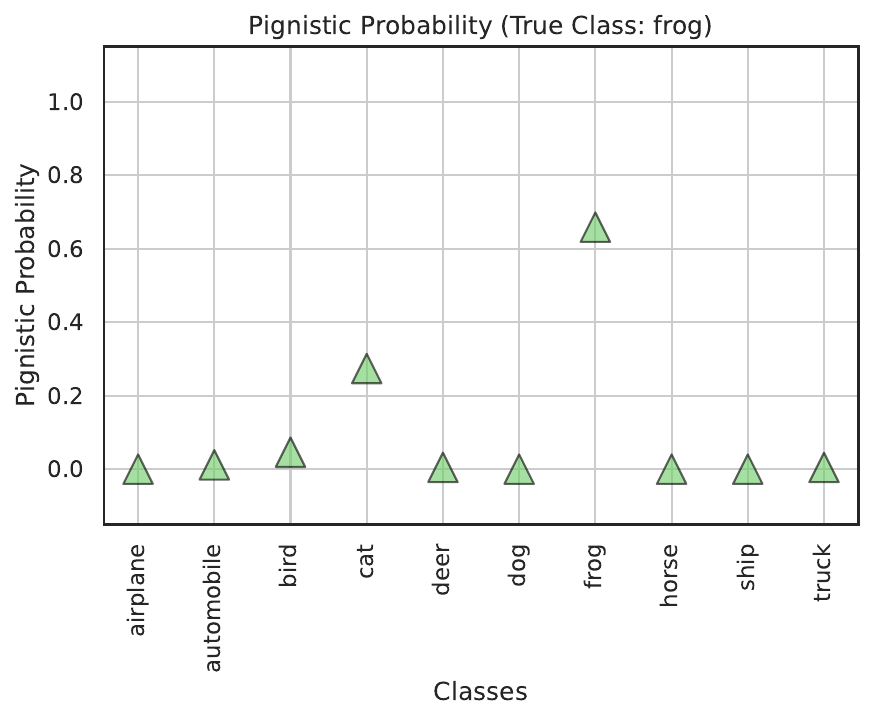}
    \caption{Visualizations of belief and mass predictions on the power-set space and its mapping to the label space using pignistic probabilities on the CIFAR-10 dataset. 
    }
    \label{fig:RS_SNN_pred}
\end{figure}

Since RS-NN predictions differ from those of other baseline models, this section presents \textbf{sample predictions} produced by RS-NN.
The predicted belief, mass values, pignistic probabilities, and entropy for two CIFAR-10 predictions are illustrated in Tab. \ref{tab:prediction-table}. In the top figure, corresponding to the true label `horse' the model makes a highly confident prediction with 99.9\% confidence and a low entropy of 0.0017. Conversely, the bottom figure, associated with the true label `cat' represents an uncertain prediction with 33.3\% confidence and a higher entropy of 2.6955. It is important to note that the second image is slightly unclear and poor in quality.

Fig. \ref{fig:RS_SNN_pred} shows belief function predictions for an input sample with the true class `\verb+frog+'. The belief function predictions and their mapping to mass functions and pignistic probabilities are illustrated.

\subsection{Performance: Accuracy \& Expected Calibration Error}\label{sec:acc}

\begin{table}[!t]
  \caption{
  Test accuracies (\%) and inference time (ms) for uncertainty estimation 
  over 5 consecutive runs across methods and datasets. Average and standard deviation are shown for each experiment.}
  \label{tab:accuracy-table}
  \centering
      \vspace{-10pt}
  \resizebox{\linewidth}{!}{
  \begin{tabular}{lllllll|l}
    \toprule
    Datasets & MNIST & CIFAR-10 & Intel Image &  CIFAR-100 & ImageNet (Top-1) & ImageNet (Top-5) & Inference time (ms) \\
    \midrule
    \textbf{RS-NN} (ours)   &  $\mathbf{99.71} \pm \mathbf{0.03}$ & $\mathbf{93.53}\pm \mathbf{0.09}$ & $\mathbf{94.22} \pm \mathbf{0.03}$ & $\mathbf{71.61} \pm \mathbf{0.07}$ & $\mathbf{79.92}$ & $\mathbf{94.47}$ & $\mathbf{1.91} \pm \mathbf{0.02}$ \\
    LB-BNN \cite{hobbhahn2022fast}  & $99.58 \pm 0.04$  &  $89.95 \pm 0.81$ & $90.49 \pm 0.42$ &$59.89 \pm 1.96$ & $72.48$ &  $90.85$ & $7.11 \pm 0.89$\\
    FSVI \cite{rudner2022tractable} & $99.18 \pm 0.03$ & $80.29 \pm 0.05$ & $88.92 \pm 0.13$ & $53.34 \pm 0.09$ & $62.56$ & $84.69$ & $340.25 \pm 0.76$\\
    DE \cite{lakshminarayanan2017simple} & $99.25 \pm 0.01$ & $92.73 \pm 0.04$ & $91.98 \pm 0.11$ & $70.53 \pm 0.07$ & $78.77$ & $94.37$ & $13163.50 \pm 3.37$\\
    ENN \cite{osband2024epistemic}  & $99.07 \pm 0.11$& $91.55 \pm 0.60$ & $91.49 \pm 0.19$  & $68.02 \pm 0.26$ & $71.82$ & $89.48$ & $3.10\pm 0.03$\\
    CNN & $99.12 \pm 0.04$   & $92.08\pm 0.42$ & $90.89 \pm 0.10$ &  $65.50 \pm 0.08$ &  $78.56$ & $94.34$ & $\mathbf{1.91} \pm \mathbf{0.03}$ \\ 
    \bottomrule
  \end{tabular}  
    }
\vspace{-4mm}
\end{table}

Tab. \ref{tab:accuracy-table} 
shows that RS-NN outperforms state-of-the-art Bayesian (LB-BNN, FSVI) and Ensemble (DE, ENN) methods in terms of test accuracy (\%) across all datasets. 
Experiments on inference time (ms) per sample, conducted over 5 runs (a single forward propagation) on the CIFAR-10 dataset, show that RS-NN and CNN have the fastest inference times among all models (Tab. \ref{tab:accuracy-table}).
{Albeit CNN and DE are close to RS-NN in performance (\textit{e.g.}, on ImageNet), CNN tends to be overconfident in incorrect predictions (Sec. {\ref{app:cnn-overconfidence}}), while}
{DE has significantly longer training (Tab. \ref{tab:trainingtime}, Sec. {\ref{app:trainingdetails}}) and inference times (Tab. \ref{tab:accuracy-table}).}
The 
pignistic
predictions
derived from predicted belief functions 
are
more reliable and accurate 
than predicted softmax probabilities of CNN and DE.
It is important to note that the comparison with standard CNN is not based on its role as an uncertainty method, but as a benchmark, underscoring RS-NN's effectiveness in achieving high accuracy compared to other uncertainty baselines.
Test accuracy for RS-NN is calculated by determining the class with the highest pignistic probability for each prediction 
and 
comparing it 
with the true class.
This proves that, by modelling the epistemic uncertainty about the prediction, we take better into account possible distribution shifts at test time - as a result, the central prediction of the set of probabilities (credal set) associated with the predicted belief function is more likely to be closer to the ground truth. 
Note that the results in the FSVI paper \cite{rudner2022tractable} are based on 
ResNet18. Although using a pre-trained ResNet50 could enhance FSVI's performance, it would provide an unfair advantage over the other baseline models in our paper, which were all trained from scratch.

We report the training times (in minutes) for all models on the CIFAR-10 dataset in Tab. \ref{tab:trainingtime}. Tab. \ref{tab:trainingtime} shows that FSVI has the longest training time, followed by RNN and DE. Compared to LB-BNN, RS-NN adds 5 minutes to the training duration, while CNN has the shortest training time. However, CNNs do not provide uncertainty estimation, and RS-NN has more reliable uncertainty and OoD metrics compared to LB-BNN.

The significant training time required for Deep Ensembles (DE) often does not justify their uncertainty estimates \citep{abe2022deep, rahaman2021uncertainty}. Despite the computational cost, DE is sometimes only as good as other uncertainty estimation models, which can be trained much faster. Other approaches, like methods with calibration techniques (\textit{e.g.}, temperature scaling or Bayesian approaches), can provide similar performance without the substantial overhead that DE entails.

\textcolor{blue}{
\begin{table}[!h]
    \centering
    \vspace{-18pt}
    \caption{Training time (\SI{}\minute) for all models on the CIFAR-10 dataset.}
        \vspace{-10pt}
    \label{tab:trainingtime}
    \resizebox{0.7\linewidth}{!}{
        \begin{tabular}{ccccccc}
        \toprule
        & RS-NN &
        LB-BNN & FSVI & DE & ENN & CNN\\
        \midrule
        Training time (\SI{}\minute) & $113.23$ & 
        $107.900$ & $1518.35$ & $426.66$ & $712.302$ & $\mathbf{85.333}$\\
        \bottomrule
         & 
    \end{tabular}
    }
    \vspace{-15pt}
\end{table}
}

As mentioned in Sec. \ref{sec:acc}, the results for FSVI are different than what was reported in \citep{rudner2022tractable}. The lower performance with ResNet-50 in our experiments may be attributed to the model being optimized specifically for ResNet-18. We noticed the unexpectedly low results and adhered to their suggested optimizer, SGD, instead of Adam (used in all other models), to achieve optimal performance, as mentioned in Tab. \ref{tab:training_hyperparameters}. Additionally, to enhance performance, we also incorporated the context points from ImageNet and CIFAR-100 as recommended in \citet{rudner2022tractable}.
Furthermore, we observed that FSVI uses more extensive data augmentation than our standardized setup. For instance, their data augmentation includes random crops and higher magnitude random flips (0.5), while we only used random horizontal/vertical flips of magnitude 0.1, potentially contributing to the performance gap.
While increasing model capacity by using ResNet-50 (25.6M parameters) over ResNet-18 (11.7M parameters) should theoretically improve performance, this is not always the case in practice. ResNet-50 requires more careful optimization and is more prone to overfitting, particularly when used with smaller datasets or suboptimal augmentation. This explains why ResNet-50 did not consistently outperform ResNet-18 in our implementation of the compared methods.

\subsection{Out-of-distribution detection}\label{sec:ood-detection}

\begin{table}[!t]
\centering
\caption{OoD detection performance and uncertainty estimation for models trained on ResNet50 on CIFAR-10 \textit{vs.} SVHN/Intel Image, MNIST \textit{vs.} F-MNIST/K-MNIST and ImageNet \textit{vs.} ImageNet-O. Evaluation metrics include AUROC/AUPRC (OoD); Entropy of predictions (uncertainty) and Expected Calibration Error (ECE).}
\label{tab:ood-table}
\vspace{-8pt}
\resizebox{\textwidth}{!}{
\begin{tabular}{lcccccccc|ccc}
\toprule
\multicolumn{6}{c}{In-distribution (iD)} &
  \multicolumn{6}{c}{Out-of-distribution (OoD)} \\ \midrule
\multirow{2}{*}{Dataset} &
  \multirow{2}{*}{Model} &
  \multirow{2}{*}{\begin{tabular}[c]{@{}c@{}}Test accuracy\\ (\%) ($\uparrow$)\end{tabular}} &
  \multicolumn{1}{c}{\multirow{2}{*}{Uncertainty measure}} &
  \multicolumn{1}{c}{\multirow{2}{*}{\begin{tabular}[c]{@{}c@{}}In-distribution  \\  Entropy ($\downarrow$)\end{tabular}}} &
    \multicolumn{1}{c}{\multirow{2}{*}{ECE ($\downarrow$)}} &
  \multicolumn{3}{c|}{SVHN} & 
  \multicolumn{3}{c}{Intel Image} \\ \cmidrule{7-12} 
 & & & & & &
  \multicolumn{1}{c}{AUROC ($\uparrow$)} &
  \multicolumn{1}{c}{AUPRC ($\uparrow$)} &
  \multicolumn{1}{c|}{Entropy ($\uparrow$)} &
  \multicolumn{1}{c}{AUROC ($\uparrow$)} &
  \multicolumn{1}{c}{AUPRC ($\uparrow$)} &
  \multicolumn{1}{c}{Entropy ($\uparrow$)} \\ 
  \midrule
\multicolumn{1}{c}{\multirow{7}{*}{CIFAR-10}} &
  RS-NN &
  $\mathbf{93.53}$ &
  Pignistic entropy & $\mathbf{0.088} \pm \mathbf{0.308}$ &
  $\mathbf{0.0484}$&
  $\mathbf{94.91}$&
  $\mathbf{93.72}$ & 
  $\mathbf{1.132} \pm \mathbf{0.855}$ &
  $\mathbf{97.39}$	&
  $\mathbf{90.27}$ &
  $\mathbf{1.517} \pm \mathbf{0.740}$ 
    \\
\multicolumn{1}{c}{} &
  LB-BNN &
  $89.95$ &
  Predictive Entropy & $0.191 \pm 0.412$ &
  $0.0585$ &
  $88.14$	&
  $81.96$ &
  $0.828 \pm 0.243$ &
  $82.21$ &
  $55.17$ &
  $0.763 \pm 0.722$ \\
\multicolumn{1}{c}{} &
  FSVI &
  $80.29$ &
  Predictive Entropy & $0.118\pm 0.563$ &
  $0.0521$ &
  $80.59$	&
  $80.84$ &
  $0.413 \pm 0.461$ &
  $74.27$ &
  $72.51$ &
  $0.289 \pm 0.670$ \\
\multicolumn{1}{c}{} &
  DE &
  $92.73$ &
  Mean Entropy & $0.154\pm 0.367$ &
  $\mathbf{0.0482}$ &
  $93.84$	&
  $91.88$ &
  $0.939 \pm 0.554$ &
  $94.25$ &
  $79.36$ &
  $ 1.166\pm 0.552$ \\
\multicolumn{1}{c}{} &
  ENN &
  $91.55$ &
  Mean Entropy & $0.126 \pm 0.323$ &
  $0.0556$ &
  $92.76$	 &
  $89.05$ &
  $0.887 \pm 0.514$ &
  $85.67$ &
  $58.09$ &
  $0.600 \pm 0.578$ 
  \\ 
\multicolumn{1}{c}{} &
  CNN &
  $92.08$ &
  Softmax Entropy & $0.114 \pm 0.304$  &
  $0.0669$ &
  $93.11$ &
  $91.0$ &
  $0.930 \pm 0.610$ &
  $87.75$	 &
  $65.54$ &
  $0.719 \pm 0.673$ 
  \\
  \cmidrule{7-12} 
\multirow{10}{*}{MNIST} &
  \multicolumn{1}{l}{} &
  \multicolumn{1}{l}{} &
  \multicolumn{1}{l}{} &    
  \multicolumn{1}{l}{} &
  \multicolumn{1}{l}{} &
  \multicolumn{3}{c|}{F-MNIST} &  
  \multicolumn{3}{c}{K-MNIST} \\ \cmidrule{7-12} 
 &
  RS-NN &
  $\mathbf{99.71}$ &
  Pignistic entropy & $0.010 \pm 0.111$  &
  $\mathbf{0.0029}$ &
  $\mathbf{93.89}$	&
  $\mathbf{93.98}$ &
 $0.530 \pm 0.770$&
  $\mathbf{96.75}$ &
  $\mathbf{96.58}$ &
  $\mathbf{0.740} \pm \mathbf{0.917}$ 
  \\
 &
  LB-BNN &
  $99.58$ &
  Predictive Entropy &  $\mathbf{0.001} \pm \mathbf{0.085}$  &
  $0.0032$ &
  $89.65$ &
  $90.36$ &
 $0.287 \pm 0.442$ &
  $95.61$ &
  $95.65$ &
  $0.540 \pm0.621$
  \\
 &
  FSVI &
  $99.18$ &
  Predictive Entropy & $0.006 \pm 0.265$ &
  $0.0047$ &
  $92.79$	&
  $91.17$ &
  $0.264 \pm 0.289$ &
  $91.65$ &
  $95.75$ &
  $0.313 \pm 0.381$ \\
 &
  DE &
  $99.25$ &
  Mean Entropy & $0.031\pm 0.155$ &
  $0.0031$ &
  $92.30$	&
  $92.05$ &
  $\mathbf{0.584} \pm \mathbf{0.587}$ &
  $95.81$ &
  $94.71$ &
  $0.564 \pm 0.715$ \\
  &
  ENN &
  $99.07$ &
  Mean Entropy &  $0.022 \pm 0.127$  &
  $0.0039$&
  $81.79$ &
  $82.92$ &
  $ 0.313 \pm 0.464$&
  $95.94$ &
  $95.45$ &
  $0.503 \pm 0.672$ 
  \\ 
\multicolumn{1}{c}{} &
  CNN &
  $98.90$ &
  Softmax Entropy & $0.023 \pm 0.135 $ &
  $0.0052$&
  $83.77$ &
  $84.14$ &
  $ 0.278 \pm 0.426$ &
  $94.46$ &
  $93.94$ &
  $0.616 \pm 0.688$ 
  \\ \cmidrule{7-12}
{\multirow{10}{*}{ImageNet}} &
  \multicolumn{1}{l}{} &
  \multicolumn{1}{l}{} &
  \multicolumn{1}{l}{} &  & &
  \multicolumn{6}{c}{ImageNet-O} \\ \cmidrule{7-12} 
 &
  \multicolumn{1}{l}{} &
  \multicolumn{1}{l}{} &
  \multicolumn{1}{l}{} &   &  &
  \multicolumn{2}{c}{AUROC} &
  \multicolumn{2}{c}{AUPRC} &
  \multicolumn{2}{c}{Entropy} 
  \\
 &
  RS-NN &
  $\mathbf{79.92}$ &
  Pignistic entropy & 
  $2.972 \pm 2.108$ &  
  $\mathbf{0.1416}$ &
  \multicolumn{2}{c}{$\mathbf{60.38}$} &
  \multicolumn{2}{c}{$\mathbf{55.16}$} &
  \multicolumn{2}{c}{$3.659 \pm 3.771$} \\
 &
  LB-BNN &
  $72.48$ &
  Predictive Entropy &   
  $2.471 \pm 2.972$&  
  $0.5812$ &
  \multicolumn{2}{c}{$41.08$} &
  \multicolumn{2}{c}{$30.99$} &
  \multicolumn{2}{c}{$1.383 \pm 0.028$} \\
 &
  FSVI &
  $62.56$ &
  Predictive Entropy &   
  $ \mathbf{1.328} \pm \mathbf{1.966}$&  
  $0.3890$ &
  \multicolumn{2}{c}{$50.55$} &
  \multicolumn{2}{c}{$49.88$} &
  \multicolumn{2}{c}{$1.637 \pm 1.328$} \\
 &
  DE &
  $78.77$ &
  Mean Entropy &   
  $1.532 \pm 1.325$&  
  $0.1940$ &
  \multicolumn{2}{c}{$55.37$} &
  \multicolumn{2}{c}{$53.20$} &
  \multicolumn{2}{c}{$1.775 \pm 1.343$} \\
 &
  ENN &
  $71.82$ &
  Mean Entropy &  
  $1.395 \pm 1.510$ &   
  $0.5961$ &
  \multicolumn{2}{c}{$54.67$} &
  \multicolumn{2}{c}{$43.73$} &
  \multicolumn{2}{c}{$1.617 \pm 1.597$} \\
 &
  CNN &
   $78.56$ &
  Softmax Entropy & $6.386 \pm 1.388$  & 
  $0.4004$ &
  \multicolumn{2}{c}{$54.28$} &
  \multicolumn{2}{c}{$48.73$} &
  \multicolumn{2}{c}{$\mathbf{6.575 \pm 1.512}$} \\
\bottomrule
\end{tabular}
} 
\end{table}

We evaluate our out-of-distribution (OoD) detection (Tab. \ref{tab:ood-table}) against baselines using AUROC and AUPRC scores. OoD detection identifies data points deviating from the in-distribution (iD) training data by measuring true and false positive rates (AUROC) and precision and recall trade-offs (AUPRC), indicating the model's ability to handle unfamiliar data
(further detailed in 
\S{\ref{app:algo-auroc}}). 
As shown in Tab. \ref{tab:ood-table}, RS-NN greatly outperforms Bayesian, Ensemble and standard CNN models in OoD detection, with significantly higher AUROC and AUPRC scores for all iD \textit{vs.} OoD datasets, especially on difficult datasets like ImageNet and ImageNet-O
{(Sec. {\ref{app:imagenet-o}}).
Even high-accuracy models like CNN and DE struggle to differentiate between ImageNet and ImageNet-O. While DE shows comparable AUROC scores on CIFAR-10 \textit{vs.}  SVHN and MNIST \textit{vs.}  F-MNIST/K-MNIST to RS-NN, it has lower AUPRC. While AUROC assesses a model's ability to distinguish between iD and OoD samples, AUPRC measures both the precision of correctly identified OoD samples and the recall of detected OoD samples, making it a more informative metric.}
Tab. \ref{tab:ood-table} also 
reports the various models' Expected Calibration Error (ECE) (detailed in \S{\ref{app:algo-ece}}). A low ECE indicates that the model's confidence scores align closely with the actual likelihood of events. RS-NN exhibits the lowest ECE indicating well-calibrated probabilistic predictions. 

\subsection{Uncertainty Estimation}\label{sec:uncertainty}

\subsubsection{Entropy of pignistic predictions}
\label{app:pignistic}\label{app:entropy}

Tab. \ref{tab:ood-table} shows the mean and variance of the entropy distributions for both iD and OoD datasets. 
Lower entropy values for iD datasets indicate a confident prediction as the model is trained on familiar data, while higher entropy for OoD datasets reflects the model's uncertainty with 
unseen data. RS-NN exhibits both low entropy for iD datasets and high entropy for OoD ones. 
The percentage mean iD/OoD shift in entropy is most pronounced in RS-NN, especially evident in CIFAR-10 \textit{vs.} SVHN/Intel Image and MNIST \textit{vs.} F-MNIST/K-MNIST.
{For ImageNet \textit{vs.} ImageNet-O, CNN exhibits high entropy for both iD and OoD datasets while RS-NN maintains a desirable iD \textit{vs.} OoD entropy ratio. 

\begin{figure}[!h]
    \centering
    \includegraphics[width=\textwidth]{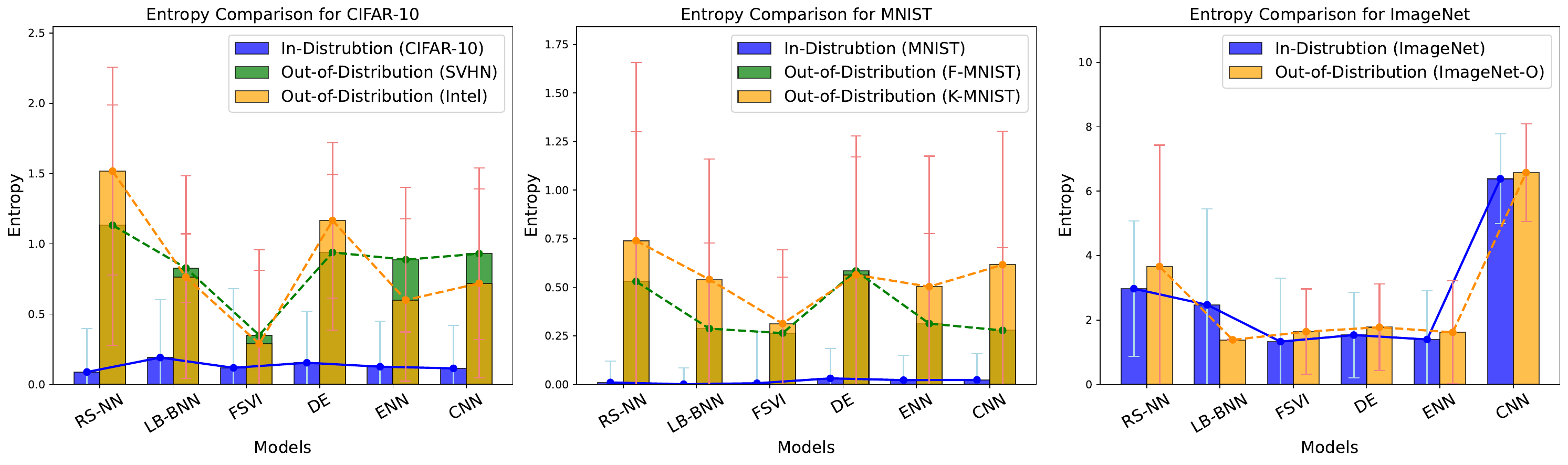}
    \vspace{-20pt}
    \caption{
    Entropy comparison of all models for iD and OoD datasets. The plots illustrate the performance of various models, with error bars indicating the standard deviation of entropy values.}
    \label{fig:entropy_comparison}
\end{figure}
Fig. \ref{fig:entropy_comparison} illustrates that RS-NN exhibits the best \textbf{iD \textit{vs.} OoD entropy ratio} among the evaluated baseline models. Specifically, RS-NN maintains significantly lower entropy values for iD datasets, indicating a strong ability to recognize familiar patterns while showing elevated entropy for OoD datasets. DE and CNN closely follow RS-NN but struggle to effectively differentiate entropy for the ImageNet dataset. LB-BNN shows more uncertainty regarding iD samples compared to OoD samples. FSVI demonstrates the weakest distinction between iD and OoD.

\begin{figure}[!h]
        \includegraphics[width=\linewidth]{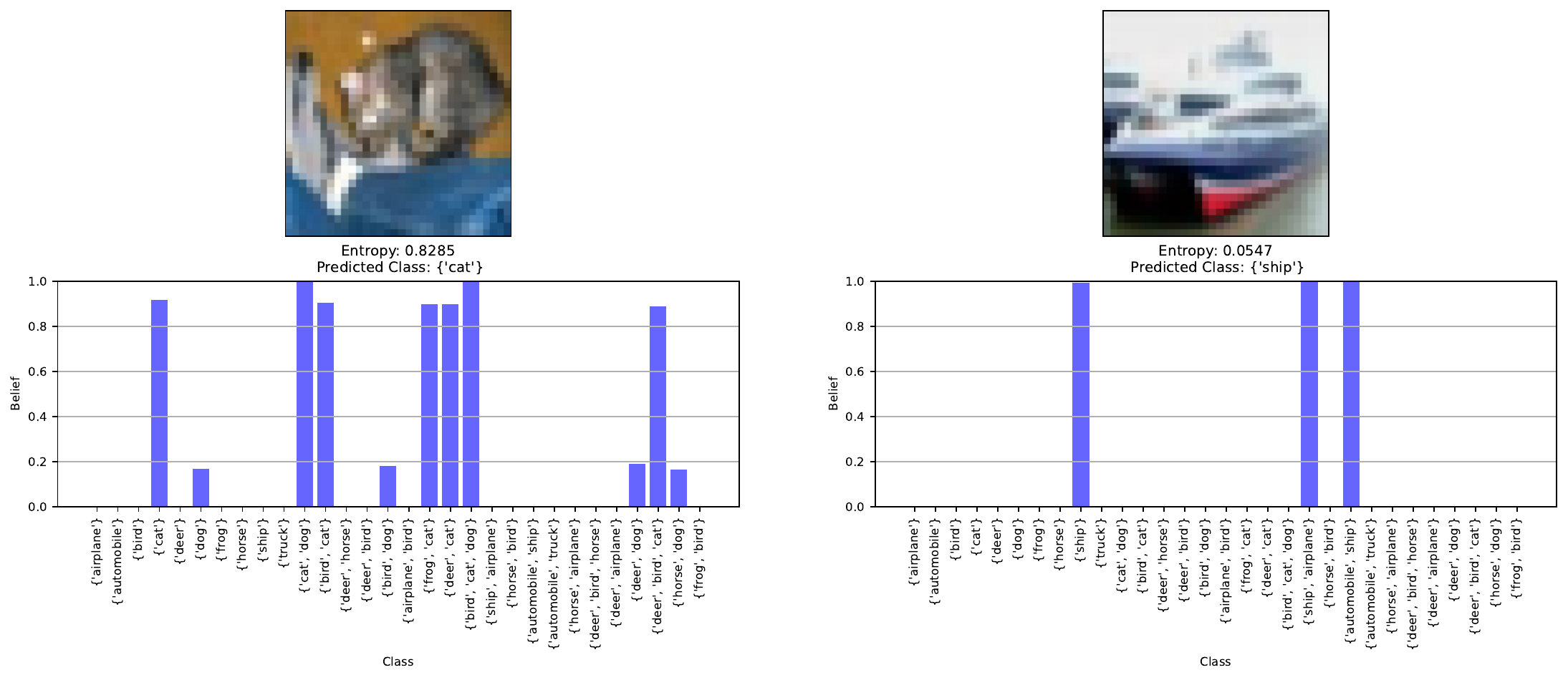} 
        \vspace{-10pt}
        \caption{Entropy, predicted class and belief value graph for two samples of CIFAR-10 dataset. 
        }
        \label{entropy_belief}
\end{figure}

Fig. \ref{entropy_belief} shows the \textbf{entropy for two samples of CIFAR-10}, one is a slightly uncertain image, whereas the other is certain with high belief values and lower entropy. Both results are shown for $K=20$ classes, additional to the 10 singletons. 

\textbf{Qualitative results of entropy.} Fig. \ref{rotated-plot} depicts entropy-based uncertainty estimates for rotated out-of-distribution (OoD) MNIST digit `$3$' samples. All models accurately predict the true class at $-30^\circ$, $0^\circ$, and $30^\circ$. RS-NN and LB-BNN consistently exhibit low entropy in these scenarios, while ENN shows higher entropy for correct classifications at these angles. As the rotation angle increases, indicating more challenging scenarios, all models predict the wrong class. Notably, LB-BNN fails to exhibit a significant increase in entropy for these incorrect predictions. ENN performs relatively better than RS-NN at $60^\circ$ and $90^\circ$ rotations but has previously demonstrated high entropy levels even for accurate predictions at relatively minor rotations of $-30^\circ$ and $+30^\circ$ degrees. Overall, RS-NN provides a more reliable measure of uncertainty.

\begin{figure}[!h]
  \centering
    \includegraphics[width=0.6\linewidth]{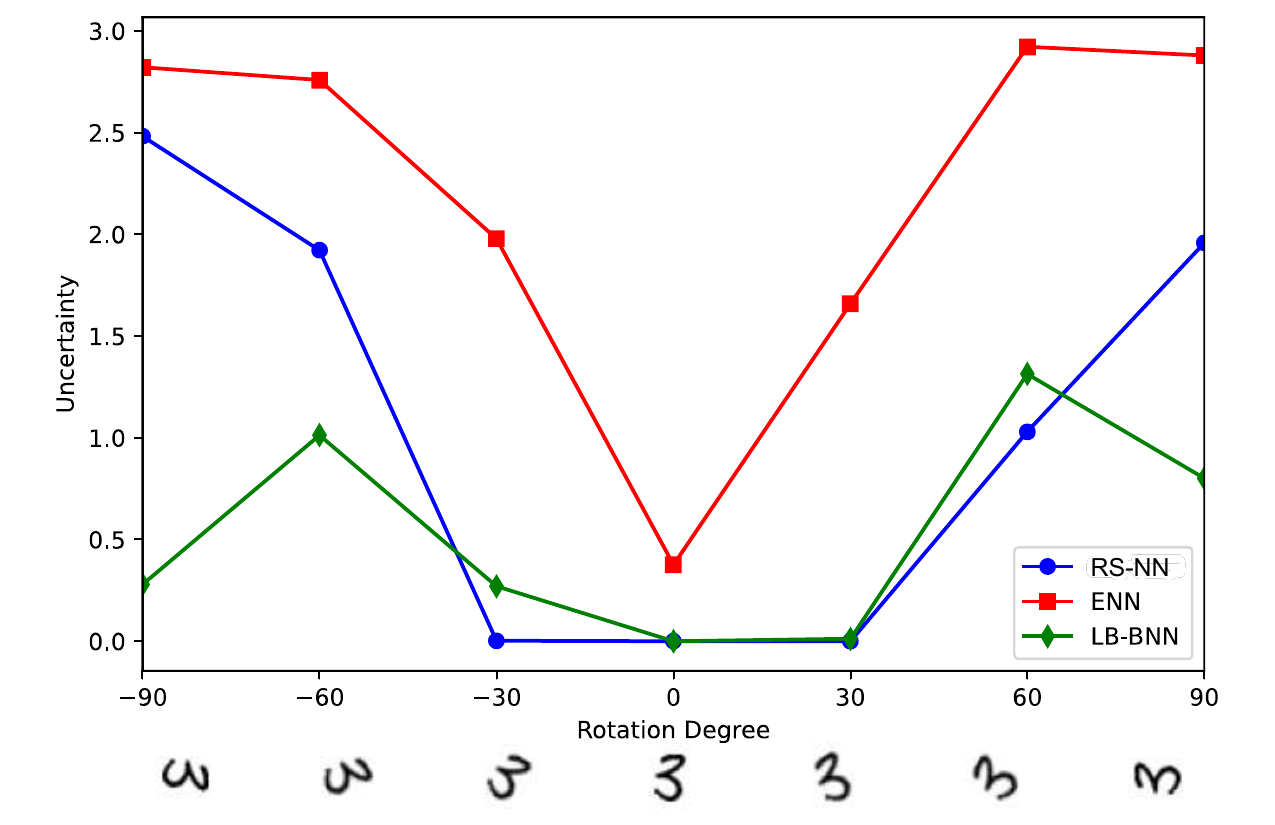}
     \caption{Uncertainty (entropy) of MNIST digit '$3$' rotated between -90 and 90 degrees. The blue line indicates RS-NN estimates low uncertainty between -30 to 30, and high uncertainty for further rotations.}
    \label{rotated-plot}
\end{figure}

\begin{figure}[!h]
\centering
    \begin{minipage}[b]{0.49\textwidth}
        \centering
        \includegraphics[width=\linewidth]{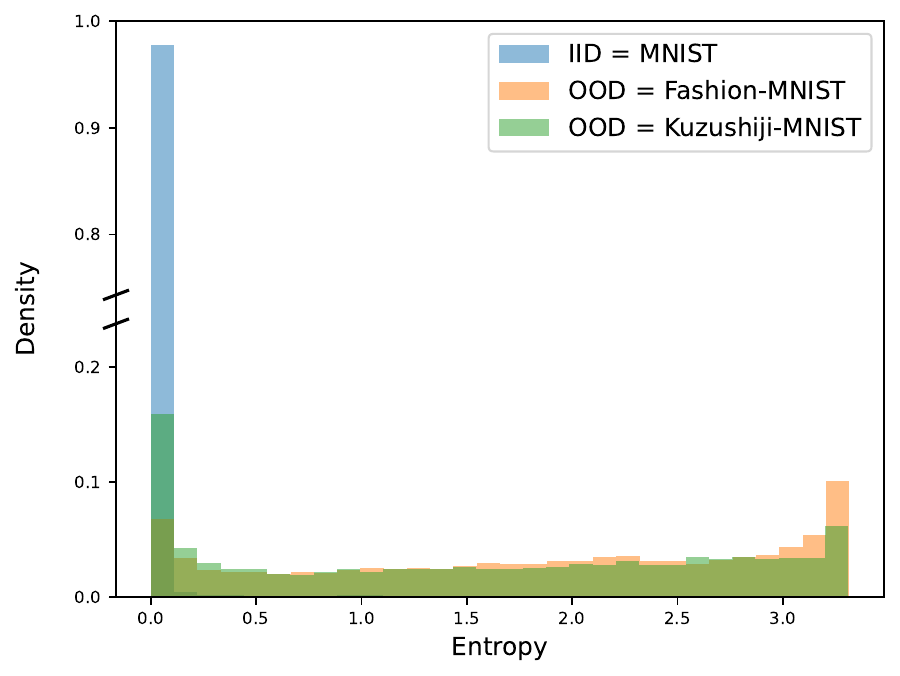}
        \caption{Entropy Distributions for RS-NN on MNIST \textit{vs.} Fashion-MNIST/Kuzushiji-MNIST.}
        \label{OoD-MNIST}
    \end{minipage}
    \hfill
    \begin{minipage}[b]{0.49\textwidth}
        \centering
        \includegraphics[width=\linewidth]{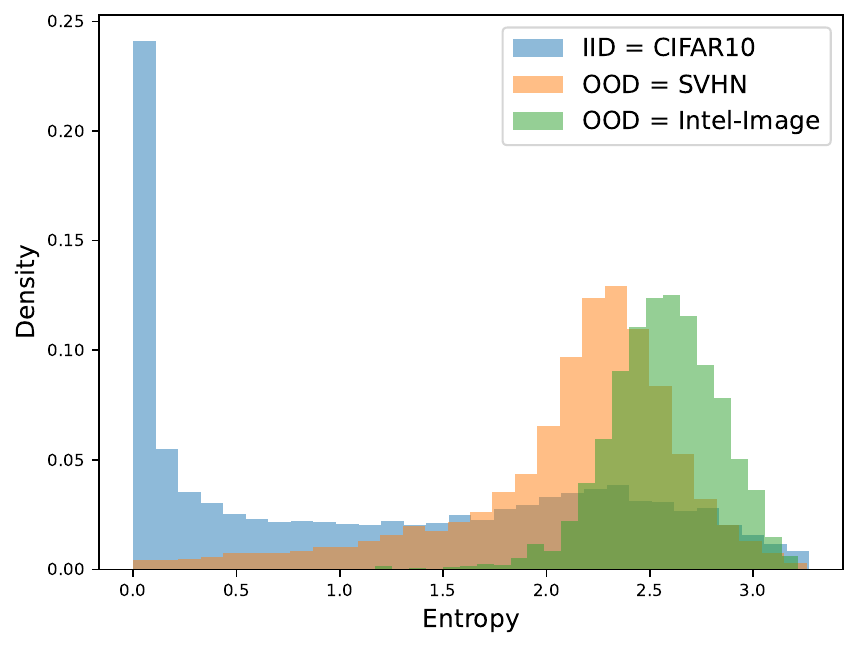}
        \caption{Entropy Distributions for RS-NN on CIFAR-10 \textit{vs.} SVHN/Intel-Image.}
        \label{OoD-CIFAR-10}
    \end{minipage}
    \vspace{-10pt}
\end{figure}

Figs. \ref{OoD-MNIST} and \ref{OoD-CIFAR-10} show the entropy distributions for RS-NN on MNIST (iD) \textit{vs.} Fashion-MNIST (OoD)/Kuzushiji-MNIST (OoD), and CIFAR-10 (iD) \textit{vs.} SVHN (OoD)/Intel-Image (OoD), respectively. There is a clear iD \textit{vs.} OoD shift in entropy for RS-NN, as detailed in Sec. \ref{sec:uncertainty}.
The plots show that RS-NN consistently exhibits larger iD \textit{vs.} OoD entropy ratio for all datasets. For iD data, the entropy values are generally lower, indicating that these models exhibit a higher level of confidence when making predictions within familiar datasets. Conversely, the OoD side shows significantly higher entropy values for most models, particularly notable with the SVHN and Intel datasets, suggesting that the models struggle to generalize when faced with unfamiliar data. 

\subsubsection{Credal set width}  \label{app:credal}

Fig. \ref{fig:credal_set_width} shows a density plot of estimated credal set widths, the difference between the lower and upper probability bounds in Eq. \ref{eq:size-credal}, for correctly and incorrectly classified samples of CIFAR-10 test data. 
Incorrect predictions (red) correlate with larger credal set widths, indicating higher epistemic uncertainty, and are more dispersed in the plot. Correct predictions (blue) concentrate around smaller values, exhibiting smaller credal intervals as expected. 

\begin{figure}[!h]
\centering
\includegraphics[width=0.5\linewidth] {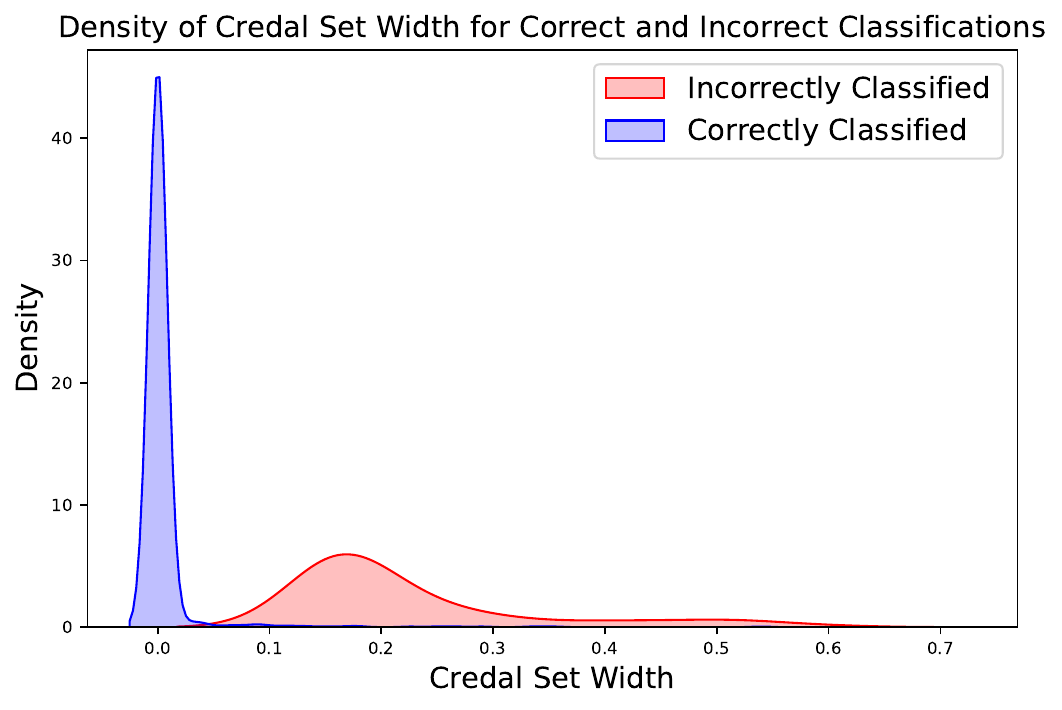}
\vspace{-5pt}
 \caption{Width of credal predictions for CIFAR-10 test data (correctly classified, blue; incorrectly classified, red).
 }
    \label{fig:credal_set_width}
     \vspace{-8pt}
\end{figure}%

Tab. \ref{tab:credal-table} reports 
the credal set widths of RS-NN predictions for iD \textit{vs.} OoD datasets. Larger 
intervals
can be observed for OoD datasets. 
{The credal set widths for ImageNet \textit{vs.} ImageNet-O are not as distinct given the nature of these datasets, as detailed in Sec. {\ref{app:imagenet-o}}.}
Note that credal set width
is not directly comparable to the entropy, variance or mutual information of other models, as such metrics have distinct semantics. 

\begin{table}[!h]
\centering
\caption{Credal set width for RS-NN on iD \textit{vs.} OoD datasets: CIFAR10 \textit{vs.} SVHN/Intel Image, MNIST \textit{vs.} F-MNIST/K-MNIST and ImageNet \textit{vs.} ImageNet-O.}
\label{tab:credal-table}
\vspace{-7pt}
\resizebox{0.7\linewidth}{!}{%
    \begin{tabular}{cccccc}
    \toprule
    \multicolumn{2}{c|}{In-distribution (iD)} & \multicolumn{4}{c}{Out-of-distribution (OoD)} \\
    \midrule
    CIFAR10 & \multicolumn{1}{l|}{$0.007 \pm 0.044$} & \multicolumn{1}{c}{SVHN} & \multicolumn{1}{c}{$0.260 \pm 0.322$} & Intel Image &  $0.587 \pm 0.367$\\
    \midrule
    MNIST & \multicolumn{1}{l|}{ $0.001 \pm 0.013$} & \multicolumn{1}{c}{F-MNIST} & \multicolumn{1}{c}{$0.070 \pm 0.167$} & K-MNIST & $0.103 \pm 0.200$ \\
    \midrule
    ImageNet & \multicolumn{1}{l|}{$0.238 \pm 0.266$} & \multicolumn{2}{c}{ImageNet-O} & $0.272 \pm 0.275$ \\
    \bottomrule
    \end{tabular}%
}
\end{table}

\begin{figure}[!h]
\centering
    \begin{minipage}[t]{0.45\linewidth}
        \centering
        \includegraphics[width=\textwidth]{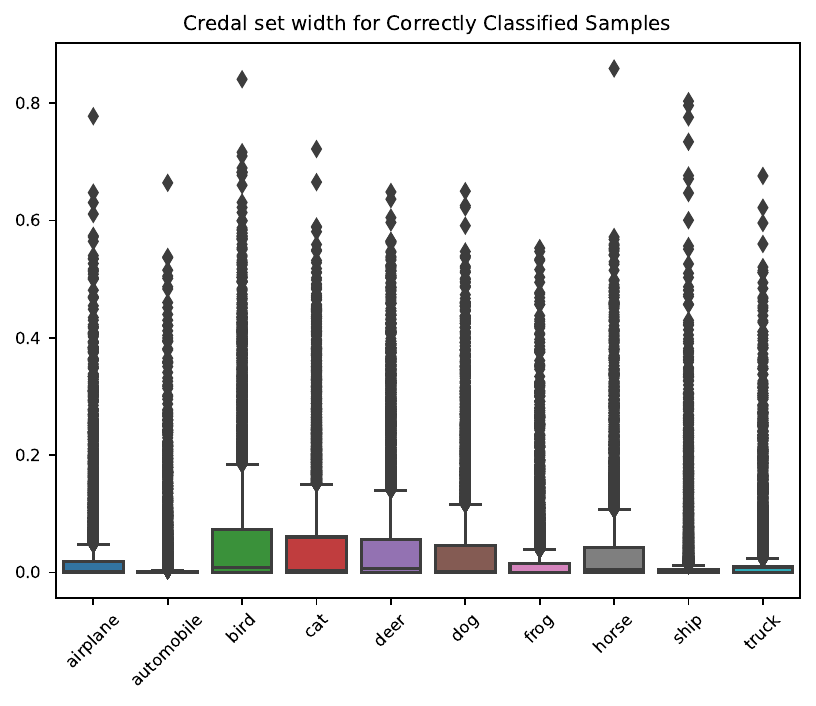}
        \label{credal_correct}
    \end{minipage}%
    \hspace{1em} 
    \begin{minipage}[t]{0.45\linewidth}
        \centering
        \includegraphics[width=\textwidth]{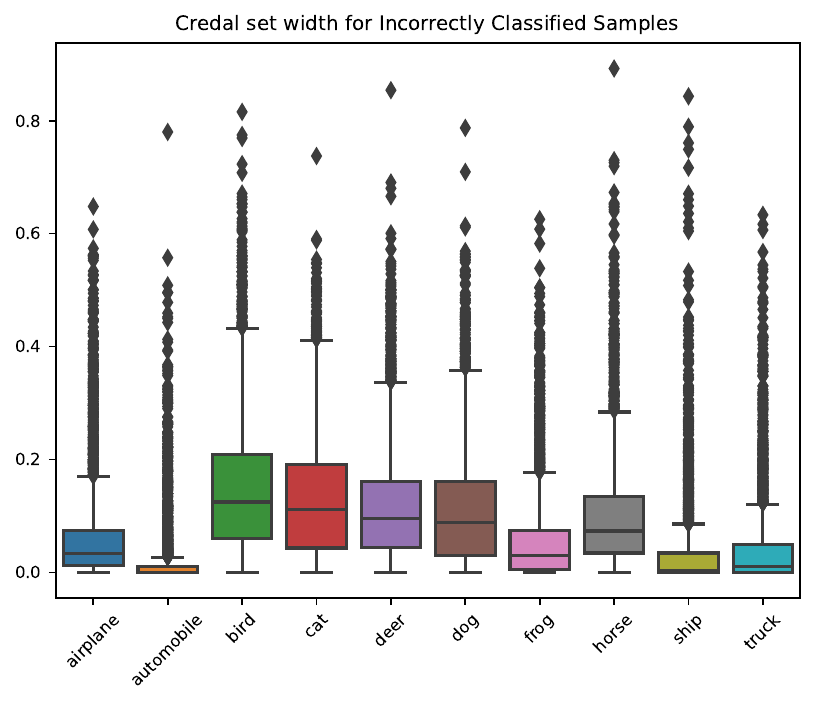}
        \label{credal_incorrect}
    \end{minipage}
    \vspace{-22pt}
\caption{Credal set widths for individual classes of CIFAR-10 dataset. For Correctly Classified samples \textit{(left)}. For Incorrectly Classified samples \textit{(right)}.}
\label{credal_width_incorrect_correct}
    \vspace{-7pt}
\end{figure}

Fig. \ref{credal_width_incorrect_correct} plots the credal set widths for {correctly and incorrectly} classified samples over each class $c$ in CIFAR-10, respectively. The box represents the interquartile range between the 25th and 75th percentiles of the data encompassing most of the data. The vertical line inside the box represents the median and the whiskers extend from the box to the minimum and maximum values within a certain range, and data points beyond this range are considered outliers and are plotted individually as dots or circles. The box plots here are shown for 10,000 samples of CIFAR-10 dataset where most samples within the incorrect classifications have higher credal set widths indicating higher uncertainty in these samples, especially for classes `\texttt{bird}', `\texttt{cat}', `\texttt{deer}', and `\texttt{dog}'.

\subsubsection{Entropy/Credal Set Width \textit{vs.} Confidence score} \label{app:ent-vs-credal}

In this section, we discuss the correlation between uncertainty measures (pignistic entropy, credal set width) and confidence scores.

Fig. \ref{subfig:entropy} shows the relationship between the entropy of RS-NN pignistic predictions and the associated confidence levels for the CIFAR-10 (left) and SVHN/Intel Image (right) datasets.
Fig. \ref{subfig:credal-set} illustrates the relationship between credal set width and confidence for the same datasets.
The distribution of iD predictions (left) for both 
tests
show a concentration at the top left, indicating high confidence and low entropy or credal set width. Conversely, OoD predictions (middle, right) exhibit a more dispersed pattern.

\begin{figure}[!h]
\centering
    \includegraphics[width=1\textwidth]{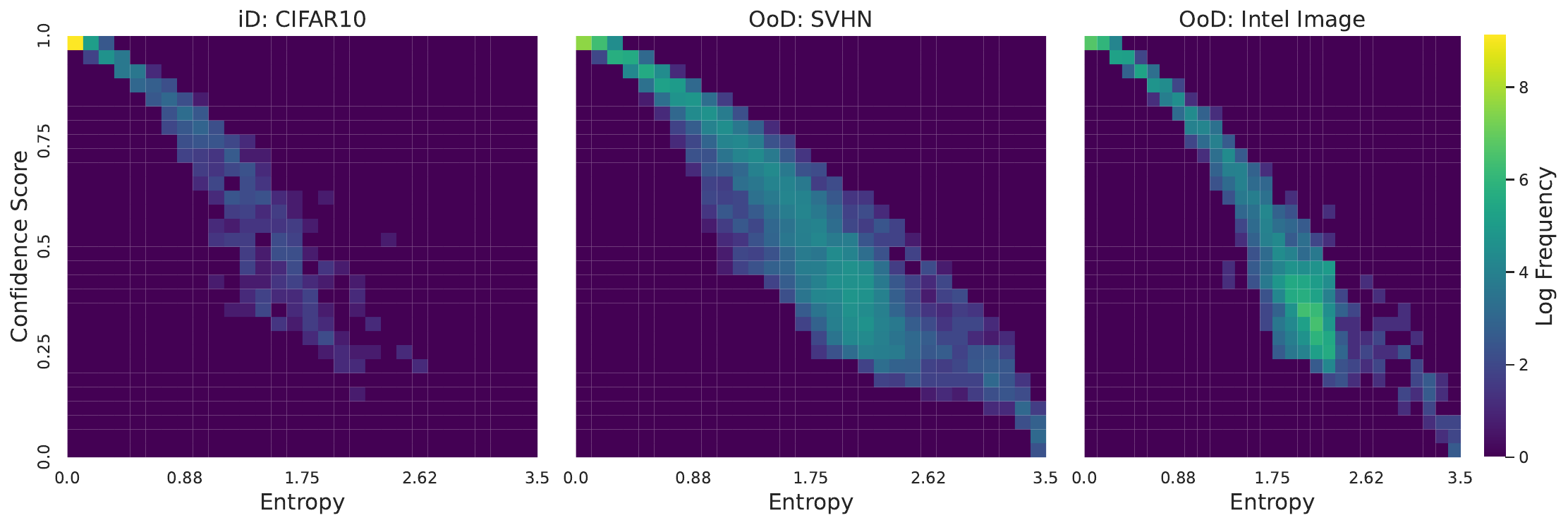}
    \caption{Entropy \textit{vs.} Confidence score on iD (left) \textit{vs.} OoD (right) datasets. For CIFAR-10, most predictions are concentrated top left of the plot indicating lower entropy and higher confidence in the predictions. For SVHN and Intel Image datasets, predictions are more distributed.}
    \label{subfig:entropy}
\end{figure}

\begin{figure}[!h]
\centering
     \vspace{-7pt}
    \includegraphics[width=1\textwidth]{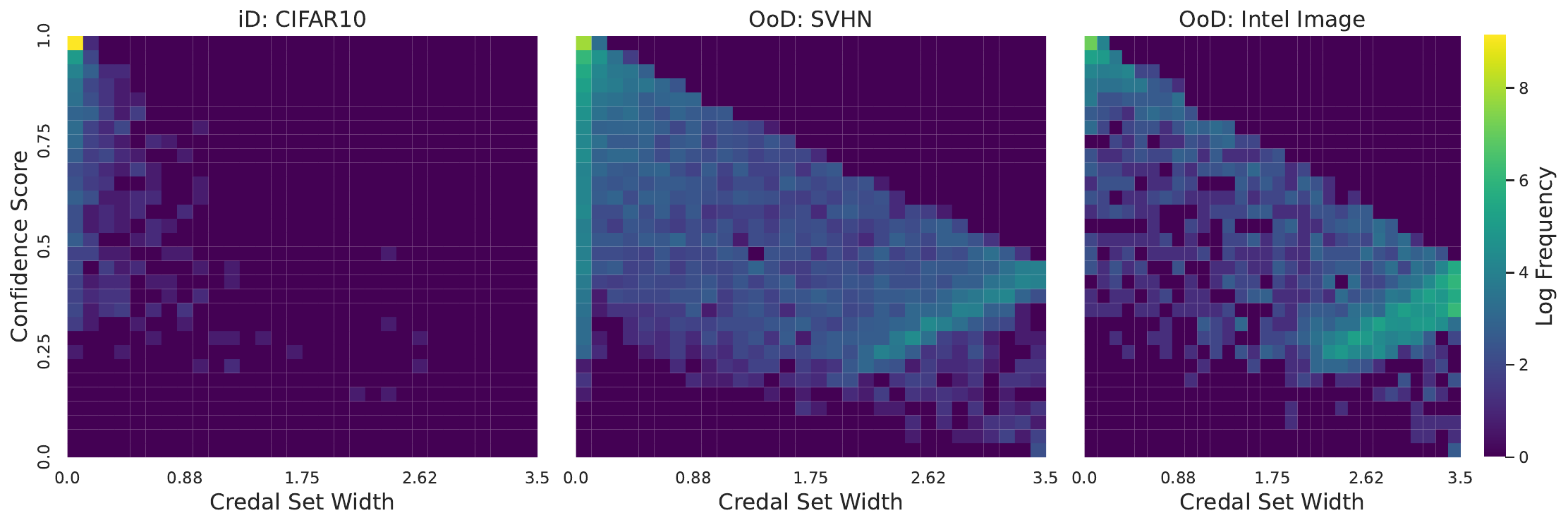}
    \caption{Credal Set Width \textit{vs.} Confidence score on iD (left) \textit{vs.} OoD (right) datasets. For CIFAR-10, confidence scores are high and credal set width is small. For SVHN and Intel Image datasets, credal set width varies for each prediction and is less reliant on confidence score.}
    \label{subfig:credal-set}
     \vspace{-7pt}
\end{figure}

Entropy, reflecting prediction uncertainty, 
is quite correlated with confidence, in both iD and OoD tests.
In contrast, 
as it considers the entire set of plausible outcomes within a belief function rather than a single prediction,
credal set width better quantifies the degree of epistemic uncertainty inherent to a prediction.
As a result, credal set width is less dependent on the concentration of predictions and is more reflective of the overall uncertainty encompassed by the model. 
Fig. \ref{subfig:credal-set} shows that, 
unlike entropy, credal with 
is clearly not correlated with confidence.

\subsection{Scalability to Large-Scale Architectures}\label{sec:scalability}

Tab. \ref{tab:scalability} shows the scalability of RS-NN to larger model architectures and its 
ability to leverage
transfer learning. RS-NN outperforms standard CNNs across various large-scale model architectures, including WideResNet28-10 (WRN-28-10), VGG16, InceptionV3 (IncV3), EfficientNetB2 (ENetB2), and Vision Transformer (ViT-Base-16), highlighting its versatility and ease in adopting different architectures, and the generality of the random-set concept.
Notably, 
using a pre-trained ResNet50 (Pre-trained R50) model with ImageNet weights, RS-NN achieves higher accuracy on CIFAR-10 compared to using only the architecture (no pre-trained weights).

\begin{table}[H]
\centering
\caption{Adaptability 
to large-scale model architectures with test accuracy (\%) and parameters (in million) reported on CIFAR10.}
\label{tab:scalability}
        \vspace{-8pt}
\resizebox{\linewidth}{!}{%
\begin{tabular}{llllllll}
\toprule
& Model & Pre-trained R50 & WRN-28-10 & VGG16 & IncepV3 & ENetB2 & ViT-Base \\ \midrule
\multirow{2}{*}{Test acc. (\%)} & RS-NN & $\mathbf{94.42}$ & $\mathbf{93.58}$ & $\mathbf{87.87}$ & $\mathbf{78.24}$ & $\mathbf{92.10}$ & $86.75$\\
& CNN    & $94.38$ & $92.79$ & $84.14$  & $76.89$ & $90.02$ & $\mathbf{87.21}$ \\ \midrule
\multirow{2}{*}{Params (M)} &
RS-NN & $2.69$  & $37.0$  & $15.12$  &  $31.22$ & $7.72$ & $9.53$ \\
& CNN    & $2.62$  & $36.7$ & $15.11$  & $31.21$ & $7.71$ & $9.52$ \\
\bottomrule
\end{tabular}%
}
\end{table}

\subsection{Overcoming the Overconfidence Problem in CNNs}\label{app:cnn-overconfidence}

\textbf{Accuracy.} RS-NN consistently achieves higher accuracy than standard CNN (ResNet50) on all datasets as shown in Tab. \ref{tab:accuracy-table}. 
This appears to attest that,
as RS-NN uses a random-set framework which does not require prior assumptions
and is more data driven,
it 
can better
adapt and capture the inherent complexity of the data (as sets), contributing to its superior performance in test accuracy across various datasets 
(Tab. \ref{tab:accuracy-table}).
The training hyperparameters for RS-NN and CNN are the same, since the training procedure is quite similar. 

\begin{table}[!h]
  \label{uncertainty-table}
  \setlength{\tabcolsep}{8pt}
  \centering
  \caption{CNN fails to perform well when tested on an noisy (rotated) sample of MNIST test data. The predicted results show how a standard CNN predicts the wrong class with a high confidence score ($99.95\%$), while the RS-NN model predicts the correct class with $49.7\%$ confidence and a high entropy of $2.1626$.}
  \vspace{-8pt}
\label{tab:rotated-8}
  \resizebox{1\linewidth}{!}{
  \begin{tabular}{lll}
    \toprule
     \textbf{True Label} =`8' & 
     \textbf{CNN Predictions} & 
     \textbf{RS-NN Predictions}\\
    \midrule
     \multirow{7}{2cm}[8ex]{\raisebox{-\totalheight}{\includegraphics[width=0.15\textwidth]{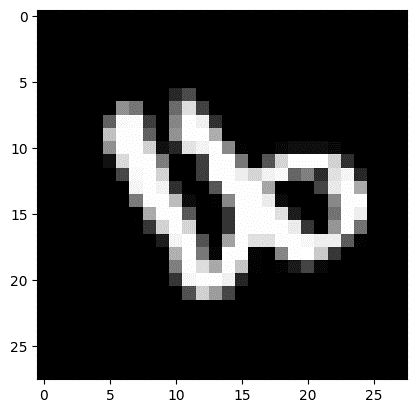}}} &
    \begin{tabular}{ll}
      \textbf{Class 6} & \textbf{0.9995} \\
      Class 5 & 0.0002 \\
      Class 8 & 0.0001 \\
      Class 0 & 1.4e-05 \\
      Class 4 & 1.1e-06
    \end{tabular} & 
    \begin{tabular}{lll}
     \begin{tabular}{ll}
  \multicolumn{2}{c}{Belief values} \\
  \{'6', '4', '8'\} & 0.7853 \\
  \{'6', '8', '1'\} & 0.7150 \\
  \{'6', '8'\} & 0.6357 \\
  \{'9', '8'\} & 0.5529 \\
  \{'8', '3'\} & 0.4092 \\
    \end{tabular} & 
    \begin{tabular}{ll}
  \multicolumn{2}{c}{Mass values} \\
  \{'6', '8'\} & 0.2079 \\
  \{'9', '8', '1'\} & 0.1999 \\
  \{'8'\} & 0.1959 \\
  \{'8', '3'\} & 0.0961 \\
  \{'6'\} & 0.05147 \\
      \end{tabular} &
        \begin{tabular}{ll}
        \multicolumn{2}{c}{Pignistic Probability} \\
        \textbf{8} & \textbf{0.4978} \\
        6 & 0.2294 \\
        9 & 0.1002  \\
        3 & 0.0502 \\
        4 & 0.0459  \\
        \end{tabular} 
    \end{tabular} \\
        \addlinespace
        & 
        \begin{tabular}{l}
        \textbf{Entropy} 0.0061
       \end{tabular} &
        \begin{tabular}{ll}
        \textbf{Entropy} 2.1626 & \textbf{Credal set width} 0.582
    \end{tabular} \\
    \addlinespace
    \bottomrule
  \end{tabular}
  }
\end{table}

\begin{figure}[!h]
      \centering
        \includegraphics[width=0.6\linewidth]{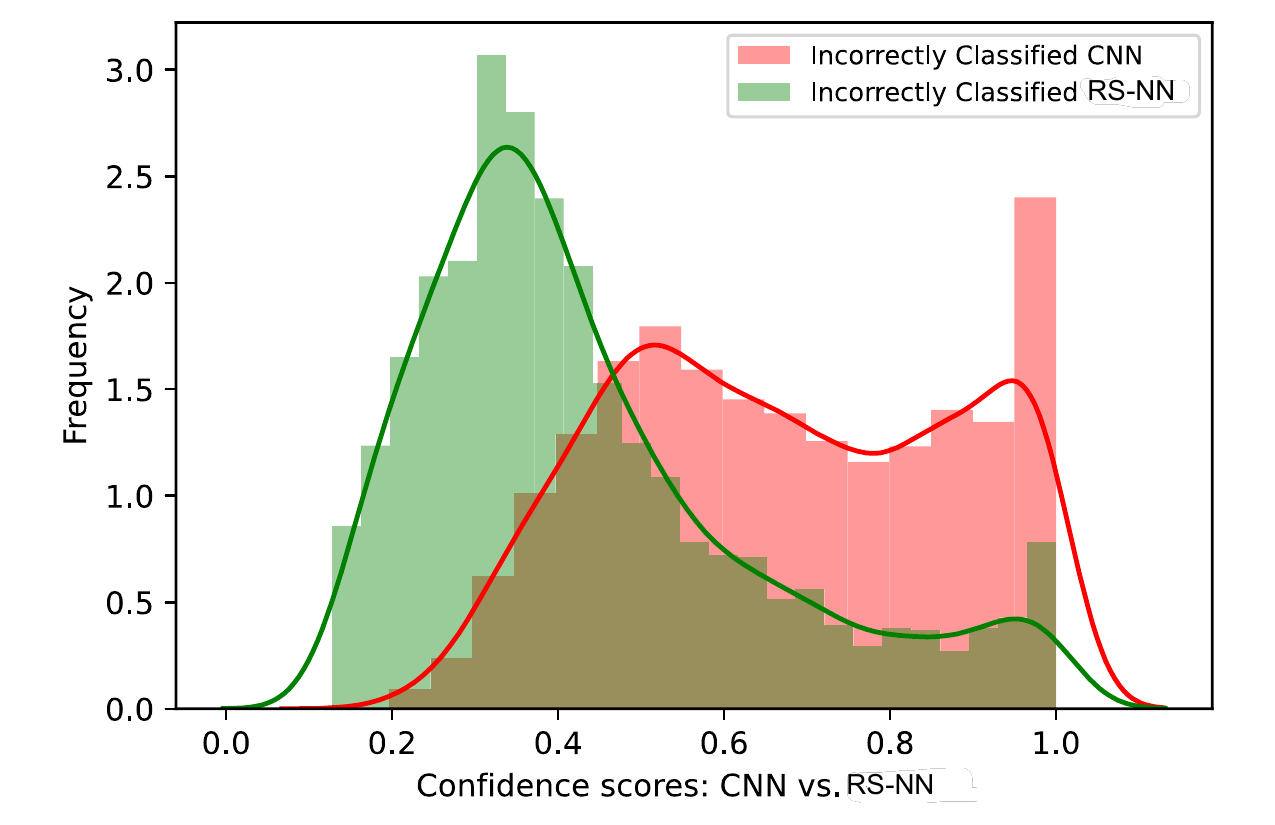}
        \vspace{-4pt}
        \caption{Confidence scores for \textit{Incorrectly Classified samples} of RS-NN and CNN}
        \label{icc}
\end{figure} 
\textbf{Confidence.} We further analyse how RS-NN overcomes the overconfidence problem in CNNs.
Tab. \ref{tab:rotated-8} shows an noisy example of a rotated MNIST image with a true label `$8$'. The standard CNN makes an incorrect prediction (`$6$') with $99.95\%$ confidence and a low entropy of $0.0061$, highlighting a key limitation of relying solely on confidence scores and the entropy of predicted probabilities.
RS-NN provides a correct prediction (class `$8$') with a confidence of $49.7\%$ and pignistic entropy of $2.1626$ and a credal set width of $0.482$. Crucially, the RS-NN's prediction process considers more than just the predicted class, taking into account both the confidence of the pignistic probability and the entropy. In this example, despite predicting the correct class (`8'), the RS-NN demonstrates a low confidence (49.7\%) and high entropy (2.1626). This holistic approach provides a more nuanced and reliable understanding of the model's uncertainty.

In Fig. \ref{icc}, we show how a standard CNN has high confidence for incorrectly classified samples whereas RS-NN exhibits lower confidence for incorrectly classified samples.

\subsection{Robustness to noisy and rotated data} \label{app:noisy}

We split the MNIST data into training and test sets, train the RS-NN model using the training test, and test the model on noisy and rotated out-of-distribution test data. This is done by adding random noise to the test set of images to obtain noisy data and rotating MNIST images with random degrees of rotation between 0 and 360. 

\begin{table}[!ht]
  \small
  \setlength{\tabcolsep}{3pt}
  \centering
      \caption{CNN fails to perform well when tested on noisy and rotated samples of MNIST test data. The predicted results show how a standard CNN predicts the wrong class with a high confidence score, whereas the RS-NN model predicts the right class with varying confidence scores.
    }
            \vspace{-8pt}
      \label{OoD-table}
  \begin{tabular}{lll}
    \toprule
      & \textbf{CNN Predictions} & \textbf{Belief RS-NN Predictions}\\
    \midrule
    \textbf{True Label = 2} & & \\
     \multirow{8}{1cm}[10ex]{\raisebox{-\totalheight}{\includegraphics[width=0.14\textwidth]{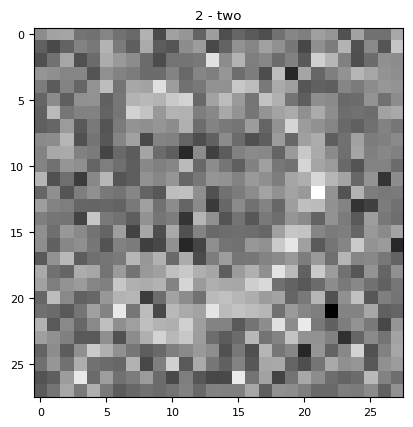}}} &
    \begin{tabular}{ll}
      \textbf{Class 0} & \textbf{0.628} \\
      Class 2 & 0.232 \\
      Class 3 & 0.117 \\
      Class 8 & 0.015 \\
    \end{tabular} & 
    \begin{tabular}{lll}
     \begin{tabular}{ll}
    \multicolumn{2}{c}{\textbf{Belief values}} \\
      \{`\texttt{2}'\} & 0.985 \\
      \{`\texttt{2}', `\texttt{0}', `\texttt{1}'\} & 0.981 \\
      \{`\texttt{2}', `\texttt{1}'\} & 0.980 \\
      \{`\texttt{2}', `\texttt{4}'\} & 0.979 \\
    \end{tabular} & 
    \begin{tabular}{ll}
    \multicolumn{2}{c}{\textbf{Mass values}} \\
      \{`\texttt{2}'\} & 0.982 \\
      \{`\texttt{0}', `\texttt{8}'\} & 0.009 \\
      \{`\texttt{7}', `\texttt{0}', `\texttt{1}'\} & 0.002 \\
      \{`\texttt{7}', `\texttt{8}'\} & 0.002 \\
      \end{tabular} &
        \begin{tabular}{ll}
        \multicolumn{2}{c}{\textbf{Pignistic}} \\
        \textbf{2} & \textbf{0.982} \\
        8 & 0.007 \\
        0 & 0.100  \\
        3 & 0.004\\
        \end{tabular} 
    \end{tabular} \\
    \addlinespace
        \midrule
     \textbf{True Label = 3 } &  & \\
     \multirow{8}{1cm}[10ex]{\raisebox{-\totalheight}{\includegraphics[width=0.14\textwidth]{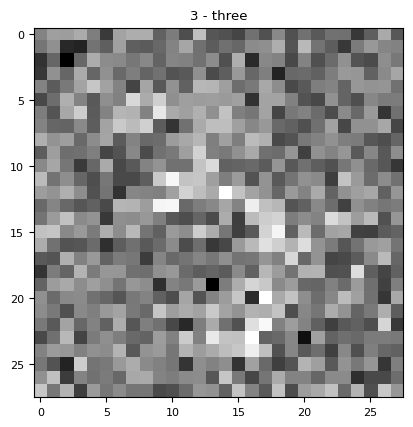}}} &
    \begin{tabular}{ll}
      \textbf{Class 8} & \textbf{0.969} \\
      Class 5 & 0.018 \\
      Class 9 & 0.006 \\
      Class 2 & 0.003 \\
    \end{tabular} & 
    \begin{tabular}{lll}
     \begin{tabular}{ll}
    \multicolumn{2}{c}{\textbf{Belief values}} \\
      \{`\texttt{3}', `\texttt{5}'\} & 0.772 \\
      \{`\texttt{6}', `\texttt{3}'\} & 0.637 \\
      \{`\texttt{6}', `\texttt{3}', `\texttt{5}'\} & 0.620 \\
      \{`\texttt{3}'\} & 0.540 
    \end{tabular} & 
    \begin{tabular}{ll}
    \multicolumn{2}{c}{\textbf{Mass values}} \\
      \{`\texttt{3}'\} & 0.361 \\
      \{`\texttt{7}', `\texttt{8}'\} & 0.134 \\
      \{`\texttt{0}', `\texttt{8}'\} & 0.104 \\
      \{`\texttt{8}'\} & 0.084 
      \end{tabular} &
        \begin{tabular}{ll}
        \multicolumn{2}{c}{\textbf{Pignistic}} \\
        \textbf{3} & \textbf{0.427} \\
        8 & 0.205 \\
        5 & 0.115  \\
        7 & 0.080
        \end{tabular} 
    \end{tabular} \\
    \addlinespace
        \midrule
     \textbf{True Label = 1 } &  & \\
     \multirow{8}{1cm}[10ex]{\raisebox{-\totalheight}{\includegraphics[width=0.14\textwidth]{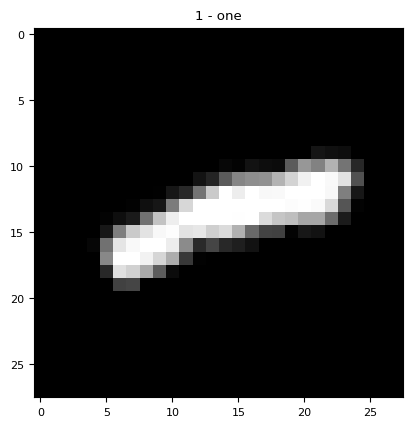}}} &
    \begin{tabular}{ll}
      \textbf{Class 2} & \textbf{1.0} \\
      Class 1 & 3.39e-08 \\
      Class 6 & 1.03e-10 \\
      Class 3 & 4.92e-11 \\
    \end{tabular} & 
    \begin{tabular}{lll}
     \begin{tabular}{ll}
    \multicolumn{2}{c}{\textbf{Belief values}} \\
      \{`\texttt{2}', `\texttt{1}'\} & 0.999 \\
      \{`\texttt{1}', `\texttt{9}'\} & 0.958 \\
      \{`\texttt{1}'\} & 0.924 \\
      \{`\texttt{1}', `\texttt{5}'\} & 0.825 \\
    \end{tabular} & 
    \begin{tabular}{ll}
    \multicolumn{2}{c}{\textbf{Mass values}} \\
      \{`\texttt{1}'\} & 0.556 \\
      \{`\texttt{2}'\} & 0.423 \\
      \{`\texttt{1}', `\texttt{9}'\} & 0.020 \\
      \{`\texttt{6}'\} & 4.31e-05 \\
      \end{tabular} &
        \begin{tabular}{ll}
        \multicolumn{2}{c}{\textbf{Pignistic}} \\
        \textbf{1} & \textbf{0.566} \\
        2 & 0.423 \\
        9 & 0.010  \\
        6 & 4.31e-05\\
        \end{tabular} 
    \end{tabular} \\
    \addlinespace
    \midrule
     \textbf{True Label = 9 } &  & \\
     \multirow{8}{1cm}[10ex]{\raisebox{-\totalheight}{\includegraphics[width=0.14\textwidth]{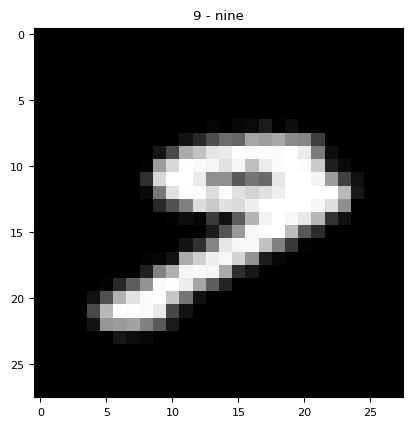}}} &
    \begin{tabular}{ll}
      \textbf{Class 5} & \textbf{0.988} \\
      Class 2 & 0.010 \\
      Class 3 & 0.0004 \\
      Class 7 & 0.0002 \\
    \end{tabular} & 
    \begin{tabular}{lll}
     \begin{tabular}{ll}
    \multicolumn{2}{c}{\textbf{Belief values}} \\
      \{`\texttt{7}', `\texttt{5}', `\texttt{9}'\} & 0.831 \\
      \{`\texttt{7}', `\texttt{9}'\} & 0.706 \\
      \{`\texttt{5}', `\texttt{9}'\} & 0.576 \\
      \{`\texttt{3}', `\texttt{9}'\} & 0.558 \\
    \end{tabular} & 
    \begin{tabular}{ll}
    \multicolumn{2}{c}{\textbf{Mass values}} \\
      \{`\texttt{7}', `\texttt{9}'\} & 0.171 \\
      \{`\texttt{3}'\} & 0.134 \\
      \{`\texttt{9}'\} & 0.132 \\
      \{`\texttt{7}', `\texttt{8}'\} & 0.102 \\
      \end{tabular} &
        \begin{tabular}{ll}
        \multicolumn{2}{c}{\textbf{Pignistic}} \\
        \textbf{9} & \textbf{0.699} \\
        7 & 0.023 \\
        5 & 0.004  \\
        3 & 0.001\\
        \end{tabular} 
    \end{tabular} \\ 
        \addlinespace
    \bottomrule
  \end{tabular}
\end{table}

Tab. \ref{OoD-table} shows predictions for noisy and rotated noisy MNIST samples. In cases where a standard CNN makes wrong predictions with high confidence scores, Random-Set NN manages to predict the correct class with varying confidences verifying that the model is not overcofident in uncertain cases.  For example, a noisy sample with true class `\texttt{3}' has a standard CNN prediction of class `\texttt{8}' with $96.9\%$ confidence, while RS-NN predicts the correct class \{`\texttt{3}'\} with $42.7\%$ confidence. Similarly, for rotated `\texttt{9}', the standard CNN predicts class `\texttt{5}' with $98.8\%$ confidence whereas RS-NN predicts the correct class \{`\texttt{9}'\} with $69.9\%$ confidence. For a rotated `\texttt{1}', the standard CNN predicts class `\texttt{2}' with $100\%$ confidence. Tab. \ref{noisy-table} shows the test accuracies for standard CNN, RS-NN, LB-BNN and ENN at different scales of random noise. RS-NN shows significantly higher test accuracy than other models as the amount of noise added to the test data increases.

\begin{table}[!h]
    \centering
    \caption{Test accuracies(\%) for RS-NN, standard CNN, LB-BNN (Bayesian) and ENN (Ensemble) on noisy samples of MNIST . `Scale' represents the standard deviation of the normal distribution from which random numbers are being generated for random noise. \label{noisy-table}}
    \vspace{-8pt}
    \resizebox{0.9\linewidth}{!}{%
    \begin{tabular}{llllllll}
     \toprule
        Noise (scale) & 0.2& 0.3 & 0.4 & 0.5 & 0.6 & 0.7 & 0.8 \\
        \midrule
        CNN & 93.40\% & 79.08\% & 79.15\% & 58.33\% & 40.19\% & 28.15\% & 28.59\% \\
        RS-NN & 96.90\% & 85.36\% & \textbf{85.91} \% & \textbf{68.33} \% & \textbf{51.46} \% & \textbf{37.18} \% & \textbf{38.05} \% \\
        LB-BNN & \textbf{98.47}\% & \textbf{95.26} \% & 80.18\% &  61.56\% & 43.28\% & 31.41\% & 24.55 \\
        ENN &97.81\% & 90.76\% & 75.41\% & 58.99\% & 45.70\% & 36.97\% & 31.51 \\
        \bottomrule
    \end{tabular}
    }
\end{table}

\begin{table}[!h]
    \centering
    \small
    \caption{Test accuracies(\%) for RS-NN, standard CNN, LB-BNN (Bayesian) and ENN (Ensemble) on Rotated MNIST out-of-distribution (OoD) samples. Rotation angle is random between the values given.}
        \vspace{-8pt}
    \label{rotation-table}
    \begin{tabular}{llllllll}
        \toprule
        Rotation (angle) & -180/-120 & -120/-60 & -60/0 & 0/60 & 60/120 & 120/180 & 0/360  \\
        \midrule
        CNN & 36.41\% & 21.86\% & 74.41\% & 80.80\% & 23.89\% & 37.53\% & 45.86\%  \\
        RS-NN & \textbf{37.84}\% & \textbf{23.54}\% & \textbf{78.44}\% & \textbf{81.46}\% & \textbf{26.31}\% & \textbf{38.48}\% & \textbf{47.71}\%  \\
        LB-BNN & 37.56\% & 20.19\% & 75.12\% &  77.67\% & 23.18\% & 37.83\% & 46.07 \\
        ENN &36.40\% & 19.43\% & 71.70\% & 78.59\% & 18.70\% & 36.38\% & 44.14 \\
        \bottomrule
    \end{tabular}
\end{table}

\begin{figure}[!h]
    \centering
        \includegraphics[width=\textwidth]{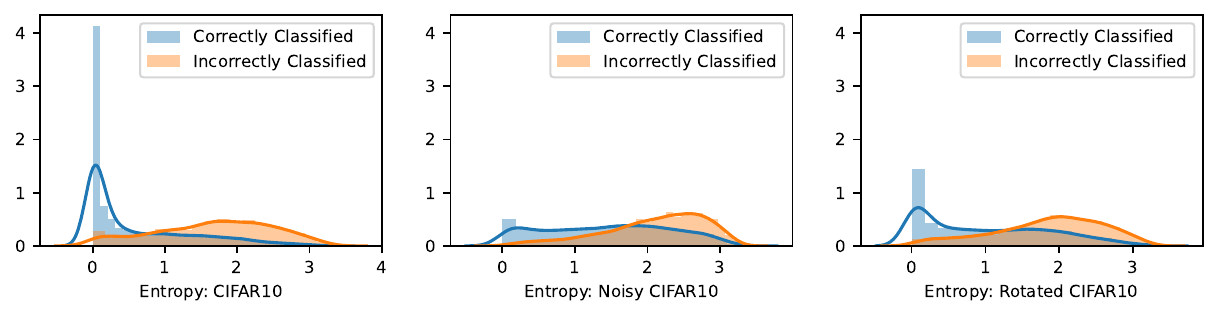}
        \vspace{-20pt}
        \caption{Entropy distribution of RS-NN on CIFAR-10, Noisy CIFAR-10, and Rotated CIFAR-10}
        \label{entropy-dist}
\end{figure}
\begin{figure}[!h]
    \centering
        \includegraphics[width=\linewidth]{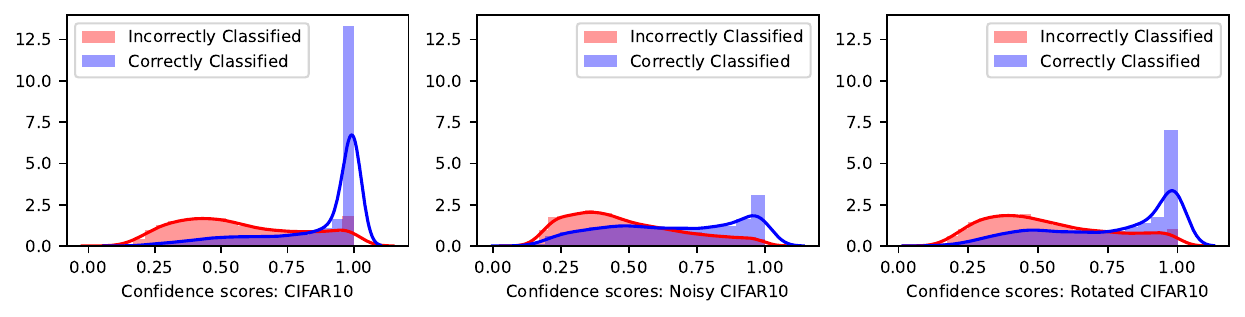}
        \vspace{-20pt}
        \caption{Confidence scores of RS-NN on CIFAR-10, Noisy CIFAR-10, and Rotated CIFAR-10 
        }
        \label{confidence-belief}
\end{figure}

The test accuracies for standard CNN, RS-NN, LB-BNN and ENN on Rotated MNIST images are shown in Tab. \ref{rotation-table}. The samples are randomly rotated every 60 degrees, -180$^{\circ}$ to -120$^{\circ}$, -120$^{\circ}$ to -60$^{\circ}$, -60$^{\circ}$ to 0$^{\circ}$, \textit{etc.} A fully random rotation between 0$^{\circ}$ and 360$^{\circ}$ also shows higher test accuracy for RS-NN at $47.71\%$ when compared to standard CNN with test accuracy $45.86\%$.

Fig. \ref{entropy-dist} displays the entropy distribution for correctly and incorrectly classified samples of CIFAR-10, including in-distribution, noisy, and rotated images. Higher entropy is observed for incorrectly classified predictions across all three cases. A distribution of confidence scores for RS-NN on the same noisy and rotated experiments are shown in Fig. \ref{confidence-belief} for both incorrectly classified and correctly classified samples.  

\subsection{Robustness to Adversarial Attacks}\label{app:fgsm}

We conducted experiments on adversarial attacks using the well-known Fast Gradient Sign Method (FGSM) \cite{goodfellow2014explaining}. FGSM is a popular approach for generating adversarial examples by perturbing input images based on the sign of the gradient of the loss function with respect to the input. Mathematically, the perturbed image $x'$ is computed as 
\begin{equation}
    x' = x + \epsilon \cdot \text{sign}(\nabla_x J(\theta, x, y)),
\end{equation}

where $x$ is the original input image, $\epsilon$ (epsilon) is a small scalar representing the magnitude of the perturbation, $J$ is the loss function, $\theta$ are the model parameters, and $y$ is the true label of the input image.

\begin{figure}[!ht]
  \centering  
  \begin{minipage}[b]{0.3\textwidth}
    \centering
    \includegraphics[width=\linewidth]{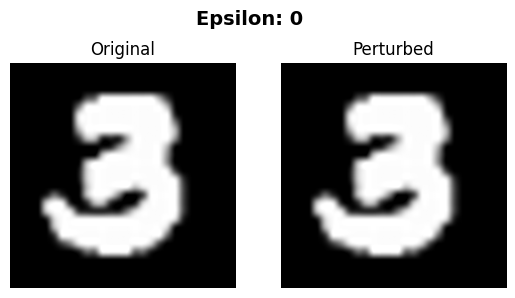}
    \label{fig:fig1}
  \end{minipage}
  \hfill
  \begin{minipage}[b]{0.3\textwidth}
    \centering
    \includegraphics[width=\linewidth]{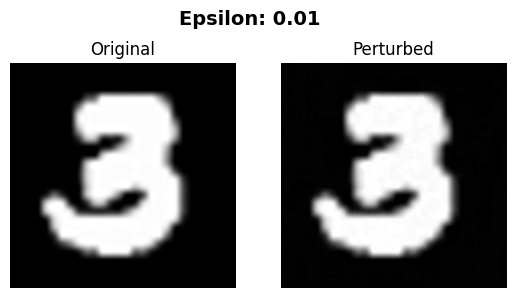}
    \label{fig:fig3}
  \end{minipage}
\hfill
  \begin{minipage}[b]{0.3\textwidth}
    \centering
    \includegraphics[width=\linewidth]{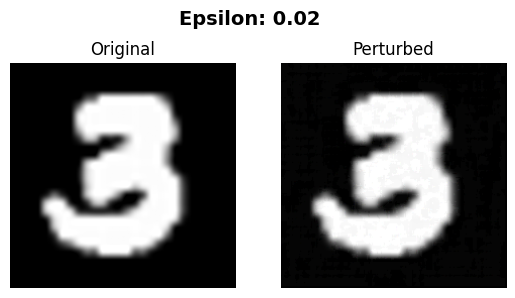}
    \label{fig:fig4}
  \end{minipage}
\\  
  \begin{minipage}[b]{0.3\textwidth}
    \centering
    \includegraphics[width=\linewidth]{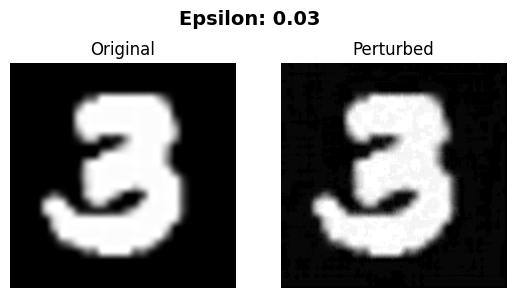}
    \label{fig:fig5}
  \end{minipage}
  \hfill
  \begin{minipage}[b]{0.3\textwidth}
    \centering
    \includegraphics[width=\linewidth]{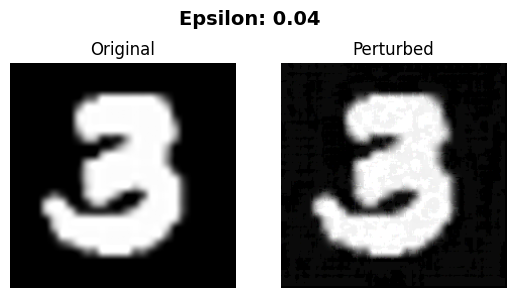}
    \label{fig:fig6}
    \end{minipage}
      \hfill
  \begin{minipage}[b]{0.3\textwidth}
    \centering
    \includegraphics[width=\linewidth]{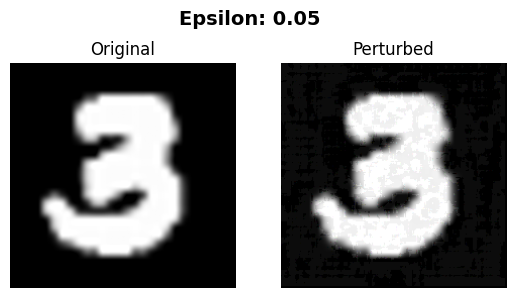}
    \label{fig:fig7}
  \end{minipage}
  \vspace{-10pt}
    \caption{Examples of perturbed images generated using FGSM adversarial attacks on the MNIST dataset for different epsilon values. Epsilon values range from 0 to 0.05.}
  \label{fig:fgsm}
\end{figure}

{In our experiments, we applied FGSM adversarial attacks on the MNIST dataset for standard CNN, Random-Set NN (RS-NN), LB-BNN (Bayesian) and ENN (Ensemble) models. This involved perturbing images from the MNIST dataset with noise generated using FGSM based on the gradient of the loss function (namely, the cross entropy loss for CNN and the $\mathcal{L}_{B-RS}$ loss, Sec. \ref{sec:loss}, Eq. (\ref{final_loss}), for RS-NN). }

Fig. \ref{fig:fgsm} shows examples of perturbed images generated using FSGM adversarial attacks applied to the MNIST dataset for various epsilon values ranging from 0 to 0.05.

\begin{table}[!htbp]
\centering
\caption{Test accuracies of CNN, RS-NN, LB-BNN and ENN models under FGSM adversarial attacks on MNIST and CIFAR-10 dataset for different epsilon values. Epsilon values range from 0 to 0.05.}
        \vspace{-8pt}
\label{tab:fgsm}
\resizebox{0.9\linewidth}{!}{%
\begin{tabular}{llllllllll}
\toprule
& \multirow{2}{*}{Dataset} & \multirow{2}{*}{Model} & \multicolumn{7}{c}{Epsilon ($\epsilon$)} \\ 
\cmidrule{4-10}
& & & 0 & 0.005 & 0.01 & 0.02 & 0.03 & 0.04 & 0.05 \\
\midrule
\multirow{8}{*}{Test acc. (\%)} & \multirow{4}{*}{MNIST} & CNN  & 99.12 & 98.25  & 94.82  & 72.62  &  49.75 & 38.08  & 32.43  \\
& & RS-NN & \textbf{99.71}
 & 98.46
  & 95.84
 & 91.90
 & \textbf{90.62}  
& \textbf{90.10} 
&  \textbf{89.72}      \\
& & LB-BNN & 99.58 & \textbf{99.07} & \textbf{98.64} & \textbf{97.15} & 80.51 & 47.65 & 34.25\\
& & ENN & 99.07 & 98.56 & 92.98 & 51.51 & 35.04 & 27.03 & 9.47
\\
\cmidrule{2-10}
& \multirow{4}{*}{CIFAR-10} & CNN  & 92.08 & 41.54 & 20.88 & 9.18 & 5.81 & 4.43 & 4.05 \\
& & RS-NN & \textbf{93.53} & \textbf{63.42} & \textbf{61.73} & \textbf{61.67} & \textbf{61.53} & \textbf{61.12} & \textbf{60.35} \\
& & LB-BNN & 89.95 & 23.95 & 10.54 & 6.17 & 5.96 & 6.26 & 6.47\\
& & ENN & 91.55 & 62.29  & 42.64  & 24.12 & 16.39 & 12.78 & 11.20\\
\bottomrule
\end{tabular} }
\end{table}

Subsequently, we evaluated the models' predictions on these perturbed images and computed their test accuracies. 
Tab. \ref{tab:fgsm} presents the test accuracies of CNN, RS-NN, LB-BNN and ENN models under FGSM adversarial attacks on the MNIST dataset for different values of $\epsilon$. 
For CIFAR-10, RS-NN demonstrates higher robustness to the attack compared to all other models with increased perturbations, while for MNIST, LB-BNN has higher accuracy at lower $\epsilon$ values but drops significantly for larger values. RS-NN circumvents the FGSM attack and shows consistent accuracy values for higher $\epsilon$. This is because FGSM for a given input image is dependent on the gradient of the loss, whereas RS-NN makes precise correct predictions with loss/gradient zero and is very confident about these predictions. 
{As a result, despite perturbations to the input image, RS-NN consistently classifies it correctly across all tested values of $\epsilon$.
}

\begin{table}[!h]
\centering
\caption{Test accuracies of CNN, RS-NN, LB-BNN and ENN models under PGD adversarial attacks on MNIST and CIFAR-10 datasets with $\alpha$ = 1/255 and number of iterations = 10. Epsilon values range from 0 to 0.05}
\vspace{-8pt}
\label{tab:pgd}
\resizebox{0.9\linewidth}{!}{%
\begin{tabular}{llllllllll}
\toprule
& \multirow{2}{*}{Dataset} & \multirow{2}{*}{Model} & \multicolumn{7}{c}{Epsilon ($\epsilon$)} \\ 
\cmidrule{4-10}
& & & 0 & 0.005 & 0.01 & 0.02 & 0.03 & 0.04 & 0.05 \\
\midrule
\multirow{8}{*}{Test acc. (\%)} & \multirow{4}{*}{MNIST}  & CNN  & 99.12 & 98.90 & 55.44 & 14.03 & 8.11 & 6.58 & 6.58 \\
& & RS-NN &  \textbf{99.71} & 99.13 & 93.39 & \textbf{91.06} & \textbf{89.25} & \textbf{89.07} & \textbf{89.02} \\
& & LB-BNN & 99.58 & \textbf{99.35} & \textbf{99.01} & 33.67 & 16.35 & 13.64 & 13.64\\
& & ENN & 99.07 & 98.43 & 78.09 & 20.64 & 9.94 & 3.78 & 0.93\\
\cmidrule{2-10}
& \multirow{4}{*}{CIFAR-10} & CNN  &  92.08 & 23.82 & 2.47 & 0.04 & 0 & 0 & 0\\
& & RS-NN &  \textbf{93.53} & \textbf{60.23} & \textbf{59.98} & \textbf{59.97} & \textbf{59.97} & \textbf{59.97} & \textbf{59.97} \\
& & LB-BNN & 89.95 & 6.28 & 0.51 & 0.55 & 0.59 & 0.59 & 0.59\\
& & ENN & 91.55 & 33.95 & 4.40 & 0.15 & 0.02 & 0 & 0\\
\bottomrule
\end{tabular} }
\end{table}

Additionally, we conducted experiments using the \textit{Projected Gradient Descent (PGD)} adversarial attack \cite{madry2017towards}, which is a more iterative and stronger attack compared to FGSM. PGD generates adversarial examples by iteratively perturbing the input image in the direction of the gradient of the loss function, followed by a projection step to ensure that the perturbation stays within a predefined range, typically defined by a norm ball around the original image. Mathematically, the perturbed image $x'$ is computed as follows:

\begin{equation} x^{(0)} = x, \quad x^{(t+1)} = \text{Proj}_{\mathcal{B}(x, \epsilon)} \left( x^{(t)} + \alpha \cdot \text{sign}(\nabla_x J(\theta, x^{(t)}, y)) \right), \end{equation}

where $x^{(t)}$ represents the image at the $t$-th iteration, $\alpha$ is the step size, $\epsilon$ is the maximum allowable perturbation (\textit{i.e.}, the size of the $\mathcal{B}(x, \epsilon)$ ball), and $\text{Proj}$ denotes the projection operation that ensures the perturbed image $x'$ stays within the $\epsilon$-ball around the original image $x$:
\begin{equation}
\mathcal{B}(x, \epsilon) = \left\{ x' : \| x' - x \| \leq \epsilon \right\}.
\end{equation}

After a number of iterations, typically denoted as $T$, the final perturbed image $x'$ is obtained. This iterative process makes PGD a more powerful adversarial attack compared to FGSM. Tab. \ref{tab:pgd} presents the test accuracies of CNN, RS-NN, LB-BNN and ENN models under the PGD adversarial attack on MNIST and CIFAR-10 datasets for different values of $\epsilon$.
{On MNIST, LB-BNN performs well} for $\epsilon=0.005$ and $\epsilon=0.01$, but CNN, LB-BNN and ENN suffer significant {decreases in} accuracy at higher $\epsilon$ values. RS-NN performs better at higher $\epsilon$ values. On CIFAR-10, RS-NN significantly outperforms all models at all $\epsilon$ values, while the other models show a significant drop in performance, even at $\epsilon=0.005$. As explained earlier, RS-NN makes precise predictions with gradient zero and, therefore, is unaffected by PGD.

\subsection{Application to Text Classification}\label{app:text-classification}

The random-set approach can be applied on any classification task. In this section, we detail experiments performed on text classification. 

\textbf{Dataset.} We chose the BBC text dataset \cite{greene06icml}, which consists of various news categories such as tech, business, sport, and entertainment, and used this data for training our model. The task is to classify the given text into one of these categories. The dataset is structured with two columns: category and text, where the category is the label, and the text is the content to be classified. The text data will be preprocessed and fed into the model to predict the category of each text.

\textbf{Model.} To train our model, we leveraged a pre-trained BERT model available on TensorFlow Hub. Specifically, we used the small BERT ($L-4_H-512_A-8$) model, which is a lightweight version of BERT. The model is designed to take text input, preprocess it with BERT's preprocessing layer, pass it through BERT's encoder to generate embeddings, and then use a dropout layer. The final layer is a fully-connected layer with sigmoid activation for the RS-NN BERT classifier (RS-NN BERT), and a fully-connected layer with softmax activation for the standard BERT classifier (CNN BERT).

\textbf{Training.}   We trained these models on the BBC text dataset, fine-tuning the BERT model as part of the training process. Both models were trained for 10 epochs using Adam optimizer, a batch size of 32, and a learning rate of 3e-5. RS-NN BERT has all the sets of classes excluding the null set and full set ($2^5$ = 32 - 2 = 30 classes) and CNN BERT has the 5 original classes.

\textbf{Out-of-distribution detection.} For OoD detection, we use the Emotion Detection from Text \cite{seyeditabari2018emotion} dataset which contains tweets annotated with emotional labels. The dataset includes three columns: tweetid, sentiment, and content, where sentiment represents the emotion behind each tweet. The dataset includes 13 emotion classes such as anger, fear, joy, love, sadness, and surprise, etc, aiming to identify emotional expressions in text.

\begin{table}[!h]
\centering
\caption{Test accuracy, AUROC, AUPRC, iD \textit{vs.} OoD entropy for RS-NN BERT and CNN BERT (both fine-tuned on BERT).}
\vspace{-8pt}
\label{tab:emotion_detection}
\resizebox{\textwidth}{!}{%
\begin{tabular}{lcccccc}
\toprule
 \multirow{2}{*}{Dataset (iD)} & \multirow{2}{*}{Model} & \multirow{2}{*}{Test Acc. (\%) ($\uparrow$)} & \multirow{2}{*}{iD Entropy ($\downarrow$)} &
\multicolumn{3}{c}{Emotion Detection (OoD)} \\ \cmidrule{5-7}
 &&& & AUROC ($\uparrow$) & AUPRC ($\uparrow$) & Entropy ($\uparrow$) \\
\midrule
\multirow{2}{*}{BBC Text (iD)} & RS-NN BERT & \textbf{96.85} & 0.541 $\pm$ 0.196 & \textbf{95.19} & \textbf{95.93} & \textbf{1.510} $\pm$ \textbf{0.611} \\
& CNN BERT & 94.15 & \textbf{0.027 $\pm$ 0.125} & 56.71 & 57.54 & 0.059 $\pm$ 0.191 \\
\bottomrule
\end{tabular} }
\end{table}

\textbf{Experimental results.}  In Tab. \ref{tab:emotion_detection} below, we show the test accuracy, out-of-distribution (OoD) metrics (AUROC, AUPRC), and the in-distribution (iD) \textit{vs.} OoD entropy for RS-NN BERT and CNN BERT. RS-NN achieves higher overall performance with a significantly higher AUROC and AUPRC scores, highlighting the efficiency of the model at differentiating between iD and OoD samples. RS-NN has higher OoD entropy than CNN, but also has higher iD entropy than CNN, which indicates that the uncertainty in predictions is higher as it is a small dataset.

\subsection{Ablation Studies}

In this section, we will address the following questions based on our experimental findings:
\begin{itemize}
    \item \textbf{Q1:} Why are most models not able to differentiate between ImageNet and ImageNet-O in terms of entropy?
    \item \textbf{Q2:} What is the importance of the mass regularizers in the loss function $\mathcal{L}_{RS}$ (Sec. \ref{sec:loss})? Why is it important to properly tune the regularization parameters $\alpha$ and $\beta$, and what impact does this have on the model’s performance? 
    \item \textbf{Q3:} What are the effects of budgeting on RS-NN? How was the number of budgeted focal sets $K$ chosen? How does budgeted RS-NN differ from standard RS-NN with all the subsets in the power set?
    \item \textbf{Q4:} How does UMAP measure as a faster alternative to t-SNE for budgeting in RS-NN? How can the budgeting procedure be adapted for continuous data streams?
\end{itemize}

\subsubsection{ImageNet vs. ImageNet-O}
\label{app:imagenet-o}

In Tab. \ref{tab:ood-table}, the clear separation between iD and OoD entropy is evident for RS-NN, yet for ImageNet \textit{vs.} ImageNet-O, all other models struggle to distinguish between the iD \textit{vs.} OoD entropy and the OoD dataset almost falls within the standard deviation of the iD samples. Also, in Tab. \ref{tab:credal-table}, the credal set widths for CIFAR-10 \textit{vs.} SVHN/Intel Image and MNIST \textit{vs.} F-MNIST/K-MNIST are distinct, but ImageNet \textit{vs.} ImageNet-O is not too different.  

To better understand this, it is important to consider how ImageNet-O is generated. ImageNet-O is a dataset of adversarially filtered examples for ImageNet out-of-distribution detectors. It is created by removing all the ImageNet-1K samples from the full ImageNet-22K dataset. The remaining samples, which do not belong to ImageNet-1K classes, are classified by a 
trained model. The samples classified by 
the model as ImageNet-1K classes with high confidence becomes ImageNet-O. Hence, in general, it is difficult for models to make the iD \textit{vs.} OoD distinction as evident by uncertainty measures in Tab. \ref{tab:ood-table}.

We conducted experiments on Credal Set Width and Pignistic Entropy measures of RS-NN using a ViT-B-16 (Vision Transformer) backbone, as shown in Tab. \ref{tab:vit-table}. 
The pignistic entropy and credal set width show good distinction between iD and OoD measures, unlike the uncertainty measures from other baseline models in Tab. \ref{tab:ood-table}. 

\begin{table}[!h]
    \centering
    \caption{Credal set width and entropy for RS-NN using ViT-B-16 model on ImageNet \textit{vs.} ImageNet-O datasets.}
        \vspace{-8pt}
    \label{tab:vit-table}
        \resizebox{\linewidth}{!}{
        \begin{tabular}{cccc|cc}
        \toprule
        & Datasets &  Credal Set Width & Pignistic Entropy & & Softmax Entropy\\
        \midrule
        \multirow{2}{*}{RS-NN} & ImageNet (iD) ($\downarrow$)  & $0.1082 \pm 0.1926$ & $2.7401 \pm 2.3678$ & \multirow{2}{*}{CNN}  & $2.3442 \pm 1.2243$\\
        \cmidrule{2-4}
        \cmidrule{6-6}
        & ImageNet-O (OoD) ($\uparrow$) & $0.1948 \pm 0.2455$ & $4.4238 \pm 2.5744$ &  \multirow{2}{*}{}  & $2.8619 \pm 1.5884$\\
        \bottomrule
    \end{tabular}
    }
\end{table}

This is specifically due to the peculiar relation between ImageNet and ImageNet-O,
and not a general problem with the models not being able to distinguish samples when they belong to the same image distribution. 

To support our claim, we conducted experiments using CIFAR-10 (iD) \textit{vs.} CIFAR-100 (OoD) as shown in Tab. \ref{tab:cifar10vs100} below, and the results indicate that our model distinguishes between these two datasets well, as evident through the AUROC, AUPRC, entropy and credal set width. Therefore, the issue seems to be specific to ImageNet and ImageNet-O, likely due to the class imbalance in ImageNet. As per the general trend for ImageNet in Tab. \ref{tab:ood-table}, all models behave uncertainly for higher number of classes which makes the difference between iD and OoD less extreme. Although, RS-NN still performs better than other models on ImageNet.

\begin{table}[!h]
    \centering
    \caption{OoD detection and uncertainty estimation performance for iD \textit{vs.} OoD dataset: CIFAR-10 \textit{vs.} CIFAR-100.}
    \vspace{-8pt}
    \label{tab:cifar10vs100}
    \resizebox{\textwidth}{!}{
    \begin{tabular}{lccccccc}
    \toprule
    \multirow{2}{*}{Dataset} &
      \multirow{2}{*}{Model} &
      \multirow{2}{*}{\begin{tabular}[c]{@{}c@{}}In-distribution\\ Entropy ($\downarrow$)\end{tabular}} &
      \multirow{2}{*}{\begin{tabular}[c]{@{}c@{}}Credal Set\\ Width ($\downarrow$)\end{tabular}} &
      \multicolumn{4}{c}{CIFAR-100} \\ \cmidrule{5-8} 
     & & 
      & 
      &
      \multicolumn{1}{c}{AUROC ($\uparrow$)} &
      \multicolumn{1}{c}{AUPRC ($\uparrow$)} &
      \multicolumn{1}{c}{Entropy ($\uparrow$)} &
      \multicolumn{1}{c}{Credal Set Width ($\uparrow$)} \\
      \midrule
    \multicolumn{1}{c}{\multirow{1}{*}{CIFAR-10}} &
      RS-NN & $0.0802 \pm 0.297$ & $0.0061 \pm 0.039$ & $88.01$ & $84.93$ & $0.5778 \pm 0.729$ & $0.0721 \pm 0.165$\\ 
      \bottomrule
    \end{tabular}
}
\end{table}

\subsubsection{Ablation study on hyperparameters $\alpha$ and $\beta$} \label{app:hyperparam}

{
The regularization serves the purpose of ensuring that our model generates valid belief functions, which are constrained by maintaining a sum of masses equal to 1 and ensuring non-negativity. These terms are designed to penalize deviations from valid belief functions. However, it is crucial not to assign too much weight to this term, as excessively penalizing deviations may hinder the model's ability to accurately classify data points. For \textit{e.g.}, in a Variational Auto Encoder (VAE), if we assign too much weight to the KL divergence term in the loss, the model may prioritize fitting the latent distribution at the expense of reconstructing the input data accurately. This imbalance can lead to poor reconstruction quality and suboptimal performance on downstream tasks. }

Hence, it is essential to properly tune the values of $\alpha$ and $\beta$ to ensure that these regularization terms play a meaningful role in the training. The ablation study on hyperparameters is conducted to examine the impact of $\alpha$ and $\beta$ on the model's accuracy. It demonstrates that small values suffice for these parameters. 

In Fig. \ref{acc_a_b}, we show the test accuracies for different values of hyperparameters $\alpha$ and $\beta$ in  $\mathcal{L}_{B-RS}$ loss function (Eq. \ref{final_loss}). Hyperparameters $\alpha$ and $\beta$ adjusts the relative significance of the regularization terms $M_r$ and $M_s$ respectively in $\mathcal{L}_{B-RS}$ loss. The test accuracies are calculated for CIFAR-10 dataset
with a fixed number of focal sets $K = 20$
and varying $\alpha$ (blue) and $\beta$ (red) values, $\alpha/\beta$ = [0.001, 0.005, 0.006, 0.009, 0.01, 0.015, 0.020, 0.025, 0.03, 0.04, 0.05]. Test accuracy is the highest when $\alpha$ and $\beta$ equals $1e-3$.

{The shape of the tuning curve in Fig. \ref{acc_a_b} shows an interesting asymmetric pattern in how accuracy responds to changes in the regularization parameters $\alpha$ and $\beta$.
In the low-regularization regime ($\alpha$, $\beta$ $\in$ [0, 0.001]), accuracy increases from approximately 92.5\% to 93.5\%.
However, beyond the point $\alpha$ = $\beta$ = 0.001, accuracy drops sharply and steadily as $\alpha$ or $\beta$ increase further, reaching significantly lower performance by 0.05. This steep decline suggests that excessive regularization overly constrains the space of feasible belief functions.
This also reinforces the idea that while $\alpha$ and $\beta$ are necessary for producing valid beliefs, they must be used sparingly, and fine-tuning them can have a measurable impact on classification performance.}

\begin{figure}[!h]
 \centering
       \includegraphics[width=0.8\linewidth]{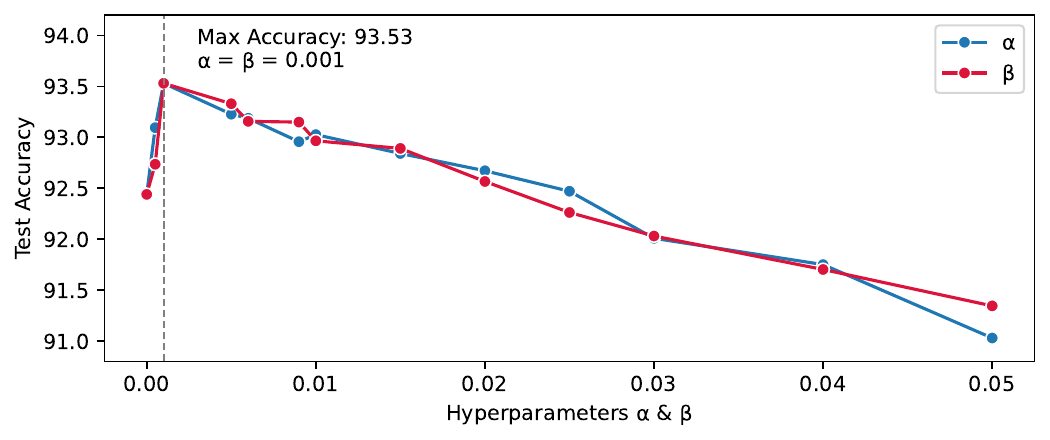}  
        \caption{Test accuracies of RS-NN 
        on the CIFAR-10 dataset using ResNet50, for a fixed $K=20$
        and different values of hyperparameters $\alpha$ and $\beta$ in the loss function.}
       \label{acc_a_b}
\end{figure}

In our experiments across various datasets and architectures, we found that a value of $1e-3$ for the hyperparameters yields satisfactory results. This includes architectures ranging from ResNet-50 to Vision Transformers (ViT-Base-16), and datasets ranging from MNIST ($\approx$ 60,000 images of 10 classes) to ImageNet ($\approx$ 1.1M images of 1000 classes). While conducting a parameter search for each dataset could potentially lead to further optimization, it is worth noting that, in many cases, this step can be omitted without sacrificing performance significantly.
For further optimization one can, of course, perform a tailored parameter search per dataset, but this is not quite necessary.

\textbf{An alternative loss with valid belief functions.} A straightforward way of enforcing the positivity of the masses is to apply softmax over the masses computed in the BCE loss. This ensures that the masses are non-negative and sum to one. After obtaining the softmaxed masses, one can compute the belief functions from these normalized masses and minimize the loss between this computed belief and the original unnormalized belief. 

The random set loss $\mathcal{L}_{RS}$ now becomes,
\[\mathcal{L}_{RS} = \mathcal{L}_{BCE} + \mathcal{L}_{BCE_{norm}},\]
where $\mathcal{L}_{BCE}$ is the BCE loss between belief function logits and ground truth, and $\mathcal{L}_{BCE_{norm}}$ is the BCE loss between the normalized belief function logits reconstructed from mass logits that have been softmaxed.

We implemented this simple method on ResNet50 RS-NN on the CIFAR-10 dataset and obtained an accuracy of $92.47\%$. This is quite close to our original accuracy of $93.53\%$ with the mass regularizations.
This mirrors results in the literature showing that soft constraints work as well as hard ones in practice \cite{marquez2017imposing}.

\subsubsection{Ablation Study on the number of focal sets} \label{app:ablation_k}

\begin{figure}[!ht]
 \centering
       \includegraphics[width=0.6\linewidth]{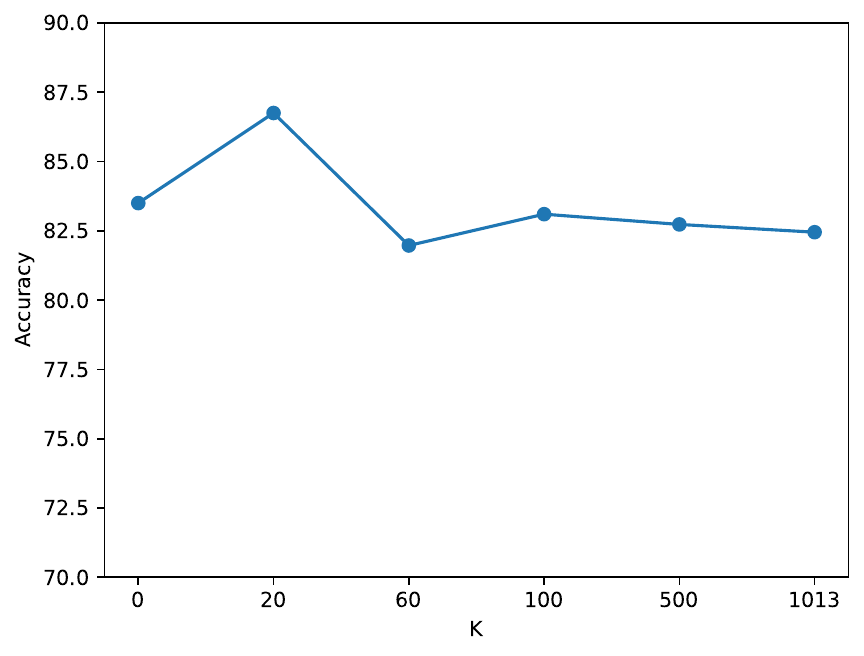}
       \vspace{-8pt}
       \caption{Ablation study on number of non-singleton focal sets $K$ on CIFAR-10 dataset using ViT-Base-16
       (with $\alpha$ = $\beta$ = $1e-3$). 
       The maximum value of K can be 1013 for 10 classes (after excluding the singletons and empty set).}
       \label{budgeting_ablation_k}
          \vspace{-8pt}
\end{figure}



The number of non-singleton focal sets $K$ to be budgeted is a hyperparameter and needs to be studied. A smaller value of $K$ can lead to more similar results to classical classification while a larger value of $K$ can increase the complexity. Therefore, we conducted an ablation study on $K$ on the CIFAR-10 dataset. We found that a small value of $K$ (comparable to the number of classes) works {best and performs} better than $K=0$ (when there are no non-singleton focal sets) (Fig. \ref{budgeting_ablation_k}). {shows a clear peak in performance at a moderate number of focal sets ($K$=20), beyond which the test accuracy drops. This behaviour suggests that a limited number of informative non-singleton focal sets is sufficient to capture useful uncertainty while preserving model generalization. As 
$K$ increases, the model may overfit to complex or overly granular belief assignments, increasing computational complexity without adding meaningful discriminatory power. }

Also, using set prediction along with the proposed budgeting algorithm not only helps induce uncertainty quantification but also improves the performance of the model. Note that performance comparison was made in terms of accuracy. Smet's Pignistic Transform \cite{SMETS2005133} was used to compute class-wise probabilities from the predicted belief function. Fig. \ref{budgeting_ablation_k} shows an ablation study on different number of focal sets $K$ for CIFAR-10 on the Vision-Transformer (ViT-Base-16) model with $\alpha = \beta = 1e-3$.

To further support our claim, we compare a budgeted RS-NN with standard RS-NN (full $2^{N}$ sets) on the CIFAR-10 datasets using ResNet50 as the backbone. 
In Tabs. \ref{tab:full-rsnn} and \ref{tab:full-rsnn-2} below, we report the test accuracy, OoD detection metrics (AUROC, AUPRC), pignistic entropy and credal set width for the standard RS-NN \textit{vs.} budgeted RS-NN. The standard RS-NN has 1024 ($2^{10}$) sets since  CIFAR-10 has 10 classes, and the budgeted RS-NN has $K = 20$ focal (non-singleton) sets, so $10 + 20 = 30$ input sets. Tab. \ref{tab:full-rsnn} shows that budgeted RS-NN performs better than the standard model, especially iD \textit{vs.} OoD entropy, and AUROC and AUPRC scores. 

\begin{table}[!h]
\centering
\caption{OoD detection performance: Comparison of accuracy, AUROC and AUPRC for standard RS-NN \textit{vs.} budgeted RS-NN on the CIFAR-10 dataset.}
\label{tab:full-rsnn}
\vspace{-8pt}
\resizebox{0.8\textwidth}{!}{
\begin{tabular}{lccccc|cc}
\toprule
\multicolumn{4}{c}{In-distribution (iD)} & \multicolumn{4}{c}{Out-of-distribution (OoD)} \\ \midrule
\multirow{2}{*}{Dataset} & \multirow{2}{*}{Model} & \multirow{2}{*}{\begin{tabular}[c]{@{}c@{}}Test accuracy\\ (\%) ($\uparrow$)\end{tabular}} & \multicolumn{1}{c}{\multirow{2}{*}{ECE ($\downarrow$)}} & \multicolumn{2}{c|}{SVHN} & \multicolumn{2}{c}{Intel Image} \\ \cmidrule{5-8} 
 & & & & \multicolumn{1}{c}{AUROC ($\uparrow$)} & \multicolumn{1}{c|}{AUPRC ($\uparrow$)} & \multicolumn{1}{c}{AUROC ($\uparrow$)} & \multicolumn{1}{c}{AUPRC ($\uparrow$)} \\ 
\midrule
\multicolumn{1}{c}{\multirow{2}{*}{CIFAR-10}} & Standard RS-NN & $92.66$ & $0.0501$ & $93.39$ & $91.38$ & $95.84$ & $89.33$\\
& Budgeted RS-NN &
          $\mathbf{93.53}$  & $\mathbf{0.0484}$ &
          $\mathbf{94.91}$&
          $\mathbf{93.72}$ & 
          $\mathbf{97.39}$	&
          $\mathbf{90.27}$ \\
\bottomrule
\end{tabular}}
\end{table}

\begin{table}[!h]
\centering
\caption{Uncertainty estimation: Comparison of pignistic entropy and credal set width for standard RS-NN \textit{vs.} budgeted RS-NN on the CIFAR-10 dataset.}
\label{tab:full-rsnn-2}
\vspace{-8pt}
\resizebox{\textwidth}{!}{
\begin{tabular}{lccccc|cc}
\toprule
\multicolumn{4}{c}{In-distribution (iD)} & \multicolumn{4}{c}{Out-of-distribution (OoD)} \\ \midrule
\multirow{2}{*}{Dataset (iD)} & \multirow{2}{*}{Model} & \multirow{2}{*}{Entropy (iD) ($\downarrow$)}
 & \multirow{2}{*}{Credal Set Width (iD) ($\downarrow$)} &  \multicolumn{2}{c|}{SVHN} & \multicolumn{2}{c}{Intel Image} \\ \cmidrule{5-8} 
& & &  & \multicolumn{1}{c}{Entropy ($\uparrow$)} & \multicolumn{1}{c|}{Credal Set Width ($\uparrow$)} & \multicolumn{1}{c}{Entropy ($\uparrow$)} & \multicolumn{1}{c}{Credal Set Width ($\uparrow$)} \\   
\cmidrule{1-8}
\multicolumn{1}{c}{\multirow{2}{*}{CIFAR-10}} & Standard RS-NN & $0.286 \pm 0.80$ & $0.048 \pm 0.15$ &$\mathbf{1.205} \pm \mathbf{1.20}$ & $\mathbf{0.408} \pm \mathbf{0.07}$  & $1.490 \pm 0.71$ & $\mathbf{0.669} \pm \mathbf{0.21}$\\
& Budgeted RS-NN  & $\mathbf{0.088} \pm \mathbf{0.308}$ & $\mathbf{0.007} \pm \mathbf{0.044}$ & 
          $1.132 \pm 0.855$ & $0.260 \pm 0.322$ & 
          $\mathbf{1.517} \pm \mathbf{0.740}$  &  $0.587 \pm 0.367$\\
\bottomrule
\end{tabular}}
\end{table}

\textit{This shows that while, in theory, using the full power set is ideal for uncertainty estimation, in practice, having all those degrees of freedom might make it more difficult for the network to learn.} When a model is confronted with an overwhelming number of input sets, it may struggle to identify meaningful patterns amidst the noise created by less relevant or redundant combinations. This excessive flexibility can hinder the network's ability to converge effectively, as it may become trapped in local minima or overfit to the training data. This can dilute the model's capacity to generalize well to unseen data, thereby complicating the learning of the underlying relationships between features and the set structure of the data.

\subsubsection{On the feasibility of budgeting of sets} \label{app:budget}

Tabs. \ref{tab:umap1} and \ref{tab:umap2} show that replacing t-SNE with UMAP for budgeting in the RS-NN model yields comparable performance in terms of test accuracy, out-of-distribution (OoD) detection, and uncertainty estimation. Both t-SNE and UMAP are capable of embedding data in a 3D space, with UMAP employing a faster algorithm that utilizes five nearest neighbors and a minimum distance of 0.9. This approach ensures that the integrity of the data structure is maintained while significantly reducing computational costs. UMAP offers an efficient alternative without compromising the model's effectiveness. To make the generation of embeddings for budgeting of focal sets more efficient, we can utilize faster alternatives like UMAP instead of t-SNE, without compromising the overall model performance. {We chose t-SNE for our main budgeting experiments as it better preserves local neighbor structure \cite{wattenberg2016use}, often resulting in tighter and more visually distinct class clusters.}

\begin{table}[!htbp]
\centering
\caption{Test accuracy and OoD detection (AUROC, AUPRC) results on RS-NN where budgeting is done using t-SNE \textit{vs.} budgeting done on UMAP.}
\label{tab:umap1}
\vspace{-8pt}
\resizebox{\textwidth}{!}{
\begin{tabular}{lccccc|cc}
\toprule
\multicolumn{4}{c}{In-distribution (iD)} &
  \multicolumn{4}{c}{Out-of-distribution (OoD)} \\ \midrule
\multirow{2}{*}{Dataset} &
  \multirow{2}{*}{Model} &
  \multirow{2}{*}{\begin{tabular}[c]{@{}c@{}}Test accuracy\\ (\%) ($\uparrow$)\end{tabular}} &
  \multicolumn{1}{c}{\multirow{2}{*}{ECE ($\downarrow$)}} &
  \multicolumn{2}{c|}{SVHN} & 
  \multicolumn{2}{c}{Intel Image} \\ \cmidrule{5-8} 
 & & & &
  \multicolumn{1}{c}{AUROC ($\uparrow$)} &
  \multicolumn{1}{c}{AUPRC ($\uparrow$)} &
  \multicolumn{1}{c}{AUROC ($\uparrow$)} &
  \multicolumn{1}{c}{AUPRC ($\uparrow$)} \\
  \midrule
\multicolumn{1}{c}{\multirow{2}{*}{CIFAR-10}} &
  RS-NN (t-SNE) &
  93.53 &
  0.0484 &
  94.91 &
  93.72 & 
  97.39 &
  90.27 
  \\
  & RS-NN (UMAP) 
  & 92.97 & 0.0482 &
  94.31 &
  92.46 &
  96.22 &
  89.66 
  \\ \bottomrule
\end{tabular}
}
\end{table}

\begin{table}[!htbp]
\centering
\caption{Entropy and credal set width results on RS-NN where budgeting is done using t-SNE \textit{vs.} budgeting done on UMAP. The goal is to minimize both metrics for in-distribution (iD) data while increasing them for out-of-distribution (OoD) data.}
\label{tab:umap2}
\vspace{-8pt}
\resizebox{\textwidth}{!}{
\begin{tabular}{lccccc|cc}
\toprule
\multicolumn{4}{c}{In-distribution (iD)} &
  \multicolumn{4}{c}{Out-of-distribution (OoD)} \\ \midrule
\multirow{2}{*}{Dataset} &
  \multirow{2}{*}{Model} &
  \multirow{2}{*}{Entropy ($\downarrow$)} &
  \multirow{2}{*}{Credal Set Width ($\downarrow$)} &
  \multicolumn{2}{c|}{SVHN} &
  \multicolumn{2}{c}{Intel Image} \\ \cmidrule{5-8} 
 & & 
  & 
  &
  \multicolumn{1}{c}{Entropy ($\uparrow$)} &
  \multicolumn{1}{c|}{Credal Set Width ($\uparrow$)} &
  \multicolumn{1}{c}{Entropy ($\uparrow$)} &
  \multicolumn{1}{c}{Credal Set Width ($\uparrow$)} \\
  \midrule
\multicolumn{1}{c}{\multirow{2}{*}{CIFAR-10}} &
  RS-NN (t-SNE) &
  0.088 $\pm$ 0.308 &
    0.007 $\pm$ 0.044 &
  1.132 $\pm$ 0.855 &
    0.260 $\pm$ 0.322 &
  1.517 $\pm$ 0.740 &
  0.587 $\pm$ 0.367 \\
& RS-NN (UMAP) &
  0.082 $\pm$ 0.305 &
    0.007 $\pm$ 0.047 &
  1.119 $\pm$ 0.887 & 
    0.341 $\pm$ 0.374 &
  1.423 $\pm$ 0.762 &
  0.582 $\pm$ 0.402 \\
  \bottomrule
\end{tabular}
}
\end{table}

\section{Discussion and Implications}

This chapter proposes a novel \emph{Random-Set Neural Network (RS-NN)} for uncertainty estimation in classification predicting belief functions.
{This random-set representation is a foundational approach that acts as a versatile wrapper, 
applicable
to any model architecture and classification task (e.g, text). 
This concept, along with budgeting for optimal selection of sets, is unprecedented.}
RS-NN outperforms both state-of-the-art uncertainty estimation models and standard CNNs in terms of predictive performance (\textbf{accuracy} and \textbf{calibration}), \textbf{out-of-distribution (OoD) detection}, and \textbf{adversarial robustness}. It provides \textbf{reliable uncertainty} quantification and \textbf{scales effectively} to large architectures and datasets.

What distinguishes RS-NN, however, is not just empirical superiority but the epistemological stance embedded in its design. These results are not incidental; they emerge directly from the fact that RS-NN treats prediction as inference over \textit{sets} of outcomes. \textbf{This reflects the central claim of this thesis:} that epistemic uncertainty must be captured through second-order frameworks which can represent both partial knowledge and ambiguity in a way that other models cannot. RS-NN achieves this by modifying the network's output space to predict belief over sets and by deriving pignistic probabilities for decision-making. This makes the abstract idea of epistemic modeling tangible, producing a concrete, scalable algorithm that generalizes across tasks and architectures. The proposed \textit{budgeting technique} enables set-valued learning to be computationally feasible, selecting informative class subsets without incurring exponential complexity. In doing so, it exemplifies how theory-driven modeling, grounded in random set theory and belief functions, can lead to practical advances in machine learning under uncertainty.

Building on the success of Random-Set Neural Networks (RS-NNs) in classification, applications of the random set framework to Large Language Models (LLMs) are demonstrated in Part \ref{part:app}, with the development of Random-Set Large Language Models (RS-LLMs) for text generation presented in Chapter \ref{ch:rsllm}, alongside other practical applications such as weather classification for autonomous racing (Chapter \ref{ch:weather}).

\chapter{Credal Set Models} 
\label{ch:cre_models} 

This chapter presents a brief description of credal set models for classification developed collaboratively within the epistemic AI framework. Although a detailed exposition of each model is beyond the scope of this thesis due to word limit constraints, their core characteristics are summarized here. {These co-authored works} have been published in \citet{wang2024creinns} and \citet{wang2024credal}. These models are evaluated in Part \ref{part:eval} and their overall performance comparison with other approaches are shown in Sec. \ref{sec:impact-epi}. The following sections provide background and a concise description of each model. 

Credal sets provide a formalism for representing uncertainty more robustly than traditional probabilistic models. In contrast to Bayesian models, which rely on precise priors, credal sets adopt a more agnostic approach by accommodating a set of plausible probability distributions. This is particularly advantageous when probability estimates are based on incomplete or conflicting sources, such as in medical diagnosis or risk assessment \cite{cozman00credal}.
\textit{Credal networks}, for example, generalize Bayesian networks by allowing imprecise conditional probability tables, leading to more cautious and reliable inferences in high-stakes domains. By capturing a range of admissible probabilities rather than a single distribution, credal sets help mitigate overconfidence in decision-making.

Models that generate \emph{credal sets} \cite{levi1980enterprise,zaffalon-treebased,cuzzolin2010credal,antonucci2010credal,cuzzolin2008credal} 
represent uncertainty in predictions by providing a set of plausible outcomes, rather than a single point estimate. 
A \emph{credal set} \cite{levi1980enterprise,zaffalon-treebased,cuzzolin2010credal,antonucci2010credal,cuzzolin2008credal} is a convex set of probability distributions on the target (class) space. 
Credal sets can be elicited, for instance, 
from predicted
probability intervals \cite{wang2024creinns, caprio2023imprecise} 
$[\hat{\underline{p}}(y), \hat{\overline{p}}(y)]$, encoding lower and upper bounds, respectively, to the probabilities of each of the classes (in list of classes $\mathbf{Y}$):
\begin{equation}\label{eq:credal_interval}
    \hat{\mathbb{C}r} (y \mid \mathbf{x}, \mathbb{D}) = \{ p \in \mathcal{P} \ | \ \hat{\underline{p}}(y) \leq p(y) \leq \hat{\overline{p}}(y), \forall y \in \mathbf{Y} \}.
\end{equation}

\noindent where $\mathcal{P}$ denotes the set of all probability distributions over $\mathbf{Y}$.


A credal set is efficiently represented by its extremal points; their number can vary, depending on the size of the class set and the complexity of the network prediction the credal set represents.


Based on these principles, we introduce two models: Cre-INN and CreDE, each providing a distinct mechanism for constructing and leveraging credal sets in deep learning.

\section{Credal Interval Neural Networks (Cre-INN)}\label{sec:creinn}

Credal-Set Interval Neural Networks (CreINNs) \cite{wang2024creinns}, as shown in Fig. \ref{FIG: CreINNConcept}, retain the fundamental structure of traditional Interval Neural Networks, capturing weight uncertainty through deterministic intervals. CreINNs are designed to predict an upper and a lower probability bound for each class, rather than a single probability value. The probability intervals can define a credal set, facilitating estimating different types of uncertainties associated with predictions. 

\begin{figure*}[ht]
\begin{center}
\includegraphics[width=\linewidth]{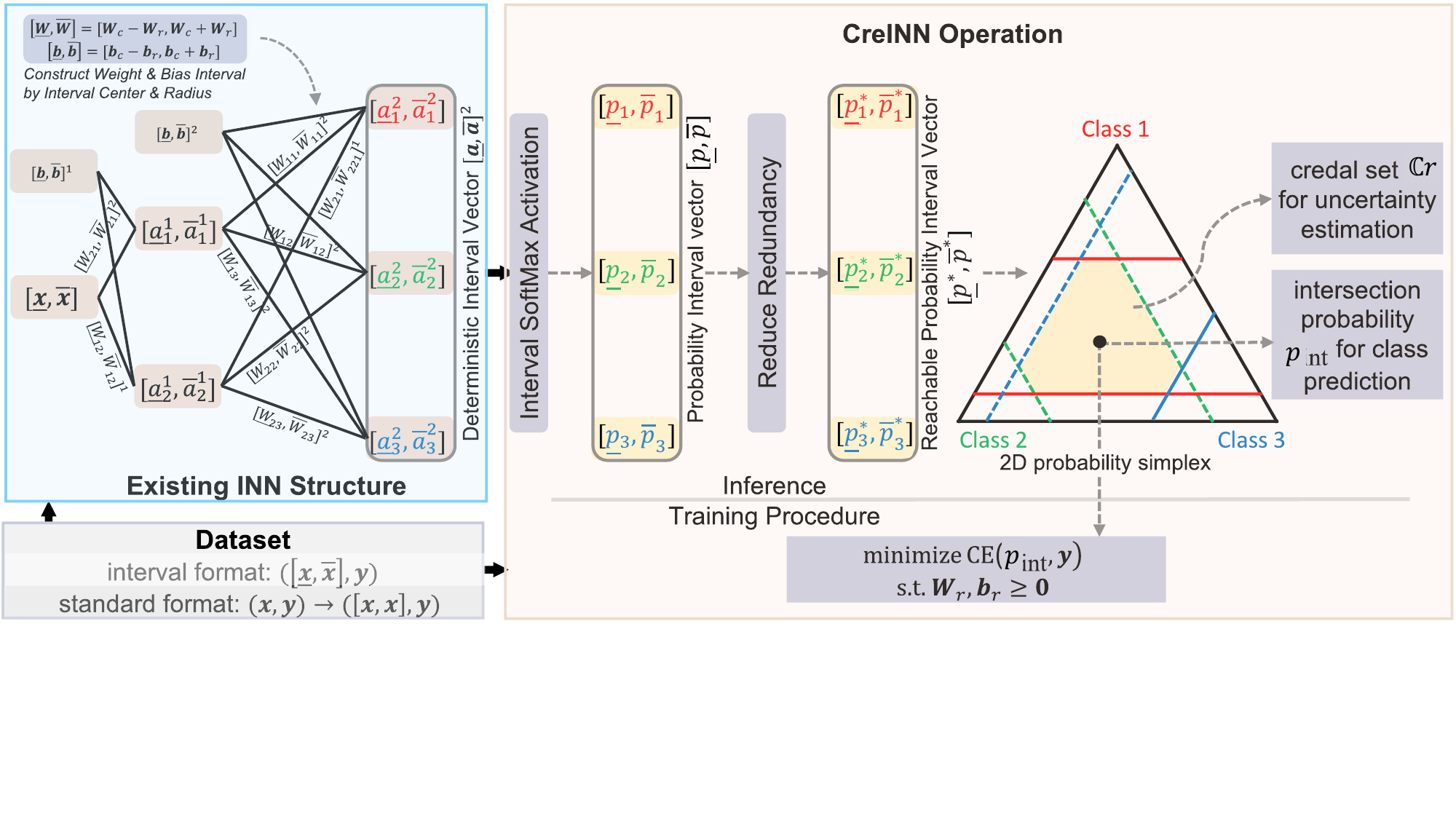}
\vspace{-75pt}
\caption[Illustration of the proposed CreINN model for a three-class classification task.]{{Illustration of the proposed CreINN model for a three-class classification task. CreINN follows the conventional INN architecture, representing inputs $[\underline{{x}}, \overline{{x}}]$, node outputs, weights, and biases (\textit{i.e.}, $[\underline{a}_z^l, \overline{a}_z^l]$ and $[\underline{w}_{jz}^l, \overline{w}_{jz}^l]$ for the $z^{th}$ node of $l^{th}$ layer and ${[\underline{{b}}, \overline{{b}}]}^l$ for the $l^{th}$ layer, respectively) as deterministic intervals. Using the proposed Interval SoftMax activation, a set of probability intervals $[\hat{\underline{p}}, \hat{\overline{p}}]\!:=\!{\{[\hat{\underline{p}}_{c_i}, \hat{\overline{p}}_{c_i}]\}}_{{i}=1}^{{i}=3}$ can derived from the outputted deterministic output interval vector. Through redundancy reduction, the resulting reachable probability interval $[\hat{\underline{p}}^*, \hat{\overline{p}}^*]$ (shown as parallel dashed lines) can define a credal set $\hat{\mathbb{C}r}$ for uncertainty estimation, depicted as the light orange convex hull within the probability simplex (a triangle representing all probability distributions over the target space). In addition, an intersection probability ${q}_{\text{int}}$ can be computed from these probability intervals for class classification purposes. Model training involves minimizing the cross-entropy (CE) loss with constraints that guarantee valid weight and bias intervals. Moreover, the proposed CreINN can handle both interval and standard format data.}}
\label{FIG: CreINNConcept}
\end{center}
\end{figure*}

CreINN introduces a novel \textit{Interval SoftMax activation} that converts output interval scores 
\begin{equation}
[\underline{{a}}^L, \overline{{a}}^L] = \{[\underline{a}_{c_i}^L, \overline{a}_{c_i}^L]\}_{{i}=1}^N,
\end{equation}

where $N$ is the number of classes and $[\underline{a}_{c_i}^L, \overline{a}_{c_i}^L]$ denotes the score interval for the ${c_i}^{th}$ class at the final layer $L$, into valid probability intervals 
\begin{equation}
[\hat{\underline{p}}, \hat{\overline{p}}] = \{[\hat{\underline{p}}_{c_i}, \hat{\overline{p}}_{c_i}]\}_{{i}=1}^N,
\end{equation}

with $[\hat{\underline{p}}_{c_i}, \hat{\overline{p}}_{c_i}]$ representing the predicted probability interval for class ${c_i}$. These intervals define a credal set 
\begin{equation}
\hat{\mathbb{C}r} = \left\{ p \in \mathcal{P} \mid p_{c_i} \in [\hat{\underline{p}}_{c_i}, \hat{\overline{p}}_{c_i}], \ \sum_{i=1}^N p_{c_i} = 1 \right\},
\end{equation}

which is a convex set of probability distributions capturing uncertainty over class assignments. The \textit{Interval SoftMax} is given by
\begin{equation}
\hat{\underline{p}}_{c_i} = \frac{\exp(\underline{a}_{c_i}^L)}{\exp(\underline{a}_{c_i}^L) + \sum_{{c_j} \neq {c_i}} \exp\left(\frac{\underline{a}_{c_j}^L + \overline{a}_{c_j}^L}{2}\right)}, \quad
\hat{\overline{p}}_{c_i} = \frac{\exp(\overline{a}_{c_i}^L)}{\exp(\overline{a}_{c_i}^L) + \sum_{{c_j} \neq {c_i}} \exp\left(\frac{\underline{a}_{c_j}^L + \overline{a}_{c_j}^L}{2}\right)},
\end{equation}

which ensures that the output intervals are valid probabilities, maintain order ($\hat{\underline{p}}_{c_i} \leq \hat{\overline{p}}_{c_i}$), and are differentiable for backpropagation.

CreINN enforces interval constraints on parameters by representing weights and biases as center-radius intervals:
\begin{equation}
[\underline{{W}}, \overline{{W}}] = [{W}_c - {W}_r, \ {W}_c + {W}_r], \quad
[\underline{{b}}, \overline{{b}}] = [{b}_c - {b}_r, \ {b}_c + {b}_r],
\end{equation}

where ${W}_c$ and ${b}_c$ are the center (midpoint) parameters, and ${W}_r \geq {0}$ and ${b}_r \geq {0}$ are the non-negative radii (half-widths) of the intervals, guaranteeing $\underline{{W}} \leq \overline{{W}}$ and $\underline{{b}} \leq \overline{{b}}$.

To prevent interval explosion in deeper architectures, an \textit{Interval Batch Normalization} method is introduced, which stabilizes the interval propagation.
Since the initial probability intervals $[\hat{\underline{p}}, \hat{\overline{p}}]$ can be redundant (\textit{i.e.}, some interval bounds are not achievable by any probability vector in $\hat{\mathbb{C}r}$, CreINN refines these using reachable bounds:
\begin{equation}
\hat{\overline{p}}_{c_i}^* = \min \left(\hat{\overline{p}}_{c_i}, \ 1 - \sum_{{c_j} \neq {c_i}} \hat{\underline{p}}_{c_j} \right), \quad
\hat{\underline{p}}_{c_i}^* = \max \left(\hat{\underline{p}}_{c_i}, \ 1 - \sum_{{c_j} \neq {c_i}} \hat{\overline{p}}_{c_j} \right),
\end{equation}

which ensure that for each class $c_i$, the refined interval $[\hat{\underline{p}}_{c_i}^*, \hat{\overline{p}}_{c_i}^*]$ is feasible within the simplex constraint. The resulting credal set formed by these reachable intervals accurately reflects the uncertainty in class probabilities.

Classification decisions can be made by considering intersection probabilities derived from these intervals, and the network is trained by minimizing a cross-entropy loss adapted to the interval-valued outputs subject to these interval constraints.


Experiments on standard multiclass and binary classification tasks demonstrate that the proposed CreINNs can achieve superior or comparable quality of uncertainty estimation compared to variational Bayesian Neural Networks (BNNs) and Deep Ensembles. Furthermore, CreINNs significantly reduce the computational complexity of variational BNNs during inference. Moreover, the effective uncertainty quantification of CreINNs is also verified when the input data are intervals.

\section{Credal Deep Ensembles (CreDE)}\label{sec:crede}

Credal Deep Ensembles (CreDEs) \cite{wang2024credal} (Fig. \ref{Fig: Concept}) build on the deep ensemble framework by incorporating credal set theory to enhance epistemic uncertainty modeling. Each ensemble member, referred to as a Credal-Set Neural Network (CreNet), is trained to output interval-valued class probabilities. The ensemble’s aggregated output defines a credal set over the label space.

Each ensemble member, a Credal-Set Neural Network (CreNet), outputs interval-valued class probabilities defined by lower and upper bounds for each class, denoted as
\begin{equation}
[\hat{\underline{p}}_{c_i}, \hat{\overline{p}}_{c_i}], \quad {i} = 1, \ldots, N,
\end{equation}

where $N$ is the number of classes. These intervals form a {credal set} as shown in Eq. \ref{eq:credal_interval}.
Training each CreNet involves a distributionally robust optimization (DRO)-inspired loss function, which encourages interval widths to reflect uncertainty due to potential dataset shift. This approach enhances the model’s robustness by simulating divergences between training and test distributions during learning.

\begin{figure}[!htpb]
\vspace{-4pt}
\begin{center}
\includegraphics[width=\textwidth]{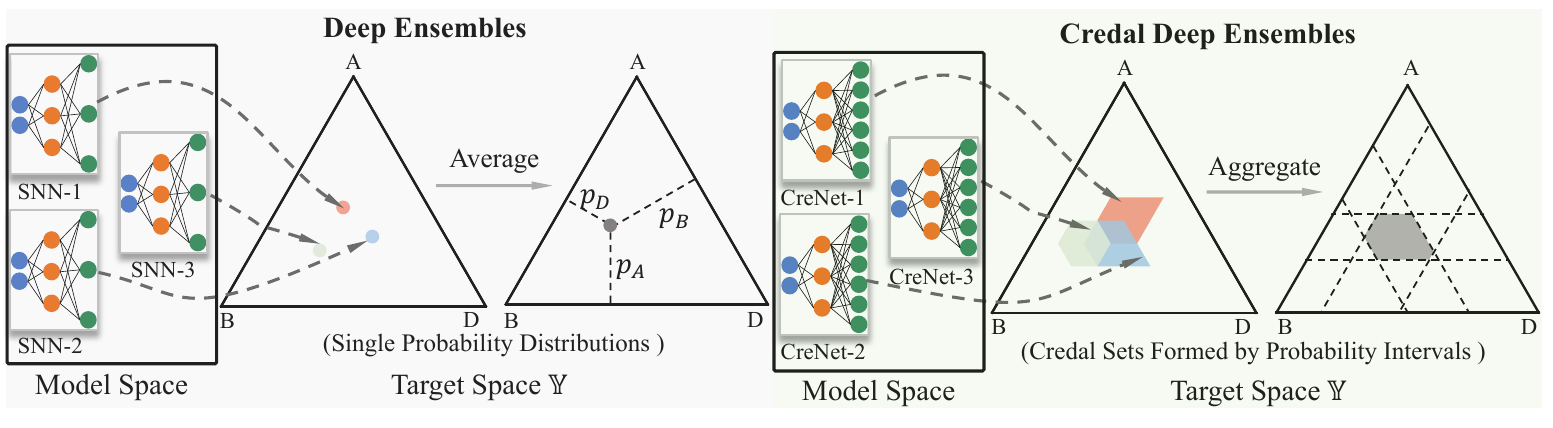}
\end{center}
\vspace{-4pt}
\caption[Comparison between the proposed Credal Deep Ensembles and traditional Deep Ensembles.]{Comparison between the proposed Credal Deep Ensembles and traditional Deep Ensembles. The former aggregate a collection of credal set predictions from CreNets as the final (credal) prediction, whereas the latter average a set of single probability distributions from standard SNNs as the outcome. For \textit{e.g.}, in the probability simplex associated with the target space $\mathbb{Y}\!=\!\{A, B, D\}$ (the triangle in the figure), a probability vector $(q_A, q_B, q_D)$ is represented as a single point. 
For each CreNet, the predicted lower and upper probabilities of each class act as constraints (parallel lines) which determine a credal prediction (in gray).} 
\label{Fig: Concept} 
\vspace{-3pt}
\end{figure}

Extensive evaluation across multiple OoD benchmarks (\textit{e.g.}, CIFAR10/100 \textit{vs.}  SVHN or Tiny-ImageNet, CIFAR10 \textit{vs.}  CIFAR10-C) and architectures (ResNet50, VGG16, ViT Base) shows that CreDEs deliver superior uncertainty estimates. Compared to traditional ensembles and variational BNNs, CreDEs yield lower expected calibration error (ECE) and higher AUROC and AUPRC on OoD data. These results confirm the efficacy of credal set modeling in capturing epistemic uncertainty more faithfully.

In Sec. \ref{sec:impact-epi},  we compare the performance of CreINN and CreDE with all other models in terms of accuracy \textit{vs.}  ECE (Fig. \ref{fig:ood-acc-ece}(b)) and OoD detection scores (Fig. \ref{fig:ood-acc-ece}(a)).
\part{{{\MakeUppercase{Evaluating Epistemic Predictions}}}}
\label{part:eval}
\thispagestyle{plain}
\chapter{Evaluation under Uncertainty} 
\label{ch:eval_unc}

As mentioned in Sec. \ref{sec:predictive-undertainty}, in 
classification, 
it is imperative to account for uncertainty inherent in the predictive process, called \emph{predictive uncertainty}. 
It can be represented using the predictive distribution $\hat{p}(y \mid \mathbf{x}, \mathbb{D})$, where $y \in \mathbf{Y}$ is a label, $\mathbf{x} \in \mathbf{X}$ an input instance, and $\mathbb{D} = \{ (\mathbf{x}_i, y_i)\}_{i=1}^\mathcal{N} \in \mathbf{X} \times \mathbf{Y}$ 
the available training set, $\mathcal{N}$ being the number of 
training instances.
In 
Bayesian models \cite{Buntine1991BayesianB, neal2012bayesian, DBLP:journals/corr/abs-2007-06823, https://doi.org/10.48550/arxiv.1312.6114}, this uncertainty is explicitly represented through posterior predictive distributions over the parameter space. 
In Ensembles \cite{lakshminarayanan2017simple}, a predictive distribution is formed by aggregating the individual predictions generated by multiple independently trained models. 
In Evidential models \cite{sensoy}, instead, predictions are parameters of a Dirichlet posterior in the label space. In deterministic \cite{mukhoti2023deep} models, predictions are point estimates obtained from a final softmax layer, similar to those of standard neural networks (CNN).
Finally, in imprecise-probabilistic models \cite{caprio2023imprecise, wang2024creinns}, predictions correspond to either credal sets \cite{levi1980enterprise, tong2021evidential}
or random sets (Chapter \ref{ch:rsnn}), 
rather than single, precise probability vectors.

\textbf{No common evaluation setting exists to date for all these diverse kinds of predictions}, making it difficult for practitioners to consistently evaluate or rank uncertainty models based on their predictive performance.

This chapter reviews the existing literature on the evaluation of uncertainty in machine learning (Sec. \ref{sec:related-eval}) and examines the forms of predictions produced by various uncertainty-aware models (Sec. \ref{sec:classes-epistemic}). These include all baseline methods introduced in Chapter \ref{ch:unc_models}, as well as the random set model developed in this thesis (Chapter \ref{ch:rsnn}) and the credal set models presented in Chapter \ref{ch:cre_models}.

\section{Current metrics for evaluation}\label{sec:related-eval}


\textbf{Aleatoric \textit{vs.}  epistemic uncertainty}.
Most scholars distinguish
\emph{aleatoric} uncertainty, caused by randomness in the data, and \emph{epistemic} uncertainty, stemming from a lack of knowledge about data distribution or model parameters \cite{kendall2017uncertainties,hullermeier2021aleatoric,manchingal2022epistemic,Zaffalon}.
Different metrics are used to quantify uncertainty across model types (see Sec. \ref{app:A}), depending on the nature of the predictions. For instance, Bayesian models often assess epistemic uncertainty using Mutual Information, while Deep Ensembles typically rely on predictive variance. As noted in Sec. \ref{app:A_BNN}, such measures are not directly comparable across models, as they operate over different value ranges and reflect uncertainty in model-specific ways.

\textbf{Metrics for evaluation under uncertainty}.
Evaluation metrics for probabilistic graphical models ( \textit{e.g.} , Bayesian networks) were detailed in \citet{MARCOT201250}, including model sensitivity analysis \cite{thogmartin2010sensitivity} 
and a model emulation strategy to enable faster evaluation using surrogate models. 
\citet{SNOWLING200117} instead, considered uncertainty for model selection, using uncertainty quantification for informed decision-making.
An `alternate hypotheses' approach was proposed by \citet{zio1996two} for evaluating model uncertainty which identifies a family of possible alternate models and probabilistically combines their predictions based on Bayesian model averaging 
\cite{droguett2008bayesian}. In 
\citet{park2011quantifying},
the model probability in the alternate hypotheses approach was further quantified by the measured deviations between data and model predictions. 
Our approach for model selection of uncertainty-aware models (Chapter \ref{ch:eval_framework}) is unique, especially as we no longer rely solely on test accuracy to select the best model, but can instead establish a trade-off between accuracy and imprecision.

\section{Classes of Epistemic Predictions}
\label{sec:classes-epistemic}

The predictions of a classifier can be plotted in the simplex (convex hull) $\mathcal{P}$ of the one-hot probability vectors assigning probability 1 to a particular class.
For instance, in a $3$-class classification scenario ($\mathbf{Y}= \{a, b, c\}$), 
the simplex would be a 2D simplex (triangle) connecting three points, each representing one of the classes, as shown in Fig. \ref{fig:simplex_preds_final}, which 
depicts all types of model predictions considered here.

\begin{figure}[!h]
    \centering
    \hspace*{-0.6cm}
\includegraphics[width=0.6\textwidth]{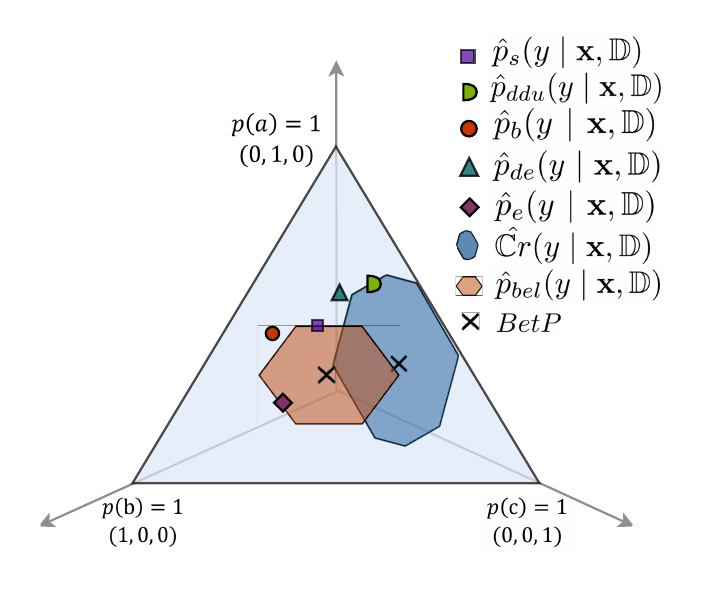}
\vspace{-20pt}
    \caption{
    Different types of uncertainty-aware model predictions, shown in a unit {simplex} of probability distributions defined on the list of classes 
    $\mathbf{Y}= \{a, b, c\}$.
    }
    \label{fig:simplex_preds_final}
    \vspace{-2mm}
\end{figure}

An overview of the predictions generated by all uncertainty-aware models is provided below. Further details on the functioning of these models can be found in Chapter \ref{ch:unc_models}.

\textbf{Traditional Neural Networks} (CNNs)
predict a vector of $N$ scores, one for each class, duly \emph{calibrated} to a probability vector representing a (discrete, categorical)
probability distribution over the list of classes $\mathbf{Y}$, 
$
    \hat{p}_{s}(y \mid \mathbf{x}, \mathbb{D}),
$
which represents the probability of observing class $y$ given the input $\mathbf{x}$.

\textbf{Bayesian Neural Networks} 
(BNNs) \citep{lampinen2001bayesian, titterington2004bayesian, goan2020bayesian, hobbhahn2022fast} 
compute a predictive distribution $\hat{p}_{b}(y \mid \mathbf{x}, \mathbb{D})$ by integrating over a learnt posterior distribution of model parameters $\theta$ given training data $\mathbb{D}$. 
This
is often infeasible due to the complexity of the posterior, leading to the use of \emph{Bayesian Model Averaging} (BMA), which approximates the predictive distribution by averaging over predictions from multiple samples.
BMA inadvertently smooths out predictive distributions, diluting the inherent uncertainty present in individual models \citep{hinne2020conceptual, graefe2015limitations} as shown in Fig. \ref{fig:BMA_vs_nonBMA}, Sec. {\ref{app:A_BNN}}}. 
\textit{When applied to classification, BMA yields
point-wise predictions.} For fair comparison and to overcome BMA's limitations, 
in our proposed method in the next Chapter \ref{ch:eval_framework}, we also use sets of prediction samples obtained from the different posterior weights \emph{before} averaging.

In \textbf{Deep Ensembles} 
(DEs) \citep{lakshminarayanan2017simple}, 
a prediction 
$\hat{p}_{de}(y \mid \mathbf{x}, \mathbb{D})$ for an input $\mathbf{x}$ is obtained by averaging the predictions of $K$ individual models:
$
   \hat{p}_{de}(y \mid \mathbf{x}, \mathbb{D}) = \frac{1}{K} \sum_{k=1}^{K} \hat{p}_k(y \mid \mathbf{x}, \mathbb{D}) ,
$
where $\hat{p}_{k}$ represents the prediction of the $k$-th model, trained independently with different initialisations or architectures. Uncertainty is then quantified as in Sec. {\ref{app:A_DE}}.

\textbf{Evidential Deep Learning} (EDL) models \citep{sensoy} make predictions $\hat{p}_{e}(y \mid \mathbf{x}, \mathbb{D})$ as parameters of a second-order Dirichlet distribution on the class space, instead of softmax probabilities. EDL uses these parameters to obtain a pointwise 
prediction.
As for BNNs, averaging may not be optimal; hence, in our proposed evaluation method (Chapter \ref{ch:eval_framework}), we consider individual prediction samples.

\textbf{Deep Deterministic Uncertainty} (DDU) \citep{mukhoti2023deep} models 
differ from other uncertainty-aware baselines as they do not represent uncertainty in the prediction space, but do so in the input space by identifying whether an input sample is in-distribution (iD) or out-of-distribution (OoD). As a result, DDU provides predictions $\hat{p}_{ddu}(y \mid \mathbf{x}, \mathbb{D})$ in the form of softmax probabilities akin to standard neural networks (CNNs).

\textbf{Credal Models}. 
Credal sets ($\hat{\mathbb{C}r} (y \mid \mathbf{x}, \mathbb{D})$) can be computed
from predicted
probability intervals \citep{wang2024creinns, caprio2023credal} 
$[\hat{\underline{p}}(y), \hat{\overline{p}}(y)]$, encoding lower and upper bounds, respectively, to the probabilities of each of the classes \citep{probability_interval_1994} as shown in Eq. \ref{eq:credal_interval}.
A credal set is efficiently represented by its extremal points; their number can vary, depending on the size of the class set and the complexity of the network prediction the credal set represents.

\textbf{Belief Function Models}. 
A predicted belief function $\hat{Bel}$ on 
$\mathbf{Y}$ is mathematically equivalent to the credal set 
\begin{equation} \label{eq:consistent}
\mathbb{C}r_{\hat{Bel}} (y \mid \mathbf{x}, \mathbb{D}) 
=
\big \{ 
p \in \mathcal{P}  \; \big | \; p(A) \geq \hat{Bel}(A) 
\big \}.
\end{equation} 
Its center of mass, termed 
\emph{pignistic probability} \citep{SMETS2005133} $BetP[\hat{Bel}]$,
assumes the role of the predictive distribution for belief function models such as RS-NN (Chapter \ref{ch:rsnn}) and E-CNN \citep{tong2021evidential}:
$
   \hat{p}_{bel}(y \mid \mathbf{x}, \mathbb{D}) = 
    BetP[\hat{Bel}]. 
$

In summary, predictive distributions play a pivotal role in all uncertainty-aware classification models, as they encapsulate both a model's uncertainty and the inherent variability in the data. 
Our proposed unified evaluation framework (Sec. \ref{sec:evaluation}) relies on mapping all such types of predictive distributions to a credal set.


\chapter{A Unified Evaluation Framework for Epistemic Predictions} 
\label{ch:eval_framework} 


This chapter of the thesis proposes a novel \emph{unified evaluation framework} (Chapter \ref{sec:evaluation})
which provides a holistic assessment of predictions produced by uncertainty-aware classifiers for model selection (Sec. \ref{sec:model-selection}).
This framework carefully balances two crucial facets: the need for accurate predictions (the distance-based ‘accuracy’ of a prediction) and the acknowledgment of the inherent precision or imprecision of such predictions (\textit{i.e.} , how ‘vague’ predictions are).

The aim is to \emph{enable the selection of the most fitting uncertainty-aware machine learning model}, 
aligning with the requirements of specific applications.
{For instance, in crop disease classification where abstention is allowed, practitioners may prefer models that prioritize accuracy even if uncertainty is higher, because uncertain cases can be sent for manual review; whereas in autonomous driving, where abstention is not an option and decisions must be made quickly, models that produce more decisive predictions with lower uncertainty are favored to ensure timely and safe actions.}
The objective here, is \emph{not} to propose yet another measure of uncertainty.

Such a unified framework (Fig. \ref{fig:simplex_KL}) harnesses the fact that predictions generated by Bayesian, Ensemble,
Evidential, Deterministic
and Imprecise-probabilistic models
can all be represented as 
\emph{credal sets}, \textit{i.e.} , convex sets of probability vectors (Sec. \ref{sec:preds_to_credal_set}), within the 
simplex of all probability distributions defined on the label space,
ensuring compatibility across various model architectures and types of uncertainty model.

\begin{figure}
\vspace{-8mm}
    \centering
    \hspace*{-0.6cm}
\includegraphics[width=0.7\textwidth]{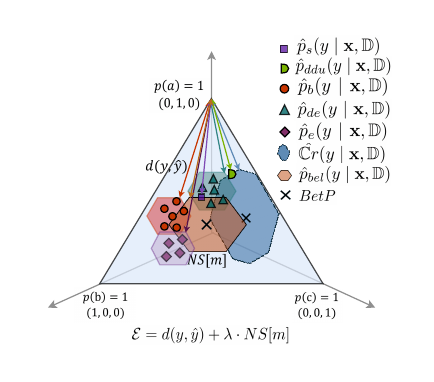}
    \vspace{-26pt}
    \caption{
    Different types of uncertainty-aware model predictions, shown in a unit {simplex} of probability distributions defined on the list of classes 
    $\mathbf{Y}= \{a, b, c\}$.
    The proposed evaluation framework uses a metric which combines, for each input $\mathbf{x}$, a distance (arrows) 
    between the corresponding ground truth (\textit{e.g.} , $(0,1,0)$) and the \textit{epistemic predictions} generated by the various models (in the form of credal sets), and a measure of the extent of the credal prediction (\emph{non-specificity}).
    }
    \label{fig:simplex_KL}
    \vspace{-2mm}
\end{figure}

\section{Key Contributions}

\begin{itemize}[label=\textbullet, leftmargin=*]
  \setlength\itemindent{10pt}
  \setlength\parindent{0pt}  

\item \textbf{Firstly,} a novel versatile \textit{{evaluation}} framework for 
\textit{{ranking}} uncertainty-aware predictions produced by a wide range of models is proposed
to facilitate model selection under uncertainty.

\item \textbf{Secondly,} instrumental to the above, a method for transforming 
predictions to credal sets in the prediction simplex is outlined.

\item \textbf{Thirdly,} a novel metric designed to assess the trade-off between accuracy and precision of predictions is proposed and shows that such metric reduces to the standard KL divergence for point-wise predictions. 

\item \textbf{Finally,} an experimental validation of the 
soundness
of the evaluation framework is provided, demonstrated 
through suitable ablation studies, especially on the main parameter determining the trade-off between accuracy and precision.
\end{itemize}

\section{Methodology: Mapping Predictions to  Credal Sets}
\label{sec:preds_to_credal_set}

Credal and belief function models directly output predictions in the form of credal sets. Traditional networks and Bayesian BMA also generate a credal set: the trivial one containing a single probability vector. For Bayesian, Ensemble, and Evidential models, we utilize the multiple predictions generated directly from the model, prior to any averaging, for a more faithful representation of the predictions.
Below, we explain how predictions from Bayesian, Ensemble and Evidential models can also be represented by credal sets, leveraging \emph{coherent lower probabilities}. 

\subsection{Coherent Lower Probabilities}
\label{app:coherent-theory}

\textbf{Coherent lower probabilities} 
\citep{pericchi1991robust, miranda2023inner, miranda2021selection, miranda2008survey} 
model partial information about a
probability distribution.
Namely, 
a \emph{lower probability} is a function $\underline{P}$ 
from the power set $\mathbb{P}(\mathbf{Y})$ of all subsets of $\mathbf{Y}$ into
$[0,1]$, that 
(i) is monotone, \textit{i.e.} , $\underline{P}(A) \leq \underline{P}(B)$ for all $A \subseteq B$; (ii)
satisfies 
$\underline{P}(\emptyset) = 0$ and $\underline{P}(\mathbf{Y}) = 1$. 

Given such a lower probability $\underline{P}$, the associated credal set $\mathbb{C}r(\underline{P})$ comprises all probability measures in $\mathcal{P}(\mathbf{Y})$ that 
have it as lower bound: 
\begin{equation} \label{eq:lower_credal}
\mathbb{C}r(\underline{P}) = \{ P \in \mathcal{P}(\mathbf{Y}) \mid P(A) \geq \underline{P}(A) \quad  \forall \ A \subseteq \mathbf{Y} \}.
\end{equation}
$\underline{P}$ is considered \emph{coherent} if it can be computed as:
$\underline{P}(A) = \min_{P \in \mathbb{C}r(\underline{P})} P(A) \quad \forall A \subseteq \mathbf{Y}$.
From (Eq. \ref{eq:consistent}), belief functions are also coherent lower probabilities.

A strong theoretical underpinning for reasoning with coherent lower probabilities (and, therefore, the corresponding credal sets) is that it allows us to comply with the coherence principle. In a Bayesian context, individual predictions (such as those of networks with specified weights) can be interpreted as subjective pieces of evidence about a fact (\textit{e.g.} , what is the true class of an input observation). Coherence ensures that one realises the full implications of such partial assessments.

For instance, assume that the available evidence is that on a {ternary classification problem is that $\underline{P}(0) = 1/4$, $\underline{P}(1) = 1/4$,  $\underline{P}(2) = 1/4$,} $\underline{P}(\{0,1\}) = 1/4$. In the behavioural interpretation of probability \citep{walley91book}, in which lower probability of an event $A$ is the upper bound to the price one is willing to pay for betting on outcome $A$, this means that one is willing to pay up to 1/4 for betting on either 0 or 1. But then one should be willing to bet 1/4 + 1/4 = 1/2 for betting on $\{0,1\}$, which is over the specified lower probability for it. Hence, the above specifications are incoherent. Incoherence means that lower probabilities (specified prices) on some events effectively imply lower probabilities on other events which are incompatible.

Now, learning a coherent lower probability from sample predictions does ensure coherence. This is explained, for instance, in \citep{bernard2005introduction}. The concept was originally introduced by Walley in his general treatment of imprecise probabilities \citep{walley91book}. Further, as observed, \textit{e.g.}  in \citep{caprio2024credal}, working with credal sets allows us to hedge against model misspecification, when the set of probability distributions considered by a statistician does not include the distribution that generated the observed data.

In some cases, it may be useful to consider the conjugate function of a lower probability, referred to as the upper probability, $\overline{P}(A) = 1 - \underline{P}(A^c)$ for every $A \subseteq \mathbf{Y}$. The upper and lower probabilities provide upper and lower bounds respectively for the true distribution $\mathbf{P}(A) \in \mathbb{C}r$. Notably, for any coherent lower probability $\underline{P}$, its conjugate upper probability $\overline{P}$ satisfies:
\begin{equation}
    \overline{P}(A) = \max_{P \leq \overline{P}} P(A) \quad \forall \ A \subseteq \mathbf{Y}
\end{equation}
This equation underscores the equivalence between the probabilistic information conveyed by the lower and upper probabilities, indicating that working with either one suffices.


\subsection{Computing lower probabilities from a sample of
probability vectors}

Given the multiple probability vectors predicted by
Bayesian, Ensemble or Evidential models, 
a lower probability can 
be obtained by computing the probability of any event $A$ (as the sum of the probabilities of the elements of A) for each probability vector and then taking the minimum of such $P(A)$ over the samples. 
When $|\mathbf{Y}| = N$ is large, it can be highly inefficient to compute probabilities for all $2^N$ events in the powerset $\mathbb{P}(\mathbf{Y})$. 
We thus
adopt the same budgeting technique used by RS-NN (Sec. \ref{sec:rsnn-budgeting}, Chapter \ref{ch:rsnn}) that provides 
a fixed budget of most relevant
subsets $A \in \mathbb{P}(\mathbf{Y})$, and compute lower probabilities only for those.


\subsection{Computing masses by Moebius inverse}

To {efficiently handle the credal set (Eq. \ref{eq:lower_credal}), we can} compute its vertices.
This can be done via the \emph{Moebius inverse}
\begin{equation}\label{eq:mobius_lower_prob}
    m_{\underline{P}}(A) = \sum_{B\subseteq A}{(-1)^{|A\setminus B|} {\underline{P}}(B)} \quad \forall A \subseteq \mathbf{Y},
\end{equation}
whose output is a \emph{mass function} 
\citep{shafer1976mathematical}, \textit{i.e.} , a set function \citep{denneberg99interaction} $m : \mathbb{P}(\mathbf{Y}) \rightarrow [0,1]$ such that $m (\emptyset) = 0$, $\sum_{A\subset\Theta} m(A) = 1$. {The Moebius inverse of a coherent lower probability that is not a belief function may yield negative values. To ensure a coherent lower probability, we set any negative mass to zero and add the universal set (i.e., the full label set $\mathbf{Y}$) as an additional focal element. This set is assigned the remaining mass needed to ensure that the total mass sums to 1. This guarantees that the resulting mass function is valid and normalized, even when starting from a lower probability that is not itself a belief function.}




\subsection{Computing the Credal Set of Predictions}\label{app:approx-credal}\label{sec:credal_set}

Once mass functions are computed, 
we can compute 
the vertices of the credal set $\mathbb{C}r(\underline{P})$ (Eq. \ref{eq:lower_credal}) identified by the computed lower probabilities.
Such a special credal set
has, as vertices, all the distributions $p^\pi$ induced by a permutation 
$\pi = \{ x_{\pi(1)}, \ldots, x_{\pi(|\mathbf{Y}|)} \}$ of the elements of the sample space $\mathbf{Y}$, 
of the form \citep{Chateauneuf89,cuzzolin2008credal} 
\begin{equation} \label{eq:prho}
p^\pi (x_{\pi(i)}) = \sum_{\substack{A \ni x_{\pi(i)}; \; A \not\ni x_{\pi(j)} \; \forall j<i}} m_{\underline{P}}(A).
\end{equation}
However, we do not need to compute all such vertices; instead, an approximation using only 
a subset of vertices
is often sufficient and even preferable. With this approximation, the method can be applied to datasets with larger number of classes.
\textbf{Approximate computation}. 
Given the prohibitively high computational complexity associated with computing permutations as in Eq. \ref{eq:prho}, especially for a large number of classes, we propose an approximation technique that offers comparable effectiveness without the need to compute all vertices.
For each class $c_i$, $i = 1, 2, \ldots, N$ in classes $\mathcal{C}$ where $N$ is the number of classes, we create two permutations. In the first permutation, $c_i$ is placed as the first element, while in the second permutation, $c_i$ is placed as the last element. By doing so, $c_i$ occupies both the first and last positions in the permutations. Due to the nature of extremal probabilities calculation (Eq. \ref{eq:prho}), when a class $c_i$ is placed as the first element in the permutation, it occupies the initial position in the subset, allowing it to be included in the maximal number of focal elements. As a result, the extremal probability for $c_i$ in this scenario is maximized because it contributes to the sum of masses for all the focal elements containing it. 

Conversely, when a class $c_i$ is placed as the last element in the permutation, 
it is excluded from the focal elements containing classes preceding it in the permutation order. Therefore, the extremal probability for $c_i$ is minimised, because it contributes to the sum of masses for fewer focal elements compared to when it is placed as the first element. This approach yields $2 \cdot N$ number of vertices, 
providing an effective approximate approach to computing all vertices.
The vertices of the credal set associated with each model are computed using mass functions as in Eq. \ref{eq:prho}, via the approximation technique detailed in this section. 


The \textbf{overall procedure} is:
(1) Given a sample of predicted probability vectors, the lower probability of each subset of classes $A$ is computed; (2) A mass function is calculated by Moebius inverse (Eq. \ref{eq:mobius_lower_prob}); (3) The vertices of the credal set associated with the lower probability (and the original predictions) are computed.

\section{Methodology: Evaluation of Epistemic Predictions}\label{sec:evaluation}

We propose a novel \emph{unified evaluation framework} for a comprehensive assessment of these uncertainty models for \textit{model selection}, extending beyond those specifically addressed in this paper, and understand the fundamental trade-off between predictive accuracy and the imprecision inherent in uncertainty models.

The evaluation metric (Fig. \ref{fig:simplex_KL}) combines a component measuring the distance in the probability simplex (in particular, the classical {Kullback-Leibler (KL) divergence}) between the ground truth and the prediction, represented by a credal set as explained in Secs. \ref{sec:classes-epistemic} and \ref{sec:credal_set}, with a \emph{non-specificity} measure \citep{dubois1987properties, dubois1993possibility, klir1987we} assessing the `imprecision' of a model, \textit{i.e.} , how far the prediction is from being a precise probability vector.

Classical performance metrics assume a single (max likelihood) class is predicted; thus, they do not act directly on the predicted probability vectors. By doing so, they discard a lot of information. 
We propose, instead, to evaluate predictions 
\emph{as they are} (either credal sets or as single probabilities, as special case).

\subsection{The Metric}\label{sec:metric}

Mathematically, the evaluation metric $\mathcal{E}$ can be expressed, 
for a single data point, as:
\begin{equation} \label{eq:metric}
    \mathcal{E} = d(y,\hat{y}) + \lambda \cdot 
    NS[m],
\end{equation}
where $d$ is the distance, however defined, between the ground truth $y$ and epistemic prediction $\hat{y}$, $\lambda$ is 
a \emph{trade-off parameter}, and 
$NS[m]$
is a non-specificity measure using as input the mass function representation (Eq. \ref{eq:mobius_lower_prob}) of the prediction.
Over a test set, the average of Eq. \ref{eq:metric} is measured. Low values are desirable.

{
Since $d$ and $NS$ are conceptually and dimensionally distinct, neither can be directly compared or combined without scaling. The trade-off parameter 
$\lambda$ serves as a dimension-matching constant that calibrates the relative importance of the non-specificity term with respect to the distance term.}
Adjusting the value of $\lambda$ allows the user to emphasize either the accuracy of predictions or the precision/imprecision of the model determined by the uncertainty estimates, based on their specific requirements or preferences.
In Sec. \ref{sec:exp_trade_off}, we show experimental results on several values of $\lambda$.

{The \textit{sum formulation} intuitively represents a weighted penalty structure, aligning well with many risk or loss functions in decision theory where distinct costs additively contribute to the total expected cost. In contrast, a multiplicative combination can disproportionately amplify the effect of one term when the other is large or near zero, potentially resulting in unstable behavior. }


\subsubsection{Distance Computation}\label{sec:kl}

To compute the distance $d$ between (epistemic) prediction and ground truth, here we adopt, in particular, the classical
Kullback-Leibler (KL) divergence
    $
    D_{KL}(y || \hat{y})=\int_{x} y(x) \log\frac{y(x)}{\hat{y}(x)}\,dx
    $
between the ground truth $y$ and the boundary of the epistemic prediction $\hat{y}$. 
Nevertheless, our framework is agnostic to that: ablation studies on different distance measures (specifically, the Jensen-Shannon (JS) \citep{menendez1997jensen} divergence) and their effect on $\mathcal{E}$ and model rankings are shown in Sec. {\ref{app:ablation-kl}}.

Whatever the chosen distance,
because of the convex nature of credal sets, this reduces to computing the 
distance to the closest vertex of the credal prediction
(Sec. {\ref{app:kl-theory}}).
For standard networks, this amounts to computing the KL (or JS) divergence between the ground truth distribution $y$ and the single prediction. 
For completeness, in our experiments KL is also applied to the pointwise predictions generated by Model Averaging, in the Bayesian, Ensemble and Evidential cases. 

\label{app:kl-theory}

\textbf{Why KL over L2 and other measures?} KL divergence measures the distance between two probability distributions whereas L2 measures the Euclidean distance between points or vectors. KL divergence has a clear interpretation in information theory as the amount of information lost when one distribution approximates the other \citep{shlens2014notes}, i.e, KL can capture the relative differences between probabilities better. Other divergence measures, such as Jensen-Shannon divergence \citep{menendez1997jensen}, are symmetric and handles small probabilities well, but KL is often preferred as it penalizes cases where the approximate distribution assigns probability to regions where the true distribution assigns none. This property ensures that both the distributions closely match in significant regions \citep{ullah1996entropy}, avoiding substantial probability mass in unlikely areas. Therefore, KL divergence can be advantageous over JS divergence, particularly in its sensitivity to differences between distributions, especially when small probabilities are involved. Unlike JS, which is symmetric and bounded, KL divergence can grow unbounded when there is a significant mismatch between the predicted and true distributions. This sensitivity allows KL to emphasize differences that JS might smooth out.

\textbf{Why use KL Divergence to the nearest vertex?} Measuring KL divergence to the nearest vertex of the credal set avoids giving an advantage to point predictions by deterministic models such as CNN. If we were to use the center of mass of a credal set formed by Bayesian predictions instead, there might be cases where the point prediction by CNN might align perfectly with this center. In such a case, the KL divergence would be identical for both the CNN prediction and the Bayesian prediction. This would favor the CNN prediction as their non-specificity is always zero, while the non-specificity of the Bayesian prediction depends on its credal set size. Hence, using the centroid would penalise imprecise predictions whose boundaries may be pretty close to the ground truth. {While better point predictions by chance should not be discounted, focusing on the nearest vertex encourages a balanced consideration of both KL divergence and non-specificity.} Now, if we were to consider the distance to farthest point in the credal set, the scoring would instead be on the basis of how wrong the {most wrong} prediction is. It acts as a {dual \citep{abellan06measures} to} the minimum distance route that we take.

{It is important to clarify that using the nearest vertex is an approximation to measuring distance to the credal set boundary. Any point on the boundary could be considered, but selecting arbitrary boundary points would require conformal optimization methods that may be computationally expensive. Thus, computing the vertices provides a tractable and interpretable proxy.}

\subsubsection{Non-Specificity}\label{sec:ns}

Assuming a mass representation $m$ is available for the prediction (Sec. \ref{sec:credal_set}),
a \emph{non-specificity} measure is used to determine how imprecise the former is: 
the greater the non-specificity, the larger the imprecision. Non-specificity, due to \citet{dubois1987properties}, \citep{dubois1993possibility, klir1987we}, is given by the degree of concentration of the mass 
assigned to focal set $A$ (sets of labels) in the label space $\mathbf{Y}$:
\begin{equation}\label{eq:non_spec}
    NS[m] = \sum_{A \subseteq \mathbf{Y}} {m(A) \log |A|}.
\end{equation}
{Here, $|A|$ denotes the cardinality (number of elements) of the focal set $A$.}
Non-specificity captures epistemic uncertainty, as it is higher for 
predictions associated with larger sets of classes (or
higher mass values for those), signifying lack of confidence in the predicted probability. 
The log function in Eq. \ref{eq:non_spec} grows more slowly than a linear function, providing diminishing returns as the set size increases. 
Non-specificity is also tied to the size of the predicted credal set, another popular measure of epistemic uncertainty \citep{hullermeier2022quantification}; as the credal set increases in size, so does non-specificity.

One of the earlier works by \citet{yager2008entropy} makes the distinction between two types of uncertainty within a credal set: conflict (also known as randomness or discord) and non-specificity. Non-specificity essentially varies with the size of the credal set \citep{kolmogorov1965three}. Since by definition, aleatoric uncertainty refers to the inherent randomness or variability in the data, while epistemic uncertainty relates to the lack of knowledge or information about the system \citep{DBLP:journals/corr/abs-1910-09457}, conflict and non-specificity directly parallel these concepts. These measures of uncertainty are axiomatically justified \citep{bronevich2008axioms}.
Several measures of non-specificity have been proposed over the years \citep{kramosil1999nonspecificity, pal1993uncertainty, huang2014evidence, smarandache2011contradiction, abellan2000non, abellan2005difference}.
Once again, our evaluation metric can be used with any suitable such measure. An ablation study is shown in Sec. {\ref{app:ablation-ns}}.

Note that, for 
pointwise predictions (including those generated by BMA and ensemble averaging) non-specificity goes to zero (as all non-singleton focal elements $|A|>1$ have mass 0, indicating perfect precision) and the credal set collapses to a single point. Thus, for precise predictions, (Eq. \ref{eq:metric}) reduces to the classical KL.

\subsection{Rationale and Interpretation}

The rationale of
Eq. \ref{eq:metric} is that
a `good' model's predictions should 
(on average)
be accurate (close to the ground truth) and specific (exhibit low uncertainty), so that the model can confidently assign probabilities to different outcomes. 
Conversely, a `bad' model is 
one whose
predictions are less accurate and/or have higher uncertainty, resulting in a wider range of possible outcomes with less confidence in their probabilities.

When evaluating models, the distinction between \emph{correct} and \emph{incorrect} predictions is crucial.
For \textbf{\textit{correct predictions}}, the ideal model exhibits low KL divergence to the ground truth and low non-specificity (uncertainty). 
{If a correct prediction is made with high KL, the model's output is far from the ground truth, indicating it may have succeeded by chance or due to overfitting. If NS is high, the model shows indecision even when it gets the answer right, which undermines reliability.}

{
For \textbf{\textit{incorrect predictions}}, the model should display high NS to reflect uncertainty, and a controlled increase in its deviation from the ground truth (low KL). This avoids the undesirable case where the prediction is too close to an incorrect class, such as being at the opposite vertex of the simplex. For this reason, we prefer low KL at all instances.
The worst kind of prediction is an incorrect (high KL) and highly confident (low NS) prediction.
}

{
This logic extends to the \textbf{\textit{out-of-distribution (OoD)}} setting: a reliable model should respond to \textit{unseen} inputs with high NS, and ideally, with KL divergence that increases as the input distribution diverges from training data.}

Here, the best model selected by our algorithm is the one that minimizes the \textit{evaluation metric} $\mathcal{E}$. We show the evaluation metric for correct and incorrect predictions (Tab. \ref{tab:cc_icc}) on CIFAR-10 (Fig. \ref{fig:kl_ns_eval}), MNIST (Fig. \ref{fig:app_kl_ns_eval_mnist}) and CIFAR-100 (Fig. \ref{fig:app_kl_ns_eval_cifar100}). 

{\textbf{\textit{Minimizing $\mathcal{E}$}} balances overconfidence and underconfidence: it penalizes being overly confident (low NS) in wrong predictions (high KL) and overly cautious (high NS) when correct (low KL). Thus, the model is encouraged to avoid being both very wrong (high KL) and highly uncertain (high NS) simultaneously, pushing for controlled uncertainty, \textit{i.e.}, enough uncertainty when wrong, but not too much to make predictions meaningless. However, this can be a drawback of the metric, since in cases like OoD detection where high uncertainty and high error (high KL, high NS) are desirable, the metric may penalize such behaviour, potentially discouraging the model from expressing full uncertainty on unfamiliar or ambiguous inputs.
In these cases, careful tuning of $\lambda$ is required to reflect the presence of OoD data. But if in-distribution and OoD data are mixed without distinction, this is a limitation of the metric and reduces its usefulness.}

\subsection{Model Selection and Scenarios}\label{sec:model-selection}

In Algorithm \ref{alg:model-selection}, we outline the process for model selection based on the evaluation metric $\mathcal{E}$. 
Using the following scenarios, we demonstrate (1) how to select the optimal $\lambda$ for a specific application, (2) its impact on the evaluation metric, and (3) how model selection differs when abstaining from a decision is allowed \textit{vs.} when a decision is mandatory.

\begin{algorithm}
\caption{Model Selection based on $\mathcal{E}$.}
   \label{alg:model-selection}
\begin{algorithmic}
   \State {\bfseries Input:} Model predictions $\hat{y}_K$ for a list of $K$ models, ground truth labels $y$, trade-off parameter $\lambda$.
    \State Select $\lambda$ based on the required precision for the task (high precision requires a high $\lambda$ and vice versa).
   \State {\bfseries Output:} Selected model $K$ with the lowest Evaluation Metric $\mathcal{E}$
   \For{each model $K$}
       \State $\circ$ Obtain predictions $\hat{y}_K$ for all available models on the test set.
       \State Compute lower probabilities $\underline{P}(\hat{y}_K)$ and mass functions $m_{\underline{P}}(\hat{y}_K)$ using (Eq. \ref{eq:mobius_lower_prob}) over the test set.
       \State $\circ$ Compute vertices of the credal set using (Eq. \ref{eq:prho}).
       \State $\circ$ Compute the minimum KL divergence between ground truth and predictions, $d(y,\hat{y}_K) = KL(y||\hat{y}_K)$ over the entire test set.
       \State $\circ$ Compute the non-specificity, $NS[m]$, of predictions on the test set using (Eq. \ref{eq:non_spec}).
       \State $\circ$ Compute $\mathcal{E} = KL(y||\hat{y}_K) + \lambda \cdot NS[m]$.
   \EndFor
   \State Select the model $K$ which has the lowest $\mathcal{E}$. 
\end{algorithmic} 
\end{algorithm}

Consider crop disease classification from drone images, aiming to categorize crops as healthy, bacterial, fungal, or viral. 
Ideally, we require a model with low KL divergence and low non-specificity. But since we can choose to abstain from making a decision, we can allow non-specificity to be high. Therefore, we assign a low value to $\lambda$ as we want to favour the KL divergence more. Finally, we select the model which has lowest $\mathcal{E}$.
If the decision-making process in this scenario required more precision (\textit{i.e.} , decision is mandatory), even at the risk of incorrect predictions, we would increase $\lambda$ to penalize models with higher uncertainty.

Alternatively, consider the following safety-critical example: an autonomous vehicle driving through a crowded urban environment. The vehicle must make split-second decisions about detecting and responding to pedestrians, traffic signals, vehicles, and other objects. In this high-stakes scenario, \textit{abstention} or making no decision is not an option because failing to act could result in accidents, injury, or property damage. 
Here, precision and timeliness in decision-making is crucial, and a model with high uncertainty (non-specificity) cannot be allowed to abstain.
Here, the parameter $\lambda$ 
would need to be set to a higher value to ensure that the model’s non-specificity is minimized. The model must be decisive, even if there is some risk of making incorrect predictions.

This illustrates why our model selection approach is superior to conventional methods that is solely based on accuracy. In crop disease classification, the choice to abstain from decisions allows us to prioritize models that balance KL divergence and non-specificity, leading to \textit{more informed interventions}. In contrast, for autonomous vehicles, where abstention is not an option, our method focuses on minimizing non-specificity and ensuring \textit{decisiveness even under uncertainty}. This approach tailors model reliability and suitability to the specific decision-making context, beyond just accuracy.

\subsection{{Practical utility of the Evaluation metric}}
\label{app:utility}

\textbf{How a practitioner determines the trade-off $\lambda$.} 

To select a suitable $\lambda$, the practitioner will consider the following task requirements:

\begin{itemize}[itemsep=0pt, topsep=2pt]
    \item Is abstention allowed?
    \item Is precision or accuracy more critical?
    \item What are the risks associated with incorrect predictions or abstention?
\end{itemize}

Here, we use the same two examples of \textbf{crop disease classification} \textit{(abstention from decision-making is allowed)}, and \textbf{autonomous driving} \textit{(abstention is not allowed)}. Consider the following toy problem involving four models (A, B, C, and D) and their respective performances under different $\lambda$ values, as shown in Tab. \ref{tab:model_comparison} below.

\begin{table}[!ht]
    \caption{Comparison of Evaluation Metric ($\mathcal{E}$) for Models A, B, C, and D under different values of $\lambda$.}
    \centering
    \resizebox{\linewidth}{!}{
    \begin{tabular}{@{}lccccc@{}}
        \toprule
        \textbf{Model} & \textbf{KL} & \textbf{NS} & \textbf{$\mathcal{E}$(Low $\lambda = 0.1$)} & \textbf{$\mathcal{E}$(Moderate $\lambda = 0.5$)} & \textbf{$\mathcal{E}$(High $\lambda = 2$)} \\ \midrule
        Model A & 0.243 & 0.166 & $0.243 + 0.1 \times 0.166 = 0.259$ & $0.243 + 0.5 \times 0.166 = 0.326$ & $0.243 + 2 \times 0.166 = 0.575$ \\ 
        Model B & 0.031 & 0.385 & $0.031 + 0.1 \times 0.385 = \textbf{0.069}$ & $0.031 + 0.5 \times 0.385 = \textbf{0.223}$ & $0.031 + 2 \times 0.385 = 0.801$ \\ 
        Model C & 0.002 & 2.267 & $0.002 + 0.1 \times 2.267 = 0.228$ & $0.002 + 0.5 \times 2.267 = 1.136$ & $0.002 + 2 \times 2.267 = 4.536$ \\ 
        Model D & 0.398 & 0.009 & $0.398 + 0.1 \times 0.009 = 0.398$ & $0.398 + 0.5 \times 0.009 = 0.402$ & $0.398 + 2 \times 0.009 = \textbf{0.416}$ \\ 
        \bottomrule
    \end{tabular}}
    \label{tab:model_comparison}
\end{table}

\textbf{Ranking:} The model selection algorithm returns the following ranking of models based on the values in Tab. \ref{tab:model_comparison}.

\begin{itemize}[itemsep=0pt, topsep=2pt]
    \item $\lambda$ = 0.1 (\textit{Low}): \textbf{B}, C, A, D
    \item $\lambda$ = 0.5 (\textit{Moderate}): \textbf{B}, A, D, C
    \item $\lambda$ = 2 (\textit{High}): \textbf{D}, A, B, C
\end{itemize}

\textbf{1. Crop disease classification}

Abstention is \textit{allowed}–can afford to abstain from making predictions for uncertain cases (\textit{e.g.} , sending ambiguous images for manual analysis).

\textbf{Why choose a low $\lambda$?} In this scenario, we have the option to abstain from making decisions for uncertain cases (\textit{e.g.} , we can send ambiguous images for manual analysis based on the model's predictions). Therefore, we do not give much {weight} to non-specificity and prefer a model which has low KL, \textit{i.e.} , it can give us a prediction more closer to the ground truth.

Note that our metric is not used for test set evaluation, it is only used for model selection. After the model is selected, we do not scale either of these terms up or down using $\lambda$. The model operates with its inherent uncertainty predictions without any further adjustments.

\textbf{Why not choose a high $\lambda$?} If we choose a high $\lambda$, we penalize models with high uncertainty. As a result, we end up selecting a model that is very precise in its predictions. This prevents us from leveraging the advantage of the abstention allowance.

\textbf{2. Autonomous driving}

Abstention is \textit{not allowed}–decision making is necessary, as abstention can lead to critical failures, such as accidents, harm to pedestrians, or collisions with other vehicles.

\textbf{Why choose a high $\lambda$?} In this scenario, a decision must be made (\textit{e.g.} , when there is a shadow on the road, we cannot afford to be uncertain or abstain; we must decide whether to stop or not). Therefore, we heavily penalize non-specificity. Models with lower non-specificity will rank first using our model selection, because they are more decisive in their predictions.

\textbf{Why not choose a low $\lambda$?} If we choose a low $\lambda$, we prioritize the models with high uncertainty, therefore, end up selecting a model that is very imprecise about its predictions. When there is no option to abstain, we do not want an indecisive model.

\textbf{How does it reflect in the rankings?
}
In the rankings for low $\lambda$ (\textbf{$\lambda$ = 0.1}), Model \textbf{B} ranks first due to its low KL. Conversely, with high $\lambda$ (\textbf{$\lambda$ = 2}), the ranking shifts to prioritize models with low non-specificity. Model \textbf{D} ranks highest because of its minimal non-specificity. This ranking penalizes uncertain models like Models C and B, pushing it to the bottom.

For crop disease classification, choosing a high $\lambda$ in this scenario negates the advantage of abstention and leads to selecting a model that is overly precise at the cost of being potentially less reliable in uncertain cases. However, in autonomous driving, abstention is not an option because hesitation in decision-making can result in accidents. Here, a high $\lambda$ is crucial to prioritize models with low non-specificity, ensuring the model makes more precise and decisive predictions.

\subsection{Evaluation Metric or Uncertainty Measure?}

The evaluation metric is \emph{not} a new uncertainty measure. The first term, the KL divergence, is unrelated to uncertainty and is more related to accuracy, whereas non-specificity \citep{yager2008entropy} is a measure of epistemic uncertainty as it is directly proportional to the size of the credal set \citep{hullermeier2021aleatoric}. Together, with the trade-off parameter $\lambda$, the evaluation metric provides a \textit{holistic assessment of the predictions}.

The value of the metric $\mathcal{E}$ is a function of the credal set. The distance measure $d(y,\hat{y})$ is the distance to a single one-hot probability vector, which has both 0 epistemic uncertainty (being a single point) and 0 aleatoric uncertainty (as it is one-hot). 
Non-specificity was originally introduced as a measure of imprecision in a random set framework in which masses are independently assigned to sets of outcomes. As discussed in \citep{hullermeier2021aleatoric}, Sec. 4.6.1, in the case of credal sets non-specificity can be considered a measure of \textit{epistemic}, rather than total uncertainty.

Therefore, the first term in the evaluation metric, KL divergence has no relation to uncertainty, while, the second term, non-specificity, can be considered a measure of epistemic uncertainty.

\section{Experiments}\label{sec:experiments}

\textbf{Firstly,} we designed a set of baseline experiments to evaluate the behaviour of the proposed composite performance metric (Sec. \ref{sec:analysis}), using a representative for each class of predictions/models. \textbf{Secondly,} we assess how these models are ranked (Sec. \ref{sec:exp_model_selection}).
We also ran \textbf{several ablations studies}: (i) on different measures of divergence (Kullback-Leibler \textit{vs.} Jensen-Shannon) (Sec. {\ref{app:ablation-kl}}), their effect on model selection (Tab. \ref{tab:model_rank_ablation}); (ii) on different non-specificity measures (Sec. {\ref{app:ablation-ns}}, Tab. \ref{tab:model_rank_ablation_ns}), and why our choices of KL and NS are better; (iii) on the trade-off parameter $\lambda$, to understand its effect on the ranking (Sec. \ref{sec:exp_trade_off}); (iv) on the effect of the number of sample predictions produced by Bayesian and Ensemble models (Sec. {\ref{sec:exp_ablation_samples}}).

\subsection{Implementation}
\label{sec:implementation-details}

\subsubsection{Datasets}

We use MNIST \citep{LeCun2005TheMD}, CIFAR-10 \citep{cifar10} and CIFAR-100 \citep{krizhevsky2009learning} as datasets. The data is split into 40000:10000:10000 samples for training, testing, and validation respectively for CIFAR-10 and CIFAR-100, and 50000:10000:10000 samples for MNIST.

\subsubsection{Baselines and Backbones}

To assess the behaviour of our evaluation metric we adopted the following baselines: (1) Traditional Neural Network (\textbf{CNN}) with no uncertainty estimation, (2) Laplace Bridge Bayesian Neural Network (\textbf{LB-BNN}) \citep{hobbhahn2022fast}, 
(3) Deep Ensemble (\textbf{DE}) \citep{lakshminarayanan2017simple},
(4) Evidential Deep Learning (\textbf{EDL}) \citep{sensoy}, (5) Deep Deterministic Uncertainty (\textbf{DDU}) \citep{mukhoti2023deep},
(6) Credal-Set Interval Neural Networks (\textbf{CreINN}) (Sec. \ref{sec:creinn}) \cite{wang2024creinns}, (7) Evidential Convolutional Neural Network (\textbf{E-CNN}) \citep{tong2021evidential}, and (8) Random-Set Neural Network (\textbf{RS-NN}) (Chapter \ref{ch:rsnn}).

{EDL} \citep{sensoy} predictions are obtained by sampling from the posterior Dirichlet distribution, and \textbf{DDU} \citep{mukhoti2023deep} predictions are softmax probabilities over classes, similar to traditional neural network \textbf{CNN}.
{CreINNs} \citep{wang2024creinns} retain the structure of traditional Interval Neural Networks to generate interval probabilities, lower and upper probabilities for each prediction, and formulate credal sets from these interval predictions. \textbf{E-CNN} is an evidential uncertainty classifier which predicts mass functions for sets of outcomes. Credal sets can directly be generated from these mass functions. ResNet50 backbone is used for all the models.

\subsubsection{Training Details}

The training details for all models are the same as in Sec. \ref{sec:training-rsnn}. We only require a CPU for evaluation of predictions on CIFAR-10 and MNIST. For CIFAR-100, we use an Nvidia A100 40GB GPU. The entire code runs in approximately 70 seconds for CIFAR-10 dataset. The credal uncertainty estimation in ablation study (Sec. {\ref{app:ablation-ns}}) takes an additional 90 seconds. 

\subsubsection{Obtaining Epistemic Predictions}

For LB-BNN, EDL and DE, we present results on both averaged predictions and credal sets generated from multiple prediction samples before averaging (100 prediction samples for LB-BNN and EDL, 15 ensembles for DE). An ablation study on number of samples \textit{vs.} evaluation metric $\mathcal{E}$ for these models is discussed in Sec. {\ref{sec:exp_ablation_samples}}.  
For CreINN, 10 samples were generated per lower and upper probability prediction. 
{E-CNN} directly predicts masses for all possible outcome sets (for CIFAR-10 with 10 classes, $2^{10} = 1024-1 = 1023$ outcomes, $\emptyset$ excluded), making it
{computationally infeasible
for large datasets (e.g, CIFAR-100} where computing $2^{100}$ scores is inefficient). 
RS-NN generates belief functions for a budgeted set of outcomes. On CIFAR-10 and MNIST, it predicts 30 class sets 
(10 classes + 20 subsets), and 300 for CIFAR-100 (100 singletons + 200 subsets).

\subsection{Analysis of the Evaluation Metric}
\label{sec:analysis}

\begin{table*}[!ht]
\caption{Comparison of Kullback-Leibler divergence (KL), Non-Specificity (NS) and Evaluation Metric ($\mathcal{E}$) for uncertainty-aware classifiers (trade-off $\lambda = 1$). Mean and standard deviation are shown for CIFAR-10, MNIST and CIFAR-100 datasets. 
}
\vspace{-8pt}
\label{tab:kl_ns_e}
\centering
\resizebox{\textwidth}{!}{
\begin{tabular}{lcccccc}
\toprule
\multirow{1}{*}{Dataset} &
  \multicolumn{1}{c}{\multirow{1}{*}{Model}}&
  \multicolumn{1}{c}{\multirow{1}{*}{Test accuracy (\%)($\uparrow$)}}&
  \multicolumn{1}{c}{\multirow{1}{*}{ECE ($\downarrow$)}}&
  \multicolumn{1}{c}{\multirow{1}{*}{KL divergence (KL)}}&
  \multicolumn{1}{c}{\multirow{1}{*}{Non-Specificity (NS)}}&
  \multicolumn{1}{c}{\multirow{1}{*}{Evaluation metric ($\mathcal{E}$)($\downarrow$)}} \\
  \midrule
\multicolumn{1}{c}{\multirow{10}{*}{CIFAR-10}}
  & LB-BNN & $89.24$ & $0.0565$ & $0.243 \pm 1.315$ & $0.166 \pm 0.398$ & $0.409 \pm 1.381$\\
  & DE & $\mathbf{93.77}$ & $\mathbf{0.0075}$ & $0.031 \pm 0.367$ & $0.385 \pm 0.715$ & $0.415 \pm 0.805$\\
  & EDL & $59.13$ & $0.0491$ & $0.002 \pm 0.011$ & $2.267 \pm 0.067$ & $ 2.270 \pm 0.066$\\
  & CreINN & $88.36$ & $0.0108$ & $0.058 \pm 0.374$ & $0.596 \pm 0.812$ & $0.654 \pm 0.892$\\
  & E-CNN & $83.5$ & $0.6497$ & $0.193 \pm 0.215$ & $1.609 \pm 0.003$ & $1.802 \pm 0.215$\\
  & RS-NN & $93.53$ & $0.0509$ & $0.398 \pm 1.895$ & $0.009 \pm 0.052$ & $\mathbf{0.407} \pm \mathbf{0.500}$\\
  \cmidrule{2-7}
  & CNN & $90.25$ & $0.0668$ & $0.481 \pm 1.797$ & $0.000 \pm 0.000$ & $0.481 \pm 1.797$\\
  & LB-BNN Avg & $89.24$ & $0.0565$  & $0.420 \pm 1.520$ & $0.000 \pm 0.000$ & $0.420 \pm 1.520$\\
   & DE Avg & $\mathbf{93.77}$ & $\mathbf{0.0075}$  & $0.195 \pm 0.763$ & $0.000 \pm 0.000$ & $\mathbf{0.195} \pm \mathbf{0.763}$\\
   & DDU & $91.34$ & $0.0439$ & $0.309 \pm 1.115$ & $0.000 \pm 0.000$ & $0.309 \pm 1.115$\\
  \midrule
\multicolumn{1}{c}{\multirow{10}{*}{MNIST}}
  & LB-BNN & $99.55$ & $0.0018$ & $0.002 \pm 0.126$ & $0.091 \pm 0.380$ & $0.093 \pm 0.401$\\
  & DE  & $99.32$ & $\mathbf{0.0012}$ & $0.002 \pm 0.072$ & $0.067 \pm 0.320$ & $0.070 \pm 0.331$\\
  & EDL & $94.42$ & $0.2418$ & $0.00007 \pm 0.002$ & $2.260 \pm	0.054$ & $2.260 \pm 0.054$\\
  & CreINN & $98.23$ & $0.0105$ & $0.071 \pm 0.609$ & $0.005 \pm 0.043$ & $0.077 \pm 0.612$\\
  & E-CNN & $99.27$ & $0.7878$ & $0.037 \pm 0.065$ & $1.608 \pm 0.004$ & $1.645 \pm 0.064$\\
  & RS-NN & $\mathbf{99.71}$ & $0.0059$ & $0.053 \pm 0.740$ & $0.001 \pm 0.016$ & $\mathbf{0.054} \pm \mathbf{0.741}$\\
  \cmidrule{2-7}
 & CNN & $98.90$ & $0.0057$ & $0.043 \pm 0.497$ & $0.000 \pm 0.000$ & $0.043 \pm 0.497$\\
    & LB-BNN Avg & $99.55$ & $0.0018$ & $0.016 \pm 0.251$ & $0.000 \pm 0.000$ & $\mathbf{0.016} \pm \mathbf{0.251}$\\
 & DE Avg & $99.32$ & $\mathbf{0.0012}$ & $0.020 \pm 0.198$ & $0.000 \pm 0.000$ & $0.020 \pm 0.198$\\
 & DDU & $99.28$ & $0.0028$ & $0.028 \pm 0.336$ & $0.000 \pm 0.000$ & $0.028 \pm 0.336$\\
  \midrule
  \multicolumn{1}{c}{\multirow{10}{*}{CIFAR-100}}
  & LB-BNN & $71.34$ & $0.1332$ & $0.146 \pm 0.504$ & $2.348 \pm 1.771$ & $2.494 \pm 1.781$\\
  & DE  & $\mathbf{74.08}$ & $\mathbf{0.0377}$ & $0.019 \pm 0.245$ & $3.182 \pm 1.909$ & $3.201 \pm 1.906$\\
  & EDL & $45.76$ & $0.3558$ & $0.010 \pm	0.192$ & $3.434 \pm 1.843$ & $3.445 \pm 1.840$\\
  & CreINN & $44.30$ & $0.1831$ & $0.723 \pm 0.646$ & $2.050 \pm	1.188$ & $2.774 \pm 0.945$\\
  & E-CNN & - & - & - & - & -\\
  & RS-NN & $71.17$ & $0.1336$ & $1.518 \pm 3.966$ & $0.569 \pm 1.164$ & $\mathbf{2.088} \pm \mathbf{4.025}$\\
  \cmidrule{2-7}
    & CNN & $65.51$ & $0.2357$ & $2.293 \pm 4.199$ & $0.000 \pm 0.000$ & $2.293 \pm 4.199$\\
  & LB-BNN Avg & $71.34$ & $0.1332$ & $1.617 \pm 2.886$ & $0.000 \pm 0.000$ & $1.617 \pm 2.886$\\
 & DE Avg &  $\mathbf{74.08}$ & $\mathbf{0.0377}$ & $1.062 \pm 1.924$ & $0.000 \pm 0.000$ & $\mathbf{1.062} \pm \mathbf{1.924}$\\
 & DDU & $73.44$ &$0.1142$ & $1.180 \pm 2.260$ & $0.000 \pm 0.000$ & $1.180 \pm 2.260$ \\
\bottomrule
\end{tabular}
} 
\vspace{-3mm}
\end{table*}

Tab. \ref{tab:kl_ns_e} shows the 
mean and standard deviation of
Kullback-Leibler divergence (KL), 
Non-Specificity (NS) (Eq. \ref{eq:non_spec}), Evaluation Metric ($\mathcal{E}$) (Eq. \ref{eq:metric}), test accuracy (\%) and Expected Calibration Error (ECE) for 
the chosen
LB-BNN, DE, EDL, CreINN, E-CNN and RS-NN models. 
We also report these for models with point predictions only, \textit{i.e.} , CNN, Bayesian Model Averaged LB-BNN ({LB-BNN Avg}), averaged ensemble predictions ({DE Avg}) and DDU, for trade-off $\lambda = 1$.

Tab. \ref{tab:kl_ns_e} shows that for CIFAR-10, DE outperforms all other models with a test accuracy of $93.77\%$, a low ECE, while DE Avg and RS-NN has the lowest $\mathcal{E}$ score. Similarly, in MNIST, RS-NN achieves the highest accuracy and the LB-BNN Avg and RS-NN has the lowest $\mathcal{E}$. Conversely, for CIFAR-100, DE has a higher test accuracy of $74.08\%$, but RS-NN has the lowest $\mathcal{E}$ score. The performance is markedly low for EDL and CreINN, with E-CNN showing no results due to computational infeasibility. This model selection in Tab. \ref{tab:kl_ns_e} is reflected in the ranking for $\lambda = 1$ in Tab. \ref{tab:model_rank}.

\begin{figure*}[!ht]
    \vspace{-8pt}
    \begin{minipage}[t]{0.49\textwidth}
    \includegraphics[width=1.02\textwidth]{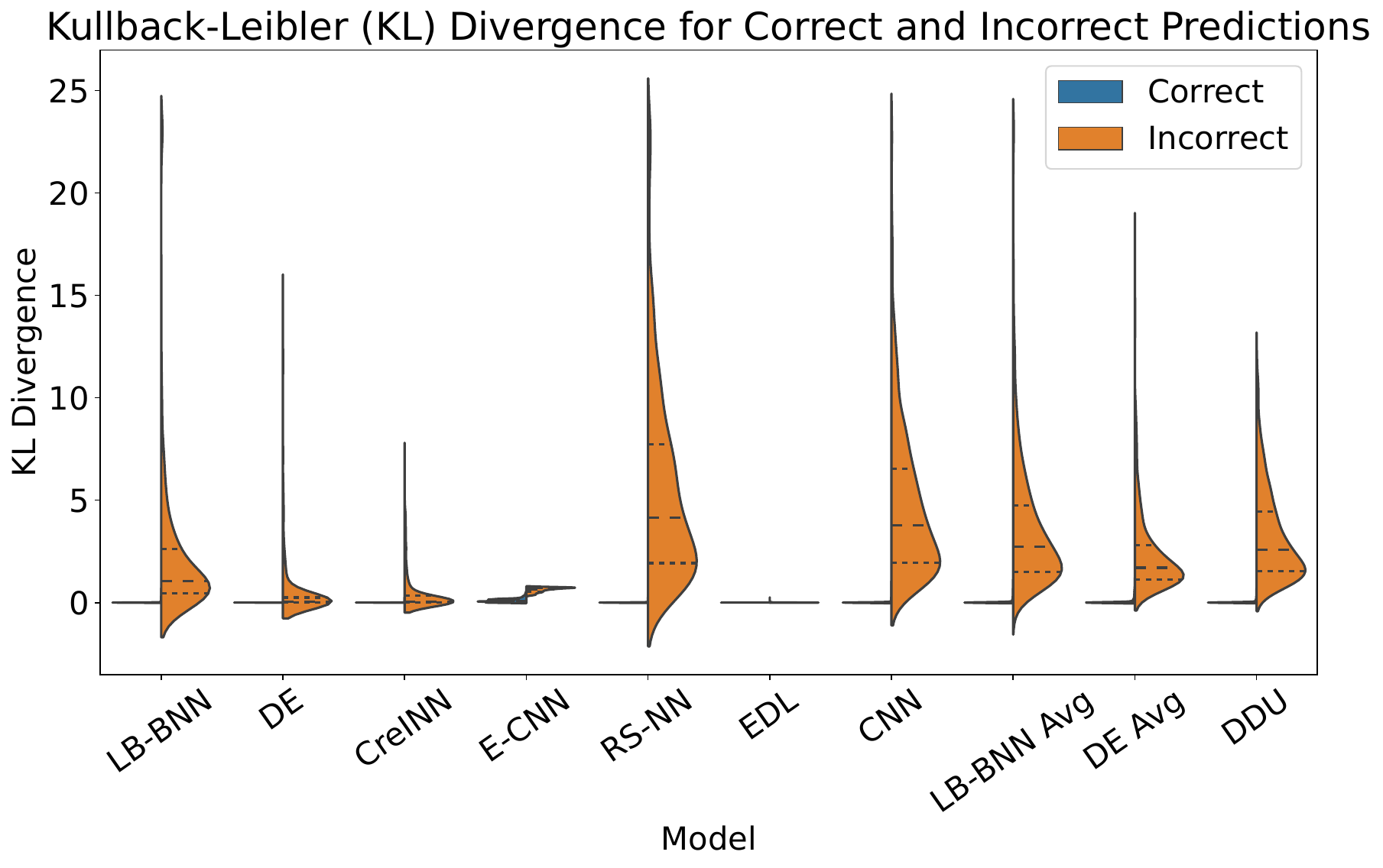}
    \end{minipage}\hspace{0.001\textwidth}
    \begin{minipage}[t]{0.48\textwidth}
    \includegraphics[width=\textwidth]{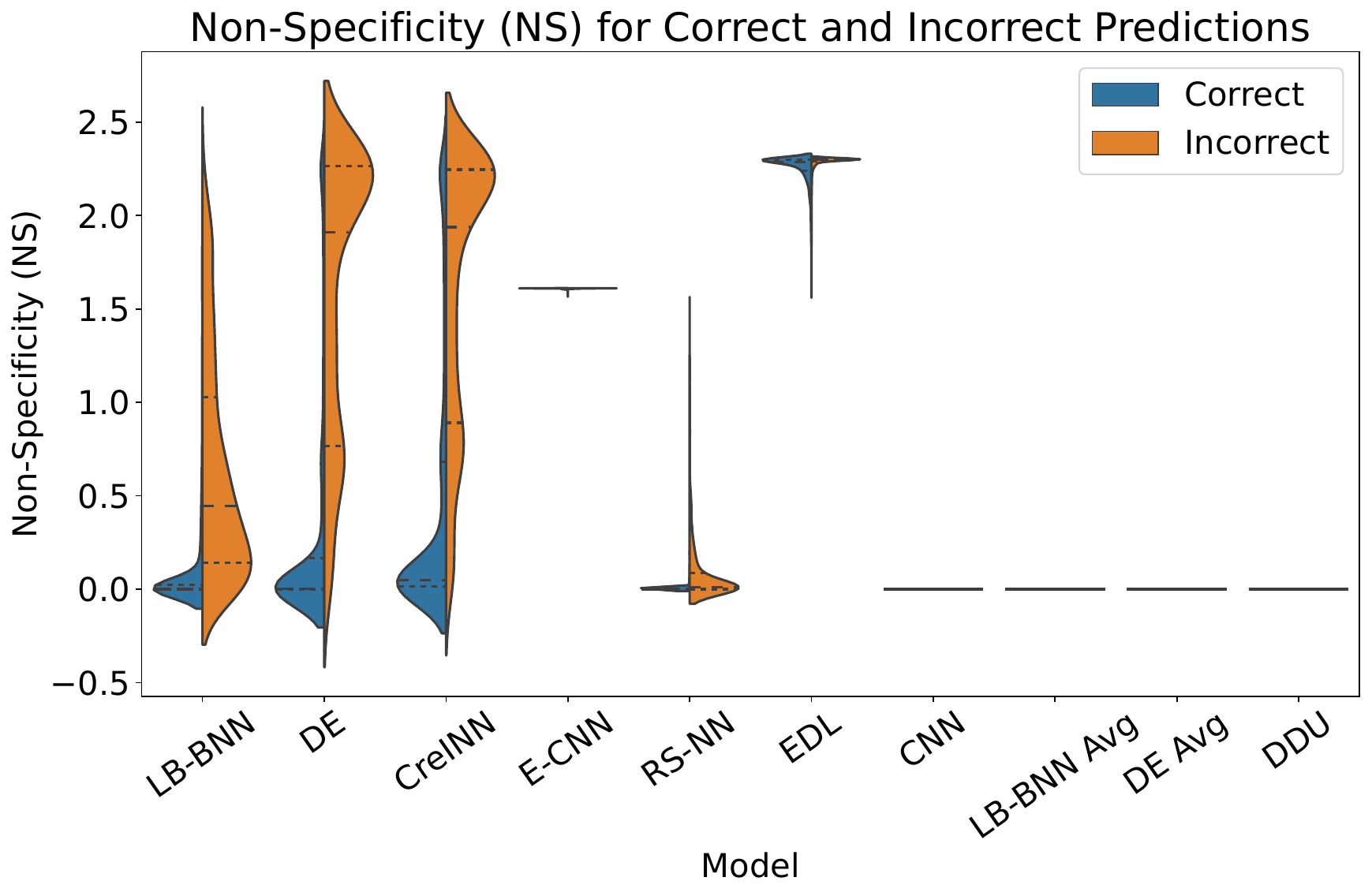}
    \end{minipage}\hspace{0.02\textwidth}
            \vspace{-7mm}
    \begin{minipage}[t]{0.47\textwidth}
    \includegraphics[width=1.02\textwidth]{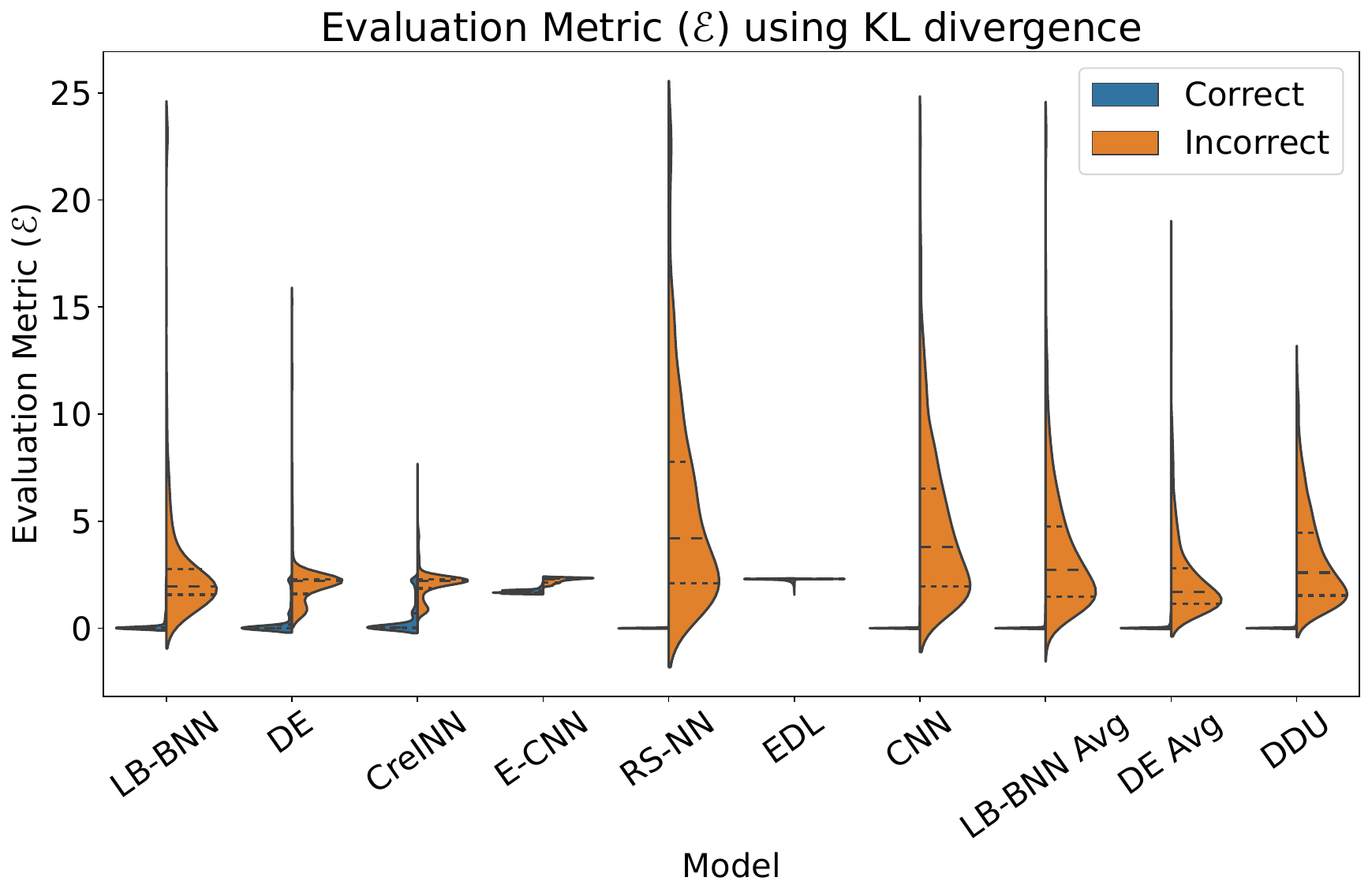}
    \end{minipage} \hspace{0.001\textwidth}
            \vspace{-5mm}
    \begin{minipage}[t]{0.51\textwidth}
    \vspace{-145pt}
    \includegraphics[width=\textwidth]{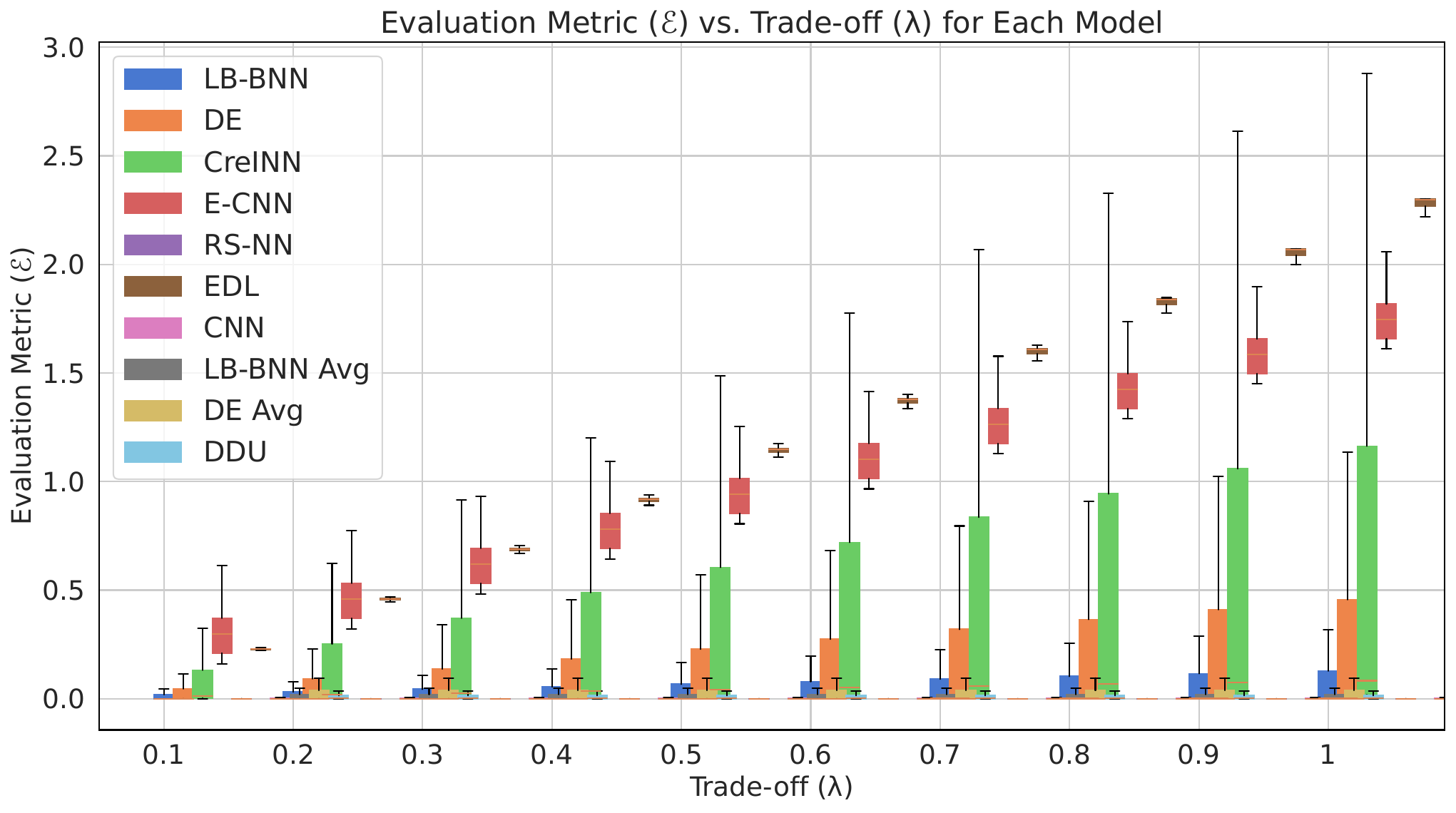}
    \vspace{-2pt}
    \end{minipage}
    \vspace{4mm}
    \caption{Measures of 
    KL divergence (a)\textit{(top left)}, (b) Non-specificity \textit{(top right)}, (c) Evaluation Metric \textit{(bottom left)} for both Correctly (CC) and Incorrectly Classified (ICC) samples from \textbf{CIFAR-10}, and (d) Evaluation metric \textit{vs.} trade-off parameter \textit{(bottom right)}, for all models, on the CIFAR-10 dataset.
    }
    \label{fig:trade_off}
    \label{fig:kl_ns_eval}
    \vspace{-3mm}
\end{figure*}

\begin{figure}[!h]
\centering
    \begin{minipage}[t]{0.49\textwidth}
    \includegraphics[width=1.02\textwidth]{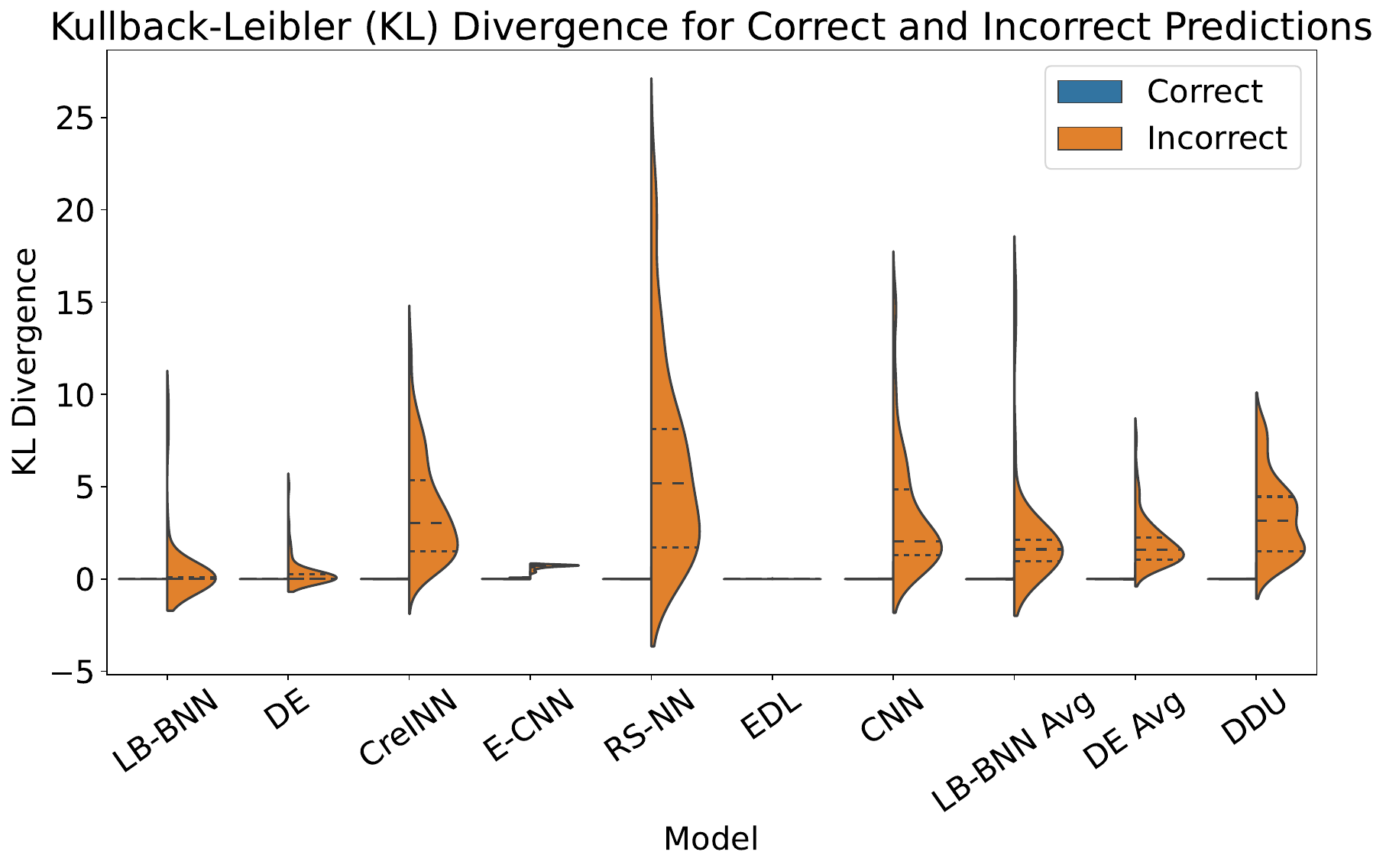}
    \vspace{-8mm}
    \label{fig:mnist_kl}
    \end{minipage}\hspace{0.001\textwidth}
    \begin{minipage}[t]{0.48\textwidth}
    \includegraphics[width=\textwidth]{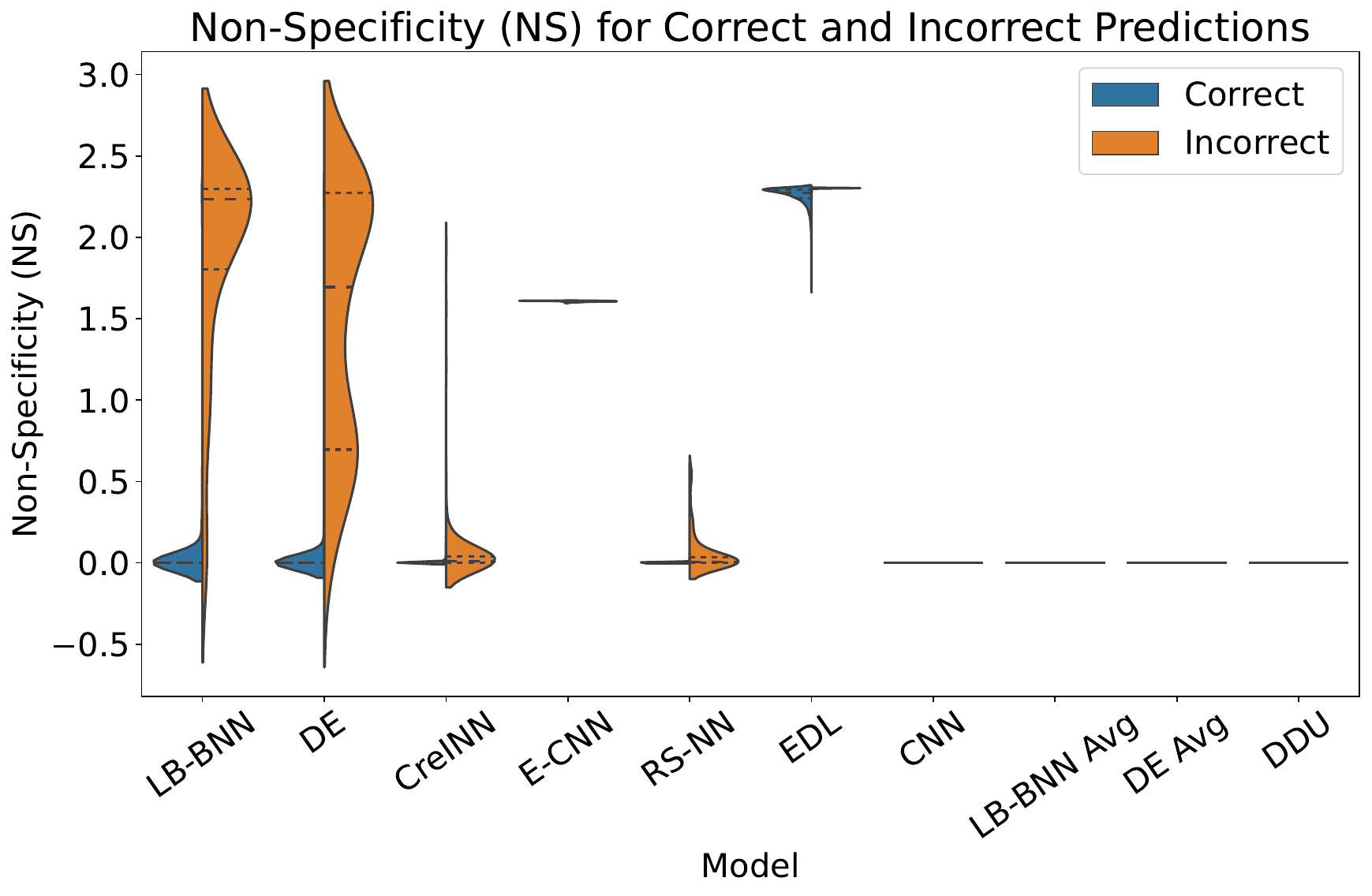}
        \vspace{-8mm}
    \label{fig:mnist_ns}
    \end{minipage}
        \begin{minipage}[t]{0.47\textwidth}
    \vspace{-46mm}
    \includegraphics[width=1.02\textwidth]{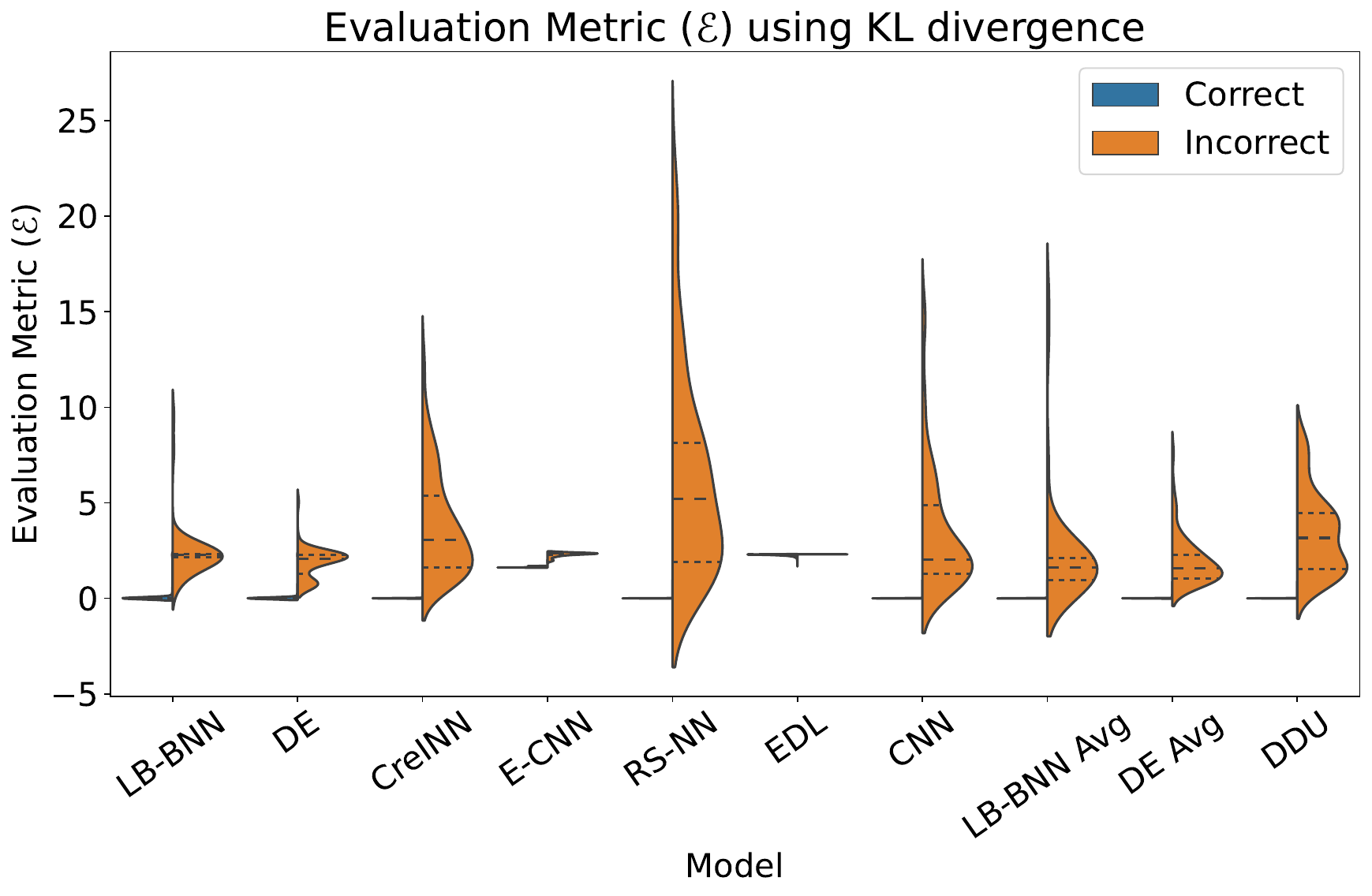}
        \vspace{-8mm}
    \label{fig:mnist_e}
    \end{minipage}
    \begin{minipage}[b]{0.51\textwidth}
    \vspace{-3mm}
    \includegraphics[width=\textwidth]{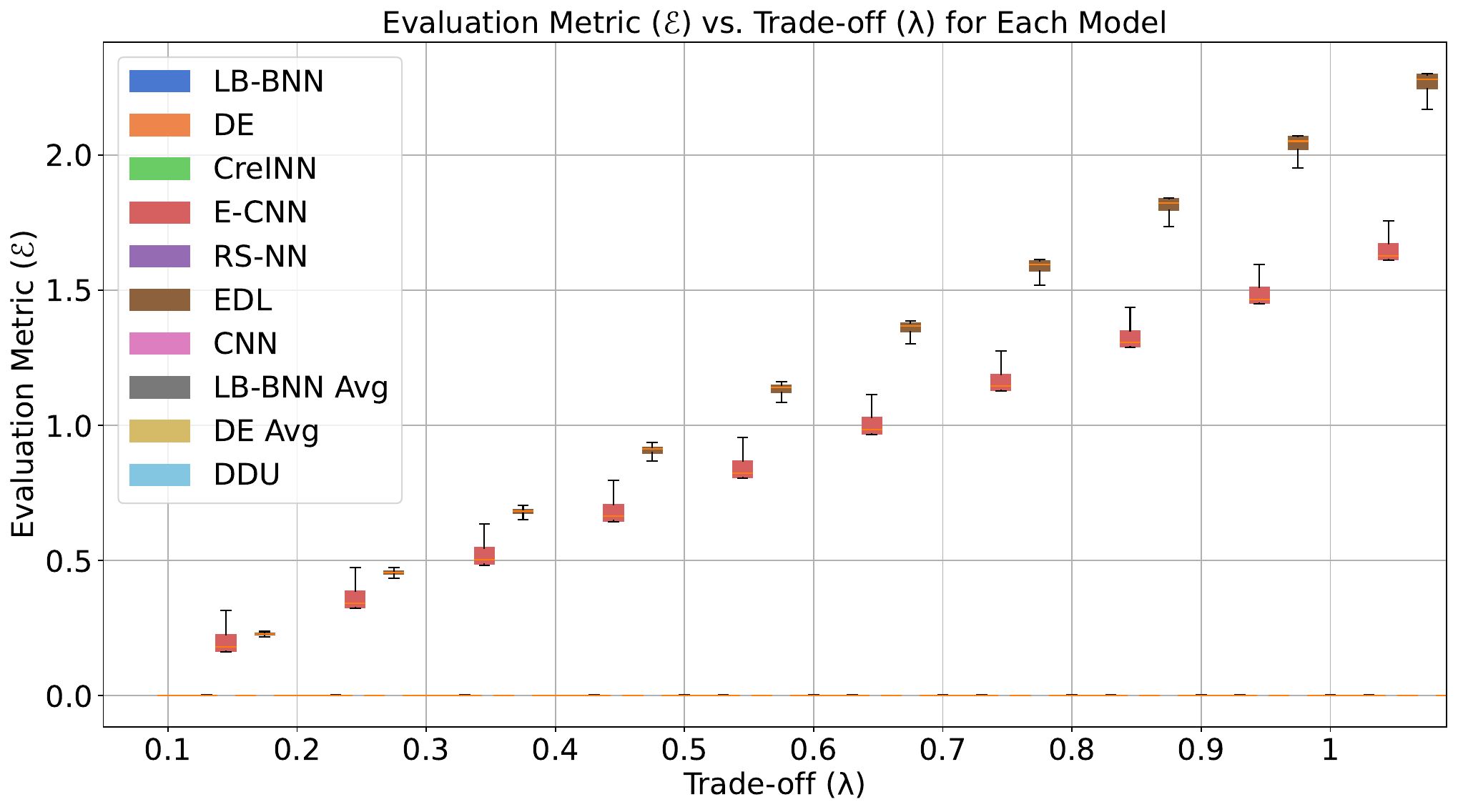}
        \vspace{-5mm}
    \label{fig:mnist_lambda}
\end{minipage} 
        \vspace{-1mm}
    \caption{Measures of (a)\textit{(top left)} Kullback-Leibler (KL) divergence, (b)\textit{(top right)} Non-specificity (NS), (c)\textit{(bottom left)} Evaluation Metric ($\mathcal{E}$) for Correctly Classified (CC) and Incorrectly Classified (ICC) samples (d)\textit{(bottom right)} Evaluation metric ($\mathcal{E}$) estimates \textit{vs.} trade-off ($\lambda$) for all models on \textbf{MNIST}.
    }\label{fig:app_kl_ns_eval_mnist}
            \vspace{-5mm}
\end{figure}

\begin{figure}[!h]
\centering
    \begin{minipage}[t]{0.49\textwidth}
    \includegraphics[width=1.03\textwidth]{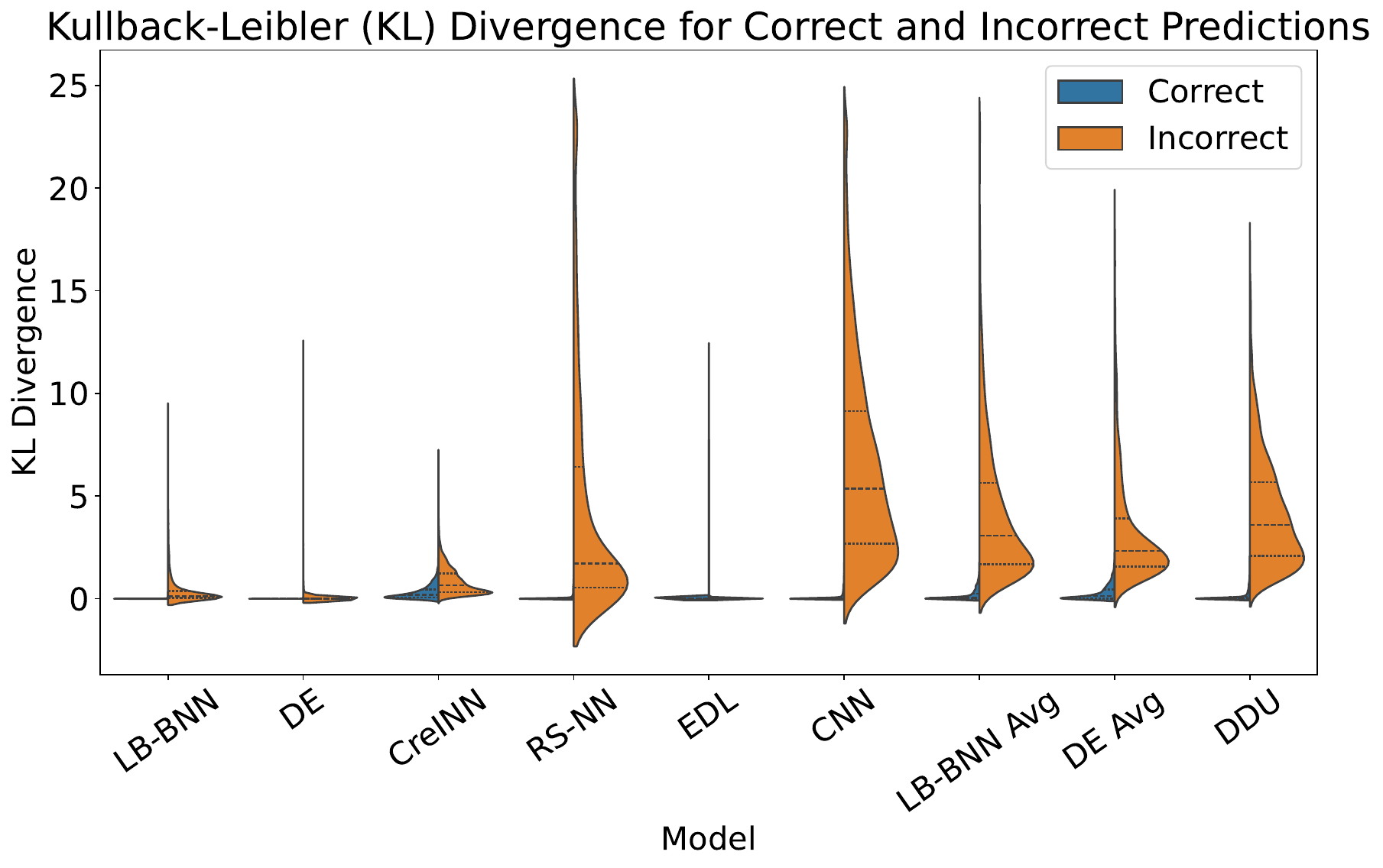}
    \vspace{-8mm}
    \label{fig:cifar100_kl}
    \end{minipage}\hspace{0.001\textwidth}
    \begin{minipage}[t]{0.49\textwidth}
    \includegraphics[width=\textwidth]{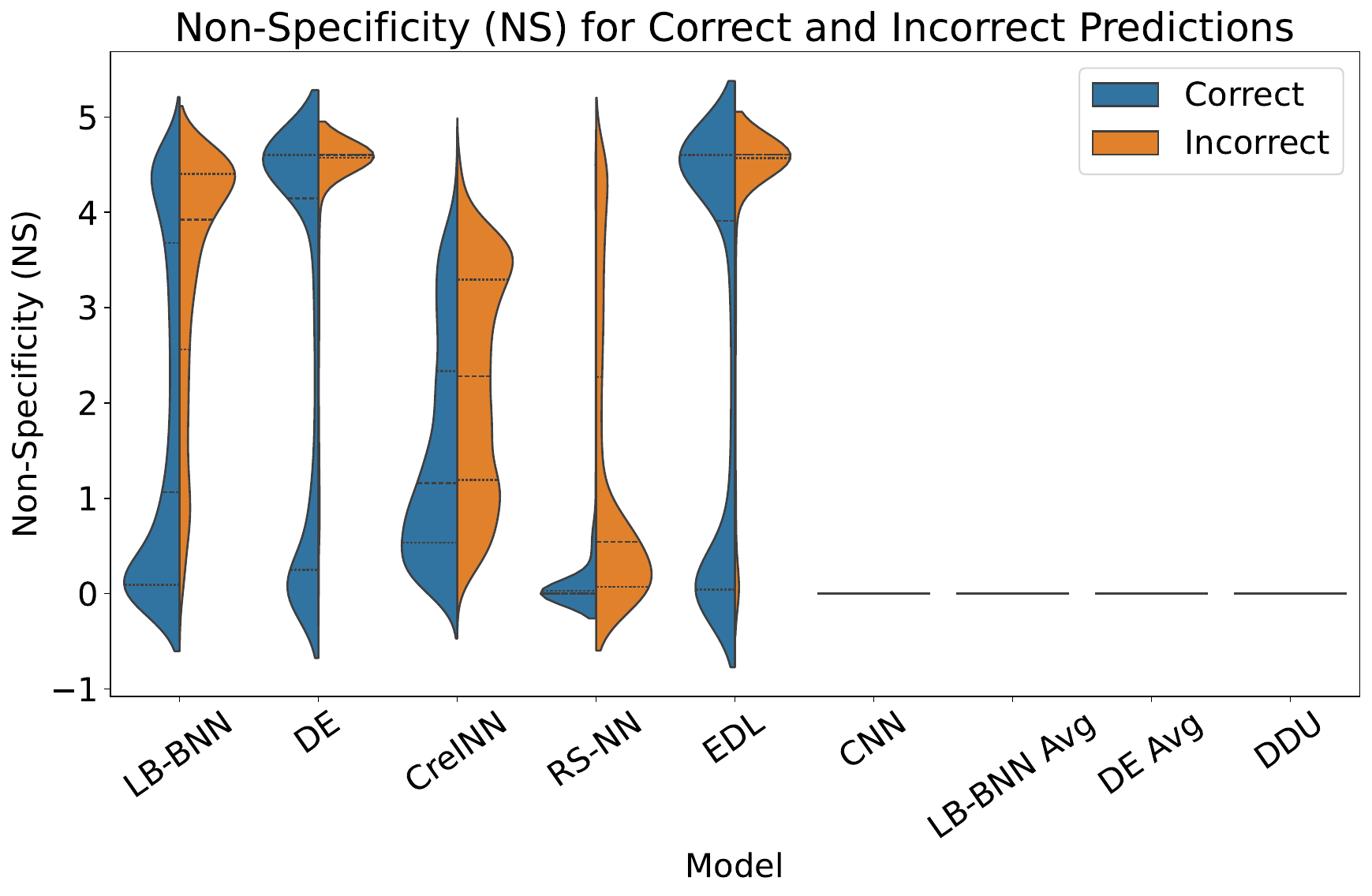}
        \vspace{-8mm}
    \label{fig:cifar100_ns}
    \end{minipage}
        \begin{minipage}[t]{0.52\textwidth}
    \includegraphics[width=1.1\textwidth]{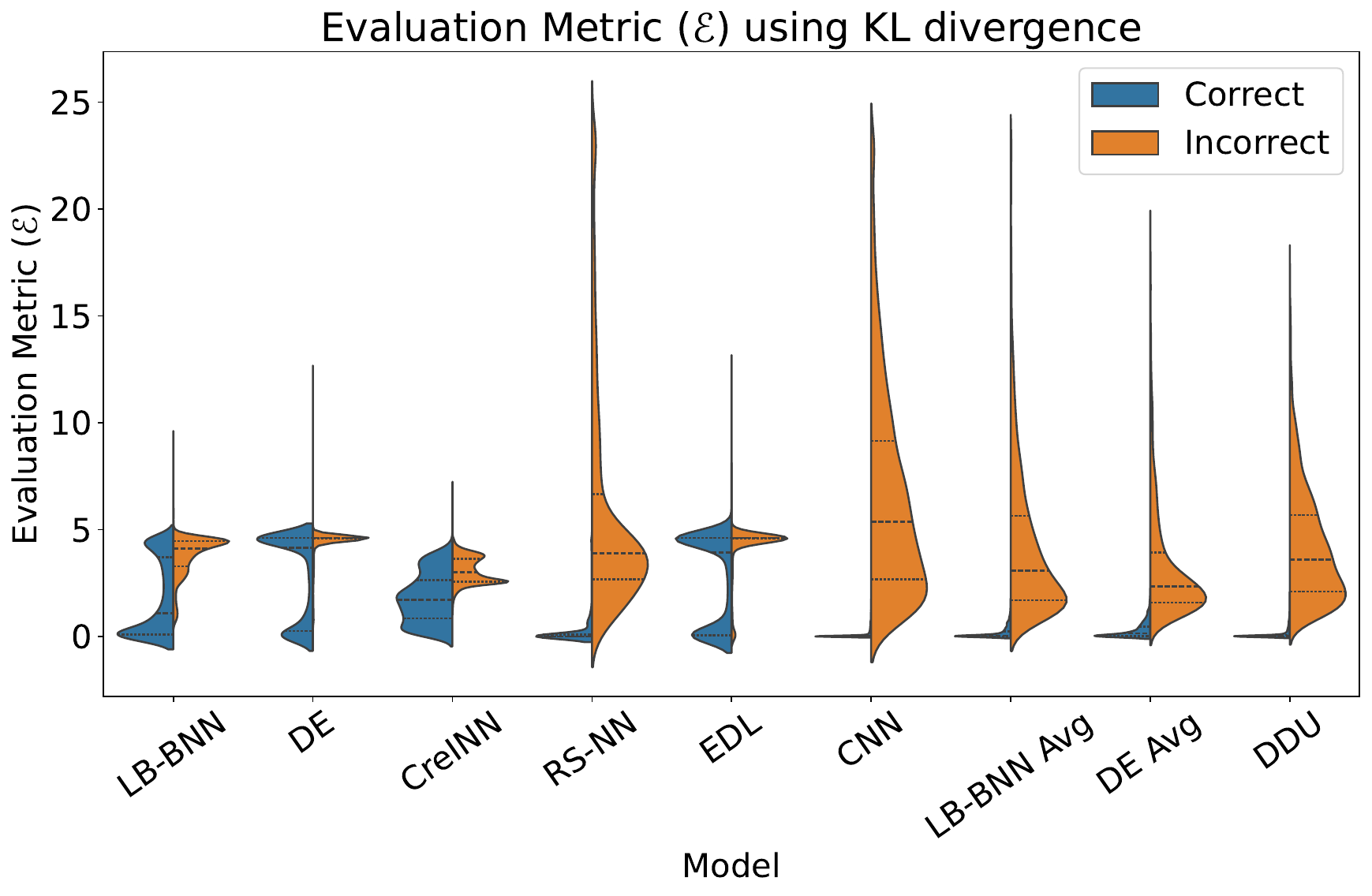}
        \vspace{-8mm}
    \label{fig:cifar100_e}
    \end{minipage} 
    \begin{minipage}[t]{\textwidth}
    \vspace{-3mm}
    \includegraphics[width=\textwidth]{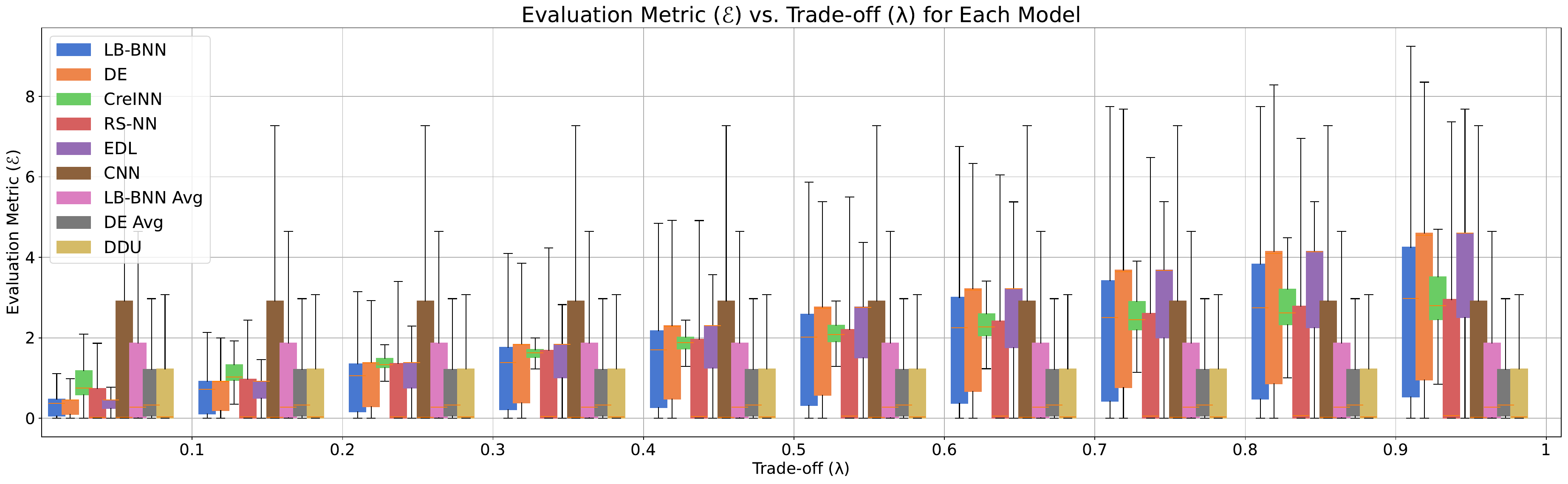}
        \vspace{-7mm}
    \label{fig:cifar100_lambda}
\end{minipage} 
        \vspace{-1mm}
    \caption{Measures of (a)\textit{(top left)} Kullback-Leibler (KL) divergence, \textit{(top right)} Non-specificity (NS), (c)\textit{(bottom left)} Evaluation Metric ($\mathcal{E}$) for Correctly Classified (CC) and Incorrectly Classified (ICC) samples (d)\textit{(bottom right)} Evaluation metric ($\mathcal{E}$) estimates \textit{vs.} trade-off ($\lambda$) for all models on \textbf{CIFAR-100}.
    }\label{fig:app_kl_ns_eval_cifar100}      
    \vspace{-4mm}
\end{figure}

\textbf{KL Divergence (KL)}. For point predictions generated by CNN, LB-BNN Avg, DE Avg and DDU, we computed the KL divergence between the ground truth and the point prediction. For all other models, we computed the KL between the ground truth and the boundary of the credal set (\textit{i.e.} , its closest vertex) (Sec. {\ref{app:kl-theory}}).
The violin plot of Fig. \ref{fig:kl_ns_eval}(a) visualises the distribution of KL divergence values for correct and incorrect predictions. 
For each model class, the KL divergence for correct predictions is represented by a single horizontal line (in blue, left), indicating that 
KL values for correct predictions are highly concentrated. 
This implies that, when the model makes correct predictions, the boundary of the predicted credal set is very close to the ground truth.
The KL divergence for incorrect predictions (in orange, right) is more broadly distributed 
for all models except EDL and E-CNN, which show consistently low KL values regardless of predictions being correct or incorrect.
{This behavior could be attributed to the high non-specificity these models demonstrate, as shown in Fig. \ref{fig:kl_ns_eval} (\textit{top right}) and Tab. \ref{tab:kl_ns_e}. Since non-specificity and credal set size are directly correlated, as illustrated in Figs. \ref{fig:app_credal_vs_non_specificity_cifar10}, \ref{fig:app_credal_vs_non_specificity_MNIST} and \ref{fig:app_credal_vs_non_specificity_cifar100}, this may be due to the large credal set size: the credal set spans much of the simplex, allowing one of its vertices to lie close to the ground truth.}

\textbf{Non-specificity (NS).} For models making pointwise predictions, Non-Specificity (NS) is trivially zero, as the credal set reduces to a single point 
(Sec. \ref{sec:credal_set}).
{it also means that models capable of expressing meaningful imprecision, such as RS-NN, EDL, and E-CNN, are inherently penalized on this metric, especially when the trade-off parameter $\lambda$ places more weight on minimizing imprecision. This should be kept in mind when interpreting comparative performance, as the unified score may favor overconfident point predictors at high $\lambda$ values.}
For all other models, 
NS (Eq. \ref{eq:non_spec}) is shown in Fig. \ref{fig:kl_ns_eval}(b),
and 
is generally broadly distributed across all models except, again, E-CNN and EDL, for both correct and incorrect predictions. Both E-CNN and EDL exhibits high non-specificity for all predictions, indicating that it is a more imprecise, less informative model. 

\textbf{Evaluation Metric ($\mathcal{E}$).} For point predictions, the Evaluation Metric ($\mathcal{E}$) reduces to the KL divergence between ground truth and the prediction. Fig. \ref{fig:kl_ns_eval}(c) shows the Evaluation Metric ($\mathcal{E}$) for all models with the trade-off parameter $\lambda = 1$. 
One can see that, when adding NS to the mix, the concentrated distributions observed in the KL divergence for correct predictions become more distributed. 
Higher NS means that the credal set is larger, which causes more variability in $\mathcal{E}$. 
EDL shows the lowest KL for both correct and incorrect predictions 
across datasets 
but struggles with non-specificity (unusually high overall), while DE and LB-BNN balance low KL with better uncertainty handling, showing higher NS for incorrect predictions and RS-NN performs consistently well on $\mathcal{E}$ for correct predictions.
A more detailed discussion can be found in Tab. \ref{tab:cc_icc}, Sec. {\ref{app:correct-incorrect}}. 

The distributions of KL-divergence, non-specificity and evaluation metric $\mathcal{E}$ for correct and incorrect samples on MNIST and CIFAR-100 datasets for all models are illustrated in Figs. \ref{fig:app_kl_ns_eval_mnist} and \ref{fig:app_kl_ns_eval_cifar100}, respectively.
For point predictions, however, the non-specificity is 0, thus the Evaluation Metric ($\mathcal{E}$) reduces to just the KL divergence between ground truth and the prediction. Figs. \ref{fig:app_kl_ns_eval_mnist}(c) and \ref{fig:app_kl_ns_eval_cifar100}(c) shows the Evaluation Metric ($\mathcal{E}$) for all models with the trade-off parameter $\lambda = 1$. All the models exhibit very concentrated distributions for KL-divergence for correct samples for both datasets, while non-specificity, and consquently, evaluation metric $\mathcal{E}$ appears more distributed. Incorrect samples, on the other hand, show more smooth distributions for all. 

\subsection{Evaluation of the trade-off parameter}
\label{sec:exp_trade_off}\label{app:trade_off_parameter}

In Fig. \ref{fig:trade_off}(d), a box plot is used to show the variation in Evaluation Metric ($\mathcal{E}$) values across different trade-off ($\lambda$) settings for each model. Each box in the plot represents the spread of $\mathcal{E}$ values for a specific model. The bottom and top edges of the box correspond to the 25\% and 75\% percentiles, respectively, while the horizontal lines (orange) inside the boxes indicate the median $\mathcal{E}$ value. 
{
Note that the y-axis range for $\mathcal{E}$ in Fig. \ref{fig:trade_off}(d) is noticeably smaller compared to the corresponding Fig. \ref{fig:trade_off}(c), despite both plots displaying $\mathcal{E}$ values. This difference arises because Fig. \ref{fig:trade_off}(d) uses box plots, which suppresses the display of outlier values. As a result, the y-axis is automatically scaled to reflect only the range of non-outlier $\mathcal{E}$ values.
In contrast, Fig. \ref{fig:trade_off}(c) visualizes the entire distribution of $\mathcal{E}$ using violin plots, which include all data points including outliers and extreme values. This results in a much broader y-axis range (up to 25), capturing the full variation in the data.}

For LB-BNN, DE, and CreINN, $\mathcal{E}$ steadily increases with $\lambda$, reflecting low KL and moderate non-specificity, also shown in Tab. \ref{tab:lambda-1}. In contrast, E-CNN and EDL show a shift in the range of $\mathcal{E}$ across $\lambda$ values while maintaining a consistent plot height, indicating very low KL and unusually high NS. This means that despite changes in the trade-off parameter $\lambda$, the variation of the evaluation metric $\mathcal{E}$ for E-CNN and EDL remains largely unchanged. {These models appear to be \textit{highly imprecise} under all conditions.}
For MNIST (Fig. \ref{fig:app_kl_ns_eval_mnist}(d)), all the models except E-CNN show relatively consistent values across all values $\lambda$, indicating that the models are very precise for this dataset. Whereas, for CIFAR-100 (Figs. \ref{fig:app_kl_ns_eval_cifar100}(d)), there is a noticeable trend of increasing $\mathcal{E}$ values as $\lambda$ grows showcasing the potential of trade-off parameter for model selection. 

\begin{table}[!h]
\caption{ 
Trade-off ($\lambda$) \textit{vs.} Evaluation Metric ($\mathcal{E}$) for different values of $\lambda$ for the CIFAR-10 dataset.} \label{tab:lambda-1}
    \vspace{-8pt}
\centering
\resizebox{\textwidth}{!}{
\begin{tabular}{lccccccc}
\toprule
\multirow{2}{*}{Dataset} &
 \multirow{2}{*}{\begin{tabular}[c]{@{}c@{}}Trade-off\\ parameter ($\lambda$)\end{tabular}} 
    & \multicolumn{5}{c}{Evaluation Metric ($\mathcal{E}$)} \\
  \cmidrule{3-8}
& & LB-BNN  & DE & EDL & CreINN  & E-CNN & RS-NN \\
  \midrule
  \multicolumn{1}{c}{\multirow{6}{*}{CIFAR-10}}
& 0.1 & $0.259 \pm 1.316$ & $\mathbf{0.069} \pm \mathbf{0.375}$ & $0.229 \pm 0.012$ & $0.117 \pm 0.382$ & $0.354 \pm 0.215$ & $0.399 \pm 1.895$\\
& 0.2 & $0.276 \pm 1.319$ & $\mathbf{0.108} \pm \mathbf{0.395}$ & $0.455 \pm 0.016$ & $0.177 \pm 0.407$ & $0.515 \pm 0.215$ & $0.400 \pm 1.896$\\
& 0.3 & $0.293 \pm 1.323$ & $\mathbf{0.146} \pm \mathbf{0.426}$ &  $0.682 \pm 0.022$& $0.237 \pm 0.445$ & $0.676 \pm 0.215$ & $0.401 \pm 1.896$\\
& 0.4 & $0.309 \pm 1.327$ & $\mathbf{0.184} \pm \mathbf{0.467}$ & $0.909 \pm 0.029$  & $0.296 \pm 0.494$ & $0.837 \pm 0.215$ & $0.402 \pm 1.897$\\
& 0.5 & $0.326 \pm 1.334$ & $\mathbf{0.223} \pm \mathbf{0.514}$ & $1.135 \pm 0.035$ & $0.356 \pm 0.550$ & $0.998 \pm 0.215$ & $0.403 \pm 1.897$\\
& 0.6 & $0.342 \pm 1.341$ & $\mathbf{0.261} \pm \mathbf{0.566}$ &  $1.362 \pm 0.042$ & $0.415 \pm 0.612$ & $1.159 \pm 0.215$ & $0.404 \pm 1.898$\\
& 0.7 & $0.359 \pm 1.349$ & $\mathbf{0.300} \pm \mathbf{0.622}$ &  $1.588 \pm 0.049$  &  $0.475 \pm 0.679$ & $1.319 \pm 0.215$ & $0.405 \pm 1.898$\\
& 0.8 & $0.376 \pm 1.359$ & $\mathbf{0.338} \pm \mathbf{0.681}$ &  $1.815 \pm 0.056$   &  $0.535 \pm 0.748$ & $1.480 \pm 0.215$ & $0.405 \pm 1.899$\\
& 0.9 & $0.392 \pm 1.369$ & $\mathbf{0.377} \pm \mathbf{0.742}$ &  $2.042 \pm 0.063$  & $0.594 \pm 0.819$ & $1.641 \pm 0.215$ & $0.406 \pm 1.899$\\
& 1 & $0.409 \pm 1.381$ & $0.415 \pm 0.805$ &  $2.268 \pm 0.070$  &$0.654 \pm 0.892$ & $1.802 \pm 0.215$ & $\mathbf{0.407} \pm \mathbf{1.900}$\\
\midrule 
\multicolumn{1}{c}{\multirow{6}{*}{MNIST}}
& 0.1 & $0.011 \pm 0.132$ & $\mathbf{0.009} \pm \mathbf{0.080}$ &  $0.226 \pm 0.005$  & $0.072 \pm 0.609$ & $0.198 \pm 0.065$ & $0.053 \pm 0.740$\\
& 0.2 & $0.020 \pm 0.147$ & $\mathbf{0.016} \pm \mathbf{0.098}$ & $0.452 \pm 0.011$   & $0.072 \pm 0.609$ & $0.359 \pm 0.065$ & $0.053 \pm 0.740$\\
& 0.3 & $0.030 \pm 0.170$ & $\mathbf{0.023} \pm \mathbf{0.123}$ &  $0.678 \pm 0.016$  & $0.073 \pm 0.610$ & $0.519 \pm 0.065$ & $0.053 \pm 0.740$\\
& 0.4 & $0.039 \pm 0.198$ & $\mathbf{0.029} \pm \mathbf{0.150}$ &  $0.905 \pm 0.021$  & $0.074 \pm 0.610$ & $0.680 \pm 0.064$ & $0.053 \pm 0.740$\\
& 0.5 & $0.048 \pm 0.229$ & $\mathbf{0.036} \pm \mathbf{0.179}$ & $1.131 \pm 0.027$   & $0.074 \pm 0.610$ & $0.841 \pm 0.064$ & $0.053 \pm 0.740$\\
& 0.6 & $0.057 \pm 0.261$ & $\mathbf{0.043} \pm \mathbf{0.208}$ & $1.357 \pm 0.032$   & $0.075 \pm 0.611$ & $1.002 \pm 0.064$ & $0.053 \pm 0.741$\\
& 0.7 & $0.066 \pm 0.295$ & $\mathbf{0.050} \pm \mathbf{0.239}$ & $1.583 \pm 0.037$   & $0.075 \pm 0.611$ & $1.163 \pm 0.064$ & $0.054 \pm 0.741$\\
& 0.8 & $0.075 \pm 0.330$ & $0.056 \pm 0.269$ & $1.809 \pm 0.043$   & $0.076 \pm 0.611$ & $1.324 \pm 0.064$ & $\mathbf{0.054} \pm \mathbf{0.741}$\\
& 0.9 & $0.084 \pm 0.366$ & $0.063 \pm 0.300$ &  $2.035 \pm 0.048$  & $0.076 \pm 0.612$ & $1.484 \pm 0.064$ & $\mathbf{0.054} \pm \mathbf{0.741}$\\
& 1 & $0.093 \pm 0.401$ & $0.070 \pm 0.331$ & $2.261 \pm 0.053$   & $0.077 \pm 0.612$ & $1.645 \pm 0.064$ & $\mathbf{0.054} \pm \mathbf{0.741}$\\
\midrule
\multicolumn{1}{c}{\multirow{6}{*}{CIFAR-100}}
& 0.1 & $0.381 \pm 0.514$ & $\mathbf{0.337} \pm \mathbf{0.299}$ &  $0.354 \pm 0.257$  & $0.929 \pm 0.582$ & - & $1.575 \pm 3.957$\\
& 0.2 & $\mathbf{0.616} \pm \mathbf{0.580}$ & $0.656 \pm 0.437$ &  $0.697 \pm 0.404$  & $1.134 \pm 0.536$ & - & $1.632 \pm 3.951$\\
& 0.3 & $\mathbf{0.850} \pm \mathbf{0.686}$ & $0.974 \pm 0.605$ & $1.041 \pm 0.573$   & $1.339 \pm 0.514$ & - & $1.689 \pm 3.948$\\
& 0.4 & $\mathbf{1.085} \pm \mathbf{0.818}$ & $1.292 \pm 0.784$ &  $1.384 \pm 0.749$ & $1.544 \pm 0.519$ & - & $1.746 \pm 3.949$\\
& 0.5 & $\mathbf{1.320} \pm \mathbf{0.964}$ & $1.610 \pm 0.967$ & $1.727 \pm 0.929$  &  $1.749 \pm 0.550$ & - & $1.803 \pm 3.953$\\
& 0.6 & $\mathbf{1.555} \pm \mathbf{1.119}$ & $1.928 \pm 1.153$ &  $2.071 \pm 1.110$  & $1.954 \pm 0.603$ & - & $1.860 \pm 3.961$\\
& 0.7 & $\mathbf{1.789} \pm \mathbf{1.280}$ & $2.247 \pm 1.340$ & $2.414 \pm 1.291$   & $2.159 \pm 0.673$ & - & $1.917 \pm 3.972$\\
& 0.8 & $2.024 \pm 1.444$ & $2.565 \pm 1.528$ & $2.758 \pm 1.474$   & $2.364 \pm 0.756$ & - & $\mathbf{1.974\mathbf} \pm \mathbf{3.986}$\\
& 0.9 & $2.259 \pm 1.612$ & $2.883 \pm 1.717$ & $3.101 \pm 1.657$   & $2.569 \pm 0.847$ & - & $\mathbf{2.031} \pm \mathbf{4.004}$\\
& 1 & $2.494 \pm 1.781$ & $3.201 \pm 1.906$ &  $3.445 \pm 1.840$  & $2.774 \pm 0.945$ & - & $\mathbf{2.088} \pm \mathbf{4.025}$\\
\bottomrule
\end{tabular}
} 
\end{table}

In Tab. \ref{tab:lambda-1}, the MNIST results show that DE has the lowest $\mathcal{E}$ for $\lambda$ values 0.1 to 0.7. therefore, the model selection will return DE as the best model. As $\lambda$ increases from 0.7 to 1, RS-NN has the lowest scores and the model selection procedure returns RS-NN as the best model. Overall, these are models with low $\mathcal{E}$ scores. These results are also reflected in Tab. \ref{tab:model_selection_mnist}.
In the CIFAR-100 dataset, the results are even more pronounced. The evaluation metric $\mathcal{E}$ start at a relatively high baseline for most models. The standard deviations also illustrate substantial variability among models, particularly at higher $\lambda$ values, where some models demonstrate higher susceptibility to the increase in non-specificity.

\subsection{Model Selection using the Evaluation Metric}
\label{sec:exp_model_selection}

\begin{table}[!h]
\centering
    \caption{Model Rankings Based on KL and NS on \textbf{CIFAR-10} for different values of $\lambda$. Model selection is based on the mean of Evaluation Metric ($\mathcal{E}$) with models with the lowest $\mathcal{E}$ ranking first.}
    \vspace{-8pt}
    \label{tab:model_rank}
    \centering
    \resizebox{\textwidth}{!}{
    \begin{tabular}{@{}c|c|c@{}}
        \toprule
        \textbf{Trade-off ($\lambda$)} & \textbf{Model Ranking} & \textbf{Evaluation ($\mathcal{E}$) Mean} \\ \midrule
        \multirow{1}{*}{0.1} & DE, CreINN, EDL, LB-BNN, E-CNN, RS-NN 
        & [0.069, 0.117, 0.229, 0.259, 0.354, 0.399]\\ 
        \multirow{1}{*}{0.2} & DE, CreINN, LB-BNN, RS-NN, EDL, E-CNN 
        & [0.108, 0.177, 0.276, 0.309, 0.456, 0.515] \\ 
        \multirow{1}{*}{0.3} & DE, CreINN, LB-BNN, RS-NN, E-CNN, EDL 
        & [0.146, 0.237, 0.293, 0.309, 0.676, 0.682] \\ 
        \multirow{1}{*}{0.4} & DE, CreINN, LB-BNN, RS-NN, E-CNN, EDL 
        & [0.184, 0.296, 0.309, 0.402, 0.837, 0.909] \\ 
        \multirow{1}{*}{0.5} & DE, LB-BNN, CreINN, RS-NN, E-CNN, EDL 
        & [0.223, 0.326, 0.356, 0.403, 0.998, 1.136] \\ 
        \multirow{1}{*}{0.6} & DE, LB-BNN, RS-NN, CreINN, E-CNN, EDL 
        & [0.261, 0.342, 0.404, 0.415, 1.159, 1.363] \\ 
        \multirow{1}{*}{0.7} & DE, LB-BNN, RS-NN, CreINN, E-CNN, EDL 
        & [0.300, 0.359, 0.405, 0.475, 1.319, 1.589] \\ 
        \multirow{1}{*}{0.8} & DE, LB-BNN, RS-NN, CreINN, E-CNN, EDL 
        & [0.338, 0.376, 0.405, 0.535, 1.480, 1.816] \\ 
        \multirow{1}{*}{0.9} & DE, LB-BNN, RS-NN, CreINN, E-CNN, EDL 
        & [0.377, 0.392, 0.406, 0.594, 1.641, 2.043] \\ 
        \multirow{1}{*}{1.0} & RS-NN, LB-BNN, DE, CreINN, E-CNN, EDL 
        & [0.407, 0.409, 0.415, 0.654, 1.802, 2.270] \\ 
        \bottomrule
    \end{tabular} 
    }
    \vspace{-2mm}
\end{table}

Tab. \ref{tab:model_rank} ranks all models on the basis of the mean value of the evaluation metric $\mathcal{E}$ 
on CIFAR-10, across different values of $\lambda$ (Tabs. \ref{tab:model_selection_mnist}, \ref{tab:model_selection_cifar100} for MNIST, CIFAR-100). Lower values indicate better model performance. Here we have excluded models predicting point estimates 
to better show the effect of the trade-off parameter on $\mathcal{E}$.

DE consistently ranks first for most $\lambda$ values, followed by CreINN and LB-BNN. As $\lambda$ increases, RS-NN shows improved performance, ranking first at $\lambda = 1.0$. E-CNN and EDL perform worse overall, especially at higher $\lambda$ values, where their $\mathcal{E}$ values are significantly higher. Models like DE and CreINN rank higher at lower $\lambda$ values (\textit{i.e.} , leaning towards imprecision in low-risk tasks like crop disease classification), while models like RS-NN and LB-BNN perform better at higher $\lambda$ values (penalizing imprecision in high-risk scenarios, such as autonomous driving). An example of practical utility of the metric is given in Sec. {\ref{app:utility}}.

\begin{table}[!h]
    \centering
    \caption{Model Selection Based on Evaluation Metric using KL and Non-Specificity on \textbf{MNIST}.}
    \vspace{-8pt}
\resizebox{\textwidth}{!}{
    \begin{tabular}{@{}c|c|c@{}}
        \toprule
        \textbf{Trade-off ($\lambda$)} & \textbf{Model Ranking} & \textbf{Evaluation ($\mathcal{E}$) Mean} \\ \midrule
        0.1 & DE, LB-BNN, RS-NN, CreINN, E-CNN, EDL & [0.009, 0.011, 0.053, 0.072, 0.198, 0.226] \\
        0.2 & DE, LB-BNN, RS-NN, CreINN, E-CNN, EDL & [0.016, 0.020, 0.053, 0.072, 0.359, 0.452] \\
        0.3 & DE, LB-BNN, RS-NN, CreINN, E-CNN, EDL & [0.023, 0.030, 0.053, 0.073, 0.519, 0.678] \\
        0.4 & DE, LB-BNN, RS-NN, CreINN, E-CNN, EDL & [0.029, 0.039, 0.053, 0.074, 0.680, 0.904] \\
        0.5 & DE, LB-BNN, RS-NN, CreINN, E-CNN, EDL & [0.036, 0.048, 0.053, 0.074, 0.841, 1.130] \\
        0.6 & DE, RS-NN, LB-BNN, CreINN, E-CNN, EDL & [0.043, 0.053, 0.057, 0.075, 1.002, 1.356] \\
        0.7 & DE, RS-NN, LB-BNN, CreINN, E-CNN, EDL & [0.050, 0.054, 0.066, 0.075, 1.163, 1.582] \\
        0.8 & RS-NN, DE, LB-BNN, CreINN, E-CNN, EDL & [0.054, 0.056, 0.075, 0.076, 1.324, 1.808] \\
        0.9 & RS-NN, DE, CreINN, LB-BNN, E-CNN, EDL & [0.054, 0.063, 0.076, 0.084, 1.484, 2.034] \\
        1.0 & RS-NN, DE, CreINN, LB-BNN, E-CNN, EDL & [0.054, 0.070, 0.077, 0.093, 1.645, 2.260] \\ \bottomrule
    \end{tabular}
    }
    \label{tab:model_selection_mnist}
\end{table}

\begin{table}[!h]
    \centering
    \small
    \caption{Model Selection Based on Evaluation Metric using KL and Non-Specificity on \textbf{CIFAR-100}.}
        \vspace{-8pt}
    \resizebox{0.9\textwidth}{!}{
    \begin{tabular}{@{}c|c|c@{}}
        \toprule
        \textbf{Trade-off ($\lambda$)} & \textbf{Model Ranking} & \textbf{Evaluation ($\mathcal{E}$) Mean} \\ \midrule
        0.1 & DE, EDL, LB-BNN, CreINN, RS-NN &
[0.337, 0.354, 0.381, 0.929, 1.575] \\
        0.2 &  LB-BNN, DE, EDL, CreINN, RS-NN &
[0.616, 0.656, 0.697, 1.134, 1.632] \\
        0.3 & LB-BNN, DE, EDL, CreINN, RS-NN &
[0.850, 0.974, 1.041, 1.339, 1.689] \\
        0.4 & LB-BNN, DE, EDL, CreINN, RS-NN &
[1.085, 1.292, 1.384, 1.544, 1.746] \\
        0.5 & LB-BNN, DE, EDL, CreINN, RS-NN &
[1.320, 1.610, 1.727, 1.749, 1.803] \\
        0.6 & LB-BNN, RS-NN, DE, CreINN, EDL &
[1.555, 1.860, 1.928, 1.954, 2.071] \\
        0.7 & LB-BNN, RS-NN, CreINN, DE, EDL &
[1.789, 1.917, 2.159, 2.247, 2.414] \\
        0.8 & RS-NN, LB-BNN, CreINN, DE, EDL &
[1.974, 2.024, 2.364, 2.565, 2.758] \\
        0.9 & RS-NN, LB-BNN, CreINN, DE, EDL &
[2.031, 2.259, 2.569, 2.883, 3.101] \\
        1.0 & RS-NN, LB-BNN, CreINN, DE, EDL &
[2.088, 2.494, 2.774, 3.201, 3.445] \\ \bottomrule
    \end{tabular}
    }
    \label{tab:model_selection_cifar100}
\end{table}

As $\lambda$ approaches $1$, RS-NN again is selected as the best model as shown in Tabs. \ref{tab:lambda-1}, \ref{tab:model_selection_mnist} and \ref{tab:model_selection_cifar100}.
The trends observed in the CIFAR-100 results further reinforce the need for careful tuning of $\lambda$ to optimize model performance. As $\lambda$ increases, the complexity of classification tasks also rises, necessitating models that can adaptively learn from both accurate and non-specific features.


\subsection{Ablation Studies}\label{app:additional_exp_main}

In this section, we will address the following questions based on our experimental findings:
\begin{itemize}
    \item \textbf{Q1:} How does the evaluation metric behave for correct vs. incorrect classifications?
    \item \textbf{Q2:} How does approximating credal set vertices affect the metric?
    \item \textbf{Q3:} How do different distance and non-specificity measures affect evaluation outcomes?
    \item \textbf{Q4:} Why is distance a better choice than confidence score in the evaluation metric?
    \item \textbf{Q5:} How does non-specificity relate to credal set size as a measure of uncertainty?
    \item \textbf{Q6:} How does the number of prediction samples impact metric stability and performance?
\end{itemize}

\subsubsection{Metric behavior for Correct (CC) \textit{vs.} Incorrect Classifications (ICC)} \label{app:correct-incorrect}

In Tab. \ref{tab:cc_icc}, a comparison of KL divergence, Non-Specificity, and Evaluation Metric ($\mathcal{E}$) with trade-off $\lambda$ = 1 is presented for Correctly Classified (CC) and Incorrectly Classified (ICC) samples across each model. Incorrect samples demonstrate a higher mean value than correct samples for KL divergence, Non-Specificity, and Evaluation Metric ($\mathcal{E}$) indicating that the models are behaving appropriately, except E-CNN which gives high non-specificity for both correct and incorrect predictions.

We can form a tentative ranking of the models with Tab. \ref{tab:cc_icc}. For correct predictions of CIFAR-10, MNIST, and CIFAR-100, we desire low KL and low non-specificity, as exhibited by DE and RS-CNN respectively. For incorrect predictions, a low KL and high non-specificity is optimal. Finally, evaluation metric $\mathcal{E}$ ($\lambda = 1$) should reflect a low value for both correct and incorrect predictions (RS-NN and DE).

\begin{table}[!h]
\caption[Comparison of KL divergence (KL), Non-Specificity (NS), and Evaluation Metric ($\mathcal{E}$) (trade-off $\lambda = 1$) for Correctly Classified (CC) and Incorrectly Classified (ICC) samples for each model on three datasets: CIFAR-10, MNIST, and CIFAR-100.]{Comparison of KL divergence (KL), Non-Specificity (NS), and Evaluation Metric ($\mathcal{E}$) (trade-off $\lambda = 1$) for Correctly Classified (CC) and Incorrectly Classified (ICC) samples for each model on three datasets: CIFAR-10, MNIST, and CIFAR-100. {\textit{It is important to note that a low $\mathcal{E}$ for incorrect predictions can result from either low KL or low NS, so $\mathcal{E}$ alone cannot distinguish between confidently wrong (low NS, high KL) and uncertain but inaccurate (high NS, low KL) cases. Thus, $\mathcal{E}$ should be interpreted alongside KL and NS.}}}
\vspace{-8pt}
\label{tab:cc_icc}
\centering
\resizebox{\textwidth}{!}{
\begin{tabular}{lccccccc}
\toprule
\multirow{2}{*}{Dataset} &
  \multirow{2}{*}{Model} &
  \multicolumn{2}{c}{KL {divergence} (KL)} &
  \multicolumn{2}{c}{Non-Specificity (NS)} &
  \multicolumn{2}{c}{Evaluation metric ($\mathcal{E}$)} \\
\cmidrule(lr){3-4} \cmidrule(lr){5-6} \cmidrule(lr){7-8}
&& CC ($\downarrow$) & ICC ($\downarrow$)  & CC ($\downarrow$) & ICC ($\uparrow$) & CC ($\downarrow$) & ICC ($\downarrow$)\\
\midrule
\multirow{6}{*}{CIFAR-10} 
& LB-BNN & $0.009 \pm 0.038$ & $2.181 \pm 3.444$ & $0.109 \pm 0.323$ & $0.637 \pm 0.600$ & $0.118 \pm 0.341$ & $2.818 \pm 3.203$\\
& DE & $0.0001 \pm 0.001$ & $0.489 \pm 1.395$ & $0.306 \pm 0.639$ & $1.566 \pm 0.756$ & $0.306 \pm 0.639$ & $2.055 \pm 1.183$\\
& EDL & $\mathbf{0.00005} \pm \mathbf{0.0002}$ & $\mathbf{0.0051} \pm \mathbf{0.017}$ & $2.253 \pm	0.080$ & $\mathbf{2.288} \pm \mathbf{0.034}$ & $ 2.253 \pm 0.08$ & $2.293 \pm 0.023$ \\
& CreINN & $0.003 \pm 0.006$ & $0.476 \pm 1.001$ & $0.465 \pm 0.727$ & $1.590 \pm 0.731$ & $0.468 \pm 0.728$ & $\mathbf{2.066} \pm \mathbf{0.744}$\\
& E-CNN & $0.108 \pm 0.093$ & $0.625 \pm 0.115$ & $1.609 \pm 0.003$ & $1.610 \pm 0.001$ & $1.717 \pm 0.092$ & $2.235 \pm 0.115$\\
& RS-NN & $0.009 \pm 0.055$ & $5.562 \pm 4.747$ & $\mathbf{0.004} \pm \mathbf{0.031}$ & $0.076 \pm 0.145$ & $\mathbf{0.013} \pm \mathbf{0.077}$ & $5.638 \pm 4.691$\\
  \cmidrule{2-8}
& CNN & $0.021 \pm 0.090$ & $4.735 \pm 3.604$ & $0.000 \pm 0.000$ & $0.000 \pm 0.000$ & $0.021 \pm 0.090$ & $4.735 \pm 3.604$\\
& LB-BNN Avg & $0.034 \pm 0.114$ & $3.628 \pm 3.145$ & $0.000 \pm 0.000$ & $0.000 \pm 0.000$ & $0.034 \pm 0.114$ & $3.628 \pm 3.145$\\
& DE Avg & $0.050 \pm 0.135$ & $2.378 \pm 2.000$ & $0.000 \pm 0.000$ & $0.000 \pm 0.000$ & $0.050 \pm 0.135$ & $2.378 \pm 2.000$\\
& DDU & $0.031 \pm	0.104$ & $3.243 \pm	2.196$ & $0.000 \pm 0.000$ & $0.000 \pm 0.000$ & $0.031 \pm	0.104$ & $3.243 \pm	2.196$\\
\midrule
\multirow{6}{*}{MNIST} 
& LB-BNN & $0.00002 \pm 0.001$ & $0.490 \pm 1.832$ & $0.083 \pm 0.359$ & $\mathbf{1.883} \pm \mathbf{0.656}$ & $0.083 \pm 0.359$ & $2.373 \pm 1.449$\\
& DE & $0.00005 \pm 0.001$ & $0.340 \pm 0.805$ & $0.058 \pm 0.291$ & $1.503 \pm 0.766$ & $0.058 \pm 0.291$ & $\mathbf{1.843} \pm \mathbf{0.758}$\\
& EDL & $\mathbf{0.00006} \pm \mathbf{0.00002}$ & $\mathbf{0.00062} \pm \mathbf{0.005}$ & $2.255 \pm 0.055$ & $2.301 \pm 0.010$ & $2.255 \pm 0.055$ &  $2.302\pm 0.007$\\
& CreINN & $0.006 \pm 0.037$ & $3.720 \pm 2.714$ & $0.004 \pm 0.032$ & $0.061 \pm 0.212$ & $0.010 \pm 0.053$ & $3.781 \pm 2.664$\\
& E-CNN & $0.032 \pm 0.033$ & $0.675 \pm 0.115$ & $1.608 \pm 0.004$ & $1.607 \pm 0.003$ & $1.641 \pm 0.032$ & $2.282 \pm 0.116$\\
& RS-NN & $0.001 \pm 0.023$ & $6.211 \pm 5.290$ & $\mathbf{0.0004} \pm \mathbf{0.010}$ & $0.056 \pm 0.120$ & $\mathbf{0.002} \pm \mathbf{0.031}$ & $6.267 \pm 5.242$\\
  \cmidrule{2-8}
& CNN & $0.004 \pm 0.040$ & $3.484 \pm 3.234$ & $0.000 \pm 0.000$ & $0.000 \pm 0.000$ & $0.004 \pm 0.040$ & $3.484 \pm 3.234$\\
& LB-BNN Avg & $0.006 \pm 0.044$ & $2.295 \pm 2.921$ & $0.000 \pm 0.000$ & $0.000 \pm 0.000$ & $0.006 \pm 0.044$ & $2.295 \pm 2.921$\\
& DE Avg & $0.007 \pm 0.048$ & $1.956 \pm 1.287$ & $0.000 \pm 0.000$ & $0.000 \pm 0.000$ & $0.007 \pm 0.048$ & $1.956 \pm 1.287$\\
& DDU & $0.003	\pm 0.032$ & $3.367 \pm	2.092$ & $0.000 \pm 0.000$ & $0.000 \pm 0.000$ & $0.003 \pm 0.032$ & $3.367 \pm	2.092$\\
\midrule
\multirow{6}{*}{CIFAR-100} 
& LB-BNN & $0.006 \pm 0.021$ & $0.399 \pm 0.783$ & $1.790 \pm 1.748$ & $3.353 \pm 1.309$ & $1.796 \pm 1.752$ & $3.752 \pm 0.947$\\
& DE & $\mathbf{0.0001} \pm \mathbf{0.001}$ & $0.074 \pm 0.477$ & $2.780 \pm 2.013$ & $\mathbf{4.331} \pm \mathbf{0.836}$ & $2.780 \pm 2.013$ & $4.405 \pm 0.690$\\
& EDL & $0.005 \pm 0.190$ & $\mathbf{0.015} \pm \mathbf{0.194}$ & $2.629 \pm 2.086$ & $4.114	\pm 1.258$ & $2.634 \pm 2.090$ & $4.129 \pm 1.238$\\
& CreINN & $0.310 \pm 0.321$ & $0.857 \pm 0.668$ & $1.483 \pm	1.132$ & $2.232 \pm	1.147$ & $1.792 \pm 1.131$ & $\mathbf{3.089} \pm \mathbf{0.599}$\\
& E-CNN & - & - & - & - & - & -\\
& RS-NN & $0.042 \pm 0.130$ & $4.539 \pm 5.857$ & $\mathbf{0.242} \pm \mathbf{0.760}$ & $1.239 \pm 1.510$ & $\mathbf{0.284} \pm \mathbf{0.783}$ & $5.777 \pm 5.276$\\
  \cmidrule{2-8}
& CNN & $0.063 \pm 0.165$ & $6.528 \pm 4.868$ & $0.000 \pm 0.000$ & $0.000 \pm 0.000$ & $\mathbf{0.063} \pm \mathbf{0.165}$ & $6.528 \pm 4.868$\\
& LB-BNN Avg & $0.167 \pm 0.255$ & $4.235 \pm 3.548$ & $0.000 \pm 0.000$ & $0.000 \pm 0.000$ & $0.167 \pm 0.255$ & $4.235 \pm 3.548$\\
& DE Avg & $0.284 \pm 0.349$ & $3.285 \pm 2.696$ & $0.000 \pm 0.000$ & $0.000 \pm 0.000$ & $0.284 \pm 0.349$ & $\mathbf{3.285} \pm \mathbf{2.696}$\\
& DDU & $0.102 \pm	0.228$ & $4.161 \pm	2.646$ & $0.000 \pm 0.000$ & $0.000 \pm 0.000$ & $0.102 \pm	0.228$ & $4.161 \pm	2.646$\\
\bottomrule
\end{tabular}
} 
\vspace{-3mm}
\end{table}

\subsubsection{The effect of approximating credal set vertices}

To demonstrate that the approximated vertices of the credal sets are sufficient (and often, even better), we consider a 4-class example of the CIFAR-10 dataset. In Tab. \ref{tab:approx_naive}, we show the evaluation metric $\mathcal{E}$ for computed using approximated credal vertices using the approach detailed above, and the naive credal set vertices computation given in Eq. \ref{eq:prho}. For the naive credal set, we use the complete 24 (4!) vertices and for the approximated credal set, we use a reduced number of 8 vertices. The values in bold are the clearly better results whereas other values are quite close to each other. For DEs, for instance, the approximated credal set is a better representation for our metric, as the approximation uses the most relevant permutations only.

\begin{table}[!ht]
\caption{Comparison of KL, Non-Specificity, Evaluation Metric ($\mathcal{E}$) calculated using approximated versus naive credal set vertices for LB-BNN and DE on the CIFAR-10 dataset.} 
\vspace{-8pt}
\label{tab:approx_naive}
\centering
\resizebox{\textwidth}{!}{
\begin{tabular}{lccccccc}
\toprule
\multirow{2}{*}{Model} &
  \multirow{2}{*}{Credal Set Vertices} &
  \multicolumn{2}{c}{KL distance (KL)} &
  \multicolumn{2}{c}{Non-Specificity (NS)} &
  \multicolumn{2}{c}{Evaluation metric ($\mathcal{E}$)} \\
\cmidrule(lr){3-4} \cmidrule(lr){5-6} \cmidrule(lr){7-8}
&& CC ($\downarrow$) & ICC ($\downarrow$)  & CC ($\downarrow$) & ICC ($\uparrow$) & CC ($\downarrow$) & ICC ($\downarrow$)\\
\midrule
\multirow{2}{*}{LB-BNN} 
& Approximated & $0.0046 \pm 0.023$ & $0.612 \pm 0.3968$ & $0.564 \pm 0.099$ & $\mathbf{0.611} \pm \mathbf{0.140}$ & $0.569 \pm 0.104$ & $1.225 \pm 0.376$\\
& Naive & $0.005 \pm 0.023$ & $0.336 \pm 0.666$ & $\mathbf{0.033} \pm \mathbf{0.103}$ & $0.286 \pm 0.216$ & $\mathbf{0.039} \pm \mathbf{0.121}$ & $\mathbf{0.365} \pm \mathbf{0.594}$\\
\midrule
\multirow{2}{*}{DE} 
& Approximated & $0.0002 \pm 0.002$ & $\mathbf{0.142} \pm \mathbf{0.260}$ & $\mathbf{0.058} \pm \mathbf{0.125}$ & $\mathbf{0.801} \pm \mathbf{0.170}$ & $\mathbf{0.058} \pm \mathbf{0.125}$ & $\mathbf{0.944} \pm \mathbf{0.265}$\\
& Naive & $0.0001 \pm 0.002$ & $0.335 \pm 0.807$ & $0.080 \pm 0.204$ & $0.656 \pm 0.219$ & $0.080 \pm 0.205$ & $0.991 \pm 0.653$\\
\bottomrule
\end{tabular}
} 
\end{table}

\begin{figure}[!ht]
    \centering
    \begin{minipage}{0.33\textwidth}
        \centering
        \includegraphics[width=\textwidth]{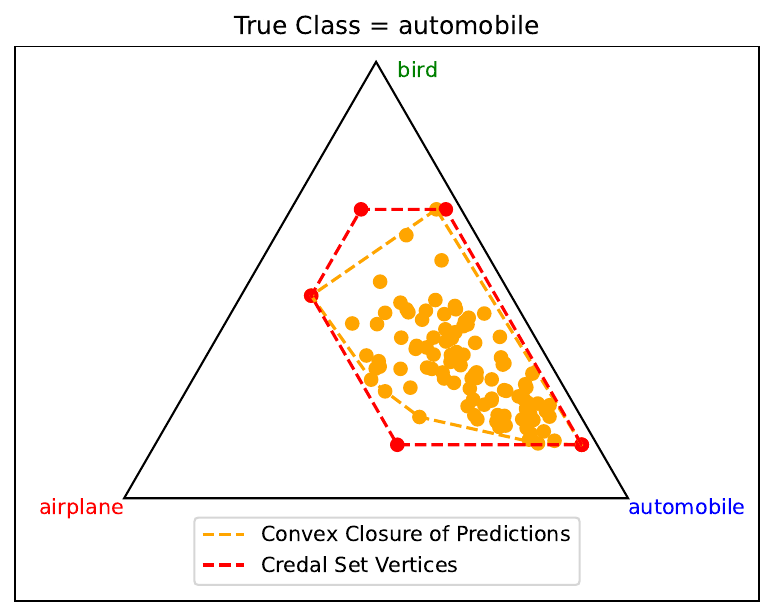}
    \end{minipage}%
    \begin{minipage}{0.33\textwidth}
        \centering
        \includegraphics[width=\textwidth]{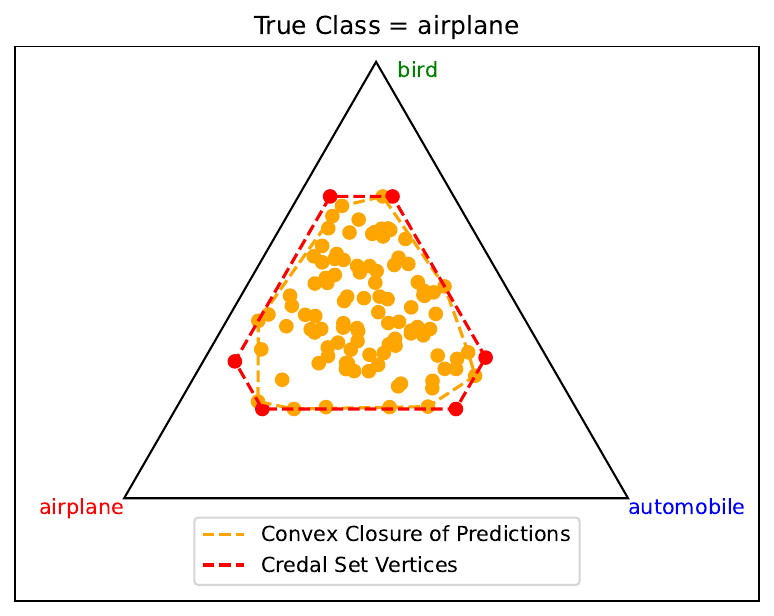}
    \end{minipage}%
    \begin{minipage}{0.33\textwidth}
        \centering
        \includegraphics[width=\textwidth]{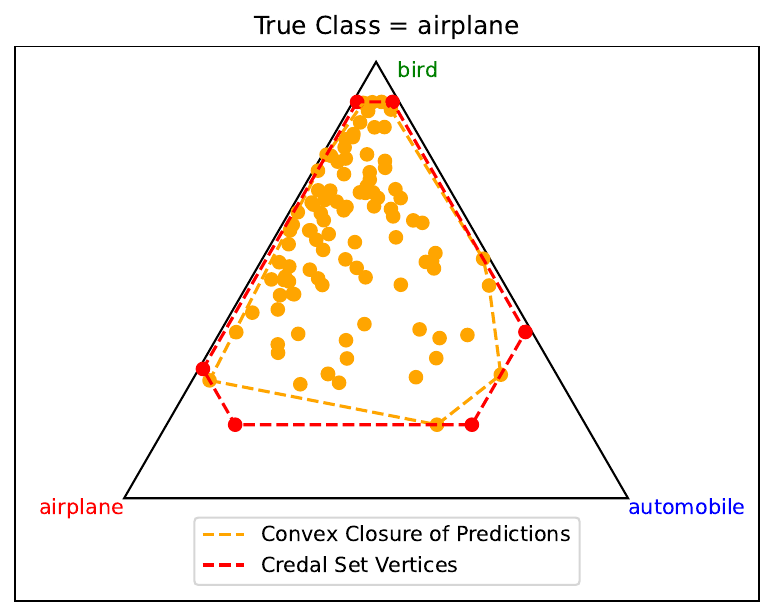} 
    \end{minipage}
    \caption{Probability simplices illustrating the convex closure of predictions and credal sets for the Bayesian model (LB-BNN) across three classes of the CIFAR-10 dataset. }
        \label{fig:convex-closure-vs-credal}
        \vspace{-8pt}
\end{figure}

\textbf{Credal set \textit{vs.} convex closure of predictions.} Since our metric uses the KL divergence between the ground truth and the closest vertex of the credal set (which is convex), the credal set representation does not affect its value as such vertices correspond to (some of) the original (\textit{e.g.}  Bayesian) predictions, as shown in Fig. \ref{fig:convex-closure-vs-credal}. Probability simplices illustrating the convex closure of predictions and credal sets for LB-BNN (Bayesian) model are shown in Fig. \ref{fig:convex-closure-vs-credal}, over three classes of CIFAR-10. The convex closure of predictions often coincides with the vertices of the credal set, indicating that some Bayesian predictions are situated exactly at the extremities of the credal set.
There is a slight difference between the convex closure of predictions and the credal set, which is not due to vertex approximations. \citep{cozman00credal} details how there is more than a single way to specify a set of distributions inducing a given lower envelope. The credal set in Fig. \ref{fig:convex-closure-vs-credal} has been approximated using coherent lower probabilities and have boundries parallel to that of the simplex.

\textbf{Computation of the credal set vertices.}
Fig. \ref{fig:credal_set_size} shows the sizes of the credal sets predicted by all models, for 50 prediction samples of the CIFAR-10 dataset. This is obtained by taking the difference between maximal and minimal extremal probability (Eq. \ref{eq:prho}) for the predicted class.  
\begin{figure}[!ht]
    \centering
    \includegraphics[width=0.9\textwidth]{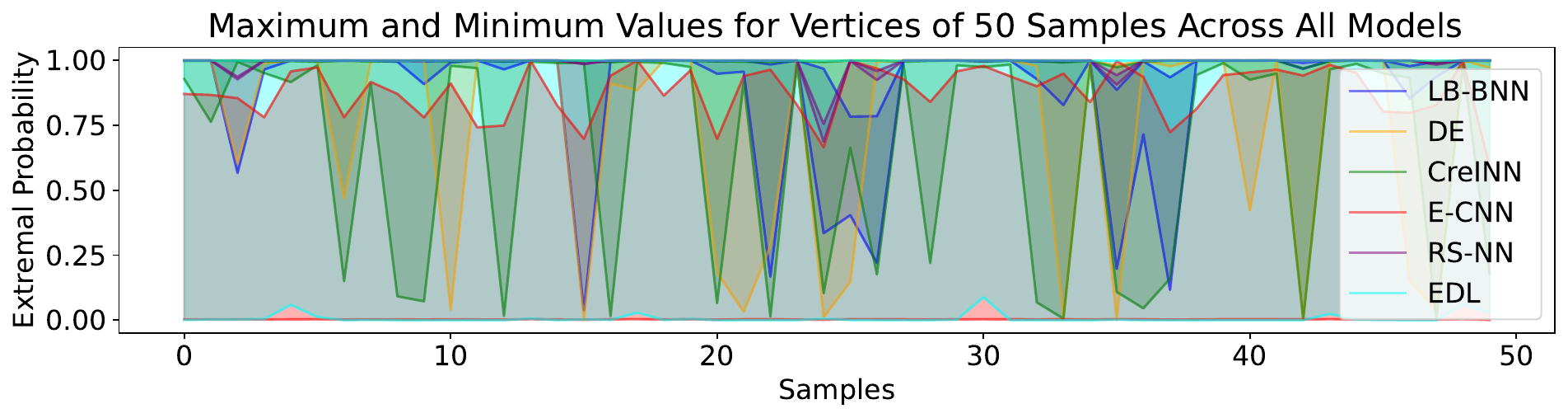}
    \vspace{-3mm}
    \caption{Credal set sizes for all models for 50 prediction samples of the CIFAR-10 dataset. Larger credal set sizes indicate a more imprecise prediction. 
    }
    \label{fig:credal_set_size}
    \vspace{-3mm}
\end{figure}
\begin{figure}[!h]
    \centering
    \includegraphics[width=0.9\textwidth]{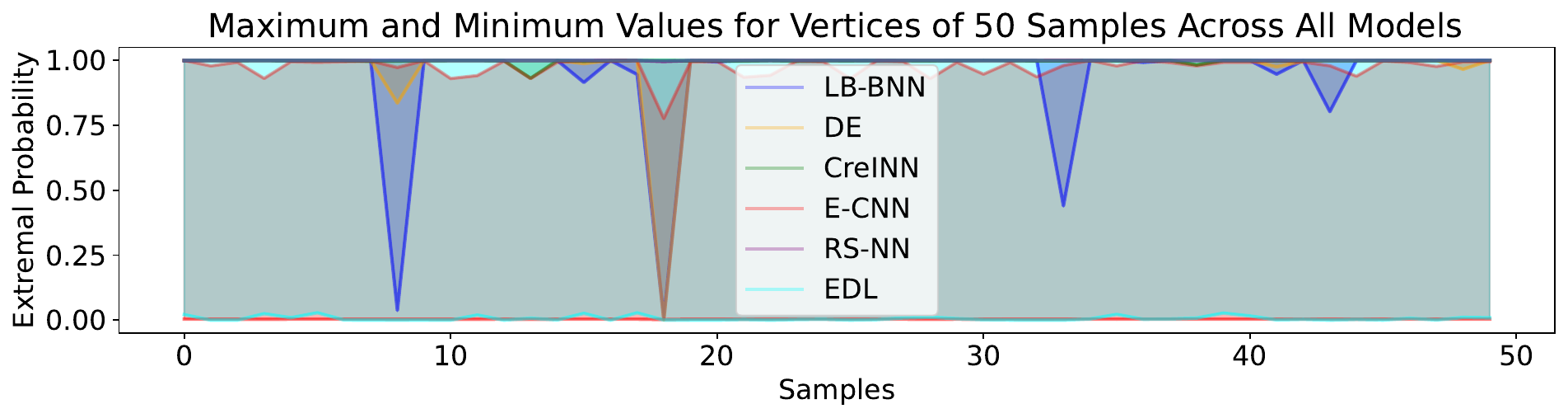}
        \vspace{-3mm}
    \caption{Credal set sizes for all models for 50 prediction samples of the MNIST dataset. Larger credal set sizes indicate a more imprecise prediction}
    \label{fig:app_credal_set_size_mnist}
        \vspace{-3mm}
\end{figure}
\begin{figure}[!htbp]
    \centering
    \includegraphics[width=0.9\textwidth]{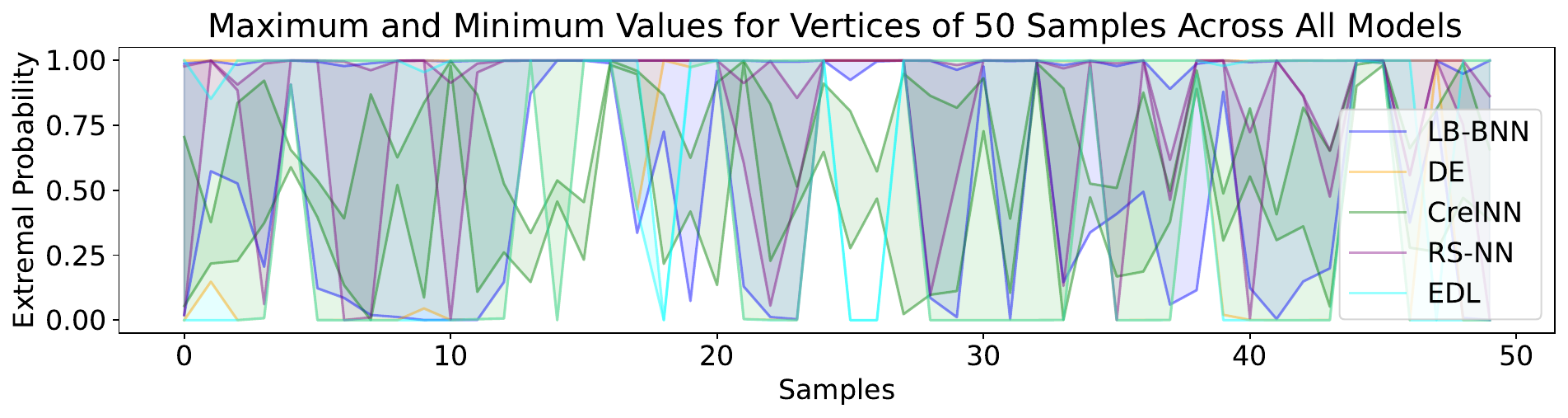}
        \vspace{-3mm}
    \caption{Credal set sizes for all models for 50 prediction samples of the CIFAR-100 dataset. Larger credal set sizes indicate a more imprecise prediction}
    \label{fig:app_credal_set_size_cifar100}
        \vspace{-3mm}
\end{figure}

Similarly, Figs. \ref{fig:app_credal_set_size_mnist} and \ref{fig:app_credal_set_size_cifar100} demonstrate the sizes of the credal
sets of the MNIST and CIFAR-100 datasets, respectively.
A larger credal set size signifies that the model is highly uncertain about its predictions. E-CNN exhibits the widest credal set for all predictions, indicating 
significant imprecision
regardless of whether the model predicts the correct or incorrect class.  This is undesirable, as it suggests a general lack of confidence in the predicted outcomes. 
Models with smaller credal set sizes demonstrate higher confidence in their predictions, implying more reliable and precise outcomes. 
Lower non-specificity values align with smaller credal widths, indicating smaller (epistemic) uncertainty and greater model confidence.

\subsubsection{Ablation study on different distance measures} \label{app:ablation-kl}

In this section, we present an ablation study comparing different distance measures, specifically Kullback-Leibler (KL) divergence and Jensen-Shannon (JS) \citep{menendez1997jensen} divergence. The goal of this analysis is to assess how these divergence measures influence the evaluation of uncertainty in our framework. While KL divergence has been chosen for its superior ability, Jensen-Shannon divergence offers a symmetric alternative that handles small probabilities more gracefully. By experimenting with both measures, we provide insights into their respective advantages and determine the most suitable metric for capturing distributional differences in uncertainty-aware model predictions.

\begin{figure}[!ht]
    \begin{minipage}[t]{0.48\textwidth}
    \includegraphics[width=\textwidth]{images/eval/CIFAR10/kl_cifar10.pdf}
    \vspace{-25pt}
    \caption*{(a)}
    \end{minipage} \hspace{0.02\textwidth}
    \begin{minipage}[t]{0.48\textwidth}
    \includegraphics[width=\textwidth]{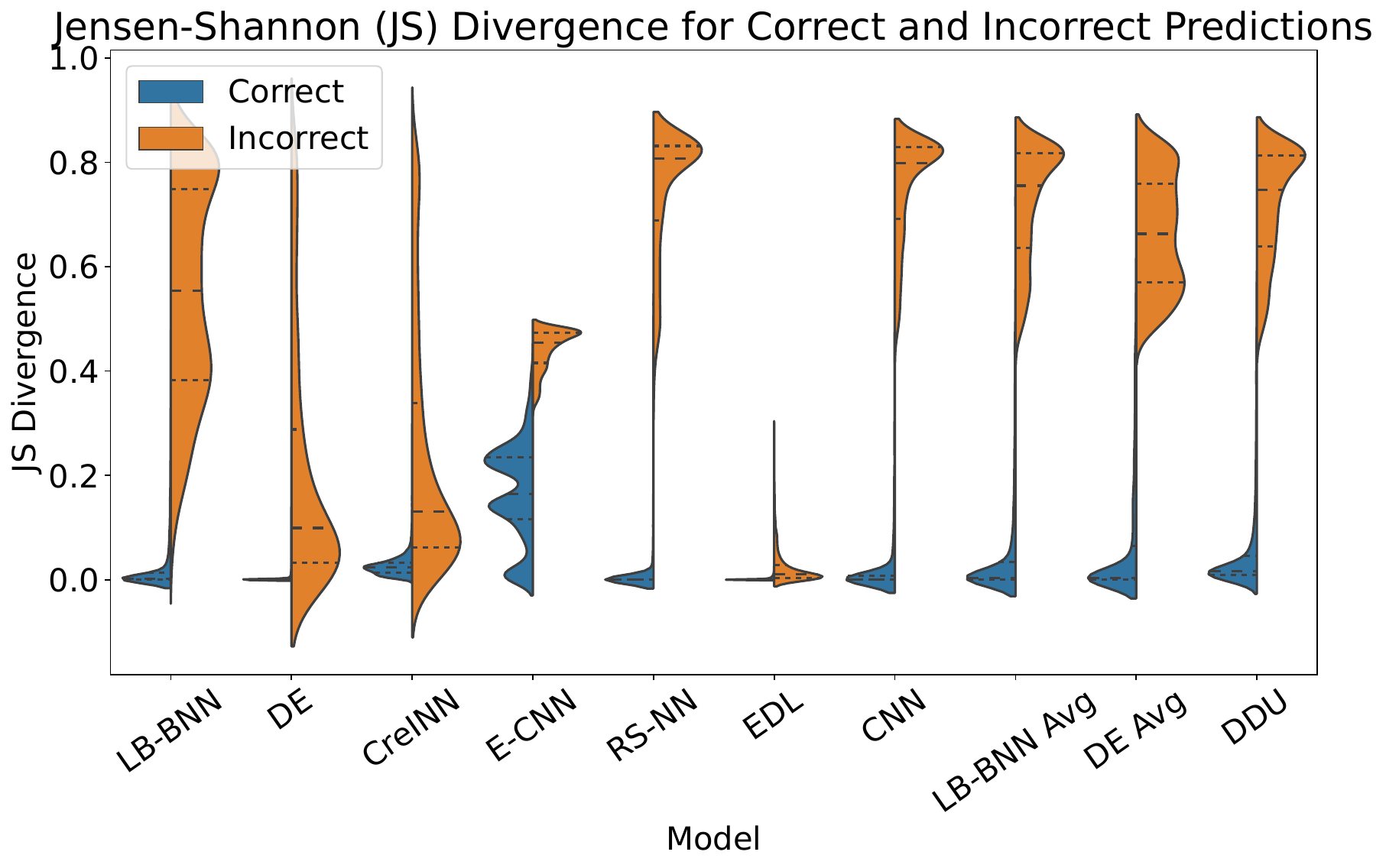}
    \vspace{-25pt}
    \caption*{(b)}
    \end{minipage}
        \vspace{-10pt}
    \caption{Comparison of (a) Kullback-Leibler (KL) divergence, and (b) Jensen-Shannon (JS) divergence for Correctly Classified (CC) and Incorrectly Classified (ICC) samples from the CIFAR-10 dataset, for all models considered here. Notably, the scales of these two measures differ significantly (Y-axis).
    }\label{fig:kl-js}
\end{figure}

Fig. \ref{fig:kl-js} illustrates the Kullback-Leibler (KL) divergence (\ref{fig:kl-js}(a)) and Jensen-Shannon (JS) divergence (\ref{fig:kl-js}(b)) between the ground truth probabilities and the vertices of the credal set on the CIFAR-10 dataset, over the entire test set. The violin plots depict distributions of the divergences for correct (blue) and incorrect (orange) predictions. Ideally, we expect the KL divergence to be larger for incorrect predictions, but not to an extent that it approaches a completely different class on the simplex of classes. This outcome indicates a preference for models that do not make severe misclassifications, which would be reflected in excessively predicting a wrong class.
The plots reveal how the KL divergence for incorrect predictions often spans a wider range, indicating greater uncertainty in these predictions compared to correct ones. 
The JS divergence is presented on a smaller scale compared to the KL divergence. 

\begin{figure}[!h]
    \includegraphics[width=\textwidth]{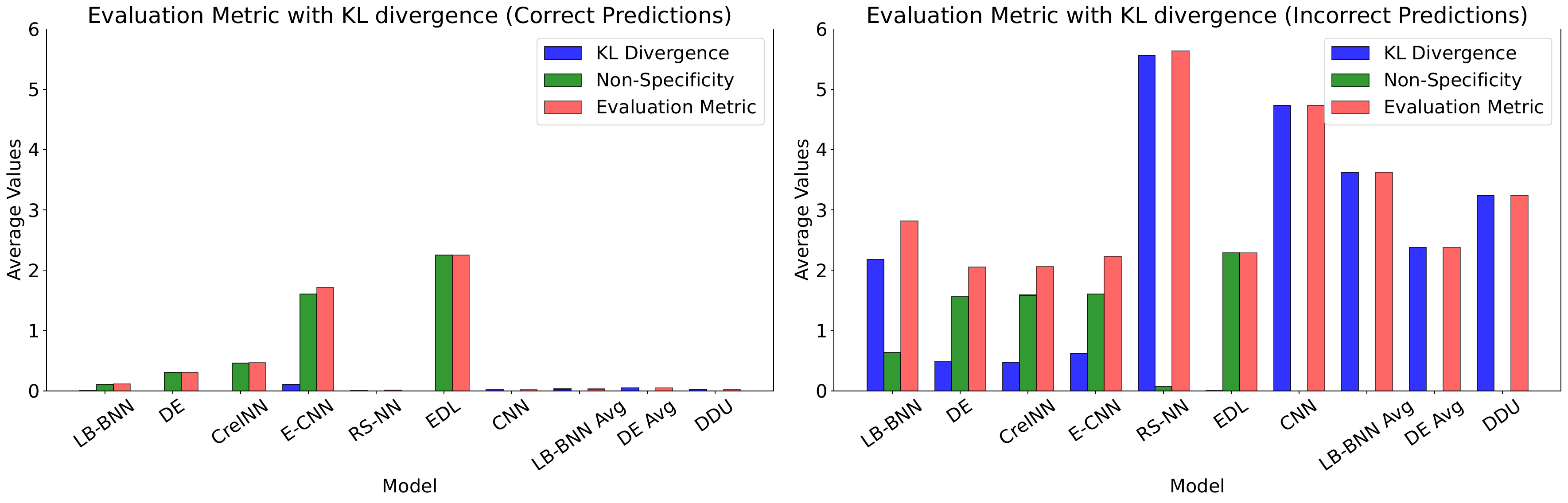}
    \includegraphics[width=\textwidth]{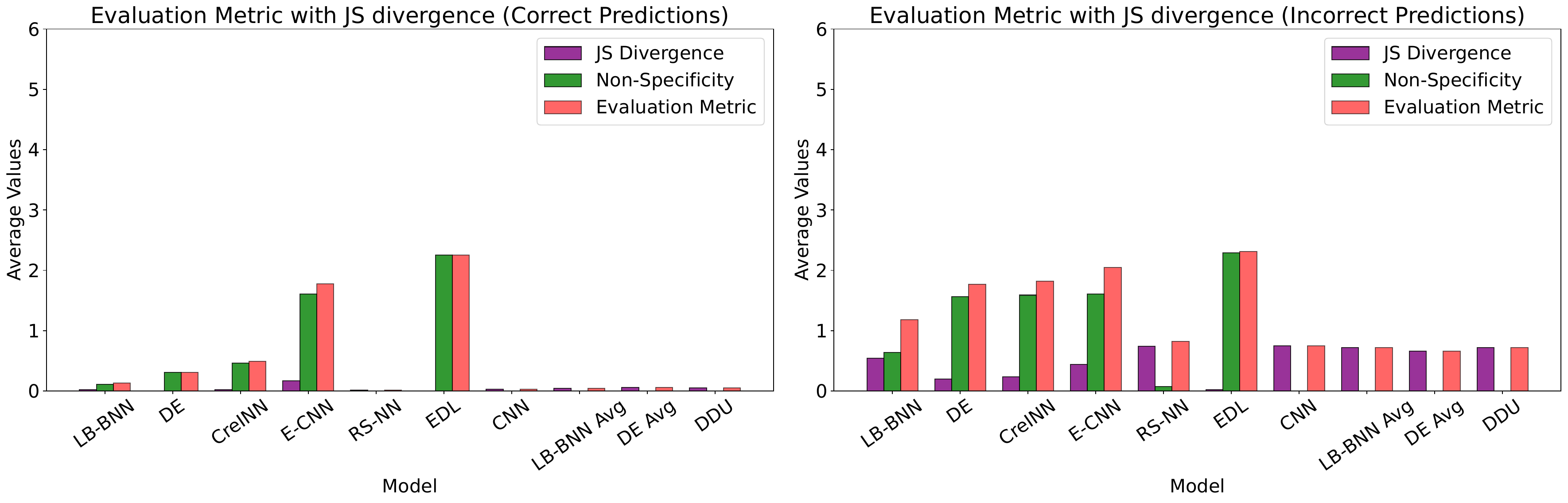}
    \vspace{-10pt}
    \caption{Comparison of mean Evaluation Metric $\mathcal{E}$ using mean \textit{Kullback Leibler} (KL) divergence (top) and mean \textit{Jensen-Shannon} (JS) divergence (bottom) for Correct (left) and Incorrect (right) predictions of the CIFAR-10 dataset.
    }\label{fig:kl-js-eval}
\end{figure}

Fig. \ref{fig:kl-js-eval} shows bar plots comparing the KL and JS divergences along with their influence on the overall evaluation metric and therefore, corresponding model rankings. All values are shown as means. Again, the differing ranges of KL and JS divergences are evident, with the Y-axis scaled equally for straightforward comparison. This discrepancy affects the overall evaluation metric $\mathcal{E}$. According to Algorithm \ref{alg:model-selection} for model selection, the model with the smallest evaluation metric should be chosen.

\textbf{Model Selection.} In Tab. \ref{tab:model_rank_ablation}, we show the ranking of all models (including the models that predict point estimates) arranged in ascending order of the mean of the evaluation metric, $\mathcal{E}$. 
As $\lambda$ increases, DE, which consistently ranked highly in KL divergence across lower $\lambda$ values, is replaced by point prediction model, DE Avg (since it has zero non-specificity). LB-BNN and RS-NN show mixed performance, typically appearing in the top half but with varying positions.

In terms of JS divergence, RS-NN stands out as a top performer, frequently occupying the top spots across most $\lambda$ values, while CNN also maintains a strong position, often ranking within the top three or four models. E-CNN and EDL tend to rank lower for both KL and JS measures. If we exclude the point prediction models from the ranking, both KL divergence and JS divergence consistently identify \textbf{RS-NN} and \textbf{DE} as the top two models in the model selection procedure.

\begin{table}[!ht]
    \centering
    \caption{Model Rankings Based on KL and JS Divergence on the CIFAR-10 dataset. Model selection is based on the mean of Evaluation Metric ($\mathcal{E}$) with the models with the lowest $\mathcal{E}$ ranking first.}
            \vspace{-8pt}
    \label{tab:model_rank_ablation}
    \resizebox{\textwidth}{!}{
    \begin{tabular}{@{}cc|l|l@{}}
        \toprule
        \textbf{Lambda} & \textbf{Metric} & \textbf{Model Ranking} & \textbf{Evaluation Metric ($\mathcal{E}$) Mean} \\ \midrule
        0.1 & KL & DE, CreINN, DE Avg, EDL, LB-BNN, DDU, E-CNN, RS-NN, LB-BNN Avg, CNN & [0.069, 0.117, 0.195, 0.229, 0.259, 0.309, 0.354, 0.399, 0.420, 0.481] \\ 
             & JS & DE, RS-NN, LB-BNN, CNN, DE Avg, DDU, CreINN, LB-BNN Avg, EDL, E-CNN & [0.053, 0.065, 0.095, 0.098, 0.099, 0.109, 0.109, 0.119, 0.238, 0.373] \\ \midrule
        0.2 & KL & DE, CreINN, DE Avg, LB-BNN, DDU, RS-NN, LB-BNN Avg, EDL, CNN, E-CNN & [0.108, 0.177, 0.195, 0.276, 0.309, 0.400, 0.420, 0.456, 0.481, 0.515] \\ 
             & JS & RS-NN, DE, CNN, DE Avg, DDU, LB-BNN, LB-BNN Avg, CreINN, EDL, E-CNN & [0.066, 0.091, 0.098, 0.099, 0.109, 0.111, 0.119, 0.169, 0.464, 0.534] \\ \midrule
        0.3 & KL & DE, DE Avg, CreINN, LB-BNN, DDU, RS-NN, LB-BNN Avg, CNN, E-CNN, EDL & [0.146, 0.195, 0.237, 0.293, 0.309, 0.401, 0.420, 0.481, 0.676, 0.682] \\ 
             & JS & RS-NN, CNN, DE Avg, DDU, LB-BNN Avg, LB-BNN, DE, CreINN, EDL, E-CNN & [0.067, 0.098, 0.099, 0.109, 0.119, 0.128, 0.130, 0.229, 0.691, 0.695] \\ \midrule
        0.4 & KL & DE, DE Avg, CreINN, DDU, LB-BNN, RS-NN, LB-BNN Avg, CNN, E-CNN, EDL & [0.184, 0.195, 0.296, 0.309, 0.309, 0.402, 0.420, 0.481, 0.837, 0.909] \\ 
             & JS & RS-NN, CNN, DE Avg, DDU, LB-BNN Avg, LB-BNN, DE, CreINN, E-CNN, EDL & [0.068, 0.098, 0.099, 0.109, 0.119, 0.145, 0.168, 0.288, 0.856, 0.918] \\ \midrule
        0.5 & KL & DE Avg, DE, DDU, LB-BNN, CreINN, RS-NN, LB-BNN Avg, CNN, E-CNN, EDL & [0.195, 0.223, 0.309, 0.326, 0.356, 0.403, 0.420, 0.481, 0.998, 1.136] \\ 
             & JS & RS-NN, CNN, DE Avg, DDU, LB-BNN Avg, LB-BNN, DE, CreINN, E-CNN, EDL & [0.069, 0.098, 0.099, 0.109, 0.119, 0.161, 0.206, 0.348, 1.017, 1.145] \\ \midrule
        0.6 & KL & DE Avg, DE, DDU, LB-BNN, RS-NN, CreINN, LB-BNN Avg, CNN, E-CNN, EDL & [0.195, 0.261, 0.309, 0.342, 0.404, 0.415, 0.420, 0.481, 1.159, 1.363] \\ 
             & JS & RS-NN, CNN, DE Avg, DDU, LB-BNN Avg, LB-BNN, DE, CreINN, E-CNN, EDL & [0.070, 0.098, 0.099, 0.109, 0.119, 0.178, 0.245, 0.407, 1.178, 1.371] \\ \midrule
        0.7 & KL & DE Avg, DE, DDU, LB-BNN, RS-NN, LB-BNN Avg, CreINN, CNN, E-CNN, EDL & [0.195, 0.300, 0.309, 0.359, 0.405, 0.420, 0.475, 0.481, 1.319, 1.589] \\ 
             & JS & RS-NN, CNN, DE Avg, DDU, LB-BNN Avg, LB-BNN, DE, CreINN, E-CNN, EDL & [0.071, 0.098, 0.099, 0.109, 0.119, 0.194, 0.283, 0.467, 1.338, 1.598] \\ \midrule
        0.8 & KL & DE Avg, DDU, DE, LB-BNN, RS-NN, LB-BNN Avg, CNN, CreINN, E-CNN, EDL & [0.195, 0.309, 0.338, 0.376, 0.405, 0.420, 0.481, 0.535, 1.480, 1.816] \\ 
             & JS & RS-NN, CNN, DE Avg, DDU, LB-BNN Avg, LB-BNN, DE, CreINN, E-CNN, EDL & [0.072, 0.098, 0.099, 0.109, 0.119, 0.211, 0.322, 0.527, 1.499, 1.825] \\ \midrule
        0.9 & KL & DE Avg, DDU, DE, LB-BNN, RS-NN, LB-BNN Avg, CNN, CreINN, E-CNN, EDL & [0.195, 0.309, 0.377, 0.392, 0.406, 0.420, 0.481, 0.594, 1.641, 2.043] \\ 
             & JS & RS-NN, CNN, DE Avg, DDU, LB-BNN Avg, LB-BNN, DE, CreINN, E-CNN, EDL & [0.072, 0.098, 0.099, 0.109, 0.119, 0.228, 0.360, 0.586, 1.660, 2.052] \\ \midrule
        1.0 & KL & DE Avg, DDU, RS-NN, LB-BNN, DE, LB-BNN Avg, CNN, CreINN, E-CNN, EDL & [0.195, 0.309, 0.407, 0.409, 0.415, 0.420, 0.481, 0.654, 1.802, 2.270] \\ 
             & JS & RS-NN, CNN, DE Avg, DDU, LB-BNN Avg, LB-BNN, DE, CreINN, E-CNN, EDL & [0.073, 0.098, 0.099, 0.109, 0.119, 0.250, 0.392, 0.681, 1.810, 2.280] \\ 
        \bottomrule
    \end{tabular} }
\end{table}

\begin{figure}[!ht]
\centering
    \includegraphics[width=0.8\textwidth]{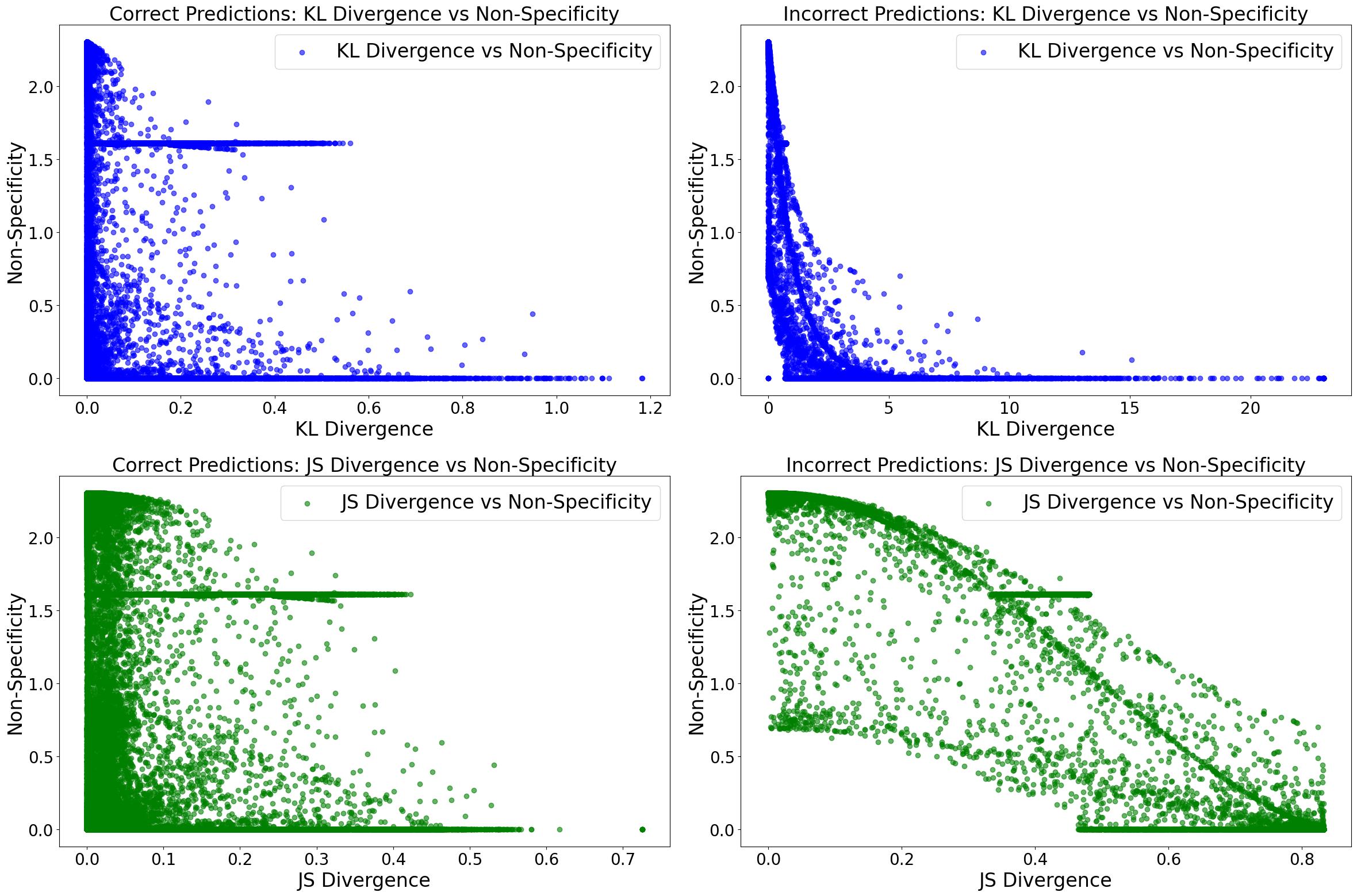}
    \caption{Scatter plots showing the relationship between uncertainty (KL and JS divergences) and non-specificity for correct and incorrect predictions across all models. {Notably, the distinct `spike’ at non-specificity just above 1.5 highlights the characteristic output of a particular model.}
    }\label{fig:kl-vs-js-vs-ns}
\end{figure}

It is worth noting that the rankings for different values of the trade-off parameter $\lambda$ vary. This variation arises from the lack of adjustment for the range of $\lambda$ across the different metrics. For instance, if the KL divergence ranges from 0 to 25, while the JS divergence spans from 0 to 1, we should consider scaling down the range of $\lambda$ for the JS divergence accordingly.
To ensure that the influence of the JS divergence is properly scaled in the overall evaluation metric, we can define a scaling factor based on their maximum values:
$\text{Scaling Factor} = \frac{\max(\text{KL})}{\max(\text{JS})} = \frac{25}{1} = 25.
$
Given that KL divergence will generally dominate due to its larger range, we should scale down $\lambda$ for the JS divergence. A suitable adjusted range for $\lambda$ in the context of JS divergence can be expressed as: $
\lambda_{JS} = \lambda \times \left(\frac{1}{\text{Scaling Factor}}\right) = \lambda \times \frac{1}{25}.
$
Thus, while $\lambda$ initially ranges from 0 to 1, the effective range for $\lambda_{JS}$ will be from 0 to $\frac{1}{25}$. This adjustment ensures that both KL and JS divergences contribute appropriately to the overall evaluation metric $\mathcal{E}$, reflecting their respective scales during model selection.


In Fig. \ref{fig:kl-vs-js-vs-ns}, the set of scatter plots visualizes the relationship between distance measures (KL and JS divergences) and non-specificity for correct (\textit{left}) and incorrect (\textit{right}) predictions {across all models}. The top row focuses on KL divergence, showing how it correlates with non-specificity for both correct and incorrect predictions. The bottom row highlights JS divergence with the same comparisons. {Both KL and JS behave similarly for correct predictions. A straight horizontal line (spike) around NS $\approx$ 1.6 appears with varying KL and JS values, likely reflecting the behavior of E-CNN and EDL.  Similarly, the concentrated spots in incorrect predictions may correspond to these models. As NS increases, KL decreases in all the plots, likely because larger credal sets at higher NS include vertices closer to the ground truth.}

\textbf{Why choose KL divergence over JS divergence?} We selected KL divergence for our measure because, as shown in Fig. \ref{fig:kl-js-eval}, JS is symmetric and bounded while KL divergence can grow unbounded when there is a significant mismatch between the predicted and true distributions, especially in small probabilities, making it more effective in distinguishing between correct and incorrect predictions. KL's greater sensitivity can provide a clearer signal when small deviations in probabilities matter, offering sharper differentiation in model evaluation. As Fig. \ref{fig:kl-js-eval} shows, KL offers a clearer separation between the two, with the plot showing the mean KL and JS values. This trend is further emphasized in Fig. \ref{fig:kl-js} where the full distributions of KL and JS are shown. Note that in Fig. \ref{fig:kl-js}, the ranges on the Y-axis differ significantly, with KL values spanning from 0 to 25, whereas JS values range only from 0 to 1.

\subsubsection{Ablation study on different non-specificity measures}  \label{app:ablation-ns}

In this section, we present an ablation study on the non-specificity (NS) measure in our evaluation metric $\mathcal{E}$. Non-specificity is in fact an epistemic uncertainty measure of the credal set, and can be replaced with other suitable uncertainty measures related to credal sets. In this ablation, we present an alternative to the \citet{dubois1987properties} equation (Eq. \ref{eq:non_spec}) that we use, in the form of a different measure of credal uncertainty (CU) \citep{wang2024credal}.

This credal uncertainty (CU) measure can be computed as the difference between the credal upper bound $\overline{H}(\hat{\mathbb{C}re})$ and lower bounds $\underline{H}(\hat{\mathbb{C}re})$ computed as follows.
Given a set of $K$ individual predictive distributions from Bayesian Neural Networks, evidential models or deep ensembles, we can obtain an upper and lower probability bound for the $c_i$-th class element, denoted as $\overline{P}_{c_i}$ and $\underline{P}_{c_i}$, respectively. Recall that $\mathbf{Y} = \{ c_i, i \}$
where $i = 1, 2, \ldots, N$, and $N$ is the number of classes in $\mathcal{C}$.

These bounds can be computed as follows: $\overline{P}_{c_i} = \max \limits_{k=1,..,K} p_{k,c_i}, \quad \underline{P}_{c_i} = \min \limits_{k=1,..,K} p_{k,c_i}$,
where $p_{k,c_i}$ denotes the $c_i$-th class element of the $k$-th single probability vector $\mathbf{p}_k$. Such probability intervals over $\mathcal{C}$ classes define a non-empty credal set $\hat{\mathbb{C}re}$:
\begin{equation}
\hat{\mathbb{C}re} = \{ \mathbf{p} \mid \underline{P}_{c_i} \leq p_{c_i} \leq \overline{P}_{c_i}, \, \forall c_i \in \mathcal{C}, i = 1, 2, \ldots, N \}
\end{equation}
with the normalization condition: $\sum_{i=1}^{N} \underline{P}_{c_i} \leq 1 \leq \sum_{i=1}^{N} \overline{P}_{c_i}$.

It can be readily shown that the probability intervals given in the previous equations satisfy this condition, as follows:
\begin{equation}
\sum_{i=1}^{N} \underline{P}_{c_i} = \sum_{i=1}^{N} \min \limits_{k=1,..,K} p_{k,{c_i}} \leq \sum_{i=1}^{N} p_{k^*,{c_i}} = 1 \leq \sum_{i=1}^{N} \max \limits_{k=1,..,K} p_{k,{c_i}} = \sum_{i=1}^{N} \overline{P}_{c_i},
\end{equation}
where $k^*$ is any index in $1, \ldots, K$ and $c_i \in \mathcal{C}$. To estimate the uncertainty using $H(\hat{\mathbb{C}re})$ and $H(\hat{\mathbb{C}re})$, we need to solve the following optimization problems:

\begin{equation}
    \overline{H}(\hat{\mathbb{C}re}) = \operatorname{maximize} \sum_{i=1}^{N} -p_{c_i} \log_2 p_{c_i}, \quad
\underline{H}(\hat{\mathbb{C}re}) = \operatorname{minimize} \sum_{i=1}^{N} -p_{c_i} \log_2 p_{c_i},
\end{equation}
subject to the constraints: $
\sum_{i=1}^{N} p_{c_i} = 1, \quad p_{c_i} \in [\underline{P}_{c_i}, \overline{P}_{c_i}], \, \forall {c_i}$.

These optimization problems can be addressed using standard solvers, such as the SciPy optimization package \citep{virtanen2020scipy}.

\begin{figure}[!ht]
    \begin{minipage}[t]{0.48\textwidth}
    \includegraphics[width=\textwidth]{images/eval/CIFAR10/ns_cifar10.pdf}
    \vspace{-25pt}
    \caption*{(a)}
    \end{minipage} \hspace{0.02\textwidth}
    \begin{minipage}[t]{0.48\textwidth}
    \includegraphics[width=\textwidth]{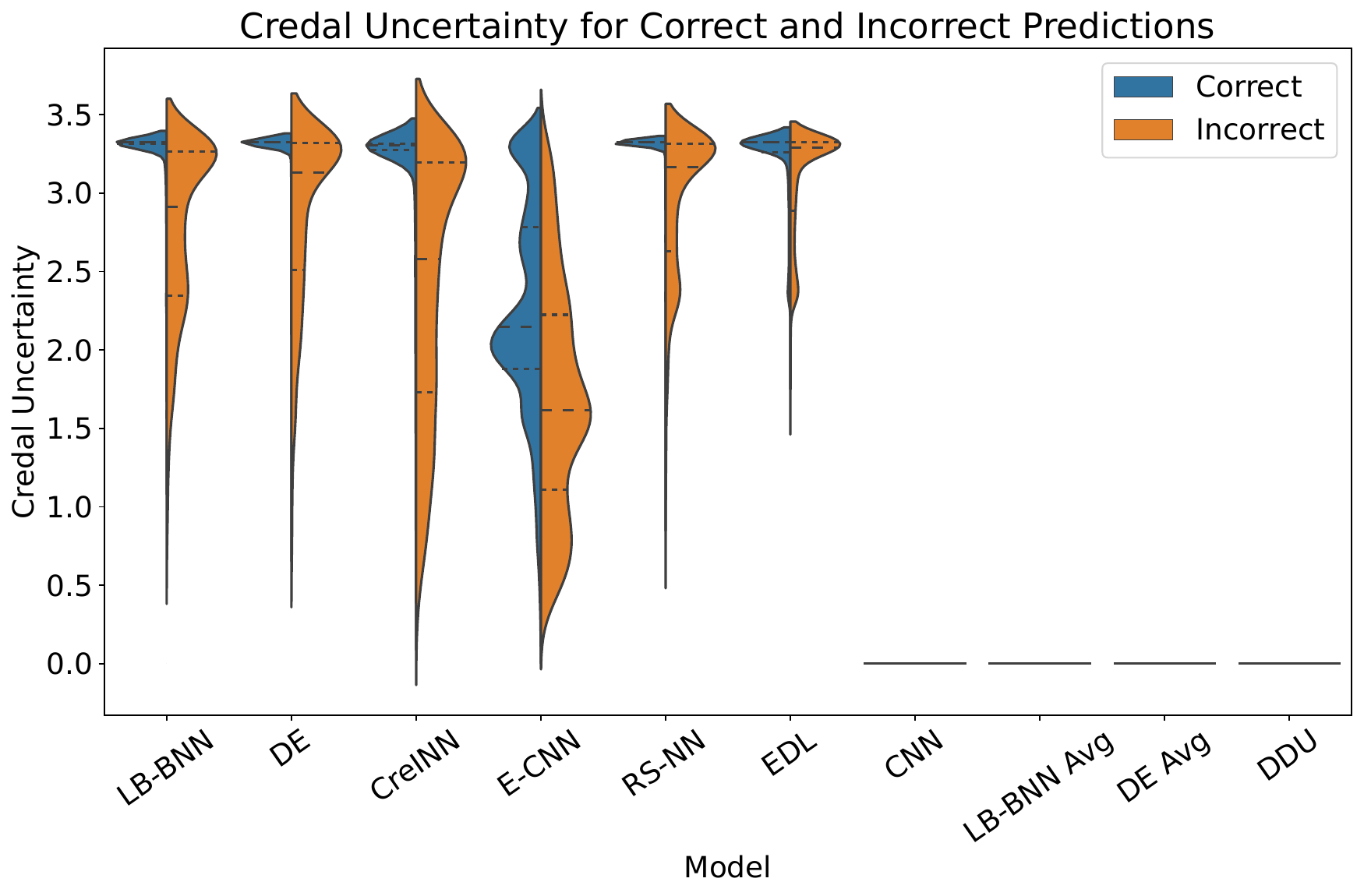}
    \vspace{-25pt}
    \caption*{(b)}
    \end{minipage}
            \vspace{-10pt}
    \caption{Comparison of (a) Non-Specificity (NS), and (b) Credal Uncertainty (CU) for Correctly Classified (CC) and Incorrectly Classified (ICC) samples from the CIFAR-10 dataset, for all models considered here. Notably, the scales of these two measures differ (Y-axis).
    }\label{fig:ns-cu}
\end{figure}

\begin{figure}[!h]
    \includegraphics[width=\textwidth]{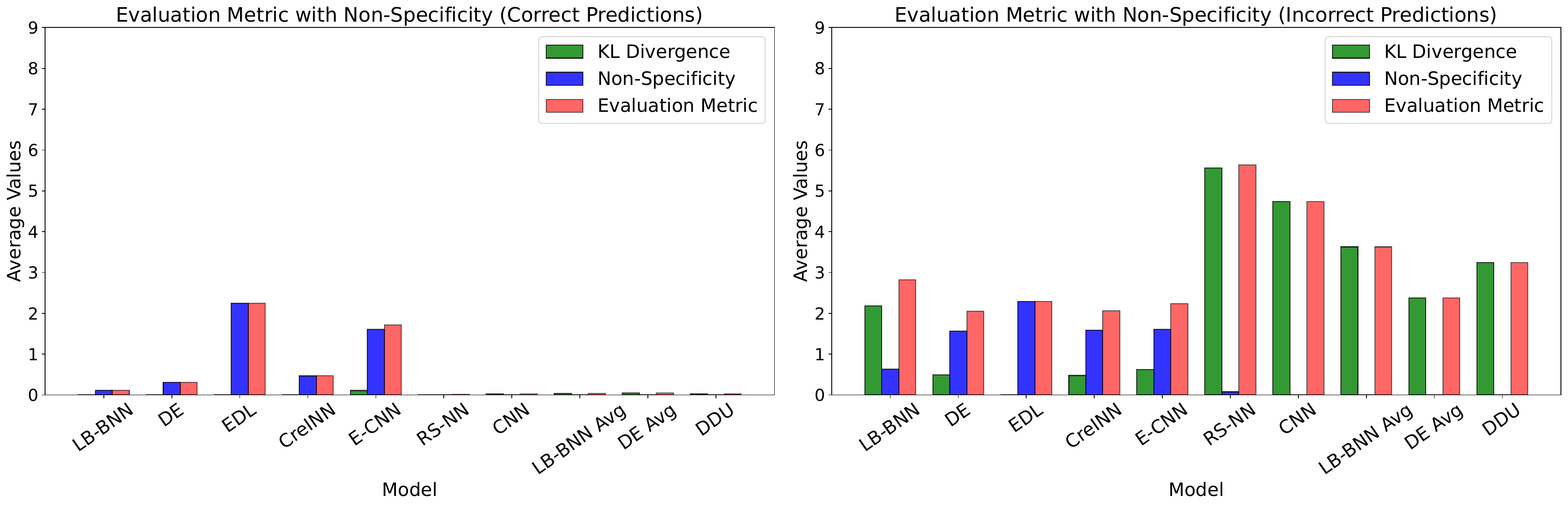}
    \includegraphics[width=\textwidth]{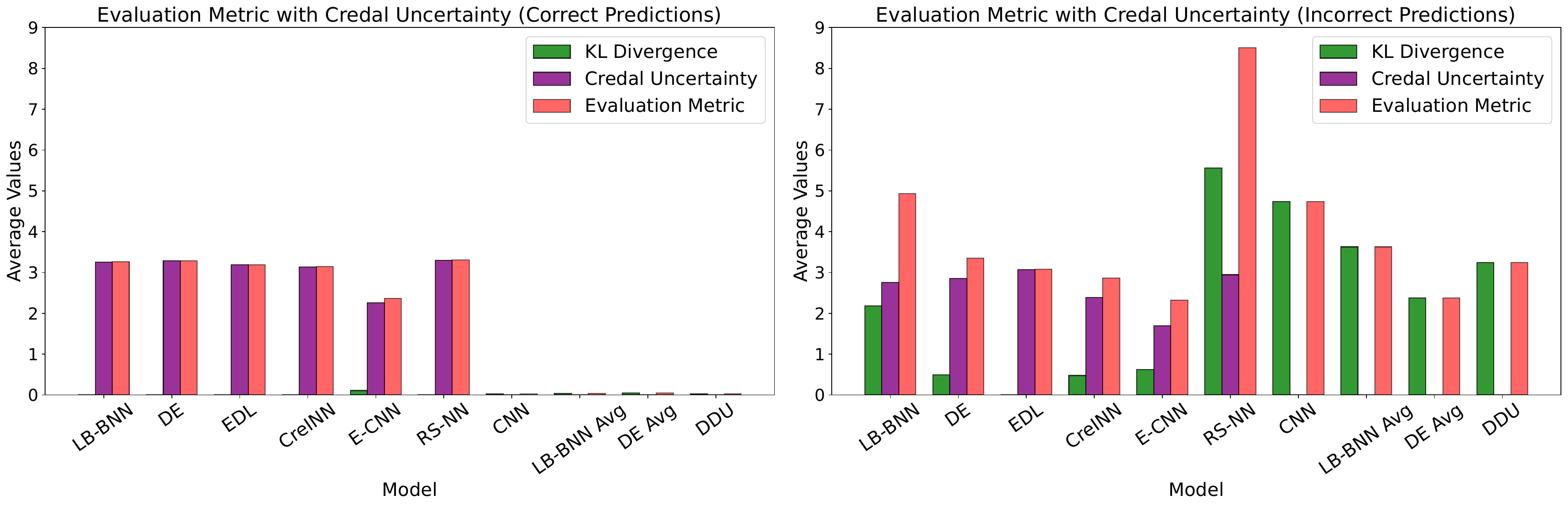}
    \vspace{-15pt}
    \caption{Comparison of mean Evaluation Metric $\mathcal{E}$ using mean Non-Specificity (top) and mean Credal Uncertainty (bottom) for Correct (left) and Incorrect (right) predictions of the CIFAR-10 dataset.
    }\label{fig:kl-ns-cu-bar}
            \vspace{-10pt}
\end{figure}

Fig. \ref{fig:ns-cu} shows the Non-Specificity (NS) (\ref{fig:ns-cu}(a)) and Credal Uncertainty (CU) (\ref{fig:ns-cu}(b)) over the entire test set. The violin plots depict distributions of the NS and CU for correct (blue) and incorrect (orange) predictions. We expect the both the measures to be larger for incorrect predictions and smaller for correct predictions. The plots show how the range of both values differ, but the trend remains the same, except for E-CNN. Non-specificity for E-CNN is high for both correct and incorrect predictions. 

Fig. \ref{fig:kl-ns-cu-bar} shows bar plots comparing Non-Specificity (NS) and Credal Uncertainty (CU) along with their influence on the overall evaluation metric. All values are shown as means. As discussed earlier, having larger upper bounds even for correct predictions often cause Credal Uncertainty to make the overall Evaluation Metric $\mathcal{E}$ larger. 

\begin{table}[!h]
    \centering
    \caption{Model Rankings Based on Non-Specificity (NS) and Credal Uncertainty (CU) on the CIFAR-10 dataset. Model selection is based on the mean of Evaluation Metric ($\mathcal{E}$) with the models with the lowest $\mathcal{E}$ ranking first. The distance metric used here is KL divergence.}
            \vspace{-8pt}
    \label{tab:model_rank_ablation_ns}
    \resizebox{\textwidth}{!}{
    \begin{tabular}{@{}cc|l|l@{}}
        \toprule
        \textbf{Lambda} & \textbf{Metric} & \textbf{Model Ranking} & \textbf{Evaluation Metric ($\mathcal{E}$) Mean} \\ \midrule
        0.1 & NS & DE, CreINN, DE Avg, EDL, LB-BNN, DDU, E-CNN, RS-NN, LB-BNN Avg, CNN & [0.069, 0.117, 0.195, 0.229, 0.259, 0.309, 0.354, 0.399, 0.420, 0.481] \\ 
             & CU & DE Avg, DDU, EDL, DE, CreINN, E-CNN, LB-BNN Avg, CNN, LB-BNN, RS-NN &
[0.195, 0.309, 0.316, 0.357, 0.363, 0.410, 0.420, 0.481, 0.563, 0.726] \\ \midrule
        0.2 & NS & DE, CreINN, DE Avg, LB-BNN, DDU, RS-NN, LB-BNN Avg, EDL, CNN, E-CNN & [0.108, 0.177, 0.195, 0.276, 0.309, 0.400, 0.420, 0.456, 0.481, 0.515] \\ 
             & CU & DE Avg, DDU, LB-BNN Avg, CNN, E-CNN, EDL, CreINN, DE, LB-BNN, RS-NN & [0.195, 0.309, 0.420, 0.481, 0.627, 0.631, 0.668, 0.683, 0.883, 1.053]
\\ \midrule
        0.3 & NS & DE, DE Avg, CreINN, LB-BNN, DDU, RS-NN, LB-BNN Avg, CNN, E-CNN, EDL & [0.146, 0.195, 0.237, 0.293, 0.309, 0.401, 0.420, 0.481, 0.676, 0.682] \\ 
             & CU & DE Avg, DDU, LB-BNN Avg, CNN, E-CNN, EDL, CreINN, DE, LB-BNN, RS-NN &
[0.195, 0.309, 0.420, 0.481, 0.844, 0.945, 0.973, 1.009, 1.202, 1.380] \\ \midrule
        0.4 & NS & DE, DE Avg, CreINN, DDU, LB-BNN, RS-NN, LB-BNN Avg, CNN, E-CNN, EDL & [0.184, 0.195, 0.296, 0.309, 0.309, 0.402, 0.420, 0.481, 0.837, 0.909] \\ 
             & CU & DE Avg, DDU, LB-BNN Avg, CNN, E-CNN, EDL, CreINN, DE, LB-BNN, RS-NN &
[0.195, 0.309, 0.420, 0.481, 1.061, 1.259, 1.279, 1.335, 1.522, 1.707] \\ \midrule
        0.5 & NS & DE Avg, DE, DDU, LB-BNN, CreINN, RS-NN, LB-BNN Avg, CNN, E-CNN, EDL & [0.195, 0.223, 0.309, 0.326, 0.356, 0.403, 0.420, 0.481, 0.998, 1.136] \\ 
             & CU & DE Avg, DDU, LB-BNN Avg, CNN, E-CNN, EDL, CreINN, DE, LB-BNN, RS-NN & [0.195, 0.309, 0.420, 0.481, 1.277, 1.573, 1.584, 1.661, 1.842, 2.034]
\\ \midrule
        0.6 & NS & DE Avg, DE, DDU, LB-BNN, RS-NN, CreINN, LB-BNN Avg, CNN, E-CNN, EDL & [0.195, 0.261, 0.309, 0.342, 0.404, 0.415, 0.420, 0.481, 1.159, 1.363] \\ 
             & CU & DE Avg, DDU, LB-BNN Avg, CNN, E-CNN, EDL, CreINN, DE, LB-BNN, RS-NN &
[0.195, 0.309, 0.420, 0.481, 1.494, 1.888, 1.889, 1.987, 2.162, 2.362] \\ \midrule
        0.7 & NS & DE Avg, DE, DDU, LB-BNN, RS-NN, LB-BNN Avg, CreINN, CNN, E-CNN, EDL & [0.195, 0.300, 0.309, 0.359, 0.405, 0.420, 0.475, 0.481, 1.319, 1.589] \\ 
             & CU & DE Avg, DDU, LB-BNN Avg, CNN, E-CNN, CreINN, EDL, DE, LB-BNN, RS-NN &
[0.195, 0.309, 0.420, 0.481, 1.711, 2.194, 2.202, 2.313, 2.482, 2.689] \\ \midrule
        0.8 & NS & DE Avg, DDU, DE, LB-BNN, RS-NN, LB-BNN Avg, CNN, CreINN, E-CNN, EDL & [0.195, 0.309, 0.338, 0.376, 0.405, 0.420, 0.481, 0.535, 1.480, 1.816] \\ 
             & CU & DE Avg, DDU, LB-BNN Avg, CNN, E-CNN, CreINN, EDL, DE, LB-BNN, RS-NN &
[0.195, 0.309, 0.420, 0.481, 1.928, 2.499, 2.516, 2.639, 2.802, 3.016] \\ \midrule
        0.9 & NS & DE Avg, DDU, DE, LB-BNN, RS-NN, LB-BNN Avg, CNN, CreINN, E-CNN, EDL & [0.195, 0.309, 0.377, 0.392, 0.406, 0.420, 0.481, 0.594, 1.641, 2.043] \\ 
             & CU & DE Avg, DDU, LB-BNN Avg, CNN, E-CNN, CreINN, EDL, DE, LB-BNN, RS-NN &
[0.195, 0.309, 0.420, 0.481, 2.145, 2.805, 2.830, 2.965, 3.122, 3.343] \\ \midrule
        1.0 & NS & DE Avg, DDU, RS-NN, LB-BNN, DE, LB-BNN Avg, CNN, CreINN, E-CNN, EDL & [0.195, 0.309, 0.407, 0.409, 0.415, 0.420, 0.481, 0.654, 1.802, 2.270] \\ 
             & CU & DE Avg, DDU, LB-BNN Avg, CNN, E-CNN, CreINN, EDL, DE, LB-BNN, RS-NN &
[0.195, 0.309, 0.420, 0.481, 2.362, 3.110, 3.144, 3.292, 3.442, 3.671]
 \\ 
        \bottomrule
    \end{tabular} }
            \vspace{-10pt}
\end{table}

\textbf{Model Selection.} Tab. \ref{tab:model_rank_ablation_ns}, presents the rankings of various models based on their performance in terms of Non-Specificity (NS) and Credal Uncertainty (CU) on the CIFAR-10 dataset. The rankings are derived from the Evaluation Metric ($\mathcal{E}$), where models are ordered by their mean $\mathcal{E}$ values. The models are assessed under different values of the parameter lambda, with lower $\mathcal{E}$ values indicating better performance. For instance, at a lambda value of 0.1, the top-ranked models for Non-Specificity are DE, CreINN, and DE Avg. Similarly, for Credal Uncertainty, DE Avg and DDU lead the rankings, emphasizing their effectiveness in managing credal uncertainty within the predictions.

As lambda increases, the rankings reveal variations in model performance across both metrics. Notably, while DE Avg and DDU consistently ranks highly across different lambda values, other models exhibit fluctuating positions. If we do not consider models that predict point predictions, since their NS and CU are zero and $\mathcal{E}$ is the same as KL, DE and E-CNN are the best ranked models for CU and DE and RS-NN are the best ranked models for NS.





\textbf{Why choose Non-Specificity (NS) over Credal Uncertainty?} We chose Non-Specificity (NS) over Credal Uncertainty (CU) for our measure because, as demonstrated in Fig. \ref{fig:ns-cu}, CU struggles to differentiate the uncertainty between correct and incorrect predictions across all models. The mean CU remains high for both cases, as highlighted in Fig. \ref{fig:kl-ns-cu-bar}. In contrast, NS offers a clearer distinction, with lower values for correct predictions and higher values for incorrect ones. CU, on the other hand, tends to exhibit a high upper bound for both correct and incorrect predictions, and remains consistently elevated for correct predictions regardless of the model's performance. This does not accurately reflect the true state of uncertainty within the credal set.

\subsubsection{Distance measure \textit{vs.} Confidence score}

The choice of using distance instead of confidence in the proposed metric is motivated by the goal to capture the true nature of the prediction within the simplex and to properly characterize the credal set.
A key issue with using confidence (the maximum predicted probability) is that many models, particularly standard neural networks, exhibit a problem of overconfidence even for wrong predictions. In contrast, models like Deep Ensembles are designed to mitigate overconfidence by averaging predictions across different models. By using a distance metric, we can better capture the spread or diversity in the predicted probabilities across the output space instead of focusing solely on the confidence of predictions.

While predictive entropy is a well-known and effective method for quantifying total uncertainty, it is not useful here as we are mainly concerned with epistemic uncertainty. For instance, if infinite amounts of data were available {(assuming this data is unbiased and free from measurement errors or artifacts)}, epistemic uncertainty (and credal set size) would reduce to zero whereas entropy will not necessarily be zero unless we have a fully confident prediction (100\% confidence).

\begin{figure}[!h]
    \centering
    \begin{minipage}[t]{0.3\textwidth}
    \includegraphics[width=\textwidth]{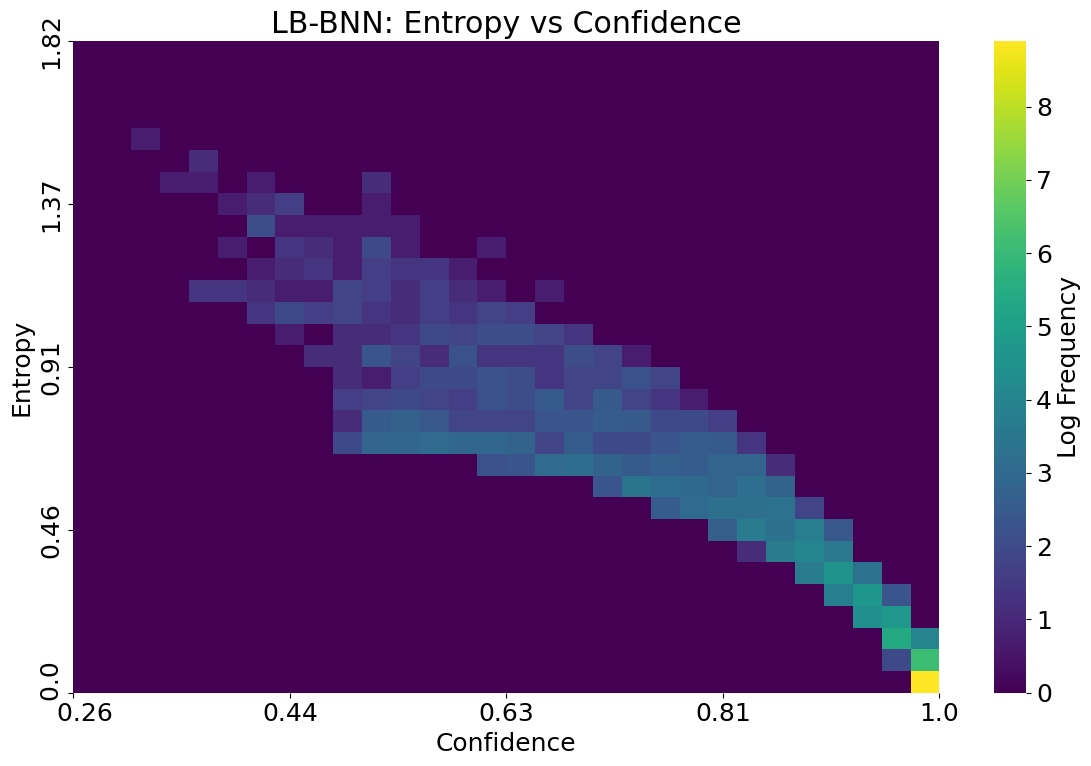}
    \end{minipage} \hspace{0.01\textwidth}
    \begin{minipage}[t]{0.3\textwidth}
    \includegraphics[width=\textwidth]{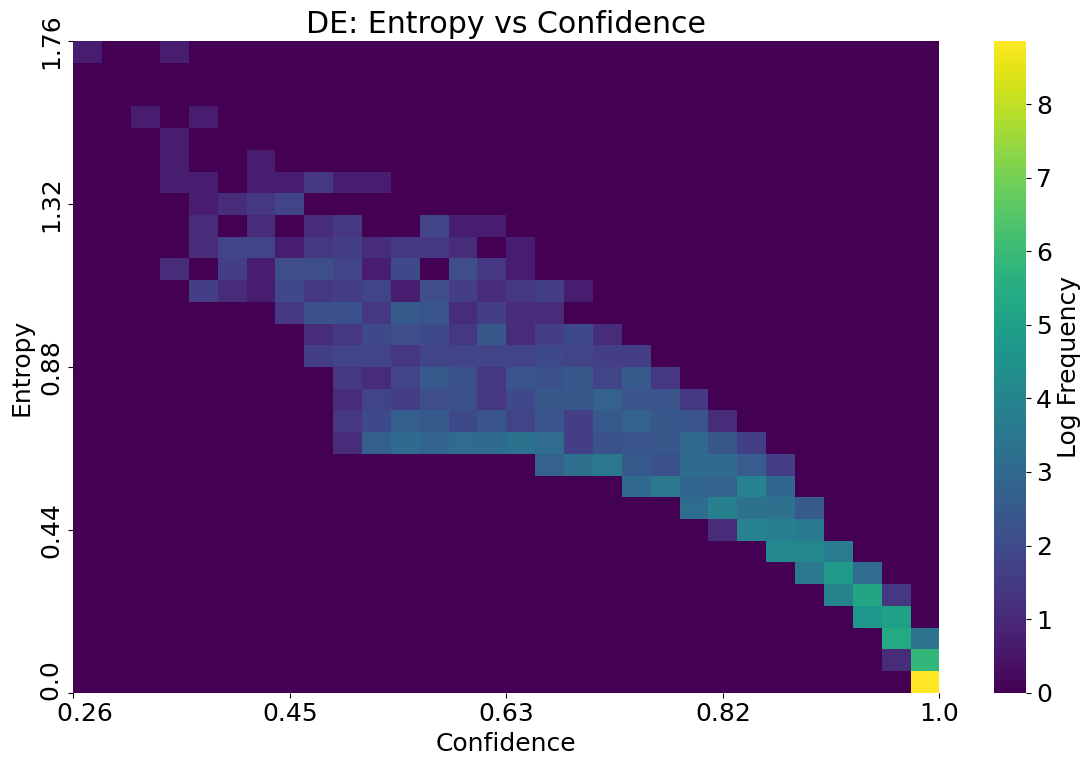}
    \end{minipage} \hspace{0.01\textwidth}
        \begin{minipage}[t]{0.3\textwidth}
    \includegraphics[width=\textwidth]{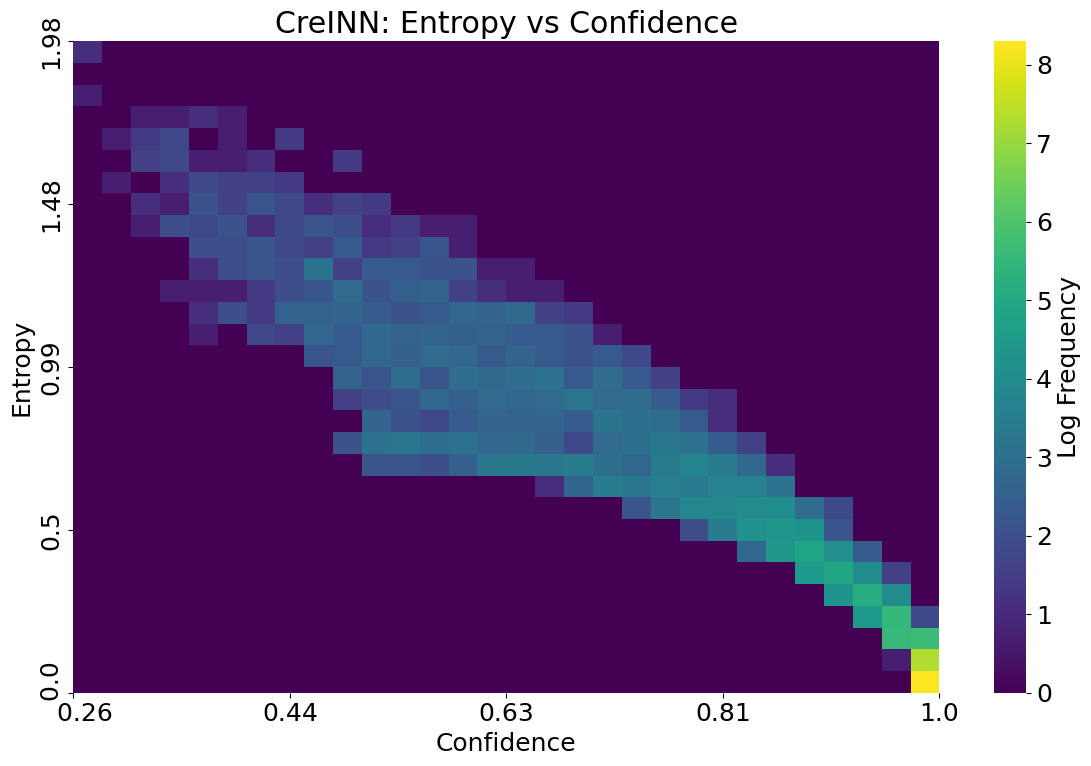}
    \end{minipage} \hspace{0.01\textwidth}
\\
    \begin{minipage}[t]{0.3\textwidth}
    \includegraphics[width=\textwidth]{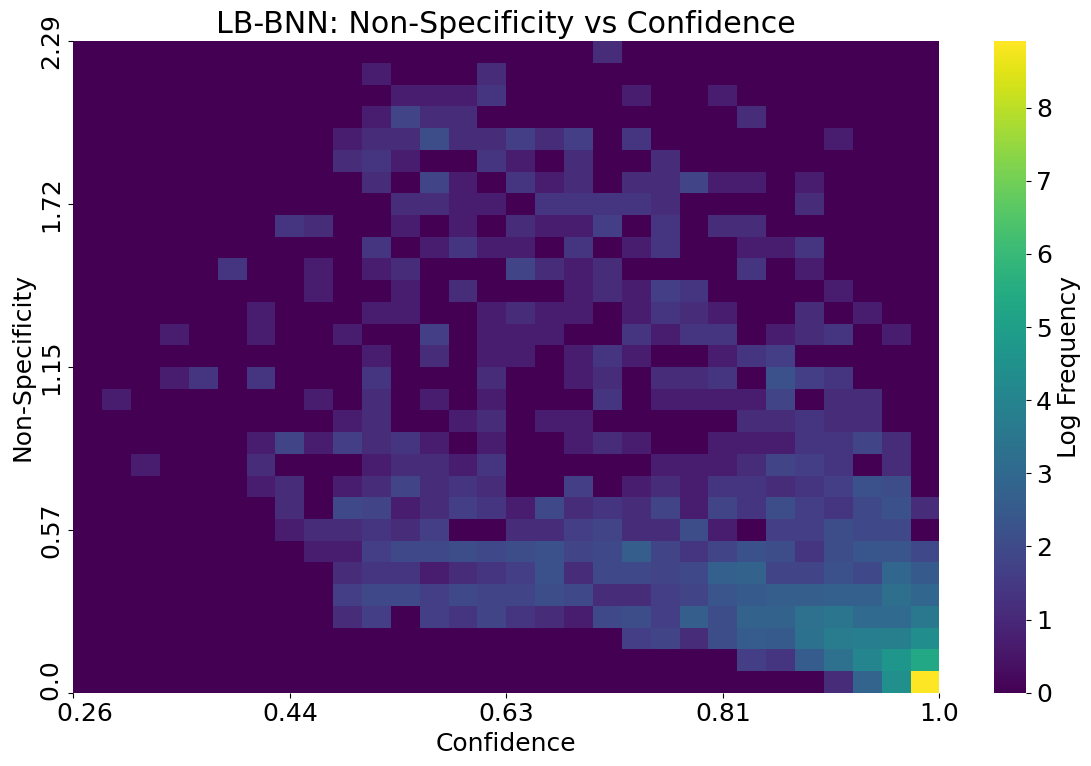}
        \caption*{(a) LB-BNN}
    \end{minipage} \hspace{0.01\textwidth}
        \begin{minipage}[t]{0.3\textwidth}
    \includegraphics[width=\textwidth]{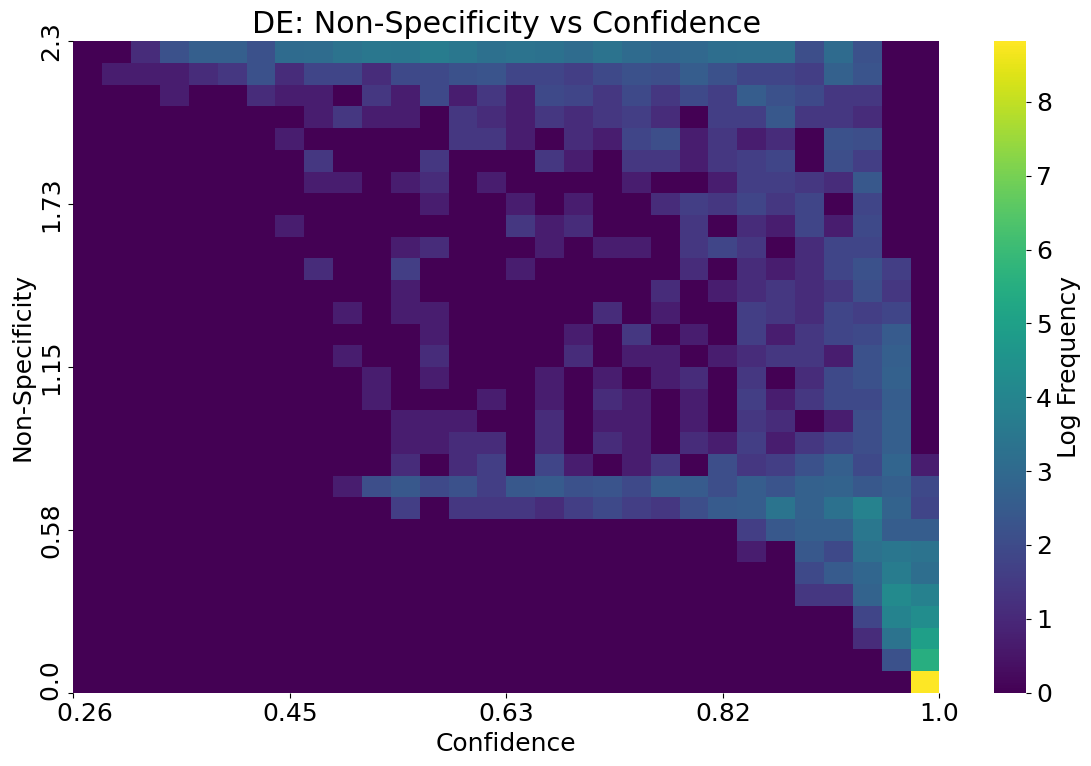}
            \caption*{(b) DE}
    \end{minipage}\hspace{0.01\textwidth}
        \begin{minipage}[t]{0.3\textwidth}
    \includegraphics[width=\textwidth]{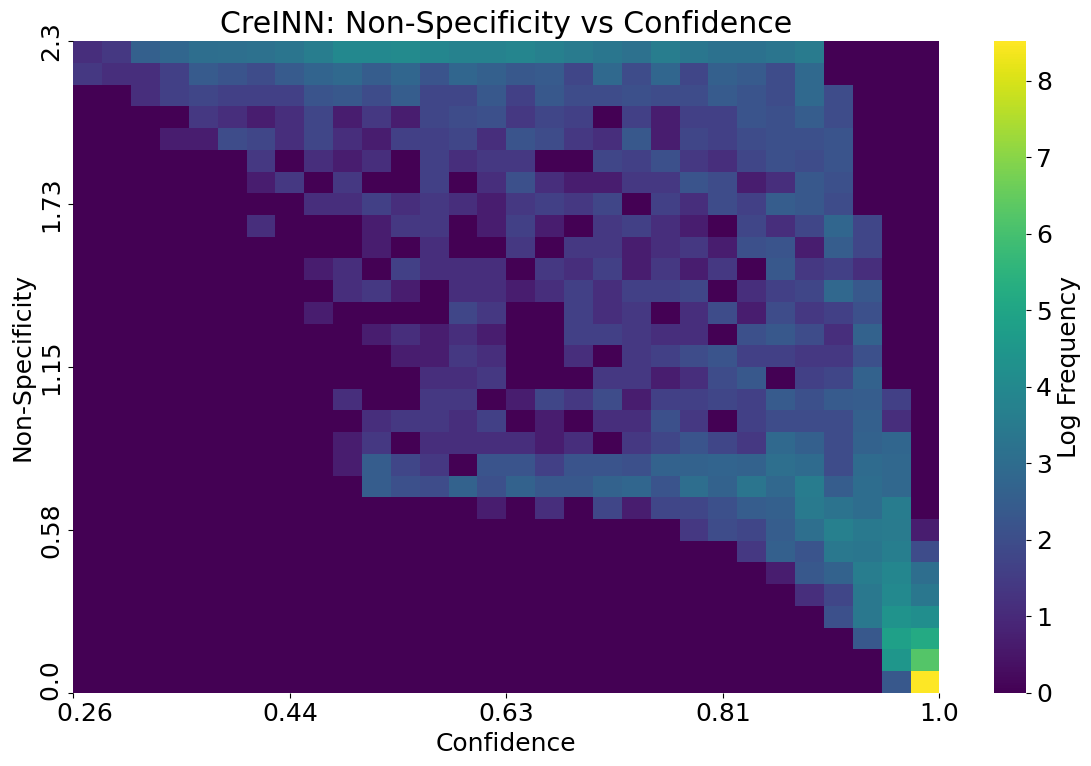}
            \caption*{(c) CreINN}
    \end{minipage}
    \\ 
        \begin{minipage}[t]{0.3\textwidth}
    \includegraphics[width=\textwidth]{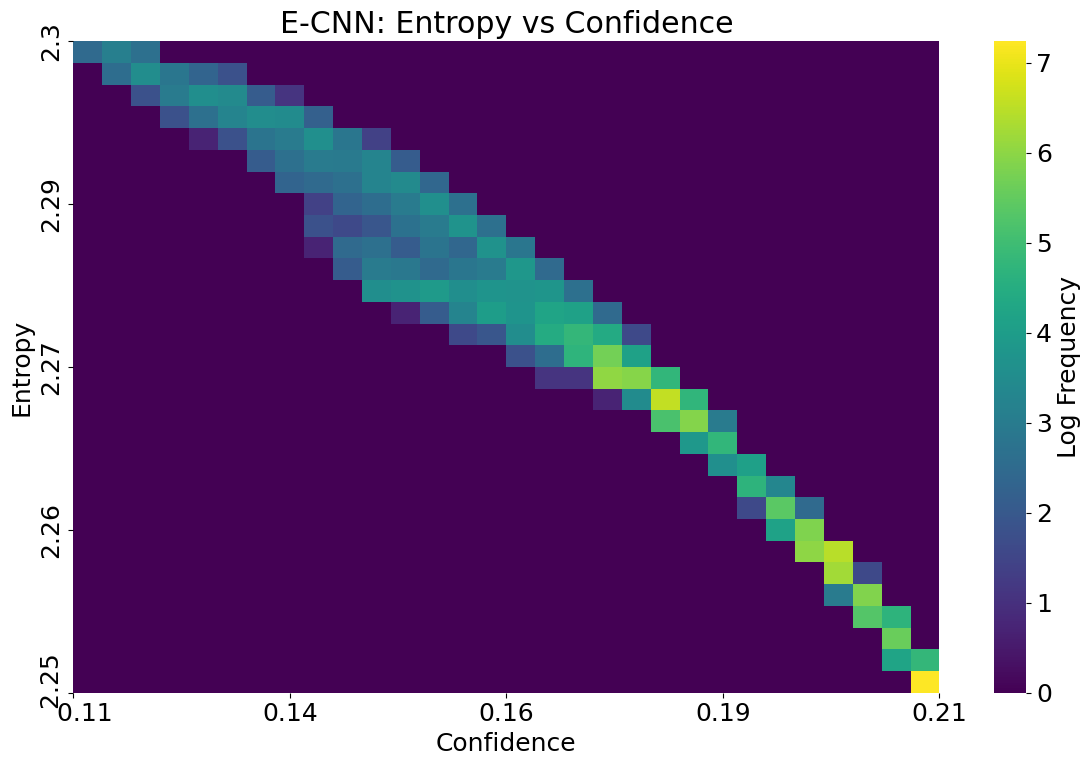}
    \end{minipage} \hspace{0.01\textwidth}
            \begin{minipage}[t]{0.3\textwidth}
    \includegraphics[width=\textwidth]{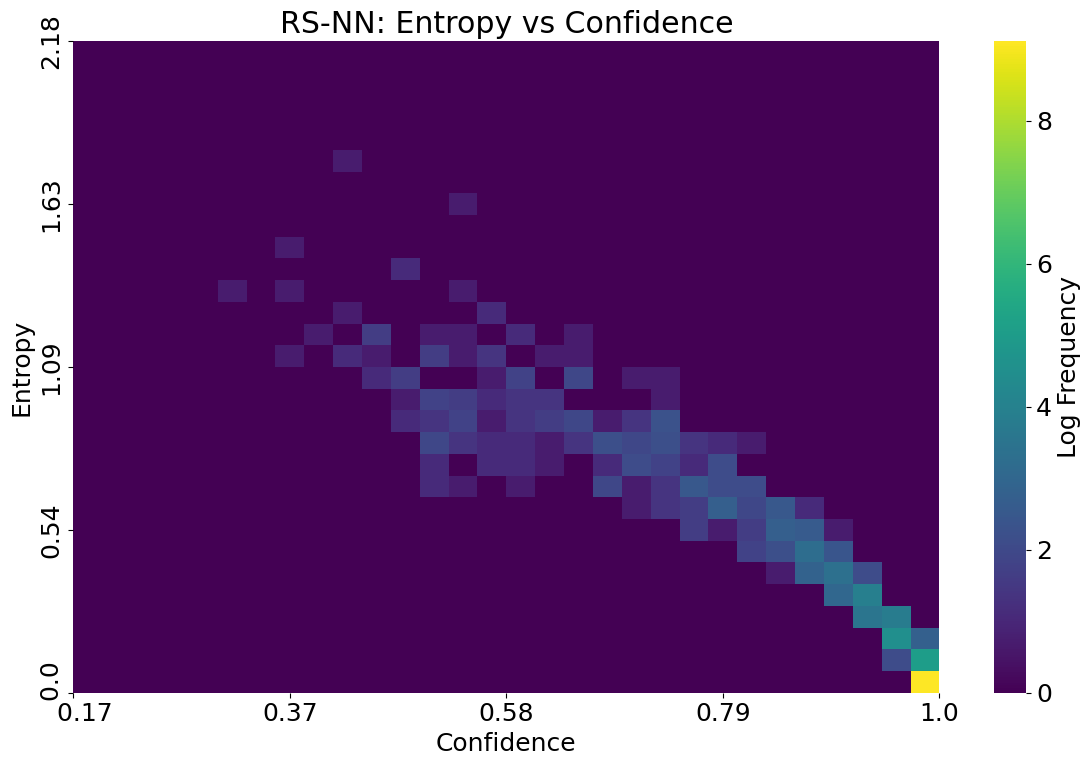}
    \end{minipage} \hspace{0.01\textwidth}
            \begin{minipage}[t]{0.3\textwidth}
    \includegraphics[width=\textwidth]{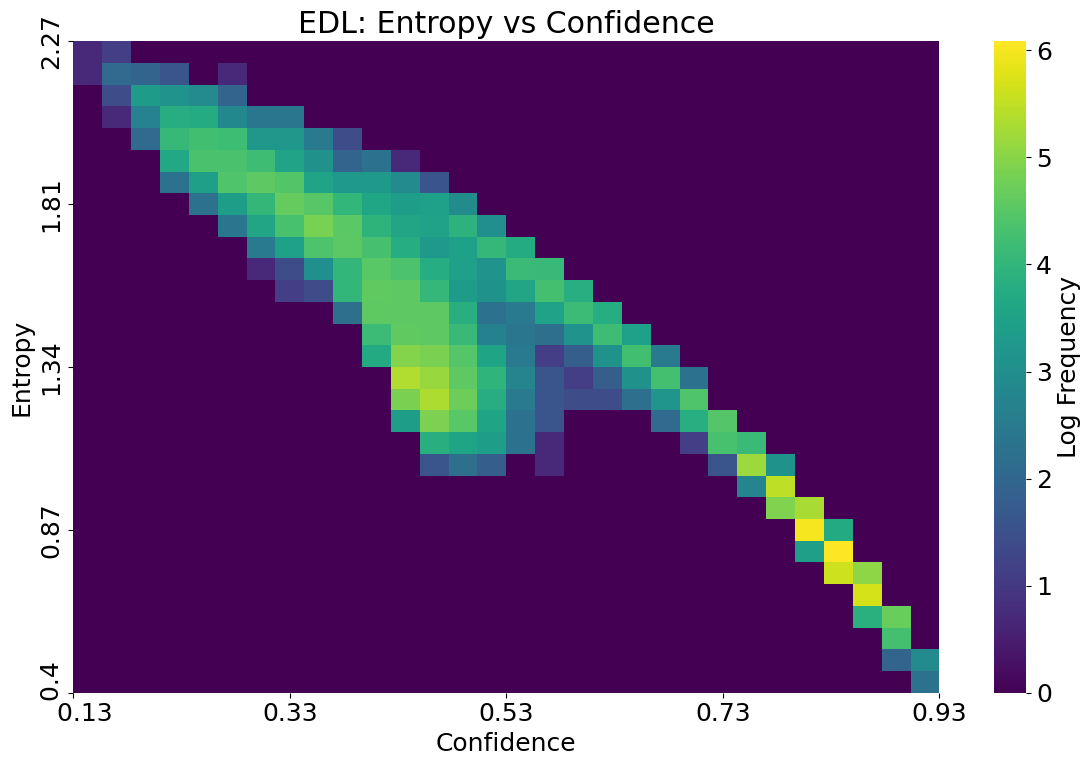}
    \end{minipage} 
    \\
        \begin{minipage}[t]{0.3\textwidth}
    \includegraphics[width=\textwidth]{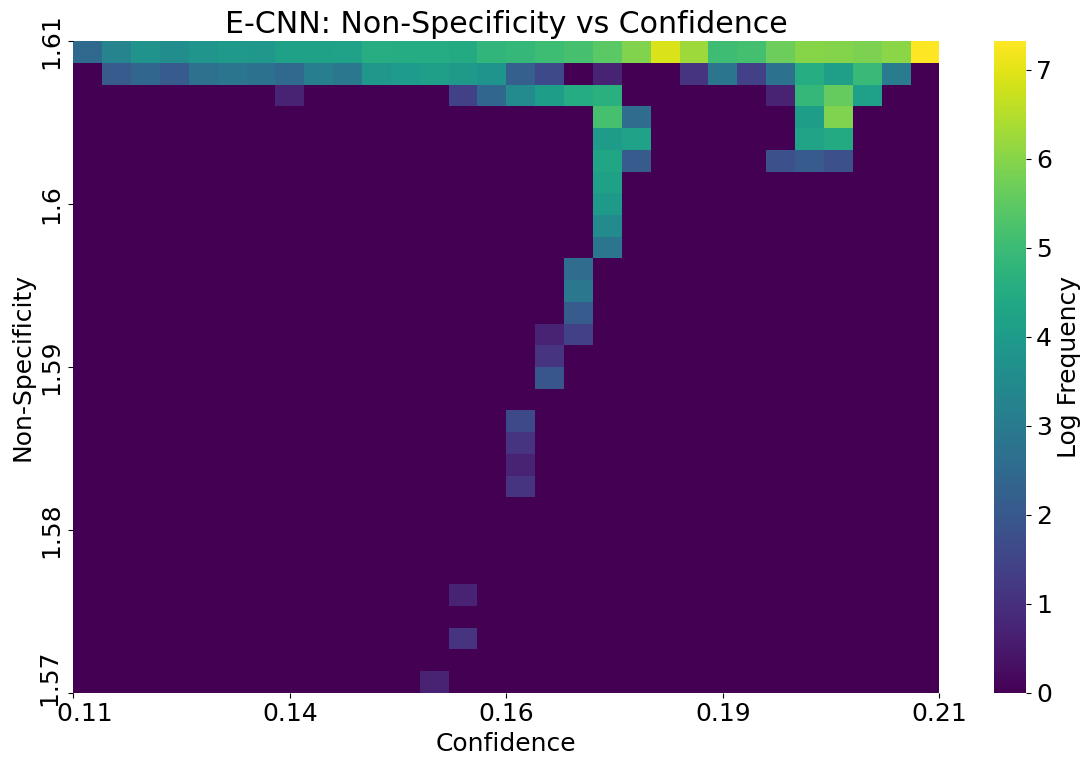}
    \caption*{(d) E-CNN}
    \end{minipage}\hspace{0.01\textwidth}
        \begin{minipage}[t]{0.3\textwidth}
    \includegraphics[width=\textwidth]{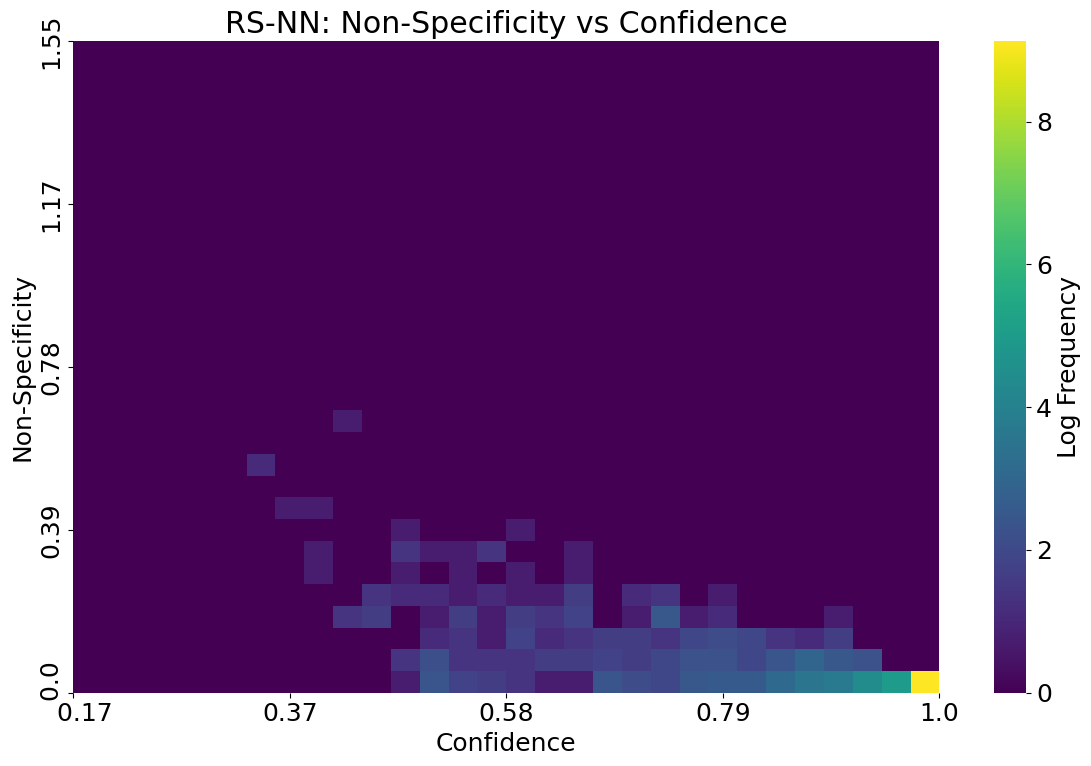}
    \caption*{(e) RS-NN}
    \end{minipage}\hspace{0.01\textwidth}
        \begin{minipage}[t]{0.3\textwidth}
    \includegraphics[width=\textwidth]{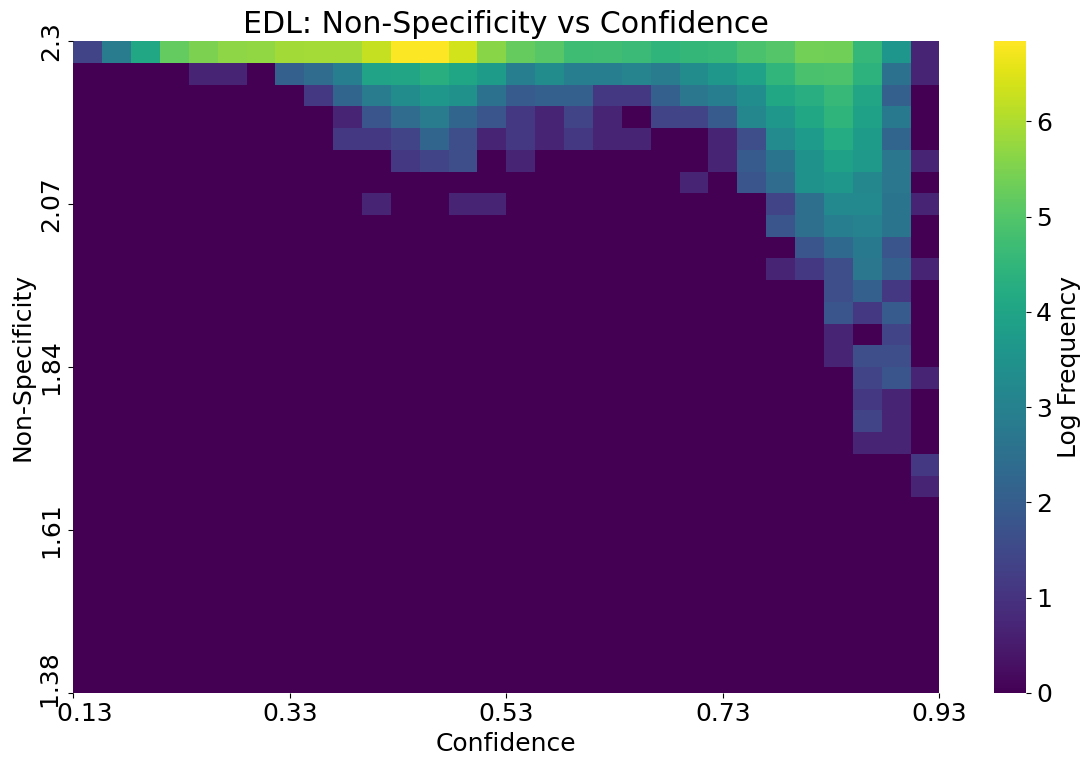}
    \caption*{(f) EDL}
    \end{minipage}
    \caption{Entropy \textit{vs.} Confidence (top) and Non-Specificity \textit{vs.} Confidence (bottom) for each of the models: (a) LB-BNN, (b) Deep Ensembles (DE), (c) CreINN, (d) E-CNN, (e) RS-NN and (f) EDL on the CIFAR-10 dataset. Entropy is strongly linked to confidence, whereas non-specificity shows minimal correlation with it.
}\label{fig:kl_vs_confidence}
\end{figure}

Furthermore, as illustrated in Fig. \ref{fig:kl_vs_confidence} below, entropy is closely tied to confidence, while non-specificity has little correlation with confidence. High confidence predictions lead to low entropy, while low confidence predictions result in higher entropy. However, when considering epistemic uncertainty, entropy alone may not provide a complete picture. It is also detailed in \citet{song2018evidence} (Sec. 3, page 5) that the entropy-based uncertainty measure mainly depends on the total distribution of probabilities without concern about the largest probability, whereas the non-specificity is related to the difference between the largest probability and other ones.

\subsubsection{Credal set size vs. Non-specificity} \label{app:credal_vs_non_specificity}

\begin{figure}[!h]
    \centering
    \begin{minipage}[t]{0.3\textwidth}
    \includegraphics[width=\textwidth]{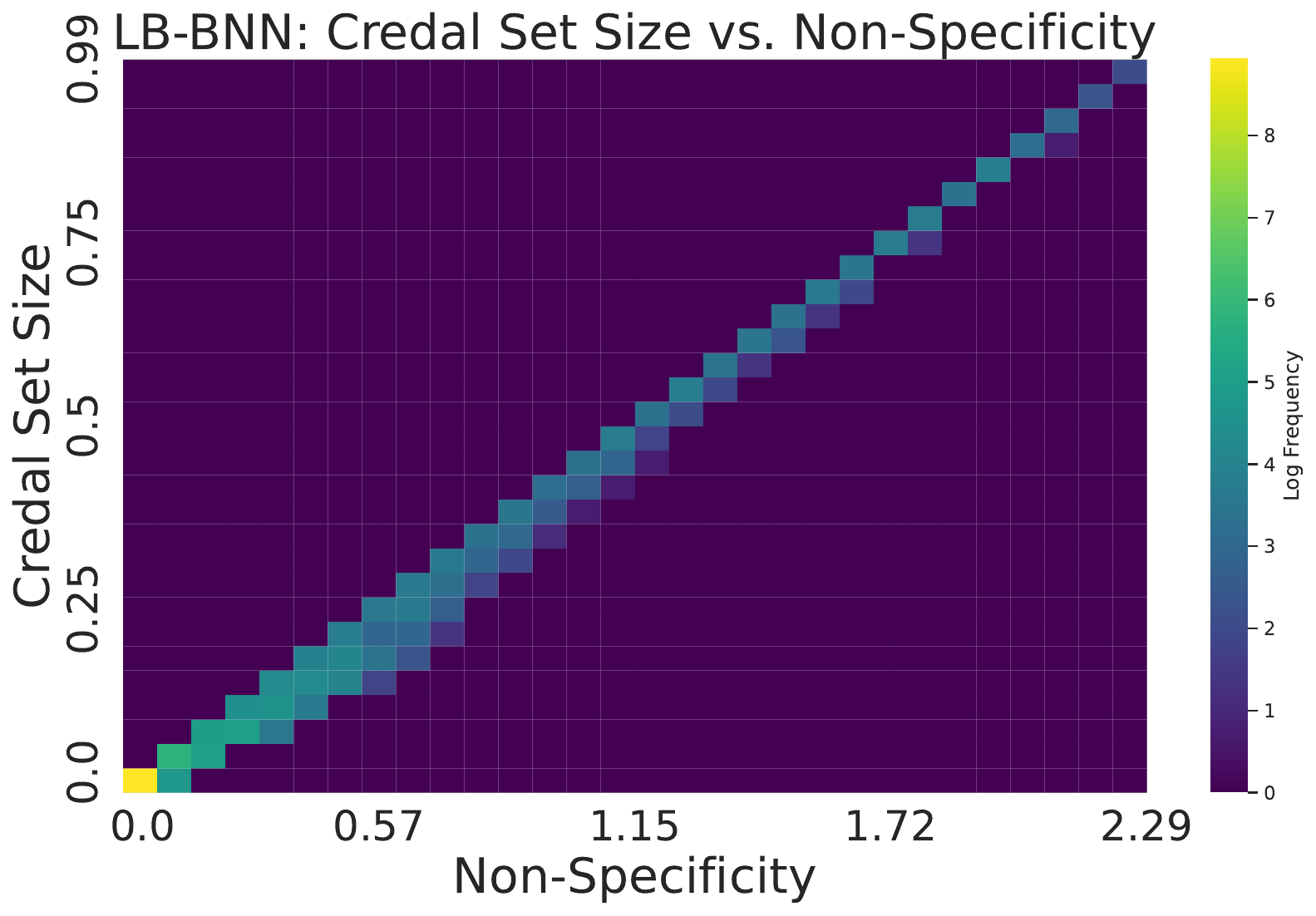}
        \vspace{-8mm}
    \caption*{(a)}
    \end{minipage} \hspace{0.01\textwidth}
    \begin{minipage}[t]{0.3\textwidth}
    \includegraphics[width=\textwidth]{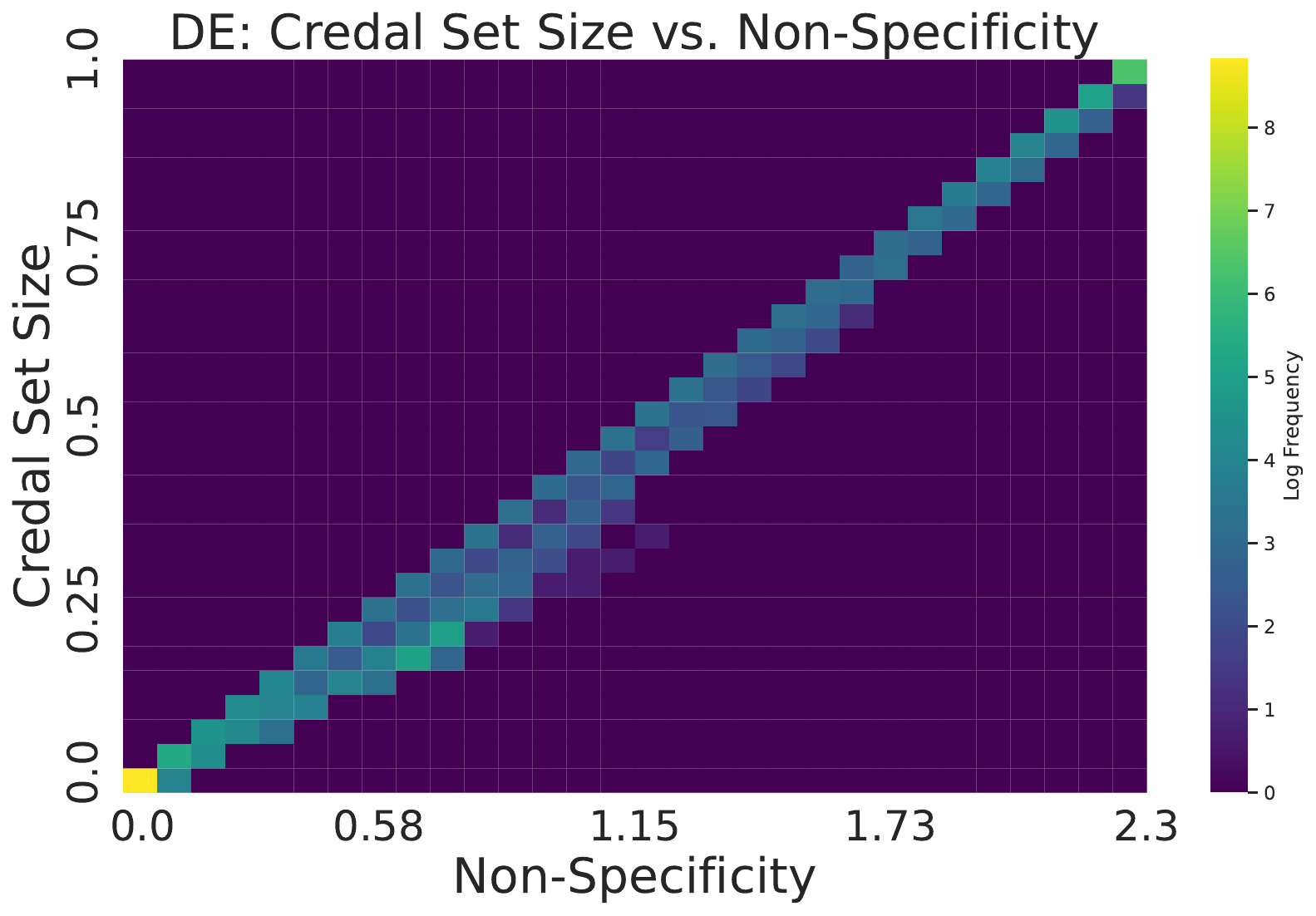}
        \vspace{-8mm}
    \caption*{(b)}
    \end{minipage} \hspace{0.01\textwidth}
        \begin{minipage}[t]{0.3\textwidth}
    \includegraphics[width=\textwidth]{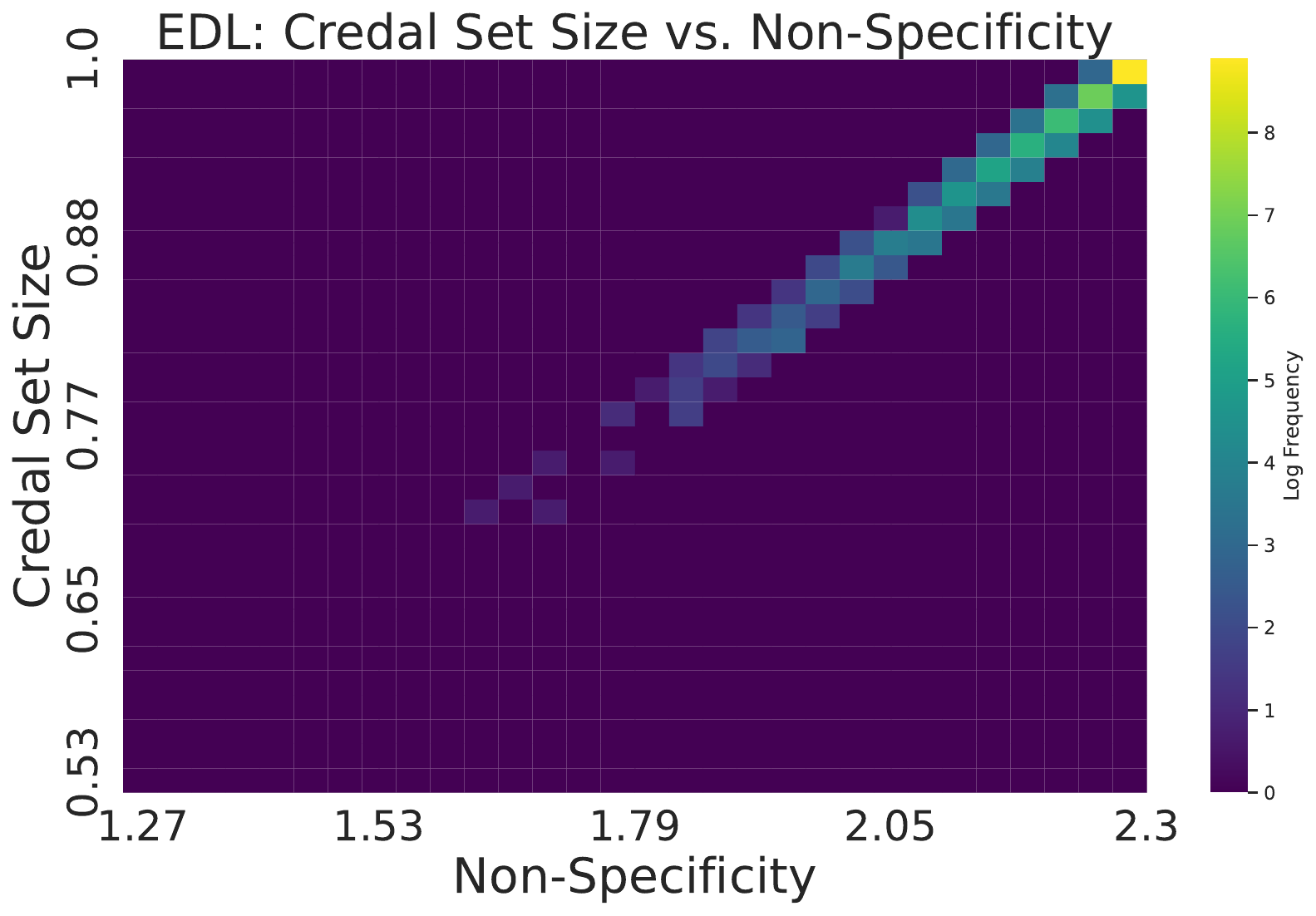}
        \vspace{-8mm}
    \caption*{(c)}
    \end{minipage}
    \\
    \begin{minipage}[t]{0.3\textwidth}
    \includegraphics[width=\textwidth]{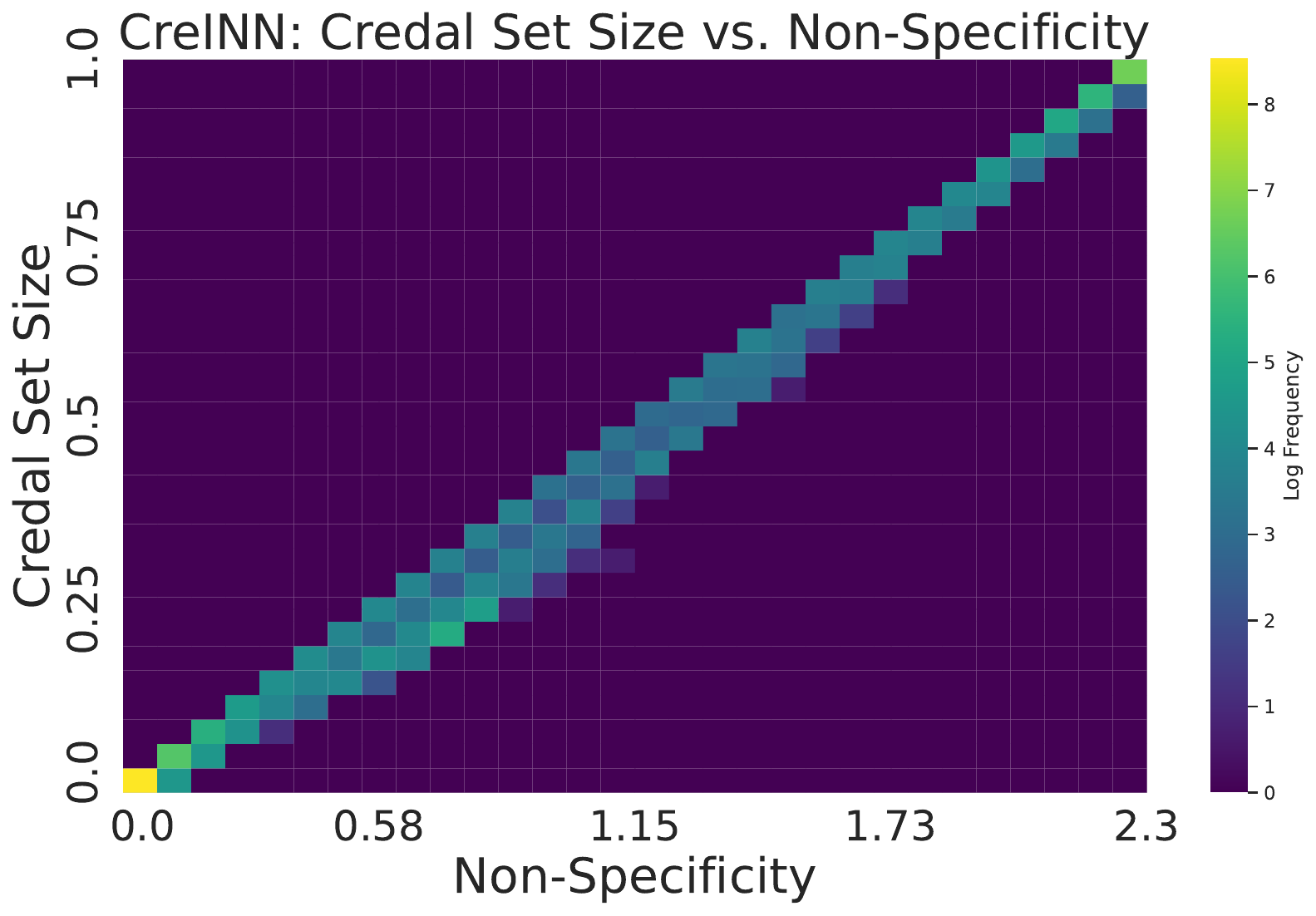}
        \vspace{-8mm}
    \caption*{(d)}
    \end{minipage} \hspace{0.01\textwidth}
        \begin{minipage}[t]{0.3\textwidth}
    \includegraphics[width=\textwidth]{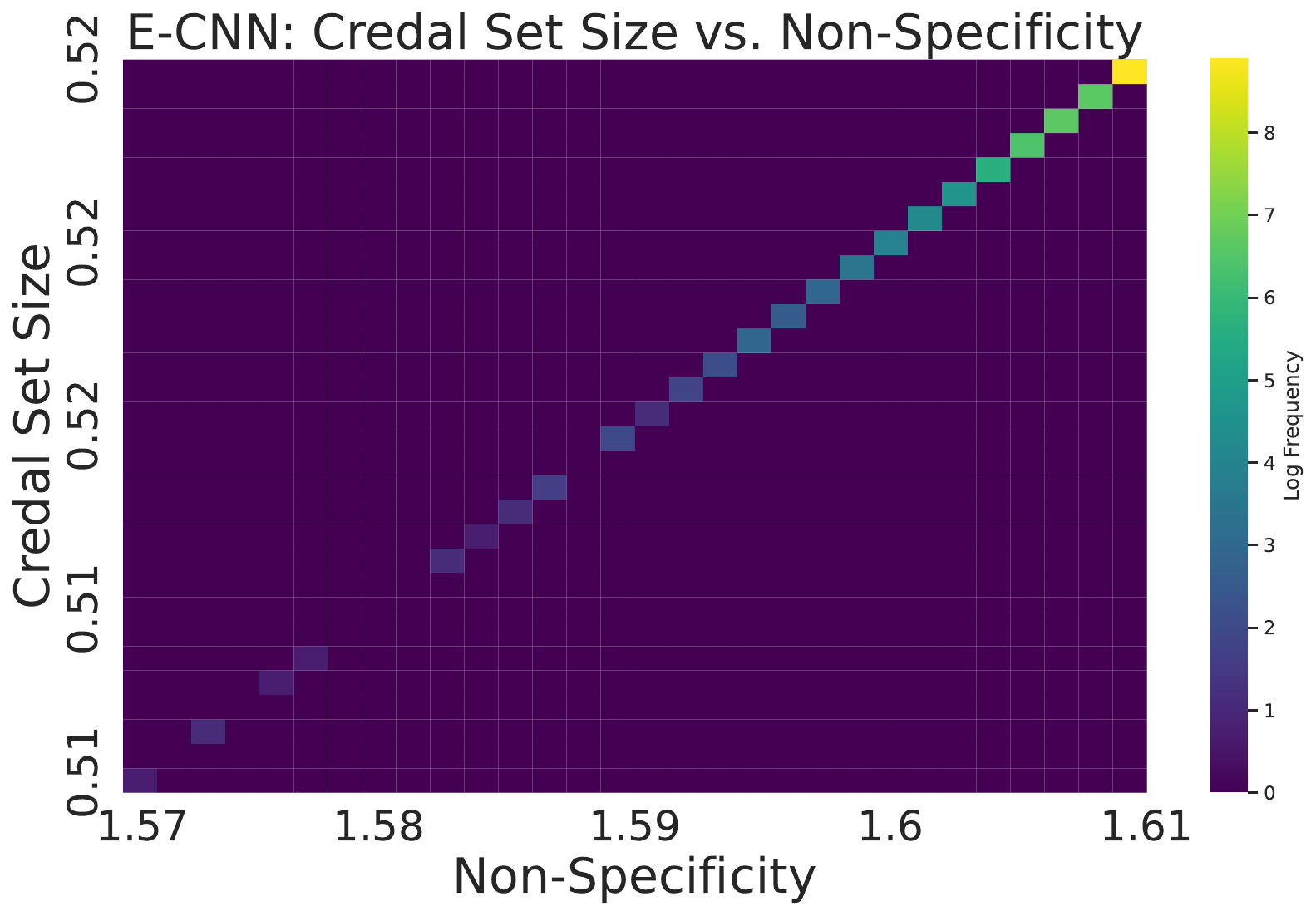}
        \vspace{-8mm}
    \caption*{(e)}
    \end{minipage}\hspace{0.01\textwidth}
            \begin{minipage}[t]{0.3\textwidth}
    \includegraphics[width=\textwidth]{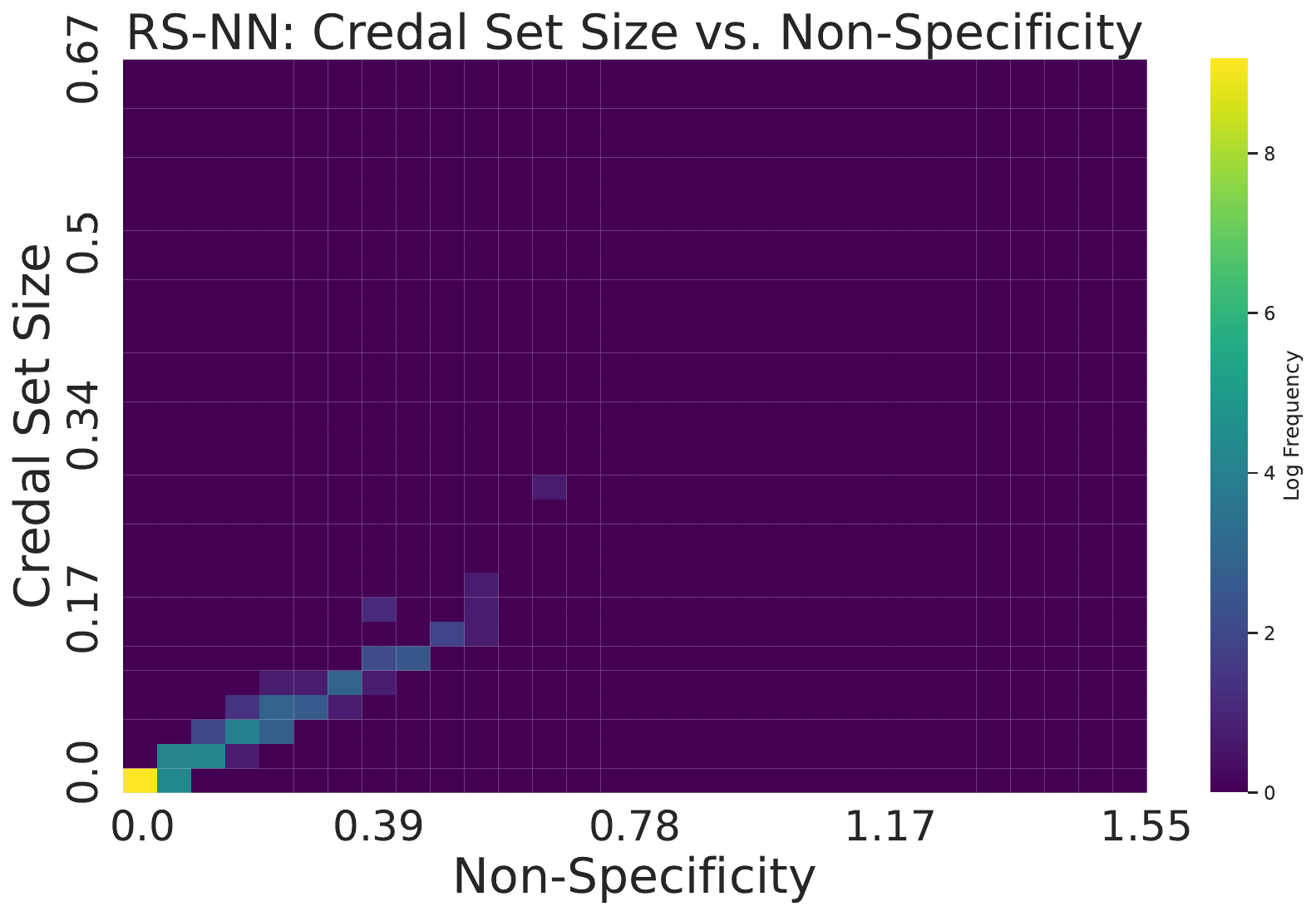}
        \vspace{-8mm}
    \caption*{(f)}
    \end{minipage}
    \vspace{-2.5mm}
    \caption{Credal Set Size vs. Non-Specificity heatmap for (a) LB-BNN, (b) Deep Ensembles (DE), (c) EDL, (d) CreINN, (e) E-CNN and (f) RS-NN on the CIFAR-10 dataset. Credal Set Size and Non-Specificity are directly correlated to each other. Log frequency is used to better showcase the trend.
}\label{fig:app_credal_vs_non_specificity_cifar10}
\end{figure}

\vspace{-4.5mm}
\begin{figure}[!h]
    \centering
    \begin{minipage}[t]{0.3\textwidth}
    \includegraphics[width=\textwidth]{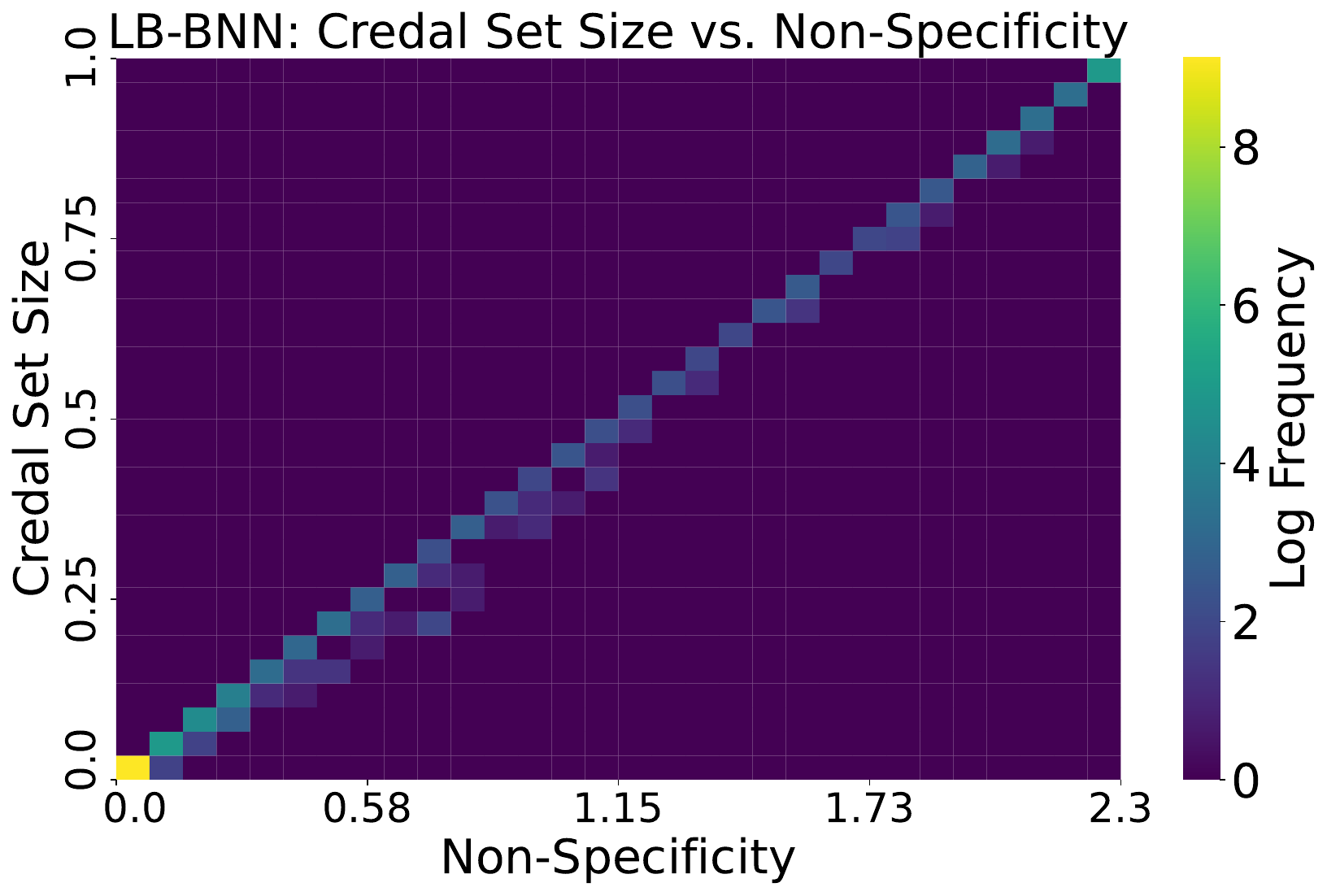}
            \vspace{-8mm}
    \caption*{(a)}
    \end{minipage} \hspace{0.01\textwidth}
    \begin{minipage}[t]{0.3\textwidth}
    \includegraphics[width=\textwidth]{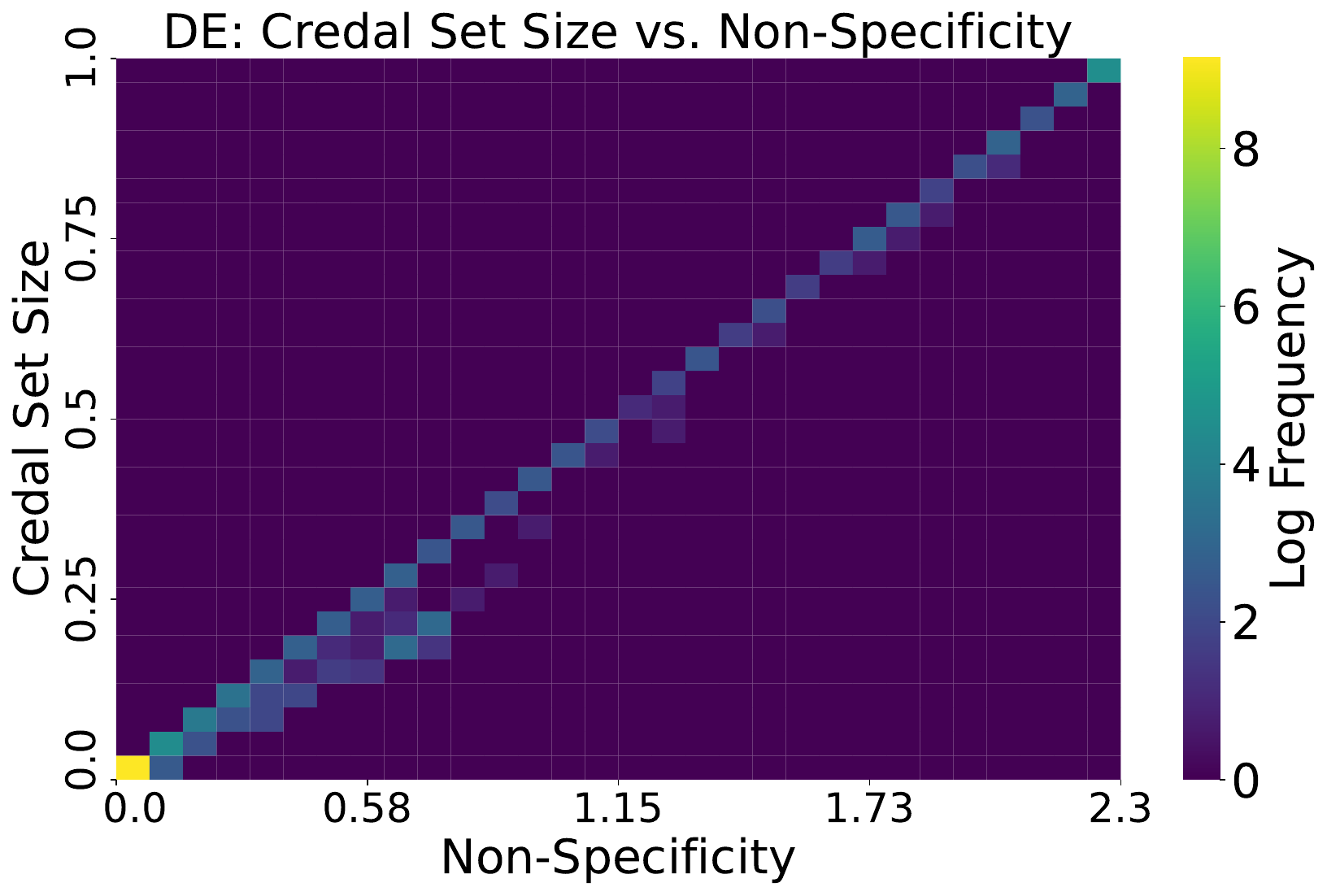}
            \vspace{-8mm}
    \caption*{(b)}
    \end{minipage} \hspace{0.01\textwidth}
        \begin{minipage}[t]{0.3\textwidth}
    \includegraphics[width=\textwidth]{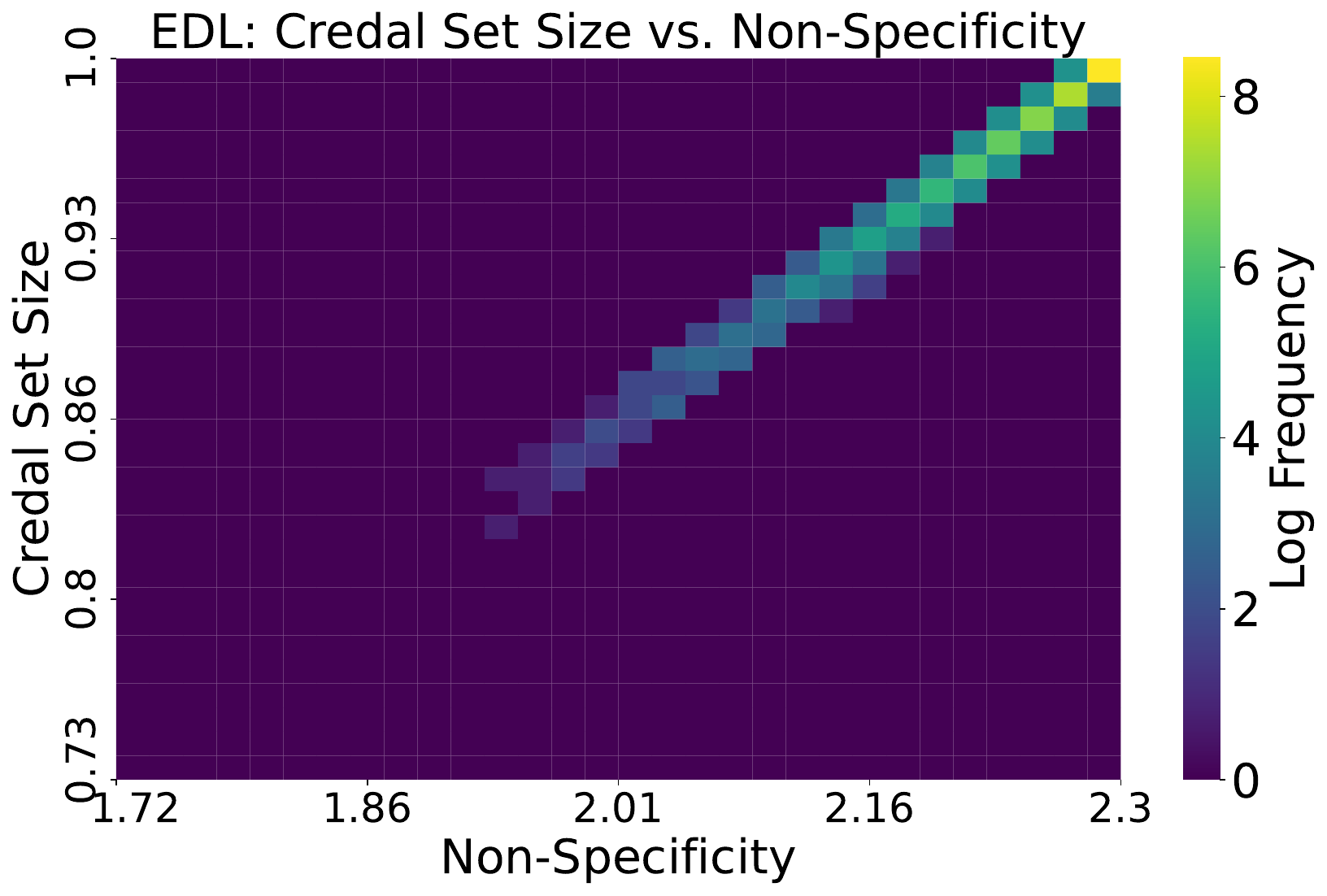}
            \vspace{-8mm}
    \caption*{(c)}
    \end{minipage}
    \\
    \begin{minipage}[t]{0.3\textwidth}
    \includegraphics[width=\textwidth]{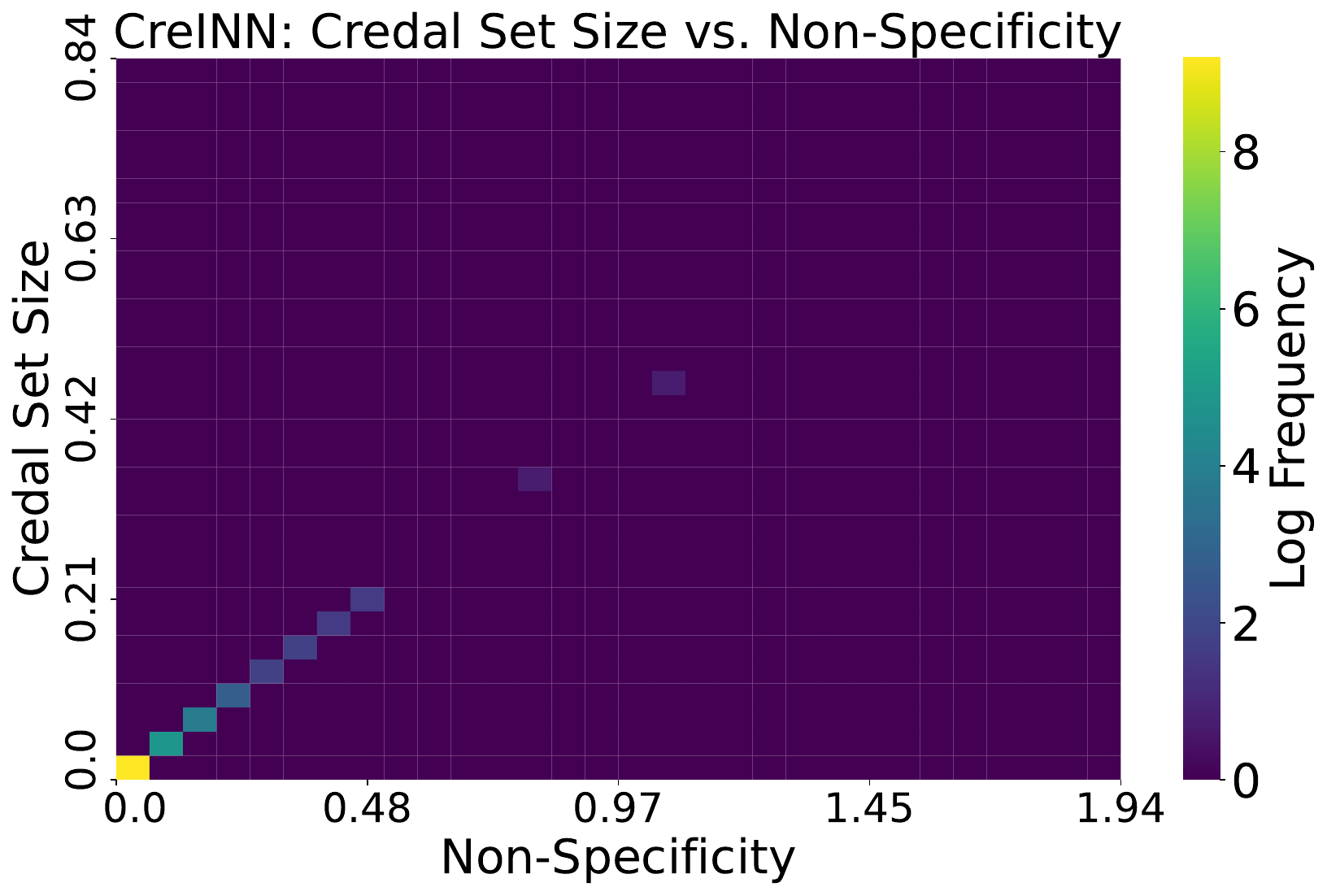}
            \vspace{-8mm}
    \caption*{(d)}
    \end{minipage} \hspace{0.01\textwidth}
        \begin{minipage}[t]{0.3\textwidth}
    \includegraphics[width=\textwidth]{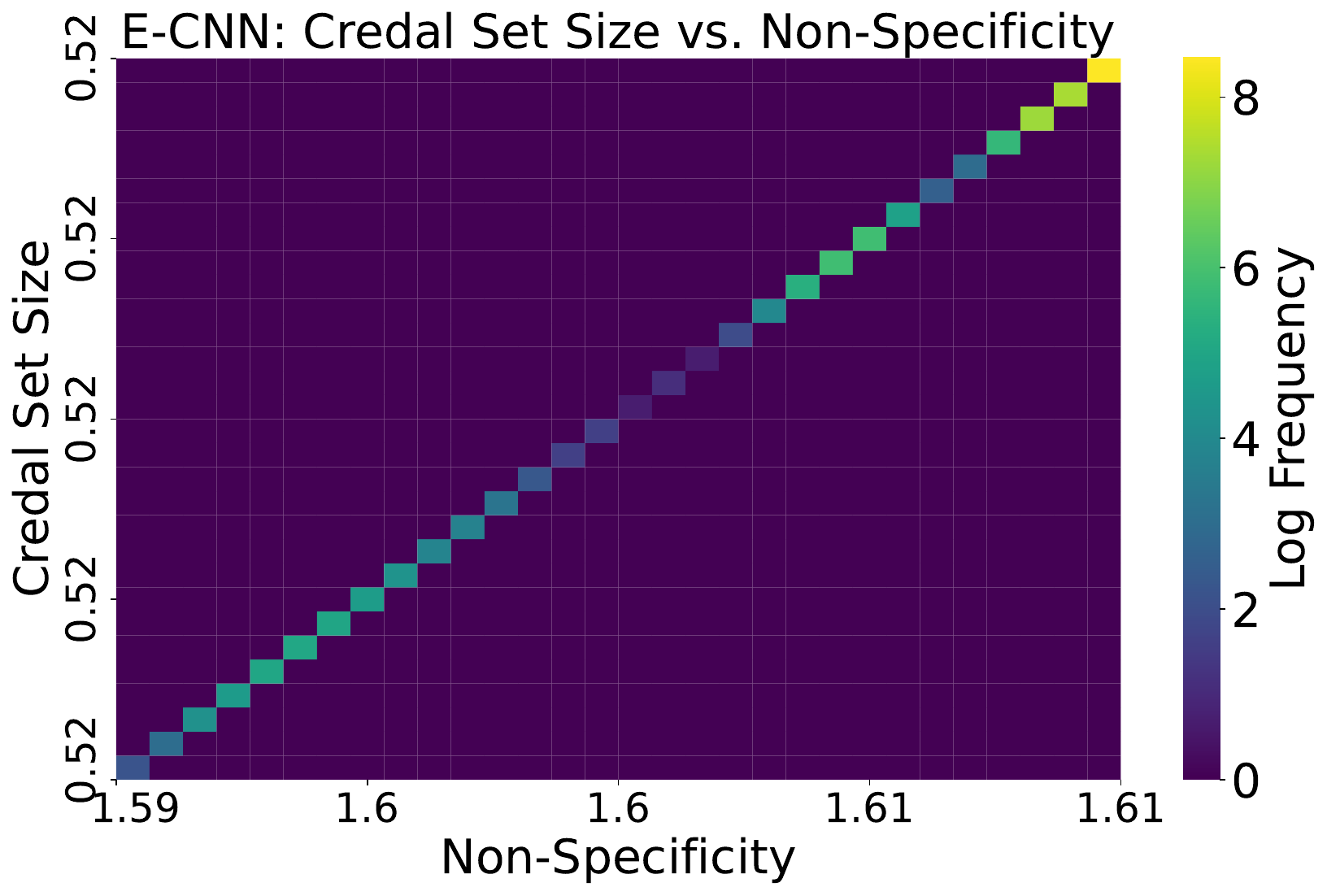}
            \vspace{-8mm}
    \caption*{(e)}
    \end{minipage}\hspace{0.01\textwidth}
            \begin{minipage}[t]{0.3\textwidth}
    \includegraphics[width=\textwidth]{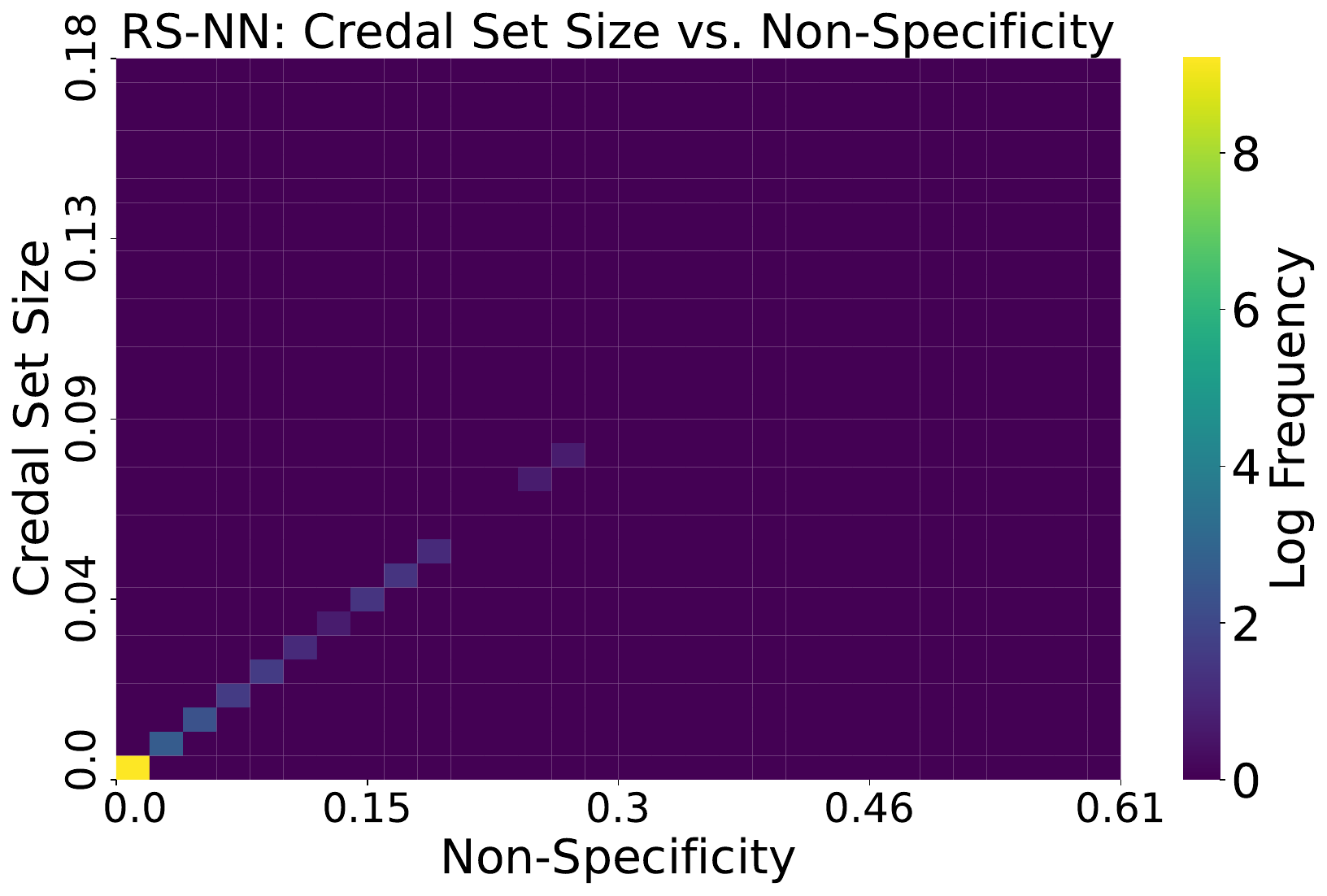}
            \vspace{-8mm}
    \caption*{(f)}
    \end{minipage}
            \vspace{-2.5mm}
    \caption{Credal Set Size vs. Non-Specificity heatmap for (a) LB-BNN, (b) Deep Ensembles (DE), (c) EDL, (d) CreINN, (e) E-CNN and (f) RS-NN on the MNIST dataset. Credal Set Size and Non-Specificity are directly correlated to each other. Log frequency is used to better showcase the trend.
}\label{fig:app_credal_vs_non_specificity_MNIST}
\end{figure}

Figs. \ref{fig:app_credal_vs_non_specificity_cifar10}, \ref{fig:app_credal_vs_non_specificity_MNIST} and \ref{fig:app_credal_vs_non_specificity_cifar100} exhibits the correlation between credal set size and non-specificity for CIFAR-10, MNIST and CIFAR-100 respectively. Here, credal set size is computed by taking the average of the difference between maximal and minimal extremal probability Eq. \ref{eq:prho} for each class, and non-specificity is obtained using Eq. \ref{eq:non_spec}. 

There exists a direct correlation between the credal set size and non-specificity. This demonstrates the efficacy of using the non-specificity measure in the evaluation metric as it captures the uncertainty associated with model's prediction making the metric more wholesome. E-CNN produces a very wide credal set over all samples visualized in MNIST dataset indicating that it's a very imprecise model. 

\begin{figure}[!h]
    \centering
    \begin{minipage}[t]{0.3\textwidth}
    \includegraphics[width=\textwidth]{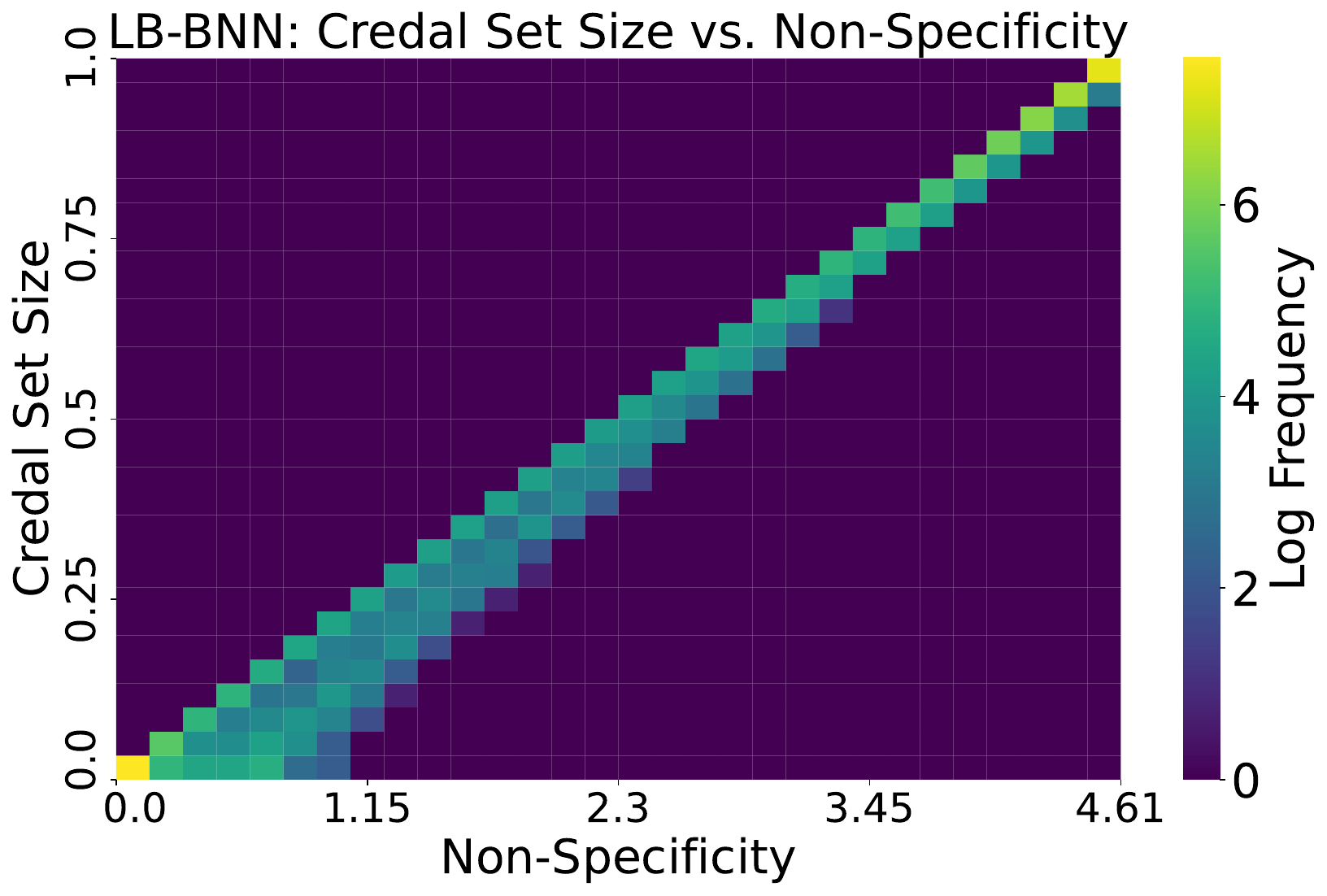}
    \caption*{(a)}
    \end{minipage} \hspace{0.01\textwidth}
    \begin{minipage}[t]{0.3\textwidth}
    \includegraphics[width=\textwidth]{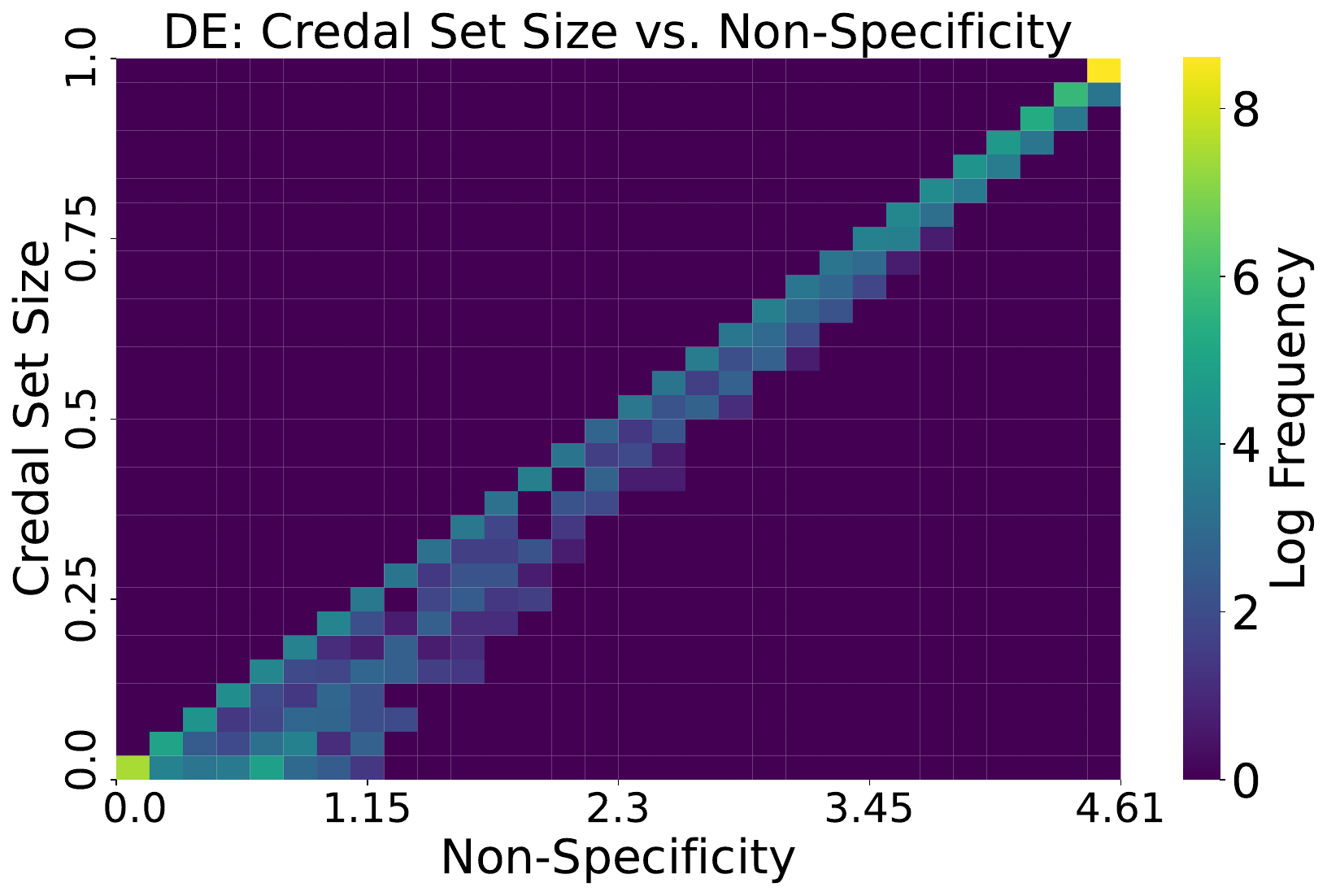}
    \caption*{(b)}
    \end{minipage} \hspace{0.01\textwidth}
        \begin{minipage}[t]{0.3\textwidth}
    \includegraphics[width=\textwidth]{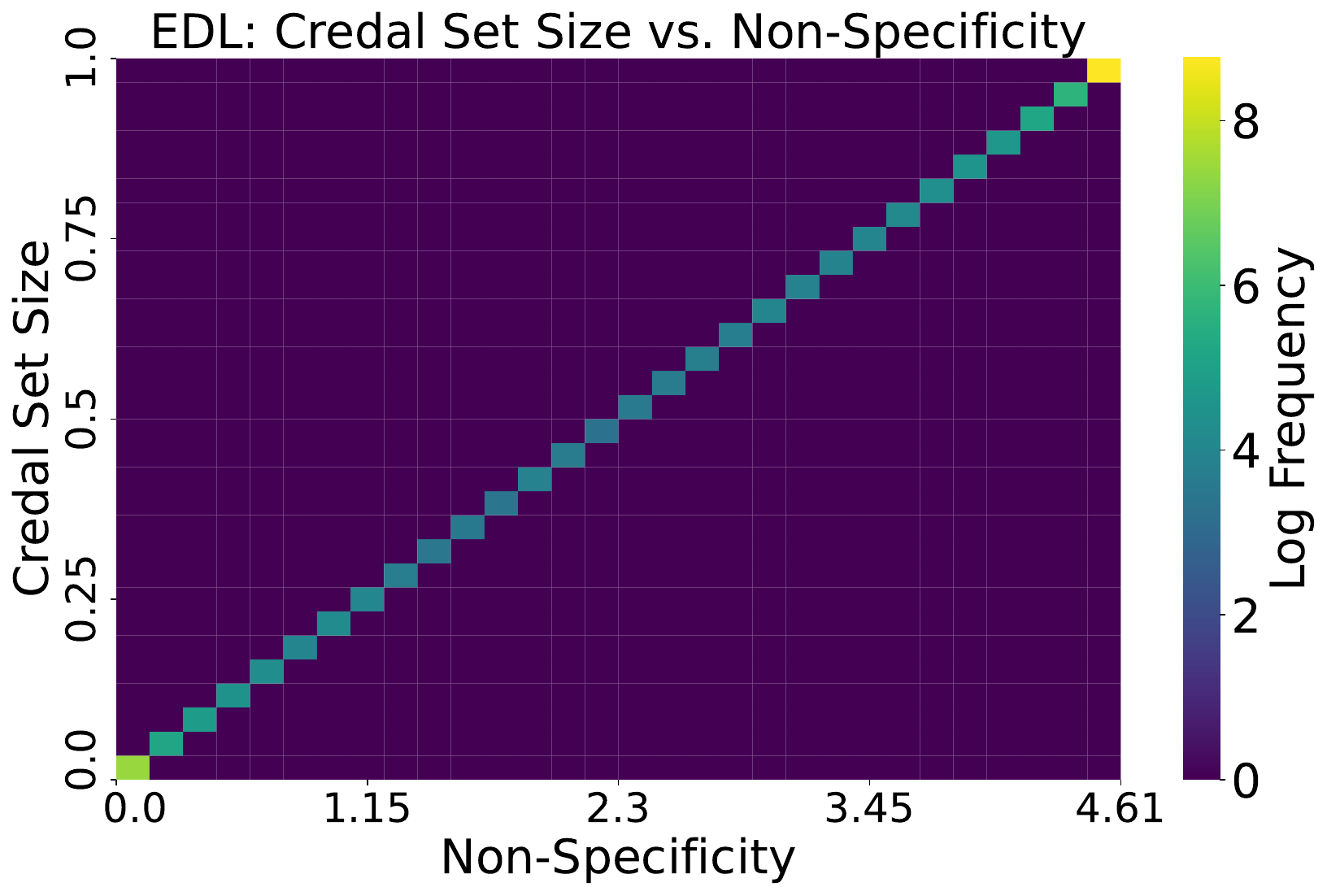}
    \caption*{(c)}
    \end{minipage}
    \\
        \begin{minipage}[t]{0.3\textwidth}
    \includegraphics[width=\textwidth]{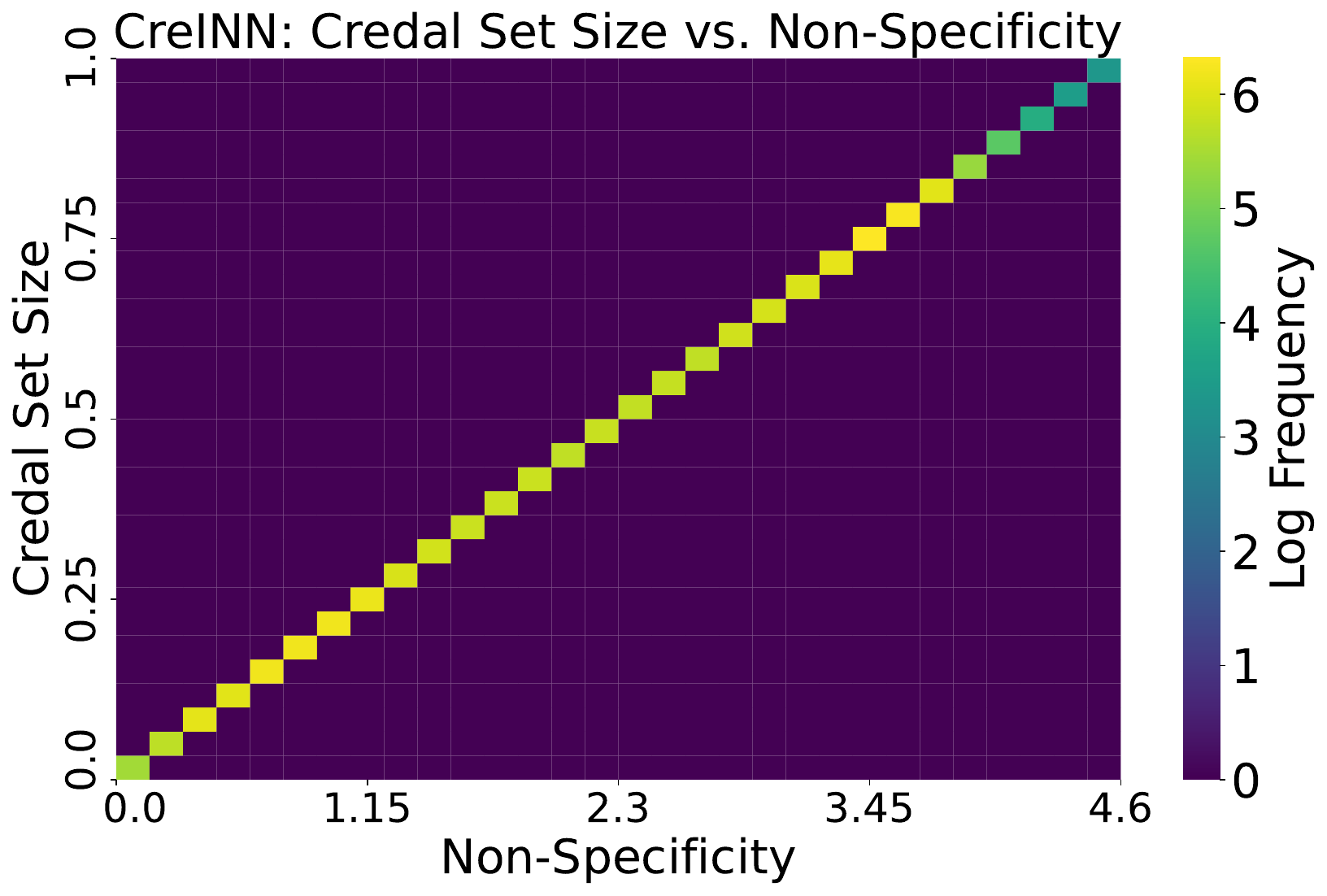}
    \caption*{(d)}
    \end{minipage}\hspace{0.01\textwidth}
            \begin{minipage}[t]{0.3\textwidth}
    \includegraphics[width=\textwidth]{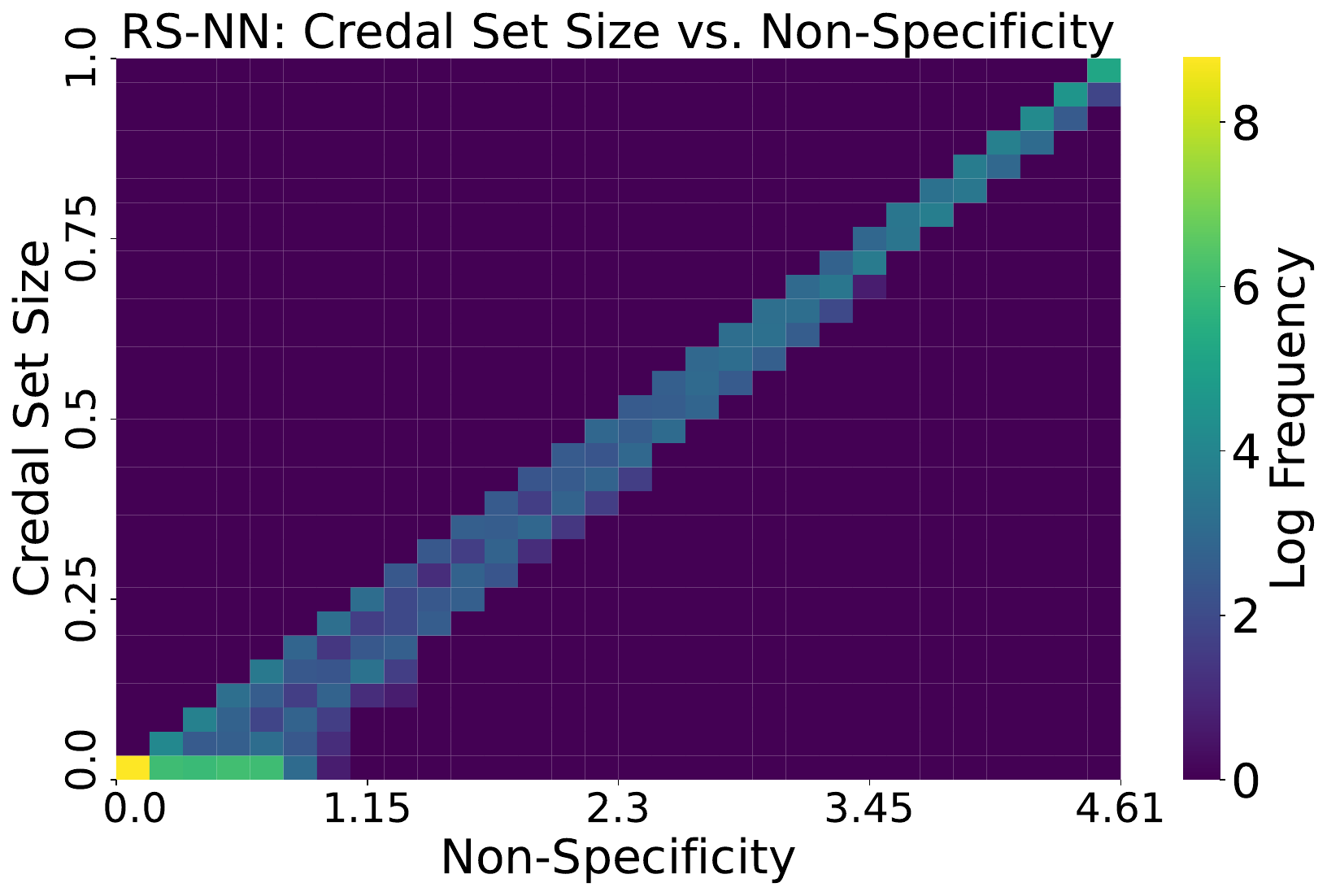}
    \caption*{(e)}
    \end{minipage}
    \caption{Credal Set Size vs. Non-Specificity heatmap for (a) LB-BNN, (b) Deep Ensembles (DE), (c) EDL, (d) CreINN and (e) RS-NN on the CIFAR-100 dataset. Credal Set Size and Non-Specificity are directly correlated to each other. Log frequency is used to better showcase the trend.
}\label{fig:app_credal_vs_non_specificity_cifar100}
\end{figure}

\subsubsection{Ablation on number of prediction samples}
\label{sec:exp_ablation_samples}

In this ablation study, we investigate the impact of varying the number of prediction samples for LB-BNN (left) and the number of ensemble members for Deep Ensembles (DE) (right) on the evaluation metric $\mathcal{E}$, as shown in Fig. \ref{fig:ablation_samples}. We observe that increasing the number of samples consistently leads to an increase in the value of $\mathcal{E}$, with the effect being more pronounced for DE compared to LB-BNN.

Specifically, for LB-BNN, we vary the number of prediction samples (from posterior) from 50 to 500, while for DE, we evaluate ensemble sizes ranging from 5 to 30 networks. As the number of samples increases, the predictive distribution becomes more expressive, leading to a more comprehensive characterization of uncertainty. This results in a broader credal set, which in turn enhances the non-specificity component of the evaluation.


\begin{figure}[!h]
    \centering
    \includegraphics[width = 0.9\linewidth]{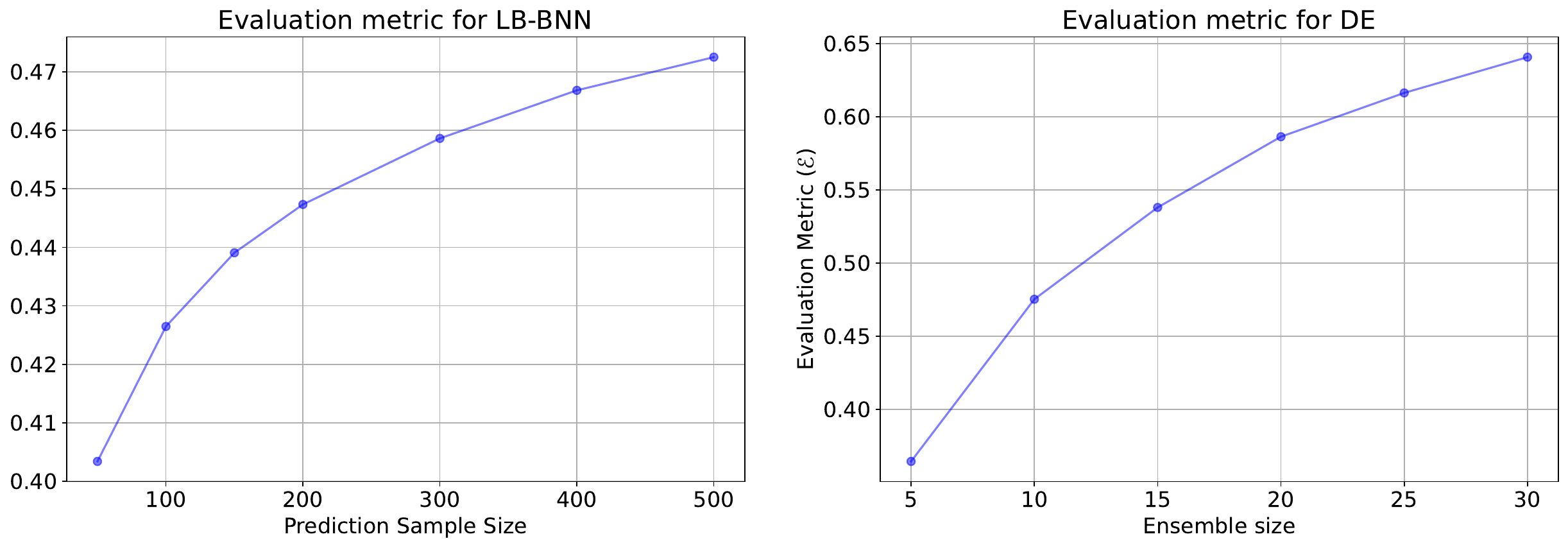} 
    \caption{Ablation study on the number of prediction samples of LB-BNN and the number of ensembles of DE with Evaluation Metric ($\mathcal{E}$). 
    } 
    \label{fig:ablation_samples}
\end{figure}

\section{Discussion and Implications}

For uncertainty-aware models, evaluation should extend beyond simple accuracy. Instead, it should utilize the full information from predictions; for instance, multiple samples from a Bayesian posterior or predictions from each model in an ensemble contain information that is often smoothed out while averaging \citep{hinne2020conceptual, graefe2015limitations}. Relying solely on correctness, using traditional metrics that assess whether the most probable class matches the ground truth, fails to capture critical information, such as the divergence of the prediction from the ground truth, the overall spread of the predictive distribution (represented by the credal set), and the robustness implied by this spread.

Moreover, constructing a compound metric by combining discrete accuracy with a continuous measure such as non-specificity may feel less principled. In contrast, both distance-based measures (\textit{e.g.} , KL divergence) and the extent of the credal set (quantified via non-specificity) operate naturally within the geometry of the probability simplex. Rather than reducing predictions to binary decisions, this evaluation framework offers a more comprehensive and informative perspective on the performance of uncertainty-aware models.

By jointly evaluating the distance to the ground truth and the extent of predictive uncertainty, this framework enables more informative comparisons between models that may share similar accuracy but differ in how they express and manage uncertainty. It facilitates a holistic evaluation of both Bayesian and evidential models, allowing for a more principled assessment of model reliability across different uncertainty paradigms.


\part{{{\MakeUppercase{Applications}}}}
\label{part:app}
\thispagestyle{plain}
\chapter{Random-Set Large Language Models}
\label{ch:rsllm}

The success of Random-Set Neural Networks (RS-NNs) in classification (Chapter \ref{ch:rsnn}) offers a strong foundation for extending the random set framework to other domains. RS-NNs demonstrated that modeling predictions as distributions over sets of outcomes, rather than over single classes, yields superior performance in accuracy, calibration, adversarial robustness, and out-of-distribution (OoD) detection. Building on this theoretical and empirical foundation, RS-NNs can be naturally extended from vision-based classification to token prediction in Large Language Models (LLMs) as Random-Set Large Language Models (RS-LLMs).

Large Language Models (LLMs) \citep{achiam2023gpt, anil2023palm, touvron2023llama} are typically trained to predict the next token from a fixed vocabulary, making next-token prediction essentially a classification problem. However, conventional LLMs output point estimates in the form of probability distributions and fail to quantify uncertainty. This becomes especially problematic when LLMs generate hallucinations, \textit{i.e.}, factually incorrect or ungrounded statements \cite{maynez2020faithfulness}. RS-LLMs resolve this by adapting RS-NN’s belief-based prediction framework to LLMs, equipping them with the ability to express second-order uncertainty over token predictions. 
{While LLMs are generally designed to produce plausible text rather than truth-grounded answers, our application focuses on supervised tasks (e.g., CoQA, OpenBookQA), where a ground truth is available. In this context, the LLM acts as a classifier over candidate answers or text spans, making the application of RS-NN possible.}

\section{Methodology}

Just as RS-NNs predict belief functions over class labels, RS-LLMs predict belief functions over sets of vocabulary tokens. These belief functions, grounded in random set theory and belief function theory \cite{shafer1976mathematical, cuzzolin2010geometry, cuzzolin2018visions}, assign mass to subsets of the vocabulary instead of individual tokens. This extension is made computationally feasible by a budgeting technique: hierarchical clustering over token embeddings \citep{mullner2011modern, ackermann2014analysis}, that selects a finite, informative set of token subsets (focal sets) over which the belief function is defined. The resulting belief function can be mapped to a credal set and its center, the pignistic distribution \cite{smets1994transferable}, is used for actual token prediction. Fig. \ref{fig:rsllm-flow} shows the training and generation flow of RS-LLM.

\begin{figure*}[t!]
\centering
\includegraphics[width = 0.9\textwidth]{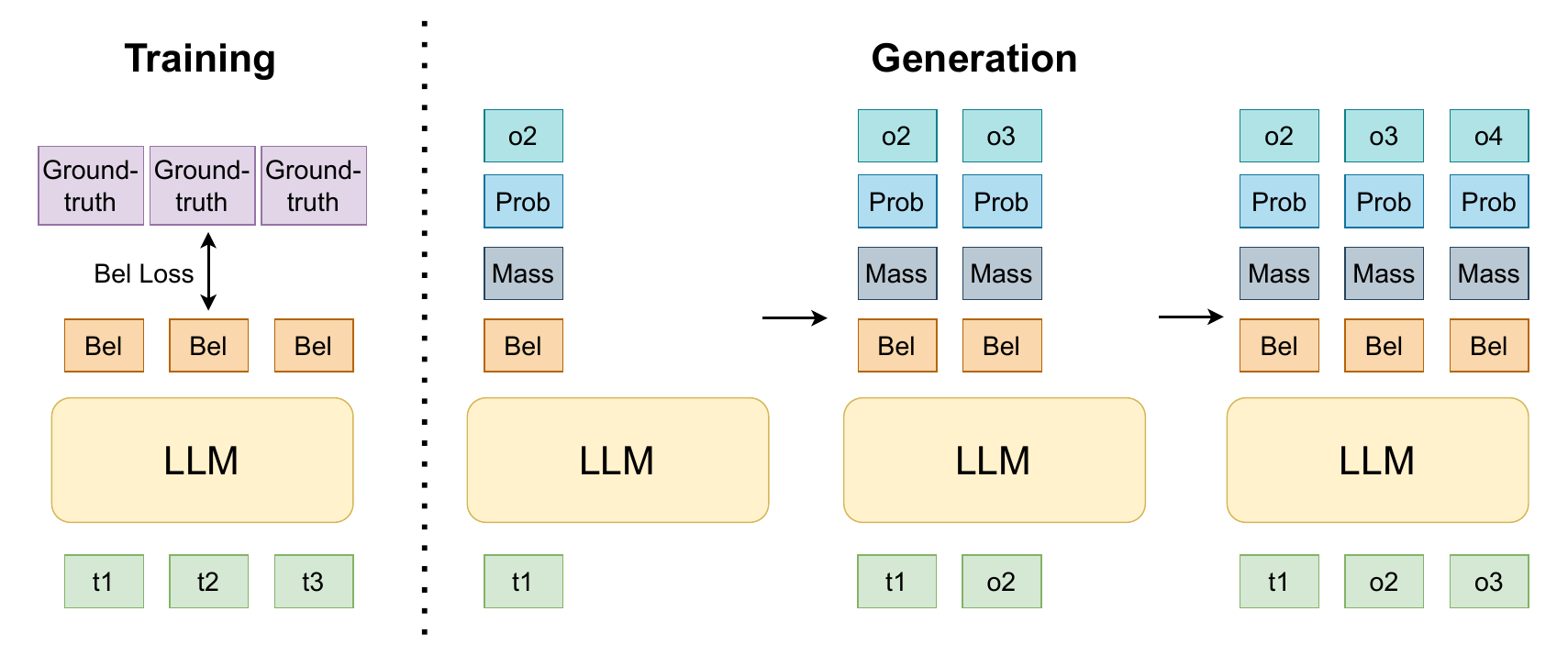}
\caption{Training and generation flow of RS-LLM. Training is performed in a parallel fashion using the teacher forcing method. Generation is done sequentially. For each token, the model predicts a belief function. Then the mass function, probability distribution and next token is subsequently computed/sampled from that belief function.
} \label{fig:rsllm-flow}
\end{figure*}

This belief-based approach offers significant interpretive advantages. For instance, when an LLM predicts that `\textit{Joe likes to play \_\_\_}' and outputs a 50-50 split between `baseball' and `basketball', it is unclear whether this reflects indecision or dual correctness. By contrast, an RS-LLM can distinguish these possibilities through different belief assignments: mass on the joint set \{baseball, basketball\} indicates epistemic uncertainty (the model is unsure), while equal mass on individual tokens suggests both are plausible answers. Thus, RS-LLMs inherit RS-NN’s ability to disambiguate types of uncertainty, bringing that clarity into the LLM domain.


Random-sets are defined on the power set of all subsets of their domain. 
However, this is computationally impossible 
when the domain cardinality is extremely large, as is the case 
for large language models where the typical size of vocabulary is $\sim$32K tokens. 
The number of subsets in this case would be an astronomical $2^{32K}$. 
Therefore, to tackle this problem, we need a method to select a budget of most appropriate focal sets and use only those as our focal sets. 
This is appropriate because, by intuition, the model will most likely be confused among tokens which are similar to each other 
and potentially have similar semantics \cite{farquhar2024detecting}.

As a first step, token embeddings are computed using a pre-trained LLM (in fact, any method for obtaining embedding of tokens can be used here). Then, hierarchical clustering is applied to cluster similar tokens. The number of clusters $K$ is a hyper-parameter which determines the granularity of the belief function.

The final budget $\mathcal{O}$ of focal sets of the belief functions to be predicted is the union of the clusters of tokens so obtained and the collection of singleton sets containing the $T$ original tokens, amounting to a total of $(K + T)$ sets.  Fig. \ref{fig:rsllm-budgeting} shows a detailed overview of the proposed budgeting method. An analysis of the focal sets obtained by this budgeting strategy is presented in Sec. \ref{sec:budgeting-analysis}.

\begin{figure*}[ht]
\centering
\includegraphics[width = 0.9\textwidth]{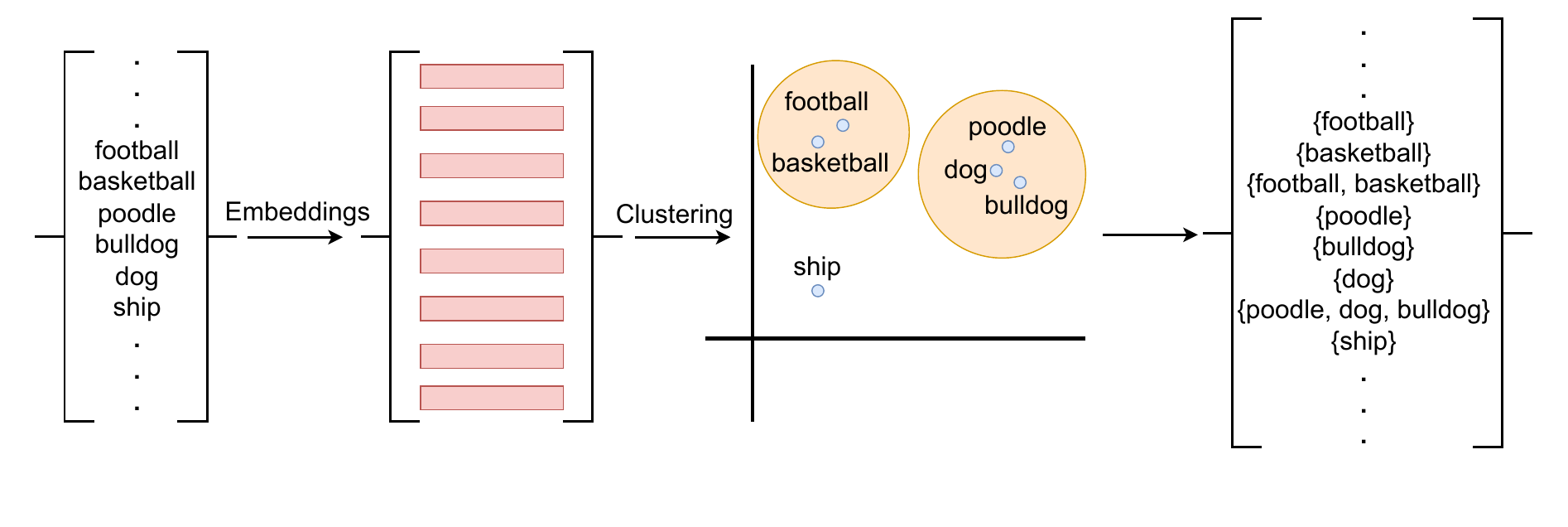}
\vspace{-8mm}
\caption{Proposed budgeting method for RS-LLM. First, embeddings are computed for all the tokens in vocabulary. Then, focal sets are computed using hierarchical clustering.} \label{fig:rsllm-budgeting}
\end{figure*}

\textbf{Training} follows the teacher forcing paradigm \citep{williams1989learning}, where the model learns to predict belief functions token by token.  During generation, belief functions are sequentially transformed into mass functions, from which pignistic probabilities are computed to generate the next token. This mirrors the RS-NN structure but adapts it to the autoregressive, sequential setting of LLMs. RS-LLM follows the loss function (Sec. \ref{sec:loss}) and uncertainty estimation techniques (Sec. \ref{sec:uncertainty}) of RS-NN.

\section{Experiments}

\subsection{Implementation}

\textbf{Datasets.} RS-LLM is evaluated on two QA benchmarks: CoQA \citep{reddy-etal-2019-coqa}, a generative dataset with free-text answers based on stories, and OBQA-Additional \citep{OpenBookQA2018}, a multiple-choice dataset in STEM with added factual context. CoQA contains 7,200 training and 500 validation samples (used as test data) across five domains. OBQA has 4,957 training and 500 test samples.

\textbf{Models and Setup.} We use Llama2-7B-hf \citep{touvron2023llama}, Mistral-7B \citep{jiang2023mistral7b}, and Phi-2 \citep{javaheripi2023phi}. As with RS-NN, any LLM can be converted into a random-set LLM (RS-LLM) by redefining its output layer to produce sets of tokens. We extract $K=8{,}000$ focal sets via budgeting, yielding final output sizes of 40K for Llama2/Mistral and 59.2K for Phi-2. For RS-Llama2, we set $\alpha = \beta = 10^{-2}$.

\textbf{Training} uses Huggingface's trl framework \citep{havrilla2023trlx}, 4-bit quantization, and rank-64 LoRA adapters \citep{yu2023low} across all layers, on A100 80GB GPUs for 5 epochs with batch size 8. Models are trained via supervised fine-tuning using the template in Fig.~\ref{fig:training_samples}, where instructions (blue) and questions (black) are provided, and the model predicts the answer (green). At inference, answers are generated greedily.

\begin{figure*}[ht!]
    \centering
    \includegraphics[width = 0.8\linewidth]{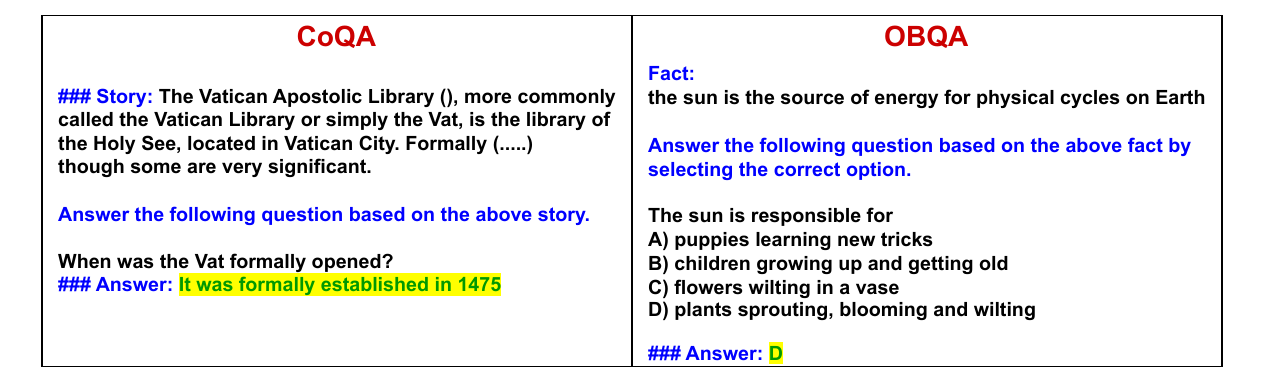}
    \vspace{-1mm}
    \caption{Training examples from CoQA and OBQA datasets. The text in black highlights the actual question, while the blue text represents prompt instructions. The model is trained to predict the text in green.}
    \label{fig:training_samples}
\end{figure*}

\subsection{Performance}

Tab. \ref{tab:correctness_performance} shows that RS-LLMs consistently outperform standard LLMs in both cosine similarity (CoQA) and accuracy (OBQA), despite identical training conditions. This supports the expressiveness of set-valued outputs.

\begin{table}[h!]
    \caption{Performance of Standard LLMs and RS-LLMs on CoQA and OBQA datasets. For CoQA, cosine similarity is reported while accuracy is reported for OBQA.}
    \vspace{-8pt}
    \centering
    \begin{tabular}{lccc}
        \toprule
        Model & Metric & CoQA & OBQA \\ 
        \midrule
        \multirow{2}{*}{Llama2} 
        & \emph{Cosine Similarity} & 0.69 & - \\
        & \emph{Accuracy} & - & 83.20 \\ 
        \multirow{2}{*}{RS-Llama2} 
        & \emph{Cosine Similarity} & \textbf{0.71} & - \\
        & \emph{Accuracy} & - & \textbf{89.60} \\
        \midrule
        \multirow{2}{*}{Mistral} 
        & \emph{Cosine Similarity} & 0.67 & - \\
        & \emph{Accuracy} & - & 91.60 \\ 
        \multirow{2}{*}{RS-Mistral} 
        & \emph{Cosine Similarity} & \textbf{0.72} & - \\
        & \emph{Accuracy} & - & \textbf{93.00} \\ 
        \midrule
        \multirow{2}{*}{Phi2} 
        & \emph{Cosine Similarity} & 0.72 & - \\
        & \emph{Accuracy} & - & 87.60 \\ 
        \multirow{2}{*}{RS-Phi2} 
        & \emph{Cosine Similarity} & \textbf{0.73} & - \\
        & \emph{Accuracy} & - & \textbf{91.80} \\ 
        \bottomrule
    \end{tabular}
    \label{tab:correctness_performance}
\end{table}

\subsection{Uncertainty Estimation} 

Fig. \ref{fig:plots} analyzes uncertainty behavior. {In CoQA, there is no obvious trend between the pignistic entropy and cosine similarity of RS-Llama2 (Fig. \ref{fig:plots}(b)), standard Llama2 (Fig. \ref{fig:plots}(a)).}
On OBQA, entropy distributions for correct/incorrect predictions overlap in both models (Figs.\ref{fig:plots}(d), \ref{fig:plots}(e)), limiting their usefulness.
More promisingly, credal width correlates negatively with correctness (Fig. \ref{fig:plots}(c)) and shows some separability across correctness in OBQA (Fig. \ref{fig:plots}(f)). 

\begin{figure}[t!]
    \includegraphics[width = \linewidth]{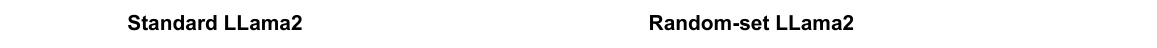}
    \centering
        \begin{subfigure}[t]{0.32\textwidth}
        \centering
        \includegraphics[width = \linewidth]{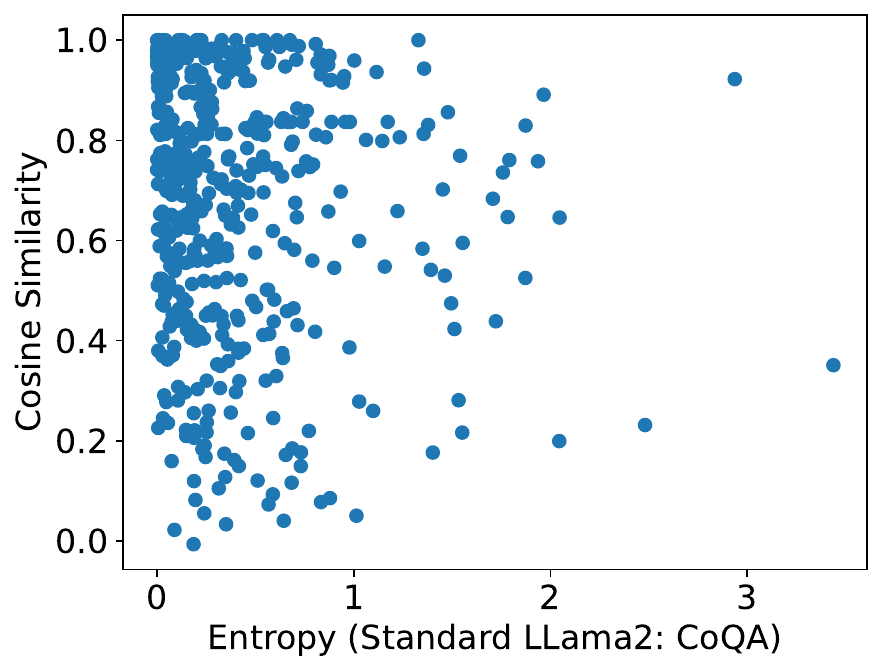}
        \caption*{(a)}
        \label{fig:Entropy_CosSim_Standard_LLama2_CoQA}
    \end{subfigure}%
    \begin{subfigure}[t]{0.32\textwidth}
        \centering
        \includegraphics[width = \linewidth]{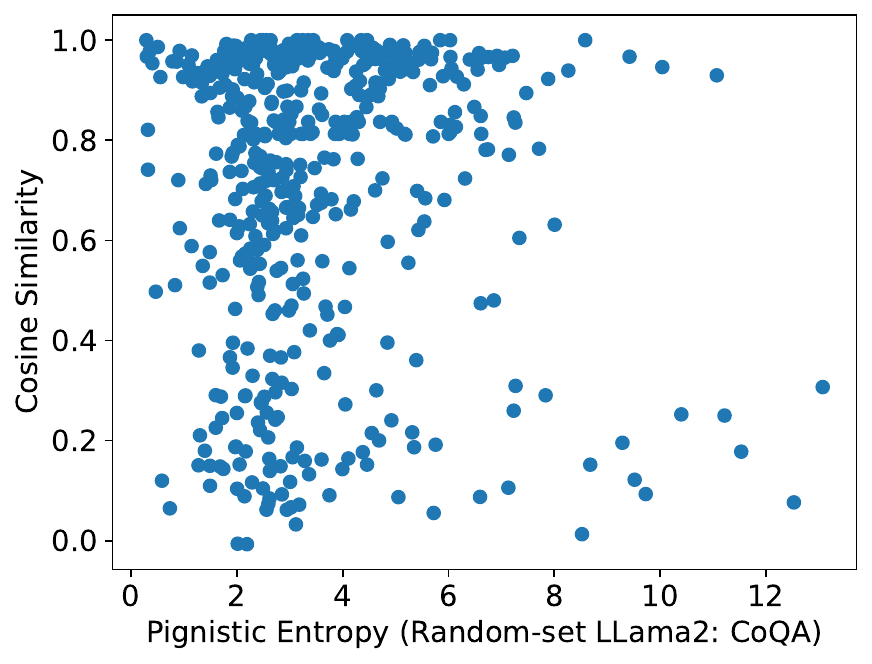}
        \caption*{(b)}
        \label{fig:Pig_Entropy_CosSim_RS_LLama2_CoQA}
    \end{subfigure}%
        \begin{subfigure}[t]{0.32\textwidth}
        \centering
        \includegraphics[width = \linewidth]{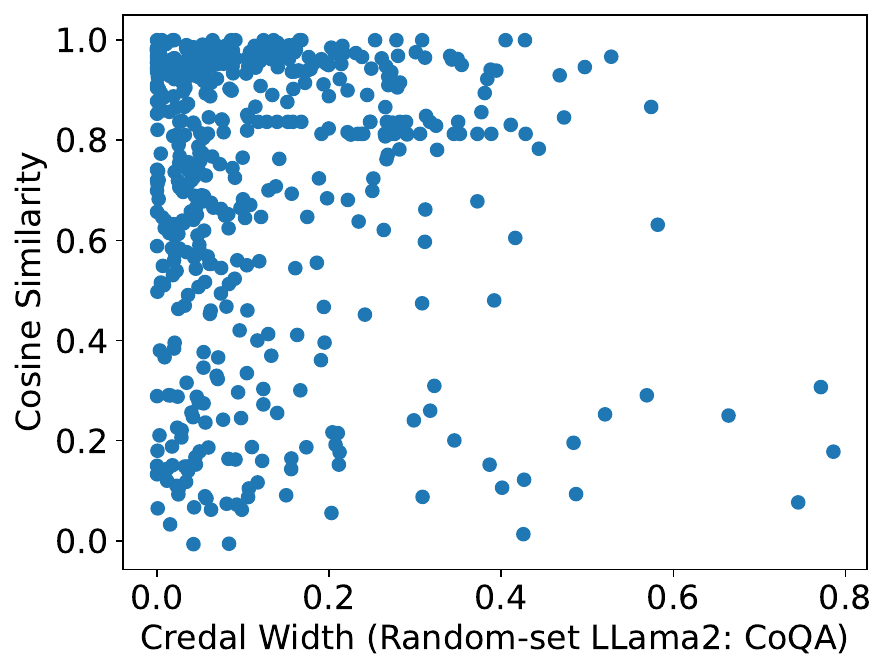}
        \caption*{(c)}
    \label{fig:Credal_width_CosSim_RS_LLama2_CoQA}
    \end{subfigure}%
    \\
    \begin{subfigure}[t]{0.32\textwidth}
        \centering
        \includegraphics[width = \linewidth]{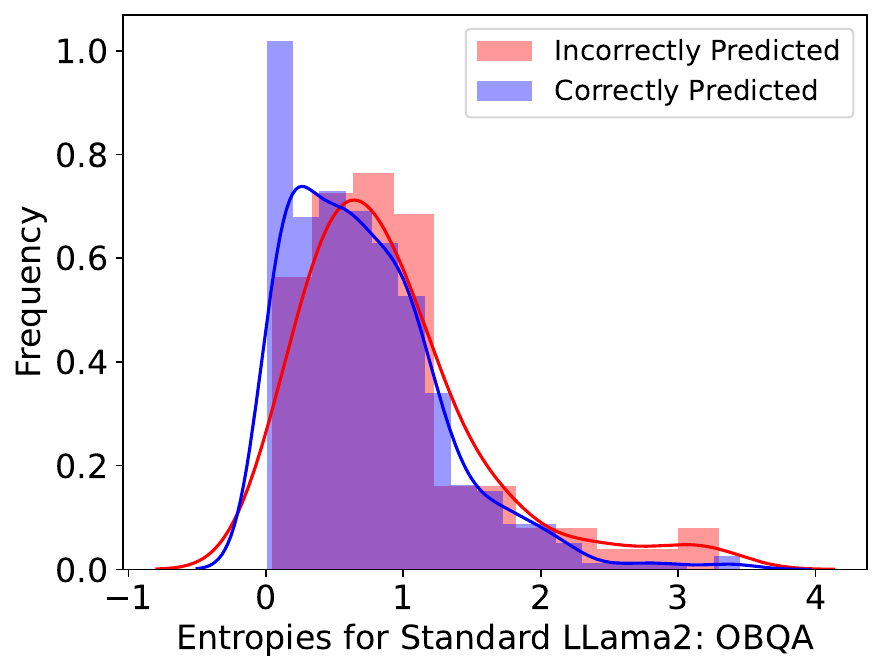}
        \caption*{(d)}
        \label{fig:Entropy_Standard_LLama2_OBQA}
    \end{subfigure}%
        \begin{subfigure}[t]{0.32\textwidth}
        \centering
        \includegraphics[width = \linewidth]{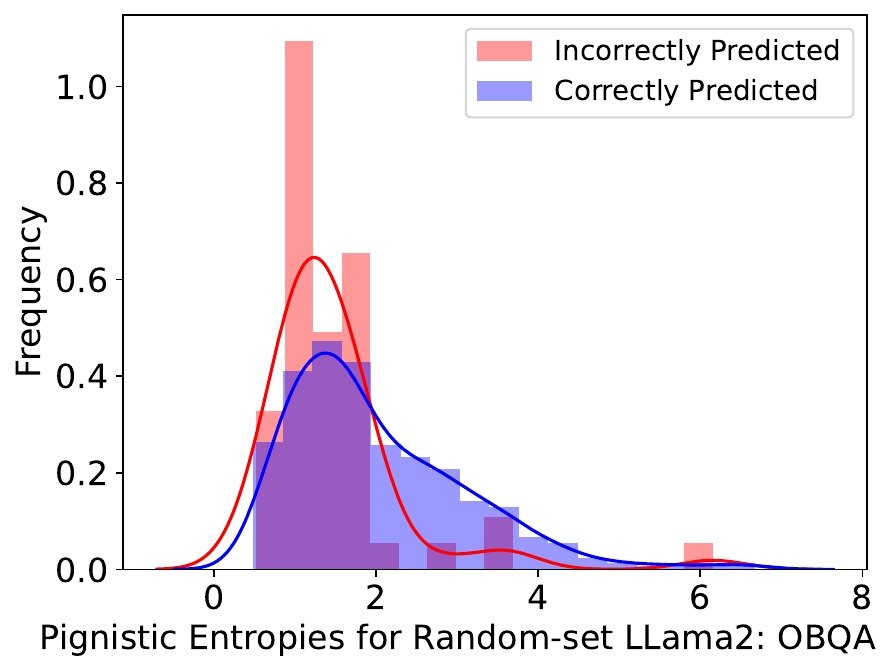}
        \caption*{(e)}
        \label{fig:Pig_Entropy_RS_LLama2_OBQA}
    \end{subfigure}%
    \begin{subfigure}[t]{0.32\textwidth}
        \centering
        \includegraphics[width = \linewidth]{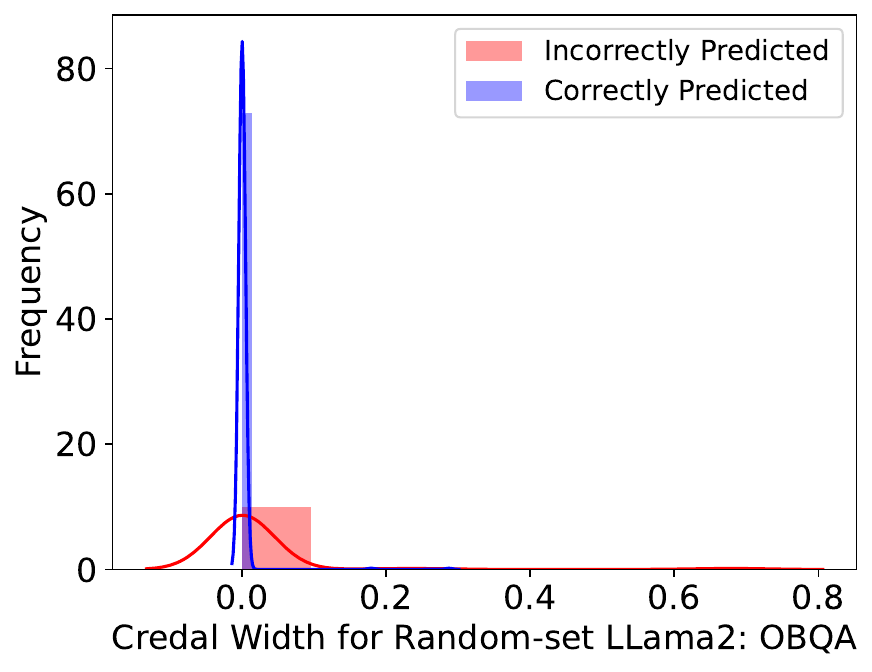}
        \caption*{(f)}
        \label{fig:Credal_width_RS_LLama2_OBQA}
    \end{subfigure}%
    \caption{Behavior of uncertainty measures of Llama2 and RS-Llama2 with respect to the correctness and closeness to the groundtruth on CoQA and OBQA datasets.}
    \label{fig:plots}
\end{figure}

\subsection{Hallucination Evaluation}

To test hallucination, we feed the models misleading contexts: for CoQA, the question is swapped with another; for OBQA, answer choices are randomized. Tab. \ref{tab:hallucination} reports uncertainty metrics under correct and incorrect contexts. Both models show increased uncertainty when context is wrong. RS-Llama2, in particular, separates the two better via credal width, reflecting second-level uncertainty modeling.

\begin{table*}[h]
    \caption{Uncertainty evaluation of Standard Llama2 and RS-Llama2 on correct and incorrect context.}
        \vspace{-8pt}
    \centering
    \begin{tabular}{cccccc}
        \toprule
         &  \multirow{2}{*}{Metric} & \multicolumn{2}{c}{Correct Context ($\downarrow$)} & \multicolumn{2}{c}{Inorrect Context($\uparrow$)} \\ 
        \cline{3-6} 
         & & CoQA & OBQA & CoQA & OBQA \\ 
        \midrule
        \multirow{1}{*}{Llama2} & \emph{Entropy} & $0.39 \pm 0.45$ &  $0.75 \pm 0.59$ & $0.90 \pm 0.80$ &$1.10 \pm 0.93$ \\ 
        RS-Llama2 & \emph{Credal Width} & $0.13 \pm 0.13$ & $0.00 \pm 0.04$ & $0.28 \pm 0.21$ & $0.02 \pm 0.08$ \\  
        \bottomrule
    \end{tabular}
    
    \label{tab:hallucination}
\end{table*}

\subsection{Budgeting Analysis} \label{sec:budgeting-analysis}

We analyze the budgeted focal sets obtained using Llama2-7b-hf (token size: 32,000) with $K = 8000$ sampled focal sets from $2^{32000}$. The sets display semantic closeness, \textit{e.g.}, `[\verb+cattle+, \verb+sheep+]', `[\verb+shame+, \verb+pity+]', `[\verb+quiet+, \verb+calm+, \verb+quietly+]', `[\verb+maintain+, \verb+retain+, \verb+maintained+, \verb+retained+]', and `[\verb+delight+, \verb+pleasure+, \verb+pleased+, \verb+proud+, \verb+pride+]'.

To evaluate this, we compute centroid distance (mean Euclidean distance from the cluster center) using sentence-transformer encodings \cite{reimers-2019-sentence-bert}. Fig. \ref{fig:budgeting_analysis}(a) shows that {the mode of the distance distribution is near zero}, indicating strong semantic coherence. Fig. \ref{fig:budgeting_analysis}(b) shows focal set sizes, where 2, 3, and 4 are most frequent (counts: 2142, 1940, and 1562, respectively). The largest set (size 162) consists of punctuation/symbols.

\begin{figure*}[h]
\centering
\begin{subfigure}[t]{0.5\textwidth}
\includegraphics[width=\linewidth]{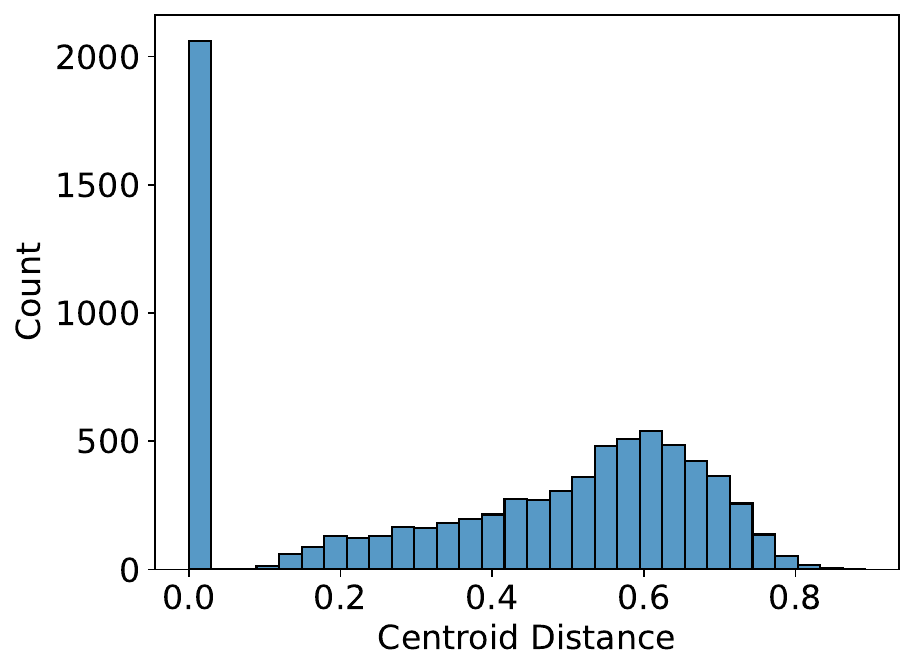}
\caption*{(a)}
\label{fig:centroid_distances}
\end{subfigure}%
\begin{subfigure}[t]{0.5\textwidth}
\includegraphics[width=\linewidth]{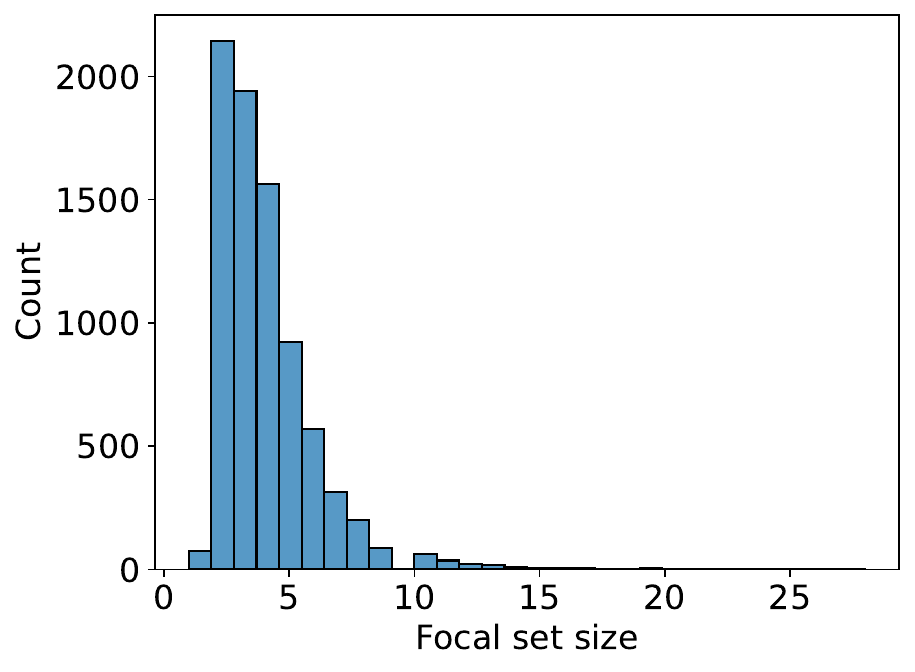}
\caption*{(b)}
\label{fig:focal_set_size}
\end{subfigure}
\caption{(a) Centroid distance distribution. (b) Focal set size distribution. Sets with size $>30$ excluded for visualization.}
\label{fig:budgeting_analysis}
\end{figure*}

\subsection{Ablation Studies}
\textbf{Hyperparameters $\alpha$ and $\beta$.}
The regularization terms $M_s$ and $M_r$ promote valid belief functions by enforcing mass normalization and non-negativity. However, excessive weighting may reduce predictive accuracy, similar to KL divergence in VAEs.

Tab. \ref{tab:ablation_hyperparameters} shows cosine similarities for different $\alpha=\beta$ values. Both $0.01$ and $0.0001$ yield best performance; we use $0.01$ to better enforce validity.

\begin{table}[h]
\centering
\caption{Cosine similarity across $\alpha = \beta$ values.}
    \vspace{-8pt}
\label{tab:ablation_hyperparameters}
\begin{tabular}{ccccc}
\toprule
$\alpha = \beta$ & 0.1 & 0.01 & 0.001 & 0.0001 \\
\midrule
Cosine Similarity & 0.66 & \textbf{0.71} & 0.69 & \textbf{0.71} \\
\bottomrule
\end{tabular}
\end{table}

\textbf{Focal Set Budget Size.}
We ablate budget size $K$ on CoQA (Tab. \ref{tab:budget_cosine_similarity}). A small $K$ behaves like classical LLMs; a large $K$ adds complexity. The best results occur at $K=8000$. The "Combined" set is the union of all budgets.

\begin{table}[h]
\centering
\caption{Cosine similarity across budget sizes on CoQA.}
    \vspace{-8pt}
\label{tab:budget_cosine_similarity}
\begin{tabular}{cccccc}
\toprule
Budget Size & 2000 & 4000 & 8000 & 16000 & Combined (56538) \\
\midrule
Cosine Similarity & 0.67 & 0.68 & \textbf{0.71} & 0.66 & 0.69 \\
\bottomrule
\end{tabular}
\end{table}

\section{Discussion and Implications}

Overall, these results substantiate the RS-NN approach’s applicability and effectiveness within large language models, demonstrating that it can successfully incorporate belief function theory to enhance uncertainty quantification and semantic representation. By efficiently budgeting focal sets, the RS-NN framework manages combinatorial complexity inherent in modeling mass functions over large token vocabularies, enabling interpretable and meaningful groupings of semantically related tokens. This balance between computational tractability and expressive power validates the RS-NN paradigm as a promising direction for neural architectures that require nuanced uncertainty modeling.

Moreover, the empirical improvements observed in downstream tasks such as question answering and summarization highlight the practical benefits of integrating RS-NN principles into state-of-the-art language models. \textit{These encouraging outcomes lay a robust foundation for further research focused on optimizing budget allocation strategies, refining focal set construction algorithms, and exploring alternative representations of belief functions within neural networks.} The integration of RS-NN with neural language models opens avenues for developing more reliable, transparent, and uncertainty-aware AI systems, thereby advancing both theoretical understanding and practical capabilities in natural language processing.

\chapter{Application to Autonomous Racing} 
\label{ch:weather} 

In this chapter, the effectiveness of Random-Set Neural Networks (RS-NN) (Chapter \ref{ch:rsnn}) on real-world \textbf{weather and cone classification} datasets is demonstrated.  RS-NN and Bayesian models are evaluated in terms of \textit{accuracy}, \textit{domain adaptation}, and \textit{uncertainty estimation}. This application requires no changes to the original RS-NN model, unlike RS-LLMs where the budgeting approach needed to be adapted to the token space.

Accurate weather classification is essential for autonomous vehicles (AVs) to handle environmental uncertainty and ensure safe operation under diverse conditions. Weather affects sensor reliability and perception accuracy, leading to increased uncertainty in object detection, scene understanding, and decision-making—key challenges for AVs, especially in edge cases. By explicitly recognizing weather conditions, AV systems can adapt their perception and planning algorithms to mitigate risks posed by adverse scenarios such as rain, fog, or low visibility. This adaptation supports epistemic AI frameworks by quantifying uncertainty related to environmental variability, enabling better generalization to previously unseen conditions. Ultimately, integrating weather classification enhances the robustness and reliability of autonomous driving systems, reducing safety-critical failures and facilitating broader deployment in real-world, dynamic environments.

\section{Experimental Setup for Weather/cone Classification}

The experiments evaluated the performance of three models: the proposed Random-Set Neural Network (RS-NN) (Chapter \ref{ch:rsnn}), a baseline standard CNN, and a Bayesian neural network (LB-BNN) \cite{hobbhahn2022fast}  on three datasets related to autonomous vehicle (AV) perception under different weather conditions:

\begin{itemize}
    \item \textbf{ROAD-WAYMO (R-WAYMO)} \cite{khan2024road}: A large, diverse \textbf{weather} dataset combining ROAD (complex road activity detection) \cite{singh2022road} and Waymo Open \cite{Sun_2020_CVPR} datasets. It contains diverse perception data from diverse geographies and conditions across the U.S. (such as San Francisco, Mountain View, Los Angeles, Detroit, Seattle, and Phoenix).

    \item \textbf{Oxford Brookes Racing-Autonomous (OBR-A)}: A smaller dataset collected from AV test tracks lined with cones, similar to Formula Student Competition settings, under various \textbf{weather} conditions including exclusive classes like Droplet and Night.

    \item \textbf{Cone-centric (CONE)}: A dataset centered on classifying \textbf{cones} by their different colors, designed to test perception models in recognizing cone color variations under various conditions.
\end{itemize}

Fig. \ref{fig:weather-dataset} presents a comprehensive summary of the various datasets employed for weather classification tasks, highlighting their origins, class distributions, and the respective sample counts allocated for training and evaluation.

\begin{figure}[!ht]
    \centering
    \includegraphics[width=0.9\linewidth]{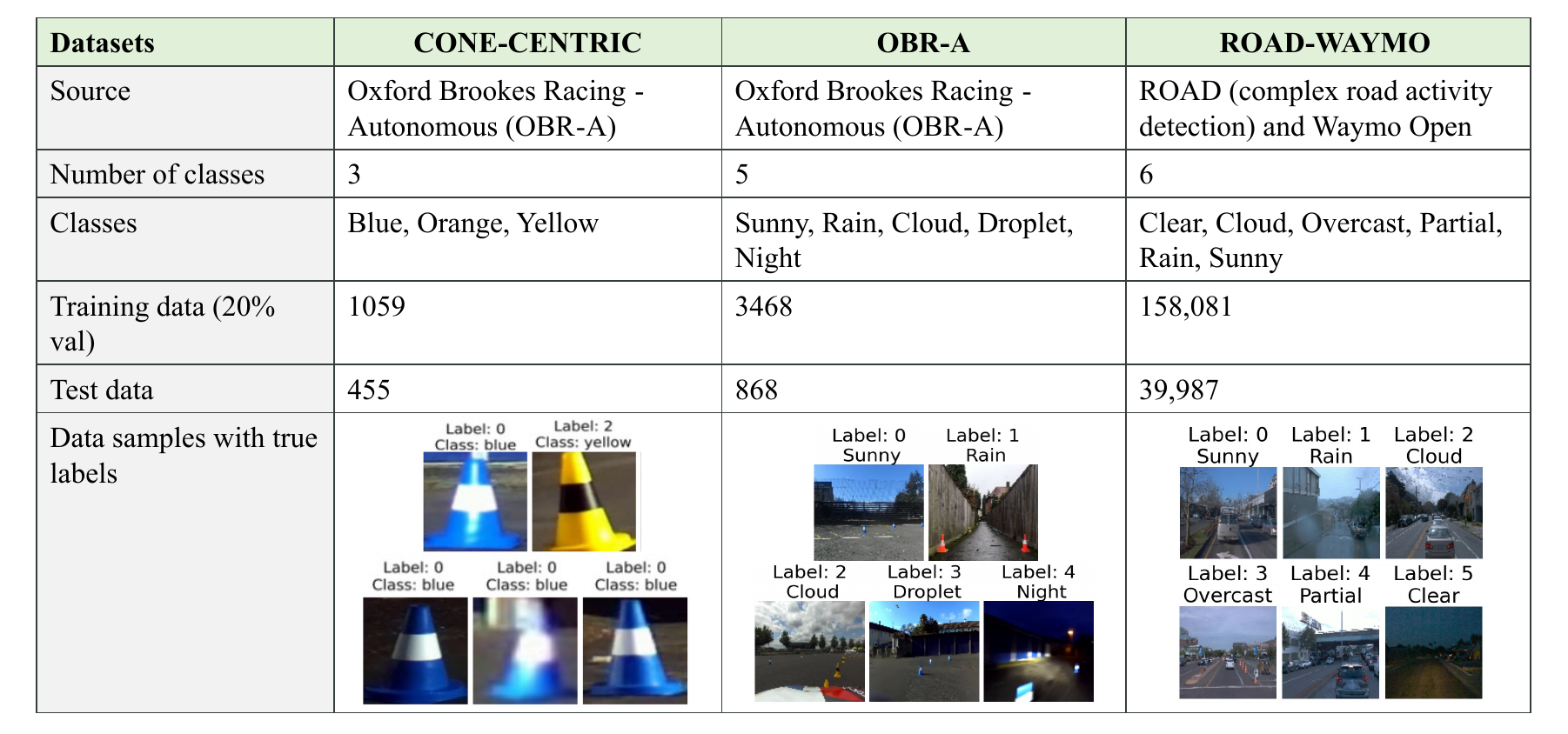}
    \vspace{-14pt}
    \caption{Overview of datasets used for weather classification, including source, number of classes, class labels, and dataset sizes for training and testing.}
    \label{fig:weather-dataset}
\end{figure}

\section{Cone classification}

Tab. \ref{pred-table} shows predictions for two representative samples from the Cone-centric dataset. For the first sample (top half of the table), the predicted belief and mass functions assign nearly all the probability mass to the class \textit{blue}, with negligible mass on other classes. This is reflected in the pignistic probability distribution, where \textit{blue} has a probability of 1.0, and the entropy is effectively zero ($1.52 \times 10^{-18}$), indicating an almost deterministic prediction with very high confidence. This corresponds to a sample that is clearly recognizable and easy to classify for the model.
In contrast, the second sample (bottom half of the table) presents a more \textit{uncertain prediction scenario}. The mass function spreads over multiple subsets, and the pignistic probabilities are more balanced among \textit{yellow} (0.506) and \textit{blue} (0.391). The entropy rises to 1.3658, reflecting this uncertainty and the ambiguity in the model’s belief.

\begin{table}[!htbp]
\setlength{\tabcolsep}{3pt}
  \caption{The predicted belief, mass values, pignistic probabilities, and entropy for two cone-centric dataset predictions.}
  \vspace{-8pt}
  \label{pred-table}
  \tiny
  \centering
  \begin{tabular}{lllllll}
    \toprule
     \textbf{Sample} & \textbf{Belief} & \textbf{Mass} & \textbf{Pignistic probability} & \textbf{Entropy}\\         
    \midrule \\
         \multirow{5}{1.9cm}[2.9ex]{\raisebox{-0.1\totalheight}{\includegraphics[width=0.12\textwidth]{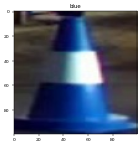}}} 
     & \begin{tabular}[t]{ll}
      \{'\verb+blue+', '\verb+orange+', '\verb+yellow+'\} & 1.0 \\
      \{'\verb+blue+', '\verb+yellow+'\} & 1.0 \\
      \{'\verb+blue+', '\verb+orange+'\} & 1.0 \\
      \{'\verb+blue+'\} & 1.0 \\
      \{'\verb+orange+', '\verb+yellow+'\} & 5.7255e-20
    \end{tabular}
    & \begin{tabular}[t]{ll}
      \{'\verb+blue+'\} & 1.0 \\
      \{'\verb+orange+', '\verb+yellow+'\} & 5.1360e-20 \\
      \{'\verb+yellow+'\} & 0.0 \\
    \end{tabular} 
   & \begin{tabular}[t]{ll}
        blue		&	1.0 \\
        yellow		&	3.15747602e-20 \\
        orange	&		2.56716262e-20 \\
    \end{tabular}
   & 	1.5213470e-18 \\\\ \addlinespace \midrule \\ 
     \multirow{5}{1.9cm}[2.9ex]{\raisebox{-0.1\totalheight}{\includegraphics[width=0.12\textwidth]{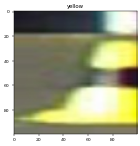}}} 
     & \begin{tabular}[t]{ll}
      \{'\verb+blue+', '\verb+orange+', '\verb+yellow+'\} & 0.999995 \\
      \{'\verb+blue+', '\verb+yellow+'\} & 0.999994 \\
      \{'\verb+orange+', '\verb+yellow+'\} & 0.903126\\
      \{'\verb+blue+', '\verb+orange+'\} & 0.84896 \\
      \{'\verb+yellow+'\} & 0.6536186
    \end{tabular}
    & \begin{tabular}[t]{ll}
      \{'\verb+yellow+'\} & 0.653616 \\
      \{'\verb+blue+'\} & 0.477563 \\
      \{'\verb+blue+', '\verb+orange+'\} & 0.371394 \\
      \{'\verb+orange+', '\verb+yellow+'\} & 0.249502\\
     \{'\verb+orange+'\} & 4.9795e-06
    \end{tabular}
   & \begin{tabular}[t]{ll}
        yellow		&	0.5058069229 \\
        blue		&	0.3906998634 \\
        orange	&		0.1034862920 \\
    \end{tabular}
   & 	1.3657757043
    \\
    \\ \addlinespace
    \bottomrule
        \vspace{-10pt}
\end{tabular}
\end{table}

Tab. \ref{tab:indomain-uncertainty-accuracy} presents the performance of RS-NN, CNN, and LB-BNN across three datasets: ROAD, OBR-A, and CONE. It includes classification accuracy, predictive entropy, and confidence metrics, specifically for incorrect (ICC) and correct classifications (CC). RS-NN demonstrates superior performance across all datasets, achieving the highest accuracy (\textit{e.g.}, 99.78\% on CONE) and the most favorable uncertainty metrics, with low ICC and high CC values. CNN also shows competitive CC scores (\textit{e.g.}, 0.977 on ROAD), but suffers from higher ICC, indicating overconfidence in incorrect predictions. LB-BNN, despite being a Bayesian model, underperforms in both accuracy and calibration, with generally higher entropy and ICC values.

The CONE dataset is relatively small in size and is included primarily to evaluate how each model performs under limited-data conditions. It serves as a stress test for generalization when training data is scarce. While RS-NN, CNN, and LB-BNN all achieve high accuracy on CONE, RS-NN maintains better uncertainty calibration, highlighting its robustness even in low-data regimes. More results on larger datasets lie in the following chapter, which presents a comprehensive evaluation on the more challenging and diverse task of weather classification.

\section{Weather classification}

The R-WAYMO dataset, adapted from the original WAYMO data for weather classification, serves as a challenging benchmark for evaluating models under diverse environmental conditions. Among the evaluated models in Tab. \ref{tab:indomain-uncertainty-accuracy}, RS-NN stands out for delivering both high classification performance and well-calibrated uncertainty estimates. It consistently shows lower uncertainty in its incorrect predictions and maintains high confidence when predictions are correct. In contrast, LB-BNN and CNN show either less accurate performance or poorer uncertainty calibration, particularly struggling with overconfident errors. Overall, RS-NN demonstrates a strong balance between accuracy and reliability, making it especially suited for complex, real-world scenarios like R-WAYMO.

Similarly, on the OBR-A dataset, RS-NN again outperforms its counterparts, offering more accurate classifications while also producing lower predictive entropy and more balanced confidence scores. While LB-BNN and CNN follow closely in accuracy, they show less consistent uncertainty behavior, particularly CNN, which tends to be overconfident even when incorrect.

\begin{table}[h]
  \caption{Performance of RS-NN, CNN, and LB-BNN on R-WAYMO, OBR-A, and CONE datasets. We report classification accuracy, predictive entropy, and confidence estimates for both incorrect classifications (ICC $\downarrow$) and correct classifications (CC $\uparrow$).}
  \label{tab:indomain-uncertainty-accuracy}
  \vspace{-8pt}
  \small
  \centering
  \resizebox{0.9\linewidth}{!}{
  \begin{tabular}{llccc}
    \toprule
    \textbf{Model} & \textbf{Dataset} & \textbf{Accuracy (\%)($\uparrow$)} & \textbf{Entropy} & \textbf{Confidence (ICC($\downarrow$) / CC($\uparrow$))} \\
    \midrule

    \multirow{3}{*}{RS-NN}
    & CONE           & \textbf{99.78} & 0.019 $\pm$ 0.207 & \textbf{0.589 $\pm$ 0.026} / 0.922 $\pm$ 0.037 \\
    & R-WAYMO           & \textbf{76.27} & 0.160 $\pm$ 0.685 & \textbf{0.554 $\pm$ 0.187} / 0.966 $\pm$ 0.103 \\
    & OBR-A          & \textbf{96.42} & 0.030 $\pm$ 0.287 & \textbf{0.692 $\pm$ 0.230} / \textbf{0.990 $\pm$ 0.042} \\

    \midrule

    \multirow{3}{*}{LB-BNN}
    & CONE           & 98.73 & 0.562 $\pm$ 0.371 & 0.787 $\pm$ 0.160 / 0.838 $\pm$ 0.156 \\
    & R-WAYMO           & 70.85 & 0.183 $\pm$ 0.366 & 0.644 $\pm$ 0.364 / 0.846 $\pm$ 0.732 \\
    & OBR-A          & 88.82 & 0.495 $\pm$ 0.053 & 0.730 $\pm$ 0.466 / 0.954 $\pm$ 0.544 \\

    \midrule

    \multirow{3}{*}{CNN}
    & CONE           & 96.92 & 0.282 $\pm$ 0.361 & 0.717 $\pm$ 0.171 / \textbf{0.958 $\pm$ 0.088} \\
    & R-WAYMO           & 71.54 & 0.285 $\pm$ 0.315 & 0.888 $\pm$ 0.152 / \textbf{0.977 $\pm$ 0.078} \\
    & OBR-A          & 83.41 & 0.210 $\pm$ 0.241 & 0.841 $\pm$ 0.164 / 0.964 $\pm$ 0.089 \\

    \bottomrule
  \end{tabular}}
  \vspace{-10pt}
\end{table}

\subsection{Domain Adaptation setting}

Both R-WAYMO and OBR-A (weather classification) datasets are collected by the respective AVs and share common weather classes— \textit{Sunny}, \textit{Rainy}, \textit{Cloudy}—while the OBR-A dataset includes exclusive classes like \textit{Droplet} and \textit{Night}, and R-Waymo includes \textit{Overcast}, \textit{Clear}, and \textit{Partial} as exclusive classes, as shown in Fig. \ref{fig:domain-adaptation}.

The models were trained on the smaller OBR-A dataset (3,468 images) and tested on the much larger, more diverse R-Waymo dataset (39,987 images).
The goal is to evaluate each model's performance in two key areas: (a) \textit{accuracy} in predicting weather conditions when faced with similar but previously unseen data from a different domain or environment (same class, different domain), and (b) the ability to estimate \textit{uncertainty} when encountering entirely new conditions that the model has not been trained on (different class, different domain). RS-NN significantly outperformed both the standard CNN and the Bayesian model in both:
\begin{itemize}[label=\textbullet, leftmargin=*]
  \setlength\itemindent{0pt}
  \setlength\parindent{0pt}  
    \item \textbf{Test Accuracy:} On the shared classes in the R-Waymo test set (approximately 30,087 images), RS-NN achieved a test accuracy of 75.51\%, significantly outperforming the CNN (63.78\%) and the Bayesian model (61.83\%). This demonstrates that RS-NN, despite being trained on a smaller dataset, is more effective at generalizing to new data, leveraging the additional information captured through random sets.
    \item \textbf{Uncertainty Estimation:} As shown in Fig. \ref{fig:domain-adaptation}, the entropy distribution of RS-NN model is wider and higher for \textit{shared classes} like Sunny, Rainy, and Cloudy, compared to the Bayesian model. {This improved performance suggests that RS-NN may have better uncertainty estimation.} Furthermore, for \textit{exclusive} R-Waymo classes that the model has never seen before, RS-NN's uncertainty of the predictions remains high, while the Bayesian model (LB-BNN) exhibits predominantly lower uncertainty, with most of its entropy values clustered around zero.
\end{itemize}

\begin{figure}[!h]
    \centering
    \vspace{-10pt}
    \includegraphics[width=\linewidth]{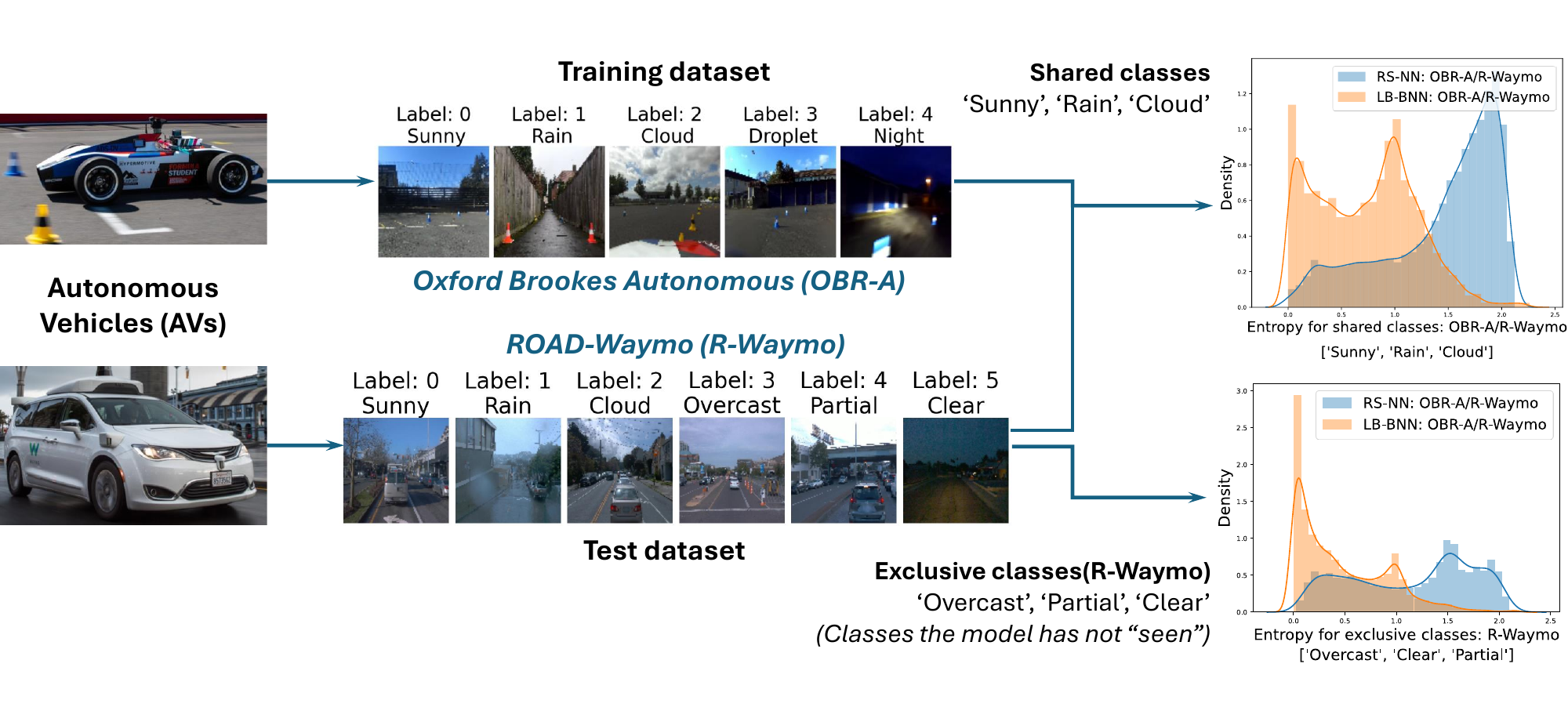}
        \vspace{-30pt}
    \caption{\textbf{Experimental setup for domain adaptation:} RS-NN and LB-BNN trained on OBR-A dataset are tested on R-Waymo. Comparison of entropy distributions (uncertainty estimates) for shared classes of both the datasets, and on the exclusive classes of the test dataset (R-Waymo) is shown.}
    \label{fig:domain-adaptation}
\end{figure}

\begin{table}[h]
\caption{Performance of RS-NN, CNN, and LB-BNN on domain adaptation tasks across OBR-A and ROAD datasets. Only classification accuracy and entropy are reported, as confidence metrics are not directly applicable in cross-domain settings.}
  \label{tab:domain-adaptation-accuracy-entropy}
    \vspace{-8pt}
  \small
  \centering
  \resizebox{0.7\linewidth}{!}{
  \begin{tabular}{llcc}
    \toprule
    \textbf{Model} & \textbf{Datasets} & \textbf{Accuracy(\%)($\uparrow$)} & \textbf{Entropy($\uparrow$)} \\
    \midrule

    \multirow{2}{*}{RS-NN}
    & OBR-A/R-WAYMO     & \textbf{75.51} & \textbf{0.659 $\pm$ 0.885} \\
    & R-WAYMO/OBR-A     & \textbf{78.47} & \textbf{1.163 $\pm$ 0.359} \\

    \midrule

    \multirow{2}{*}{LB-BNN}
    & OBR-A/R-WAYMO     & 61.83 & 0.443 $\pm$ 0.464 \\
    & R-WAYMO/OBR-A     & 28.57 & 0.287 $\pm$ 0.414 \\

    \midrule

    \multirow{2}{*}{CNN}
    & OBR-A/R-WAYMO     & 63.78 & 0.370 $\pm$ 0.295 \\
    & R-WAYMO/OBR-A     & 38.09 & 0.498 $\pm$ 0.381 \\

    \bottomrule
  \end{tabular}}
  \vspace{-4pt}
\end{table}

Tab. \ref{tab:domain-adaptation-accuracy-entropy} reports classification accuracy and entropy for domain adaptation tasks across OBR-A and R-WAYMO. RS-NN shows the best generalization in both directions, while CNN and LB-BNN struggle, especially when adapting to unseen domains. Entropy values reflect RS-NN’s stronger calibration under distribution shifts.

\section{Discussion}

The results presented in this chapter highlight the practical strengths of Random-Set Neural Networks (RS-NNs) in real-world AV perception tasks, particularly under uncertainty and domain shift. Across cone and weather classification datasets, RS-NN consistently outperforms both standard CNNs and Bayesian neural networks (LB-BNN) in terms of accuracy, uncertainty calibration, and domain generalization. These findings validate the core hypothesis of this work: RS-NNs provide a robust and uncertainty-aware framework that improves predictive reliability in safety-critical environments like autonomous driving.

\chapter{Conclusions} 
\label{ch:conclusion} 

\section{Thesis Summary}
\label{sec:thesis-summary} 

This thesis provides concrete contributions that directly address each of the research questions outlined in Sec. \ref{sec:research-ques}. In doing so, it lays foundational groundwork for the emerging field of \textbf{Epistemic AI}, where the principled modeling of ignorance is a central priority.

\textit{To answer how epistemic uncertainty can be modeled effectively and robustly}, at the core of this thesis is the development of \textbf{\textit{Random-Set Neural Networks (RS-NNs)}} (Chapter \ref{ch:rsnn}), designed to address critical challenges in epistemic uncertainty modeling. RS-NNs model uncertainty over sets of outcomes, effectively capturing both partial knowledge and ambiguity. They demonstrate robustness against previously unseen and adversarial inputs by producing higher entropy and wider credal sets in regions of uncertainty. This capability results in improved out-of-distribution (OoD) detection (Sec. \ref{sec:ood-detection}) and enhanced classification performance under adversarial perturbations (Sec. \ref{app:fgsm}) across multiple benchmarks, including ImageNet. \textbf{\textit{Credal set models}}, which predict probability intervals to represent uncertainty through sets of plausible distributions, are briefly outlined and compared to other approaches in Sec. \ref{sec:impact-epi}.

\textit{To fairly and rigorously compare epistemic uncertainty predictions from fundamentally different output structures}, ranging from scalar probabilities to belief functions, this thesis introduces a \textbf{\textit{unified evaluation framework}} (Chapter \ref{ch:eval_framework}). The framework standardizes diverse uncertainty outputs by transforming them into credal sets and applies principled metrics such as KL divergence and non-specificity within the geometry of the probability simplex. This enables comprehensive, task-relevant evaluation across Bayesian, ensemble, evidential, credal set and random-set models without declaring a universally superior approach, but instead identifying the best fit based on the precision and imprecision needs of each application.

\textit{To answer how to address computational challenges in random-set uncertainty modeling for scalability}, the thesis tackles the combinatorial complexity by proposing a \textit{budgeting} (Sec. \ref{sec:rsnn-budgeting}) technique that efficiently selects informative focal sets, reducing combinatorial complexity and making RS-NNs scalable to large datasets and architectures. Building on these advances, the framework is extended to \textbf{\textit{Random-Set Large Language Models (RS-LLMs)}} (Chapter \ref{ch:rsllm}), which improve token-level uncertainty quantification and downstream performance in natural language processing tasks such as question answering. Additionally, the applicability of RS-NNs is demonstrated in \textbf{\textit{autonomous racing}} (Chapter \ref{ch:weather}) environments, further validating the robustness and scalability of the approach in real-world dynamic domains.

The following are the \textbf{key findings} of this work:

\begin{itemize}
    \item RS-NNs attained \textbf{state-of-the-art accuracy} across benchmarks, outperforming Bayesian and ensemble baselines by up to 4\% on CIFAR-10 and surpassing competitors by as much as 7\% on the challenging, large-scale ImageNet dataset.
    
    \item Extensions to \textbf{RS-LLMs }improved token-level prediction accuracy by up to 7\% on NLP tasks such as CoQA and OBQA, while enhancing uncertainty quantification.

    \item RS-NNs demonstrated up to 40\% better accuracy under \textbf{domain shifts in weather classification} tasks, showing especially strong robustness and generalization across diverse and ambiguous real-world radar conditions such as fog, rain, and snow. RS-NNs generalize well to unseen classes, as evidenced in R-WAYMO exclusive classes where they maintain high entropy and wide credal widths, outperforming LB-BNN.

    \item RS-NNs outperform competing models by achieving up to 6\% higher performance on \textbf{out-of-distribution (OoD) detection}. They handle \textbf{unseen samples} more effectively by consistently exhibiting lower entropy for in-distribution (iD) samples and higher entropy for OoD samples, resulting in a pronounced entropy gap (e.g., 0.45 on CIFAR-10 vs SVHN), surpassing DE and LB-BNN. Additionally, RS-NNs show larger credal set widths for OoD samples, reflecting stronger epistemic uncertainty (e.g., iD width of 0.007 vs OoD width of 0.260 on CIFAR-10 vs SVHN).

    \item RS-NNs achieve up to 20\% better \textbf{calibration} than Bayesian models, maintaining a low Expected Calibration Error (ECE < 0.05) even under dataset shifts. Their uncertainty is also increased by up to 15\% for more challenging OoD samples.

    \item Under FGSM and PGD \textbf{adversarial attacks}, RS-NN maintained up to 50\% higher accuracy compared to baselines, demonstrating resilience that supports the claim of reliable performance under adversarial conditions.
        
    \item Most importantly, RS-NNs demonstrated \textbf{efficient inference times} (Tab. \ref{tab:uq-comparison-table}) of approximately \textbf{1.91 ms per sample}, vastly outperforming Bayesian models like FSVI (340.25 ms/sample) and ensembles such as Deep Ensembles (DE) (13,163.50 ms/sample). DE achieves a CIFAR-10 accuracy of 92.73\%, RS-NNs surpass this with 93.53\% accuracy (an improvement of 0.8\%). However, the extremely high inference time of DEs makes them impractical for real-world applications, highlighting the scalability and deployment advantages of RS-NNs.

    \item Using the \textbf{unified evaluation} framework introduced in this thesis, which balances distance-based accuracy and imprecision via a trade-off parameter, RS-NNs are consistently selected as the best-performing model at the \textbf{trade-off value of 1}. This indicates that when equal weight is given to accuracy and imprecision, RS-NNs provide the most favorable balance. {However, RS-NN shows poorer performance at lower trade-off values. Additionally, the interpretation of the metric is based on several underlying assumptions.}
\end{itemize}

\section{Impact on the Epistemic AI paradigm}\label{sec:impact-epi}

\begin{figure}[!ht]
    \centering
    \begin{minipage}[b]{0.49\textwidth}
        \centering
        \includegraphics[width=\textwidth]{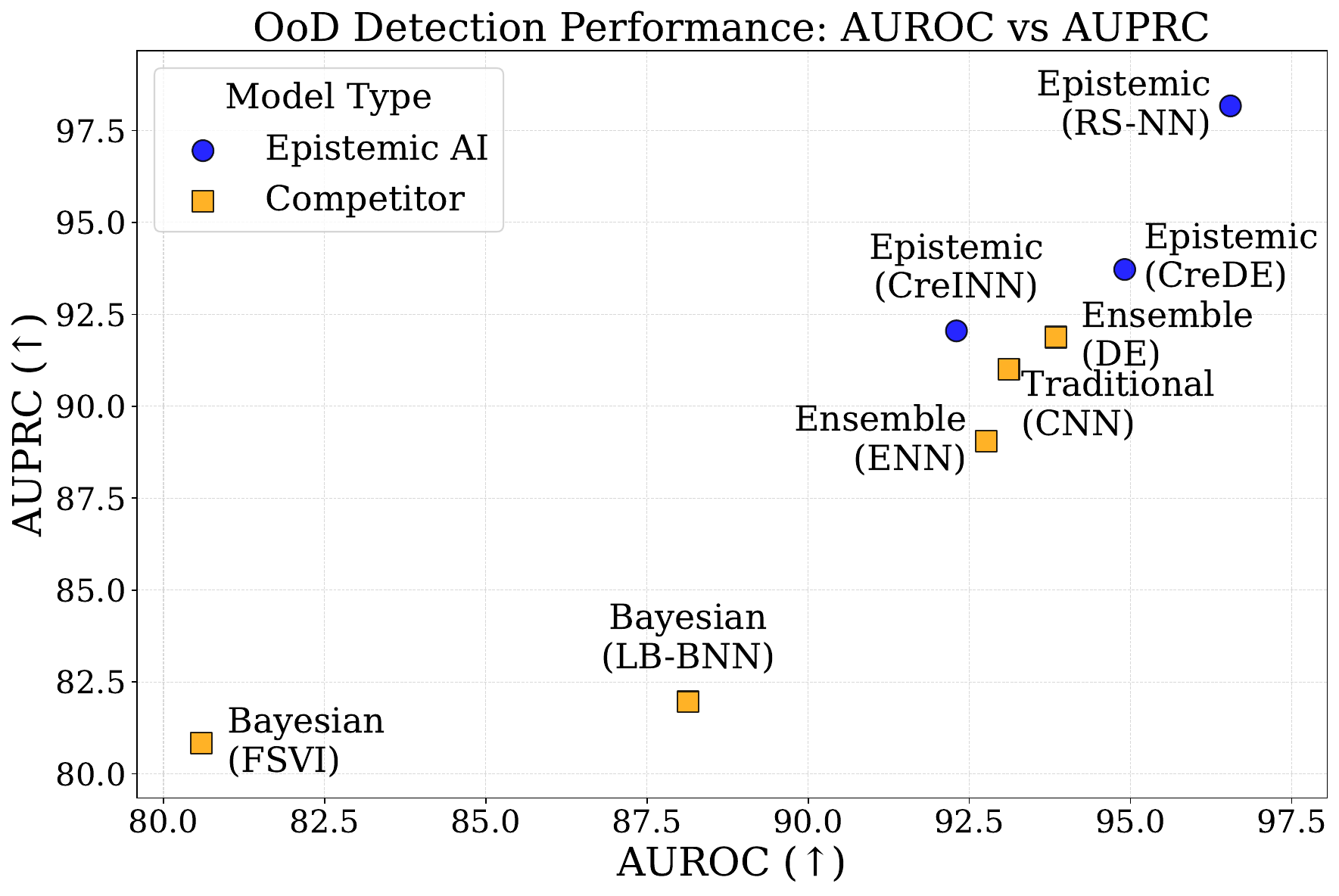}
        \vspace{-15pt}
        \caption*{(a)}
    \end{minipage}
    \hfill
    \begin{minipage}[b]{0.49\textwidth}
        \centering
        \includegraphics[width=\textwidth]{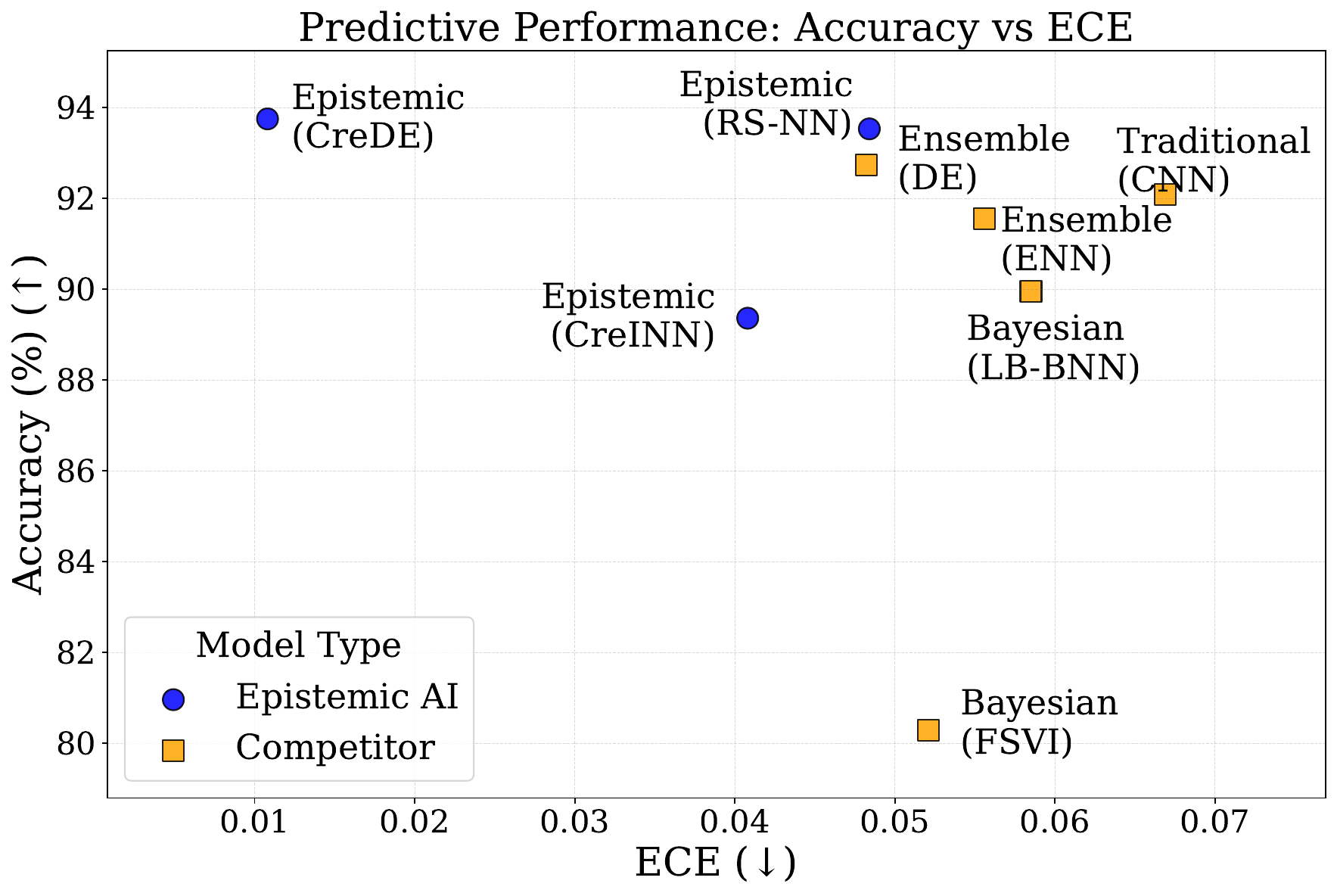}
        \vspace{-15pt}
        \caption*{(b)}
    \end{minipage}
    \vspace{-8pt}
    \caption[Comparison of {Epistemic AI models} (circles) and {competitor models} (squares) on CIFAR-10: 
    (a) {OoD detection performance (AUROC vs.\ AUPRC)},
    (b) Predictive performance (Accuracy vs.\ ECE).]{Comparison of \textbf{Epistemic AI models} (circles) and \textbf{competitor models} (squares) on CIFAR-10. 
    (a) \textbf{OoD detection performance (AUROC vs.\ AUPRC)}. Epistemic AI models cluster in the top-right (\textit{high separability}) while {competitor} methods show a much greater spread (\textit{lower performance}.
    (b) \textbf{Predictive performance (Accuracy vs.\ ECE).} {Epistemic AI} models cluster in the top-left (\textit{high accuracy, low calibration error}) while {competitor} methods show poorer trade-offs (\textit{weaker calibration}).}
    \label{fig:ood-acc-ece}
    \vspace{-8pt}
\end{figure}

This section highlights the overall impact of the models introduced in this thesis within the Epistemic AI framework. 
This thesis presents three models developed under the Epistemic AI paradigm using second-order uncertainty measures: RS-NN (Chapter \ref{ch:rsnn}), the primary contribution, and two collaborative works: CreINN (Sec. \ref{sec:creinn}) \citep{wang2024creinns} and CreDE (Sec. \ref{sec:crede}) \citep{wang2024credal}. These models demonstrated superior performance over leading competitor approaches, including \textit{Bayesian} models such as LB-BNN \citep{hobbhahn2022fast} and FSVI \citep{rudner2022tractable}, and \textit{Ensemble}-based models like Deep Ensembles (DE) \citep{lakshminarayanan2017simple} and Epistemic Neural Networks (ENN) \citep{osband2024epistemic}. These models were used as baselines in the experimental evaluations in Chapters \ref{ch:rsnn} and \ref{ch:eval_framework}. Evaluated on large-scale benchmarks such as ImageNet, the proposed Epistemic AI models achieved notable improvements in accuracy, robustness, uncertainty quantification, and out-of-distribution (OoD) detection, enhancing the ability to recognize unfamiliar or anomalous inputs.

As shown in Fig. \ref{fig:ood-acc-ece}(a), the Epistemic AI models (depicted as circles) cluster in the top-right region on CIFAR-10, indicating consistently strong OoD detection performance, attributed to their capacity to explicitly preserve and express ignorance. In contrast, competitor models (squares) exhibit lower and more variable performance, underscoring the theoretical advantage of second-order uncertainty under distribution shift. Furthermore, Fig. \ref{fig:ood-acc-ece}(b) illustrates that Epistemic AI models strike a superior balance between accuracy and calibration, dominating the top-left region of the plot with both high predictive accuracy and low Expected Calibration Error (ECE). 


\begin{table}[!h]
\caption{Training (in minutes; per 100 epochs) and inference time (in milliseconds; per sample) comparison of uncertainty estimation methods on the CIFAR-10 dataset.}
\label{tab:uq-comparison-table}
\vspace{-5pt}
\begin{center}
\resizebox{\textwidth}{!}{%
\begin{sc}
\begin{tabular}{lccc}
\toprule
\textbf{Model}         & \textbf{Training Time (100 epochs) (min)} & \textbf{Inference Time (ms/sample)} \\
\midrule
Traditional                      & 85.33                  & 1.91 $\pm$ 0.7                                            \\
Deterministic (DDU) \citep{mukhoti2023deep} &  243.85    &   59.35   $\pm$ 0.40   \\
Bayesian (LB-BNN) \citep{hobbhahn2022fast}  & 107.90                          & 7.11 $\pm$ 0.89                                            \\
Bayesian (FSVI) \citep{rudner2022tractable}   & 1518.35                         & 340.25 $\pm$ 0.76                                               \\
Ensemble (DE) \citep{lakshminarayanan2017simple}  & 426.66                     & 13163.50 $\pm$ 3.37                                             \\
Ensemble (ENN) \citep{osband2024epistemic}      & 712.30                          & 3.10 $\pm$ 0.03                                                 \\
Evidential (EDL) \citep{sensoy} & 188.57 & 6.12 $\pm$ 0.01\\
Evidential (Hyper-Opinion EDL) \citep{qu2024hyper} & 186.56  & 23.01 $\pm$ 0.15\\
Epistemic AI (CreDE) \citep{wang2024credal}                  & 122.95                   & 63.0 $\pm$ 1.1                                          \\
Epistemic AI (RS-NN)          & 113.23                                   & 1.91 $\pm$ 0.02                      \\
\bottomrule
\end{tabular}
\end{sc}
}
\end{center}
\end{table}

In Tab. \ref{tab:uq-comparison-table}, we present the training and inference times (computational costs) for the uncertainty methods discussed in Sec. \ref{sec:uncertainty-models} and Fig. \ref{fig:uncertainty-models}. Two examples of each model type are shown, all trained on the ResNet50 backbone. More training details are given below.
These models differ not only in their uncertainty modeling principles but also in computational costs (see Tab. \ref{tab:uq-comparison-table}), reflecting a spectrum of trade-offs between performance and efficiency. For instance, function-space Bayesian methods provide uncertainty but at a high computational cost \citep{rudner2022tractable}, while RS-NN offer competitive accuracy with efficient inference.

\section{Limitations}
\label{sec:limitations} 

While the application of second-order uncertainty measures, such as those based on sets of distributions including credal and random-set models, offers powerful means to represent epistemic uncertainty, it also introduces notable challenges. These methods often require computationally intensive, especially during decision-making processes \citep{augustin2014introduction, augustin2022statistics}. In this thesis, we use \textit{budgeting} techniques (Sec. \ref{sec:rsnn-budgeting}) for RS-NN and credal set approximations for other methods. However, extending these efficient approaches to a broader class of second-order representations remains an open research direction.

Regarding Random-Set Neural Networks (RS-NNs) (Chapter \ref{ch:rsnn}) specifically, the budgeting step designed to reduce the combinatorial complexity of focal set selection can sometimes be time-consuming for very large datasets, although it is a one-time pre-training procedure for a given dataset. For large-scale datasets, this process can be made more tractable by applying dimensionality reduction techniques (e.g., t-SNE, autoencoders \citep{ghasedi2017deep, guo2017deep}, or PCA \citep{abdi2010principal}) on representative samples rather than the entire dataset, significantly decreasing computational overhead. Moreover, the current method requires manual tuning of the number of focal elements ($K$), which could be improved by developing dynamic strategies that adapt $K$ based on focal set overlap or other criteria. The budgeting technique itself also holds promise for extension to multi-source or streaming data settings, but these applications require further development. Alternative approaches for subset selection, such as leveraging sparse mass functions \citep{itkina2020evidential, chen2021evidential}, may provide a more principled, quantitative basis for budgeting and remain a valuable area for exploration.

\textbf{Random-Set Neural Networks (RS-NN).} {Future work} includes exploring alternative ways to ensure valid belief functions without relying on mass regularization such as integrating Semantic Probabilistic Layers \citep{ahmed2022semantic} or applying softmax over masses (see Appendix~\ref{app:hyperparam}). Additionally, extending RS-NNs to parameter-level representations and regression tasks will be pursued.
The budgeting procedure can also be modified to accommodate multiple data sources or a continuous stream of data.
t-SNE can be replaced by an autoencoder (this will need to be trained first) or PCA to add continual inference capabilities in the dimensionality reduction step. Similarly, Gaussian-based dynamic probabilistic clustering (GDPC) \cite{diaz2018clustering} can be used instead of GMM clustering to handle a continuous stream of data. GDPC works by first initialising a GMM and then updating its parameters as it sees new data. In alternative, \cite{kulis2011revisiting} have proposed an interesting approach to learn a GMM from multiple sources of data by maintaining a local and a global mixture model. Both these approaches can be plugged into our budgeting framework 
with little modifications 
to add continual learning capabilities to it. 
The 
computation of cluster overlaps
remains the same, so overlapping scores and the resultant focal sets 
will need to 
be updated as the clusters evolve. However, a cluster tracking strategy \cite{barbara2001tracking} can be employed so that the overlap assessment step is only re-done when a sufficient drift has been detected in the clusters. 
All the proposed changes are efficient enough to not have a significant effect on the overall time of budgeting.  Random-Set Neural Networks may also support regression, {e.g.}, for object detection, by predicting Dirichlet distributions over Borel closed intervals \citep{strat1984continuous} for bounding box coordinates. Class labels can be modeled as sets, enabling robust uncertainty in both spatial and categorical outputs.

The proposed evaluation metric (Chapter \ref{ch:eval_framework}) is a function of various design choices: (i) the choice of mapping all predictions to credal sets using lower probabilities, (ii) the use of specific distance and non-specificity measures, and (iii) the metric being a linear combination of them. While these choices are theoretically motivated and supported by ablation studies (Secs. \ref{app:ablation-kl} and \ref{app:ablation-ns}), alternative formulations and broader validation are important avenues for future work. Additionally, practical computation of this metric depends heavily on the model class and dataset; for example, training E-CNN \citep{tong2021evidential} (an imprecise model) on more complex datasets like CIFAR-100 proved infeasible, highlighting ongoing computational and architectural constraints.

Overall, this thesis provides a strong contribution toward addressing core uncertainty quantification challenges, while several important avenues remain to be explored in computational efficiency, evaluation standardization, and practical scalability.

\section{Broader challenges within Epistemic Deep Learning}
\label{sec:challenges} 

Despite significant contributions presented in this thesis, several key challenges remain in advancing Epistemic AI and second-order uncertainty quantification.

\textbf{Scaling up.} Most existing evidential approaches \citep{sensoy} struggle to scale beyond medium-sized datasets due to computational and representational bottlenecks. The budgeting method proposed in this thesis addresses this by enabling the tractable use of random-set representations on large-scale datasets such as ImageNet and architectures like Vision Transformers. This opens up further directions, such as employing Dirichlet mixture models \citep{yin2014dirichlet} or dynamic clustering techniques \citep{shafeeq2012dynamic} to support scalability and continual adaptation. However, a central question remains: \textit{Can epistemic representations scale to foundation models and truly massive datasets?} Notably, quantum approaches offer new promise, with recent developments in belief representation \citep{zhou2023bf}, combination mechanisms \citep{zhou2024combining}, and their integration into quantum circuits \citep{wu2024novel}.

\textbf{From one-off to continual learning.} Most deep learning pipelines assume static training datasets, yet real-world systems must learn continuously from streaming data with shifting distributions. This thesis focuses primarily on static settings, but adapting Epistemic AI for continual learning presents a major open challenge. Existing continual learning work largely centers on avoiding catastrophic forgetting \citep{kirkpatrick2017overcoming}, using priors, task-specific parameters, or replay buffers \citep{rolnick2019experience}. More recent developments include domain-incremental learning \citep{van2019three}, and online convex optimization techniques \citep{dall2020optimization} that offer robustness via regret minimization. Although some recent methods explore this direction for uncertainty-aware models \citep{zheng2021continual, jha2024npcl}, a unified framework that links continual learning and second-order epistemic uncertainty remains largely unexplored.

\textbf{Learning and symbolic reasoning under uncertainty.}
Epistemic uncertainty can be reduced by collecting more data or incorporating prior knowledge, such as symbolic information (\textit{e.g.}, Snorkel \citep{ratner2017snorkel}), but data alone does not guarantee better performance, as seen in autonomous vehicle failures. \emph{Neurosymbolic AI} integrates symbolic reasoning with deep learning to regularize predictions and enable knowledge transfer across domains \citep{d2009neural, mao2019neuro}.
Current NeurAI frameworks enforce symbolic constraints but struggle with assessing output frequency or scaling to large knowledge bases \citep{aditya2019integrating}. Approaches like DeepProbLog \citep{manhaeve2018deepproblog} and DL2 \citep{fischer2019dl2} leverage fuzzy and probabilistic semantics but lack epistemic uncertainty modeling. Potential solutions include designing epistemic semantic losses
or using logical circuits like trigger graphs 
\citep{tsamoura2021materializing} to extend DeepProbLog-style reasoning.

\textbf{Establishing statistical guarantees.} Epistemic models introduced in this thesis offer rich uncertainty representations, but like most approaches in the field, they do not yet provide frequentist statistical guarantees. However, RS-NNs can be combined with conformal prediction techniques (\S\ref{app:statistical}) to yield distribution-free coverage guarantees. Broader efforts to generalize conformal learning to second-order settings are ongoing. Notably, extensions of confidence intervals to belief functions, such as confidence structures \citep{denoeux2018frequency}, enable frequency-calibrated belief representations. Within the random-set framework, Inferential Models (IMs) provide belief functions with strong frequentist properties \citep{martin2015inferential}, while predictive belief functions \citep{denoeux2006constructing} can express epistemic uncertainty that is provably less committed than the true distribution with a specified probability. These approaches could help bridge the gap between formal guarantees and expressive uncertainty modeling.

{
\textbf{Adaptive budgeting.} In this thesis, we tackled scalability for medium to large datasets by introducing a budgeting mechanism (Sec. \ref{sec:rsnn-budgeting}, Chapter \ref{ch:rsnn}) that constrains the number of focal sets in RS-NN, making computation feasible on datasets like ImageNet and architectures such as Vision Transformers. However, when extending these epistemic approaches to \textit{foundation models}, which are vastly larger and trained on massive datasets, new challenges emerge. The number of focal sets can grow exponentially, making budgeting essential but also requiring more advanced pruning and summarization techniques to keep inference and training tractable. Computational and memory demands increase significantly, necessitating highly optimized algorithms and hardware-aware solutions. Additionally, foundation models often undergo continual learning and fine-tuning, which calls for adaptive budget management strategies to efficiently update uncertainty representations. Moreover, the complex architectures of foundation models incorporating multi-modal inputs and sophisticated attention mechanisms pose challenges for effectively representing and propagating uncertainty without excessive overhead. Thus, while budgeting has proven useful at smaller scales, scaling RSNNs to foundation models will require new methods for scalable focal set management, adaptive budgeting, and efficient uncertainty propagation to maintain epistemic expressiveness at massive scale. }

\chapter{Future Directions}
\label{sec:future-directions} \label{sec:opportunities}

This thesis has introduced novel approaches to epistemic deep learning, notably through Random-Set Neural Networks and a unified evaluation framework. However, much remains to be explored to make epistemic models more expressive, scalable, and impactful, particularly in high-stakes scientific and engineering domains.

\textbf{Advancing Random-Set Learning.} This thesis has unlocked the power of random-set learning across image classification, text classification, and language generation with LLMs, revealing its versatile potential. The next exciting frontier is pushing this approach into truly multimodal realms, such as image generation in diffusion models by quantizing images into discrete tokens. Although high-resolution images pose challenges, exploring this could revolutionize how uncertainty and ambiguity are modeled in complex visual data.

From a modeling standpoint, removing the need for explicit mass regularization by integrating semantic probabilistic layers \citep{ahmed2022semantic} or applying a softmax over mass functions could enforce belief validity more naturally. The budgeting procedure itself can be made more dynamic and scalable by enabling continual learning: replacing t-SNE with an autoencoder or PCA for online dimensionality reduction, and adopting stream-compatible clustering methods such as Gaussian-based Dynamic Probabilistic Clustering (GDPC) \citep{diaz2018clustering} or mixture modeling across multiple sources \citep{kulis2011revisiting}.
Efficient cluster tracking techniques \citep{barbara2001tracking} can then trigger overlap recomputation only when significant drift is detected, maintaining both accuracy and efficiency. RS-NNs also present a pathway to regression tasks, such as object detection, by predicting Dirichlet distributions over Borel closed intervals \citep{strat1984continuous}, thus enabling uncertainty modeling in both class labels and spatial coordinates.

The next frontier for the random-set approach lies in extending from target-space to parameter-space representations. \textit{How can Bayesian posteriors be translated into random-set posteriors over model weights?}
An approach is to transform Bayesian posterior distributions into random-set posteriors using Shafer’s mapping \citep{shafer1976mathematical}, efficiently encoded via Dirichlet distributions over parameter intervals. 

\textbf{Towards a More Unified Evaluation Framework.
}The proposed evaluation metric opens avenues not only for benchmarking epistemic predictions but also for training models via loss-based optimization. Future work can explore its integration as a training objective, helping models internalize and optimize their uncertainty representations. Extending this framework to structured outputs and multi-label classification could broaden its utility. Future work includes applying the evaluation framework on a real-world decision-making task (as a classification problem), moving beyond synthetic benchmarks. Additionally, deeper analysis will be done to assess model performance without overly coarse aggregations. 

\textit{A central question is whether a single epistemic model can generalize effectively across a wide range of tasks}, or whether specific applications require fundamentally different uncertainty structures. This also raises the critical challenge of whether a principled, task-aware trade-off between models can be computed easily and meaningfully for practical deployment. 

\textbf{Towards an Epistemic Generative AI.}
Generative AI systems, particularly LLMs, face critical trust challenges due to hallucinations and overconfidence. Current solutions like Laplace-LORA \citep{yang2023bayesian}, Bayesian fine-tuning \citep{wang2024blob}, and ensemble-based methods address some of these issues but remain either computationally expensive or weakly grounded in second-order uncertainty.

Epistemic AI introduces a \textit{novel lens}: instead of modeling pointwise likelihoods, we can model belief over sets of possible outputs. Random-Set LLMs (RS-LLMs) predict belief functions over vocabulary tokens, enabling the model to express ambiguity and ignorance explicitly. Hierarchical token embeddings can define meaningful focal sets, while belief-based sampling processes enhance both diversity and robustness of generations. This is especially useful in polysemous or synonym-rich languages like Japanese or Arabic.
A fundamental challenge remains: \textit{How do we teach a model that the examples it sees are only samples from an incredibly rich set of possibilities?}

\textbf{Opportunities in Science and Engineering.}
Epistemic AI presents a major opportunity to advance fields such as drug discovery, materials science, and astronomy. DeepMind’s Alphafold \citep{jumper2021highly} transformed protein structure prediction, but models like Alphafold and those used in weather forecasting often lack uncertainty modeling, which is crucial for real-world tasks like climate change, additive manufacturing, and \textbf{nuclear fusion} plasma simulation. Recent developments with Alphafoldv3 and neural operator (NO) models \citep{magnani2022approximate, garg2023vb, zou2025uncertainty}, applied to nuclear fusion and climate prediction, demonstrate promise but require better uncertainty quantification to improve accuracy and efficiency.

{Neural Operators} are powerful surrogate models for solving PDE-governed problems across science and engineering. Despite advances in Bayesian methods \citep{magnani2022approximate} and conformal prediction (CP) \citep{gray2025guaranteed}, NOs face challenges in uncertainty due to limited data and PDE misspecification. CP offers calibrated uncertainty but depends on costly calibration data. Since neural operators learn mappings between functions, such as boundary conditions and PDE solutions, \textit{how can epistemic learning quantify uncertainty in these functional spaces?} This extends classical neural network uncertainty modeling. Epistemic AI will also help bridge low- and high-fidelity simulation gaps, as shown in fusion plasma edge modeling \citep{faza2024interval}. Given the wide range of NO applications, from climate science to materials research, epistemic methods have significant potential impact.

Climate change is a critical example, altering global weather cycles and increasing extreme events like floods and droughts \citep{samaniego2018anthropogenic}. Accurate prediction depends on correctly representing Earth system compartments, including atmosphere, ocean, and land, and their complex interactions. These compartments evolve dynamically, making reliable long-term forecasts a challenge. \textit{How can models trained on insufficient and sparse data quantify epistemic uncertainty to avoid forecasting errors?} Beyond simple adaptation to streaming data, this uncertainty quantification is essential to improve decision-making with important societal and scientific consequences.


\textbf{Final Remarks.}
Modern AI systems are growing in complexity: \textit{larger} models, \textit{multimodal} inputs, and feedback loops where predictions influence future data. In this evolving landscape, \textit{epistemic uncertainty quantification (UQ) cannot remain a post hoc correction or afterthought}. Instead, it must be embedded at the architectural level, fundamentally shaping how models learn, adapt, and communicate what they know, and crucially, what they do not know, in an interpretable and actionable manner.

Uncertainty estimation in generative foundation models for natural language tasks poses distinct challenges and opportunities. LLMs sometimes demonstrate calibrated confidence scores, yet the mechanisms enabling this calibration remain incompletely understood. Furthermore, the transfer of epistemic AI capabilities to \textbf{vision-language models} (VLMs), especially those fine-tuned with instruction, remains an open question. Future work must dissect how input data properties, training dynamics, and architectural choices affect epistemic uncertainty predictions, addressing existing biases and limitations to improve model calibration, interpretability, and trustworthiness. Progress in this direction is vital for producing more reliable and \textbf{context-aware outputs from generative AI}.

Moreover, most performance measures for these models depend heavily on the \textbf{quality of the model’s predictions} themselves. This dependency means that better uncertainty quantification and epistemic modeling are not only critical for trustworthy predictions but also for the meaningful assessment of model performance. Exploration is needed to develop assessment criteria that truly reflect a model’s knowledge and ignorance, beyond just predictive accuracy.

Beyond technical advances, epistemic UQ must engage with \textbf{human trust and decision-making} processes. Uncertainty is not only a mathematical construct but a critical communication tool. How uncertainty is presented, calibrated, and interpreted deeply influences its practical value. This necessitates integrating epistemic AI research with human-computer interaction, interpretability, and ethical considerations. Building trustworthy AI, especially in high-stakes domains like healthcare and public policy, depends on uncertainties that are both technically sound and socially comprehensible.

Ultimately, epistemic uncertainty quantification is not a peripheral specialty but a foundational pillar of trustworthy AI. It provides a lens for understanding model limitations, the nature of data, and the contours of knowledge itself. Expanding this perspective mathematically, methodologically, and philosophically, will empower the creation of \textbf{\textit{AI systems that are not just intelligent, but wise; systems that `know when they do not know'}}.

\bookmarksetup{startatroot}

\part{{{\MakeUppercase{Appendix}}}}
\label{part:appendix}
\thispagestyle{plain}

\appendix 


\chapter{Algorithms} \label{app:algo}  \label{app:B}
\vspace{-8pt}

This section outlines the algorithms integral to the implementation and evaluation of budgeting (\S{\ref{app:algo-budget}}), expected calibration error (ECE) (\S{\ref{app:algo-ece}}), area under the receiver operating characteristic curve (AUROC) and area under the precision-recall curve (AUPRC) (\S{\ref{app:algo-auroc}}). 

\subsection{Algorithm for Budgeting} \label{app:algo-budget}

For RS-NN with $N$ classes, generating $2^N$ outputs is computationally infeasible due to exponential complexity. Instead, we choose $K$ relevant subsets ($A_K$ focal sets) from the $2^N$ possibilities.

To obtain these $K$ focal subsets, we extract feature vectors from the penultimate layer of a trained standard CNN with $N$ outputs. We then apply t-SNE for dimensionality reduction to 3 dimensions. Note that our approach is agnostic, as t-SNE could be replaced with any other dimensionality reduction technique, including autoencoders. 

Next, we fit a Gaussian Mixture Model (GMM) to the reduced feature vectors of each class. Using the eigenvectors and eigenvalues of the covariance matrix $\Sigma_c$ and the mean vector $\mu_c$ for each class $c$, we define an ellipsoid covering 95\% of the data. The lengths of the ellipsoid's principal axes are computed as $length_{c,i} = 2 \sqrt{7.815 \lambda_i}$, where $\lambda_i$ is the $i^{th}$ eigenvalue. The scalar 7.815 corresponds to a 95\% confidence interval for a chi-square distribution with 3 degrees of freedom.
The class ellipsoids 
are plotted in a 3D space and the overlap of each subset in the power set of $N$ is computed. As it is computationally infeasible to compute overlap for all $2^N$ subsets, we start doing so from cardinality 2
and use early stopping when increasing the cardinality further does not alter the list of most-overlapping sets of classes.
We choose the top-$K$ subsets ($A_K$) with the highest overlapping ratio, computed as the intersection over union for each subset.

\begin{algorithm}[!h]
  \caption{Budgeting Algorithm}
  \label{alg:budgeting}
  \begin{algorithmic}[1]
    \State \textbf{Input:} $\mathcal{D}$ -- Training data with $N$ classes, $\mathcal{C}$ -- The set of classes, $K$ -- Number of non-singleton focal sets
    \State \textbf{Output:} $\mathcal{O}$ -- Set containing $N + K$ focal sets
    
    \State \textit{Initialization}
    \State Extract feature vectors using a trained CNN
    \State Apply t-SNE for dimensionality reduction to 3 dimensions
    
    \For{each class $c$}
      \State Fit GMM to the reduced feature vectors for that class to obtain $\mu_c$ and $\Sigma_c$
      \State Define an ellipsoid covering 95\% data using $\mu_c$ and $\Sigma_c$
    \EndFor
    
    \State ${\text{most\_overlapping\_sets}} \leftarrow$ Initialize an empty list for non-singleton focal sets $A_1, \ldots, A_K$
    \State Set $current\_cardinality \leftarrow 2$
    
    \While{$current\_cardinality \leq N$}
      \State \textit{Compute overlaps for subsets of cardinality $current\_cardinality$}
      \State $\text{overlap}(A) = \frac{\cap_{c \in A}A^c}{\cup_{c \in A}A^c}$
        
      \State \textit{Select top-$K$ subsets with highest overlap}
      \State Update ${\text{most\_overlapping\_sets}}$
      
      \If{no change in ${\text{most\_overlapping\_sets}}$}
        \State \textbf{break}
      \EndIf
      
      \State $current\_cardinality \leftarrow current\_cardinality + 1$
    \EndWhile
    
    \State \textit{Combine selected non-singleton focal sets with $N$ singleton sets}
    \State $\mathcal{O} \leftarrow \mathcal{C} \cup \{ A_1, \ldots, A_K \}$
    
    \State \textbf{return} $\mathcal{O}$
  \end{algorithmic}
\end{algorithm}

\subsection{Algorithm for ECE} \label{app:algo-ece}

\begin{algorithm}[tb]
  \caption{Expected Calibration Error (ECE)}
  \label{alg:ece}

  \begin{algorithmic}[1]
    \Require
      \textit{confidences}: List or array of confidence scores predicted by the model\;
      \textit{predictions}: List or array of predicted class labels\;
      \textit{true\_labels}: List or array of true class labels\;
      $B$: Number of bins for binning confidence scores\;

    \Ensure
      \textit{ECE}: Expected Calibration Error\;

    \State \textbf{Step 1: Normalize Confidences}\;
    Normalize the confidence scores to ensure they are in the range [0, 1].

    \State \textbf{Step 2: Binning}\;
    Divide the confidence scores into \textit{num\_bins} bins.

    \State \textbf{Step 3: Initialize Arrays}\;
    Initialize arrays \textit{bin\_accuracy}, \textit{bin\_confidence}, and \textit{weights} with zeros.

    \State \textbf{Step 4: Populate Arrays}\;
    \For{each bin}
      \State Identify instances falling into the bin\;
      \State Calculate mean accuracy and mean confidence within the bin\;
      \State Update \textit{bin\_accuracy}, \textit{bin\_confidence}, and \textit{weights}\;
    \EndFor

    \State \textbf{Step 5: Calculate ECE}\;
    Calculate the Expected Calibration Error using the populated arrays.
    \begin{equation*}
        ECE = \sum_{i=1}^{B} |(bin\_accuracy_i
        - bin\_confidence_i)| \times weights_i
            \end{equation*}
  \end{algorithmic}
\end{algorithm}

Expected Calibration Error (ECE) is computed by comparing the average confidence and accuracy within each bin. The steps involved are as follows: 
  \begin{enumerate}
    \item For each bin, calculate the absolute difference between the average confidence and the accuracy.  
    \item Weight each difference by the proportion of instances in that bin compared to the total number of instances.
    \item Sum up these weighted differences across all bins to get the final ECE value.
  \end{enumerate}
  The formula for ECE is given by: 
  \[ \text{ECE} = \sum_{i=1}^{B} |(\text{Accuracy}_i - \text{Confidence}_i)| \times \text{Weight}_i \]
  where: 
  \begin{itemize}
    \item $\text{Accuracy}_i$ is the accuracy within the $i$-th bin.
    \item $\text{Confidence}_i$ is the average confidence within the $i$-th bin.
    \item $\text{Weight}_i$ is the proportion of instances in the $i$-th bin compared to the total number of instances.
    \item $B$ is the total number of bins.
  \end{itemize}
The final ECE value represents how much the model's estimated probabilities differ from the true (observed) probabilities. A low ECE indicates well-calibrated probabilistic predictions, demonstrating that the model's confidence scores align closely with the actual likelihood of events. RS-NN exhibits the lowest ECE as shown in Tab. \ref{tab:ood-table}.

\subsection{Algorithm for AUROC, AUPRC} \label{app:algo-auroc}

\vspace{-20pt}
\begin{algorithm}[!ht]
  \caption{Algorithm for AUROC, AUPRC}
  \label{alg:auroc-auprc}
  \begin{algorithmic}[1]
    \Require
      $\textit{uncertainty\_iid}$: Uncertainty scores for in-distribution samples.
      $\textit{uncertainty\_ood}$: Uncertainty scores for out-of-distribution samples.
    \Ensure
      $(\text{fpr, tpr, thresholds})$: ROC curve metrics. \\
      $(\text{precision, recall, prc\_thresholds})$: Precision-Recall curve metrics. \\
      $\text{auroc}$: Area Under the Receiver Operating Characteristic curve. \\
      $\text{auprc}$: Area Under the Precision-Recall curve.

    \State \textbf{Step 1: Concatenate uncertainties; } \\ $\text{uncertainties} \gets \text{concatenate}(\text{uncertainty\_iid}, \text{uncertainty\_ood})$

    \State \textbf{Step 2: Create and combine labels; } \\$\text{in\_labels} \gets \text{zeros}(\text{uncertainty\_iid.shape}[0])$
    $\text{ood\_labels} \gets \text{ones}(\text{uncertainty\_ood.shape}[0])$
    $\text{labels} \gets \text{concatenate}(\text{in\_labels}, \text{ood\_labels})$

    \State \textbf{Step 4: Calculate ROC curve; } \\ $(\text{fpr, tpr, thresholds}) \gets \text{roc\_curve}(\text{labels}, \text{uncertainties})$

    \State \textbf{Step 5: Calculate AUROC; } \\ $\text{auroc} \gets \text{roc\_auc\_score}(\text{labels}, \text{uncertainties})$

    \State \textbf{Step 6: Calculate Precision-Recall curve; }\\  $(\text{precision, recall, prc\_thresholds}) \gets \text{precision\_recall\_curve}(\text{labels}, \text{uncertainties})$

    \State \textbf{Step 7: Calculate AUPRC; } \\ $\text{auprc} \gets \text{average\_precision\_score}(\text{labels}, \text{uncertainties})$
  \end{algorithmic}
\end{algorithm}

\vspace{20pt}
Out-of-distribution (OoD) detection involves the identification of data points that deviate from the in-distribution (iD) data on which a model was trained. This process relies on assessing the model's uncertainty, particularly its epistemic uncertainty, which reflects the model's lack of knowledge or confidence in making predictions. When exposed to OoD data, which differs significantly from the training data, the model tends to exhibit higher epistemic uncertainty.

AUROC (Area Under the Receiver Operating Characteristic Curve) and AUPRC (Area Under the Precision-Recall Curve) are evaluation metrics commonly used to assess the performance of binary classification models, providing insights into their ability to distinguish between positive and negative instances. In the context of evaluating uncertainty estimation in machine learning models, these metrics quantify how well the model separates iD samples from OoD samples. AUROC is derived from the Receiver Operating Characteristic (ROC) curve, which illustrates the trade-off between True Positive Rate (TPR) and False Positive Rate (FPR) across various classification thresholds. 

The AUROC value represents the area under this curve and ranges from 0 to 1, with higher values indicating better discriminative performance. AUPRC is based on the Precision-Recall curve, which plots Precision against Recall at different classification thresholds. Precision measures the accuracy of positive predictions, while Recall quantifies the ability to capture all positive instances. AUPRC calculates the area under this curve and provides a complementary perspective, particularly valuable when dealing with imbalanced datasets.

\begin{figure}[!ht]
    \centering
    \includegraphics[width=\textwidth]{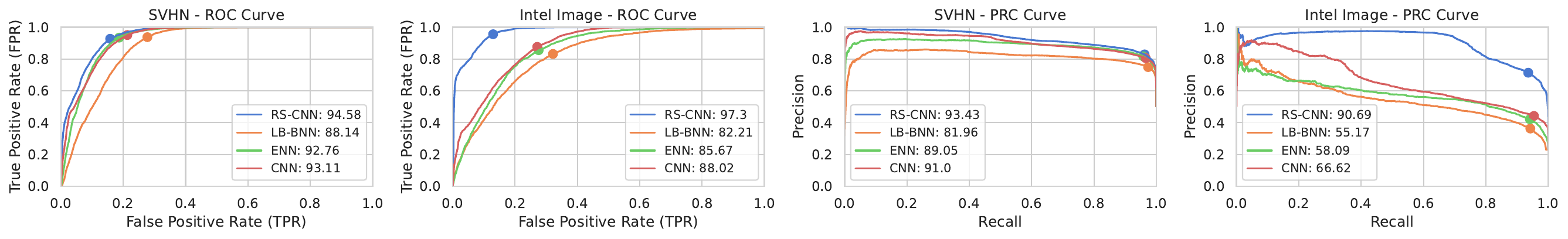}
    \caption{Receiver Operating Characteristic (ROC) and Precision-Recall Characteristic (PRC) curves for RS-NN, LB-BNN, ENN, and CNN evaluated on the SVHN and Intel Image OoD datasets for CIFAR-10.}
    \label{fig:roc-prc-curve-cifar10}
    \includegraphics[width=\textwidth]{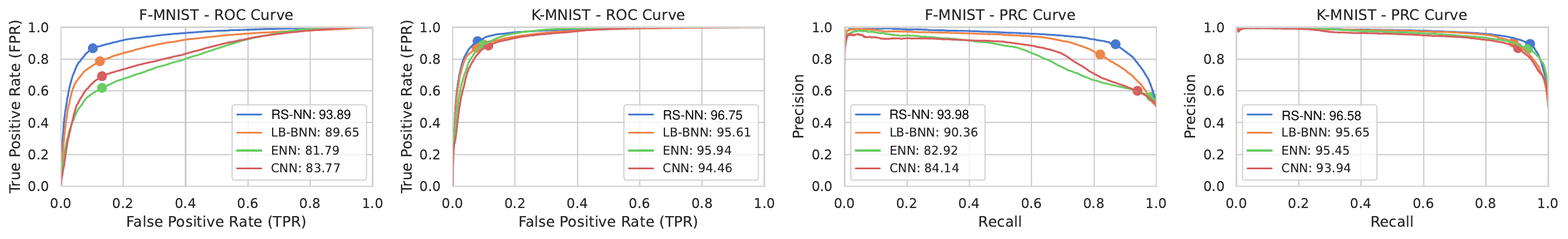}
    \caption{Receiver Operating Characteristic (ROC) and Precision-Recall Characteristic (PRC) curves for RS-NN, LB-BNN, ENN, and CNN evaluated on the SVHN and Intel Image OoD datasets for MNIST.}
    \label{fig:roc-prc-curve-mnist}
\end{figure}

Figs. \ref{fig:roc-prc-curve-cifar10} and \ref{fig:roc-prc-curve-mnist} plot each model's performance, illustrating the trade-offs between True Positive Rate and False Positive Rate (ROC curve), Precision and Recall (PRC curve) for CIFAR-10 (Fig. \ref{fig:roc-prc-curve-cifar10}) and MNIST (Fig. \ref{fig:roc-prc-curve-mnist}). The left two plots depict the AUROC curves, where the blue curve representing RS-NN outperforms others, indicating superior discrimination between true positive and false positive rates. Similarly, on the two right plots displaying Precision-Recall Curve (PRC), RS-NN exhibits the highest curve, emphasizing its precision and recall performance. These results showcase RS-NN's effectiveness in distinguishing in-distribution and out-of-distribution samples.

In real-world scenarios, encountering OoD instances is inevitable, making reliable OoD detection essential for safety-critical applications to avoid erroneous decisions on unfamiliar data. Recognising unfamiliar data signals the model about situations beyond its training, allowing it to acknowledge its own limitations and ignorance, which in turn enhances its uncertainty estimation.

\chapter{Statistical guarantees}
\label{app:statistical}
To complement the extensive discussion in the main paper, we provide here
an in-depth discussion on how statistical guarantees can be provided in an RS-NN framework.


\textbf{Plugging RS-NN into conformal learning.}
Indeed, RS-NN can be employed as the `underlying model' in an inductive conformal learning framework\footnote{\url{https://cml.rhul.ac.uk/copa2017/presentations/CP_Tutorial_2017.pdf}}, which builds an empirical cumulative distribution of the `non-conformity' scores of a set of calibration samples, and at test time outputs the set of labels whose empirical CDF is above a desired significance level $\epsilon$ (e.g., 95\%). 
\\
Given a test input $x$ and the associated predictive belief function $\hat{Bel}(c|x)$ (the output of RS-NN), we could, for instance, set as non-conformity score
\begin{equation} \label{eq:non_conformity}
s(x,c) \doteq 1 - \hat{Pl}(c|x) = \hat{Bel}(\mathcal{C} \setminus \{c\} | x)
\end{equation}

(i.e., a label $c$ is 
`non-conformal'
if its predicted \emph{plausibility}, which is defined as $Pl(A) = 1 - Bel(\Theta \setminus A)$ and has the semantic of an upper probability bound \citep{shafer1976mathematical}, is low), and compute predictive regions in the usual way:
\[
\Gamma (x) = \{ c \in \mathcal{C} : p^c > \epsilon \},
\]
where 
\[
\begin{array}{lll}
p^c & = & \displaystyle \frac{|(x_j,c_j) : s(x_j,c_j) > s(x,c)|}{q+1} +
u \cdot \frac{|(x_j,c_j) : s(x_j,c_j) = s(x,c)|}{q+1},
\end{array}
\] 
$(x_j,c_j)$ is the $j$-th calibration point, $q$ is the number of calibration points, and
$u \sim \mathcal{U}(0,1)$ (the uniform distribution on the interval $(0,1)$).

\textbf{Experiments on conformal guarantees.} Thus, we conducted experiments on CIFAR-10 dataset to provide conformal prediction guarantees with non-conformity scores as given in Eq. \ref{eq:non_conformity}. The goal is to assess the reliability of classification predictions by determining a threshold of non-conformity scores that covers 95\% of the data.

To achieve this, non-conformity scores are obtained for all classes, and a threshold is calculated such that it contains 95\% of the data. This threshold determines the coverage of the classification, which refers to the proportion of predictions that actually contain the true outcome.

This prediction threshold is determined by finding the percentile corresponding to $1 - \alpha$, where $\alpha$ is set to $0.05$ to achieve 95\% coverage. Fig. \ref{fig:CP_threshold} 
shows the non-conformity scores for all the calibration data, 
$j = 1000$ \citep{angelopoulos2021gentle}
{(using the following split: 40000:10000:9000:1000 training, testing, validation and calibration data points respectively for CIFAR-10)}, with a cut-off threshold that ensures 95\% coverage. 

The conformal prediction sets for two sample inputs are shown in Fig. \ref{fig:CPsets}, along with their predicted probabilities. 

\begin{figure}[!ht]
    \centering
    \includegraphics[width = 1\linewidth]{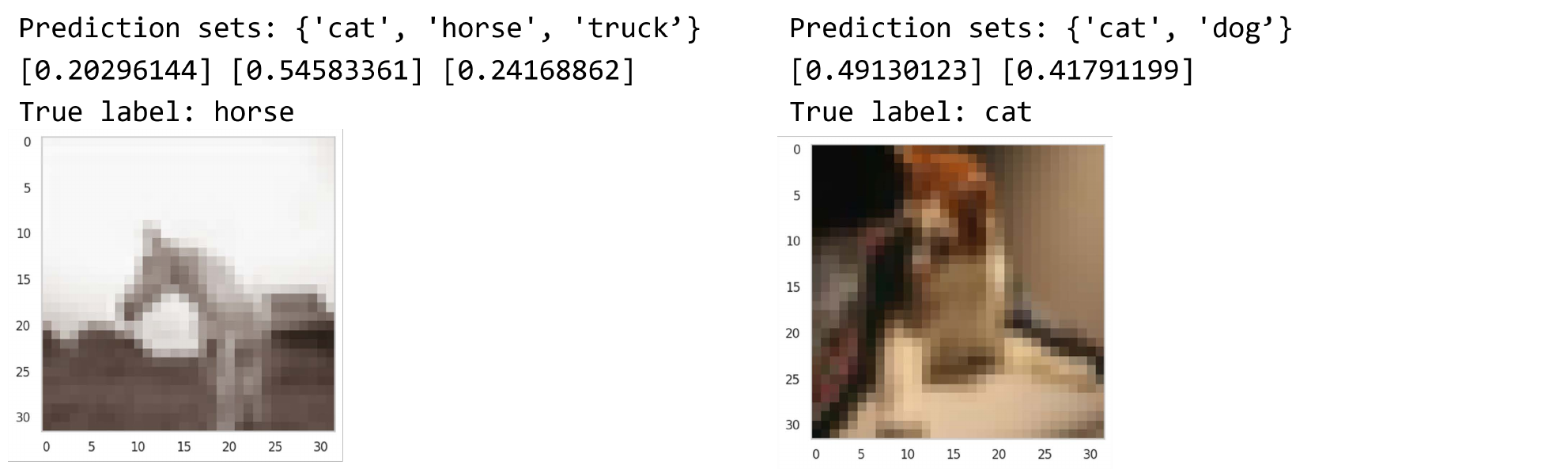} 
    \caption{Conformal prediction sets by RS-NN on the CIFAR-10 dataset. The predicted probabilities for each class is shown. 
    } 
    \label{fig:CPsets}
    \vspace{20pt}
\end{figure}

\begin{figure}[!ht]
\centering
\begin{minipage}[b]{0.7\textwidth}
    \centering
    \vspace{-15pt}
    \includegraphics[width = 0.85\linewidth]{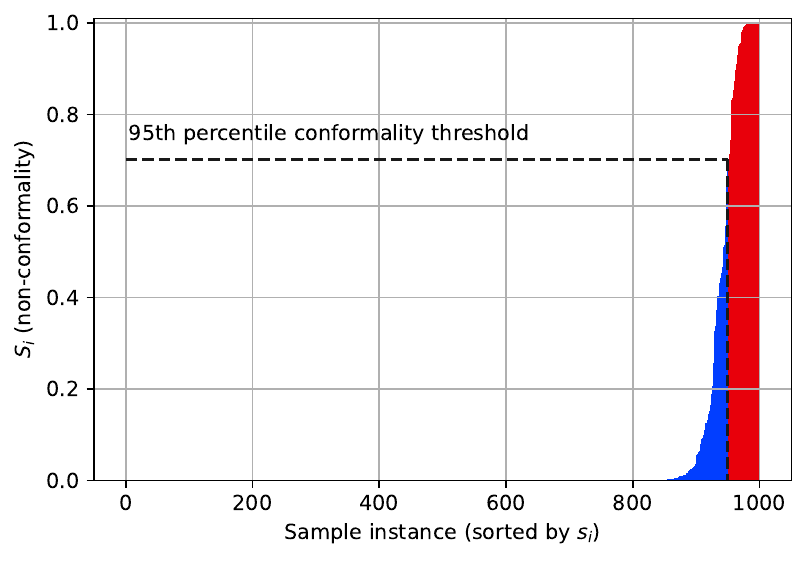}
    \caption{Conformal prediction threshold for CIFAR-10 indicating the boundary beyond which predictions are considered non-conformal.}
    \label{fig:CP_threshold}
\end{minipage}
\begin{minipage}[b]{0.47\textwidth}
    \centering
    \vspace{20pt}
    \includegraphics[width =1\linewidth]{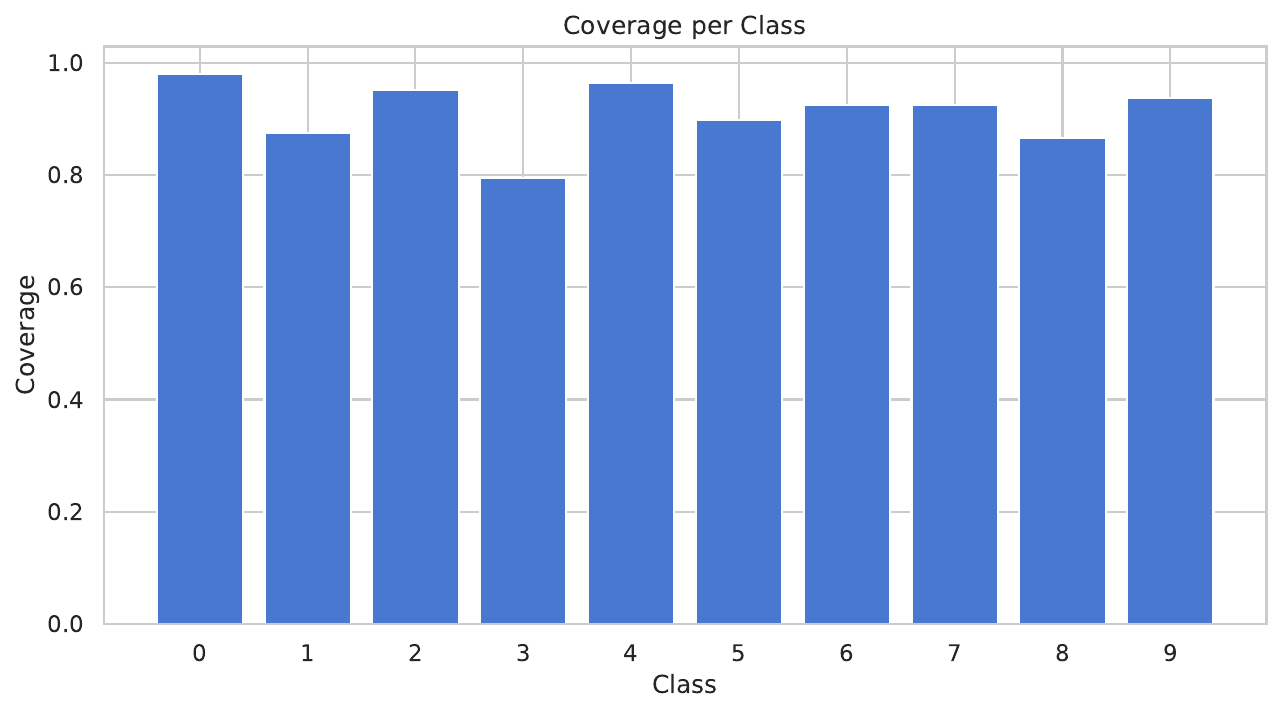} 
            \vspace{-10pt}
    \caption{Coverage per class of RS-NN for CIFAR-10 dataset.}
    \label{fig:CP_coverage}
\end{minipage} \hspace{0.6cm}
\begin{minipage}[b]{0.47\textwidth}
    \centering
        \vspace{20pt}
    \includegraphics[width = 1\linewidth]{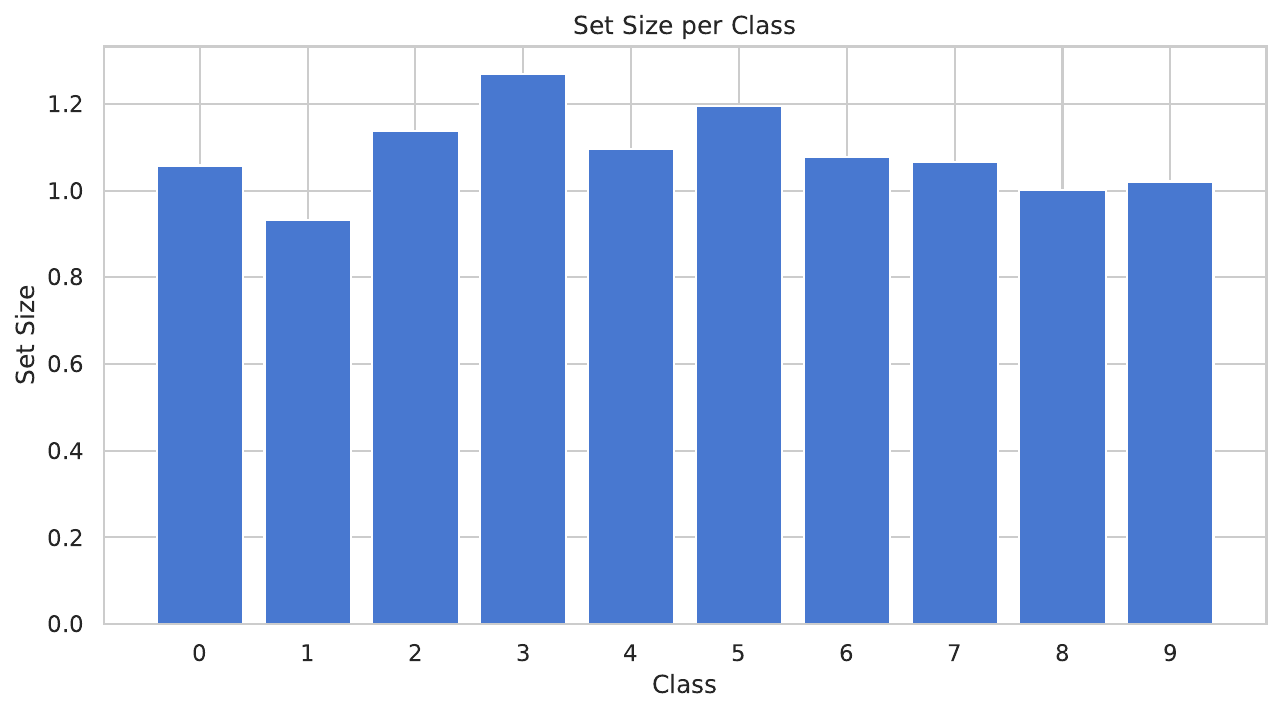}
            \vspace{-10pt}
    \caption{Average size of the predictive sets generated for each class of CIFAR-10.}
    \label{fig:CP_set_size}
\end{minipage}
\end{figure}

Figs. \ref{fig:CP_coverage} and \ref{fig:CP_set_size} show the coverage and average set size for each class on the test dataset. 
Coverage is the proportion of prediction sets that actually contain the true outcome, and average set size is the average number of predicted classes per instance. Higher numbers represent more overlap between the classification regions of different classes.

{Summarizing the results, (1) the coverage averages across classes over the desired threshold. 
This indicates that the conformal prediction method is providing reliable prediction intervals that contain the true class label with the specified confidence level; (2) The spread or variability in set sizes across different samples provides insights into how well the prediction sets adapt to the difficulty of the samples.  Ideally, the prediction sets should dynamically adjust based on the difficulty of each example, with larger sets for more challenging inputs and smaller sets for easier ones. 
Fig. \ref{fig:CP_set_size} shows that the average set size for classes 1, 2, 3, 4, and 6 is considerably larger than for the rest, indicating that the model found samples from these as challenging inputs. 
}



\textbf{Alternative approaches to statistical guarantees.}
In the future we plan to explore the possibility of generalising the empirical CDF at the basis of conformal learning in a belief function / random set representation, aiming to retain statistical guarantees. Given the complexity of this enterprise, we are working towards this in a separate paper.

Finally, an intriguing alternative approach (and one that is more achievable in the short term) is the study of confidence intervals in a belief functions representation, such as the one we employ in RS-NNs. In fact, recent studies have been looking at extending the notion of confidence interval (\url{https://en.wikipedia.org/wiki/Confidence_interval}) to belief functions, under the name of \emph{confidence structures} (\url{https://hal.science/hal-01576356v3}), which generalise standard confidence distributions and generate ``frequency-calibrated" belief functions.

Liu and Martin, in particular, have developed an Inferential Model (IM) approach which produces 
belief functions with well-defined frequentist properties \citep{martin2013inferential, martin2015inferential}.  
An alternative approach relies on the notion of ``predictive" belief function \citep{denoeux2006constructing}, which, under repeated sampling, is less committed than the true probability distribution of interest with some prescribed probability.



\bibliography{biblio/deduplicated_file}

\begin{thebibliography}{353}
\providecommand{\natexlab}[1]{#1}
\providecommand{\url}[1]{\texttt{#1}}
\expandafter\ifx\csname urlstyle\endcsname\relax
  \providecommand{\doi}[1]{doi: #1}\else
  \providecommand{\doi}{doi: \begingroup \urlstyle{rm}\Url}\fi

\bibitem[Abdi and Williams(2010)]{abdi2010principal}
Herv{\'e} Abdi and Lynne~J Williams.
\newblock Principal component analysis.
\newblock \emph{Wiley interdisciplinary reviews: computational statistics}, 2\penalty0 (4):\penalty0 433--459, 2010.

\bibitem[Abe et~al.(2022)Abe, Buchanan, Pleiss, Zemel, and Cunningham]{abe2022deep}
Taiga Abe, Estefany~Kelly Buchanan, Geoff Pleiss, Richard Zemel, and John~P Cunningham.
\newblock Deep ensembles work, but are they necessary?
\newblock \emph{Advances in Neural Information Processing Systems}, 35:\penalty0 33646--33660, 2022.

\bibitem[Abell{\'a}n and Moral(2000)]{abellan2000non}
Joaqu{\'\i}n Abell{\'a}n and Seraf{\'\i}n Moral.
\newblock A non-specificity measure for convex sets of probability distributions.
\newblock \emph{International journal of uncertainty, fuzziness and knowledge-based systems}, 8\penalty0 (03):\penalty0 357--367, 2000.

\bibitem[Abell{\'a}n and Moral(2005)]{abellan2005difference}
Joaqu{\'\i}n Abell{\'a}n and Seraf{\'\i}n Moral.
\newblock Difference of entropies as a non-specificity function on credal sets.
\newblock \emph{International journal of general systems}, 34\penalty0 (3):\penalty0 201--214, 2005.

\bibitem[Abell\'an and G\'omez(2006)]{abellan06measures}
Joaquín Abell\'an and Manuel G\'omez.
\newblock Measures of divergence on credal sets.
\newblock \emph{Fuzzy Sets and Systems}, 157\penalty0 (11):\penalty0 1514--1531, 2006.

\bibitem[Achiam et~al.(2023)Achiam, Adler, Agarwal, Ahmad, Akkaya, Aleman, Almeida, Altenschmidt, Altman, Anadkat, et~al.]{achiam2023gpt}
Josh Achiam, Steven Adler, Sandhini Agarwal, Lama Ahmad, Ilge Akkaya, Florencia~Leoni Aleman, Diogo Almeida, Janko Altenschmidt, Sam Altman, Shyamal Anadkat, et~al.
\newblock Gpt-4 technical report.
\newblock \emph{arXiv preprint arXiv:2303.08774}, 2023.

\bibitem[Ackermann et~al.(2014)Ackermann, Bl{\"o}mer, Kuntze, and Sohler]{ackermann2014analysis}
Marcel~R Ackermann, Johannes Bl{\"o}mer, Daniel Kuntze, and Christian Sohler.
\newblock Analysis of agglomerative clustering.
\newblock \emph{Algorithmica}, 69:\penalty0 184--215, 2014.

\bibitem[Aditya et~al.(2019)Aditya, Yang, and Baral]{aditya2019integrating}
Somak Aditya, Yezhou Yang, and Chitta Baral.
\newblock Integrating knowledge and reasoning in image understanding.
\newblock \emph{arXiv preprint arXiv:1906.09954}, 2019.

\bibitem[Aggarwal et~al.(2014)Aggarwal, Kong, Gu, Han, and Philip]{aggarwal2014active}
Charu~C Aggarwal, Xiangnan Kong, Quanquan Gu, Jiawei Han, and S~Yu Philip.
\newblock Active learning: A survey.
\newblock In \emph{Data Classification: Algorithms and Applications}, pages 571--605. CRC Press, 2014.

\bibitem[Ahmed et~al.(2022)Ahmed, Teso, Chang, Van~den Broeck, and Vergari]{ahmed2022semantic}
Kareem Ahmed, Stefano Teso, Kai-Wei Chang, Guy Van~den Broeck, and Antonio Vergari.
\newblock Semantic probabilistic layers for neuro-symbolic learning.
\newblock \emph{Advances in Neural Information Processing Systems}, 35:\penalty0 29944--29959, 2022.

\bibitem[Albuquerque et~al.(2019)Albuquerque, Monteiro, Darvishi, Falk, and Mitliagkas]{albuquerque2019generalizing}
Isabela Albuquerque, Jo{\~a}o Monteiro, Mohammad Darvishi, Tiago~H Falk, and Ioannis Mitliagkas.
\newblock Generalizing to unseen domains via distribution matching.
\newblock \emph{arXiv preprint arXiv:1911.00804}, 2019.

\bibitem[Alemi et~al.(2018)Alemi, Fischer, and Dillon]{alemi2018uncertainty}
Alexander~A Alemi, Ian Fischer, and Joshua~V Dillon.
\newblock Uncertainty in the variational information bottleneck.
\newblock \emph{arXiv preprint arXiv:1807.00906}, 2018.

\bibitem[Alijani et~al.(2024)Alijani, Fayyad, and Najjaran]{alijani2024vision}
Shadi Alijani, Jamil Fayyad, and Homayoun Najjaran.
\newblock Vision transformers in domain adaptation and domain generalization: a study of robustness.
\newblock \emph{Neural Computing and Applications}, 36\penalty0 (29):\penalty0 17979--18007, 2024.

\bibitem[Amodei et~al.(2016)Amodei, Olah, Steinhardt, Christiano, Schulman, and Man{\'e}]{amodei2016concrete}
Dario Amodei, Chris Olah, Jacob Steinhardt, Paul Christiano, John Schulman, and Dan Man{\'e}.
\newblock Concrete problems in ai safety.
\newblock \emph{arXiv preprint arXiv:1606.06565}, 2016.

\bibitem[Angelopoulos and Bates(2021)]{angelopoulos2021gentle}
Anastasios~N Angelopoulos and Stephen Bates.
\newblock A gentle introduction to conformal prediction and distribution-free uncertainty quantification.
\newblock \emph{arXiv preprint arXiv:2107.07511}, 2021.

\bibitem[Anil et~al.(2023)Anil, Dai, Firat, Johnson, Lepikhin, Passos, Shakeri, Taropa, Bailey, Chen, et~al.]{anil2023palm}
Rohan Anil, Andrew~M Dai, Orhan Firat, Melvin Johnson, Dmitry Lepikhin, Alexandre Passos, Siamak Shakeri, Emanuel Taropa, Paige Bailey, Zhifeng Chen, et~al.
\newblock Palm 2 technical report.
\newblock \emph{arXiv preprint arXiv:2305.10403}, 2023.

\bibitem[Antonucci and Corani(2017)]{ANTONUCCI2017320}
Alessandro Antonucci and Giorgio Corani.
\newblock The multilabel {Naive Credal Classifier}.
\newblock \emph{International Journal of Approximate Reasoning}, 83:\penalty0 320--336, 2017.
\newblock ISSN 0888-613X.
\newblock \doi{https://doi.org/10.1016/j.ijar.2016.10.006}.
\newblock URL \url{https://www.sciencedirect.com/science/article/pii/S0888613X16301773}.

\bibitem[Antonucci and Cuzzolin(2010)]{antonucci2010credal}
Alessandro Antonucci and Fabio Cuzzolin.
\newblock Credal sets approximation by lower probabilities: application to credal networks.
\newblock In \emph{Computational Intelligence for Knowledge-Based Systems Design: 13th International Conference on Information Processing and Management of Uncertainty, IPMU 2010, Dortmund, Germany, June 28-July 2, 2010. Proceedings 13}, pages 716--725. Springer, 2010.

\bibitem[Augustin(2022)]{augustin2022statistics}
Thomas Augustin.
\newblock Statistics with imprecise probabilities—a short survey.
\newblock \emph{Uncertainty in Engineering Introduction to Methods and Applications}, 67, 2022.

\bibitem[Augustin et~al.(2014)Augustin, Coolen, De~Cooman, and Troffaes]{augustin2014introduction}
Thomas Augustin, Frank~PA Coolen, Gert De~Cooman, and Matthias~CM Troffaes.
\newblock \emph{Introduction to imprecise probabilities}, volume 591.
\newblock John Wiley \& Sons, 2014.

\bibitem[Awais et~al.(2021)Awais, Zhou, Xu, Hong, Luo, Bae, and Li]{awais2021adversarial}
Muhammad Awais, Fengwei Zhou, Hang Xu, Lanqing Hong, Ping Luo, Sung-Ho Bae, and Zhenguo Li.
\newblock Adversarial robustness for unsupervised domain adaptation.
\newblock In \emph{Proceedings of the IEEE/CVF International Conference on Computer Vision}, pages 8568--8577, 2021.

\bibitem[Azar et~al.(2017)Azar, Osband, and Munos]{azar2017minimax}
Mohammad~Gheshlaghi Azar, Ian Osband, and R{\'e}mi Munos.
\newblock Minimax regret bounds for reinforcement learning.
\newblock In \emph{International conference on machine learning}, pages 263--272. PMLR, 2017.

\bibitem[Balasubramanian et~al.(2014)Balasubramanian, Ho, and Vovk]{balasubramanian2014conformal}
Vineeth Balasubramanian, Shen-Shyang Ho, and Vladimir Vovk.
\newblock \emph{Conformal prediction for reliable machine learning: theory, adaptations and applications}.
\newblock Newnes, 2014.

\bibitem[Bansal(2019)]{intelimage}
Puneet Bansal.
\newblock Intel image classification.
\newblock \emph{Available on https://www. kaggle. com/puneet6060/intel-image-classification, Online}, 2019.

\bibitem[Barbara and Chen(2001)]{barbara2001tracking}
Daniel Barbara and Ping Chen.
\newblock Tracking clusters in evolving data sets.
\newblock In \emph{FLAIRS}, pages 239--243, 2001.

\bibitem[Baron(1987)]{baron1987second}
Jonathan Baron.
\newblock Second-order probabilities and belief functions.
\newblock \emph{Theory and Decision}, 23:\penalty0 25--36, 1987.

\bibitem[Bates et~al.(2023)Bates, Cand{\`e}s, Lei, Romano, and Sesia]{bates2023testing}
Stephen Bates, Emmanuel Cand{\`e}s, Lihua Lei, Yaniv Romano, and Matteo Sesia.
\newblock Testing for outliers with conformal p-values.
\newblock \emph{The Annals of Statistics}, 51\penalty0 (1):\penalty0 149--178, 2023.

\bibitem[Beluch et~al.(2018)Beluch, Genewein, N{\"u}rnberger, and K{\"o}hler]{beluch2018power}
William~H Beluch, Tim Genewein, Andreas N{\"u}rnberger, and Jan~M K{\"o}hler.
\newblock The power of ensembles for active learning in image classification.
\newblock In \emph{Proceedings of the IEEE conference on computer vision and pattern recognition}, pages 9368--9377, 2018.

\bibitem[Bender et~al.(2021)Bender, Gebru, McMillan-Major, and Shmitchell]{bender2021dangers}
Emily~M Bender, Timnit Gebru, Angelina McMillan-Major, and Shmargaret Shmitchell.
\newblock On the dangers of stochastic parrots: Can language models be too big?
\newblock In \emph{Proceedings of the 2021 ACM conference on fairness, accountability, and transparency}, pages 610--623, 2021.

\bibitem[Bengs et~al.(2022)Bengs, H{\"u}llermeier, and Waegeman]{bengs2022pitfalls}
Viktor Bengs, Eyke H{\"u}llermeier, and Willem Waegeman.
\newblock Pitfalls of epistemic uncertainty quantification through loss minimisation.
\newblock \emph{Advances in Neural Information Processing Systems}, 35:\penalty0 29205--29216, 2022.

\bibitem[Berger(2013)]{berger2013statistical}
James~O Berger.
\newblock \emph{Statistical decision theory and Bayesian analysis}.
\newblock Springer Science \& Business Media, 2013.

\bibitem[Bernard(2005)]{bernard2005introduction}
Jean-Marc Bernard.
\newblock An introduction to the imprecise dirichlet model for multinomial data.
\newblock \emph{International Journal of Approximate Reasoning}, 39\penalty0 (2-3):\penalty0 123--150, 2005.

\bibitem[Betancourt and Muhanna(2022)]{betancourt2022interval}
David Betancourt and Rafi~L Muhanna.
\newblock Interval deep learning for computational mechanics problems under input uncertainty.
\newblock \emph{Probabilistic Engineering Mechanics}, 70:\penalty0 103370, 2022.

\bibitem[Blanchard et~al.(2011)Blanchard, Lee, and Scott]{blanchard2011generalizing}
Gilles Blanchard, Gyemin Lee, and Clayton Scott.
\newblock Generalizing from several related classification tasks to a new unlabeled sample.
\newblock \emph{Advances in neural information processing systems}, 24, 2011.

\bibitem[Blundell et~al.(2015)Blundell, Cornebise, Kavukcuoglu, and Wierstra]{blundell2015weight}
Charles Blundell, Julien Cornebise, Koray Kavukcuoglu, and Daan Wierstra.
\newblock Weight uncertainty in neural network.
\newblock In \emph{International conference on machine learning}, pages 1613--1622. PMLR, 2015.

\bibitem[Bojarski(2016)]{bojarski2016end}
Mariusz Bojarski.
\newblock End to end learning for self-driving cars.
\newblock \emph{arXiv preprint arXiv:1604.07316}, 2016.

\bibitem[Bronevich and Klir(2008)]{bronevich2008axioms}
Andrey Bronevich and George~J Klir.
\newblock Axioms for uncertainty measures on belief functions and credal sets.
\newblock In \emph{NAFIPS 2008-2008 Annual Meeting of the North American Fuzzy Information Processing Society}, pages 1--6. IEEE, 2008.

\bibitem[Brown et~al.(2020)Brown, Mann, Ryder, Subbiah, Kaplan, Dhariwal, Neelakantan, Shyam, Sastry, Askell, et~al.]{brown2020language}
Tom Brown, Benjamin Mann, Nick Ryder, Melanie Subbiah, Jared~D Kaplan, Prafulla Dhariwal, Arvind Neelakantan, Pranav Shyam, Girish Sastry, Amanda Askell, et~al.
\newblock Language models are few-shot learners.
\newblock \emph{Advances in neural information processing systems}, 33:\penalty0 1877--1901, 2020.

\bibitem[Buntine and Weigend(1991)]{Buntine1991BayesianB}
Wray~L. Buntine and Andreas~S. Weigend.
\newblock Bayesian back-propagation.
\newblock \emph{Complex Syst.}, 5, 1991.
\newblock URL \url{https://api.semanticscholar.org/CorpusID:14814125}.

\bibitem[Burger et~al.(2006)Burger, Aran, and Caplier]{burger}
Thomas Burger, Oya Aran, and Alice Caplier.
\newblock Modeling hesitation and conflict: a belief-based approach for multi-class problems.
\newblock In \emph{2006 5th International Conference on Machine Learning and Applications (ICMLA'06)}, pages 95--100. IEEE, 2006.

\bibitem[Campos et~al.(2024)Campos, Farinhas, Zerva, Figueiredo, and Martins]{campos2024conformal}
Margarida Campos, Ant{\'o}nio Farinhas, Chrysoula Zerva, M{\'a}rio~AT Figueiredo, and Andr{\'e}~FT Martins.
\newblock Conformal prediction for natural language processing: A survey.
\newblock \emph{Transactions of the Association for Computational Linguistics}, 12:\penalty0 1497--1516, 2024.

\bibitem[Cao et~al.(2024)Cao, Wang, Wang, Xu, and Wang]{cao2024interval}
Yang Cao, Xiaojun Wang, Yi~Wang, Lianming Xu, and Yifei Wang.
\newblock An interval neural network method for identifying static concentrated loads in a population of structures.
\newblock \emph{Aerospace}, 11\penalty0 (9):\penalty0 770, 2024.

\bibitem[Caprio et~al.(2023{\natexlab{a}})Caprio, Dutta, Jang, Lin, Ivanov, Sokolsky, and Lee]{caprio2023credal}
Michele Caprio, Souradeep Dutta, Kuk Jang, Vivian Lin, Radoslav Ivanov, Oleg Sokolsky, and Insup Lee.
\newblock Credal bayesian deep learning.
\newblock \emph{arXiv preprint arXiv:2302.09656}, 2023{\natexlab{a}}.

\bibitem[Caprio et~al.(2023{\natexlab{b}})Caprio, Dutta, Jang, Lin, Ivanov, Sokolsky, and Lee]{caprio2023imprecise}
Michele Caprio, Souradeep Dutta, Kuk~Jin Jang, Vivian Lin, Radoslav Ivanov, Oleg Sokolsky, and Insup Lee.
\newblock Imprecise {Bayesian} neural networks.
\newblock \emph{arXiv preprint arXiv:2302.09656}, 2023{\natexlab{b}}.

\bibitem[Caprio et~al.(2024)Caprio, Sultana, Elia, and Cuzzolin]{caprio2024credal}
Michele Caprio, Maryam Sultana, Eleni Elia, and Fabio Cuzzolin.
\newblock Credal learning theory.
\newblock \emph{arXiv preprint arXiv:2402.00957}, 2024.

\bibitem[Charpentier et~al.(2020)Charpentier, Z{\"u}gner, and G{\"u}nnemann]{charpentier2020posterior}
Bertrand Charpentier, Daniel Z{\"u}gner, and Stephan G{\"u}nnemann.
\newblock Posterior network: Uncertainty estimation without ood samples via density-based pseudo-counts.
\newblock \emph{Advances in neural information processing systems}, 33:\penalty0 1356--1367, 2020.

\bibitem[Chateauneuf and Jaffray(1989)]{Chateauneuf89}
Alain Chateauneuf and Jean-Yves Jaffray.
\newblock Some characterizations of lower probabilities and other monotone capacities through the use of {Möbius inversion}.
\newblock \emph{Mathematical Social Sciences}, 17\penalty0 (3):\penalty0 263--283, 1989.
\newblock ISSN 0165-4896.
\newblock \doi{https://doi.org/10.1016/0165-4896(89)90056-5}.
\newblock URL \url{https://www.sciencedirect.com/science/article/pii/0165489689900565}.

\bibitem[Chaudhary et~al.(2022)Chaudhary, Leit{\~a}o, Donauer, D’Aronco, Perraudin, Obozinski, Perez-Cruz, Schindler, Wegner, and Russo]{chaudhary2022flood}
Priyanka Chaudhary, Jo{\~a}o~P Leit{\~a}o, Tabea Donauer, Stefano D’Aronco, Nathana{\"e}l Perraudin, Guillaume Obozinski, Fernando Perez-Cruz, Konrad Schindler, Jan~Dirk Wegner, and Stefania Russo.
\newblock Flood uncertainty estimation using deep ensembles.
\newblock \emph{Water}, 14\penalty0 (19):\penalty0 2980, 2022.

\bibitem[Chen et~al.(2016)Chen, Carlson, Gan, Li, and Carin]{chen2016bridging}
Changyou Chen, David Carlson, Zhe Gan, Chunyuan Li, and Lawrence Carin.
\newblock Bridging the gap between stochastic gradient mcmc and stochastic optimization.
\newblock In \emph{Artificial Intelligence and Statistics}, pages 1051--1060. PMLR, 2016.

\bibitem[Chen et~al.(2021)Chen, Itkina, Senanayake, and Kochenderfer]{chen2021evidential}
Phil Chen, Mikhal Itkina, Ransalu Senanayake, and Mykel~J Kochenderfer.
\newblock Evidential softmax for sparse multimodal distributions in deep {Generative} models.
\newblock \emph{Advances in Neural Information Processing Systems}, 34:\penalty0 11565--11576, 2021.

\bibitem[Ciosek et~al.(2019)Ciosek, Fortuin, Tomioka, Hofmann, and Turner]{ciosek2019conservative}
Kamil Ciosek, Vincent Fortuin, Ryota Tomioka, Katja Hofmann, and Richard Turner.
\newblock Conservative uncertainty estimation by fitting prior networks.
\newblock In \emph{International Conference on Learning Representations}, 2019.

\bibitem[Clanuwat et~al.(2018)Clanuwat, Bober-Irizar, Kitamoto, Lamb, Yamamoto, and Ha]{clanuwat2018deep}
Tarin Clanuwat, Mikel Bober-Irizar, Asanobu Kitamoto, Alex Lamb, Kazuaki Yamamoto, and David Ha.
\newblock Deep learning for classical {Japanese} literature.
\newblock \emph{arXiv preprint arXiv:1812.01718}, 2018.

\bibitem[Corani and Zaffalon(2008)]{corani2008learning}
Giorgio Corani and Marco Zaffalon.
\newblock Learning reliable classifiers from small or incomplete data sets: The naive credal classifier 2.
\newblock \emph{Journal of Machine Learning Research}, 9\penalty0 (4), 2008.

\bibitem[Corani et~al.(2012)Corani, Antonucci, and Zaffalon]{corani2012bayesian}
Giorgio Corani, Alessandro Antonucci, and Marco Zaffalon.
\newblock {Bayesian} networks with imprecise probabilities: Theory and application to classification.
\newblock \emph{Data Mining: Foundations and Intelligent Paradigms: Volume 1: Clustering, Association and Classification}, pages 49--93, 2012.

\bibitem[Cozman(2000)]{cozman00credal}
Fabio~Gagliardi Cozman.
\newblock Credal networks.
\newblock \emph{Artificial Intelligence}, 120\penalty0 (2):\penalty0 199--233, 2000.

\bibitem[Cuzzolin(2003)]{cuzzolin2003geometry}
Fabio Cuzzolin.
\newblock Geometry of {Upper Probabilities}.
\newblock In \emph{ISIPTA}, pages 188--203, 2003.

\bibitem[Cuzzolin(2004{\natexlab{a}})]{cuzzolin04ipmu}
Fabio Cuzzolin.
\newblock Simplicial complexes of finite fuzzy sets.
\newblock In \emph{Proceedings of the 10th International Conference on Information Processing and Management of Uncertainty (IPMU'04)}, volume~4, pages 4--9, 2004{\natexlab{a}}.

\bibitem[Cuzzolin(2004{\natexlab{b}})]{cuzzolin04smcb}
Fabio Cuzzolin.
\newblock Geometry of {D}empster's rule of combination.
\newblock \emph{IEEE Transactions on Systems, Man and Cybernetics part B}, 34\penalty0 (2):\penalty0 961--977, 2004{\natexlab{b}}.

\bibitem[Cuzzolin(2007)]{cuzzolin07smcb}
Fabio Cuzzolin.
\newblock Two new {B}ayesian approximations of belief functions based on convex geometry.
\newblock \emph{IEEE Transactions on Systems, Man, and Cybernetics - Part B}, 37\penalty0 (4):\penalty0 993--1008, 2007.

\bibitem[Cuzzolin(2008{\natexlab{a}})]{cuzzolin2008credal}
Fabio Cuzzolin.
\newblock On the credal structure of consistent probabilities.
\newblock In \emph{European Workshop on Logics in Artificial Intelligence}, pages 126--139. Springer, 2008{\natexlab{a}}.

\bibitem[Cuzzolin(2008{\natexlab{b}})]{cuzzolin2008geometric}
Fabio Cuzzolin.
\newblock A geometric approach to the theory of evidence.
\newblock \emph{IEEE Transactions on Systems, Man, and Cybernetics, Part C (Applications and Reviews)}, 38\penalty0 (4):\penalty0 522--534, 2008{\natexlab{b}}.

\bibitem[Cuzzolin(2009)]{cuzzolin09ecsqaru}
Fabio Cuzzolin.
\newblock Complexes of outer consonant approximations.
\newblock In \emph{Proceedings of the 10th European Conference on Symbolic and Quantitative Approaches to Reasoning with Uncertainty (ECSQARU'09)}, pages 275--286, 2009.

\bibitem[Cuzzolin(2010{\natexlab{a}})]{cuzzolin10ida}
Fabio Cuzzolin.
\newblock Three alternative combinatorial formulations of the theory of evidence.
\newblock \emph{Intelligent Data Analysis}, 14\penalty0 (4):\penalty0 439--464, 2010{\natexlab{a}}.

\bibitem[Cuzzolin(2010{\natexlab{b}})]{cuzzolin2010credal}
Fabio Cuzzolin.
\newblock {Credal semantics of {{Bayesian}} transformations in terms of probability intervals}.
\newblock \emph{IEEE Transactions on Systems, Man, and Cybernetics, Part B: Cybernetics}, 40\penalty0 (2):\penalty0 421--432, 2010{\natexlab{b}}.

\bibitem[Cuzzolin(2010{\natexlab{c}})]{cuzzolin2010geometry}
Fabio Cuzzolin.
\newblock The geometry of consonant belief functions: simplicial complexes of necessity measures.
\newblock \emph{Fuzzy Sets and Systems}, 161\penalty0 (10):\penalty0 1459--1479, 2010{\natexlab{c}}.

\bibitem[Cuzzolin(2013)]{cuzzolin2013l}
Fabio Cuzzolin.
\newblock $ l\_ $\{$p$\}$ $ consonant approximations of belief functions.
\newblock \emph{IEEE Transactions on Fuzzy Systems}, 22\penalty0 (2):\penalty0 420--436, 2013.

\bibitem[Cuzzolin(2018{\natexlab{a}})]{cuzzolin18belief-maxent}
Fabio Cuzzolin.
\newblock Generalised max entropy classifiers.
\newblock In S{\'e}bastien Destercke, Thierry Den{\oe}ux, Fabio Cuzzolin, and Arnaud Martin, editors, \emph{Belief Functions: Theory and Applications}, pages 39--47, Cham, 2018{\natexlab{a}}. Springer International Publishing.

\bibitem[Cuzzolin(2018{\natexlab{b}})]{cuzzolin2018visions}
Fabio Cuzzolin.
\newblock Visions of a generalized probability theory.
\newblock \emph{arXiv preprint arXiv:1810.10341}, 2018{\natexlab{b}}.

\bibitem[Cuzzolin(2020)]{cuzzolin2020geometry}
Fabio Cuzzolin.
\newblock \emph{The Geometry of Uncertainty: The Geometry of Imprecise Probabilities}.
\newblock Artificial Intelligence: Foundations, Theory, and Algorithms. Springer International Publishing, 2020.
\newblock ISBN 9783030631536.
\newblock URL \url{https://books.google.co.uk/books?id=jNQPEAAAQBAJ}.

\bibitem[Cuzzolin(2024)]{cuzzolin2024uncertainty}
Fabio Cuzzolin.
\newblock Uncertainty measures: A critical survey.
\newblock \emph{Information Fusion}, page 102609, 2024.

\bibitem[Cuzzolin(July 2011)]{cuzzolin11isipta-consonant}
Fabio Cuzzolin.
\newblock Lp consonant approximations of belief functions in the mass space.
\newblock In \emph{Proceedings of the 7th International Symposium on Imprecise Probability: Theory and Applications (ISIPTA'11)}, July 2011.

\bibitem[Cuzzolin and Frezza(2001)]{cuzzolin2001geometric}
Fabio Cuzzolin and Ruggero Frezza.
\newblock Geometric analysis of belief space and conditional subspaces.
\newblock In \emph{ISIPTA}, pages 122--132, 2001.

\bibitem[Dall'Anese et~al.(2020)Dall'Anese, Simonetto, Becker, and Madden]{dall2020optimization}
Emiliano Dall'Anese, Andrea Simonetto, Stephen Becker, and Liam Madden.
\newblock Optimization and learning with information streams: Time-varying algorithms and applications.
\newblock \emph{IEEE Signal Processing Magazine}, 37\penalty0 (3):\penalty0 71--83, 2020.

\bibitem[Daxberger et~al.(2021)Daxberger, Kristiadi, Immer, Eschenhagen, Bauer, and Hennig]{daxberger2021laplace}
Erik Daxberger, Agustinus Kristiadi, Alexander Immer, Runa Eschenhagen, Matthias Bauer, and Philipp Hennig.
\newblock Laplace redux-effortless {Bayesian} deep learning.
\newblock \emph{Advances in Neural Information Processing Systems}, 34:\penalty0 20089--20103, 2021.

\bibitem[De~Campos et~al.(1994)De~Campos, Huete, and Moral]{probability_interval_1994}
Luis~M. De~Campos, Juan~F. Huete, and Serafin Moral.
\newblock Probability intervals: A tool for uncertain reasoning.
\newblock \emph{International Journal of Uncertainty, Fuzziness and Knowledge-Based Systems}, 02\penalty0 (02):\penalty0 167--196, June 1994.
\newblock ISSN 0218-4885, 1793-6411.
\newblock \doi{10.1142/S0218488594000146}.

\bibitem[de~Finetti(1974)]{DeFinetti74}
Bruno de~Finetti.
\newblock \emph{Theory of Probability}.
\newblock Wiley, London, 1974.

\bibitem[de~Jong et~al.(2024)de~Jong, Sburlea, and Valdenegro-Toro]{de2024disentangled}
Ivo~Pascal de~Jong, Andreea~Ioana Sburlea, and Matias Valdenegro-Toro.
\newblock How disentangled are your classification uncertainties?
\newblock \emph{arXiv preprint arXiv:2408.12175}, 2024.

\bibitem[Dempster(1967)]{Dempster67}
Arthur~P. Dempster.
\newblock Upper and lower probability inferences based on a sample from a finite univariate population.
\newblock \emph{Biometrika}, 54\penalty0 (3-4):\penalty0 515--528, 1967.

\bibitem[Dempster(2008)]{dempster2008upper}
Arthur~P Dempster.
\newblock Upper and lower probabilities induced by a multivalued mapping.
\newblock In \emph{Classic works of the {Dempster-Shafer} theory of belief functions}, pages 57--72. Springer, 2008.

\bibitem[Deng et~al.(2009)Deng, Dong, Socher, Li, Li, and Fei-Fei]{deng2009imagenet}
Jia Deng, Wei Dong, Richard Socher, Li-Jia Li, Kai Li, and Li~Fei-Fei.
\newblock Imagenet: A large-scale hierarchical image database.
\newblock In \emph{2009 IEEE Conference on Computer Vision and Pattern Recognition}, pages 248--255. Ieee, 2009.

\bibitem[Denneberg and Grabisch(1999)]{denneberg99interaction}
Dieter Denneberg and Michel Grabisch.
\newblock Interaction transform of set functions over a finite set.
\newblock \emph{Information Sciences}, 121\penalty0 (1-2):\penalty0 149--170, 1999.

\bibitem[Denoeux(2000)]{denoeux}
Thierry. Denoeux.
\newblock A neural network classifier based on {{Dempster-Shafer}} theory.
\newblock \emph{IEEE Transactions on Systems, Man, and Cybernetics - Part A: Systems and Humans}, 30\penalty0 (2):\penalty0 131--150, 2000.
\newblock \doi{10.1109/3468.833094}.

\bibitem[Den{\oe}ux(2006)]{denoeux2006constructing}
Thierry Den{\oe}ux.
\newblock Constructing belief functions from sample data using multinomial confidence regions.
\newblock \emph{International Journal of Approximate Reasoning}, 42\penalty0 (3):\penalty0 228--252, 2006.

\bibitem[Den{\oe}ux(2008)]{Denoeux2008classic}
Thierry Den{\oe}ux.
\newblock A k-nearest neighbor classification rule based on {D}empster--{S}hafer theory.
\newblock In Roland~R. Yager and Liping Liu, editors, \emph{Classic Works of the {Dempster-Shafer} Theory of Belief Functions}, volume 219 of \emph{Studies in Fuzziness and Soft Computing}, pages 737--760. Springer, 2008.

\bibitem[Denoeux(2021)]{denoeux2021distributed}
Thierry Denoeux.
\newblock Distributed combination of belief functions.
\newblock \emph{Information Fusion}, 65:\penalty0 179--191, 2021.

\bibitem[Den{\oe}ux(2022)]{https://doi.org/10.48550/arxiv.2208.00647}
Thierry Den{\oe}ux.
\newblock An evidential neural network model for regression based on random fuzzy numbers.
\newblock In \emph{International Conference on Belief Functions}, pages 57--66. Springer, 2022.

\bibitem[Den{\oe}ux and Li(2018)]{denoeux2018frequency}
Thierry Den{\oe}ux and Shoumei Li.
\newblock Frequency-calibrated belief functions: review and new insights.
\newblock \emph{International Journal of Approximate Reasoning}, 92:\penalty0 232--254, 2018.

\bibitem[Depeweg et~al.(2018)Depeweg, Hernandez-Lobato, Doshi-Velez, and Udluft]{depeweg2018decomposition}
Stefan Depeweg, Jose-Miguel Hernandez-Lobato, Finale Doshi-Velez, and Steffen Udluft.
\newblock Decomposition of uncertainty in bayesian deep learning for efficient and risk-sensitive learning.
\newblock In \emph{International conference on machine learning}, pages 1184--1193. PMLR, 2018.

\bibitem[Derman et~al.(2020)Derman, Mankowitz, Mann, and Mannor]{derman2020bayesian}
Esther Derman, Daniel Mankowitz, Timothy Mann, and Shie Mannor.
\newblock A bayesian approach to robust reinforcement learning.
\newblock In \emph{Uncertainty in Artificial Intelligence}, pages 648--658. PMLR, 2020.

\bibitem[Deshmukh et~al.(2019)Deshmukh, Lei, Sharma, Dogan, Cutler, and Scott]{deshmukh2019generalization}
Aniket~Anand Deshmukh, Yunwen Lei, Srinagesh Sharma, Urun Dogan, James~W Cutler, and Clayton Scott.
\newblock A generalization error bound for multi-class domain generalization.
\newblock \emph{arXiv preprint arXiv:1905.10392}, 2019.

\bibitem[Diaz-Rozo et~al.(2018)Diaz-Rozo, Bielza, and Larra{\~n}aga]{diaz2018clustering}
Javier Diaz-Rozo, Concha Bielza, and Pedro Larra{\~n}aga.
\newblock Clustering of data streams with dynamic gaussian mixture models: An iot application in industrial processes.
\newblock \emph{IEEE Internet of Things Journal}, 5\penalty0 (5):\penalty0 3533--3547, 2018.

\bibitem[Droguett and Mosleh(2008)]{droguett2008bayesian}
Enrique~L{\'o}pez Droguett and Ali Mosleh.
\newblock Bayesian methodology for model uncertainty using model performance data.
\newblock \emph{Risk Analysis: An International Journal}, 28\penalty0 (5):\penalty0 1457--1476, 2008.

\bibitem[Dubois and Prade(1987)]{dubois1987properties}
Didier Dubois and Henri Prade.
\newblock Properties of measures of information in evidence and possibility theories.
\newblock \emph{Fuzzy sets and systems}, 24\penalty0 (2):\penalty0 161--182, 1987.

\bibitem[Dubois and Prade(1990)]{Dubois90}
Didier Dubois and Henri Prade.
\newblock Consonant approximations of belief functions.
\newblock \emph{International Journal of Approximate Reasoning}, 4:\penalty0 419--449, 1990.

\bibitem[Dubois and Prade(2012)]{dubois2012possibility}
Didier Dubois and Henri Prade.
\newblock \emph{Possibility theory: an approach to computerized processing of uncertainty}.
\newblock Springer Science \& Business Media, 2012.

\bibitem[Dubois et~al.(1993)Dubois, Prade, and Sandri]{dubois1993possibility}
Didier Dubois, Henri Prade, and Sandra Sandri.
\newblock On possibility/probability transformations.
\newblock In \emph{Fuzzy logic: State of the art}, pages 103--112. Springer, 1993.

\bibitem[Dusenberry et~al.(2020)Dusenberry, Jerfel, Wen, Ma, Snoek, Heller, Lakshminarayanan, and Tran]{dusenberry2020efficient}
Michael Dusenberry, Ghassen Jerfel, Yeming Wen, Yian Ma, Jasper Snoek, Katherine Heller, Balaji Lakshminarayanan, and Dustin Tran.
\newblock Efficient and scalable bayesian neural nets with rank-1 factors.
\newblock In \emph{International conference on machine learning}, pages 2782--2792. PMLR, 2020.

\bibitem[d’Avila Garcez et~al.(2009)d’Avila Garcez, Lamb, and Gabbay]{d2009neural}
Artur~S d’Avila Garcez, Lu{\'\i}s~C Lamb, and Dov~M Gabbay.
\newblock \emph{Neural-symbolic learning systems}.
\newblock Springer, 2009.

\bibitem[Elouedi et~al.(Madrid, 2000)Elouedi, Mellouli, and Smets]{elouedi00decision}
Zied Elouedi, Khaled Mellouli, and Philippe Smets.
\newblock Decision trees using the belief function theory.
\newblock In \emph{Proceedings of the Eighth International Conference on Information Processing and Management of Uncertainty in Knowledge-based Systems (IPMU 2000)}, volume~1, pages 141--148, Madrid, 2000.

\bibitem[Elouedi1 et~al.(2000)Elouedi1, Mellouli1, and Smets]{elouedi}
Zied Elouedi1, Khaled Mellouli1, and Philippe Smets.
\newblock Classification with belief decision trees.
\newblock In \emph{International Conference on Artificial Intelligence: Methodology, Systems, and Applications}, pages 80--90. Springer, 2000.

\bibitem[Farquhar et~al.(2024)Farquhar, Kossen, Kuhn, and Gal]{farquhar2024detecting}
Sebastian Farquhar, Jannik Kossen, Lorenz Kuhn, and Yarin Gal.
\newblock Detecting hallucinations in large language models using semantic entropy.
\newblock \emph{Nature}, 630\penalty0 (8017):\penalty0 625--630, 2024.

\bibitem[Fay et~al.(2006)Fay, Schwenker, Thiel, and Palm]{10.1007/11829898_18}
Rebecca Fay, Friedhelm Schwenker, Christian Thiel, and G{\"u}nther Palm.
\newblock Hierarchical neural networks utilising dempster-shafer evidence theory.
\newblock In Friedhelm Schwenker and Simone Marinai, editors, \emph{Artificial Neural Networks in Pattern Recognition}, pages 198--209, Berlin, Heidelberg, 2006. Springer Berlin Heidelberg.
\newblock ISBN 978-3-540-37952-2.

\bibitem[Faza et~al.(2024)Faza, Shariatmadar, Hallez, and Moens]{faza2024interval}
Ghifari~A Faza, Keivan Shariatmadar, Hans Hallez, and David Moens.
\newblock Interval reduced order surrogate modelling framework for uncertainty quantification.
\newblock In \emph{AIAA Scitech 2024 Forum}, page 0387, 2024.

\bibitem[Fischer et~al.(2019)Fischer, Balunovic, Drachsler-Cohen, Gehr, Zhang, and Vechev]{fischer2019dl2}
Marc Fischer, Mislav Balunovic, Dana Drachsler-Cohen, Timon Gehr, Ce~Zhang, and Martin Vechev.
\newblock Dl2: training and querying neural networks with logic.
\newblock In \emph{International Conference on Machine Learning}, pages 1931--1941. PMLR, 2019.

\bibitem[Fort and Jastrzebski(2019)]{fort2019large}
Stanislav Fort and Stanislaw Jastrzebski.
\newblock Large scale structure of neural network loss landscapes.
\newblock \emph{Advances in Neural Information Processing Systems}, 32, 2019.

\bibitem[Fortuin(2022)]{https://doi.org/10.48550/arxiv.2105.06868}
Vincent Fortuin.
\newblock Priors in {Bayesian Deep Learning: A Review}.
\newblock \emph{International Statistical Review}, 90, 05 2022.
\newblock \doi{10.1111/insr.12502}.

\bibitem[Freedman(1999)]{freedman1999wald}
David Freedman.
\newblock Wald lecture: On the {Bernstein-von Mises} theorem with infinite-dimensional parameters.
\newblock \emph{The Annals of Statistics}, 27\penalty0 (4):\penalty0 1119--1141, 1999.

\bibitem[ga~Liu et~al.(2012)ga~Liu, Dezert, Mercier, and Pan]{Liu2012291}
Zhun ga~Liu, Jean Dezert, Gr\'egoire Mercier, and Quan Pan.
\newblock Belief {C}-means: {A}n extension of fuzzy {C}-means algorithm in belief functions framework.
\newblock \emph{Pattern Recognition Letters}, 33\penalty0 (3):\penalty0 291--300, 2012.

\bibitem[Gal and Ghahramani(2015)]{https://doi.org/10.48550/arxiv.1506.02158}
Yarin Gal and Zoubin Ghahramani.
\newblock {Bayesian} convolutional neural networks with {Bernoulli} approximate variational inference.
\newblock \emph{arXiv preprint arXiv:1506.02158}, 2015.

\bibitem[Gal and Ghahramani(2016)]{gal2016dropout}
Yarin Gal and Zoubin Ghahramani.
\newblock Dropout as a {{Bayesian}} approximation: Representing model uncertainty in deep learning.
\newblock In \emph{International Conference on Machine Learning}, pages 1050--1059. PMLR, 2016.

\bibitem[Ganin and Lempitsky(2015)]{ganin2015unsupervised}
Yaroslav Ganin and Victor Lempitsky.
\newblock Unsupervised domain adaptation by backpropagation.
\newblock In \emph{International conference on machine learning}, pages 1180--1189. PMLR, 2015.

\bibitem[Gao et~al.(2024)Gao, Chen, Xiang, and Xu]{gao2024comprehensive}
Junyu Gao, Mengyuan Chen, Liangyu Xiang, and Changsheng Xu.
\newblock A comprehensive survey on evidential deep learning and its applications.
\newblock \emph{arXiv preprint arXiv:2409.04720}, 2024.

\bibitem[Garczarczyk(2000)]{Garczarczyk2000}
Z.A. Garczarczyk.
\newblock Interval neural networks.
\newblock In \emph{Proceedings of the IEEE International Symposium on Circuits and Systems}, volume~3, pages 567--570 vol.3, 2000.

\bibitem[Garg and Chakraborty(2023)]{garg2023vb}
Shailesh Garg and Souvik Chakraborty.
\newblock Vb-deeponet: A bayesian operator learning framework for uncertainty quantification.
\newblock \emph{Engineering Applications of Artificial Intelligence}, 118:\penalty0 105685, 2023.

\bibitem[Gasperini et~al.(2021)Gasperini, Haug, Mahani, Marcos-Ramiro, Navab, Busam, and Tombari]{gasperini2021certainnet}
Stefano Gasperini, Jan Haug, Mohammad-Ali~Nikouei Mahani, Alvaro Marcos-Ramiro, Nassir Navab, Benjamin Busam, and Federico Tombari.
\newblock Certainnet: Sampling-free uncertainty estimation for object detection.
\newblock \emph{IEEE Robotics and Automation Letters}, 7\penalty0 (2):\penalty0 698--705, 2021.

\bibitem[Geman and Geman(1984)]{4767596}
Stuart Geman and Donald Geman.
\newblock Stochastic relaxation, gibbs distributions, and the {{Bayesian}} restoration of images.
\newblock \emph{IEEE Transactions on Pattern Analysis and Machine Intelligence}, PAMI-6\penalty0 (6):\penalty0 721--741, 1984.
\newblock \doi{10.1109/TPAMI.1984.4767596}.

\bibitem[Ghasedi~Dizaji et~al.(2017)Ghasedi~Dizaji, Herandi, Deng, Cai, and Huang]{ghasedi2017deep}
Kamran Ghasedi~Dizaji, Amirhossein Herandi, Cheng Deng, Weidong Cai, and Heng Huang.
\newblock Deep clustering via joint convolutional autoencoder embedding and relative entropy minimization.
\newblock In \emph{Proceedings of the IEEE international conference on computer vision}, pages 5736--5745, 2017.

\bibitem[Giunchiglia et~al.(2023)Giunchiglia, Stoian, Khan, Cuzzolin, and Lukasiewicz]{giunchiglia2023road}
Eleonora Giunchiglia, Mihaela~C{\u{a}}t{\u{a}}lina Stoian, Salman Khan, Fabio Cuzzolin, and Thomas Lukasiewicz.
\newblock Road-r: The autonomous driving dataset with logical requirements.
\newblock \emph{Machine Learning}, pages 1--31, 2023.

\bibitem[Goan and Fookes(2020)]{goan2020bayesian}
Ethan Goan and Clinton Fookes.
\newblock Bayesian neural networks: An introduction and survey.
\newblock \emph{Case Studies in Applied Bayesian Data Science: CIRM Jean-Morlet Chair, Fall 2018}, pages 45--87, 2020.

\bibitem[Gong and Cuzzolin(2017)]{gong2017belief}
Wenjuan Gong and Fabio Cuzzolin.
\newblock A belief-theoretical approach to example-based pose estimation.
\newblock \emph{IEEE Transactions on Fuzzy Systems}, 26\penalty0 (2):\penalty0 598--611, 2017.

\bibitem[Goodfellow et~al.(2014)Goodfellow, Shlens, and Szegedy]{goodfellow2014explaining}
Ian~J Goodfellow, Jonathon Shlens, and Christian Szegedy.
\newblock Explaining and harnessing adversarial examples.
\newblock \emph{arXiv preprint arXiv:1412.6572}, 2014.

\bibitem[Grabisch et~al.(2000)Grabisch, Sugeno, and Murofushi]{grabisch2000book}
Michel Grabisch, Michio Sugeno, and Toshiaki Murofushi.
\newblock \emph{Fuzzy measures and integrals: theory and applications}.
\newblock New York: Springer, 2000.

\bibitem[Graefe et~al.(2015)Graefe, K{\"u}chenhoff, Stierle, and Riedl]{graefe2015limitations}
Andreas Graefe, Helmut K{\"u}chenhoff, Veronika Stierle, and Bernhard Riedl.
\newblock Limitations of ensemble {{Bayesian}} model averaging for forecasting social science problems.
\newblock \emph{International Journal of Forecasting}, 31\penalty0 (3):\penalty0 943--951, 2015.

\bibitem[Graves(2011)]{NIPS2011_7eb3c8be}
Alex Graves.
\newblock Practical variational inference for neural networks.
\newblock In J.~Shawe-Taylor, R.~Zemel, P.~Bartlett, F.~Pereira, and K.Q. Weinberger, editors, \emph{Advances in Neural Information Processing Systems}, volume~24. Curran Associates, Inc., 2011.
\newblock URL \url{https://proceedings.neurips.cc/paper/2011/file/7eb3c8be3d411e8ebfab08eba5f49632-Paper.pdf}.

\bibitem[Gray et~al.(2025)Gray, Gopakumar, Rousseau, and Destercke]{gray2025guaranteed}
Ander Gray, Vignesh Gopakumar, Sylvain Rousseau, and S{\'e}bastien Destercke.
\newblock Guaranteed confidence-band enclosures for pde surrogates.
\newblock \emph{arXiv preprint arXiv:2501.18426}, 2025.

\bibitem[Greene and Cunningham(2006)]{greene06icml}
Derek Greene and P\'{a}draig Cunningham.
\newblock Practical solutions to the problem of diagonal dominance in kernel document clustering.
\newblock In \emph{Proc. 23rd International Conference on Machine learning (ICML'06)}, pages 377--384. ACM Press, 2006.

\bibitem[Gulrajani and Lopez-Paz(2020)]{gulrajani2020search}
Ishaan Gulrajani and David Lopez-Paz.
\newblock In search of lost domain generalization.
\newblock \emph{arXiv preprint arXiv:2007.01434}, 2020.

\bibitem[Guo et~al.(2017{\natexlab{a}})Guo, Pleiss, Sun, and Weinberger]{guo2017calibration}
Chuan Guo, Geoff Pleiss, Yu~Sun, and Kilian~Q Weinberger.
\newblock On calibration of modern neural networks.
\newblock In \emph{International conference on machine learning}, pages 1321--1330. PMLR, 2017{\natexlab{a}}.

\bibitem[Guo et~al.(2017{\natexlab{b}})Guo, Liu, Zhu, and Yin]{guo2017deep}
Xifeng Guo, Xinwang Liu, En~Zhu, and Jianping Yin.
\newblock Deep clustering with convolutional autoencoders.
\newblock In \emph{Neural Information Processing: 24th International Conference, ICONIP 2017, Guangzhou, China, November 14-18, 2017, Proceedings, Part II 24}, pages 373--382. Springer, 2017{\natexlab{b}}.

\bibitem[Gustafsson et~al.(2020)Gustafsson, Danelljan, and Schon]{gustafsson2020evaluating}
Fredrik~K Gustafsson, Martin Danelljan, and Thomas~B Schon.
\newblock Evaluating scalable bayesian deep learning methods for robust computer vision.
\newblock In \emph{Proceedings of the IEEE/CVF conference on computer vision and pattern recognition workshops}, pages 318--319, 2020.

\bibitem[Halpern(2017)]{halpern03book}
Joseph~Y. Halpern.
\newblock \emph{Reasoning About Uncertainty}.
\newblock MIT Press, 2017.

\bibitem[Hastings(1970)]{hastings1970monte}
W.~K. Hastings.
\newblock Monte carlo sampling methods using markov chains and their applications.
\newblock \emph{Biometrika}, 57\penalty0 (1):\penalty0 97--109, 1970.
\newblock ISSN 00063444, 14643510.
\newblock URL \url{http://www.jstor.org/stable/2334940}.

\bibitem[Havrilla et~al.(2023)Havrilla, Zhuravinskyi, Phung, Tiwari, Tow, Biderman, Anthony, and Castricato]{havrilla2023trlx}
Alexander Havrilla, Maksym Zhuravinskyi, Duy Phung, Aman Tiwari, Jonathan Tow, Stella Biderman, Quentin Anthony, and Louis Castricato.
\newblock trlx: A framework for large scale reinforcement learning from human feedback.
\newblock In \emph{Proceedings of the 2023 Conference on Empirical Methods in Natural Language Processing}, pages 8578--8595, 2023.

\bibitem[He et~al.(2020)He, Lakshminarayanan, and Teh]{he2020bayesian}
Bobby He, Balaji Lakshminarayanan, and Yee~Whye Teh.
\newblock Bayesian deep ensembles via the neural tangent kernel.
\newblock \emph{Advances in neural information processing systems}, 33:\penalty0 1010--1022, 2020.

\bibitem[He et~al.(2016)He, Zhang, Ren, and Sun]{He_2016_CVPR}
Kaiming He, Xiangyu Zhang, Shaoqing Ren, and Jian Sun.
\newblock Deep residual learning for image recognition.
\newblock In \emph{Proceedings of the IEEE Conference on Computer Vision and Pattern Recognition (CVPR)}, June 2016.

\bibitem[Hendrycks and Gimpel(2016)]{hendrycks2016baseline}
Dan Hendrycks and Kevin Gimpel.
\newblock A baseline for detecting misclassified and out-of-distribution examples in neural networks.
\newblock \emph{arXiv preprint arXiv:1610.02136}, 2016.

\bibitem[Hendrycks et~al.(2021)Hendrycks, Zhao, Basart, Steinhardt, and Song]{hendrycks2021natural}
Dan Hendrycks, Kevin Zhao, Steven Basart, Jacob Steinhardt, and Dawn Song.
\newblock Natural adversarial examples.
\newblock In \emph{Proceedings of the IEEE/CVF Conference on Computer Vision and Pattern Recognition}, pages 15262--15271, 2021.

\bibitem[Hinne et~al.(2020)Hinne, Gronau, van~den Bergh, and Wagenmakers]{hinne2020conceptual}
Max Hinne, Quentin~F Gronau, Don van~den Bergh, and Eric-Jan Wagenmakers.
\newblock A conceptual introduction to {{Bayesian}} model averaging.
\newblock \emph{Advances in Methods and Practices in Psychological Science}, 3\penalty0 (2):\penalty0 200--215, 2020.

\bibitem[Hobbhahn et~al.(2022)Hobbhahn, Kristiadi, and Hennig]{hobbhahn2022fast}
Marius Hobbhahn, Agustinus Kristiadi, and Philipp Hennig.
\newblock Fast predictive uncertainty for classification with {{Bayesian}} deep networks.
\newblock In \emph{Uncertainty in Artificial Intelligence}, pages 822--832. PMLR, 2022.

\bibitem[Hoeting et~al.(1999)Hoeting, Madigan, Raftery, and Volinsky]{hoeting1999bayesian}
Jennifer~A Hoeting, David Madigan, Adrian~E Raftery, and Chris~T Volinsky.
\newblock Bayesian model averaging: a tutorial (with comments by m. clyde, david draper and ei george, and a rejoinder by the authors.
\newblock \emph{Statistical science}, 14\penalty0 (4):\penalty0 382--417, 1999.

\bibitem[Hoffman et~al.(2014)Hoffman, Gelman, et~al.]{hoffman2014no}
Matthew~D Hoffman, Andrew Gelman, et~al.
\newblock {The No-U-Turn sampler: adaptively setting path lengths in Hamiltonian Monte Carlo}.
\newblock \emph{J. Mach. Learn. Res.}, 15\penalty0 (1):\penalty0 1593--1623, 2014.

\bibitem[Hu et~al.(2020)Hu, Zhang, Chen, and Chan]{hu2020domain}
Shoubo Hu, Kun Zhang, Zhitang Chen, and Laiwan Chan.
\newblock Domain generalization via multidomain discriminant analysis.
\newblock In \emph{Uncertainty in artificial intelligence}, pages 292--302. PMLR, 2020.

\bibitem[Huang et~al.(2014)Huang, Zang, and Zheng]{huang2014evidence}
Allen~H Huang, Amy~Y Zang, and Rong Zheng.
\newblock Evidence on the information content of text in analyst reports.
\newblock \emph{The Accounting Review}, 89\penalty0 (6):\penalty0 2151--2180, 2014.

\bibitem[H{\"{u}}llermeier and Waegeman(2019)]{DBLP:journals/corr/abs-1910-09457}
Eyke H{\"{u}}llermeier and Willem Waegeman.
\newblock Aleatoric and epistemic uncertainty in machine learning: {A} tutorial introduction.
\newblock \emph{CoRR}, abs/1910.09457, 2019.
\newblock URL \url{http://arxiv.org/abs/1910.09457}.

\bibitem[H{\"u}llermeier and Waegeman(2021)]{hullermeier2021aleatoric}
Eyke H{\"u}llermeier and Willem Waegeman.
\newblock Aleatoric and epistemic uncertainty in machine learning: An introduction to concepts and methods.
\newblock \emph{Machine Learning}, 110\penalty0 (3):\penalty0 457--506, 2021.

\bibitem[H{\"u}llermeier et~al.(2022)H{\"u}llermeier, Destercke, and Shaker]{hullermeier2022quantification}
Eyke H{\"u}llermeier, S{\'e}bastien Destercke, and Mohammad~Hossein Shaker.
\newblock Quantification of credal uncertainty in machine learning: A critical analysis and empirical comparison.
\newblock In \emph{Proceedings of the Uncertainty in Artificial Intelligence}, pages 548--557. PMLR, 2022.

\bibitem[Ishibuchi et~al.(1993)Ishibuchi, Tanaka, and Okada]{Ishibuchi1993}
Hisao Ishibuchi, Hideo Tanaka, and Hidehiko Okada.
\newblock An architecture of neural networks with interval weights and its application to fuzzy regression analysis.
\newblock \emph{Fuzzy Sets and Systems}, 57\penalty0 (1):\penalty0 27--39, 1993.

\bibitem[Itkina et~al.(2020)Itkina, Ivanovic, Senanayake, Kochenderfer, and Pavone]{itkina2020evidential}
Masha Itkina, Boris Ivanovic, Ransalu Senanayake, Mykel~J Kochenderfer, and Marco Pavone.
\newblock Evidential sparsification of multimodal latent spaces in conditional variational autoencoders.
\newblock \emph{Advances in Neural Information Processing Systems}, 33:\penalty0 10235--10246, 2020.

\bibitem[Javaheripi et~al.(2023)Javaheripi, Bubeck, Abdin, Aneja, Bubeck, Mendes, Chen, Del~Giorno, Eldan, Gopi, et~al.]{javaheripi2023phi}
Mojan Javaheripi, S{\'e}bastien Bubeck, Marah Abdin, Jyoti Aneja, Sebastien Bubeck, Caio C{\'e}sar~Teodoro Mendes, Weizhu Chen, Allie Del~Giorno, Ronen Eldan, Sivakanth Gopi, et~al.
\newblock Phi-2: The surprising power of small language models.
\newblock \emph{Microsoft Research Blog}, 1\penalty0 (3):\penalty0 3, 2023.

\bibitem[Javanmardi et~al.(2024)Javanmardi, Stutz, and H{\"u}llermeier]{Javanmardi2024Conformalized}
Alireza Javanmardi, David Stutz, and Eyke H{\"u}llermeier.
\newblock Conformalized credal set predictors.
\newblock \emph{arXiv preprint arXiv:2402.10723}, 2024.

\bibitem[Jeffreys(1998)]{jeffreys1998theory}
Harold Jeffreys.
\newblock \emph{The theory of probability}.
\newblock OuP Oxford, 1998.

\bibitem[Jha et~al.(2024)Jha, Gong, Zhao, and Yao]{jha2024npcl}
Saurav Jha, Dong Gong, He~Zhao, and Lina Yao.
\newblock Npcl: Neural processes for uncertainty-aware continual learning.
\newblock \emph{Advances in Neural Information Processing Systems}, 36, 2024.

\bibitem[Jiang et~al.(2023)Jiang, Sablayrolles, Mensch, Bamford, Chaplot, de~las Casas, Bressand, Lengyel, Lample, Saulnier, Lavaud, Lachaux, Stock, Scao, Lavril, Wang, Lacroix, and Sayed]{jiang2023mistral7b}
Albert~Q. Jiang, Alexandre Sablayrolles, Arthur Mensch, Chris Bamford, Devendra~Singh Chaplot, Diego de~las Casas, Florian Bressand, Gianna Lengyel, Guillaume Lample, Lucile Saulnier, Lélio~Renard Lavaud, Marie-Anne Lachaux, Pierre Stock, Teven~Le Scao, Thibaut Lavril, Thomas Wang, Timothée Lacroix, and William~El Sayed.
\newblock Mistral 7b, 2023.
\newblock URL \url{https://arxiv.org/abs/2310.06825}.

\bibitem[J{\o}sang(2016)]{josang2016subjective}
Audun J{\o}sang.
\newblock \emph{Subjective logic}, volume~3.
\newblock Springer, 2016.

\bibitem[Jospin et~al.(2022)Jospin, Laga, Boussaid, Buntine, and Bennamoun]{DBLP:journals/corr/abs-2007-06823}
Laurent~Valentin Jospin, Hamid Laga, Farid Boussaid, Wray Buntine, and Mohammed Bennamoun.
\newblock Hands-on bayesian neural networks—a tutorial for deep learning users.
\newblock \emph{IEEE Computational Intelligence Magazine}, 17\penalty0 (2):\penalty0 29--48, 2022.

\bibitem[Jumper et~al.(2021)Jumper, Evans, Pritzel, Green, Figurnov, Ronneberger, Tunyasuvunakool, Bates, {\v{Z}}{\'\i}dek, Potapenko, et~al.]{jumper2021highly}
John Jumper, Richard Evans, Alexander Pritzel, Tim Green, Michael Figurnov, Olaf Ronneberger, Kathryn Tunyasuvunakool, Russ Bates, Augustin {\v{Z}}{\'\i}dek, Anna Potapenko, et~al.
\newblock Highly accurate protein structure prediction with alphafold.
\newblock \emph{nature}, 596\penalty0 (7873):\penalty0 583--589, 2021.

\bibitem[Kass and Wasserman(1996)]{kass1996selection}
Robert~E Kass and Larry Wasserman.
\newblock The selection of prior distributions by formal rules.
\newblock \emph{Journal of the American statistical Association}, 91\penalty0 (435):\penalty0 1343--1370, 1996.

\bibitem[Kendall and Gal(2017)]{kendall2017uncertainties}
Alex Kendall and Yarin Gal.
\newblock What uncertainties do we need in {{Bayesian}} deep learning for computer vision?
\newblock \emph{arXiv:1703.04977}, 2017.

\bibitem[Kendall(1974)]{Kendall74foundations}
David~G. Kendall.
\newblock Foundations of a theory of random sets.
\newblock In E.~F. Harding and D.~G. Kendall, editors, \emph{Stochastic Geometry}, pages 322--376. Wiley, London, 1974.

\bibitem[Keynes(2013)]{keynes2013treatise}
John~Maynard Keynes.
\newblock \emph{A treatise on probability}.
\newblock Courier Corporation, 2013.

\bibitem[Khan et~al.(2024)Khan, Teeti, Alitappeh, Stoian, Giunchiglia, Singh, Bradley, and Cuzzolin]{khan2024road}
Salman Khan, Izzeddin Teeti, Reza~Javanmard Alitappeh, Mihaela~C Stoian, Eleonora Giunchiglia, Gurkirt Singh, Andrew Bradley, and Fabio Cuzzolin.
\newblock Road-waymo: Action awareness at scale for autonomous driving.
\newblock \emph{arXiv preprint arXiv:2411.01683}, 2024.

\bibitem[Khosravi et~al.(2011)Khosravi, Nahavandi, Creighton, and Atiya]{Khosrav2011DataDrivenIntervals}
Abbas Khosravi, Saeid Nahavandi, Doug Creighton, and Amir~F. Atiya.
\newblock Lower upper bound estimation method for construction of neural network-based prediction intervals.
\newblock \emph{IEEE Transactions on Neural Networks}, 22\penalty0 (3):\penalty0 337--346, 2011.

\bibitem[Kingma and Welling(2013)]{https://doi.org/10.48550/arxiv.1312.6114}
Diederik~P Kingma and Max Welling.
\newblock Auto-encoding {Variational Bayes}.
\newblock \emph{arXiv preprint arXiv:1312.6114}, 2013.

\bibitem[Kingma et~al.(2015)Kingma, Salimans, and Welling]{https://doi.org/10.48550/arxiv.1506.02557}
Durk~P Kingma, Tim Salimans, and Max Welling.
\newblock Variational dropout and the local {Reparameterization} trick.
\newblock \emph{Advances in Neural Information Processing Systems}, 28, 2015.

\bibitem[Kirkpatrick et~al.(2017)Kirkpatrick, Pascanu, Rabinowitz, Veness, Desjardins, Rusu, Milan, Quan, Ramalho, Grabska-Barwinska, et~al.]{kirkpatrick2017overcoming}
James Kirkpatrick, Razvan Pascanu, Neil Rabinowitz, Joel Veness, Guillaume Desjardins, Andrei~A Rusu, Kieran Milan, John Quan, Tiago Ramalho, Agnieszka Grabska-Barwinska, et~al.
\newblock Overcoming catastrophic forgetting in neural networks.
\newblock \emph{Proceedings of the national academy of sciences}, 114\penalty0 (13):\penalty0 3521--3526, 2017.

\bibitem[Kleijn and Van~der Vaart(2012)]{kleijn2012bernstein}
Bas~JK Kleijn and Aad~W Van~der Vaart.
\newblock The bernstein-von-mises theorem under misspecification.
\newblock 2012.

\bibitem[Klir(1987)]{klir1987we}
George~J Klir.
\newblock Where do we stand on measures of uncertainty, ambiguity, fuzziness, and the like?
\newblock \emph{Fuzzy sets and systems}, 24\penalty0 (2):\penalty0 141--160, 1987.

\bibitem[Koh et~al.(2021)Koh, Sagawa, Marklund, Xie, Zhang, Balsubramani, Hu, Yasunaga, Phillips, Gao, et~al.]{koh2021wilds}
Pang~Wei Koh, Shiori Sagawa, Henrik Marklund, Sang~Michael Xie, Marvin Zhang, Akshay Balsubramani, Weihua Hu, Michihiro Yasunaga, Richard~Lanas Phillips, Irena Gao, et~al.
\newblock Wilds: A benchmark of in-the-wild distribution shifts.
\newblock In \emph{International conference on machine learning}, pages 5637--5664. PMLR, 2021.

\bibitem[Kolmogorov(1965)]{kolmogorov1965three}
Andrei~N Kolmogorov.
\newblock Three approaches to the quantitative definition ofinformation’.
\newblock \emph{Problems of information transmission}, 1\penalty0 (1):\penalty0 1--7, 1965.

\bibitem[Kolmogorov and Bharucha-Reid(2018)]{kolmogorov}
Andrei~N. Kolmogorov and Albert~T. Bharucha-Reid.
\newblock \emph{Foundations of the theory of probability: Second English Edition}.
\newblock Courier Dover Publications, 2018.

\bibitem[Kopetzki et~al.(2021)Kopetzki, Charpentier, Z{\"u}gner, Giri, and G{\"u}nnemann]{kopetzki2021evaluating}
Anna-Kathrin Kopetzki, Bertrand Charpentier, Daniel Z{\"u}gner, Sandhya Giri, and Stephan G{\"u}nnemann.
\newblock Evaluating robustness of predictive uncertainty estimation: Are dirichlet-based models reliable?
\newblock In \emph{International Conference on Machine Learning}, pages 5707--5718. PMLR, 2021.

\bibitem[Kowalski and Kulczycki(2017)]{kowalski2017interval}
Piotr~A Kowalski and Piotr Kulczycki.
\newblock Interval probabilistic neural network.
\newblock \emph{Neural Computing and Applications}, 28\penalty0 (4):\penalty0 817--834, 2017.

\bibitem[Kramosil(1999)]{kramosil1999nonspecificity}
Ivan Kramosil.
\newblock Nonspecificity degrees of basic probability assignments in {D}empster--{S}hafer theory.
\newblock \emph{Computing and Informatics}, 18\penalty0 (6):\penalty0 559--574, 1999.

\bibitem[Krizhevsky(2012)]{krizhevsky2009learning}
Alex Krizhevsky.
\newblock Learning multiple layers of features from tiny images.
\newblock \emph{University of Toronto}, 2012.

\bibitem[Krizhevsky et~al.(2009)Krizhevsky, Nair, and Hinton]{cifar10}
Alex Krizhevsky, Vinod Nair, and Geoffrey Hinton.
\newblock {CIFAR-10 (Canadian Institute For Advanced Research)}.
\newblock Technical report, CIFAR, 2009.
\newblock URL \url{https://www.cs.toronto.edu/~kriz/cifar.html}.

\bibitem[Krizhevsky et~al.(2012)Krizhevsky, Sutskever, and Hinton]{NIPS2012_c399862d}
Alex Krizhevsky, Ilya Sutskever, and Geoffrey~E Hinton.
\newblock Imagenet classification with deep convolutional neural networks.
\newblock In F.~Pereira, C.J. Burges, L.~Bottou, and K.Q. Weinberger, editors, \emph{Advances in Neural Information Processing Systems}, volume~25. Curran Associates, Inc., 2012.
\newblock URL \url{https://proceedings.neurips.cc/paper_files/paper/2012/file/c399862d3b9d6b76c8436e924a68c45b-Paper.pdf}.

\bibitem[Kuleshov et~al.(2018)Kuleshov, Fenner, and Ermon]{kuleshov2018accurate}
Volodymyr Kuleshov, Nathan Fenner, and Stefano Ermon.
\newblock Accurate uncertainties for deep learning using calibrated regression.
\newblock In \emph{International conference on machine learning}, pages 2796--2804. PMLR, 2018.

\bibitem[Kulis and Jordan(2011)]{kulis2011revisiting}
Brian Kulis and Michael~I Jordan.
\newblock Revisiting k-means: New algorithms via bayesian nonparametrics.
\newblock \emph{arXiv preprint arXiv:1111.0352}, 2011.

\bibitem[Kwon et~al.(2020)Kwon, Won, Kim, and Paik]{kwon2020uncertainty}
Yongchan Kwon, Joong-Ho Won, Beom~Joon Kim, and Myunghee~Cho Paik.
\newblock {Uncertainty quantification using Bayesian neural networks in classification: Application to biomedical image segmentation}.
\newblock \emph{Computational Statistics \& Data Analysis}, 142:\penalty0 106816, 2020.

\bibitem[Laanaya et~al.(2010)Laanaya, Martin, Aboutajdine, and Khenchaf]{Laanaya2010338}
Hicham Laanaya, Arnaud Martin, Driss Aboutajdine, and Ali Khenchaf.
\newblock Support vector regression of membership functions and belief functions -- {A}pplication for pattern recognition.
\newblock \emph{Information Fusion}, 11\penalty0 (4):\penalty0 338--350, 2010.

\bibitem[Lai et~al.(2022)Lai, Shi, Han, Shao, Qi, and Li]{lai2022exploring}
Yuandu Lai, Yucheng Shi, Yahong Han, Yunfeng Shao, Meiyu Qi, and Bingshuai Li.
\newblock Exploring uncertainty in regression neural networks for construction of prediction intervals.
\newblock \emph{Neurocomputing}, 481:\penalty0 249--257, 2022.

\bibitem[Lakshminarayanan et~al.(2017)Lakshminarayanan, Pritzel, and Blundell]{lakshminarayanan2017simple}
Balaji Lakshminarayanan, Alexander Pritzel, and Charles Blundell.
\newblock Simple and {Scalable Predictive Uncertainty Estimation using Deep Ensembles}.
\newblock \emph{Advances in Neural Information Processing Systems}, 30, 2017.

\bibitem[Lambrou et~al.(2010)Lambrou, Papadopoulos, and Gammerman]{lambrou2010reliable}
Antonis Lambrou, Harris Papadopoulos, and Alex Gammerman.
\newblock Reliable confidence measures for medical diagnosis with evolutionary algorithms.
\newblock \emph{IEEE Transactions on Information Technology in Biomedicine}, 15\penalty0 (1):\penalty0 93--99, 2010.

\bibitem[Lampinen and Vehtari(2001)]{lampinen2001bayesian}
Jouko Lampinen and Aki Vehtari.
\newblock Bayesian approach for neural networks—review and case studies.
\newblock \emph{Neural networks}, 14\penalty0 (3):\penalty0 257--274, 2001.

\bibitem[LeCun and Cortes(2005)]{LeCun2005TheMD}
Yann LeCun and Corinna Cortes.
\newblock The {MNIST} database of handwritten digits.
\newblock In \emph{Proceedings of the IEEE Conference on Computer Vision and Pattern Recognition (CVPR)}, pages 1--9, 2005.
\newblock \doi{10.1109/CVPR.2005.177}.
\newblock URL \url{https://ieeexplore.ieee.org/document/1467284}.

\bibitem[LeCun et~al.(2015)LeCun, Bengio, and Hinton]{lecun2015deep}
Yann LeCun, Yoshua Bengio, and Geoffrey Hinton.
\newblock Deep learning.
\newblock \emph{Nature}, 521\penalty0 (7553):\penalty0 436--444, 2015.

\bibitem[Levi(1980)]{levi1980enterprise}
Isaac Levi.
\newblock \emph{The enterprise of knowledge: An essay on knowledge, credal probability, and chance}.
\newblock MIT press, 1980.

\bibitem[Liu et~al.(2020)Liu, Lin, Padhy, Tran, Bedrax~Weiss, and Lakshminarayanan]{liu2020simple}
Jeremiah Liu, Zi~Lin, Shreyas Padhy, Dustin Tran, Tania Bedrax~Weiss, and Balaji Lakshminarayanan.
\newblock Simple and principled uncertainty estimation with deterministic deep learning via distance awareness.
\newblock \emph{Advances in neural information processing systems}, 33:\penalty0 7498--7512, 2020.

\bibitem[Liu et~al.(2019)Liu, Liu, Dezert, and Cuzzolin]{liu2019evidence}
Zhun-Ga Liu, Yu~Liu, Jean Dezert, and Fabio Cuzzolin.
\newblock Evidence combination based on credal belief redistribution for pattern classification.
\newblock \emph{IEEE Transactions on Fuzzy Systems}, 28\penalty0 (4):\penalty0 618--631, 2019.

\bibitem[Louizos and Welling(2017)]{louizos2017multiplicative}
Christos Louizos and Max Welling.
\newblock Multiplicative normalizing flows for variational bayesian neural networks.
\newblock In \emph{International Conference on Machine Learning}, pages 2218--2227. PMLR, 2017.

\bibitem[Luo et~al.(2022)Luo, Bhatnagar, Bai, Zhao, Wang, Xiong, Savarese, Ermon, Schmerling, and Pavone]{luo2022local}
Rachel Luo, Aadyot Bhatnagar, Yu~Bai, Shengjia Zhao, Huan Wang, Caiming Xiong, Silvio Savarese, Stefano Ermon, Edward Schmerling, and Marco Pavone.
\newblock Local calibration: metrics and recalibration.
\newblock In \emph{Uncertainty in Artificial Intelligence}, pages 1286--1295. PMLR, 2022.

\bibitem[MacKay(1992)]{10.1162/neco.1992.4.3.448}
David J.~C. MacKay.
\newblock {A Practical {{Bayesian}} Framework for Backpropagation Networks}.
\newblock \emph{Neural Computation}, 4\penalty0 (3):\penalty0 448--472, 05 1992.
\newblock ISSN 0899-7667.
\newblock \doi{10.1162/neco.1992.4.3.448}.
\newblock URL \url{https://doi.org/10.1162/neco.1992.4.3.448}.

\bibitem[Ma{\'c}kiewicz and Ratajczak(1993)]{mackiewicz1993principal}
Andrzej Ma{\'c}kiewicz and Waldemar Ratajczak.
\newblock Principal components analysis (pca).
\newblock \emph{Computers \& Geosciences}, 19\penalty0 (3):\penalty0 303--342, 1993.

\bibitem[M{a}dry et~al.(2017)M{a}dry, Makelov, Schmidt, Tsipras, and Vladu]{madry2017towards}
Aleksander M{a}dry, Aleksandar Makelov, Ludwig Schmidt, Dimitris Tsipras, and Adrian Vladu.
\newblock Towards deep learning models resistant to adversarial attacks.
\newblock \emph{stat}, 1050\penalty0 (9), 2017.

\bibitem[Magnani et~al.(2022)Magnani, Kr{\"a}mer, Eschenhagen, Rosasco, and Hennig]{magnani2022approximate}
Emilia Magnani, Nicholas Kr{\"a}mer, Runa Eschenhagen, Lorenzo Rosasco, and Philipp Hennig.
\newblock Approximate bayesian neural operators: Uncertainty quantification for parametric pdes.
\newblock \emph{arXiv preprint arXiv:2208.01565}, 2022.

\bibitem[Malinin and Gales(2018)]{malinin2018predictive}
Andrey Malinin and Mark Gales.
\newblock Predictive uncertainty estimation via prior networks.
\newblock \emph{Advances in neural information processing systems}, 31, 2018.

\bibitem[Malinin and Gales(2019)]{malinin2019reverse}
Andrey Malinin and Mark Gales.
\newblock Reverse {KL}-divergence training of prior networks: Improved uncertainty and adversarial robustness.
\newblock \emph{Advances in Neural Information Processing Systems}, 32, 2019.

\bibitem[Malinin et~al.(2019)Malinin, Mlodozeniec, and Gales]{malinin2019ensemble}
Andrey Malinin, Bruno Mlodozeniec, and Mark Gales.
\newblock Ensemble distribution distillation.
\newblock In \emph{International Conference on Learning Representations}, 2019.

\bibitem[Manchingal and Cuzzolin(2022)]{manchingal2022epistemic}
Shireen~Kudukkil Manchingal and Fabio Cuzzolin.
\newblock Epistemic deep learning.
\newblock \emph{arXiv preprint arXiv:2206.07609}, 2022.

\bibitem[Manchingal et~al.(2025{\natexlab{a}})Manchingal, Mubashar, Wang, and Cuzzolin]{manchingal2025unifiedevaluationframeworkepistemic}
Shireen~Kudukkil Manchingal, Muhammad Mubashar, Kaizheng Wang, and Fabio Cuzzolin.
\newblock A unified evaluation framework for epistemic predictions, 2025{\natexlab{a}}.
\newblock URL \url{https://arxiv.org/abs/2501.16912}.

\bibitem[Manchingal et~al.(2025{\natexlab{b}})Manchingal, Mubashar, Wang, Shariatmadar, and Cuzzolin]{manchingal2025randomsetneuralnetworksrsnn}
Shireen~Kudukkil Manchingal, Muhammad Mubashar, Kaizheng Wang, Keivan Shariatmadar, and Fabio Cuzzolin.
\newblock Random-set neural networks.
\newblock In \emph{The Thirteenth International Conference on Learning Representations}, 2025{\natexlab{b}}.
\newblock URL \url{https://openreview.net/forum?id=pdjkikvCch}.

\bibitem[Manhaeve et~al.(2018)Manhaeve, Dumancic, Kimmig, Demeester, and De~Raedt]{manhaeve2018deepproblog}
Robin Manhaeve, Sebastijan Dumancic, Angelika Kimmig, Thomas Demeester, and Luc De~Raedt.
\newblock Deepproblog: Neural probabilistic logic programming.
\newblock \emph{Advances in neural information processing systems}, 31, 2018.

\bibitem[Mao et~al.(2019)Mao, Gan, Kohli, Tenenbaum, and Wu]{mao2019neuro}
Jiayuan Mao, Chuang Gan, Pushmeet Kohli, Joshua~B Tenenbaum, and Jiajun Wu.
\newblock The neuro-symbolic concept learner: Interpreting scenes, words, and sentences from natural supervision.
\newblock \emph{arXiv preprint arXiv:1904.12584}, 2019.

\bibitem[Marcot(2012)]{MARCOT201250}
Bruce~G. Marcot.
\newblock Metrics for evaluating performance and uncertainty of bayesian network models.
\newblock \emph{Ecological Modelling}, 230:\penalty0 50--62, 2012.
\newblock ISSN 0304-3800.
\newblock \doi{https://doi.org/10.1016/j.ecolmodel.2012.01.013}.
\newblock URL \url{https://www.sciencedirect.com/science/article/pii/S0304380012000245}.

\bibitem[M{\'a}rquez-Neila et~al.(2017)M{\'a}rquez-Neila, Salzmann, and Fua]{marquez2017imposing}
Pablo M{\'a}rquez-Neila, Mathieu Salzmann, and Pascal Fua.
\newblock Imposing hard constraints on deep networks: Promises and limitations.
\newblock \emph{arXiv preprint arXiv:1706.02025}, 2017.

\bibitem[Martin and Liu(2013)]{martin2013inferential}
Ryan Martin and Chuanhai Liu.
\newblock {Inferential models: A framework for prior-free posterior probabilistic inference}.
\newblock \emph{Journal of the American Statistical Association}, 108\penalty0 (501):\penalty0 301--313, 2013.

\bibitem[Martin and Liu(2015)]{martin2015inferential}
Ryan Martin and Chuanhai Liu.
\newblock \emph{Inferential models: reasoning with uncertainty}.
\newblock CRC Press, 2015.

\bibitem[Matheron(1975)]{Matheron75}
Georges Matheron.
\newblock \emph{Random sets and integral geometry}.
\newblock Wiley Series in Probability and Mathematical Statistics, New York, 1975.

\bibitem[Maynez et~al.(2020)Maynez, Narayan, Bohnet, and McDonald]{maynez2020faithfulness}
Joshua Maynez, Shashi Narayan, Bernd Bohnet, and Ryan McDonald.
\newblock On faithfulness and factuality in abstractive summarization.
\newblock \emph{arXiv preprint arXiv:2005.00661}, 2020.

\bibitem[McInnes et~al.(2018)McInnes, Healy, and Melville]{mcinnes2018umap}
Leland McInnes, John Healy, and James Melville.
\newblock Umap: Uniform manifold approximation and projection for dimension reduction.
\newblock \emph{arXiv preprint arXiv:1802.03426}, 2018.

\bibitem[Men{\'e}ndez et~al.(1997)Men{\'e}ndez, Pardo, Pardo, and Pardo]{menendez1997jensen}
Mar{\'\i}a~Luisa Men{\'e}ndez, JA~Pardo, L~Pardo, and MC~Pardo.
\newblock The jensen-shannon divergence.
\newblock \emph{Journal of the Franklin Institute}, 334\penalty0 (2):\penalty0 307--318, 1997.

\bibitem[Mihaylov et~al.(2018)Mihaylov, Clark, Khot, and Sabharwal]{OpenBookQA2018}
Todor Mihaylov, Peter Clark, Tushar Khot, and Ashish Sabharwal.
\newblock Can a suit of armor conduct electricity? a new dataset for open book question answering.
\newblock In \emph{EMNLP}, 2018.

\bibitem[Minderer et~al.(2021)Minderer, Djolonga, Romijnders, Hubis, Zhai, Houlsby, Tran, and Lucic]{minderer2021revisiting}
Matthias Minderer, Josip Djolonga, Rob Romijnders, Frances Hubis, Xiaohua Zhai, Neil Houlsby, Dustin Tran, and Mario Lucic.
\newblock Revisiting the calibration of modern neural networks.
\newblock \emph{Advances in Neural Information Processing Systems}, 34:\penalty0 15682--15694, 2021.

\bibitem[Miranda(2008)]{miranda2008survey}
Enrique Miranda.
\newblock A survey of the theory of coherent lower previsions.
\newblock \emph{International Journal of Approximate Reasoning}, 48\penalty0 (2):\penalty0 628--658, 2008.

\bibitem[Miranda et~al.(2021)Miranda, Montes, and Vicig]{miranda2021selection}
Enrique Miranda, Ignacio Montes, and Paolo Vicig.
\newblock On the selection of an optimal outer approximation of a coherent lower probability.
\newblock \emph{Fuzzy Sets and Systems}, 424:\penalty0 1--36, 2021.

\bibitem[Miranda et~al.(2023)Miranda, Montes, and Presa]{miranda2023inner}
Enrique Miranda, Ignacio Montes, and Andr{\'e}s Presa.
\newblock Inner approximations of coherent lower probabilities and their application to decision making problems.
\newblock \emph{Annals of Operations Research}, pages 1--39, 2023.

\bibitem[Mitchell(1980)]{mitchell1980need}
Tom~M Mitchell.
\newblock The need for biases in learning generalizations.
\newblock 1980.

\bibitem[Molchanov(2017)]{Molchanov_2017}
Ilya Molchanov.
\newblock Random sets and random functions.
\newblock \emph{Theory of Random Sets}, pages 451--552, 2017.

\bibitem[Molchanov(2005)]{Molchanov05}
Ilya~S Molchanov.
\newblock \emph{Theory of random sets}, volume~19.
\newblock Springer, 2005.

\bibitem[Mortier et~al.(2021)Mortier, Wydmuch, Dembczy{\'n}ski, H{\"u}llermeier, and Waegeman]{Mortier}
Thomas Mortier, Marek Wydmuch, Krzysztof Dembczy{\'n}ski, Eyke H{\"u}llermeier, and Willem Waegeman.
\newblock Efficient set-valued prediction in multi-class classification.
\newblock \emph{Data Mining and Knowledge Discovery}, 35\penalty0 (4):\penalty0 1435--1469, 2021.

\bibitem[Muandet et~al.(2013)Muandet, Balduzzi, and Sch\"{o}lkopf]{10.5555/3042817.3042820}
Krikamol Muandet, David Balduzzi, and Bernhard Sch\"{o}lkopf.
\newblock Domain generalization via invariant feature representation.
\newblock In \emph{Proceedings of the 30th International Conference on International Conference on Machine Learning - Volume 28}, ICML'13, page I–10–I–18. JMLR.org, 2013.

\bibitem[Mucs{\'a}nyi et~al.(2024)Mucs{\'a}nyi, Kirchhof, and Oh]{mucsanyi2024benchmarking}
B{\'a}lint Mucs{\'a}nyi, Michael Kirchhof, and Seong~Joon Oh.
\newblock Benchmarking uncertainty disentanglement: Specialized uncertainties for specialized tasks.
\newblock \emph{Advances in neural information processing systems}, 37:\penalty0 50972--51038, 2024.

\bibitem[Mukhoti et~al.(2020)Mukhoti, Kulharia, Sanyal, Golodetz, Torr, and Dokania]{mukhoti2020calibrating}
Jishnu Mukhoti, Viveka Kulharia, Amartya Sanyal, Stuart Golodetz, Philip Torr, and Puneet Dokania.
\newblock Calibrating deep neural networks using focal loss.
\newblock \emph{Advances in Neural Information Processing Systems}, 33:\penalty0 15288--15299, 2020.

\bibitem[Mukhoti et~al.(2023)Mukhoti, Kirsch, van Amersfoort, Torr, and Gal]{mukhoti2023deep}
Jishnu Mukhoti, Andreas Kirsch, Joost van Amersfoort, Philip~HS Torr, and Yarin Gal.
\newblock Deep deterministic uncertainty: A new simple baseline.
\newblock In \emph{Proceedings of the IEEE/CVF Conference on Computer Vision and Pattern Recognition}, pages 24384--24394, 2023.

\bibitem[M{\"u}llner(2011)]{mullner2011modern}
Daniel M{\"u}llner.
\newblock Modern hierarchical, agglomerative clustering algorithms.
\newblock \emph{arXiv preprint arXiv:1109.2378}, 2011.

\bibitem[Naeini et~al.(2015)Naeini, Cooper, and Hauskrecht]{naeini2015obtaining}
Mahdi~Pakdaman Naeini, Gregory Cooper, and Milos Hauskrecht.
\newblock Obtaining well calibrated probabilities using bayesian binning.
\newblock In \emph{Proceedings of the AAAI conference on artificial intelligence}, volume~29, 2015.

\bibitem[Neal(1992)]{NIPS1992_f29c21d4}
Radford Neal.
\newblock {Bayesian} learning via stochastic dynamics.
\newblock In S.~Hanson, J.~Cowan, and C.~Giles, editors, \emph{Advances in Neural Information Processing Systems}, volume~5. Morgan-Kaufmann, 1992.
\newblock URL \url{https://proceedings.neurips.cc/paper/1992/file/f29c21d4897f78948b91f03172341b7b-Paper.pdf}.

\bibitem[Neal(2012)]{neal2012bayesian}
Radford~M Neal.
\newblock \emph{{Bayesian} learning for neural networks}, volume 118.
\newblock Springer Science \& Business Media, 2012.

\bibitem[Neal et~al.(2011)]{neal2011mcmc}
Radford~M Neal et~al.
\newblock {MCMC using Hamiltonian dynamics}.
\newblock \emph{{Handbook of Markov Chain Monte Carlo}}, 2\penalty0 (11):\penalty0 2, 2011.

\bibitem[Netzer et~al.(2011)Netzer, Wang, Coates, Bissacco, Wu, Ng, et~al.]{netzer2011reading}
Yuval Netzer, Tao Wang, Adam Coates, Alessandro Bissacco, Baolin Wu, Andrew~Y Ng, et~al.
\newblock Reading digits in natural images with unsupervised feature learning.
\newblock In \emph{NIPS workshop on deep learning and unsupervised feature learning}, volume 2011, page~7. Granada, Spain, 2011.

\bibitem[Nguyen et~al.(2015)Nguyen, Yosinski, and Clune]{Nguyen_2015_CVPR}
Anh Nguyen, Jason Yosinski, and Jeff Clune.
\newblock Deep neural networks are easily fooled: High confidence predictions for unrecognizable images.
\newblock In \emph{Proceedings of the IEEE Conference on Computer Vision and Pattern Recognition (CVPR)}, June 2015.

\bibitem[Nguyen(1978)]{Nguyen78}
Hung~T. Nguyen.
\newblock On random sets and belief functions.
\newblock \emph{Journal of Mathematical Analysis and Applications}, 65:\penalty0 531--542, 1978.

\bibitem[Nguyen et~al.(2018)Nguyen, Destercke, Masson, and Hüllermeier]{ijcai2018-706}
Vu-Linh Nguyen, Sébastien Destercke, Marie-Hélène Masson, and Eyke Hüllermeier.
\newblock Reliable multi-class classification based on pairwise epistemic and aleatoric uncertainty.
\newblock In \emph{Proceedings of the Twenty-Seventh International Joint Conference on Artificial Intelligence, {IJCAI-18}}, pages 5089--5095. International Joint Conferences on Artificial Intelligence Organization, 7 2018.
\newblock \doi{10.24963/ijcai.2018/706}.
\newblock URL \url{https://doi.org/10.24963/ijcai.2018/706}.

\bibitem[Nixon et~al.(2019)Nixon, Dusenberry, Zhang, Jerfel, and Tran]{nixon2019measuring}
Jeremy Nixon, Michael~W Dusenberry, Linchuan Zhang, Ghassen Jerfel, and Dustin Tran.
\newblock Measuring calibration in deep learning.
\newblock In \emph{CVPR workshops}, volume~2, 2019.

\bibitem[Nouretdinov et~al.(2001)Nouretdinov, Melluish, and Vovk]{Nouretdinov01ridgeregression}
Ilia Nouretdinov, Tom Melluish, and Volodya Vovk.
\newblock Ridge regression confidence machine.
\newblock In \emph{Proceedings of the Eighteenth International Conference on Machine Learning}, pages 385--392. Morgan Kaufmann, 2001.

\bibitem[Oala et~al.(2021)Oala, Hei{\ss}, Macdonald, M{\"a}rz, Kutyniok, and Samek]{oala2021}
Luis Oala, Cosmas Hei{\ss}, Jan Macdonald, Maximilian M{\"a}rz, Gitta Kutyniok, and Wojciech Samek.
\newblock Detecting failure modes in image reconstructions with interval neural network uncertainty.
\newblock \emph{International Journal of Computer Assisted Radiology and Surgery}, 16\penalty0 (12):\penalty0 2089--2097, 2021.

\bibitem[Ojeda et~al.(2023)Ojeda, Jansen, Thi{\'e}ry, Blankenberg, Weimar, Schmid, and Ziegler]{ojeda2023calibrating}
Francisco~M Ojeda, Max~L Jansen, Alexandre Thi{\'e}ry, Stefan Blankenberg, Christian Weimar, Matthias Schmid, and Andreas Ziegler.
\newblock Calibrating machine learning approaches for probability estimation: A comprehensive comparison.
\newblock \emph{Statistics in Medicine}, 42\penalty0 (29):\penalty0 5451--5478, 2023.

\bibitem[Osawa et~al.(2019)Osawa, Swaroop, Khan, Jain, Eschenhagen, Turner, and Yokota]{osawa2019practical}
Kazuki Osawa, Siddharth Swaroop, Mohammad Emtiyaz~E Khan, Anirudh Jain, Runa Eschenhagen, Richard~E Turner, and Rio Yokota.
\newblock Practical deep learning with bayesian principles.
\newblock \emph{Advances in neural information processing systems}, 32, 2019.

\bibitem[Osband et~al.(2024)Osband, Wen, Asghari, Dwaracherla, Ibrahimi, Lu, and Van~Roy]{osband2024epistemic}
Ian Osband, Zheng Wen, Seyed~Mohammad Asghari, Vikranth Dwaracherla, Morteza Ibrahimi, Xiuyuan Lu, and Benjamin Van~Roy.
\newblock Epistemic neural networks.
\newblock \emph{Advances in Neural Information Processing Systems}, 36, 2024.

\bibitem[Ovadia et~al.(2019)Ovadia, Fertig, Ren, Nado, Sculley, Nowozin, Dillon, Lakshminarayanan, and Snoek]{ovadia2019can}
Yaniv Ovadia, Emily Fertig, Jie Ren, Zachary Nado, David Sculley, Sebastian Nowozin, Joshua Dillon, Balaji Lakshminarayanan, and Jasper Snoek.
\newblock Can you trust your model's uncertainty? evaluating predictive uncertainty under dataset shift.
\newblock \emph{Advances in neural information processing systems}, 32, 2019.

\bibitem[Pal et~al.(1993)Pal, Bezdek, and Hemasinha]{pal1993uncertainty}
Nikhil~R Pal, James~C Bezdek, and Rohan Hemasinha.
\newblock Uncertainty measures for evidential reasoning ii: A new measure of total uncertainty.
\newblock \emph{International Journal of Approximate Reasoning}, 8\penalty0 (1):\penalty0 1--16, 1993.

\bibitem[Papadopoulos et~al.(2002{\natexlab{a}})Papadopoulos, Proedrou, Vovk, and Gammerman]{10.1007/3-540-36755-1_29}
Harris Papadopoulos, Kostas Proedrou, Volodya Vovk, and Alex Gammerman.
\newblock Inductive confidence machines for regression.
\newblock In Tapio Elomaa, Heikki Mannila, and Hannu Toivonen, editors, \emph{Machine Learning: ECML 2002}, pages 345--356, Berlin, Heidelberg, 2002{\natexlab{a}}. Springer Berlin Heidelberg.
\newblock ISBN 978-3-540-36755-0.

\bibitem[Papadopoulos et~al.(2002{\natexlab{b}})Papadopoulos, Vovk, and Gammerman]{ICP}
Harris Papadopoulos, Vladimir Vovk, and Alexander Gammerman.
\newblock Qualified prediction for large data sets in the case of pattern recognition.
\newblock In \emph{ICMLA}, pages 159--163, 2002{\natexlab{b}}.

\bibitem[Papadopoulos et~al.(2008)Papadopoulos, Gammerman, and Vovk]{10.5555/1712759.1712773}
Harris Papadopoulos, Alex Gammerman, and Volodya Vovk.
\newblock Normalized nonconformity measures for regression conformal prediction.
\newblock In \emph{Proceedings of the 26th IASTED International Conference on Artificial Intelligence and Applications}, AIA '08, page 64–69, USA, 2008. ACTA Press.
\newblock ISBN 9780889867109.

\bibitem[Papernot et~al.(2016)Papernot, McDaniel, Jha, Fredrikson, Celik, and Swami]{papernot2016limitations}
Nicolas Papernot, Patrick McDaniel, Somesh Jha, Matt Fredrikson, Z~Berkay Celik, and Ananthram Swami.
\newblock The limitations of deep learning in adversarial settings.
\newblock In \emph{2016 IEEE European Symposium on Security and Privacy (EuroS\&P)}, pages 372--387, 2016.

\bibitem[Park and Grandhi(2011)]{park2011quantifying}
Inseok Park and Ramana~V Grandhi.
\newblock Quantifying multiple types of uncertainty in physics-based simulation using bayesian model averaging.
\newblock \emph{AIAA journal}, 49\penalty0 (5):\penalty0 1038--1045, 2011.

\bibitem[Pawlak(1982)]{Pawlak1982}
Zdzislaw Pawlak.
\newblock Rough sets.
\newblock \emph{International Journal of Computer and Information Sciences}, 11\penalty0 (5):\penalty0 341--356, 1982.

\bibitem[Pearce et~al.(2018)Pearce, Brintrup, Zaki, and Neely]{pearce2018high}
Tim Pearce, Alexandra Brintrup, Mohamed Zaki, and Andy Neely.
\newblock High-quality prediction intervals for deep learning: A distribution-free, ensembled approach.
\newblock In \emph{International conference on machine learning}, pages 4075--4084. PMLR, 2018.

\bibitem[Pearce et~al.(2020)Pearce, Leibfried, and Brintrup]{pearce2020uncertainty}
Tim Pearce, Felix Leibfried, and Alexandra Brintrup.
\newblock Uncertainty in neural networks: Approximately bayesian ensembling.
\newblock In \emph{International conference on artificial intelligence and statistics}, pages 234--244. PMLR, 2020.

\bibitem[Pericchi and Walley(1991)]{pericchi1991robust}
Luis~Raul Pericchi and Peter Walley.
\newblock Robust bayesian credible intervals and prior ignorance.
\newblock \emph{International Statistical Review/Revue Internationale de Statistique}, pages 1--23, 1991.

\bibitem[Pinto et~al.(2017)Pinto, Davidson, Sukthankar, and Gupta]{pinto2017robust}
Lerrel Pinto, James Davidson, Rahul Sukthankar, and Abhinav Gupta.
\newblock Robust adversarial reinforcement learning.
\newblock In \emph{International conference on machine learning}, pages 2817--2826. PMLR, 2017.

\bibitem[Platt et~al.(1999)]{platt1999probabilistic}
John Platt et~al.
\newblock Probabilistic outputs for support vector machines and comparisons to regularized likelihood methods.
\newblock \emph{Advances in large margin classifiers}, 10\penalty0 (3):\penalty0 61--74, 1999.

\bibitem[Popper(1959)]{popper1959propensity}
Karl~R Popper.
\newblock The propensity interpretation of probability.
\newblock \emph{The British journal for the philosophy of science}, 10\penalty0 (37):\penalty0 25--42, 1959.

\bibitem[Postels et~al.(2021)Postels, Segu, Sun, Sieber, Van~Gool, Yu, and Tombari]{postels2021practicality}
Janis Postels, Mattia Segu, Tao Sun, Luca Sieber, Luc Van~Gool, Fisher Yu, and Federico Tombari.
\newblock On the practicality of deterministic epistemic uncertainty.
\newblock \emph{arXiv preprint arXiv:2107.00649}, 2021.

\bibitem[Proedrou et~al.(2002)Proedrou, Nouretdinov, Vovk, and Gammerman]{10.1007/3-540-36755-1_32}
Kostas Proedrou, Ilia Nouretdinov, Volodya Vovk, and Alex Gammerman.
\newblock Transductive confidence machines for pattern recognition.
\newblock In Tapio Elomaa, Heikki Mannila, and Hannu Toivonen, editors, \emph{Machine Learning: ECML 2002}, pages 381--390, Berlin, Heidelberg, 2002. Springer.
\newblock ISBN 978-3-540-36755-0.

\bibitem[Qu et~al.(2024)Qu, Chen, Yue, Fu, and Huang]{qu2024hyper}
Jingen Qu, Yufei Chen, Xiaodong Yue, Wei Fu, and Qiguang Huang.
\newblock Hyper-opinion evidential deep learning for out-of-distribution detection.
\newblock \emph{Advances in Neural Information Processing Systems}, 37:\penalty0 84645--84668, 2024.

\bibitem[Radford et~al.(2021)Radford, Kim, Hallacy, Ramesh, Goh, Agarwal, Sastry, Askell, Mishkin, Clark, et~al.]{radford2021learning}
Alec Radford, Jong~Wook Kim, Chris Hallacy, Aditya Ramesh, Gabriel Goh, Sandhini Agarwal, Girish Sastry, Amanda Askell, Pamela Mishkin, Jack Clark, et~al.
\newblock Learning transferable visual models from natural language supervision.
\newblock In \emph{International conference on machine learning}, pages 8748--8763. PMLR, 2021.

\bibitem[Rahaman et~al.(2021)]{rahaman2021uncertainty}
Rahul Rahaman et~al.
\newblock Uncertainty quantification and deep ensembles.
\newblock \emph{Advances in neural information processing systems}, 34:\penalty0 20063--20075, 2021.

\bibitem[Ramesh et~al.(2022)Ramesh, Dhariwal, Nichol, Chu, and Chen]{ramesh2022hierarchical}
Aditya Ramesh, Prafulla Dhariwal, Alex Nichol, Casey Chu, and Mark Chen.
\newblock Hierarchical text-conditional image generation with clip latents.
\newblock \emph{arXiv preprint arXiv:2204.06125}, 1\penalty0 (2):\penalty0 3, 2022.

\bibitem[Ratner et~al.(2017)Ratner, Bach, Ehrenberg, Fries, Wu, and R{\'e}]{ratner2017snorkel}
Alexander Ratner, Stephen~H Bach, Henry Ehrenberg, Jason Fries, Sen Wu, and Christopher R{\'e}.
\newblock Snorkel: Rapid training data creation with weak supervision.
\newblock In \emph{Proceedings of the VLDB endowment. International conference on very large data bases}, volume~11, page 269. NIH Public Access, 2017.

\bibitem[Reddy et~al.(2019)Reddy, Chen, and Manning]{reddy-etal-2019-coqa}
Siva Reddy, Danqi Chen, and Christopher~D. Manning.
\newblock {C}o{QA}: A conversational question answering challenge.
\newblock \emph{Transactions of the Association for Computational Linguistics}, 7:\penalty0 249--266, 2019.
\newblock \doi{10.1162/tacl_a_00266}.
\newblock URL \url{https://aclanthology.org/Q19-1016}.

\bibitem[Reimers and Gurevych(2019)]{reimers-2019-sentence-bert}
Nils Reimers and Iryna Gurevych.
\newblock Sentence-bert: Sentence embeddings using siamese bert-networks.
\newblock In \emph{Proceedings of the 2019 Conference on Empirical Methods in Natural Language Processing}. Association for Computational Linguistics, 11 2019.
\newblock URL \url{https://arxiv.org/abs/1908.10084}.

\bibitem[Ritter et~al.(2018)Ritter, Botev, and Barber]{ritter2018scalable}
Hippolyt Ritter, Aleksandar Botev, and David Barber.
\newblock A scalable laplace approximation for neural networks.
\newblock In \emph{6th international conference on learning representations, ICLR 2018-conference track proceedings}, volume~6. International Conference on Representation Learning, 2018.

\bibitem[Rogova(2008)]{ROGOVA1994777}
Galina Rogova.
\newblock Combining the results of several neural network classifiers.
\newblock \emph{Classic Works of the Dempster-Shafer Theory of Belief Functions}, pages 683--692, 2008.

\bibitem[Rolnick et~al.(2019)Rolnick, Ahuja, Schwarz, Lillicrap, and Wayne]{rolnick2019experience}
David Rolnick, Arun Ahuja, Jonathan Schwarz, Timothy Lillicrap, and Gregory Wayne.
\newblock Experience replay for continual learning.
\newblock \emph{Advances in neural information processing systems}, 32, 2019.

\bibitem[Rosenfeld et~al.(2022)Rosenfeld, Ravikumar, and Risteski]{rosenfeld2022online}
Elan Rosenfeld, Pradeep Ravikumar, and Andrej Risteski.
\newblock An online learning approach to interpolation and extrapolation in domain generalization.
\newblock In \emph{International Conference on Artificial Intelligence and Statistics}, pages 2641--2657. PMLR, 2022.

\bibitem[Rudner et~al.(2022)Rudner, Chen, Teh, and Gal]{rudner2022tractable}
Tim~GJ Rudner, Zonghao Chen, Yee~Whye Teh, and Yarin Gal.
\newblock Tractable function-space variational inference in {{Bayesian}} neural networks.
\newblock \emph{Advances in Neural Information Processing Systems}, 35:\penalty0 22686--22698, 2022.

\bibitem[Russell et~al.(2015)Russell, Dewey, and Tegmark]{russell2015research}
Stuart Russell, Daniel Dewey, and Max Tegmark.
\newblock Research priorities for robust and beneficial artificial intelligence.
\newblock \emph{AI magazine}, 36\penalty0 (4):\penalty0 105--114, 2015.

\bibitem[S.~Salem et~al.(2020)S.~Salem, Langseth, and Ramampiaro]{pmlrsalem20a}
T\'arik S.~Salem, Helge Langseth, and Heri Ramampiaro.
\newblock Prediction intervals: Split normal mixture from quality-driven deep ensembles.
\newblock In Jonas Peters and David Sontag, editors, \emph{Proceedings of the 36th Conference on Uncertainty in Artificial Intelligence (UAI)}, volume 124 of \emph{Proceedings of Machine Learning Research}, pages 1179--1187. PMLR, 03--06 Aug 2020.

\bibitem[Sale et~al.(2023)Sale, Caprio, and H{\"u}llermeier]{sale2023volume}
Yusuf Sale, Michele Caprio, and Eyke H{\"u}llermeier.
\newblock Is the volume of a credal set a good measure for epistemic uncertainty?
\newblock In \emph{Uncertainty in Artificial Intelligence}, pages 1795--1804. PMLR, 2023.

\bibitem[Samaniego et~al.(2018)Samaniego, Thober, Kumar, Wanders, Rakovec, Pan, Zink, Sheffield, Wood, and Marx]{samaniego2018anthropogenic}
Luis Samaniego, Stephan Thober, Rohini Kumar, Niko Wanders, Oldrich Rakovec, Ming Pan, Matthias Zink, Justin Sheffield, Eric~F Wood, and Andreas Marx.
\newblock Anthropogenic warming exacerbates european soil moisture droughts.
\newblock \emph{Nature Climate Change}, 8\penalty0 (5):\penalty0 421--426, 2018.

\bibitem[Saunders et~al.(1999)Saunders, Gammerman, and Vovk]{Saunders1999TransductionWC}
Craig Saunders, Alexander Gammerman, and Volodya Vovk.
\newblock Transduction with confidence and credibility.
\newblock In \emph{Proceedings of the Sixteenth International Joint Conference on Artificial Intelligence}, IJCAI '99, page 722–726, San Francisco, CA, USA, 1999. Morgan Kaufmann Publishers Inc.
\newblock ISBN 1558606130.

\bibitem[Schwarting et~al.(2018)Schwarting, Alonso-Mora, and Rus]{schwarting2018planning}
Wilko Schwarting, Javier Alonso-Mora, and Daniela Rus.
\newblock Planning and decision-making for autonomous vehicles.
\newblock \emph{Annual Review of Control, Robotics, and Autonomous Systems}, 1\penalty0 (1):\penalty0 187--210, 2018.

\bibitem[Scott(2012)]{bceloss}
Clayton Scott.
\newblock {Calibrated asymmetric surrogate losses}.
\newblock \emph{Electronic Journal of Statistics}, 6\penalty0 (none):\penalty0 958 -- 992, 2012.
\newblock \doi{10.1214/12-EJS699}.
\newblock URL \url{https://doi.org/10.1214/12-EJS699}.

\bibitem[Senge et~al.(2014)Senge, B{\"o}sner, Dembczy{\'n}ski, Haasenritter, Hirsch, Donner-Banzhoff, and H{\"u}llermeier]{senge2014reliable}
Robin Senge, Stefan B{\"o}sner, Krzysztof Dembczy{\'n}ski, J{\"o}rg Haasenritter, Oliver Hirsch, Norbert Donner-Banzhoff, and Eyke H{\"u}llermeier.
\newblock Reliable classification: Learning classifiers that distinguish aleatoric and epistemic uncertainty.
\newblock \emph{Information Sciences}, 255:\penalty0 16--29, 2014.

\bibitem[Sensoy et~al.(2018)Sensoy, Kaplan, and Kandemir]{sensoy}
Murat Sensoy, Lance Kaplan, and Melih Kandemir.
\newblock Evidential deep learning to quantify classification uncertainty.
\newblock In \emph{Proceedings of the 32nd International Conference on Neural Information Processing Systems}, NIPS'18, page 3183–3193, Red Hook, NY, USA, 2018. Curran Associates Inc.

\bibitem[Seyeditabari et~al.(2018)Seyeditabari, Tabari, and Zadrozny]{seyeditabari2018emotion}
Armin Seyeditabari, Narges Tabari, and Wlodek Zadrozny.
\newblock Emotion detection in text: a review.
\newblock \emph{arXiv preprint arXiv:1806.00674}, 2018.

\bibitem[Shafeeq and Hareesha(2012)]{shafeeq2012dynamic}
Ahamed Shafeeq and KS~Hareesha.
\newblock Dynamic clustering of data with modified k-means algorithm.
\newblock In \emph{Proceedings of the 2012 conference on information and computer networks}, pages 221--225, 2012.

\bibitem[Shafer(1976)]{shafer1976mathematical}
Glenn Shafer.
\newblock \emph{A mathematical theory of evidence}, volume~42.
\newblock Princeton university press, 1976.

\bibitem[Shafer(1978)]{shafer1981b}
Glenn Shafer.
\newblock Two theories of probability.
\newblock In \emph{PSA: Proceedings of the Biennial Meeting of the Philosophy of Science Association}, volume 1978, pages 441--465. Philosophy of Science Association, 1978.

\bibitem[Shafer(1984)]{shafer84tech}
Glenn Shafer.
\newblock The combination of evidence.
\newblock Technical Report 162, School of Business, University of Kansas, 1984.

\bibitem[Shafer and Vovk(2008)]{shafer2008tutorial}
Glenn Shafer and Vladimir Vovk.
\newblock A tutorial on {Conformal Prediction}.
\newblock \emph{Journal of Machine Learning Research}, 9\penalty0 (3), 2008.

\bibitem[Shaker and H{\"u}llermeier(2021)]{shaker2021ensemble}
Mohammad~Hossein Shaker and Eyke H{\"u}llermeier.
\newblock Ensemble-based uncertainty quantification: {Bayesian} versus credal inference.
\newblock In \emph{PROCEEDINGS 31. WORKSHOP COMPUTATIONAL INTELLIGENCE}, volume~25, page~63, 2021.

\bibitem[Shlens(2014)]{shlens2014notes}
Jonathon Shlens.
\newblock Notes on kullback-leibler divergence and likelihood.
\newblock \emph{arXiv preprint arXiv:1404.2000}, 2014.

\bibitem[Singh et~al.(2024)Singh, Chau, Bouabid, and Muandet]{singh2024domain}
Anurag Singh, Siu~Lun Chau, Shahine Bouabid, and Krikamol Muandet.
\newblock Domain generalisation via imprecise learning.
\newblock \emph{arXiv preprint arXiv:2404.04669}, 2024.

\bibitem[Singh et~al.(2022)Singh, Akrigg, Di~Maio, Fontana, Alitappeh, Khan, Saha, Jeddisaravi, Yousefi, Culley, et~al.]{singh2022road}
Gurkirt Singh, Stephen Akrigg, Manuele Di~Maio, Valentina Fontana, Reza~Javanmard Alitappeh, Salman Khan, Suman Saha, Kossar Jeddisaravi, Farzad Yousefi, Jacob Culley, et~al.
\newblock Road: The road event awareness dataset for autonomous driving.
\newblock \emph{IEEE transactions on pattern analysis and machine intelligence}, 45\penalty0 (1):\penalty0 1036--1054, 2022.

\bibitem[Smarandache et~al.(2011)Smarandache, Martin, and Osswald]{smarandache2011contradiction}
Florentin Smarandache, Arnaud Martin, and Christophe Osswald.
\newblock Contradiction measures and specificity degrees of basic belief assignments.
\newblock In \emph{14th International Conference on Information Fusion}, pages 1--8. IEEE, 2011.

\bibitem[Smets(2005)]{SMETS2005133}
Philippe Smets.
\newblock Decision making in the {TBM}: the necessity of the pignistic transformation.
\newblock \emph{International Journal of Approximate Reasoning}, 38\penalty0 (2):\penalty0 133--147, 2005.
\newblock ISSN 0888-613X.
\newblock \doi{https://doi.org/10.1016/j.ijar.2004.05.003}.
\newblock URL \url{https://www.sciencedirect.com/science/article/pii/S0888613X04000593}.

\bibitem[Smets(July 1986)]{smets86bayes}
Philippe Smets.
\newblock Bayes' theorem generalized for belief functions.
\newblock In \emph{Proceedings of the 7th European Conference on Artificial Intelligence (ECAI-86)}, volume~2, pages 169--171, July 1986.

\bibitem[Smets and Kennes(1994)]{smets1994transferable}
Philippe Smets and Robert Kennes.
\newblock The transferable belief model.
\newblock \emph{Artificial intelligence}, 66\penalty0 (2):\penalty0 191--234, 1994.

\bibitem[Snee(1990)]{snee1990statistical}
Ronald~D. Snee.
\newblock Statistical thinking and its contribution to total quality.
\newblock \emph{The American Statistician}, 44\penalty0 (2):\penalty0 116--121, 1990.

\bibitem[Snowling and Kramer(2001)]{SNOWLING200117}
S.D Snowling and J.R Kramer.
\newblock Evaluating modelling uncertainty for model selection.
\newblock \emph{Ecological Modelling}, 138\penalty0 (1):\penalty0 17--30, 2001.
\newblock ISSN 0304-3800.
\newblock \doi{https://doi.org/10.1016/S0304-3800(00)00390-2}.
\newblock URL \url{https://www.sciencedirect.com/science/article/pii/S0304380000003902}.

\bibitem[Song et~al.(2018)Song, Wang, Wu, Quan, and Huang]{song2018evidence}
Yafei Song, Xiaodan Wang, Wenhua Wu, Wen Quan, and Wenlong Huang.
\newblock Evidence combination based on credibility and non-specificity.
\newblock \emph{Pattern Analysis and Applications}, 21:\penalty0 167--180, 2018.

\bibitem[Spruyt(2013)]{spruyt2014draw}
Vincent Spruyt.
\newblock How to draw a covariance error ellipse. 2014.
\newblock \emph{Computer Vision for Dummies.[Available Online-http://www. visiondummy. com/2014/04/drawerror-ellipse-representing-covariance-matrix/]}, 2013.

\bibitem[Srivastava et~al.(2014)Srivastava, Hinton, Krizhevsky, Sutskever, and Salakhutdinov]{srivastava2014dropout}
Nitish Srivastava, Geoffrey Hinton, Alex Krizhevsky, Ilya Sutskever, and Ruslan Salakhutdinov.
\newblock Dropout: a simple way to prevent neural networks from overfitting.
\newblock \emph{The journal of machine learning research}, 15\penalty0 (1):\penalty0 1929--1958, 2014.

\bibitem[Stadler et~al.(2021)Stadler, Charpentier, Geisler, Z{\"u}gner, and G{\"u}nnemann]{stadler2021graph}
Maximilian Stadler, Bertrand Charpentier, Simon Geisler, Daniel Z{\"u}gner, and Stephan G{\"u}nnemann.
\newblock Graph posterior network: Bayesian predictive uncertainty for node classification.
\newblock \emph{Advances in Neural Information Processing Systems}, 34:\penalty0 18033--18048, 2021.

\bibitem[Strat(1984)]{strat1984continuous}
Thomas~M Strat.
\newblock Continuous belief functions for evidential reasoning.
\newblock In \emph{AAAI}, pages 308--313, 1984.

\bibitem[Sun et~al.(2020)Sun, Kretzschmar, Dotiwalla, Chouard, Patnaik, Tsui, Guo, Zhou, Chai, Caine, Vasudevan, Han, Ngiam, Zhao, Timofeev, Ettinger, Krivokon, Gao, Joshi, Zhang, Shlens, Chen, and Anguelov]{Sun_2020_CVPR}
Pei Sun, Henrik Kretzschmar, Xerxes Dotiwalla, Aurelien Chouard, Vijaysai Patnaik, Paul Tsui, James Guo, Yin Zhou, Yuning Chai, Benjamin Caine, Vijay Vasudevan, Wei Han, Jiquan Ngiam, Hang Zhao, Aleksei Timofeev, Scott Ettinger, Maxim Krivokon, Amy Gao, Aditya Joshi, Yu~Zhang, Jonathon Shlens, Zhifeng Chen, and Dragomir Anguelov.
\newblock Scalability in perception for autonomous driving: Waymo open dataset.
\newblock In \emph{Proceedings of the IEEE/CVF Conference on Computer Vision and Pattern Recognition (CVPR)}, June 2020.

\bibitem[Sun et~al.(2019)Sun, Zhang, Shi, and Grosse]{sun2019functional}
Shengyang Sun, Guodong Zhang, Jiaxin Shi, and Roger Grosse.
\newblock Functional variational bayesian neural networks.
\newblock \emph{arXiv preprint arXiv:1903.05779}, 2019.

\bibitem[Swaminathan and Joachims(2015)]{swaminathan2015self}
Adith Swaminathan and Thorsten Joachims.
\newblock The self-normalized estimator for counterfactual learning.
\newblock \emph{advances in neural information processing systems}, 28, 2015.

\bibitem[Tao et~al.(2023)Tao, Dong, and Xu]{tao2023dual}
Linwei Tao, Minjing Dong, and Chang Xu.
\newblock Dual focal loss for calibration.
\newblock In \emph{International Conference on Machine Learning}, pages 33833--33849. PMLR, 2023.

\bibitem[Teeti et~al.(2023)Teeti, Bhargav, Singh, Bradley, Banerjee, and Cuzzolin]{teeti2023temporal}
Izzeddin Teeti, Rongali~Sai Bhargav, Vivek Singh, Andrew Bradley, Biplab Banerjee, and Fabio Cuzzolin.
\newblock Temporal dino: A self-supervised video strategy to enhance action prediction.
\newblock In \emph{Proceedings of the IEEE/CVF International Conference on Computer Vision}, pages 3281--3291, 2023.

\bibitem[Thogmartin(2010)]{thogmartin2010sensitivity}
Wayne~E Thogmartin.
\newblock Sensitivity analysis of north american bird population estimates.
\newblock \emph{Ecological Modelling}, 221\penalty0 (2):\penalty0 173--177, 2010.

\bibitem[Titterington(2004)]{titterington2004bayesian}
D~Michael Titterington.
\newblock Bayesian methods for neural networks and related models.
\newblock \emph{Statistical science}, pages 128--139, 2004.

\bibitem[Tong et~al.(2021)Tong, Xu, and Denoeux]{tong2021evidential}
Zheng Tong, Philippe Xu, and Thierry Denoeux.
\newblock An evidential classifier based on {Dempster-Shafer} theory and deep learning.
\newblock \emph{Neurocomputing}, 450:\penalty0 275--293, 2021.

\bibitem[Touvron et~al.(2023)Touvron, Martin, Stone, Albert, Almahairi, Babaei, Bashlykov, Batra, Bhargava, Bhosale, et~al.]{touvron2023llama}
Hugo Touvron, Louis Martin, Kevin Stone, Peter Albert, Amjad Almahairi, Yasmine Babaei, Nikolay Bashlykov, Soumya Batra, Prajjwal Bhargava, Shruti Bhosale, et~al.
\newblock Llama 2: Open foundation and fine-tuned chat models.
\newblock \emph{arXiv preprint arXiv:2307.09288}, 2023.

\bibitem[Trabelsi et~al.(2011)Trabelsi, Elouedi, and Lingras]{Trabelsi2011ClassificationSB}
Salsabil Trabelsi, Zied Elouedi, and Pawan Lingras.
\newblock Classification systems based on rough sets under the belief function framework.
\newblock \emph{Int. J. Approx. Reason.}, 52:\penalty0 1409--1432, 2011.

\bibitem[Tran et~al.(2020)Tran, Rossi, Milios, and Filippone]{tran2020all}
Ba-Hien Tran, Simone Rossi, Dimitrios Milios, and Maurizio Filippone.
\newblock All you need is a good functional prior for {{Bayesian}} deep learning.
\newblock \emph{arXiv preprint arXiv:2011.12829}, 2020.

\bibitem[Tretiak et~al.(2023)Tretiak, Schollmeyer, and Ferson]{tretiak2023neural}
Krasymyr Tretiak, Georg Schollmeyer, and Scott Ferson.
\newblock Neural network model for imprecise regression with interval dependent variables.
\newblock \emph{Neural Networks}, 161:\penalty0 550--564, 2023.

\bibitem[Troffaes(2007)]{troffaes07}
Matthias Troffaes.
\newblock Decision making under uncertainty using imprecise probabilities.
\newblock \emph{International Journal of Approximate Reasoning}, 45\penalty0 (1):\penalty0 17--29, 2007.

\bibitem[Tsamoura et~al.(2021)Tsamoura, Carral, Malizia, and Urbani]{tsamoura2021materializing}
Efthymia Tsamoura, David Carral, Enrico Malizia, and Jacopo Urbani.
\newblock Materializing knowledge bases via trigger graphs.
\newblock \emph{arXiv preprint arXiv:2102.02753}, 2021.

\bibitem[Ullah(1996)]{ullah1996entropy}
Aman Ullah.
\newblock Entropy, divergence and distance measures with econometric applications.
\newblock \emph{Journal of Statistical Planning and Inference}, 49\penalty0 (1):\penalty0 137--162, 1996.

\bibitem[Ulmer et~al.(2023)Ulmer, Hardmeier, and Frellsen]{ulmer2023prior}
Dennis~Thomas Ulmer, Christian Hardmeier, and Jes Frellsen.
\newblock Prior and posterior networks: A survey on evidential deep learning methods for uncertainty estimation.
\newblock \emph{Transactions on Machine Learning Research}, 2023.

\bibitem[Van~Amersfoort et~al.(2020)Van~Amersfoort, Smith, Teh, and Gal]{van2020uncertainty}
Joost Van~Amersfoort, Lewis Smith, Yee~Whye Teh, and Yarin Gal.
\newblock Uncertainty estimation using a single deep deterministic neural network.
\newblock In \emph{International conference on machine learning}, pages 9690--9700. PMLR, 2020.

\bibitem[Van~de Ven and Tolias(2019)]{van2019three}
Gido~M Van~de Ven and Andreas~S Tolias.
\newblock Three scenarios for continual learning.
\newblock \emph{arXiv preprint arXiv:1904.07734}, 2019.

\bibitem[van~der Maaten and Hinton(2008)]{JMLR:v9:vandermaaten08a}
Laurens van~der Maaten and Geoffrey Hinton.
\newblock Visualizing data using t-sne.
\newblock \emph{Journal of Machine Learning Research}, 9\penalty0 (86):\penalty0 2579--2605, 2008.
\newblock URL \url{http://jmlr.org/papers/v9/vandermaaten08a.html}.

\bibitem[Vannoorenberghe and Smets(2005)]{10.1007/11518655_80}
Patrick Vannoorenberghe and Philippe Smets.
\newblock Partially supervised learning by a credal em approach.
\newblock In Llu{\'i}s Godo, editor, \emph{Symbolic and Quantitative Approaches to Reasoning with Uncertainty}, pages 956--967, Berlin, Heidelberg, 2005. Springer Berlin Heidelberg.
\newblock ISBN 978-3-540-31888-0.

\bibitem[Vega and Todd(2022)]{vega2022variational}
Manuel~A Vega and Michael~D Todd.
\newblock {A variational Bayesian neural network for structural health monitoring and cost-informed decision-making in miter gates}.
\newblock \emph{Structural Health Monitoring}, 21\penalty0 (1):\penalty0 4--18, 2022.

\bibitem[Virtanen et~al.(2020)Virtanen, Gommers, Oliphant, Haberland, Reddy, Cournapeau, Burovski, Peterson, Weckesser, Bright, et~al.]{virtanen2020scipy}
Pauli Virtanen, Ralf Gommers, Travis~E Oliphant, Matt Haberland, Tyler Reddy, David Cournapeau, Evgeni Burovski, Pearu Peterson, Warren Weckesser, Jonathan Bright, et~al.
\newblock Scipy 1.0: fundamental algorithms for scientific computing in python.
\newblock \emph{Nature methods}, 17\penalty0 (3):\penalty0 261--272, 2020.

\bibitem[Vovk(2012)]{crossconformal}
Vladimir Vovk.
\newblock Cross-conformal predictors.
\newblock \emph{Annals of Mathematics and Artificial Intelligence}, 74, 08 2012.
\newblock \doi{10.1007/s10472-013-9368-4}.

\bibitem[Vovk and Petej(2012)]{https://doi.org/10.48550/arxiv.1211.0025}
Vladimir Vovk and Ivan Petej.
\newblock Venn-abers predictors.
\newblock \emph{arXiv preprint arXiv:1211.0025}, 2012.

\bibitem[Vovk et~al.(2003)Vovk, Shafer, and Nouretdinov]{NIPS2003_10c66082}
Vladimir Vovk, Glenn Shafer, and Ilia Nouretdinov.
\newblock Self-calibrating probability forecasting.
\newblock In S.~Thrun, L.~Saul, and B.~Sch\"{o}lkopf, editors, \emph{Advances in Neural Information Processing Systems}, volume~16. MIT Press, 2003.
\newblock URL \url{https://proceedings.neurips.cc/paper/2003/file/10c66082c124f8afe3df4886f5e516e0-Paper.pdf}.

\bibitem[Vovk et~al.(2005)Vovk, Gammerman, and Shafer]{vovk2005algorithmic}
Vladimir Vovk, Alexander Gammerman, and Glenn Shafer.
\newblock \emph{Algorithmic learning in a random world}.
\newblock Springer Science \& Business Media, 2005.

\bibitem[Walley(1991)]{walley91book}
Peter Walley.
\newblock \emph{Statistical Reasoning with Imprecise Probabilities}.
\newblock Chapman and Hall, New York, 1991.

\bibitem[Walley(1996)]{walley1996inferences}
Peter Walley.
\newblock Inferences from multinomial data: learning about a bag of marbles.
\newblock \emph{Journal of the Royal Statistical Society Series B: Statistical Methodology}, 58\penalty0 (1):\penalty0 3--34, 1996.

\bibitem[Walley(2000)]{walley2000towards}
Peter Walley.
\newblock Towards a unified theory of imprecise probability.
\newblock \emph{International Journal of Approximate Reasoning}, 24\penalty0 (2-3):\penalty0 125--148, 2000.

\bibitem[Wallner(2005)]{wallner2005}
Anton Wallner.
\newblock Maximal number of vertices of polytopes defined by f-probabilities.
\newblock In Fabio~Gagliardi Cozman, R.~Nau, and Teddy Seidenfeld, editors, \emph{Proceedings of the Fourth International Symposium on Imprecise Probabilities and Their Applications (ISIPTA 2005)}, pages 388--395, 2005.

\bibitem[Wang et~al.(2024{\natexlab{a}})Wang, Cuzzolin, Shariatmadar, Moens, and Hallez]{wang2024credalwrapper}
Kaizheng Wang, Fabio Cuzzolin, Keivan Shariatmadar, David Moens, and Hans Hallez.
\newblock Credal wrapper of model averaging for uncertainty estimation on out-of-distribution detection.
\newblock \emph{arXiv preprint arXiv:2405.15047}, 2024{\natexlab{a}}.

\bibitem[Wang et~al.(2024{\natexlab{b}})Wang, Cuzzolin, Shariatmadar, Moens, Hallez, et~al.]{wang2024credal}
Kaizheng Wang, Fabio Cuzzolin, Keivan Shariatmadar, David Moens, Hans Hallez, et~al.
\newblock Credal deep ensembles for uncertainty quantification.
\newblock \emph{Advances in Neural Information Processing Systems}, 37:\penalty0 79540--79572, 2024{\natexlab{b}}.

\bibitem[Wang et~al.(2024{\natexlab{c}})Wang, Shariatmadar, Manchingal, Cuzzolin, Moens, and Hallez]{wang2024creinns}
Kaizheng Wang, Keivan Shariatmadar, Shireen~Kudukkil Manchingal, Fabio Cuzzolin, David Moens, and Hans Hallez.
\newblock Creinns: Credal-set interval neural networks for uncertainty estimation in classification tasks.
\newblock \emph{arXiv preprint arXiv:2401.05043}, 2024{\natexlab{c}}.

\bibitem[Wang et~al.(2024{\natexlab{d}})Wang, Shi, Han, Metaxas, and Wang]{wang2024blob}
Yibin Wang, Haizhou Shi, Ligong Han, Dimitris Metaxas, and Hao Wang.
\newblock Blob: Bayesian low-rank adaptation by backpropagation for large language models.
\newblock \emph{arXiv preprint arXiv:2406.11675}, 2024{\natexlab{d}}.

\bibitem[Wattenberg et~al.(2016)Wattenberg, Vi{\'e}gas, and Johnson]{wattenberg2016use}
Martin Wattenberg, Fernanda Vi{\'e}gas, and Ian Johnson.
\newblock How to use t-sne effectively.
\newblock \emph{Distill}, 1\penalty0 (10):\penalty0 e2, 2016.

\bibitem[Welling and Teh(2011)]{welling2011bayesian}
Max Welling and Yee~W Teh.
\newblock Bayesian learning via stochastic gradient langevin dynamics.
\newblock In \emph{Proceedings of the 28th international conference on machine learning (ICML-11)}, pages 681--688. Citeseer, 2011.

\bibitem[Wen et~al.(2020)Wen, Tran, and Ba]{wen2020batchensemble}
Yeming Wen, Dustin Tran, and Jimmy Ba.
\newblock Batchensemble: an alternative approach to efficient ensemble and lifelong learning.
\newblock \emph{arXiv preprint arXiv:2002.06715}, 2020.

\bibitem[Williams and Zipser(1989)]{williams1989learning}
Ronald~J Williams and David Zipser.
\newblock A learning algorithm for continually running fully recurrent neural networks.
\newblock \emph{Neural computation}, 1\penalty0 (2):\penalty0 270--280, 1989.

\bibitem[Wu and Xiao(2024)]{wu2024novel}
Keming Wu and Fuyuan Xiao.
\newblock A novel quantum belief entropy for uncertainty measure in complex evidence theory.
\newblock \emph{Information Sciences}, 652:\penalty0 119744, 2024.

\bibitem[Wu and Goodman(2020)]{wu2020simple}
Mike Wu and Noah Goodman.
\newblock A simple framework for uncertainty in contrastive learning.
\newblock \emph{arXiv preprint arXiv:2010.02038}, 2020.

\bibitem[Xiao et~al.(2017)Xiao, Rasul, and Vollgraf]{xiao2017fashion}
Han Xiao, Kashif Rasul, and Roland Vollgraf.
\newblock Fashion-{MNIST}: a novel image dataset for benchmarking machine learning algorithms.
\newblock \emph{arXiv preprint arXiv:1708.07747}, 2017.

\bibitem[Xu et~al.(1992)Xu, Krzyzak, and Suen]{155943}
L.~Xu, A.~Krzyzak, and C.Y. Suen.
\newblock Methods of combining multiple classifiers and their applications to handwriting recognition.
\newblock \emph{IEEE Transactions on Systems, Man, and Cybernetics}, 22\penalty0 (3):\penalty0 418--435, 1992.
\newblock \doi{10.1109/21.155943}.

\bibitem[Yager(2008)]{yager2008entropy}
Ronald~R Yager.
\newblock Entropy and specificity in a mathematical theory of evidence.
\newblock \emph{Classic works of the Dempster-Shafer theory of belief functions}, pages 291--310, 2008.

\bibitem[Yang et~al.(2023)Yang, Robeyns, Wang, and Aitchison]{yang2023bayesian}
Adam~X Yang, Maxime Robeyns, Xi~Wang, and Laurence Aitchison.
\newblock Bayesian low-rank adaptation for large language models.
\newblock \emph{arXiv preprint arXiv:2308.13111}, 2023.

\bibitem[Yin and Wang(2014)]{yin2014dirichlet}
Jianhua Yin and Jianyong Wang.
\newblock A dirichlet multinomial mixture model-based approach for short text clustering.
\newblock In \emph{Proceedings of the 20th ACM SIGKDD international conference on Knowledge discovery and data mining}, pages 233--242, 2014.

\bibitem[Yu et~al.(2023)Yu, Yang, Kolehmainen, Shivakumar, Gu, Ren, Luo, Gourav, Chen, Liu, et~al.]{yu2023low}
Yu~Yu, Chao-Han~Huck Yang, Jari Kolehmainen, Prashanth~G Shivakumar, Yile Gu, Sungho Ryu~Roger Ren, Qi~Luo, Aditya Gourav, I-Fan Chen, Yi-Chieh Liu, et~al.
\newblock Low-rank adaptation of large language model rescoring for parameter-efficient speech recognition.
\newblock In \emph{2023 IEEE Automatic Speech Recognition and Understanding Workshop (ASRU)}, pages 1--8. IEEE, 2023.

\bibitem[Zaffalon(2002)]{Zaffalon}
Marco Zaffalon.
\newblock The {Naive Credal Classifier}.
\newblock \emph{Journal of Statistical Planning and Inference - J STATIST PLAN INFER}, 105:\penalty0 5--21, 06 2002.
\newblock \doi{10.1016/S0378-3758(01)00201-4}.

\bibitem[Zaffalon and Fagiuoli(2003)]{zaffalon-treebased}
Marco Zaffalon and Enrico Fagiuoli.
\newblock Tree-based credal networks for classification.
\newblock \emph{Reliable computing}, 9\penalty0 (6):\penalty0 487--509, 2003.

\bibitem[Zagoruyko and Komodakis(2016)]{zagoruyko2016wide}
Sergey Zagoruyko and Nikos Komodakis.
\newblock Wide residual networks.
\newblock \emph{arXiv preprint arXiv:1605.07146}, 2016.

\bibitem[Zhang et~al.(2018)Zhang, Sun, Duvenaud, and Grosse]{zhang2018noisy}
Guodong Zhang, Shengyang Sun, David Duvenaud, and Roger Grosse.
\newblock Noisy natural gradient as variational inference.
\newblock In \emph{International conference on machine learning}, pages 5852--5861. PMLR, 2018.

\bibitem[Zhang et~al.(2021)Zhang, Lou, Wang, Wu, Lu, and Jia]{zhang2021evaluating}
Jindi Zhang, Yang Lou, Jianping Wang, Kui Wu, Kejie Lu, and Xiaohua Jia.
\newblock Evaluating adversarial attacks on driving safety in vision-based autonomous vehicles.
\newblock \emph{IEEE Internet of Things Journal}, 9\penalty0 (5):\penalty0 3443--3456, 2021.

\bibitem[Zheng et~al.(2021)Zheng, Yu, Li, Shi, and Haake]{zheng2021continual}
Ervine Zheng, Qi~Yu, Rui Li, Pengcheng Shi, and Anne Haake.
\newblock A continual learning framework for uncertainty-aware interactive image segmentation.
\newblock In \emph{Proceedings of the AAAI Conference on Artificial Intelligence}, volume~35, pages 6030--6038, 2021.

\bibitem[Zhou et~al.(2023)Zhou, Tian, and Deng]{zhou2023bf}
Qianli Zhou, Guojing Tian, and Yong Deng.
\newblock Bf-qc: Belief functions on quantum circuits.
\newblock \emph{Expert Systems with Applications}, 223:\penalty0 119885, 2023.

\bibitem[Zhou et~al.(2024)Zhou, Luo, Boss{\'e}, and Deng]{zhou2024combining}
Qianli Zhou, Hao Luo, {\'E}loi Boss{\'e}, and Yong Deng.
\newblock Why combining belief functions on quantum circuits?
\newblock In \emph{International Conference on Belief Functions}, pages 161--170. Springer, 2024.

\bibitem[Zio and Apostolakis(1996)]{zio1996two}
Enrico Zio and GE~Apostolakis.
\newblock Two methods for the structured assessment of model uncertainty by experts in performance assessments of radioactive waste repositories.
\newblock \emph{Reliability Engineering \& System Safety}, 54\penalty0 (2-3):\penalty0 225--241, 1996.

\bibitem[Zou et~al.(2025)Zou, Meng, and Karniadakis]{zou2025uncertainty}
Zongren Zou, Xuhui Meng, and George~Em Karniadakis.
\newblock Uncertainty quantification for noisy inputs--outputs in physics-informed neural networks and neural operators.
\newblock \emph{Computer Methods in Applied Mechanics and Engineering}, 433:\penalty0 117479, 2025.

\end{thebibliography}


\end{document}